\documentclass[11pt]{article}%
\usepackage{amssymb}
\usepackage[leqno]{amsmath}
\usepackage{amsfonts}
\usepackage[numbers,square]{natbib}
\usepackage{graphicx}
\usepackage{fancyhdr}
\usepackage{lineno}
\usepackage[letterpaper]{hyperref}%
\UseRawInputEncoding
\providecommand{\U}[1]{\protect\rule{.1in}{.1in}}
\newtheorem{theorem}{Theorem}[section]

\newtheorem{axiom}[theorem]{Axiom}

\newtheorem{corollary}[theorem]{Corollary}

\newtheorem{definition}[theorem]{Definition}

\newtheorem{lemma}[theorem]{Lemma}

\newtheorem{remark}[theorem]{Remark}

\newenvironment{proof}[1][Proof]{\noindent\textbf{#1.} }{\ \rule{0.5em}{0.5em}}
\begin{document}

\title{Fundamental Laws of Binary Classification\thanks{Regularization methods
presented in this paper appeared in\textit{\ WIREs Computational Statistics},
3: 204 - 215, 2011.}}
\author{Denise M. Reeves\thanks{Principal Eigenloci, LLC, Burke, Virginia,
(dmreeves22@verizon.net)}}
\date{}
\maketitle

\begin{abstract}
Finding discriminant functions of minimum risk binary classification systems
is a novel geometric locus problem---which requires solving a system of
fundamental locus equations of binary classification---subject to deep-seated
statistical laws. We show that a discriminant function of a minimum risk
binary classification system is the solution of a locus equation that
represents the geometric locus of the decision boundary of the system, wherein
the discriminant function is connected to the decision boundary by an
exclusive principal eigen-coordinate system---at which point the discriminant
function is represented by a geometric locus of a novel principal
eigenaxis---structured as a dual locus of likelihood components and principal
eigenaxis components. We demonstrate that a minimum risk binary classification
system acts to jointly minimize its eigenenergy and risk by locating a point
of equilibrium, at which point critical minimum eigenenergies exhibited by the
system are symmetrically concentrated in such a manner that the novel
principal eigenaxis of the system exhibits symmetrical dimensions and
densities, so that counteracting and opposing forces and influences of the
system are symmetrically balanced with each other---about the geometric center
of the locus of the novel principal eigenaxis---whereon the statistical
fulcrum of the system is located. Thereby, a minimum risk binary
classification system satisfies a state of statistical equilibrium---so that
the total allowed eigenenergy and the expected risk exhibited by the system
are jointly minimized within the decision space of the system---at which point
the system exhibits the minimum probability of classification error.

\textbf{Key words.} fundamental laws of binary classification, direct problem
of the binary classification of random vectors, inverse problem of the binary
classification of random vectors, likelihood ratio tests, minimum risk
classification systems, minimum probability of classification error, minimum
expected risk, total allowed eigenenergy, critical minimum eigenenergies,
statistical equilibrium, eigenaxis of symmetry, geometric locus of a novel
principal eigenaxis, coordinate geometry, vector algebra locus equations,
novel geometric locus methods, novel principal eigen-coordinate transforms,
novel principal eigen-coordinate transform algorithms, machine learning
algorithms, constrained optimization algorithms, system identification
problems, ill-posed direct problems, ill-posed inverse problems, data-driven
mathematical models, vector-valued cost functions, reproducing kernel Hilbert
spaces, supervised learning no free lunch theorems, Bayes' decision rule,
support vector machines

\end{abstract}

\section{Finding Discriminant Functions}

Finding discriminant functions of minimum risk classification systems is a
long-standing and deep-seated problem in both machine learning and
statistics---situated far beneath the surface---such that neither statistical
learning theory nor the approaches of statistical decision theory, including
Bayesian decision theory, have resolved the fundamental problem of how to find
discriminant functions of minimum risk binary classification systems that
exhibit the minimum probability of classification error.

Generally, statistical learning theory recommends using machine learning
algorithms called \textquotedblleft support vector machines\textquotedblright%
\ (SVMs) for finding indicator functions and separating hyperplanes of binary
classification systems
\citep{Bennett2000,Boser1992,Cortes1995,Cristianini2000,Scholkopf2002}%
, whereas Bayesian decision theory recommends using modifications of Bayes'
theorem called \textquotedblleft Bayes' decision rule\textquotedblright\ for
selecting likelihood ratios and decision thresholds of binary classification
systems
\citep
{Duda2001,Fukunaga1990,Kay1998fundamentals,Poor1996,Srinath1996,stark1994probability,VanTrees1968}%
.%

\pagestyle{fancy}
\lhead{}%

Regardless of the method used for finding or selecting discriminant functions,
the effectiveness of any given classification system is primarily evaluated by
its probability of classification error---which is the lowest possible error
rate of the system. Bayes' decision rule is known for minimizing the
probability of classification error, better known as \textquotedblleft Bayes'
error,\textquotedblright\ since Bayes' decision rule minimizes the
\textquotedblleft Bayes' risk\textquotedblright\ of binary and multiclass
classification systems
\citep
{Duda2001,Fukunaga1990,Poor1996,Srinath1996,stark1994probability,VanTrees1968}%
. On the other hand, SVM classifiers are known for minimizing the mean error
rate since SVM classifiers minimize the \textquotedblleft empirical
risk\textquotedblright\ of binary and multiclass classification systems for
certain data samples
\citep{Bennett2000,Boser1992,Cortes1995,Cristianini2000,Scholkopf2002}%
.

Despite the fact that SVM classifiers are widely reported to perform well on
classification tasks, SVM classifiers are largely determined by arbitrary or
ill-suited hyperparameters
\citep{Byun2002,Eitrich2006,Liang2011}%
. For example, arbitrary regularization parameters are conflated with
ill-suited slack variables
\citep{Reeves2011,Reeves2009}%
, while the selection of a nonlinear kernel and its hyperparameter for a
classification task is considered a research problem
\citep{Burges1998}%
.

Thereby, we have concluded that SVM classifiers and the empirical risk of SVM
classifiers are both determined by inconsistent and arbitrary criteria.

Moreover, even though Bayes' decision rule is considered the gold standard for
classification tasks
\citep
{Duda2001,Fukunaga1990,Kay1998fundamentals,Poor1996,Srinath1996,stark1994probability,VanTrees1968}%
, we have determined that Bayes' decision rule is mathematically inconsistent
with the conditions of Bayes' theorem.

Thereby, we have concluded that Bayes' decision rule and Bayes' risk are both
determined by inconsistent and arbitrary criteria.

As of now, no statistical laws have been established that determine the
overall statistical structure and behavior and properties of minimum risk
classification systems that exhibit the minimum probability of classification
error. Since Bayes' decision rule and Bayes' risk are both determined by
inconsistent and arbitrary criteria, it follows that the statistical structure
and behavior and properties of Bayes' minimum risk classification systems are
ill-defined and ambiguous.

Furthermore, no statistical laws have been established that determine the
generalization behavior of machine learning algorithms that find target
functions of minimum risk classification systems---since the \emph{target
functions} are largely \emph{unknown}
\citep{Breiman1991,Burges1998,Duda2001,Geman1992}%
.

Moreover, given the supervised learning no free lunch theorems
\citep{Duda2001,Wolpert2001supervised,Wolpert2020}%
, we realize that unless one can somehow prove, \emph{from first principles},
that a \emph{target function} of a minimum risk classification system has a
\emph{certain statistical structure}---then one cannot prove that a particular
machine learning algorithm will be aligned with the target function---and one
cannot prove anything concerning how well that learning algorithm generalizes.

The process of using observed data to determine a mathematical model of a
system is essential in science and engineering
\citep{Linz1979,ljung1998system,ljung1998systemid}%
. We realize that determining the generalization behavior of certain machine
learning algorithms involves solving a \emph{system identification problem},
so that the overall statistical structure and behavior and properties of a
given system are determined by transforming a collection of observations into
a data-driven mathematical model that represents fundamental aspects of the
system
\citep{Reeves2011,Reeves2009,Reeves2015resolving,Reeves2018design}%
.

It will be seen that \emph{data-driven} \emph{mathematical models} of systems
are \emph{driven} by \emph{data}---that \emph{satisfy} certain
\emph{equations} of \emph{mathematical laws}---which \emph{determine} the
overall structure and behavior and properties of a \emph{given system}.

Solving a system identification problem involves formulating and solving an
\emph{inverse problem}, so that a collection of actual observations are used
to infer the values of the parameters characterizing a given system
\citep{Linz1979,ljung1998system,ljung1998systemid}%
.

We realize, however, that solving a system identification problem also
involves formulating and solving a \emph{direct or forward problem}---which
entails formulating and solving a \emph{fully specified} mathematical
\emph{model} of a given system---whose solution is \emph{used} to
\emph{predict} some type of \emph{system behavior}
\citep{Linz1979,Linz2003}%
.

It is well known that an inverse problem is an \emph{ill-posed} problem in the
event that distinct causes for a given system account for the same effect
\citep
{Engl2000,Groetsch1984,Groetsch1993,Hansen1998,Linz1979,Linz2003,Wahba1987}%
. Generally, a problem is \emph{well-posed} when the problem has a solution
that exists, is unique, and is stable. If one or more of these criteria are
not satisfied, the problem is ill-posed
\citep{Groetsch1993,Linz1979,Linz2003}%
.

We have determined that finding discriminant functions of minimum risk
classification systems involves resolving two separate---but
related---ill-posed problems, so that we need to formulate a well-posed
\emph{direct} problem \emph{and} a well-posed \emph{inverse} problem.

It is also well known that the structure and function of biological organisms
are \emph{intimately intertwined}
\citep
{blakeslee2008body,lipton2011biology,mate2011body,pert1997molecules,ravella2022silent,sternberg2001balance,van2014body}%
. Accordingly, we realize that the overall structure and behavior and
properties of any given system are also intimately intertwined, so that
solving a system identification problem generally involves finding the
solution to some deep-seated statistical dilemma---situated far beneath the surface.

It has also long been recognized that biological organisms act to
\emph{minimize energy}, so that a biological organism satisfies a state of
equilibrium known as \emph{homeostasis}---at which point internal, physical
and chemical conditions of the organism are \emph{stable}---such that the
organism is composed of \emph{resilient} \emph{interconnections}, so that the
organism exhibits \emph{resilience }%
\citep{pert1997molecules,ravella2022silent,sternberg2001balance}%
.

Moreover, it has long been recognized that physical systems act to
\emph{minimize energy}, so that a physical system locates a \textquotedblleft
point\textquotedblright\ or \textquotedblleft position\textquotedblright\ of
\emph{equilibrium}---at which point the overall structure and behavior and
properties of the system are \emph{stable}---such that the physical system is
composed of \emph{resilient} \emph{interconnections,} so that the system
exhibits \emph{resilience}. Thereby, physical systems exhibit less risky
behavior and are less likely to be subject to catastrophic system failure
\citep
{fowler1980statistical,gibbs2010elementary,hill1956statistical,jackson2000equilibrium,strang1986introduction,strang2007computational}%
.

We recognize that formulating and solving certain system identification
problems involves determining \emph{how} and \emph{why} a particular system
\emph{locates} a \emph{point} of \emph{equilibrium}---so that the energy
exhibited by the system is minimized in such a manner that the system
satisfies a state of equilibrium---at which point the overall structure and
behavior and properties of the system exhibit a maximum amount of stability,
such that the system is composed of resilient interconnections. Thereby, the
system exhibits resilience and a minimum amount of risky behavior.

So, how might we find statistical laws that determine the overall statistical
structure and behavior and properties of minimum risk classification systems
that exhibit the minimum probability of classification error?

Equally important, how might we find statistical laws that determine the
generalization behavior of machine learning algorithms that find target
functions of minimum risk classification systems?

We realize that the discovery of such laws involves the \emph{discovery} of a
mathematical \emph{system} that \emph{models} fundamental \emph{aspects} of a
minimum risk binary classification system.

To see this, take any given binary classification system that is subject
random vectors. Generally, we know that the\ binary classification system has
two main components: $\left(  1\right)  $ a discriminant function that assigns
random vectors to one of two possible classes; and $\left(  2\right)  $ a
decision boundary that partitions the decision space of the system into two,
disjoint decision regions---which may be contiguous or non-contiguous
\citep{Duda2001,Fukunaga1990,Srinath1996}%
.

We realize, however, that a fundamental \emph{component} is \emph{missing}.
Since a decision boundary is a \emph{geometric figure}, we realize that a
binary classification system must contain some kind of \emph{intrinsic
coordinate system}.

Thereby, we also realize that a decision boundary of a binary classification
system is essentially a novel type of \emph{geometric locus}---whose shape and
fundamental properties are regulated by certain statistical laws
\citep{Reeves2015resolving,Reeves2018design}%
.%

\citet{Naylor1971}
noted that: \textquotedblleft A truly amazing number of problems in science
and engineering can be fruitfully treated with geometric methods in Hilbert
spaces.\textquotedblright

Geometric methods in Hilbert spaces include a class of geometric methods known
as \textquotedblleft\emph{coordinate geometry},\textquotedblright\ wherein
certain types of geometric problems are treated by a \emph{system of
coordinates}, such that each point of a geometric figure is uniquely specified
by a set of numerical coordinates, so that any given point of the geometric
figure satisfies certain conditions that are determined by an intrinsic
coordinate system---which is an inherent part of an algebraic equation.

Thereby, \emph{algebraic methods} are a means to the \emph{solution} of
certain types of geometric problems known as \emph{geometric locus problems},
where a geometric locus is a certain curve or surface that is formed by
specific points---each of which possesses some uniform property that is common
to all points that lie on the curve or surface---and \emph{no other points }%
\citep{Eisenhart1939,Hewson2009, Nichols1893,Tanner1898}%
.

We resolve the fundamental problem of finding discriminant functions of
minimum risk binary classification systems by devising novel geometric
\emph{locus methods} in Hilbert\emph{\ }spaces---\emph{within statistical
frameworks}---that fruitfully treat fundamental locus problems in binary
classification, where the Hilbert spaces are reproducing kernel Hilbert spaces
that have certain reproducing kernels.

In this treatise, we devise a \emph{mathematical system} whose statistical
structure and behavior and properties \emph{models} fundamental aspects of a
minimum risk binary classification system---which is subject to random
vectors. The model represents a discriminant function, a decision boundary, an
exclusive principal eigen-coordinate system and an eigenaxis of
symmetry---that spans the decision space---of a minimum risk binary
classification system, so that the exclusive principal eigen-coordinate system
connects the discriminant function to the decision boundary of the system, at
which point the discriminant function, the exclusive principal
eigen-coordinate system and the eigenaxis of symmetry are each represented by
a geometric locus of a novel principal eigenaxis---which has the structure of
a dual locus of likelihood components and principal eigenaxis components.

We use the model of a minimum risk binary classification system that is
outlined above to explain how a discriminant function \emph{extrapolates}%
---and thereby \emph{generalizes} in a \emph{significant manner}. We also use
the model to explain how a minimum risk binary classification system acts to
\emph{minimize} its \emph{risk}. Even more, we use the model to \emph{predict
error rates} exhibited by minimum risk binary classification systems.

Equally important, we use the model to \emph{predict behavior} that we have
not been aware of. We use the model to predict that a minimum risk binary
classification system acts to jointly minimize\ its eigenenergy and risk by
locating a point of equilibrium, at which point critical minimum eigenenergies
exhibited by the system are symmetrically concentrated in such a manner that
the dual locus of the discriminant function of the system is \emph{in}
statistical equilibrium---\emph{at} the geometric locus of the decision
boundary of the system, so that counteracting and opposing forces and
influences of the system are symmetrically balanced with each other---about
the geometric center of the locus of the novel principal eigenaxis of the
system---whereon the statistical fulcrum of the system is located.

Most importantly, we use the model to devise a \emph{mathematical framework}
for both the direct problem and the inverse problem of the binary
classification of random vectors. Accordingly, we use the model to formulate a
well-posed direct problem and a well-posed inverse problem---for the binary
classification of random vectors.

We devise a\ theoretical model and an applied model of a minimum risk binary
classification system that are both determined by a general locus formula for
finding discriminant functions of minimum risk binary classification systems,
so that a discriminant function of a minimum risk binary classification system
has a certain statistical structure and exhibits certain statistical behavior
and properties.

It will be seen that the theoretical model expresses fundamental laws of
binary classification, whereas the applied model explains and executes these laws.

We develop a general locus formula for finding discriminant functions of
minimum risk binary classification systems that has the general form of a
system of fundamental locus equations of binary classification, subject to
distinctive geometrical and statistical conditions for a minimum risk binary
classification system in statistical equilibrium, so that certain random
vectors have coordinates that are solutions of the locus equations.

Thereby, we formulate the direct problem of the binary classification of
random vectors according to a theoretical model---based on first
principles---that expresses fundamental laws of binary classification that
discriminant functions of minimum risk binary classification systems are
subject to. Accordingly, we derive the statistical structure of a target
function of a minimum risk binary classification system.

It will be seen that the general \emph{locus formula} for finding discriminant
functions determines mathematical conditions that \emph{statistically}
\emph{pre-wire} important \emph{generalizations} within the \emph{discriminant
function} of a minimum risk binary classification system---so that the
discriminant function generalizes and thereby \emph{extrapolates} in a
significant manner.

Most surprisingly, we derive the general locus formula that resolves the
direct problem of the binary classification of random vectors by enlarging the
complexity of a likelihood ratio test---that is based on the \emph{maximum
likelihood criterion}---at which point the likelihood ratio test constitutes a
\emph{well-posed variant }of \emph{\textquotedblleft Bayes' decision
rule\textquotedblright}\ for binary classification systems.

We also develop a constrained optimization algorithm that uses certain random
vectors to infer the values of the parameters characterizing a discriminant
function of a minimum risk binary classification system. Most remarkably, the
constrained optimization algorithm \emph{finds a system }of fundamental
\emph{locus equations} of binary classification, subject to distinctive
geometrical and statistical conditions for a minimum risk binary
classification system in statistical equilibrium---that is satisfied by
certain random vectors---such that data-driven versions of the general forms
of the fundamental locus equations are determined by distinctive algebraic and
geometric interconnections between all of the random vectors and the
components of the minimum risk binary classification system.

Equally remarkable, the constrained optimization algorithm finds discriminant
functions---of minimum risk binary classification systems---by
\emph{executing} a \emph{novel principal eigen-coordinate transform algorithm}.

Thereby, we formulate the inverse problem of the binary classification of
random vectors according to a constructive proof that demonstrates how a
well-posed constrained optimization algorithm executes the fundamental laws of
binary classification expressed in the direct problem---at which point the
constrained optimization algorithm executes precise mathematical conditions
that \emph{statistically pre-wire} important \emph{generalizations} within the
\emph{discriminant function} of a minimum risk binary classification
system---so that the discriminant function \emph{generalizes} and thereby
extrapolates in a significant manner.

Accordingly, we derive the process by which a well-posed constrained
optimization algorithm determines the\ statistical structure of a target
function of a minimum risk binary classification system.

Most surprisingly, we derive the general locus formula that resolves the
inverse problem of the binary classification of random vectors by identifying
novel and extremely unobvious processes---which include a novel principal
eigen-coordinate transform algorithm---that are executed by a \emph{well-posed
variant} of the constrained optimization \emph{algorithm} that is \emph{used}
by \emph{support vector machines} to learn nonlinear decision boundaries.

As a final point,
\citet{Keener2000}
noted that: \textquotedblleft For many of the problems we encounter in the
sciences, there is a natural way to represent the solution that transforms the
problem into a substantially easier one.\textquotedblright

We demonstrate that the constrained optimization algorithm that resolves the
inverse problem of the binary classification of random vectors executes novel
and elegant processes---which include a novel principal eigen-coordinate
transform algorithm---that represent the solution for finding discriminant
functions of minimum risk binary classification systems, at which point the
direct problem is transformed into a feasible one.

Thereby, we demonstrate that the general problem of the binary classification
of random vectors is essentially a deep-seated locus problem in binary
classification---situated far beneath the surface---at which point underlying
aspects of the general problem are subtle and extremely unobvious conditions.

\subsection{Outline of the Paper}

We treat the direct problem of the binary classification of random vectors in
Sections \ref{Section 2} - \ref{Section 10} of our treatise. In Section
\ref{Section 2}, we identify inconsistencies in Bayes' decision rule. We
express these inconsistencies by Theorem \ref{Bayes' Decision Rule Theorem}
and Corollaries \ref{Bayes' Multiclass Corollary} -
\ref{Bayes' Risk Multiclass Corollary}. In Section \ref{Section 3}, we develop
first principles of binary classification systems. We express these principles
by Axioms \ref{Likelihood Value Axiom} -
\ref{Minimum Risk Decision Rule Axiom}, Theorem
\ref{Basis of Locus Formula Theorem} and Corollaries
\ref{Form of Decision Boundary Corollary} -
\ref{Secondary Integral Equation Corollary}. In Section \ref{Section 4}, we
identify novel geometric locus problems in binary classification. In Section
\ref{Section 5}, we begin the development of novel geometric locus methods
that fruitfully treat locus problems in binary classification.

In Section \ref{Section 6}, we continue the development of novel geometric
locus methods that fruitfully treat fundamental locus problems in binary
classification---in accordance with certain mathematical aspects of exclusive
principal eigen-coordinate systems---that are inherent parts of vector algebra
locus equations. We express these mathematical aspects by Lemma
\ref{Equivalent Form of Algebraic Equation Lemma} and Theorems
\ref{Vector Algebra Equation of Linear Loci Theorem} -
\ref{Vector Algebra Equation of Spherical Loci Theorem}.

In Sections \ref{Section 7} and \ref{Section 8}, we identify how to represent
the solution of a fundamental and deep-seated locus problem in binary
classification---which we express by Theorem
\ref{Principal Eigen-coordinate System Theorem}, Corollary
\ref{Principal Eigen-coordinate System Corollary} and Theorem
\ref{Geometric Locus of a Novel Principal Eigenaxis Theorem}. In Section
\ref{Section 8}, we also consider the algebraic and geometrical significance
of reproducing kernels---which are seen to be fundamental components of
minimum risk binary classification systems. We use these results to develop a
novel principal eigen-coordinate transform algorithm that we use to find
discriminant functions of minimum risk binary classification systems.

In Section \ref{Section 9}, we outline the process by which a well-posed
constrained optimization algorithm resolves what we consider to be the most
difficult problem in binary classification---at which point a novel principal
eigen-coordinate transform algorithm is used to find discriminant functions of
minimum risk binary classification systems. In Section \ref{Section 10}, we
develop locus equations of binary classification. In Section \ref{Section 11},
we present a general locus formula that resolves the direct problem of the
binary classification of random vectors---in terms of an existence
theorem---which we express by Theorem
\ref{Direct Problem of Binary Classification Theorem}.

We treat the inverse problem of the binary classification of random vectors in
Sections \ref{Section 12} - \ref{Section 23} of our treatise. In Section
\ref{Section 12}, we present a detailed overview of the constrained
optimization algorithm that resolves the inverse problem of the binary
classification of random vectors. In Section \ref{Section 13}, we demonstrate
that solutions of the constrained optimization algorithm---which are based on
eigenstructure deficiencies---are generally ill-posed and ill-conditioned, so
the algorithm must be constrained in a certain manner.

In Section \ref{Section 14}, we present an overview of statistical
relations---within Hilbert spaces and reproducing kernel Hilbert spaces---that
determine pointwise covariance statistics, joint covariance statistics\ and
conditional distributions for individual random vectors. In Section
\ref{Section 15}, we examine the core of the machine learning algorithm that
finds discriminant functions of minimum risk binary classification systems. In
Section \ref{Section 16}, we examine how the decision space of a minimum risk
binary classification system is partitioned. In Section \ref{Section 17}, we
examine elegant statistical balancing acts---inside a certain principal
eigenspace---that are coincident with a minimum risk binary classification
system acting to jointly minimize its eigenenergy and risk.

In Section \ref{Section 18}, we examine dual capacities of discriminant
functions of minimum risk binary classification systems. In Section
\ref{Section 19}, we examine how a discriminant function extrapolates---and
thereby generalizes in a very nontrivial manner. In Section \ref{Section 20},
we examine the action taken by a minimum risk binary classification system to
jointly minimize its eigenenergy and risk. In Section \ref{Section 21}, we
identify critical interconnections---between the intrinsic components of a
minimum risk binary classification system---that determine the statistical
structure and the functionality of the discriminant function of the system. We
express these critical interconnections by Theorem
\ref{Principal Eigenstructures Theorem}.

In Section \ref{Section 22}, we present an overview of a constructive proof
that demonstrates how a well-posed constrained optimization algorithm executes
the fundamental laws of binary classification expressed by Theorem
\ref{Direct Problem of Binary Classification Theorem}. In Section
\ref{Section 23}, we present a general locus formula that resolves the inverse
problem of the binary classification of random vectors---which we have
obtained by a constructive proof---that we express by Theorem
\ref{Inverse Problem of Binary Classification Theorem}.

We summarize our major findings for the fundamental problem of the binary
classification of random vectors in Section \ref{Section 24}. Lastly, in
Section \ref{Section 25}, we present new insights into fundamental issues in
data-driven modeling and machine learning applications.

\subsection{Preliminary Remarks}

We have discovered most of the findings that are presented in this treatise
throughout the past decade. Our findings are based on constructing useful
combinations between known and recently discovered mathematical
entities---that reveal unsuspected relations between certain elements borrowed
from widely separated domains---where certain well-known elements from the
widely separated domains are wrongly believed to be unrelated to each other.
It is worth noting that our findings are related to one another in an
interdependent hierarchy, so that interdependent previous results are used to
obtain successive results.

We have one last remark before we begin our treatise on the binary
classification of random vectors. Given the deep cultural divide between the
\textquotedblleft data modeling\textquotedblright\ community\ and the
\textquotedblleft algorithmic modeling\textquotedblright\ community\ that is
described by Leo Breiman in
\citep{Breiman1991}%
, we consider it to be markedly renewing that statistical modeling\ approaches
of Bayesian decision theory\ and algorithmic modeling\ approaches of
statistical learning theory \emph{meet on significant points of binary
classification}, at which point fundamental laws of binary
classification---that have been determined by enlarging the complexity of a
well-posed variant of Bayes' decision rule---are effectively executed by a
well-posed variant of the constrained optimization algorithm---that is used by
support vector machines to learn nonlinear decision boundaries.

We begin our treatise on the binary classification of random vectors by
identifying inconsistencies in Bayes' decision rule for binary and multiclass
classification systems, such that identical random vectors generated by
distinct probability density functions account for the same effect exhibited
by a binary classification system.

Thereby, we demonstrate that Bayes' decision rule constitutes an ill-posed
rule for the direct problem of the binary classification of random
vectors---at which point the direct problem is recognized to be an ill-posed problem.

\section{\label{Section 2}Inconsistencies in Bayes' Decision Rule}

Bayes' decision rule for binary classification systems%
\begin{equation}
\Lambda\left(  \mathbf{x}\right)  \triangleq\frac{p\left(  \mathbf{x}%
|\omega_{1}\right)  }{p\left(  \mathbf{x}|\omega_{2}\right)  }\overset{\omega
_{1}}{\underset{\omega_{2}}{\gtrless}}\frac{P\left(  \omega_{2}\right)
\left(  C_{12}-C_{22}\right)  }{P\left(  \omega_{1}\right)  \left(
C_{21}-C_{11}\right)  } \tag{2.1}\label{Bayes' Decision Rule}%
\end{equation}
is widely-known for minimizing the probability of classification error for two
classes $\omega_{1}$ and $\omega_{2}$ of random vectors $\mathbf{x\in}$ $%
\mathbb{R}
^{d}$, where $\omega_{1}$ or $\omega_{2}$ is the true category, $P\left(
\omega_{1}\right)  $ and $P\left(  \omega_{2}\right)  $ are prior
probabilities of class $\omega_{1}$ and class $\omega_{2}$, the scalars
$C_{12}$, $C_{22}$, $C_{21}$, and $C_{11}$ denote costs for right and wrong
decisions, and $\Lambda\left(  \mathbf{x}\right)  $ denotes the likelihood
ratio $\frac{p\left(  \mathbf{x}|\omega_{1}\right)  }{p\left(  \mathbf{x}%
|\omega_{2}\right)  }$ of the system, wherein $p\left(  \mathbf{x}|\omega
_{1}\right)  $ and $p\left(  \mathbf{x}|\omega_{2}\right)  $ are
\textquotedblleft class-conditional\textquotedblright\ probability density
functions of the two classes of random vectors
\citep
{Duda2001,Fukunaga1990,Poor1996,Srinath1996,stark1994probability,VanTrees1968}%
.

Any given probability density function $p\left(  \mathbf{x}|\omega_{1}\right)
$ or $p\left(  \mathbf{x}|\omega_{2}\right)  $ in Bayes' decision rule
$\frac{p\left(  \mathbf{x}|\omega_{1}\right)  }{p\left(  \mathbf{x}|\omega
_{2}\right)  }\overset{\omega_{1}}{\underset{\omega_{2}}{\gtrless}}%
\frac{P\left(  \omega_{2}\right)  \left(  C_{12}-C_{22}\right)  }{P\left(
\omega_{1}\right)  \left(  C_{21}-C_{11}\right)  }$ represents a certain
probability law that governs how random vectors $\mathbf{x}$ $\in\omega_{1}$
and $\mathbf{x}$ $\in\omega_{2}$ generated by each respective probability
density function $\mathbf{x\sim}$ $p\left(  \mathbf{x}|\omega_{1}\right)  $
and $\mathbf{x\sim}$ $p\left(  \mathbf{x}|\omega_{2}\right)  $ are distributed
within certain regions $\mathcal{R}_{1}$ and $\mathcal{R}_{2}$ of Euclidean
space $%
\mathbb{R}
^{d}$, such that the regions $\mathcal{R}_{1}$ and $\mathcal{R}_{2}$ are
either overlapping with each other in some manner $\mathcal{R}_{1}%
\cap\mathcal{R}_{2}\neq\emptyset$, or the regions $\mathcal{R}_{1}$ and
$\mathcal{R}_{2}$ are disjoint $\mathcal{R}_{1}\cap\mathcal{R}_{2}=\emptyset$
\citep{parzen1962stochastic,Parzen1960}%
.

Bayes' risk for the binary classification system in
(\ref{Bayes' Decision Rule}) is given by the integral%
\begin{align}
\mathfrak{R}_{\mathfrak{B}}\left(  \Lambda\left(  \mathbf{x}\right)  \right)
&  \triangleq P\left(  \omega_{1}\right)  \left(  C_{21}-C_{11}\right)
\int_{-\infty}^{\eta}p\left(  \mathbf{x}|\omega_{1}\right)  d\mathbf{x}%
\tag{2.2}\label{Bayes' Risk}\\
&  \mathbf{+}P\left(  \omega_{2}\right)  \left(  C_{12}-C_{22}\right)
\int_{\eta}^{\infty}p\left(  \mathbf{x}|\omega_{2}\right)  d\mathbf{x}%
\text{,}\nonumber
\end{align}
over the decision space $Z=Z_{1}\cup Z_{2}$ of the system, where the decision
space $Z$ is defined over the interval $\left[  -\infty,\infty\right]  $, the
decision regions $Z_{1}$ and $Z_{2}$ are defined over the respective intervals
$\left[  \eta,\infty\right]  $ and $\left[  -\infty,\eta\right]  $, and $\eta$
is the decision threshold of the system, wherein $\eta=\frac{P\left(
\omega_{2}\right)  \left(  C_{12}-C_{22}\right)  }{P\left(  \omega_{1}\right)
\left(  C_{21}-C_{11}\right)  }$. The integral in (\ref{Bayes' Risk})
calculates the total probability that Bayes' decision rule chooses the wrong
class---which is also known as Bayes' error
\citep
{Duda2001,Fukunaga1990,Poor1996,Srinath1996,stark1994probability,VanTrees1968}%
.

Bayes' decision rule for binary classification systems is based on
modifications of Bayes' rule---also known as Bayes' theorem or Bayes' formula.
Bayes' theorem uses certain laws of probability to describe probabilities of
events that are possible outcomes of random experiments. Accordingly, a sample
space is defined to be a set of possible outcomes of a random experiment,
wherein an event is a subset of the sample space, such that any given event is
a collection of outcomes in the sample space
\citep{Hewson2009,Ross1998,ross2007introduction}%
.

\subsection{Bayes' Theorem}

Let the true categories $\omega_{1}$ and $\omega_{2}$ in Bayes' decision rule
be denoted by $\omega_{i}$ and $\omega_{j}$ respectively. The binary
classification rule in (\ref{Bayes' Decision Rule}) is based on modifications
of Bayes' rule---also known as Bayes' theorem or Bayes' formula%
\begin{align}
P\left(  \omega_{i}|\mathbf{x}\right)   &  =\frac{P\left(  \mathbf{x}%
|\omega_{i}\right)  P\left(  \omega_{i}\right)  }{P\left(  \mathbf{x}\right)
}\tag{2.3}\label{Bayes' Theorem}\\
&  =\frac{P\left(  \mathbf{x}|\omega_{i}\right)  P\left(  \omega_{i}\right)
}{\sum\nolimits_{j=1}^{2}P\left(  \mathbf{x}|\omega_{j}\right)  P\left(
\omega_{j}\right)  }\text{,}\nonumber
\end{align}
where $\omega_{i}$ and $\omega_{j}$ are nonoverlapping sets of events that
partition a sample space $S$ in such a manner that the events $\omega_{i}$ and
$\omega_{j}$ satisfy the following conditions: $\left(  1\right)  $
$S=\omega_{i}\cup\omega_{j}$; $\left(  2\right)  $ $\omega_{i}\cap\omega
_{j}=\emptyset$; and $\left(  3\right)  $ $P\left(  \omega_{i}\right)  >0$ and
$P\left(  \omega_{j}\right)  >0$.

Accordingly, only one of the events $\omega_{i}$ and $\omega_{j}$ in the
sample space $S$ occurs, so that sets of the events $\omega_{i}$ and
$\omega_{j}$ have no overlap $%
{\textstyle\bigcap\limits_{i=1}^{2}}
\omega_{i}=\emptyset$ and cover all possible outcomes in the sample space $S$,
at which point $P\left(  \omega_{i}\right)  +P\left(  \omega_{j}\right)  =1$.

Additionally, $\mathbf{x}$ is an event in the sample space $S$, at which point
a certain amount of overlap exists between the event $\mathbf{x}$ and each of
the individual pieces $\omega_{i}$ and $\omega_{j}$ forming the partition of
the sample space $S$. Accordingly, let $\mathbf{x}\cap\omega_{i}$ and
$\mathbf{x}\cap\omega_{j}$ denote the overlap between the event $\mathbf{x}$
and each of the individual events $\omega_{i}$ and $\omega_{j}$ in the sample
space $S$.

It follows that outcomes of the event $\mathbf{x}$ satisfy the condition $%
{\textstyle\bigcap\nolimits_{j=1}^{2}}
\mathbf{x}\omega_{j}=\emptyset$, at which point only one of the events
$\omega_{i}$ and $\omega_{j}$ occurs, so that outcomes $\mathbf{x}$ of the
events $\omega_{i}$ and $\omega_{j}$ have no overlap $%
{\textstyle\bigcap\limits_{i=1}^{2}}
\mathbf{x}\omega_{i}=\emptyset$ in the sample space $S$
\citep
{bernardo2009,berrar2018bayes,Bishop2006,Hewson2009,papoulis2002probability,rice1995mathematical,Ross1998,ross2007introduction}%
.

\subsection{The Law of Total Probability}

By Bayes' theorem in (\ref{Bayes' Theorem}), $P\left(  \omega_{i}%
|\mathbf{x}\right)  $ is the conditional probability of an event $\omega_{i}$
in a partition of a sample space $S$, given an event $\mathbf{x}$, such that
any given event $\mathbf{x}$ in the sample space $S$ satisfies the \emph{law
of total probability}%
\begin{equation}
P\left(  \mathbf{x}\right)  =%
{\textstyle\sum\nolimits_{i=1}^{2}}
P\left(  \mathbf{x}\cap\omega_{i}\right)  =%
{\textstyle\sum\nolimits_{i=1}^{2}}
P\left(  \mathbf{x}|\omega_{i}\right)  P\left(  \omega_{i}\right)  \text{,}
\tag{2.4}\label{The Law of Total Probability}%
\end{equation}
wherein sets of the events $\omega_{i}$ and $\omega_{j}$ have no overlap $%
{\textstyle\bigcap\limits_{i=1}^{2}}
\omega_{i}=\emptyset$, so that exactly one---and \emph{only one}---of the
events $\omega_{i}$ and $\omega_{j}$ occurs, at which point the event
$\mathbf{x}$ has a certain amount of overlap $\mathbf{x}\cap\omega_{i}$ and
$\mathbf{x}\cap\omega_{j}$ with each piece $\omega_{i}$ and $\omega_{j}$ of
the sample space $S$, such that outcomes $\mathbf{x}$ of the events
$\omega_{i}$ and $\omega_{j}$ have no overlap $%
{\textstyle\bigcap\limits_{i=1}^{2}}
\mathbf{x}\omega_{i}=\emptyset$ in the sample space $S$
\citep
{berrar2018bayes,papoulis2002probability,rice1995mathematical,Ross1998,ross2007introduction}%
.

We can use the law of total probability in (\ref{The Law of Total Probability}%
) to determine the probability $P\left(  \mathbf{x}\right)  $ of an event
$\mathbf{x}$ by evaluating the partition of the sample space $S$ that the
event $\mathbf{x}$ occurs in. Accordingly, the probability $P\left(
\mathbf{x}\right)  $ of the event $\mathbf{x}$ is determined by the
probability of the overlap $P\left(  \mathbf{x}\cap\omega_{i}\right)  $ and
$P\left(  \mathbf{x}\cap\omega_{j}\right)  $ between the event $\mathbf{x}$
and each of the individual pieces $\omega_{i}$ and $\omega_{j}$ forming the
partition of the sample space $S$.

Thus, we add the amount of probabilities $P\left(  \mathbf{x}|\omega
_{i}\right)  P\left(  \omega_{i}\right)  $ and $P\left(  \mathbf{x}|\omega
_{j}\right)  P\left(  \omega_{j}\right)  $ of the event $\mathbf{x}$ that fall
within each piece $\omega_{i}$ and $\omega_{j}$ of the sample space $S$, such
that the probability $P\left(  \mathbf{x}\right)  $ of the event $\mathbf{x}$
is given by the equation $P\left(  \mathbf{x}\right)  =%
{\textstyle\sum\nolimits_{i=1}^{2}}
P\left(  \mathbf{x}|\omega_{i}\right)  P\left(  \omega_{i}\right)  $.

\subsection{Using Bayes' Formula}

If the events $\omega_{i}$ and $\omega_{j}$ in Bayes' theorem are regarded as
\textquotedblleft causes,\textquotedblright\ then Bayes' formula in
(\ref{Bayes' Theorem}) can be regarded as a formula for the probability that
an event $\mathbf{x}$---which has occurred---is the result of a certain
cause.\ Therefore, if the event $\omega_{i}$ is the cause of an event
$\mathbf{x}$, it follows that the event $\omega_{j}$ cannot be the\emph{
}cause of the event $\mathbf{x}$
\citep{Parzen1960}%
.

Accordingly, Bayes' formula in (\ref{Bayes' Theorem}) has been interpreted as
a formula for the probabilities of \textquotedblleft causes\textquotedblright%
\ or \textquotedblleft hypotheses.\textquotedblright\ The problem with this
interpretation, however, is that in many contexts, the probabilities in Bayes'
formula are rarely known, especially the unconditional probabilities $P\left(
\omega_{i}\right)  $ of the causes,\ which enter into the right hand side of
(\ref{Bayes' Theorem})
\citep{cowan2007data,Parzen1960}%
.

Even so, Bayes' theorem has important practical uses, especially in medical
applications that involve certain diagnostic tests, wherein conditional
probabilities of diagnostic tests for certain diseases or causes\ and
unconditional probabilities of the causes\ are both known
\citep{Parzen1960,Ross1998,ross2007introduction,stark1994probability}%
. Bayes' theorem, however, is difficult for clinicians to use---since the
theorem is so abstract
\citep{TIEMENS2020320}%
.

Bayes' theorem also has important practical uses in military applications. For
example, Alan Turing and other researchers used Bayes' theorem to crack the
Enigma Code---generated by a famous encryption machine---that was used by the
Germans during WWII to transmit coded messages. Bayes' theorem has also been
used to find Russian submarines
\citep{mcgrayne2011theory}%
.

Bayes' theorem has provoked much philosophical speculation---and has also been
the source of much controversy
\citep{cousins1995isn,cowan2007data,efron2005modern}%
. Therefore, any given application of Bayes' theorem must always be made with
conscious knowledge of just what model of reality Bayes' formula represents
\citep{drake1967fundamentals,Parzen1960}%
.

We realize that Bayes' theorem is a true theorem of mathematical probability.
However, we also realize that---before we apply Bayes' theorem---we need to
adhere to what
\citet{Parzen1960}
called the fundamental principle of applied probability: \textquotedblleft
Before applying a theorem, one must carefully ponder whether the hypotheses of
the theorem may be assumed to be satisfied.\textquotedblright

We have carefully considered whether the conditions of Bayes' theorem in
(\ref{Bayes' Theorem}) are satisfied by Bayes' decision rule in \textbf{(}%
\ref{Bayes' Decision Rule}\textbf{)}. Our findings are presented next.

\subsection{Warning of Inconsistencies in Bayes' Decision Rule}

We realize that Bayes' decision rule in \textbf{(}\ref{Bayes' Decision Rule}%
\textbf{)} is inconsistent with the conditions of Bayes' theorem in
(\ref{Bayes' Theorem}): Theorem \ref{Bayes' Decision Rule Theorem} provides us
with a warning of this inconsistency.

\subsubsection{Notation and Assumptions}

Let $\mathbf{x\sim}$ $p\left(  \mathbf{x}|\omega_{1}\right)  $ or
$\mathbf{x\sim}$ $p\left(  \mathbf{x}|\omega_{2}\right)  $ denote an event
that a random vector $\mathbf{x\in}$ $%
\mathbb{R}
^{d}$ is generated by a respective class-conditional probability density
function $p\left(  \mathbf{x}|\omega_{1}\right)  $ or $p\left(  \mathbf{x}%
|\omega_{2}\right)  $, such that the output $\mathbf{x\sim}$ $p\left(
\mathbf{x}|\omega_{1}\right)  $ or $\mathbf{x\sim}$ $p\left(  \mathbf{x}%
|\omega_{2}\right)  $ of the probability density function $p\left(
\mathbf{x}|\omega_{1}\right)  $ or $p\left(  \mathbf{x}|\omega_{2}\right)  $
is caused by an event $\omega_{1}$ or $\omega_{2}$ that occurs whenever the
corresponding event $\mathbf{x\sim}$ $p\left(  \mathbf{x}|\omega_{1}\right)  $
or $\mathbf{x\sim}$ $p\left(  \mathbf{x}|\omega_{2}\right)  $ occurs.

Accordingly, let any given output $\mathbf{x\sim}$ $p\left(  \mathbf{x}%
|\omega_{1}\right)  $ of the probability density function $p\left(
\mathbf{x}|\omega_{1}\right)  $ be caused by an event $\omega_{1}$ that
occurs, at which point the event $\omega_{1}$ occurs whenever the event
$\mathbf{x\sim}$ $p\left(  \mathbf{x}|\omega_{1}\right)  $ occurs.
Correspondingly, let any given output $\mathbf{x\sim}$ $p\left(
\mathbf{x}|\omega_{2}\right)  $ of the probability density function $p\left(
\mathbf{x}|\omega_{2}\right)  $ be caused by an event $\omega_{2}$ that
occurs, at which point the event $\omega_{2}$ occurs whenever the event
$\mathbf{x\sim}$ $p\left(  \mathbf{x}|\omega_{2}\right)  $ occurs.

\begin{theorem}
\label{Bayes' Decision Rule Theorem}Take any given binary classification
system subject to two categories $\omega_{1}$ and $\omega_{2}$ of random
vectors $\mathbf{x\in}$ $%
\mathbb{R}
^{d}$, such that the event $\omega_{1}$ or $\omega_{2}$ occurs whenever the
corresponding event $\mathbf{x\sim}$ $p\left(  \mathbf{x}|\omega_{1}\right)  $
or $\mathbf{x\sim}$ $p\left(  \mathbf{x}|\omega_{2}\right)  $ occurs, that is
determined by Bayes' decision rule $\frac{p\left(  \mathbf{x}|\omega
_{1}\right)  }{p\left(  \mathbf{x}|\omega_{2}\right)  }\overset{\omega
_{1}}{\underset{\omega_{2}}{\gtrless}}\frac{P\left(  \omega_{2}\right)
\left(  C_{12}-C_{22}\right)  }{P\left(  \omega_{1}\right)  \left(
C_{21}-C_{11}\right)  }$, where $p\left(  \mathbf{x}|\omega_{1}\right)  $ and
$p\left(  \mathbf{x}|\omega_{2}\right)  $ are class-conditional probability
density functions of the two categories $\omega_{1}$ and $\omega_{2}$ of
random vectors $\mathbf{x\in}$ $%
\mathbb{R}
^{d}$, $\omega_{1}$ or $\omega_{2}$ is the true category, $P\left(  \omega
_{1}\right)  $ and $P\left(  \omega_{2}\right)  $ are prior probabilities of
class $\omega_{1}$ and class $\omega_{2}$, and the scalars $C_{12}$, $C_{22}$,
$C_{21}$, and $C_{11}$ denote costs for right and wrong decisions.

Let $C_{11}=C_{22}=0$ and $C_{21}=C_{12}=1$, so that right decisions have a
value or a cost of $0$, and wrong decisions have a unit value or a cost of
$1$, at which point Bayes' decision rule has the form $\frac{p\left(
\mathbf{x}|\omega_{1}\right)  }{p\left(  \mathbf{x}|\omega_{2}\right)
}\overset{\omega_{1}}{\underset{\omega_{2}}{\gtrless}}\frac{P\left(
\omega_{2}\right)  }{P\left(  \omega_{1}\right)  }$, where $P\left(
\omega_{1}\right)  +P\left(  \omega_{2}\right)  =1$.

Bayes' decision rule $\frac{p\left(  \mathbf{x}|\omega_{1}\right)  }{p\left(
\mathbf{x}|\omega_{2}\right)  }\overset{\omega_{1}}{\underset{\omega
_{2}}{\gtrless}}\frac{P\left(  \omega_{2}\right)  }{P\left(  \omega
_{1}\right)  }$ does not satisfy the law of total probability%
\begin{align*}
P\left(  \mathbf{x}\right)   &  =%
{\textstyle\sum\nolimits_{i=1}^{2}}
P\left(  \mathbf{x}\cap\omega_{i}\right) \\
&  =%
{\textstyle\sum\nolimits_{i=1}^{2}}
P\left(  \mathbf{x|}\omega_{i}\right)  P\left(  \omega_{i}\right)
\end{align*}
since the probability $P\left(  \mathbf{x}\right)  $ of any given event
$\mathbf{x}$, such that $%
{\textstyle\bigcap\limits_{i=1}^{2}}
\mathbf{x}\omega_{i}\neq\emptyset$, is determined by the equation%
\begin{align*}
P\left(  \mathbf{x}\right)   &  =P\left(
{\textstyle\bigcup\limits_{i=1}^{2}}
\mathbf{x}\omega_{i}\right) \\
&  =%
{\textstyle\sum\nolimits_{i=1}^{2}}
P\left(  \mathbf{x}\cap\omega_{i}\right)  -P\left(
{\textstyle\bigcap\limits_{i=1}^{2}}
\mathbf{x}\omega_{i}\right)  \text{,}%
\end{align*}
at which point%
\[
P\left(  \mathbf{x}\right)  \neq%
{\textstyle\sum\nolimits_{i=1}^{2}}
P\left(  \mathbf{x|}\omega_{i}\right)  P\left(  \omega_{i}\right)  \text{,}%
\]
so the probability $P\left(  \mathbf{x}\right)  $ of the event $\mathbf{x}$
does not satisfy the law of total probability $P\left(  \mathbf{x}\right)  =%
{\textstyle\sum\nolimits_{i=1}^{2}}
P\left(  \mathbf{x|}\omega_{i}\right)  P\left(  \omega_{i}\right)  $.

Therefore, Bayes' decision rule $\frac{p\left(  \mathbf{x}|\omega_{1}\right)
}{p\left(  \mathbf{x}|\omega_{2}\right)  }\overset{\omega_{1}%
}{\underset{\omega_{2}}{\gtrless}}\frac{P\left(  \omega_{2}\right)  }{P\left(
\omega_{1}\right)  }$ does not satisfy the conditions of Bayes' theorem
$P\left(  \omega_{i}|\mathbf{x}\right)  =\frac{P\left(  \mathbf{x}|\omega
_{i}\right)  P\left(  \omega_{i}\right)  }{\sum\nolimits_{j=1}^{2}P\left(
\mathbf{x}|\omega_{j}\right)  P\left(  \omega_{j}\right)  }$ since%
\[
P\left(  \omega_{i}|\mathbf{x}\right)  \neq\frac{P\left(  \mathbf{x}%
|\omega_{i}\right)  P\left(  \omega_{i}\right)  }{\sum\nolimits_{j=1}%
^{2}P\left(  \mathbf{x}|\omega_{j}\right)  P\left(  \omega_{j}\right)
}\text{,}%
\]
as well as the conditions of the modification of Bayes' theorem $P\left(
\omega_{i}|\mathbf{x}\right)  =\frac{p\left(  \mathbf{x}|\omega_{i}\right)
P\left(  \omega_{i}\right)  }{\sum\nolimits_{j=1}^{2}p\left(  \mathbf{x}%
|\omega_{j}\right)  P\left(  \omega_{j}\right)  }$ since%
\[
P\left(  \omega_{i}|\mathbf{x}\right)  \neq\frac{p\left(  \mathbf{x}%
|\omega_{i}\right)  P\left(  \omega_{i}\right)  }{\sum\nolimits_{j=1}%
^{2}p\left(  \mathbf{x}|\omega_{j}\right)  P\left(  \omega_{j}\right)
}\text{.}%
\]

\end{theorem}

\begin{proof}
Take Bayes' decision rule $\frac{p\left(  \mathbf{x}|\omega_{1}\right)
}{p\left(  \mathbf{x}|\omega_{2}\right)  }\overset{\omega_{1}%
}{\underset{\omega_{2}}{\gtrless}}\frac{P\left(  \omega_{2}\right)  \left(
C_{12}-C_{22}\right)  }{P\left(  \omega_{1}\right)  \left(  C_{21}%
-C_{11}\right)  }$ for any given binary classification system that
discriminates between two classes $\omega_{1}$ and $\omega_{2}$ of random
vectors $\mathbf{x\in}$ $%
\mathbb{R}
^{d}$, where $\omega_{1}$ or $\omega_{2}$ is the true category, $P\left(
\omega_{1}\right)  $ and $P\left(  \omega_{2}\right)  $ are prior
probabilities of class $\omega_{1}$ and class $\omega_{2}$, the scalars
$C_{12}$, $C_{22}$, $C_{21}$, and $C_{11}$ denote costs for right and wrong
decisions, and $\frac{p\left(  \mathbf{x}|\omega_{1}\right)  }{p\left(
\mathbf{x}|\omega_{2}\right)  }$ denotes the likelihood ratio of the system,
wherein $p\left(  \mathbf{x}|\omega_{1}\right)  $ and $p\left(  \mathbf{x}%
|\omega_{2}\right)  $ are class-conditional probability density functions of
the random vectors $\mathbf{x\in}$ $%
\mathbb{R}
^{d}$.

Let $C_{11}=C_{22}=0$\ and $C_{21}=C_{12}=1$, so that right decisions have a
value or a cost of $0$, and wrong decisions have a unit value or a cost of
$1$, at which point Bayes' decision rule has the form $\frac{p\left(
\mathbf{x}|\omega_{1}\right)  }{p\left(  \mathbf{x}|\omega_{2}\right)
}\overset{\omega_{1}}{\underset{\omega_{2}}{\gtrless}}\frac{P\left(
\omega_{2}\right)  }{P\left(  \omega_{1}\right)  }$, where $P\left(
\omega_{1}\right)  +P\left(  \omega_{2}\right)  =1$.

Now let the event that a random vector $\mathbf{x}$ is generated by $p\left(
\mathbf{x}|\omega_{1}\right)  $ or $p\left(  \mathbf{x}|\omega_{2}\right)  $
be expressed by $\mathbf{x\sim}$ $p\left(  \mathbf{x}|\omega_{1}\right)  $ or
$\mathbf{x\sim}$ $p\left(  \mathbf{x}|\omega_{2}\right)  $ respectively, at
which point any given output $\mathbf{x\sim}$ $p\left(  \mathbf{x}|\omega
_{1}\right)  $ or $\mathbf{x\sim}$ $p\left(  \mathbf{x}|\omega_{2}\right)  $
of each respective probability density function $p\left(  \mathbf{x}%
|\omega_{1}\right)  $ or $p\left(  \mathbf{x}|\omega_{2}\right)  $ is caused
by an event $\omega_{1}$ or $\omega_{2}$ that occurs whenever the
corresponding event $\mathbf{x\sim}$ $p\left(  \mathbf{x}|\omega_{1}\right)  $
or $\mathbf{x\sim}$ $p\left(  \mathbf{x}|\omega_{2}\right)  $ occurs\textbf{.}

Accordingly, let $\omega_{1}$ and $\omega_{2}$ be sets of events within the
decision space $Z=Z_{1}\cup Z_{2}$ of the binary classification system
$\frac{p\left(  \mathbf{x}|\omega_{1}\right)  }{p\left(  \mathbf{x}|\omega
_{2}\right)  }\overset{\omega_{1}}{\underset{\omega_{2}}{\gtrless}}%
\frac{P\left(  \omega_{2}\right)  }{P\left(  \omega_{1}\right)  }$, where the
decision space $Z$ is defined over the interval $\left[  -\infty
,\infty\right]  $, and the decision regions $Z_{1}$ and $Z_{2}$ are defined
over the respective intervals $\left[  \eta,\infty\right]  $ and $\left[
-\infty,\eta\right]  $, wherein $\eta=\frac{P\left(  \omega_{2}\right)
}{P\left(  \omega_{1}\right)  }$.

Next, let $p\left(  \mathbf{x}|\omega_{1}\right)  $ and $p\left(
\mathbf{x}|\omega_{2}\right)  $ determine overlapping distributions of random
points $\mathbf{x\in}$ $%
\mathbb{R}
^{d}$, such that $\mathbf{x}\cap\omega_{1}=\mathbf{x}\cap\omega_{2}$ in the
event that $\mathbf{x\sim}$ $p\left(  \mathbf{x}|\omega_{1}\right)
\equiv\mathbf{x\sim}$ $p\left(  \mathbf{x}|\omega_{2}\right)  $ in Euclidean
space $%
\mathbb{R}
^{d}$, at which point both of the events $\omega_{1}$ and $\omega_{2}$ are
causes of an identical event $\mathbf{x\sim}$ $p\left(  \mathbf{x}|\omega
_{1}\right)  \equiv\mathbf{x\sim}$ $p\left(  \mathbf{x}|\omega_{2}\right)  $
that occurs.

It follows that both of the events $\omega_{1}$ and $\omega_{2}$ occur in such
a manner that sets of the events $\omega_{1}$ and $\omega_{2}$ are overlapping
$%
{\textstyle\bigcap\limits_{i=1}^{2}}
\omega_{i}\neq\emptyset$ within the decision space $Z=Z_{1}\cup Z_{2}$ of the
binary classification system $\frac{p\left(  \mathbf{x}|\omega_{1}\right)
}{p\left(  \mathbf{x}|\omega_{2}\right)  }\overset{\omega_{1}%
}{\underset{\omega_{2}}{\gtrless}}\frac{P\left(  \omega_{2}\right)  }{P\left(
\omega_{1}\right)  }$, at which point sets of the events $\omega_{1}$ and
$\omega_{2}$ have overlapping outcomes $%
{\textstyle\bigcap\limits_{i=1}^{2}}
\mathbf{x}\omega_{i}\neq\emptyset$ within the decision space $Z=Z_{1}\cup
Z_{2}$.

Since sets of the events $\omega_{1}$ and $\omega_{2}$ are not mutually
exclusive $%
{\textstyle\bigcap\limits_{i=1}^{2}}
\omega_{i}\neq\emptyset$, it follows that%
\[
P\left(
{\textstyle\bigcup\limits_{i=1}^{2}}
\omega_{i}\right)  =\sum\nolimits_{i=1}^{2}P\left(  \omega_{i}\right)
-P\left(
{\textstyle\bigcap\limits_{i=1}^{2}}
\omega_{i}\right)  \text{,}%
\]
at which point%
\[
P\left(
{\textstyle\bigcup\limits_{i=1}^{2}}
\omega_{i}\right)  \neq\sum\nolimits_{i=1}^{2}P\left(  \omega_{i}\right)
\text{.}%
\]

Therefore, sets of the events $\omega_{1}$ and $\omega_{2}$ do not form a
partition of the decision space $Z=Z_{1}\cup Z_{2}$ of the binary
classification system $\frac{p\left(  \mathbf{x}|\omega_{1}\right)  }{p\left(
\mathbf{x}|\omega_{2}\right)  }\overset{\omega_{1}}{\underset{\omega
_{2}}{\gtrless}}\frac{P\left(  \omega_{2}\right)  }{P\left(  \omega
_{1}\right)  }$, such that sets of the events $\omega_{1}$ and $\omega_{2}$
have no overlap within the decision space $Z=Z_{1}\cup Z_{2}$%
\[%
{\textstyle\bigcap\limits_{i=1}^{2}}
\omega_{i}=\emptyset\text{,}%
\]
and sets of the events $\omega_{1}$ and $\omega_{2}$ collectively cover all
possible outcomes within the decision space $Z=Z_{1}\cup Z_{2}$%
\[%
{\textstyle\bigcup\limits_{i=1}^{2}}
\omega_{i}=\omega_{1}\cup\omega_{2}\text{,}%
\]
since sets of the events $\omega_{1}$ and $\omega_{2}$ have overlap within the
decision space $Z=Z_{1}\cup Z_{2}$%
\[%
{\textstyle\bigcap\limits_{i=1}^{2}}
\omega_{i}\neq\emptyset\text{,}%
\]
and sets of the events $\omega_{1}$ and $\omega_{2}$ do not collectively cover
all possible outcomes within the decision space $Z=Z_{1}\cup Z_{2}$%
\[%
{\textstyle\bigcup\limits_{i=1}^{2}}
\omega_{i}\neq\omega_{1}\cup\omega_{2}\text{.}%
\]

Moreover, since sets of the events $\omega_{1}$ and $\omega_{2}$ have
overlapping outcomes $%
{\textstyle\bigcap\limits_{i=1}^{2}}
\mathbf{x}\omega_{i}\neq\emptyset$ within the decision space $Z=Z_{1}\cup
Z_{2}$ of the binary classification system $\frac{p\left(  \mathbf{x}%
|\omega_{1}\right)  }{p\left(  \mathbf{x}|\omega_{2}\right)  }\overset{\omega
_{1}}{\underset{\omega_{2}}{\gtrless}}\frac{P\left(  \omega_{2}\right)
}{P\left(  \omega_{1}\right)  }$, it follows that the probability $P\left(
\mathbf{x}\right)  $ of any given event $\mathbf{x}$, such that $%
{\textstyle\bigcap\limits_{i=1}^{2}}
\mathbf{x}\omega_{i}\neq\emptyset$, is determined by the equation%
\begin{align*}
P\left(  \mathbf{x}\right)   &  =P\left(
{\textstyle\bigcup\limits_{i=1}^{2}}
\mathbf{x}\omega_{i}\right) \\
&  =\sum\nolimits_{i=1}^{2}P\left(  \mathbf{x}\cap\omega_{i}\right)  -P\left(
%
{\textstyle\bigcap\limits_{i=1}^{2}}
\mathbf{x}\omega_{i}\right)  \text{,}%
\end{align*}
at which point the probability $P\left(  \mathbf{x}\right)  $ of the event
$\mathbf{x}$ does not satisfy the law of total probability $P\left(
\mathbf{x}\right)  =%
{\textstyle\sum\nolimits_{i=1}^{2}}
P\left(  \mathbf{x|}\omega_{i}\right)  P\left(  \omega_{i}\right)  $.

Therefore, Bayes' decision rule $\frac{p\left(  \mathbf{x}|\omega_{1}\right)
}{p\left(  \mathbf{x}|\omega_{2}\right)  }\overset{\omega_{1}%
}{\underset{\omega_{2}}{\gtrless}}\frac{P\left(  \omega_{2}\right)  }{P\left(
\omega_{1}\right)  }$ does not satisfy the law of total probability%
\[
P\left(  \mathbf{x}\right)  =\sum\nolimits_{i=1}^{2}P\left(  \mathbf{x}%
\cap\omega_{i}\right)  =%
{\textstyle\sum\nolimits_{i=1}^{2}}
P\left(  \mathbf{x|}\omega_{i}\right)  P\left(  \omega_{i}\right)
\]
since for any given event $\mathbf{x}$, such that $%
{\textstyle\bigcap\limits_{i=1}^{2}}
\mathbf{x}\omega_{i}\neq\emptyset$, it follows that%
\[
P\left(
{\textstyle\bigcup\limits_{i=1}^{2}}
\mathbf{x}\omega_{i}\right)  \neq\sum\nolimits_{i=1}^{2}P\left(
\mathbf{x}\cap\omega_{i}\right)  \text{,}%
\]
at which point%
\begin{align*}
P\left(  \mathbf{x}\right)   &  \neq\sum\nolimits_{i=1}^{2}P\left(
\mathbf{x}\cap\omega_{i}\right) \\
&  \neq%
{\textstyle\sum\nolimits_{i=1}^{2}}
P\left(  \mathbf{x|}\omega_{i}\right)  P\left(  \omega_{i}\right)  \text{,}%
\end{align*}
so the probability $P\left(  \mathbf{x}\right)  $ of the event $\mathbf{x}$
does not satisfy the law of total probability $P\left(  \mathbf{x}\right)  =%
{\textstyle\sum\nolimits_{i=1}^{2}}
P\left(  \mathbf{x|}\omega_{i}\right)  P\left(  \omega_{i}\right)  $.

Therefore, the conditions of Bayes' theorem $P\left(  \omega_{i}%
|\mathbf{x}\right)  =\frac{P\left(  \mathbf{x}|\omega_{i}\right)  P\left(
\omega_{i}\right)  }{\sum\nolimits_{j=1}^{2}P\left(  \mathbf{x}|\omega
_{j}\right)  P\left(  \omega_{j}\right)  }$ are not satisfied since%
\[
P\left(  \omega_{i}|\mathbf{x}\right)  \neq\frac{P\left(  \mathbf{x}%
|\omega_{i}\right)  P\left(  \omega_{i}\right)  }{\sum\nolimits_{j=1}%
^{2}P\left(  \mathbf{x}|\omega_{j}\right)  P\left(  \omega_{j}\right)
}\text{.}%
\]

Moreover, the conditions of the modification of Bayes' theorem $P\left(
\omega_{i}|\mathbf{x}\right)  =\frac{p\left(  \mathbf{x}|\omega_{i}\right)
P\left(  \omega_{i}\right)  }{\sum\nolimits_{j=1}^{2}p\left(  \mathbf{x}%
|\omega_{j}\right)  P\left(  \omega_{j}\right)  }$ are not satisfied either
since%
\[
P\left(  \omega_{i}|\mathbf{x}\right)  \neq\frac{p\left(  \mathbf{x}%
|\omega_{i}\right)  P\left(  \omega_{i}\right)  }{\sum\nolimits_{j=1}%
^{2}p\left(  \mathbf{x}|\omega_{j}\right)  P\left(  \omega_{j}\right)
}\text{.}%
\]

Thereby, it is concluded that Bayes' decision rule $\frac{p\left(
\mathbf{x}|\omega_{1}\right)  }{p\left(  \mathbf{x}|\omega_{2}\right)
}\overset{\omega_{1}}{\underset{\omega_{2}}{\gtrless}}\frac{P\left(
\omega_{2}\right)  }{P\left(  \omega_{1}\right)  }$ does not satisfy the
conditions of Bayes' theorem, as well as the conditions of the modification of
Bayes' theorem.
\end{proof}

Figure $1$ illustrates the basis of the inconsistencies expressed by Theorem
\ref{Bayes' Decision Rule Theorem}, such that the events $\omega_{1}$ and
$\omega_{2}$ in Bayes' decision rule $\frac{p\left(  \mathbf{x}|\omega
_{1}\right)  }{p\left(  \mathbf{x}|\omega_{2}\right)  }\overset{\omega
_{1}}{\underset{\omega_{2}}{\gtrless}}\frac{P\left(  \omega_{2}\right)
}{P\left(  \omega_{1}\right)  }$ are both causes of identical events
$\mathbf{x\sim}$ $p\left(  \mathbf{x}|\omega_{1}\right)  \equiv\mathbf{x\sim}$
$p\left(  \mathbf{x}|\omega_{2}\right)  $ that occur, at which point sets of
the events $\omega_{1}$ and $\omega_{2}$ are overlapping $%
{\textstyle\bigcap\limits_{i=1}^{2}}
\omega_{i}\neq\emptyset$ in such a manner that the events $\omega_{1}$ and
$\omega_{2}$ have overlapping outcomes $%
{\textstyle\bigcap\nolimits_{i=1}^{2}}
\mathbf{x}\omega_{i}\neq\emptyset$ in the event that $\mathbf{x\sim}$
$p\left(  \mathbf{x}|\omega_{1}\right)  \equiv\mathbf{x\sim}$ $p\left(
\mathbf{x}|\omega_{2}\right)  $.

Figure $1$ also illustrates how Bayes' decision rule $\frac{p\left(
\mathbf{x}|\omega_{1}\right)  }{p\left(  \mathbf{x}|\omega_{2}\right)
}\overset{\omega_{1}}{\underset{\omega_{2}}{\gtrless}}\frac{P\left(
\omega_{2}\right)  }{P\left(  \omega_{1}\right)  }$ is an ill-posed rule of
binary classification, wherein identical random vectors $\mathbf{x\sim}$
$p\left(  \mathbf{x}|\omega_{1}\right)  \equiv\mathbf{x\sim}$ $p\left(
\mathbf{x}|\omega_{2}\right)  $ generated or caused by distinct probability
density functions $p\left(  \mathbf{x}|\omega_{1}\right)  $ and $p\left(
\mathbf{x}|\omega_{2}\right)  $ account for the same effect exhibited by a
binary classification system that is subject to two categories $\omega_{1}$
and $\omega_{2}$ of random vectors $\mathbf{x\in}$ $%
\mathbb{R}
^{d}$.%
\begin{figure}[h]%
\centering
\includegraphics[
height=2.5374in,
width=5.5988in
]%
{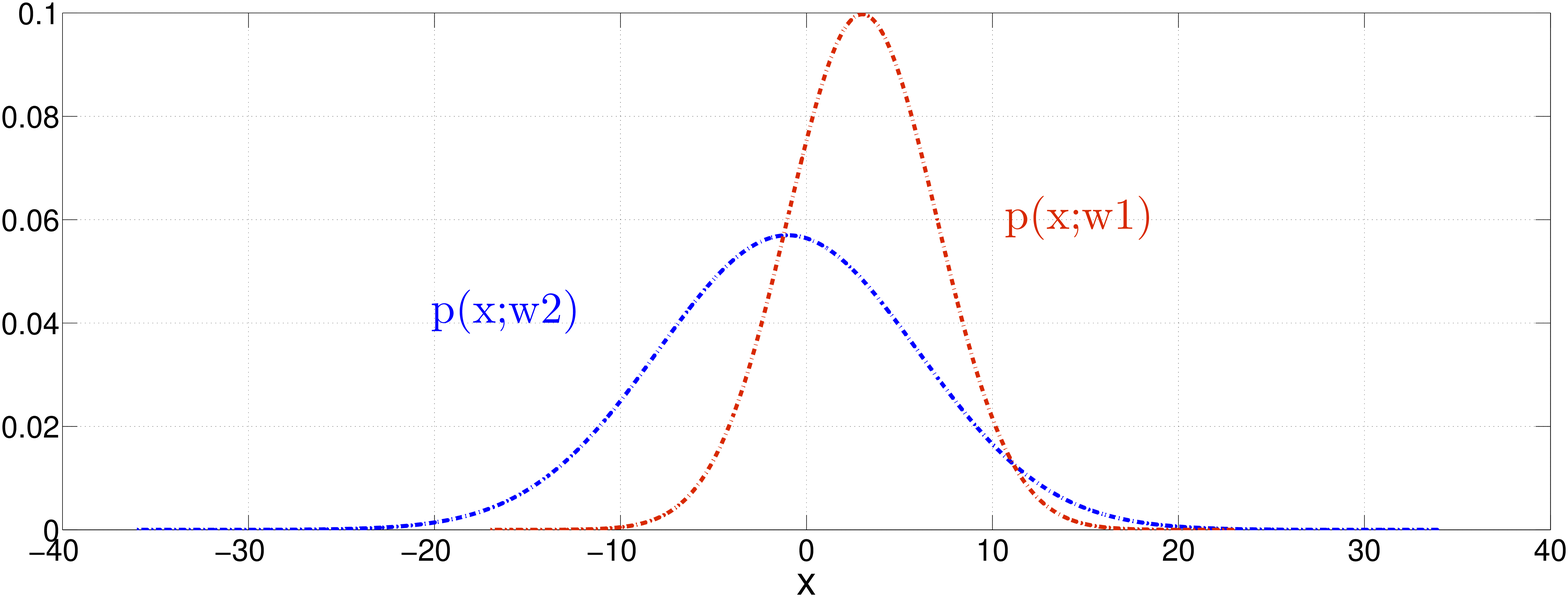}%
\caption{Take any given overlapping distributions of random points
$\mathbf{x\in}$ $\mathbb{R} ^{d}$, such that the events $\omega_{1}$ and
$\omega_{2}$ in Bayes' decision rule $\frac{p\left(  \mathbf{x}|\omega
_{1}\right)  }{p\left(  \mathbf{x}|\omega_{2}\right)  }\protect\overset{\omega
_{1}}{\protect\underset{\omega_{2}}{\gtrless}}\frac{P\left(  \omega
_{2}\right)  }{P\left(  \omega_{1}\right)  }$ are both causes of identical
events $\mathbf{x\sim}$ $p\left(  \mathbf{x}|\omega_{1}\right)  \equiv
\mathbf{x\sim}$ $p\left(  \mathbf{x}|\omega_{2}\right)  $ that occur. It
follows that sets of the events $\omega_{1}$ and $\omega_{2}$ are overlapping
with each other ${\textstyle\protect\bigcap\limits_{i=1}^{2}}\omega_{i}%
\neq\emptyset$ , at which point the events $\omega_{1}$ and $\omega_{2}$ have
overlapping outcomes ${\textstyle\protect\bigcap\nolimits_{i=1}^{2}%
}\mathbf{x}\omega_{i}\neq\emptyset$ whenever $\mathbf{x\sim}$ $p\left(
\mathbf{x}|\omega_{1}\right)  \equiv\mathbf{x\sim}$ $p\left(  \mathbf{x}%
|\omega_{2}\right)  $.}%
\end{figure}

\subsection{An Ill-posed Rule of Binary Classification}

By Theorem \ref{Bayes' Decision Rule Theorem}, we have demonstrated that
Bayes' decision rule constitutes an \emph{ill-posed} rule for the direct
problem of the binary classification of random vectors $\mathbf{x\in}$ $%
\mathbb{R}
^{d}$.

Thereby, we conclude that the direct problem of the binary classification of
random vectors $\mathbf{x\in}$ $%
\mathbb{R}
^{d}$ is essentially an \emph{ill-posed} \emph{problem}, such that identical
random observations $\mathbf{x\sim}$ $p\left(  \mathbf{x}|\omega_{1}\right)
\equiv\mathbf{x\sim}$ $p\left(  \mathbf{x}|\omega_{2}\right)  $ generated or
caused by distinct probability density functions $p\left(  \mathbf{x}%
|\omega_{1}\right)  $ and $p\left(  \mathbf{x}|\omega_{2}\right)  $ account
for the same effect exhibited by a binary classification system.

\subsection{Generalization of Results to Multiclass Systems}

Corollary \ref{Bayes' Multiclass Corollary} generalizes the results presented
in Theorem \ref{Bayes' Decision Rule Theorem} to multiclass classification systems.

\begin{corollary}
\label{Bayes' Multiclass Corollary}Take any given $M$-class classification
system, subject to $M$ categories of random vectors $\mathbf{x\in}$ $%
\mathbb{R}
^{d}$ generated by $M$ classes of probability density functions $\mathbf{x\sim
}$ $p\left(  \mathbf{x}|\omega_{i}\right)  :i=1,\ldots,M$, that is determined
by $M$ ensembles of $M-1$ Bayes' decision rules $\frac{p\left(  \mathbf{x}%
|\omega_{i}\right)  }{p\left(  \mathbf{x}|\omega_{j}\right)  }\overset{\omega
_{i}}{\underset{\omega_{j}}{\gtrless}}\frac{P\left(  \omega_{j}\right)
}{P\left(  \omega_{i}\right)  }$, such that Bayes' decision rule has the form%
\[
\Lambda\left(  \mathbf{x}\right)  \triangleq\sum\limits_{i=1}^{M}%
\sum\limits_{j=1}^{M-1}\frac{p\left(  \mathbf{x}|\omega_{i}\right)  }{p\left(
\mathbf{x}|\omega_{j}\right)  }\overset{\omega_{i}}{\underset{\omega
_{j}}{\gtrless}}\frac{P\left(  \omega_{j}\right)  }{P\left(  \omega
_{i}\right)  }\text{,}%
\]
wherein one class is compared with all of the other $M-1$ classes in each
ensemble of $M-1$ binary classifiers.

The $M$ ensembles of the $M-1$ Bayes' decision rules $\sum\limits_{i=1}%
^{M}\sum\limits_{j=1}^{M-1}\frac{p\left(  \mathbf{x}|\omega_{i}\right)
}{p\left(  \mathbf{x}|\omega_{j}\right)  }\overset{\omega_{i}%
}{\underset{\omega_{j}}{\gtrless}}\frac{P\left(  \omega_{j}\right)  }{P\left(
\omega_{i}\right)  }$ do not satisfy the law of total probability since none
of the $M-1$ Bayes' decision rules $\frac{p\left(  \mathbf{x}|\omega
_{i}\right)  }{p\left(  \mathbf{x}|\omega_{j}\right)  }\overset{\omega
_{i}}{\underset{\omega_{j}}{\gtrless}}\frac{P\left(  \omega_{j}\right)
}{P\left(  \omega_{i}\right)  }$ in each and every one of the $M$ ensembles
satisfies the law of total probability.

Thereby, Bayes' decision rule for $M$-class classification systems does not
satisfy the law of total probability.

Therefore, Bayes' decision rule for $M$-class classification systems does not
satisfy the conditions of Bayes' theorem, as well as the conditions of the
modification of Bayes' theorem.
\end{corollary}

\begin{proof}
Corollary \ref{Bayes' Multiclass Corollary} is proved by using conditions
expressed by Theorem \ref{Bayes' Decision Rule Theorem}, along with the
superposition principle
\citep{Lathi1998}%
.
\end{proof}

\subsection{Warning of Ill-suited and Irrelevant Parameters}

Theorem \ref{Bayes' Decision Rule Theorem} and Corollary
\ref{Bayes' Multiclass Corollary} demonstrate that Bayes' decision rule is
ill-defined. Corollary \ref{Bayes' Decision Rule Corollary} provides us with a
warning that Bayes' decision rule for binary classification systems is
determined by ill-suited and irrelevant parameters.

\begin{corollary}
\label{Bayes' Decision Rule Corollary}Take any given binary classification
system, subject to two categories $\omega_{1}$ and $\omega_{2}$ of random
vectors $\mathbf{x\in}$ $%
\mathbb{R}
^{d}$, that is determined by Bayes' decision rule $\frac{p\left(
\mathbf{x}|\omega_{1}\right)  }{p\left(  \mathbf{x}|\omega_{2}\right)
}\overset{\omega_{1}}{\underset{\omega_{2}}{\gtrless}}\frac{P\left(
\omega_{2}\right)  \left(  C_{12}-C_{22}\right)  }{P\left(  \omega_{1}\right)
\left(  C_{21}-C_{11}\right)  }$, where $p\left(  \mathbf{x}|\omega
_{1}\right)  $ and $p\left(  \mathbf{x}|\omega_{2}\right)  $ are
class-conditional probability density functions of the two categories
$\omega_{1}$ and $\omega_{2}$ of random vectors $\mathbf{x\in}$ $%
\mathbb{R}
^{d}$, $\omega_{1}$ or $\omega_{2}$ is the true category, $P\left(  \omega
_{1}\right)  $ and $P\left(  \omega_{2}\right)  $ are prior probabilities of
class $\omega_{1}$ and class $\omega_{2}$, and the scalars $C_{12}$, $C_{22}$,
$C_{21}$, and $C_{11}$ denote costs for right and wrong decisions.

The decision threshold $\frac{P\left(  \omega_{2}\right)  \left(
C_{12}-C_{22}\right)  }{P\left(  \omega_{1}\right)  \left(  C_{21}%
-C_{11}\right)  }$ of the binary classification system $\frac{p\left(
\mathbf{x}|\omega_{1}\right)  }{p\left(  \mathbf{x}|\omega_{2}\right)
}\overset{\omega_{1}}{\underset{\omega_{2}}{\gtrless}}\frac{P\left(
\omega_{2}\right)  \left(  C_{12}-C_{22}\right)  }{P\left(  \omega_{1}\right)
\left(  C_{21}-C_{11}\right)  }$ is ill-defined since the prior probabilities
$P\left(  \omega_{1}\right)  $ and $P\left(  \omega_{2}\right)  $ and the
scalars $C_{12}$, $C_{22}$, $C_{21}$, and $C_{11}$ are ill-suited and irrelevant.
\end{corollary}

\begin{proof}
By Theorem \ref{Bayes' Decision Rule Theorem}, it follows that the prior
probabilities $P\left(  \omega_{1}\right)  $ and $P\left(  \omega_{2}\right)
$ in Bayes' decision rule $\frac{p\left(  \mathbf{x}|\omega_{1}\right)
}{p\left(  \mathbf{x}|\omega_{2}\right)  }\overset{\omega_{1}%
}{\underset{\omega_{2}}{\gtrless}}\frac{P\left(  \omega_{2}\right)  }{P\left(
\omega_{1}\right)  }$ are ill-suited and irrelevant.

Moreover, there is no statistical basis to choose numerical values for four
scalars $C_{12}$, $C_{22}$, $C_{21}$, and $C_{11}$ that determine costs
associated with all of the right and wrong decisions for any given binary
classification system $\frac{p\left(  \mathbf{x}|\omega_{1}\right)  }{p\left(
\mathbf{x}|\omega_{2}\right)  }\overset{\omega_{1}}{\underset{\omega
_{2}}{\gtrless}}\eta$, such that $\eta=\frac{P\left(  \omega_{2}\right)
\left(  C_{12}-C_{22}\right)  }{P\left(  \omega_{1}\right)  \left(
C_{21}-C_{11}\right)  }$.

It follows that the four scalars $C_{12}$, $C_{22}$, $C_{21}$, and $C_{11}$
that appear in the decision threshold $\frac{P\left(  \omega_{2}\right)
\left(  C_{12}-C_{22}\right)  }{P\left(  \omega_{1}\right)  \left(
C_{21}-C_{11}\right)  }$ of Bayes' decision rule $\frac{p\left(
\mathbf{x}|\omega_{1}\right)  }{p\left(  \mathbf{x}|\omega_{2}\right)
}\overset{\omega_{1}}{\underset{\omega_{2}}{\gtrless}}\frac{P\left(
\omega_{2}\right)  \left(  C_{12}-C_{22}\right)  }{P\left(  \omega_{1}\right)
\left(  C_{21}-C_{11}\right)  }$ are ill-suited and irrelevant.

Therefore, it is concluded that the decision threshold $\eta$ of any given
binary classification system $\frac{p\left(  \mathbf{x}|\omega_{1}\right)
}{p\left(  \mathbf{x}|\omega_{2}\right)  }\overset{\omega_{1}%
}{\underset{\omega_{2}}{\gtrless}}\eta$ that has the form $\eta=\frac{P\left(
\omega_{2}\right)  \left(  C_{12}-C_{22}\right)  }{P\left(  \omega_{1}\right)
\left(  C_{21}-C_{11}\right)  }$ is ill-defined.
\end{proof}

Corollary \ref{Bayes' Decision Rule Corollary} is readily generalized to
multiclass classification systems. Corollary
\ref{Bayes' Decision Rule Multiclass Corollary} provides us with a warning
that Bayes' decision rule for multiclass classification systems is determined
by ill-suited and irrelevant parameters.

\begin{corollary}
\label{Bayes' Decision Rule Multiclass Corollary}Take any given multiclass
classification system, subject to $M$ categories of random vectors
$\mathbf{x\in}$ $%
\mathbb{R}
^{d}$ generated by $M$ classes of probability density functions $\mathbf{x\sim
}$ $p\left(  \mathbf{x}|\omega_{i}\right)  $, that is determined by Bayes'
decision rule%
\[
\Lambda\left(  \mathbf{x}\right)  \triangleq\sum\limits_{i=1}^{M}%
\sum\limits_{j=1}^{M-1}\frac{p\left(  \mathbf{x}|\omega_{i}\right)  }{p\left(
\mathbf{x}|\omega_{j}\right)  }\overset{\omega_{i}}{\underset{\omega
_{j}}{\gtrless}}\frac{P\left(  \omega_{j}\right)  \left(  C_{ij}%
-C_{jj}\right)  }{P\left(  \omega_{i}\right)  \left(  C_{ji}-C_{ii}\right)
}\text{,}%
\]
such that one class is compared with all of the other $M-1$ classes in each
ensemble of $M-1$ binary classifiers.

The decision threshold $\frac{P\left(  \omega_{j}\right)  \left(
C_{ij}-C_{jj}\right)  }{P\left(  \omega_{i}\right)  \left(  C_{ji}%
-C_{ii}\right)  }$ of any given binary classification system $\frac{p\left(
\mathbf{x}|\omega_{i}\right)  }{p\left(  \mathbf{x}|\omega_{j}\right)
}\overset{\omega_{1}}{\underset{\omega_{2}}{\gtrless}}\frac{P\left(
\omega_{j}\right)  \left(  C_{ij}-C_{jj}\right)  }{P\left(  \omega_{i}\right)
\left(  C_{ji}-C_{ii}\right)  }$ in any given ensemble%
\[
\sum\limits_{j=1}^{M-1}\frac{p\left(  \mathbf{x}|\omega_{i}\right)  }{p\left(
\mathbf{x}|\omega_{j}\right)  }\overset{\omega_{i}}{\underset{\omega
_{j}}{\gtrless}}\frac{P\left(  \omega_{j}\right)  \left(  C_{ij}%
-C_{jj}\right)  }{P\left(  \omega_{i}\right)  \left(  C_{ji}-C_{ii}\right)  }%
\]
of $M-1$ Bayes' decision rules is ill-defined since the prior probabilities
$P\left(  \omega_{i}\right)  $ and $P\left(  \omega_{j}\right)  $ and the
scalars $C_{ij}$, $C_{jj}$, $C_{ji}$, and $C_{ii}$ are ill-suited and irrelevant.
\end{corollary}

\begin{proof}
Corollary \ref{Bayes' Decision Rule Multiclass Corollary} is proved by using
conditions expressed by Corollary \ref{Bayes' Decision Rule Corollary}, along
with the superposition principle
\citep{Lathi1998}%
.
\end{proof}

\subsection{Warning of Inconsistent and Arbitrary Measures}

Theorem \ref{Bayes' Decision Rule Theorem} and Corollaries
\ref{Bayes' Multiclass Corollary} -
\ref{Bayes' Decision Rule Multiclass Corollary} demonstrate that Bayes' risk
\ and Bayes' error are ill-defined measures of the expected risk and the
probability of classification error. Corollary \ref{Bayes' Risk Corollary}
provides us with a warning that Bayes' risk \ and Bayes' error are
inconsistent and arbitrary measures of the expected risk and the probability
of classification error exhibited by Bayes' decision rule for binary
classification systems, subject to two categories of random vectors.

\begin{corollary}
\label{Bayes' Risk Corollary}Bayes' risk and Bayes' error are inconsistent and
arbitrary measures of the expected risk and the probability of classification
error exhibited by Bayes' decision rule for any given binary classification
system $\frac{p\left(  \mathbf{x}|\omega_{1}\right)  }{p\left(  \mathbf{x}%
|\omega_{2}\right)  }\overset{\omega_{1}}{\underset{\omega_{2}}{\gtrless}%
}\frac{P\left(  \omega_{2}\right)  \left(  C_{12}-C_{22}\right)  }{P\left(
\omega_{1}\right)  \left(  C_{21}-C_{11}\right)  }$, subject to two categories
$\omega_{1}$ and $\omega_{2}$ of random vectors $\mathbf{x\in}$ $%
\mathbb{R}
^{d}$ generated by two classes of probability density functions $\mathbf{x\sim
}$ $p\left(  \mathbf{x}|\omega_{1}\right)  $ and $\mathbf{x\sim}$ $p\left(
\mathbf{x}|\omega_{2}\right)  $, such that the integral%
\begin{align*}
\mathfrak{R}_{\mathfrak{B}}\left(  \Lambda\left(  \mathbf{x}\right)  \right)
&  \triangleq P\left(  \omega_{1}\right)  \left(  C_{21}-C_{11}\right)
\int_{-\infty}^{\eta}p\left(  \mathbf{x}|\omega_{1}\right)  d\mathbf{x}\\
&  \mathbf{+}P\left(  \omega_{2}\right)  \left(  C_{12}-C_{22}\right)
\int_{\eta}^{\infty}p\left(  \mathbf{x}|\omega_{2}\right)  d\mathbf{x}\text{,}%
\end{align*}
where $\eta=\frac{P\left(  \omega_{2}\right)  \left(  C_{12}-C_{22}\right)
}{P\left(  \omega_{1}\right)  \left(  C_{21}-C_{11}\right)  }$, is
inconsistent and arbitrary, at which point the prior probabilities $P\left(
\omega_{1}\right)  $ and $P\left(  \omega_{2}\right)  $ and the scalars
$C_{12}$, $C_{22}$, $C_{21}$, and $C_{11}$ are ill-suited and irrelevant, and
the decision threshold $\frac{P\left(  \omega_{2}\right)  \left(
C_{12}-C_{22}\right)  }{P\left(  \omega_{1}\right)  \left(  C_{21}%
-C_{11}\right)  }$ is ill-defined, such that $\eta\neq\frac{P\left(
\omega_{2}\right)  \left(  C_{12}-C_{22}\right)  }{P\left(  \omega_{1}\right)
\left(  C_{21}-C_{11}\right)  }$.
\end{corollary}

\begin{proof}
Corollary \ref{Bayes' Risk Corollary} is proved by generalizing conditions
expressed by Theorem \ref{Bayes' Decision Rule Theorem} and Corollary
\ref{Bayes' Decision Rule Corollary}.
\end{proof}

Corollary \ref{Bayes' Risk Corollary} is readily generalized to multiclass
classification systems. Corollary \ref{Bayes' Risk Multiclass Corollary}
provides us with a warning that Bayes' risk and Bayes' error are inconsistent
and arbitrary measures of the expected risk and the probability of
classification error exhibited by Bayes' decision rule for multiclass
classification systems, subject to $M$ categories of random vectors.

\begin{corollary}
\label{Bayes' Risk Multiclass Corollary}Bayes' risk and Bayes' error are
inconsistent and arbitrary measures of the expected risk and the probability
of classification error exhibited by Bayes' decision rule for any given
multiclass classification system, subject to $M$ categories of random vectors
$\mathbf{x\in}$ $%
\mathbb{R}
^{d}$ generated by $M$ classes of probability density functions $\mathbf{x\sim
}$ $p\left(  \mathbf{x}|\omega_{i}\right)  $, wherein Bayes' decision rule has
the form%
\[
\Lambda\left(  \mathbf{x}\right)  \triangleq\sum\limits_{i=1}^{M}%
\sum\limits_{j=1}^{M-1}\frac{p\left(  \mathbf{x}|\omega_{i}\right)  }{p\left(
\mathbf{x}|\omega_{j}\right)  }\overset{\omega_{i}}{\underset{\omega
_{j}}{\gtrless}}\frac{P\left(  \omega_{j}\right)  \left(  C_{ij}%
-C_{jj}\right)  }{P\left(  \omega_{i}\right)  \left(  C_{ji}-C_{ii}\right)
}\text{,}%
\]
such that any given integral that contributes to Bayes' risk and Bayes' error%
\begin{align*}
&  P\left(  \omega_{i}\right)  \left(  C_{ji}-C_{ii}\right)  \int_{-\infty
}^{\eta}p\left(  \mathbf{x}|\omega_{i}\right)  d\mathbf{x}\\
&  \mathbf{+}P\left(  \omega_{j}\right)  \left(  C_{ij}-C_{jj}\right)
\int_{\eta}^{\infty}p\left(  \mathbf{x}|\omega_{j}\right)  d\mathbf{x}\text{,}%
\end{align*}
where $\eta=\frac{P\left(  \omega_{j}\right)  \left(  C_{ij}-C_{jj}\right)
}{P\left(  \omega_{i}\right)  \left(  C_{ji}-C_{ii}\right)  }$, is
inconsistent and arbitrary, at which point the prior probabilities $P\left(
\omega_{i}\right)  $ and $P\left(  \omega_{j}\right)  $ and the scalars
$C_{ij}$, $C_{jj}$, $C_{ji}$, and $C_{ii}$ are ill-suited and irrelevant, and
the decision threshold $\frac{P\left(  \omega_{j}\right)  \left(
C_{ij}-C_{jj}\right)  }{P\left(  \omega_{i}\right)  \left(  C_{ji}%
-C_{ii}\right)  }$ is ill-defined, such that $\eta\neq\frac{P\left(
\omega_{j}\right)  \left(  C_{ij}-C_{jj}\right)  }{P\left(  \omega_{i}\right)
\left(  C_{ji}-C_{ii}\right)  }$.
\end{corollary}

\begin{proof}
Corollary \ref{Bayes' Risk Multiclass Corollary} is proved by using conditions
expressed by Corollaries \ref{Bayes' Multiclass Corollary} and
\ref{Bayes' Decision Rule Multiclass Corollary}.
\end{proof}

In conclusion, by Theorem \ref{Bayes' Decision Rule Theorem} and Corollaries
\ref{Bayes' Multiclass Corollary} -
\ref{Bayes' Decision Rule Multiclass Corollary}, we have demonstrated that
Bayes' decision rule is ill-defined for any given binary or multiclass
classification system that is subject to random vectors.

Furthermore, by Corollaries \ref{Bayes' Risk Corollary} -
\ref{Bayes' Risk Multiclass Corollary}, we have demonstrated that Bayes' risk
and Bayes' error are ill-defined measures of the expected risk and the
probability of classification error exhibited by Bayes' decision rule for any
given binary or multiclass classification system that is subject to random vectors.

\subsection{Formulation of a Well-posed Direct Problem}

We recognize that the inconsistencies expressed by Theorem
\ref{Bayes' Decision Rule Theorem} and Corollaries
\ref{Bayes' Multiclass Corollary} - \ref{Bayes' Risk Multiclass Corollary} are
representative of an ill-posed problem, such that identical random
observations $\mathbf{x\sim}$ $p\left(  \mathbf{x}|\omega_{1}\right)
\equiv\mathbf{x\sim}$ $p\left(  \mathbf{x}|\omega_{2}\right)  $ generated by
distinct probability density functions $p\left(  \mathbf{x}|\omega_{1}\right)
$ and $p\left(  \mathbf{x}|\omega_{2}\right)  $ account for the same effect
exhibited by a binary classification system.

Thereby, we conclude that the direct problem of the binary classification of
random vectors is an ill-posed problem. We also conclude that Bayes' decision
rule constitutes an ill-posed rule for the direct problem of the binary
classification of random vectors. Accordingly, we are motivated to derive a
minimum risk decision rule---\emph{from first principles}---for the
fundamental problem of the binary classification of random vectors.

In the next section, we begin deriving a minimum risk decision rule from first principles.

\section{\label{Section 3}First Principles of Minimum Risk Classifiers}

We now develop first principles---that provide the basis of a minimum risk
decision rule---for the fundamental problem of the binary classification of
random vectors. We express these first principles by Definition $3.1$, Axioms
\ref{Likelihood Value Axiom} - \ref{Minimum Risk Decision Rule Axiom}, Theorem
\ref{Basis of Locus Formula Theorem} and Corollaries
\ref{Form of Decision Boundary Corollary} -
\ref{Secondary Integral Equation Corollary}.

We begin by considering the idea of a likelihood ratio test---which is the
fundamental basis of a decision rule
\citep
{melsa1978decision,Poor1996,Srinath1996,stark1994probability,VanTrees1968}%
.

\subsection{Likelihood Ratio Tests}

Bayes' decision rule $\Lambda\left(  \mathbf{x}\right)  \triangleq
\frac{p\left(  \mathbf{x}|\omega_{1}\right)  }{p\left(  \mathbf{x}|\omega
_{2}\right)  }\overset{\omega_{1}}{\underset{\omega_{2}}{\gtrless}}%
\frac{P\left(  \omega_{2}\right)  \left(  C_{12}-C_{22}\right)  }{P\left(
\omega_{1}\right)  \left(  C_{21}-C_{11}\right)  }$ is considered a likelihood
ratio test, such that the likelihood ratio%
\[
\Lambda\left(  \mathbf{x}\right)  \triangleq\frac{p\left(  \mathbf{x}%
|\omega_{1}\right)  }{p\left(  \mathbf{x}|\omega_{2}\right)  }%
\]
and the decision threshold%
\[
\eta\triangleq\frac{P\left(  \omega_{2}\right)  \left(  C_{12}-C_{22}\right)
}{P\left(  \omega_{1}\right)  \left(  C_{21}-C_{11}\right)  }%
\]
determine a decision rule%
\[
\Lambda\left(  \mathbf{x}\right)  \overset{\omega_{1}}{\underset{\omega
_{2}}{\gtrless}}\eta
\]
that divides an observation space $Z=Z_{1}\cup Z_{2}$ into two regions $Z_{1}$
and $Z_{2}$---by assigning each point $\mathbf{x}$ in region $Z_{1}$ to
hypothesis $\omega_{1}$ and each point $\mathbf{x}$ in region $Z_{2}$ to
hypothesis $\omega_{2}$
\citep
{hippenstiel2017detection,melsa1978decision,Poor1996,Srinath1996,stark1994probability,VanTrees1968}%
.

\subsection{Fragmented Components of Likelihood Ratio Tests}

Take any given likelihood ratio test that has the form of Bayes' decision rule
$\frac{p\left(  \mathbf{x}|\omega_{1}\right)  }{p\left(  \mathbf{x}|\omega
_{2}\right)  }\overset{\omega_{1}}{\underset{\omega_{2}}{\gtrless}}%
\frac{P\left(  \omega_{2}\right)  \left(  C_{12}-C_{22}\right)  }{P\left(
\omega_{1}\right)  \left(  C_{21}-C_{11}\right)  }$. We realize that the
likelihood ratio $\Lambda\left(  \mathbf{x}\right)  =\frac{p\left(
\mathbf{x}|\omega_{1}\right)  }{p\left(  \mathbf{x}|\omega_{2}\right)  }%
$---which determines likelihood values and likely locations of random
observations $\mathbf{x\sim}$ $p\left(  \mathbf{x}|\omega_{1}\right)  $ and
$\mathbf{x\sim}$ $p\left(  \mathbf{x}|\omega_{2}\right)  $ within the decision
space $Z=Z_{1}\cup Z_{2}$ of Bayes' decision system $\frac{p\left(
\mathbf{x}|\omega_{1}\right)  }{p\left(  \mathbf{x}|\omega_{2}\right)
}\overset{\omega_{1}}{\underset{\omega_{2}}{\gtrless}}\frac{P\left(
\omega_{2}\right)  \left(  C_{12}-C_{22}\right)  }{P\left(  \omega_{1}\right)
\left(  C_{21}-C_{11}\right)  }$---is \emph{not connected} to the boundary
$\eta\triangleq\frac{P\left(  \omega_{2}\right)  \left(  C_{12}-C_{22}\right)
}{P\left(  \omega_{1}\right)  \left(  C_{21}-C_{11}\right)  }$ of Bayes'
decision system $\frac{p\left(  \mathbf{x}|\omega_{1}\right)  }{p\left(
\mathbf{x}|\omega_{2}\right)  }\overset{\omega_{1}}{\underset{\omega
_{2}}{\gtrless}}\eta$.

We recognize that the likelihood ratio is \emph{not connected} to the boundary
of Bayes' decision system \emph{because} the processing involved in computing
the likelihood ratio $\frac{p\left(  \mathbf{x}|\omega_{1}\right)  }{p\left(
\mathbf{x}|\omega_{2}\right)  }$ is \emph{not related} to computing the cost
assignments $C_{12}$, $C_{22}$, $C_{21}$, $C_{11}$ and the prior probabilities
$P\left(  \omega_{1}\right)  $ and $P\left(  \omega_{2}\right)  $ that appear
in the decision threshold $\frac{P\left(  \omega_{2}\right)  \left(
C_{12}-C_{22}\right)  }{P\left(  \omega_{1}\right)  \left(  C_{21}%
-C_{11}\right)  }$ of Bayes' decision system $\frac{p\left(  \mathbf{x}%
|\omega_{1}\right)  }{p\left(  \mathbf{x}|\omega_{2}\right)  }\overset{\omega
_{1}}{\underset{\omega_{2}}{\gtrless}}\frac{P\left(  \omega_{2}\right)
\left(  C_{12}-C_{22}\right)  }{P\left(  \omega_{1}\right)  \left(
C_{21}-C_{11}\right)  }$
\citep{Srinath1996,VanTrees1968}%
.

Indeed, the decision threshold $\eta=\frac{P\left(  \omega_{2}\right)  \left(
C_{12}-C_{22}\right)  }{P\left(  \omega_{1}\right)  \left(  C_{21}%
-C_{11}\right)  }$ of Bayes' decision system $\frac{p\left(  \mathbf{x}%
|\omega_{1}\right)  }{p\left(  \mathbf{x}|\omega_{2}\right)  }\overset{\omega
_{1}}{\underset{\omega_{2}}{\gtrless}}\frac{P\left(  \omega_{2}\right)
\left(  C_{12}-C_{22}\right)  }{P\left(  \omega_{1}\right)  \left(
C_{21}-C_{11}\right)  }$ is considered a \emph{variable quantity} that
accommodates changes in cost assignments and prior probabilities---at which
point \emph{values} for cost assignments and prior probabilities are
frequently \emph{\textquotedblleft educated guesses\textquotedblright}\
\citep{Srinath1996,VanTrees1968}%
.

Since prior probabilities and cost assignments are difficult to determine, the
decision threshold $\eta=\frac{P\left(  \omega_{2}\right)  \left(
C_{12}-C_{22}\right)  }{P\left(  \omega_{1}\right)  \left(  C_{21}%
-C_{11}\right)  }$ can simply be assigned the value of $1$, wherein
$C_{11}=C_{22}=0$, $C_{21}=C_{12}=1$, and $P\left(  \omega_{1}\right)
=P\left(  \omega_{2}\right)  =0.5$.

Now, take any given criterion---i.e., the Bayes' criterion, the Neyman-Pearson
criterion, the minimum probability of error criterion, the min-max criterion,
or the maximum likelihood criterion---that is used to determine a value for
the threshold $\eta$ of a decision rule $\Lambda\left(  \mathbf{x}\right)
\overset{\omega_{1}}{\underset{\omega_{2}}{\gtrless}}\eta$.

For all of the criterion listed above, we realize that the likelihood ratio
$\Lambda\left(  \mathbf{x}\right)  =\frac{p\left(  \mathbf{x}|\omega
_{1}\right)  }{p\left(  \mathbf{x}|\omega_{2}\right)  }$ of the decision
system $\Lambda\left(  \mathbf{x}\right)  \overset{\omega_{1}%
}{\underset{\omega_{2}}{\gtrless}}\eta$ is \emph{not connected} to the
boundary $\eta$ of the decision system $\Lambda\left(  \mathbf{x}\right)
\overset{\omega_{1}}{\underset{\omega_{2}}{\gtrless}}\eta$ because the
processing involved in computing the \emph{likelihood ratio} $\frac{p\left(
\mathbf{x}|\omega_{1}\right)  }{p\left(  \mathbf{x}|\omega_{2}\right)  }$ is
\emph{not related} to computing the variables that appear in the
\emph{decision} \emph{threshold} $\eta$ of the decision rule $\Lambda\left(
\mathbf{x}\right)  \overset{\omega_{1}}{\underset{\omega_{2}}{\gtrless}}\eta$.

\subsection{The Maximum Likelihood Criterion}

Suppose that we let the decision threshold $\eta$ of a decision system
$\frac{p\left(  \mathbf{x}|\omega_{1}\right)  }{p\left(  \mathbf{x}|\omega
_{2}\right)  }\overset{\omega_{1}}{\underset{\omega_{2}}{\gtrless}}\eta$ be
assigned the value of $1$. If the decision threshold $\eta\triangleq1$, then
the likelihood ratio test%
\[
\frac{p\left(  \mathbf{x}|\omega_{1}\right)  }{p\left(  \mathbf{x}|\omega
_{2}\right)  }\overset{\omega_{1}}{\underset{\omega_{2}}{\gtrless}}1
\]
is based on the \emph{maximum likelihood criterion}. The basic idea behind the
maximum likelihood criterion is to select the class $\omega_{1}$ or
$\omega_{2}$ that a random vector $\mathbf{x}$ likely belongs to
\citep{hippenstiel2017detection,melsa1978decision}%
. Thereby, if we know the probability density functions $p\left(
\mathbf{x}|\omega_{1}\right)  $ and $p\left(  \mathbf{x}|\omega_{2}\right)  $
for two classes $\omega_{1}$ and $\omega_{2}$ of random vectors $\mathbf{x\in
}$ $%
\mathbb{R}
^{d}$, then, given a particular observation $\mathbf{x}_{0}$, we can compute
the likelihood values $p\left(  \mathbf{x}_{0}|\omega_{1}\right)  $ and
$p\left(  \mathbf{x}_{0}|\omega_{2}\right)  $, and select the most likely
cause of the observation.

The maximum likelihood criterion is considered to be the \emph{simplest} of
all of the techniques that have been used to determine decision rules, which
include $\left(  1\right)  $ the Bayes' criterion---where a decision threshold
$\eta\triangleq\frac{P\left(  \omega_{2}\right)  \left(  C_{12}-C_{22}\right)
}{P\left(  \omega_{1}\right)  \left(  C_{21}-C_{11}\right)  }$ is selected to
minimize the cost and risk; $\left(  2\right)  $ the Neyman-Pearson
criterion---where a decision threshold $\lambda$ is selected to maximize the
probability of detection (the power) for a given level of significance;
$\left(  3\right)  $ the minimum probability of error criterion---where
decision regions $Z_{1}$ and $Z_{2}$ are selected to minimize the total
probability of error; $\left(  4\right)  $ the maximum a posteriori (MAP)
criterion---which is identical to the minimum probability of error criterion;
$\left(  5\right)  $ the min-max criterion---which is based on a version of
Bayes' decision rule, where the average cost is maximum for certain decisions;
and $\left(  6\right)  $ the maximum likelihood criterion, where the decision
threshold $\eta\triangleq1$
\citep{hippenstiel2017detection,melsa1978decision}%
.

Now---consider whether a decision rule $\Lambda\left(  \mathbf{x}\right)
\overset{\omega_{1}}{\underset{\omega_{2}}{\gtrless}}\eta$ is determined by
the Bayes' criterion, the Neyman-Pearson criterion, the minimum probability of
error criterion, the MAP criterion, the min-max criterion, or the maximum
likelihood criterion.

We realize that any given decision rule $\Lambda\left(  \mathbf{x}\right)
\overset{\omega_{1}}{\underset{\omega_{2}}{\gtrless}}\eta$ is determined by
the \emph{same} likelihood ratio $\Lambda\left(  \mathbf{x}\right)
\triangleq\frac{p\left(  \mathbf{x}|\omega_{1}\right)  }{p\left(
\mathbf{x}|\omega_{2}\right)  }$ since each criterion simply \emph{determines}
the \emph{value} of the decision threshold $\eta$ of the decision rule
$\frac{p\left(  \mathbf{x}|\omega_{1}\right)  }{p\left(  \mathbf{x}|\omega
_{2}\right)  }\overset{\omega_{1}}{\underset{\omega_{2}}{\gtrless}}\eta$.

Moreover, for any given criterion, we realize that the value of the decision
threshold $\eta$ of any given decision rule $\frac{p\left(  \mathbf{x}%
|\omega_{1}\right)  }{p\left(  \mathbf{x}|\omega_{2}\right)  }\overset{\omega
_{1}}{\underset{\omega_{2}}{\gtrless}}\eta$ is largely determined by \emph{ad
hoc methods}. Indeed, we are unaware of any established principle that can be
used to determine the decision threshold $\eta$ of a decision system
$\frac{p\left(  \mathbf{x}|\omega_{1}\right)  }{p\left(  \mathbf{x}|\omega
_{2}\right)  }\overset{\omega_{1}}{\underset{\omega_{2}}{\gtrless}}\eta$.

Equally important, we realize that the criterion for any given
technique---that is used to determine the decision threshold $\eta$ of a
decision rule $\Lambda\left(  \mathbf{x}\right)  \overset{\omega
_{1}}{\underset{\omega_{2}}{\gtrless}}\eta$---cannot be used to determine the
likelihood ratio $\Lambda\left(  \mathbf{x}\right)  \triangleq\frac{p\left(
\mathbf{x}|\omega_{1}\right)  }{p\left(  \mathbf{x}|\omega_{2}\right)  }$ of a
decision system $\frac{p\left(  \mathbf{x}|\omega_{1}\right)  }{p\left(
\mathbf{x}|\omega_{2}\right)  }\overset{\omega_{1}}{\underset{\omega
_{2}}{\gtrless}}\eta$. Indeed, the likelihood ratio is usually \emph{unknown}
and must be \emph{estimated} in some manner
\citep
{Duda2001,Fukunaga1990,hippenstiel2017detection,melsa1978decision,Poor1996,Srinath1996,stark1994probability,VanTrees1968}%
.

We regard the\emph{ }likelihood ratio to be an essential component of any
given decision rule $\Lambda\left(  \mathbf{x}\right)  \overset{\omega
_{1}}{\underset{\omega_{2}}{\gtrless}}\eta$ since the likelihood ratio is
determined by \emph{probability laws} that govern how random observations
$\mathbf{x\sim}$ $p\left(  \mathbf{x}|\omega_{1}\right)  $ and $\mathbf{x\sim
}$ $p\left(  \mathbf{x}|\omega_{2}\right)  $ are distributed within certain
regions $\mathcal{R}_{1}$ and $\mathcal{R}_{2}$ of Euclidean space $%
\mathbb{R}
^{d}$, such that $\mathcal{R}_{1}\cap\mathcal{R}_{2}=\emptyset$ or
$\mathcal{R}_{1}\cap\mathcal{R}_{2}\neq\emptyset$.

\subsection{Probability Laws of Distributions}

Each probability density function $p\left(  \mathbf{x}|\omega_{1}\right)  $
and $p\left(  \mathbf{x}|\omega_{2}\right)  $ in the likelihood ratio
$\frac{p\left(  \mathbf{x}|\omega_{1}\right)  }{p\left(  \mathbf{x}|\omega
_{2}\right)  }$ of any given decision system $\frac{p\left(  \mathbf{x}%
|\omega_{1}\right)  }{p\left(  \mathbf{x}|\omega_{2}\right)  }\overset{\omega
_{1}}{\underset{\omega_{2}}{\gtrless}}\eta$ constitutes a \textquotedblleft
probability law\textquotedblright\ for a respective class $\omega_{1}$ and
$\omega_{2}$ of random vectors $\mathbf{x\in}$ $%
\mathbb{R}
^{d}$
\citep{parzen1962stochastic,Parzen1960}%
.

Thereby, we realize that any given probability density function $p\left(
\mathbf{x}|\omega_{1}\right)  $ or $p\left(  \mathbf{x}|\omega_{2}\right)  $
represents a certain probability law that determines how random observations
$\mathbf{x\sim}$ $p\left(  \mathbf{x}|\omega_{1}\right)  $ or $\mathbf{x\sim}$
$p\left(  \mathbf{x}|\omega_{2}\right)  $ generated by the respective
probability density function $p\left(  \mathbf{x}|\omega_{1}\right)  $ or
$p\left(  \mathbf{x}|\omega_{2}\right)  $ are \emph{distributed} within
\emph{certain regions} $\mathcal{R}_{1}$ and $\mathcal{R}_{2}$ of Euclidean
space $%
\mathbb{R}
^{d}$, such that the regions $\mathcal{R}_{1}$ and $\mathcal{R}_{2}$ are
either \emph{overlapping} with each other in some manner $\mathcal{R}_{1}%
\cap\mathcal{R}_{2}\neq\emptyset$, or the regions $\mathcal{R}_{1}$ and
$\mathcal{R}_{2}$ are \emph{disjoint} $\mathcal{R}_{1}\cap\mathcal{R}%
_{2}=\emptyset$.

We also recognize that the \emph{error rate} of any given decision system
$\frac{p\left(  \mathbf{x}|\omega_{1}\right)  }{p\left(  \mathbf{x}|\omega
_{2}\right)  }\overset{\omega_{1}}{\underset{\omega_{2}}{\gtrless}}\eta$ is a
\emph{function} of both likelihood values \emph{and} likely locations of
random observations $\mathbf{x\sim}$ $p\left(  \mathbf{x}|\omega_{1}\right)  $
and $\mathbf{x\sim}$ $p\left(  \mathbf{x}|\omega_{2}\right)  $ within the
decision space $Z=Z_{1}\cup Z_{2}$ of the system $\frac{p\left(
\mathbf{x}|\omega_{1}\right)  }{p\left(  \mathbf{x}|\omega_{2}\right)
}\overset{\omega_{1}}{\underset{\omega_{2}}{\gtrless}}\eta$.

\subsection{Regulating the Locus of a Decision Boundary}

Since the error rate of any given decision system $\frac{p\left(
\mathbf{x}|\omega_{1}\right)  }{p\left(  \mathbf{x}|\omega_{2}\right)
}\overset{\omega_{1}}{\underset{\omega_{2}}{\gtrless}}\eta$ is a
\emph{function} of both likelihood values \emph{and} likely locations of
random observations $\mathbf{x\sim}$ $p\left(  \mathbf{x}|\omega_{1}\right)  $
and $\mathbf{x\sim}$ $p\left(  \mathbf{x}|\omega_{2}\right)  $ within the
decision space $Z=Z_{1}\cup Z_{2}$ of the system $\frac{p\left(
\mathbf{x}|\omega_{1}\right)  }{p\left(  \mathbf{x}|\omega_{2}\right)
}\overset{\omega_{1}}{\underset{\omega_{2}}{\gtrless}}\eta$, we realize that
the decision threshold $\eta$ of any given decision system $\frac{p\left(
\mathbf{x}|\omega_{1}\right)  }{p\left(  \mathbf{x}|\omega_{2}\right)
}\overset{\omega_{1}}{\underset{\omega_{2}}{\gtrless}}\eta$ must somehow
\emph{account for} likelihood values and likely locations of the random
observations $\mathbf{x\sim}$ $p\left(  \mathbf{x}|\omega_{1}\right)  $ and
$\mathbf{x\sim}$ $p\left(  \mathbf{x}|\omega_{2}\right)  $ generated by the
respective probability density functions $p\left(  \mathbf{x}|\omega
_{1}\right)  $ and $p\left(  \mathbf{x}|\omega_{2}\right)  $ of the system.

Indeed, Corollary \ref{Form of Decision Boundary Corollary} demonstrates that
the \emph{locus} of the boundary $\eta$ of any given decision system
$\frac{p\left(  \mathbf{x}|\omega_{1}\right)  }{p\left(  \mathbf{x}|\omega
_{2}\right)  }\overset{\omega_{1}}{\underset{\omega_{2}}{\gtrless}}\eta$ is
\emph{regulated} by its relationship with the likelihood ratio $\Lambda\left(
\mathbf{x}\right)  \triangleq\frac{p\left(  \mathbf{x}|\omega_{1}\right)
}{p\left(  \mathbf{x}|\omega_{2}\right)  }$ of the decision system
$\frac{p\left(  \mathbf{x}|\omega_{1}\right)  }{p\left(  \mathbf{x}|\omega
_{2}\right)  }\overset{\omega_{1}}{\underset{\omega_{2}}{\gtrless}}\eta$.

We now turn our attention to essential criteria of a minimum risk decision system.

\subsection{Essential Criteria of a Minimum Risk Decision System}

The simplicity of the maximum likelihood criterion is considered its
\emph{weakness}---since the maximum likelihood criterion is too simple to
adequately represent realistic problems
\citep{melsa1978decision}%
.

We agree that the maximum likelihood criterion is too simple---and thereby is
an insufficient criterion to determine a minimum risk decision rule. However,
we also consider the Bayes' criterion, the Neyman-Pearson criterion, the
minimum probability of error criterion, the MAP criterion, and the min-max
criterion to be \emph{insufficient} \emph{criterion} that are \emph{too
simple}---since none of the above-mentioned criterion embody \emph{essential
criteria} that can be \emph{used} to \emph{determine} the overall statistical
structure and behavior and properties of the \emph{likelihood ratio} of a
minimum risk binary classification system that exhibits the minimum
probability of classification error.

We use Occam's razor to motivate \emph{starting from} the simplest decision
rule $\frac{p\left(  \mathbf{x}|\omega_{1}\right)  }{p\left(  \mathbf{x}%
|\omega_{2}\right)  }\overset{\omega_{1}}{\underset{\omega_{2}}{\gtrless}}%
1$---which provides the \emph{simplest possible explanation}---of a minimum
risk decision system $\frac{p\left(  \mathbf{x}|\omega_{1}\right)  }{p\left(
\mathbf{x}|\omega_{2}\right)  }\overset{\omega_{1}}{\underset{\omega
_{2}}{\gtrless}}\eta$ that is subject to random vectors $\mathbf{x\in}$ $%
\mathbb{R}
^{d}$, such that $\mathbf{x\sim}$ $p\left(  \mathbf{x}|\omega_{1}\right)  $
and $\mathbf{x\sim}$ $p\left(  \mathbf{x}|\omega_{2}\right)  $ are generated
by the respective probability density functions $p\left(  \mathbf{x}%
|\omega_{1}\right)  $ and $p\left(  \mathbf{x}|\omega_{2}\right)  $ of the system.

Then, given the \emph{principle of parsimony}---that is expressed by Occam's
razor---we enlarge the complexity of the decision rule $\frac{p\left(
\mathbf{x}|\omega_{1}\right)  }{p\left(  \mathbf{x}|\omega_{2}\right)
}\overset{\omega_{1}}{\underset{\omega_{2}}{\gtrless}}1$ in such a manner that
the \emph{increased capacity} of the minimum risk decision system
$\frac{p\left(  \mathbf{x}|\omega_{1}\right)  }{p\left(  \mathbf{x}|\omega
_{2}\right)  }\overset{\omega_{1}}{\underset{\omega_{2}}{\gtrless}}1$
\emph{reveals} fundamental \emph{laws} of binary classification---that are
exhibited by discriminant functions of minimum risk binary classification
systems $\frac{p\left(  \mathbf{x}|\omega_{1}\right)  }{p\left(
\mathbf{x}|\omega_{2}\right)  }\overset{\omega_{1}}{\underset{\omega
_{2}}{\gtrless}}1$ that are subject to random vectors $\mathbf{x\in}$ $%
\mathbb{R}
^{d}$, such that random vectors $\mathbf{x\sim}$ $p\left(  \mathbf{x}%
|\omega_{1}\right)  $ and $\mathbf{x\sim}$ $p\left(  \mathbf{x}|\omega
_{2}\right)  $ are generated by the respective probability density functions
$p\left(  \mathbf{x}|\omega_{1}\right)  $ and $p\left(  \mathbf{x}|\omega
_{2}\right)  $ of the system.

\subsection{Application of Occam's Razor}

Occam's razor, also known as the principle of parsimony or the law of
parsimony, is a problem-solving principle which states that \textquotedblleft
entities should not be multiplied beyond necessity\textquotedblright\
\citep{braithwaite2007occam,Duda2001,mitchell1997machine}%
. According to Occam's razor, we should not make unnecessary assumptions.
Correspondingly, we should remove any aspect of a theory that cannot be
objectively observed or measured or whose case cannot be argued on logical grounds.

\subsection{The Essence of Occam's Razor}%

\citet{braithwaite2007occam}
has sized up the essence of Occam's razor: \textquotedblleft The essence of
the point, in its proper context, is to start from the simplest possible
explanation and make it more complex only if, and when, absolutely
necessary.\textquotedblright

In like manner, Isaac Newton
\citep{newton1999principia}
stated that: \textquotedblleft We are to admit no more causes of natural
things, than such as are both true and sufficient to explain their
appearances.\textquotedblright\ Thus, according to Newton's first rule of
philosophical reasoning, theories and hypotheses should be as simple as they
can be while still accounting for the observed facts.

\subsection{Irrelevant Entities of Minimum Risk Decision Systems}

Given the law of parsimony and Newton's first rule of philosophical reasoning,
we regard cost assignments, prior probabilities, and significance levels
associated with detection probabilities and false alarm probabilities---each
of which has been used to determine the \emph{value} of the threshold $\eta$
of a minimum risk decision system $\Lambda\left(  \mathbf{x}\right)
\overset{\omega_{1}}{\underset{\omega_{2}}{\gtrless}}\eta$---to be
\emph{irrelevant entities} of a minimum risk decision system that is subject
to random vectors $\mathbf{x\in}$ $%
\mathbb{R}
^{d}$.

\subsection{Regulating the Structure of a Decision System}

In this section of our treatise, we demonstrate that the \emph{structure} of
the \emph{locus} of the boundary $\eta$ of any given minimum risk decision
system $\Lambda\left(  \mathbf{x}\right)  \overset{\omega_{1}%
}{\underset{\omega_{2}}{\gtrless}}\eta$---that is subject to random vectors
$\mathbf{x\in}$ $%
\mathbb{R}
^{d}$---is determined by its \emph{relationship} with the \emph{likelihood
ratio} $\Lambda\left(  \mathbf{x}\right)  $ of the system $\Lambda\left(
\mathbf{x}\right)  \overset{\omega_{1}}{\underset{\omega_{2}}{\gtrless}}\eta$.
Correspondingly, in Sections \ref{Section 7} and \ref{Section 8}, we
demonstrate that the \emph{structure} of the \emph{locus} of the likelihood
ratio $\Lambda\left(  \mathbf{x}\right)  $ of any given minimum risk decision
system $\Lambda\left(  \mathbf{x}\right)  \overset{\omega_{1}%
}{\underset{\omega_{2}}{\gtrless}}\eta$ is determined by its relationship with
the locus of the boundary $\eta$ of the system.

We reconsider the simple decision rule $\frac{p\left(  \mathbf{x}|\omega
_{1}\right)  }{p\left(  \mathbf{x}|\omega_{2}\right)  }\overset{\omega
_{1}}{\underset{\omega_{2}}{\gtrless}}1$ that was initially motivated by the
maximum likelihood criterion. Consequently, we enlarge the complexity of the
decision rule $\frac{p\left(  \mathbf{x}|\omega_{1}\right)  }{p\left(
\mathbf{x}|\omega_{2}\right)  }\overset{\omega_{1}}{\underset{\omega
_{2}}{\gtrless}}1$ in such a manner that the increased capacity of the minimum
risk decision system $\frac{p\left(  \mathbf{x}|\omega_{1}\right)  }{p\left(
\mathbf{x}|\omega_{2}\right)  }\overset{\omega_{1}}{\underset{\omega
_{2}}{\gtrless}}1$ \emph{reveals} fundamental \emph{laws} of binary
classification---that are exhibited by discriminant functions of minimum risk
binary classification systems $\frac{p\left(  \mathbf{x}|\omega_{1}\right)
}{p\left(  \mathbf{x}|\omega_{2}\right)  }\overset{\omega_{1}%
}{\underset{\omega_{2}}{\gtrless}}1$ that are subject to random vectors
$\mathbf{x\in}$ $%
\mathbb{R}
^{d}$.

Thereby, we derive essential criteria---\emph{from first principles}---for the
fundamental problem of the binary classification of random vectors. We use
these first principles to develop fundamental statistical laws that determine
the overall statistical structure and behavior and properties of minimum risk
classification systems that exhibit the minimum probability of classification
error. Consequently, we use these statistical laws to develop a data-driven
theoretical blueprint that provides fundamental statistical laws that
determine the generalization behavior of machine learning algorithms that find
target functions of minimum risk classification systems.

We begin by defining the notion of a minimum risk binary classification
system. Let probability density functions of random vectors $\mathbf{x\in}$ $%
\mathbb{R}
^{d}$ that belong to class $\omega_{1}$ or $\omega_{2}$ be denoted by
$p\left(  \mathbf{x};\omega_{1}\right)  $ or $p\left(  \mathbf{x};\omega
_{2}\right)  $ respectively.

\begin{definition}
Any given\ binary classification system, subject to two categories $\omega
_{1}$ and $\omega_{2}$ of random vectors $\mathbf{x\in}$ $%
\mathbb{R}
^{d}$ such that $\mathbf{x\sim}$ $p\left(  \mathbf{x};\omega_{1}\right)  $ and
$\mathbf{x\sim}$ $p\left(  \mathbf{x};\omega_{2}\right)  $, where
distributions of the random vectors $\mathbf{x}$ are determined by certain
probability density functions $p\left(  \mathbf{x};\omega_{1}\right)  $ and
$p\left(  \mathbf{x};\omega_{2}\right)  $, is said to be a minimum risk binary
classification system if and only if the binary classification system exhibits
the lowest possible error rate for any given random vectors $\mathbf{x}$ such
that $\mathbf{x\sim}$ $p\left(  \mathbf{x};\omega_{1}\right)  $ and
$\mathbf{x\sim}$ $p\left(  \mathbf{x};\omega_{2}\right)  $.
\end{definition}

\subsection{Likelihood Values of Random Vectors}

We recognize that a probability density function $p\left(  \mathbf{x}\right)
$ of a random vector $\mathbf{x\in}$ $%
\mathbb{R}
^{d}$ is essentially a curve or surface that is formed by a distribution of
likelihood values of random vectors $\mathbf{x}$---such that likelihood values
of the random vectors $\mathbf{x}$ are determined by distributions of the
random vectors $\mathbf{x}$\textbf{---}that are conditional on statistical
distributions of random vectors $\mathbf{x}$ determined by the probability
density function $p\left(  \mathbf{x}\right)  $. Accordingly, we realize that
each point on the curve or surface of a probability density function $p\left(
\mathbf{x}\right)  $ represents a likelihood that a corresponding random
vector $\mathbf{x}$ will be observed, such that the point determines a
likelihood value for the random vector $\mathbf{x}$. Axiom
\ref{Likelihood Value Axiom} expresses these conditions.

\begin{axiom}
\label{Likelihood Value Axiom}Any given probability density function of a
random vector is essentially a curve or surface that is formed by a
distribution of likelihood values of random vectors, such that each point on
the curve or surface represents a likelihood value of a corresponding random
vector, wherein each likelihood value on the curve or surface is determined by
a distribution of a random vector that is conditional on statistical
distributions of random vectors determined by the probability density function.

Thereby, every point on the curve or surface of a probability density function
represents a likelihood that a corresponding random vector will be observed,
such that the point determines a likelihood value for the random vector.
\end{axiom}

As a concrete example of Axiom \ref{Likelihood Value Axiom}, take any given
normally distributed random vector $\mathbf{x}$ such that $\mathbf{x\sim}$
$p\left(  \mathbf{x};\boldsymbol{\mu},\mathbf{\Sigma}\right)  $, where the
density function $p\left(  \mathbf{x};\boldsymbol{\mu},\mathbf{\Sigma}\right)
$ for the general normal distribution is represented by the vector algebra
expression%
\[
p\left(  \mathbf{x};\boldsymbol{\mu},\mathbf{\Sigma}\right)  =\mathbf{x}%
^{T}\mathbf{\Sigma}^{-1}\mathbf{x}-2\mathbf{x}^{T}\mathbf{\Sigma}%
^{-1}\boldsymbol{\mu}+\boldsymbol{\mu}^{T}\mathbf{\Sigma}^{-1}\boldsymbol{\mu
}-\ln\left(  \left\vert \mathbf{\Sigma}\right\vert \right)  \text{,}%
\]
where $\mathbf{x}$ is a $d$-component normal random vector, $\boldsymbol{\mu}$
is a $d$-component mean vector, $\mathbf{\Sigma}$ is a $d$-by-$d$ covariance
matrix, and $\mathbf{\Sigma}^{-1}$ and $\left\vert \mathbf{\Sigma}\right\vert
$ denote the inverse and the determinant of the covariance matrix.

By Axiom \ref{Likelihood Value Axiom}, it follows that the likelihood value of
the random vector $\mathbf{x}$ is determined by a distribution of the random
vector $\mathbf{x}$ that is conditional on statistical distributions of normal
random vectors $\mathbf{x\sim}$ $p\left(  \mathbf{x};\boldsymbol{\mu
},\mathbf{\Sigma}\right)  $ determined by the statistical expressions
$\mathbf{\Sigma}^{-1}$ and $-2\mathbf{\Sigma}^{-1}\boldsymbol{\mu}$, such that
the likelihood value of the random vector $\mathbf{x}$ is determined by the
value of the expression $\mathbf{x}^{T}\mathbf{\Sigma}^{-1}\mathbf{x}%
-2\mathbf{x}^{T}\mathbf{\Sigma}^{-1}\boldsymbol{\mu}$, along with the value of
the statistical expression $\boldsymbol{\mu}^{T}\mathbf{\Sigma}^{-1}%
\boldsymbol{\mu}-\ln\left(  \left\vert \mathbf{\Sigma}\right\vert \right)  $.

By Axiom \ref{Likelihood Value Axiom}, we recognize that a minimum risk
decision rule for the binary classification problem is based on conditional
likelihood values of random vectors.

\subsection{Conditional Likelihood Values}

We realize that a conditional likelihood value of a random vector is
determined by an output value of a probability density function, given the
input value of the random vector. Axiom \ref{Minimum Risk Decision Rule Axiom}
expresses how conditional likelihood values of random vectors are the basis of
a minimum risk binary classification rule.

\begin{axiom}
\label{Minimum Risk Decision Rule Axiom}Let $p\left(  \mathbf{x};\omega
_{1}\right)  $ and $p\left(  \mathbf{x};\omega_{2}\right)  $ be any given
probability density functions for two classes $\omega_{1}$ and $\omega_{2}$ of
random vectors $\mathbf{x\in}$ $%
\mathbb{R}
^{d}$ such that $\mathbf{x\sim}$ $p\left(  \mathbf{x};\omega_{1}\right)  $ and
$\mathbf{x\sim}$ $p\left(  \mathbf{x};\omega_{2}\right)  $.

Now take any given random vector $\mathbf{x}$ such that either $\mathbf{x\sim
}$ $p\left(  \mathbf{x};\omega_{1}\right)  $ or $\mathbf{x\sim}$ $p\left(
\mathbf{x};\omega_{2}\right)  $.

Next, let $p\left(  \mathbf{x};\omega_{1}|\mathbf{x}\right)  $ and $p\left(
\mathbf{x};\omega_{2}|\mathbf{x}\right)  $ be output values of $p\left(
\mathbf{x};\omega_{1}\right)  $ and $p\left(  \mathbf{x};\omega_{2}\right)  $,
given the input value of the random vector $\mathbf{x}$.

It follows that the output values $p\left(  \mathbf{x};\omega_{1}%
|\mathbf{x}\right)  $ and $p\left(  \mathbf{x};\omega_{2}|\mathbf{x}\right)  $
of the probability density functions $p\left(  \mathbf{x};\omega_{1}\right)  $
and $p\left(  \mathbf{x};\omega_{2}\right)  $, given the input value of the
random vector $\mathbf{x}$, determine conditional likelihood values $p\left(
\mathbf{x};\omega_{1}|\mathbf{x}\right)  $ and $p\left(  \mathbf{x};\omega
_{2}|\mathbf{x}\right)  $ of the random vector $\mathbf{x}$ that indicate
which class $\omega_{1}$ or $\omega_{2}$ the random vector $\mathbf{x}$ likely
belongs to.

Thereby, if $p\left(  \mathbf{x};\omega_{1}|\mathbf{x}\right)  >p\left(
\mathbf{x};\omega_{2}|\mathbf{x}\right)  $, then $\mathbf{x}$ likely belongs
to class $\omega_{1}$, whereas if $p\left(  \mathbf{x};\omega_{1}%
|\mathbf{x}\right)  <p\left(  \mathbf{x};\omega_{2}|\mathbf{x}\right)  $, then
$\mathbf{x} $ likely belongs to class $\omega_{2}$.
\end{axiom}

\subsection{Conditional Probability Values}

It is important to note that conditional likelihood values of random vectors
are not equivalent to conditional probability values of the random vectors.

\begin{remark}
Take any given random vector $\mathbf{x}$ such that $\mathbf{x\sim}$ $p\left(
\mathbf{x};\omega_{1}\right)  $ or $\mathbf{x\sim}$ $p\left(  \mathbf{x}%
;\omega_{2}\right)  $, where $p\left(  \mathbf{x};\omega_{1}\right)  $ and
$p\left(  \mathbf{x};\omega_{2}\right)  $ are certain probability density
functions for two classes $\omega_{1}$ and $\omega_{2}$ of random vectors
$\mathbf{x\in}$ $%
\mathbb{R}
^{d}$. Conditional likelihood values $p\left(  \mathbf{x};\omega
_{1}|\mathbf{x}\right)  $ and $p\left(  \mathbf{x};\omega_{2}|\mathbf{x}%
\right)  $ of the random vector $\mathbf{x}$ are not equivalent to conditional
probability values $P\left(  \mathbf{x};\omega_{1}|\mathbf{x}\right)  $ and
$P\left(  \mathbf{x};\omega_{2}|\mathbf{x}\right)  $ of the random vector
$\mathbf{x}$.

Rather, the conditional probability of observing any given random vector
$\mathbf{x\sim}$ $p\left(  \mathbf{x};\omega_{i}\right)  $, where either $i=1$
or $i=2$, is determined by an integral $P\left(  \mathbf{x};\omega
_{i}|\mathbf{x}\right)  =\int\nolimits_{\mathbf{x}-\triangle\mathbf{x}%
}^{\mathbf{x}+\triangle\mathbf{x}}p\left(  \mathbf{x};\omega_{i}%
|\mathbf{x}\right)  d\mathbf{x}$ over a region $\mathcal{R}$ of Euclidean
space $%
\mathbb{R}
^{d}$, where the size of the region $\mathcal{R}$ accounts for an expected
value and a variability of the random vector $\mathbf{x}$ that is conditional
on the distributions of the random vectors $\mathbf{x}$ determined by the
probability density function $p\left(  \mathbf{x};\omega_{i}\right)  $.
\end{remark}

\subsection{General Form of Binary Classification Systems}

Theorem \ref{Basis of Locus Formula Theorem} expresses the general form of a
minimum risk binary classification system---that will be seen to provide the
basis of a general locus formula for finding discriminant functions of minimum
risk binary classification systems that are subject to random vectors
$\mathbf{x\in}$ $%
\mathbb{R}
^{d}$.

\begin{theorem}
\label{Basis of Locus Formula Theorem}Let $d\left(  \mathbf{x}\right)
\triangleq\frac{p\left(  \mathbf{x};\omega_{1}\right)  }{p\left(
\mathbf{x};\omega_{2}\right)  }$ be the discriminant function of any given
minimum risk binary classification system that is subject to random inputs
$\mathbf{x\in}$ $%
\mathbb{R}
^{d}$ such that $\mathbf{x\sim}$ $p\left(  \mathbf{x};\omega_{1}\right)  $ and
$\mathbf{x\sim}$ $p\left(  \mathbf{x};\omega_{2}\right)  $, where $p\left(
\mathbf{x};\omega_{1}\right)  $ and $p\left(  \mathbf{x};\omega_{2}\right)  $
are certain probability density functions for two classes $\omega_{1}$ and
$\omega_{2}$ of random vectors $\mathbf{x\in}$ $%
\mathbb{R}
^{d}$.

The discriminant function $d\left(  \mathbf{x}\right)  =\frac{p\left(
\mathbf{x};\omega_{1}\right)  }{p\left(  \mathbf{x};\omega_{2}\right)  }$ of
the minimum risk binary classification system satisfies the inequality
relation%
\begin{align}
&  \frac{p\left(  \mathbf{x};\omega_{1}\right)  }{p\left(  \mathbf{x}%
;\omega_{2}\right)  }\overset{\omega_{1}}{\underset{\omega_{2}}{\gtrless}%
}1\tag{3.1}\label{Minimum Risk Decision Rule}\\
&  =p\left(  \mathbf{x};\omega_{1}\right)  \overset{\omega_{1}%
}{\underset{\omega_{2}}{\gtrless}}p\left(  \mathbf{x};\omega_{2}\right)
\text{,}\nonumber
\end{align}
where $\omega_{1}$ or $\omega_{2}$ is the true category, at which point the
discriminant function is the solution of the equation $p\left(  \mathbf{x}%
;\omega_{1}\right)  =p\left(  \mathbf{x};\omega_{2}\right)  $ at the decision
threshold of the system.
\end{theorem}

\begin{proof}
Take any given random vectors $\mathbf{x\in}$ $%
\mathbb{R}
^{d}$ such that $\mathbf{x\sim}$ $p\left(  \mathbf{x};\omega_{1}\right)  $ and
$\mathbf{x\sim}$ $p\left(  \mathbf{x};\omega_{2}\right)  $, where the
probability density functions $p\left(  \mathbf{x};\omega_{1}\right)  $ and
$p\left(  \mathbf{x};\omega_{2}\right)  $ determine distributions of two
classes $\omega_{1}$ and $\omega_{2}$ of random vectors $\mathbf{x\in}$ $%
\mathbb{R}
^{d}$.

Now take any given random vector $\mathbf{x}$ such that $\mathbf{x\sim}$
$p\left(  \mathbf{x};\omega_{1}\right)  $ or $\mathbf{x\sim}$ $p\left(
\mathbf{x};\omega_{2}\right)  $.

By Axiom \ref{Minimum Risk Decision Rule Axiom}, it follows that the output
values $p\left(  \mathbf{x};\omega_{1}|\mathbf{x}\right)  $ and $p\left(
\mathbf{x};\omega_{2}|\mathbf{x}\right)  $ of $p\left(  \mathbf{x};\omega
_{1}\right)  $ and $p\left(  \mathbf{x};\omega_{2}\right)  $, given the input
value of the random vector $\mathbf{x}$, determine conditional likelihood
values $p\left(  \mathbf{x};\omega_{1}|\mathbf{x}\right)  $ and $p\left(
\mathbf{x};\omega_{2}|\mathbf{x}\right)  $ of the random vector $\mathbf{x}$
that indicate which class the random vector $\mathbf{x}$ likely belongs to.

Therefore, if $p\left(  \mathbf{x};\omega_{1}|\mathbf{x}\right)  >p\left(
\mathbf{x};\omega_{2}|\mathbf{x}\right)  $, then $\mathbf{x}$ likely belongs
to class $\omega_{1}$, whereas if $p\left(  \mathbf{x};\omega_{1}%
|\mathbf{x}\right)  <p\left(  \mathbf{x};\omega_{2}|\mathbf{x}\right)  $, then
$\mathbf{x} $ likely belongs to class $\omega_{2}$.

Thereby, if%
\[
p\left(  \mathbf{x};\omega_{1}|\mathbf{x}\right)  \geq p\left(  \mathbf{x}%
;\omega_{2}|\mathbf{x}\right)  \text{,}%
\]
then $\mathbf{x}$ is assigned to class $\omega_{1}$; otherwise, $\mathbf{x}$
is assigned to class $\omega_{2}$.

Thus, it is concluded that the discriminant function $d\left(  \mathbf{x}%
\right)  \triangleq\frac{p\left(  \mathbf{x};\omega_{1}\right)  }{p\left(
\mathbf{x};\omega_{2}\right)  }$ of any given minimum risk binary
classification system that is subject to random vectors $\mathbf{x\sim}$
$p\left(  \mathbf{x};\omega_{1}\right)  $ and $\mathbf{x\sim}$ $p\left(
\mathbf{x};\omega_{2}\right)  $ satisfies the inequality relation
$\frac{p\left(  \mathbf{x};\omega_{1}\right)  }{p\left(  \mathbf{x};\omega
_{2}\right)  }\overset{\omega_{1}}{\underset{\omega_{2}}{\gtrless}}1$, at
which point the discriminant function is the solution of the equation
$p\left(  \mathbf{x};\omega_{1}\right)  =p\left(  \mathbf{x};\omega
_{2}\right)  $ at the decision threshold of the system.
\end{proof}

Figure $2$ illustrates how a discriminant function $d\left(  \mathbf{x}%
\right)  =\frac{p\left(  \mathbf{x};\omega_{1}\right)  }{p\left(
\mathbf{x};\omega_{2}\right)  }$ of a minimum risk binary classification
system $\frac{p\left(  \mathbf{x};\omega_{1}\right)  }{p\left(  \mathbf{x}%
;\omega_{2}\right)  }\overset{\omega_{1}}{\underset{\omega_{2}}{\gtrless}}1$
operates, such that the conditional likelihood values $p\left(  \mathbf{x}%
;\omega_{1}|\mathbf{x}\right)  $ and $p\left(  \mathbf{x};\omega
_{2}|\mathbf{x}\right)  $ of any given random vector $\mathbf{x\sim}$
$p\left(  \mathbf{x};\omega_{1}\right)  $ or $\mathbf{x\sim}$ $p\left(
\mathbf{x};\omega_{2}\right)  $ being classified are compared with each other
relative to the decision threshold of the system, at which point the
discriminant function is the solution of the equation $p\left(  \mathbf{x}%
;\omega_{1}\right)  =p\left(  \mathbf{x};\omega_{2}\right)  $.%

\begin{figure}[h]%
\centering
\includegraphics[
height=1.8741in,
width=5.5988in
]%
{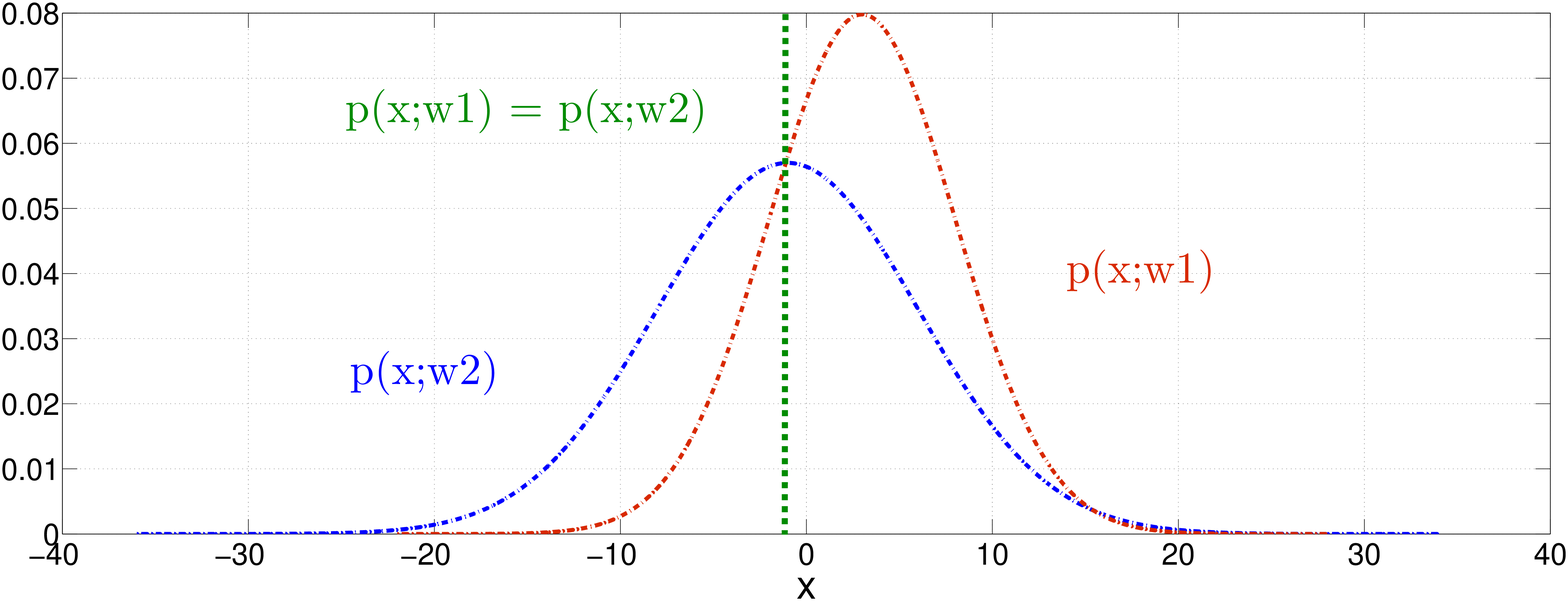}%
\caption{Conditional likelihood values $p\left(  \mathbf{x};\omega
_{1}|\mathbf{x}\right)  $ and $p\left(  \mathbf{x};\omega_{2}|\mathbf{x}%
\right)  $ of any given random vectors $\mathbf{x\sim}$ $p\left(
\mathbf{x};\omega_{1}\right)  $ or $\mathbf{x\sim}$ $p\left(  \mathbf{x}%
;\omega_{2}\right)  $ being classified are compared with each other relative
to the decision threshold of a minimum risk binary classification system
$p\left(  \mathbf{x};\omega_{1}\right)  \protect\overset{\omega_{1}%
}{\protect\underset{\omega_{2}}{\gtrless}}p\left(  \mathbf{x};\omega
_{2}\right)  $, at which point the discriminant function is the solution of
the equation $p\left(  \mathbf{x};\omega_{1}\right)  =p\left(  \mathbf{x}%
;\omega_{2}\right)  $.}%
\end{figure}

\subsection{General Form of Decision Boundaries}

Corollary \ref{Form of Decision Boundary Corollary} expresses the general form
of the decision boundary of any given minimum risk binary classification
system that is subject to random vectors $\mathbf{x\in}$ $%
\mathbb{R}
^{d}$.

\begin{corollary}
\label{Form of Decision Boundary Corollary}Let $\frac{p\left(  \mathbf{x}%
;\omega_{1}\right)  }{p\left(  \mathbf{x};\omega_{2}\right)  }\overset{\omega
_{1}}{\underset{\omega_{2}}{\gtrless}}1$ be any given minimum risk binary
classification system that is subject to random inputs $\mathbf{x\in}$ $%
\mathbb{R}
^{d}$ such that $\mathbf{x\sim}$ $p\left(  \mathbf{x};\omega_{1}\right)  $ and
$\mathbf{x\sim}$ $p\left(  \mathbf{x};\omega_{2}\right)  $, where $p\left(
\mathbf{x};\omega_{1}\right)  $ and $p\left(  \mathbf{x};\omega_{2}\right)  $
are certain probability density functions for two classes $\omega_{1}$ and
$\omega_{2}$ of random vectors $\mathbf{x\in}$ $%
\mathbb{R}
^{d}$.

The decision boundary of the minimum risk binary classification system
$\frac{p\left(  \mathbf{x};\omega_{1}\right)  }{p\left(  \mathbf{x};\omega
_{2}\right)  }\overset{\omega_{1}}{\underset{\omega_{2}}{\gtrless}}1$ is
determined by the equation%
\begin{equation}
\frac{p\left(  \mathbf{x};\omega_{1}\right)  }{p\left(  \mathbf{x};\omega
_{2}\right)  }=1\text{,} \tag{3.2}\label{Equation1 of Decision Boundary}%
\end{equation}
at which point the discriminant function $\frac{p\left(  \mathbf{x};\omega
_{1}\right)  }{p\left(  \mathbf{x};\omega_{2}\right)  }$ of the minimum risk
binary classification system $\frac{p\left(  \mathbf{x};\omega_{1}\right)
}{p\left(  \mathbf{x};\omega_{2}\right)  }\overset{\omega_{1}%
}{\underset{\omega_{2}}{\gtrless}}1$ is the solution of the equation
$\frac{p\left(  \mathbf{x};\omega_{1}\right)  }{p\left(  \mathbf{x};\omega
_{2}\right)  }=1$ that represents the decision boundary of the system.
\end{corollary}

\begin{proof}
Corollary \ref{Form of Decision Boundary Corollary} is proved by generalizing
conditions expressed by Theorem \ref{Basis of Locus Formula Theorem}.
\end{proof}

By Corollary \ref{Form of Decision Boundary Corollary}, we realize that the
structure and the locus of the decision boundary of any given minimum risk
binary classification system is regulated by its relationship with the
discriminant function of the system.

\subsection{Equilibrium Requirement}

Corollary \ref{Equilibrium Requirement Corollary} expresses an equilibrium
requirement for the discriminant function of a minimum risk binary
classification system that is satisfied at the decision boundary of the system.

\begin{corollary}
\label{Equilibrium Requirement Corollary}Let $\frac{p\left(  \mathbf{x}%
;\omega_{1}\right)  }{p\left(  \mathbf{x};\omega_{2}\right)  }\overset{\omega
_{1}}{\underset{\omega_{2}}{\gtrless}}1$ be any given minimum risk binary
classification system that is subject to random inputs $\mathbf{x\in}$ $%
\mathbb{R}
^{d}$ such that $\mathbf{x\sim}$ $p\left(  \mathbf{x};\omega_{1}\right)  $ and
$\mathbf{x\sim}$ $p\left(  \mathbf{x};\omega_{2}\right)  $, where $p\left(
\mathbf{x};\omega_{1}\right)  $ and $p\left(  \mathbf{x};\omega_{2}\right)  $
are certain probability density functions for two classes $\omega_{1}$ and
$\omega_{2}$ of random vectors $\mathbf{x\in}$ $%
\mathbb{R}
^{d}$.

The discriminant function $d\left(  \mathbf{x}\right)  \triangleq
\frac{p\left(  \mathbf{x};\omega_{1}\right)  }{p\left(  \mathbf{x};\omega
_{2}\right)  }$ of the minimum risk binary classification system
$\frac{p\left(  \mathbf{x};\omega_{1}\right)  }{p\left(  \mathbf{x};\omega
_{2}\right)  }\overset{\omega_{1}}{\underset{\omega_{2}}{\gtrless}}1$ is the
solution of the equation%
\begin{equation}
p\left(  \mathbf{x};\omega_{1}\right)  =p\left(  \mathbf{x};\omega_{2}\right)
\tag{3.3}\label{Discriminant Function in Equilibrium}%
\end{equation}
at the decision boundary $\frac{p\left(  \mathbf{x};\omega_{1}\right)
}{p\left(  \mathbf{x};\omega_{2}\right)  }=1$ of the system, so that points on
the curves or surfaces of the probability density functions $p\left(
\mathbf{x};\omega_{1}\right)  $ and $p\left(  \mathbf{x};\omega_{2}\right)  $
that represent likelihood values of corresponding random vectors
$\mathbf{x\sim}$ $p\left(  \mathbf{x};\omega_{1}\right)  $ and $\mathbf{x\sim
}$ $p\left(  \mathbf{x};\omega_{2}\right)  $ are symmetrically balanced with
each other, at which point the discriminant function of the system is in
statistical equilibrium at the decision boundary of the system.
\end{corollary}

\begin{proof}
Corollary \ref{Equilibrium Requirement Corollary} is proved by generalizing
conditions expressed by Axiom \ref{Likelihood Value Axiom}, Theorem
\ref{Basis of Locus Formula Theorem} and Corollary
\ref{Form of Decision Boundary Corollary}.
\end{proof}

\subsection{Regulation of Expected Risk}

Corollary \ref{Primary Integral Equation Corollary} expresses the requirement
that a discriminant function of a minimum risk binary classification system is
the solution of an integral equation over the decision space $Z=Z_{1}\cup
Z_{2}$ of the system, so that the expected risk exhibited by the system is
regulated by the equilibrium requirement in
(\ref{Discriminant Function in Equilibrium})---on the discriminant function at
the decision boundary of the system---expressed by Corollary
\ref{Equilibrium Requirement Corollary}, at which point counter risks\ and
risks exhibited by the system are symmetrically balanced with each other
throughout the decision regions $Z_{1}$ and $Z_{2}$ of the system.

\paragraph{Counter Risks and Risks}

We realize that minimum risk binary classification systems have properties
that we have named \textquotedblleft counter risks\textquotedblright\ and
\textquotedblleft risks,\textquotedblright\ such that right decisions made by
a minimum risk binary classification system are associated with a property of
the system that we have named counter risks, whereas wrong decisions made by a
minimum risk binary classification system are associated with a property of
the system that we have named risks.

\begin{corollary}
\label{Primary Integral Equation Corollary}Let $\frac{p\left(  \mathbf{x}%
;\omega_{1}\right)  }{p\left(  \mathbf{x};\omega_{2}\right)  }\overset{\omega
_{1}}{\underset{\omega_{2}}{\gtrless}}1$ be any given minimum risk binary
classification system that is subject to random inputs $\mathbf{x\in}$ $%
\mathbb{R}
^{d}$ such that $\mathbf{x\sim}$ $p\left(  \mathbf{x};\omega_{1}\right)  $ and
$\mathbf{x\sim}$ $p\left(  \mathbf{x};\omega_{2}\right)  $, where $p\left(
\mathbf{x};\omega_{1}\right)  $ and $p\left(  \mathbf{x};\omega_{2}\right)  $
are certain probability density functions for two classes $\omega_{1}$ and
$\omega_{2}$ of random vectors $\mathbf{x\in}$ $%
\mathbb{R}
^{d}$.

The discriminant function $d\left(  \mathbf{x}\right)  \triangleq
\frac{p\left(  \mathbf{x};\omega_{1}\right)  }{p\left(  \mathbf{x};\omega
_{2}\right)  }$ of the minimum risk binary classification system
$\frac{p\left(  \mathbf{x};\omega_{1}\right)  }{p\left(  \mathbf{x};\omega
_{2}\right)  }\overset{\omega_{1}}{\underset{\omega_{2}}{\gtrless}}1$ is the
solution of the integral equation%
\begin{align}
f_{1}\left(  d\left(  \mathbf{x}\right)  \right)   &  :%
{\displaystyle\int\nolimits_{Z_{1}}}
p\left(  \mathbf{x};\omega_{1}\right)  d\mathbf{x}+%
{\displaystyle\int\nolimits_{Z_{2}}}
p\left(  \mathbf{x};\omega_{1}\right)  d\mathbf{x}+C_{1} \tag{3.4}%
\label{Primary Integral Equation}\\
&  =%
{\displaystyle\int\nolimits_{Z_{1}}}
p\left(  \mathbf{x};\omega_{2}\right)  d\mathbf{x}+%
{\displaystyle\int\nolimits_{Z_{2}}}
p\left(  \mathbf{x};\omega_{2}\right)  d\mathbf{x}+C_{2}\text{,}\nonumber
\end{align}
over the decision space $Z=Z_{1}\cup Z_{2}$ of the system, where $C_{1}$ and
$C_{2}$ are certain integration constants, so that the expected risk exhibited
by the system is regulated by the equilibrium requirement $p\left(
\mathbf{x};\omega_{1}\right)  =p\left(  \mathbf{x};\omega_{2}\right)  $ on the
discriminant function at the decision boundary $\frac{p\left(  \mathbf{x}%
;\omega_{1}\right)  }{p\left(  \mathbf{x};\omega_{2}\right)  }=1$ of the
system, at which point counter risks exhibited by the system---determined by
integrated likelihood values of random points $\mathbf{x\sim}$ $p\left(
\mathbf{x};\omega_{1}\right)  $ located throughout the decision region $Z_{1}%
$, along with risks exhibited by the system---determined by integrated
likelihood values of random points $\mathbf{x\sim}$ $p\left(  \mathbf{x}%
;\omega_{1}\right)  $ located throughout the decision region $Z_{2}$, are
symmetrically balanced with risks exhibited by the system---determined by
integrated likelihood values of random points $\mathbf{x\sim}$ $p\left(
\mathbf{x};\omega_{2}\right)  $ located throughout the decision region $Z_{1}%
$, along with counter risks exhibited by the system---determined by integrated
likelihood values of random points $\mathbf{x\sim}$ $p\left(  \mathbf{x}%
;\omega_{2}\right)  $ located throughout the decision region $Z_{2}$.
\end{corollary}

\begin{proof}
Corollary \ref{Primary Integral Equation Corollary} is proved by generalizing
conditions expressed by Corollary \ref{Equilibrium Requirement Corollary}.
\end{proof}

\subsection{Symmetrical Partitioning of Decision Spaces}

Corollary \ref{Partitioning of Decision Space Corollary} expresses how the
decision boundary of any given minimum risk binary classification system
symmetrically partitions the decision space of the system.

\begin{corollary}
\label{Partitioning of Decision Space Corollary}Take the decision boundary
$\frac{p\left(  \mathbf{x};\omega_{1}\right)  }{p\left(  \mathbf{x};\omega
_{2}\right)  }=1$ of any given minimum risk binary classification system
$\frac{p\left(  \mathbf{x};\omega_{1}\right)  }{p\left(  \mathbf{x};\omega
_{2}\right)  }\overset{\omega_{1}}{\underset{\omega_{2}}{\gtrless}}1$ that is
subject to random inputs $\mathbf{x\in}$ $%
\mathbb{R}
^{d}$ such that $\mathbf{x\sim}$ $p\left(  \mathbf{x};\omega_{1}\right)  $ and
$\mathbf{x\sim}$ $p\left(  \mathbf{x};\omega_{2}\right)  $, where $p\left(
\mathbf{x};\omega_{1}\right)  $ and $p\left(  \mathbf{x};\omega_{2}\right)  $
are certain probability density functions for two classes $\omega_{1}$ and
$\omega_{2}$ of random vectors $\mathbf{x\in}$ $%
\mathbb{R}
^{d}$.

The decision boundary $\frac{p\left(  \mathbf{x};\omega_{1}\right)  }{p\left(
\mathbf{x};\omega_{2}\right)  }=1$ divides the decision space $Z=Z_{1}\cup
Z_{2}$ of the minimum risk binary classification system $\frac{p\left(
\mathbf{x};\omega_{1}\right)  }{p\left(  \mathbf{x};\omega_{2}\right)
}\overset{\omega_{1}}{\underset{\omega_{2}}{\gtrless}}1$ into symmetrical
decision regions $Z_{1}$ and $Z_{2}$ that cover the decision space
$Z=Z_{1}\cup Z_{2}$ in a symmetrically balanced manner, such that if $p\left(
\mathbf{x};\omega_{1}\right)  $ and $p\left(  \mathbf{x};\omega_{2}\right)  $
determine overlapping distributions of random points $\mathbf{x\in}$ $%
\mathbb{R}
^{d}$, wherein $\bigcap\nolimits_{i=1}^{2}\omega_{i}\neq\emptyset$ and
$\bigcap\nolimits_{i=1}^{2}\mathbf{x}\omega_{i}\neq\emptyset$, then locations
of random points $\mathbf{x\sim}$ $p\left(  \mathbf{x};\omega_{1}\right)  $
and $\mathbf{x\sim}$ $p\left(  \mathbf{x};\omega_{2}\right)  $ are
symmetrically distributed throughout both decision regions $Z_{1}$ and $Z_{2}$
of the decision space $Z=Z_{1}\cup Z_{2}$, whereas if $p\left(  \mathbf{x}%
;\omega_{1}\right)  $ and $p\left(  \mathbf{x};\omega_{2}\right)  $ determine
non-overlapping distributions of random points $\mathbf{x\in}$ $%
\mathbb{R}
^{d}$, wherein $\bigcap\nolimits_{i=1}^{2}\omega_{i}=\emptyset$ and
$\bigcap\nolimits_{i=1}^{2}\mathbf{x}\omega_{i}=\emptyset$, then locations of
random points $\mathbf{x\sim}$ $p\left(  \mathbf{x};\omega_{1}\right)  $
within the decision region $Z_{1}$ along with locations of random points
$\mathbf{x\sim}$ $p\left(  \mathbf{x};\omega_{2}\right)  $ within the decision
region $Z_{2}$ are positioned at symmetrical distances from the decision
boundary $\frac{p\left(  \mathbf{x};\omega_{1}\right)  }{p\left(
\mathbf{x};\omega_{2}\right)  }=1$.

Thereby, the expected risk of the minimum risk binary classification system
$\frac{p\left(  \mathbf{x};\omega_{1}\right)  }{p\left(  \mathbf{x};\omega
_{2}\right)  }\overset{\omega_{1}}{\underset{\omega_{2}}{\gtrless}}1$ is
minimized within the decision space $Z=Z_{1}\cup Z_{2}$ of the system in such
a manner that likely locations of random points---associated with right and
wrong decisions made by the system---are symmetrically balanced with each
other throughout both decision regions $Z_{1}$ and $Z_{2}$ of the decision
space $Z=Z_{1}\cup Z_{2}$.
\end{corollary}

\begin{proof}
Corollary \ref{Partitioning of Decision Space Corollary} is proved by
generalizing certain conditions expressed by Corollary
\ref{Primary Integral Equation Corollary}.
\end{proof}

\subsection{State of Statistical Equilibrium}

Corollary \ref{Secondary Integral Equation Corollary} expresses the
requirement that a discriminant function of a minimum risk binary
classification system minimize an integral equation over the decision regions
$Z_{1}$ and $Z_{2}$ of the system, so that the expected risk exhibited by the
system is minimized within the decision space $Z=Z_{1}\cup Z_{2}$ of the
system in such a manner that the system satisfies a state of statistical
equilibrium, at which point counter risks exhibited by the system are
symmetrically balanced with risks exhibited by the system. The integral
equation expressed by Corollary \ref{Secondary Integral Equation Corollary} is
derived from the integral equation in (\ref{Primary Integral Equation})
expressed by Corollary \ref{Primary Integral Equation Corollary}.

\begin{corollary}
\label{Secondary Integral Equation Corollary}Let $\frac{p\left(
\mathbf{x};\omega_{1}\right)  }{p\left(  \mathbf{x};\omega_{2}\right)
}\overset{\omega_{1}}{\underset{\omega_{2}}{\gtrless}}1$ be any given minimum
risk binary classification system that is subject to random inputs
$\mathbf{x\in}$ $%
\mathbb{R}
^{d}$ such that $\mathbf{x\sim}$ $p\left(  \mathbf{x};\omega_{1}\right)  $ and
$\mathbf{x\sim}$ $p\left(  \mathbf{x};\omega_{2}\right)  $, where $p\left(
\mathbf{x};\omega_{1}\right)  $ and $p\left(  \mathbf{x};\omega_{2}\right)  $
are certain probability density functions for two classes $\omega_{1}$ and
$\omega_{2}$ of random vectors $\mathbf{x\in}$ $%
\mathbb{R}
^{d}$.

The discriminant function $d\left(  \mathbf{x}\right)  \triangleq
\frac{p\left(  \mathbf{x};\omega_{1}\right)  }{p\left(  \mathbf{x};\omega
_{2}\right)  }$ of the minimum risk binary classification system
$\frac{p\left(  \mathbf{x};\omega_{1}\right)  }{p\left(  \mathbf{x};\omega
_{2}\right)  }\overset{\omega_{1}}{\underset{\omega_{2}}{\gtrless}}1$
minimizes the integral equation%
\begin{align}
f_{2}\left(  d\left(  \mathbf{x}\right)  \right)   &  :%
{\displaystyle\int\nolimits_{Z_{1}}}
p\left(  \mathbf{x};\omega_{1}\right)  d\mathbf{x-}%
{\displaystyle\int\nolimits_{Z_{1}}}
p\left(  \mathbf{x};\omega_{2}\right)  d\mathbf{x}+C_{1} \tag{3.5}%
\label{Secondary Integral Equation}\\
&  =%
{\displaystyle\int\nolimits_{Z_{2}}}
p\left(  \mathbf{x};\omega_{2}\right)  d\mathbf{x-}%
{\displaystyle\int\nolimits_{Z_{2}}}
p\left(  \mathbf{x};\omega_{1}\right)  d\mathbf{x}+C_{2}\text{,}\nonumber
\end{align}
over the decision regions $Z_{1}$ and $Z_{2}$ of the system, where $C_{1}$ and
$C_{2}$ are certain integration constants, so that the system satisfies a
state of statistical equilibrium such that the expected risk exhibited by the
system is minimized within the decision space $Z=Z_{1}\cup Z_{2}$ of the
system, at which point counter risks and risks exhibited by the
system---located throughout the decision region $Z_{1}$, are symmetrically
balanced with counter risks and risks exhibited by the system---located
throughout the decision region $Z_{2}$.

Thereby, the minimum risk binary classification system exhibits the minimum
probability of classification error for any given random vectors
$\mathbf{x\in}$ $%
\mathbb{R}
^{d}$ such that $\mathbf{x\sim}$ $p\left(  \mathbf{x};\omega_{1}\right)  $ and
$\mathbf{x\sim}$ $p\left(  \mathbf{x};\omega_{2}\right)  $.
\end{corollary}

\begin{proof}
Corollary \ref{Secondary Integral Equation Corollary} is proved by
generalizing conditions expressed by Corollaries
\ref{Equilibrium Requirement Corollary} and
\ref{Primary Integral Equation Corollary}.
\end{proof}

In conclusion, Theorem \ref{Basis of Locus Formula Theorem} and Corollaries
\ref{Form of Decision Boundary Corollary} -
\ref{Secondary Integral Equation Corollary} express fundamental statistical
laws of binary classification---for the fundamental problem of the binary
classification of random vectors. Thereby, we are now in a position to
consider novel geometrical and statistical problems---related to
unconventional algebraic problems---in binary classification.

We now turn our attention to novel geometric locus problems in binary classification.

\section{\label{Section 4}Novel Locus Problems in Binary Classification}

Take the discriminant function of any given minimum risk binary classification
system that satisfies the inequality relation in
(\ref{Minimum Risk Decision Rule}) expressed by Theorem
\ref{Basis of Locus Formula Theorem}. Corollary
\ref{Form of Decision Boundary Corollary} reveals that the discriminant
function of the system is the solution of the equation in
(\ref{Equation1 of Decision Boundary}) that represents the decision boundary
of the system.

This relationship raises an underlying dilemma in binary classification:
\emph{How} is the discriminant function of a minimum risk binary
classification system \emph{connected} to the decision boundary of the system?
We realize that this dilemma is a novel geometrical and statistical
problem---that is related to an unconventional algebraic problem---in binary classification.

\subsection{Prediction of a Critical Statistical Bond}

The conditions expressed by Corollaries
\ref{Equilibrium Requirement Corollary},
\ref{Primary Integral Equation Corollary} and
\ref{Secondary Integral Equation Corollary} reveal \emph{conditions of
statistical equilibrium} that are satisfied by a \emph{discriminant function}
of a minimum risk binary classification system \emph{at} the \emph{decision
boundary} of the system. Indeed, the equilibrium equation in
(\ref{Discriminant Function in Equilibrium}) expressed by Corollary
\ref{Equilibrium Requirement Corollary} and the integral equation in
(\ref{Primary Integral Equation}) expressed by Corollary
\ref{Primary Integral Equation Corollary} both reveal that the discriminant
function of a minimum risk binary classification system is in statistical
equilibrium at the decision boundary of the system, whereas the integral
equation in (\ref{Secondary Integral Equation}) expressed by Corollary
\ref{Secondary Integral Equation Corollary} reveals that the discriminant
function minimizes an integral equation in such a manner that the system
satisfies a state of statistical equilibrium---at which point the expected
risk exhibited by the system is minimized within the decision space of the system.

We realize that Corollaries \ref{Form of Decision Boundary Corollary} -
\ref{Secondary Integral Equation Corollary} predict the existence of a
critical statistical bond---between the discriminant function and the decision
boundary of any given minimum risk binary classification system---which
enables surprising statistical balancing acts that raise fundamental dilemmas
in binary classification.

\subsection{Underlying Dilemmas in Binary Classification}

Corollaries \ref{Equilibrium Requirement Corollary} -
\ref{Secondary Integral Equation Corollary} express surprising statistical
balancing acts---exhibited by discriminant functions and decision boundaries
of minimum risk binary classification systems---that raise the following
dilemmas in binary classification:

\begin{enumerate}
\item \emph{How is the discriminant function of a minimum risk binary
classification system connected to the decision boundary of the system?}

\item \emph{How does the decision boundary account for likelihood values of
the random vectors generated by the respective probability density functions
of the system?}

\item \emph{How does the decision boundary account for likely locations of the
random vectors generated by the respective probability density functions of
the system?}

\item \emph{How is the discriminant function of a minimum risk binary
classification system in statistical equilibrium at the decision boundary of
the system?}

\item \emph{What are the counteracting and opposing forces and influences of a
minimum risk binary classification system?}

\item \emph{How are the forces and influences of a minimum risk binary
classification system related to each other?}

\item \emph{How are the forces and influences of a minimum risk binary
classification system related to the discriminant function of the system?}

\item \emph{How are the forces and influences of a minimum risk binary
classification system related to the decision boundary of the system?}

\item \emph{How are the counteracting and opposing forces and influences of a
minimum risk binary classification system symmetrically balanced with each
other?}

\item \emph{How does a minimum risk binary classification system satisfy a
state of statistical equilibrium?}
\end{enumerate}

We realize that all of the above dilemmas are novel geometrical and
statistical problems---which are related to unconventional algebraic
problems---in binary classification. Moreover, we have determined that each
dilemma is a novel geometric locus problem in binary classification.

In this treatise, we demonstrate that each novel geometric locus problem in
binary classification that is outlined above is fruitfully treated by novel
geometric locus methods in Hilbert spaces---within statistical
frameworks---where the Hilbert spaces are reproducing kernel Hilbert spaces
that have certain reproducing kernels.

By way of motivation, we first consider how classic locus problems are solved.

\subsection{Solving Classic Locus Problems}

The general idea of a curve or surface which at any point of it exhibits some
uniform property is expressed in geometry by the term \textquotedblleft
locus.\textquotedblright\ Generally speaking, a locus is a curve or surface
formed by specific points---each of which possesses some uniform property that
is common to all points on the locus---and no other points
\citep{Eisenhart1939,Nichols1893,Tanner1898}%
.

Classic locus problems are solved by finding algebraic equations of conic
sections or quadratic surfaces, so that the uniform property exhibited by any
given locus of points is identified---relative to an intrinsic coordinate
system---that is an inherent part of an algebraic equation, at which point the
graph of the algebraic equation determines all of the points that lie on the
locus of the algebraic equation.

We can choose whatever coordinate system we prefer to describe a locus of
points---since any given locus of points is independent of the coordinate
system that is used to describe it. Therefore, for any given locus of points,
we can choose the \emph{most natural} coordinate system---for the given
locus---so that each axis of a Cartesian coordinate system is rotated in an
appropriate manner
\citep{Eisenhart1939,Hewson2009,Nichols1893,Tanner1898}%
.

Since we can choose the most suitable coordinate system for any given locus of
points, it follows that conic sections and quadratic surfaces are both subject
to \emph{distinctive} geometric conditions---with respect to and in relation
to---coordinate axes of \emph{various} intrinsic coordinate systems.
Accordingly, since the form of an algebraic equation of a geometric locus is
determined by the form of the intrinsic coordinate system of the locus, it
follows that the positions of the axes of coordinates---to which a given locus
of points is referenced---\emph{are arbitrary}
\citep{Eisenhart1939,Nichols1893,Tanner1898}%
.

Finding the form of an algebraic equation for a locus of points can be a
difficult problem. The primary locus problem involves identifying the uniform
property exhibited by a locus of points relative to the mathematical structure
of an intrinsic coordinate system---that is an inherent part of an algebraic
equation---such that the form of the algebraic equation is determined by the
mathematical structure of the intrinsic coordinate system.

The inverse locus problem involves determining the form of an algebraic
equation---of a locus of points that has been described geometrically---whose
graph determines the coordinates of each and every point on the given locus,
at which point no other points, other than points on the given locus, have
coordinates that satisfy the graph of the algebraic equation. Inverse locus
problems are usually more complex
\citep{Eisenhart1939,Nichols1893,Tanner1898}%
.

Finally, the determination of the form of the algebraic equation of a locus of
points, along with the identification of the uniform property exhibited by all
of the points that lie on the locus, can be greatly simplified by changing the
positions of the coordinate axes to which the locus of points is referenced:

Since changing the positions of the coordinate axes transforms both the form
of the algebraic equation of the locus and the coordinates of all of the
points that lie on the locus
\citep{Eisenhart1939,Nichols1893,Tanner1898}%
.

In order to devise novel locus methods for binary classification---within
statistical frameworks---we need a clear definition of a geometric locus.

\subsection{General Idea of a Geometric Locus}

\begin{definition}
\label{Geometric Locus Definition}A definite curve or surface that contains
specific points is said to be a geometric locus if and only if each point on
the curve or surface possesses a certain uniform property relative to an
intrinsic coordinate system that is an inherent part of an algebraic equation,
so that all of the points that lie on the curve or surface have coordinates
that are solutions of the graph of the algebraic equation, at which point no
other points than those on the curve or surface have coordinates that satisfy
the graph of the algebraic equation.
\end{definition}

Usually, a geometric locus is referred to as a locus of points. In this
treatise, when we consider the geometric aspects of a given locus of points to
be prominent, we refer to the locus as a geometric locus.

\subsection{Equation of a Locus}

Any given locus of points is determined by an equation, such that an equation
of a locus is the location of all of the points---and only those
points---whose coordinates are solutions of the graph of the equation
\citep{Eisenhart1939,Nichols1893,Tanner1898}%
. The following definition expresses the general notion of an equation of a locus.

\begin{definition}
\label{Algebraic Equation of a Locus Definition}An algebraic equation is said
to be an equation of a locus if and only if the equation has an algebraic form
that determines the shape of a curve or surface, along with the conditions
that are satisfied by the coordinates of all of the points that lie on the
curve or surface---relative to an intrinsic coordinate system that is an
inherent part of the algebraic equation---so that the algebraic equation
relates certain uniform geometric conditions satisfied by coordinates of
points that are solutions of the algebraic equation to certain uniform
geometric conditions satisfied by points that lie on the curve or surface of
the algebraic equation, at which point no other points than those that lie on
the curve or surface have coordinates that satisfy the graph of the algebraic equation.
\end{definition}

\subsection{Representation of a Geometric Locus of Points}

It is important to distinguish between the coordinates of the points that
\emph{lie on} the geometric locus of \emph{a certain curve or surface} and the
coordinates of the points that are \emph{solutions} of the \emph{algebraic
equation} of the graph of the curve or surface. Indeed, the points that lie on
the geometric locus of a certain curve or surface are \emph{not} the same
points that are solutions of the locus equation of the curve or surface. The
following definition expresses the idea of the representation of a locus of points.

\begin{definition}
\label{Representation of a Locus of Points Definition}An algebraic equation
represents a locus of points if and only if all of the points that lie on the
locus have coordinates that satisfy the graph of the algebraic equation, where
no other points than those on the locus have coordinates that satisfy the
graph of the algebraic equation, whereas all of the points that are solutions
of the algebraic equation have coordinates that satisfy a certain intrinsic
coordinate system---which is an inherent part of the algebraic equation.
\end{definition}

By Definitions \ref{Geometric Locus Definition},
\ref{Algebraic Equation of a Locus Definition} and
\ref{Representation of a Locus of Points Definition}, we recognize that the
\emph{positions} of the \emph{coordinate axes} of any given intrinsic
coordinate system---to which a certain locus of points is
referenced---\emph{determines} the \emph{coordinates} of all of the
\emph{points} that lie \emph{on} the geometric \emph{locus}, along with the
\emph{form} of the algebraic \emph{equation} of the \emph{locus}. Thereby, we
recognize that the \emph{shape} of any given geometric locus is independent of
the coordinate system that is used to describe it. This essential relationship
is expressed by Axiom \ref{Rotation of Intrinsic Coordinate Axes Axiom}.

\subsection{Changing the Form of a Locus Equation}

Axiom \ref{Rotation of Intrinsic Coordinate Axes Axiom} expresses how the form
of an algebraic equation of a locus can be changed, along with the
\emph{coordinates} of all of the points that lie on the locus.

\begin{axiom}
\label{Rotation of Intrinsic Coordinate Axes Axiom}Let an algebraic equation
be an equation of a locus, so that the form of the algebraic equation is
determined by the mathematical structure of an intrinsic coordinate system of
the locus, at which point the positions of the coordinate axes of the
intrinsic coordinate system determines the coordinates of all of the points
that lie on the locus.

Changing the positions of the coordinate axes of the intrinsic coordinate
system of the locus changes the form of the algebraic equation of the locus,
along with the coordinates of all of the points that lie on the locus.
\end{axiom}

We use Axiom \ref{Rotation of Intrinsic Coordinate Axes Axiom} to motivate how
we devise a general locus formula for the binary classification of random
vectors $\mathbf{x\in}$ $%
\mathbb{R}
^{d}$.

\section{\label{Section 5}Novel Locus Methods in Binary Classification}

Given the minimum risk decision rule in (\ref{Minimum Risk Decision Rule})
expressed by Theorem \ref{Basis of Locus Formula Theorem}, along with
properties of the natural logarithm, it follows that the natural logarithm of
the minimum risk decision rule in (\ref{Minimum Risk Decision Rule})
determines an inequality relation%
\begin{equation}
\ln p\left(  \mathbf{x};\omega_{1}\right)  -\ln p\left(  \mathbf{x};\omega
_{2}\right)  \overset{\omega_{1}}{\underset{\omega_{2}}{\gtrless}}0
\tag{5.1}\label{Transformed Decision Rule}%
\end{equation}
that is satisfied by the discriminant function of any given minimum risk
binary classification system that is subject to random inputs $\mathbf{x\in}$
$%
\mathbb{R}
^{d}$, such that $\mathbf{x\sim}$ $p\left(  \mathbf{x};\omega_{1}\right)  $
and $\mathbf{x\sim}$ $p\left(  \mathbf{x};\omega_{2}\right)  $, where
$p\left(  \mathbf{x};\omega_{1}\right)  $ and $p\left(  \mathbf{x};\omega
_{2}\right)  $ are certain probability density functions for two classes
$\omega_{1}$ and $\omega_{2}$ of random vectors $\mathbf{x\in}$ $%
\mathbb{R}
^{d}$, wherein the discriminant function of the system is represented by the
statistical expression $d\left(  \mathbf{x}\right)  =\ln p\left(
\mathbf{x};\omega_{1}\right)  -\ln p\left(  \mathbf{x};\omega_{2}\right)  $.

Given the general equation of a decision boundary in
(\ref{Equation1 of Decision Boundary}) expressed by Corollary
\ref{Form of Decision Boundary Corollary}, it follows that the decision
boundary of any given minimum risk binary classification system in
(\ref{Transformed Decision Rule}) is determined by the equation%
\begin{equation}
\ln p\left(  \mathbf{x};\omega_{1}\right)  -\ln p\left(  \mathbf{x};\omega
_{2}\right)  =0\text{,} \tag{5.2}\label{Locus of a Decision Boundary}%
\end{equation}
at which point the discriminant function $\ln p\left(  \mathbf{x};\omega
_{1}\right)  -\ln p\left(  \mathbf{x};\omega_{2}\right)  $ of the system is
the solution of the equation $\ln p\left(  \mathbf{x};\omega_{1}\right)  -\ln
p\left(  \mathbf{x};\omega_{2}\right)  =0$ that represents the decision
boundary of the system, such that the graph of
(\ref{Locus of a Decision Boundary}) constitutes a decision boundary that is a
\emph{data-driven} locus of points.

\subsection{Tractable Classification Systems}

We now consider concrete examples of both (\ref{Transformed Decision Rule})
and (\ref{Locus of a Decision Boundary}), such that tractable minimum risk
binary classification systems are subject to multivariate normal data, so that
any given system is determined by a closed-form solution.

Substitution of the statistical expressions for the probability density
functions $p\left(  \mathbf{x};\boldsymbol{\mu}_{1},\mathbf{\Sigma}%
_{1}\right)  $ and $p\left(  \mathbf{x};\boldsymbol{\mu}_{2},\mathbf{\Sigma
}_{2}\right)  $ for the general normal distribution into
(\ref{Transformed Decision Rule}) produces an inequality relation that is
satisfied by the discriminant function of any given minimum risk binary
classification system that is subject to multivariate normal data%
\begin{align}
d\left(  \mathbf{x}\right)   &  \triangleq\mathbf{x}^{T}\mathbf{\Sigma}%
_{1}^{-1}\mathbf{x}-2\mathbf{x}^{T}\mathbf{\Sigma}_{1}^{-1}\boldsymbol{\mu
}_{1}+\boldsymbol{\mu}_{1}^{T}\mathbf{\Sigma}_{1}^{-1}\boldsymbol{\mu}_{1}%
-\ln\left(  \left\vert \mathbf{\Sigma}_{1}\right\vert \right)  \tag{5.3}%
\label{Gaussian Rule}\\
&  -\mathbf{x}^{T}\mathbf{\Sigma}_{2}^{-1}\mathbf{x}+2\mathbf{x}%
^{T}\mathbf{\Sigma}_{2}^{-1}\boldsymbol{\mu}_{2}\mathbf{-}\boldsymbol{\mu}%
_{2}^{T}\mathbf{\Sigma}_{2}^{-1}\boldsymbol{\mu}_{2}+\ln\left(  \left\vert
\mathbf{\Sigma}_{2}\right\vert \right)  \overset{\omega_{1}}{\underset{\omega
_{2}}{\gtrless}}0\text{,}\nonumber
\end{align}
where $\mathbf{x}$ is a $d$-component normal random vector such that
$\mathbf{x\sim}$ $p\left(  \mathbf{x};\boldsymbol{\mu}_{1},\mathbf{\Sigma}%
_{1}\right)  $ or $\mathbf{x\sim}$ $p\left(  \mathbf{x};\boldsymbol{\mu}%
_{2},\mathbf{\Sigma}_{2}\right)  $, $\boldsymbol{\mu}_{1}$ and
$\boldsymbol{\mu}_{2}$ are $d$-component mean vectors, $\mathbf{\Sigma}_{1}$
and $\mathbf{\Sigma}_{2}$ are $d$-by-$d$ covariance matrices, $\mathbf{\Sigma
}^{-1}$ and $\left\vert \mathbf{\Sigma}\right\vert $ denote the inverse and
the determinant of a covariance matrix, and $\omega_{1}$ or $\omega_{2}$ is
the true category
\citep{Duda2001,VanTrees1968}%
.

Set the statistical expression that represents the discriminant function in
(\ref{Gaussian Rule}) equal to zero in accordance with
(\ref{Locus of a Decision Boundary}). It follows that the geometric locus of
the decision boundary of any given minimum risk binary classification system
that is subject to multivariate normal inputs $\mathbf{x\in}$ $%
\mathbb{R}
^{d}$ is represented by a \emph{vector algebra locus equation}%
\begin{align}
d\left(  \mathbf{x}\right)   &  :\mathbf{x}^{T}\mathbf{\Sigma}_{1}%
^{-1}\mathbf{x}-2\mathbf{x}^{T}\mathbf{\Sigma}_{1}^{-1}\boldsymbol{\mu}%
_{1}+\boldsymbol{\mu}_{1}^{T}\mathbf{\Sigma}_{1}^{-1}\boldsymbol{\mu}_{1}%
-\ln\left(  \left\vert \mathbf{\Sigma}_{1}\right\vert \right)  \tag{5.4}%
\label{Norm_Dec_Bound}\\
&  -\mathbf{x}^{T}\mathbf{\Sigma}_{2}^{-1}\mathbf{x}+2\mathbf{x}%
^{T}\mathbf{\Sigma}_{2}^{-1}\boldsymbol{\mu}_{2}\mathbf{-}\boldsymbol{\mu}%
_{2}^{T}\mathbf{\Sigma}_{2}^{-1}\boldsymbol{\mu}_{2}+\ln\left(  \left\vert
\mathbf{\Sigma}_{2}\right\vert \right) \nonumber\\
&  =0\text{,}\nonumber
\end{align}
wherein the graph of the vector algebra locus equation of
(\ref{Norm_Dec_Bound}) constitutes the geometric locus of the decision
boundary of the system, at which point the discriminant function of the system%
\begin{align*}
d\left(  \mathbf{x}\right)   &  =\mathbf{x}^{T}\mathbf{\Sigma}_{1}%
^{-1}\mathbf{x}-2\mathbf{x}^{T}\mathbf{\Sigma}_{1}^{-1}\boldsymbol{\mu}%
_{1}-\mathbf{x}^{T}\mathbf{\Sigma}_{2}^{-1}\mathbf{x}+2\mathbf{x}%
^{T}\mathbf{\Sigma}_{2}^{-1}\boldsymbol{\mu}_{2}\\
&  +\boldsymbol{\mu}_{1}^{T}\mathbf{\Sigma}_{1}^{-1}\boldsymbol{\mu}%
_{1}\mathbf{-}\boldsymbol{\mu}_{2}^{T}\mathbf{\Sigma}_{2}^{-1}\boldsymbol{\mu
}_{2}+\ln\left(  \left\vert \mathbf{\Sigma}_{2}\right\vert \right)
-\ln\left(  \left\vert \mathbf{\Sigma}_{1}\right\vert \right)
\end{align*}
is the solution of the vector algebra locus equation of (\ref{Norm_Dec_Bound}%
), such that normal random vectors $\mathbf{x\sim}$ $p\left(  \mathbf{x}%
;\boldsymbol{\mu}_{1},\mathbf{\Sigma}_{1}\right)  $ and $\mathbf{x\sim}$
$p\left(  \mathbf{x};\boldsymbol{\mu}_{2},\mathbf{\Sigma}_{2}\right)  $ have
coordinates that are solutions of (\ref{Norm_Dec_Bound}), so that the graph of
the vector algebra locus equation (\ref{Norm_Dec_Bound}) determines the
geometric locus of a decision boundary that is either a data-driven quadratic
surface that is a hyperplane, hypersphere, hyperellipsoid, hyperparaboloid or
hyperhyperboloid, or a data-driven conic section that is a line, circle,
ellipse, parabola or hyperbola
\citep{Duda2001,VanTrees1968}%
.

\subsection{The Importance of the Normal Probability Law}

The probability density functions $p\left(  \mathbf{x};\boldsymbol{\mu}%
_{1},\mathbf{\Sigma}_{1}\right)  $ and $p\left(  \mathbf{x};\boldsymbol{\mu
}_{2},\mathbf{\Sigma}_{2}\right)  $ for the general normal distribution in
(\ref{Gaussian Rule}) and (\ref{Norm_Dec_Bound}) represent probability laws
that have played a significant role in probability theory since the early
eighteenth century. This significance derives from random phenomena that obey
a normal probability law $p\left(  \mathbf{x};\mu,\sigma\right)  =$ $\frac
{1}{2}\exp-\left(  x-\mu^{2}\right)  /2\sigma^{2}$, where $\mu$ is the mean
and $\sigma^{2}$ is the variance of the normal probability law. One example of
such a phenomenon is a molecule, with mass $M$, in a gas at absolute temp $T$
that---according to Maxwell's law of velocities---obeys a normal probability
law with parameters $\mu=0$ and $\sigma^{2}=M/kT$, where $\mu$ is the mean and
$\sigma^{2}$ is the variance of the normal probability law, and $k$ is the
physical constant called Boltzmann's constant
\citep{Parzen1960}%
.

Only certain random phenomena obey a normal probability law precisely.
Instead, normal probability laws derive their importance from the fact that
under various conditions, they \emph{closely approximate} other probability
laws. For example normal probability laws have been used to approximate
height, IQ, birth weight and income distribution
\citep{Parzen1960}%
.

Thus, we examine tractable minimum risk binary classification systems that are
subject to normal random vectors $\mathbf{x\in}$ $%
\mathbb{R}
^{d}$, such that $\mathbf{x\sim}$ $p\left(  \mathbf{x};\boldsymbol{\mu}%
_{1},\mathbf{\Sigma}_{1}\right)  $ or $\mathbf{x\sim}$ $p\left(
\mathbf{x};\boldsymbol{\mu}_{2},\mathbf{\Sigma}_{2}\right)  $, that obey
certain normal probability laws precisely.

We now turn our attention to the locus equation in (\ref{Norm_Dec_Bound}).

\subsection{Locus Equation of a Decision Boundary}

We recognize (\ref{Norm_Dec_Bound}) as an equation of a locus that is subject
to distinctive geometrical and statistical conditions for a discriminant
function of a minimum risk binary classification system in statistical
equilibrium---at the decision boundary of the system---relative to an
intrinsic coordinate system, such that the \emph{form} of the vector algebra
locus equation of (\ref{Norm_Dec_Bound}) is \emph{determined} by the
\emph{intrinsic coordinate system} of the geometric locus of the
\emph{decision boundary} of the system, at which point the positions of the
coordinate axes of the intrinsic coordinate system determines the coordinates
of the random points that satisfy the vector algebra locus equation of
(\ref{Norm_Dec_Bound}), along with the coordinates of the points that lie on
the graph of the vector algebra locus equation of (\ref{Norm_Dec_Bound}%
)---namely the geometric locus of the decision boundary of the system.

We now identify the \emph{statistical nexus} of a minimum risk binary
classification system---at which point the discriminant function of the system
is \emph{connected} to the decision boundary of the system.

\subsection{Statistical Nexus of a Binary Classification System}

Take the discriminant function of any given minimum risk binary classification
system that is subject to multivariate normal inputs $\mathbf{x\in}$ $%
\mathbb{R}
^{d}$ such that $\mathbf{x\sim}$ $p\left(  \mathbf{x};\boldsymbol{\mu}%
_{1},\mathbf{\Sigma}_{1}\right)  $ and $\mathbf{x\sim}$ $p\left(
\mathbf{x};\boldsymbol{\mu}_{2},\mathbf{\Sigma}_{2}\right)  $, where $p\left(
\mathbf{x};\boldsymbol{\mu}_{1},\mathbf{\Sigma}_{1}\right)  $ and $p\left(
\mathbf{x};\boldsymbol{\mu}_{2},\mathbf{\Sigma}_{2}\right)  $ are certain
probability density functions.

Since the discriminant function is the solution of the vector algebra locus
equation of (\ref{Norm_Dec_Bound})---that represents the geometric locus of
the decision boundary of the minimum risk binary classification system---by
Definitions \ref{Geometric Locus Definition} and
\ref{Algebraic Equation of a Locus Definition}, we have determined that the
discriminant function of the system is connected to the geometric locus of the
decision boundary of the system \emph{by} the \emph{intrinsic coordinate
system} of the geometric locus of the decision boundary, such that the
mathematical structure of the intrinsic coordinate system is an inherent part
of the \emph{form} of the vector algebra locus equation of
(\ref{Norm_Dec_Bound}).

It follows that any given normal random points $\mathbf{x\sim}$ $p\left(
\mathbf{x};\boldsymbol{\mu}_{1},\mathbf{\Sigma}_{1}\right)  $ and
$\mathbf{x\sim}$ $p\left(  \mathbf{x};\boldsymbol{\mu}_{2},\mathbf{\Sigma}%
_{2}\right)  $ that are \emph{solutions} of the vector algebra locus equation
of (\ref{Norm_Dec_Bound})---\emph{satisfy} \emph{both} the discriminant
function and the intrinsic coordinate system of the geometric locus of the
decision boundary of the minimum risk binary classification system.

Thereby, we recognize that the discriminant function and the intrinsic
coordinate system of the geometric locus of the decision boundary are
\emph{dual components} of the minimum risk binary classification
\emph{system}---that have different \emph{functions} and \emph{properties}.

\subsection{Dual Components of Decision Systems}

We have determined that the discriminant function and the intrinsic coordinate
system of the geometric locus of the decision boundary of any given minimum
risk binary classification system are \emph{dual components} of the system, at
which point the mathematical structure of the discriminant function and the
intrinsic coordinate system are determined by \emph{identical} vector algebra
locus equations in such a manner that the discriminant function and the
intrinsic coordinate system have \emph{different }functions and properties.
This insight will enable us to develop a mathematical model of a minimum risk
binary classification that is both inherently complex and surprisingly elegant.

We now turn out attention to the idea of the locus of a point---which we
originally defined in our working papers
\citep{Reeves2015resolving}
and
\citep{Reeves2018design}%
.

\subsection{The Locus of a Point}

The most elemental and ubiquitous component of a minimum risk binary
classification system is the locus of a point. Definition
\ref{Locus of a Point Definition} and Axiom \ref{Locus of a Point Axiom}
express the idea of the locus of a point.

\begin{definition}
\label{Locus of a Point Definition}A directed line segment that is formed by
two points $P_{\mathbf{0}}\triangleq$ $P_{\mathbf{0}}%
\begin{pmatrix}
0, & \cdots, & 0
\end{pmatrix}
$ and $P_{\mathbf{x}}\triangleq P_{\mathbf{x}}%
\begin{pmatrix}
x_{1}, & \cdots, & x_{d}%
\end{pmatrix}
$ is said to be the locus of a point $\mathbf{x}=%
\begin{pmatrix}
x_{1}, & \cdots, & x_{d}%
\end{pmatrix}
^{T}$ if and only if the distance between $P_{\mathbf{0}}$ and $P_{\mathbf{x}%
}$ is determined by the relation $\left\Vert \mathbf{x}\right\Vert =\left(
x_{1}^{2}+\cdots+x_{d}^{2}\right)  ^{1/2}$, where $\left\Vert \mathbf{x}%
\right\Vert $ is the length of the vector $\mathbf{x}$, so that each point
coordinate $x_{i}$ on the locus of $\mathbf{x}$ is located at a signed
distance $\left\Vert \mathbf{x}\right\Vert \cos\mathbb{\alpha}_{ij}$ from the
origin $P_{\mathbf{0}}$, along the direction of an orthonormal coordinate axis
$\mathbf{e}_{j}$, where $\mathbf{e}_{j}$ is a standard basis vector that
belongs to the set $\left\{  \mathbf{e}_{1}=\left(  1,0,\ldots,0\right)
,\ldots,\mathbf{e}_{d}=\left(  0,0,\ldots,1\right)  \right\}  $, and
$\mathbb{\alpha}_{ij}$ is the angle between $\mathbf{x}$ and $\mathbf{e}_{j}$.
\end{definition}

\begin{axiom}
\label{Locus of a Point Axiom}The locus $\mathbf{x}=%
\begin{pmatrix}
x_{1}, & \cdots, & x_{d}%
\end{pmatrix}
^{T}$ of any given point $\mathbf{x\in}$ $%
\mathbb{R}
^{d}$ and corresponding vector $\mathbf{x\in}$ $%
\mathbb{R}
^{d}$ is formed by $d$ point coordinates $\left\{  x_{i}\right\}  _{i=1}^{d}$
in such a manner that each point coordinate $x_{i}$ on the locus of
$\mathbf{x}$ is determined by an inner product relation between the locus of
$\mathbf{x}$ and an orthonormal coordinate axis $\mathbf{e}_{j}$%
\[
\mathbf{x}^{T}\mathbf{e}_{j}=\left\Vert \mathbf{x}\right\Vert \left\Vert
\mathbf{e}_{j}\right\Vert \cos\mathbb{\alpha}_{\mathbf{e}_{j}\mathbf{x}%
}=\left\Vert \mathbf{x}\right\Vert \cos\mathbb{\alpha}_{\mathbf{e}%
_{j}\mathbf{x}}\text{,}%
\]
where $\mathbf{e}_{j}$ belongs to the basis $\left\{  \mathbf{e}_{1}=\left(
1,0,\ldots,0\right)  ,\ldots,\mathbf{e}_{d}=\left(  0,0,\ldots,1\right)
\right\}  $, and $\mathbb{\alpha}_{ij}$ is the angle between $\mathbf{x}$ and
$\mathbf{e}_{j}$, at which point the $d$ point coordinates and the $d$ vector
components on the locus of $\mathbf{x}$ are both determined by an ordered
collection of $d$ components%
\[
\mathbf{x=}%
\begin{pmatrix}
\left\Vert \mathbf{x}\right\Vert \cos\mathbb{\alpha}_{\mathbf{e}_{1}%
\mathbf{x}}, & \left\Vert \mathbf{x}\right\Vert \cos\mathbb{\alpha
}_{\mathbf{e}_{2}\mathbf{x}}, & \cdots, & \left\Vert \mathbf{x}\right\Vert
\cos\mathbb{\alpha}_{\mathbf{e}_{d}\mathbf{x}}%
\end{pmatrix}
^{T}\text{.}%
\]

\end{axiom}

By Definition \ref{Locus of a Point Definition} and Axiom
\ref{Locus of a Point Axiom}, points and vectors will both be denoted by
$\mathbf{x}$, and the terms point and vector will be used according to context.

\subsection{Data Sources of Expected Risk}

It is well known that any given minimum risk binary classification system%
\[
\ln p\left(  \mathbf{x};\omega_{1}\right)  -\ln p\left(  \mathbf{x};\omega
_{2}\right)  \overset{\omega_{1}}{\underset{\omega_{2}}{\gtrless}}0\text{,}%
\]
where $p\left(  \mathbf{x};\omega_{1}\right)  $ and $p\left(  \mathbf{x}%
;\omega_{2}\right)  $ are certain probability density functions of two classes
$\omega_{1}$ and $\omega_{2}$ of random vectors $\mathbf{x\in}$ $%
\mathbb{R}
^{d}$, is subject to an inherent amount of uncertainty, such that the expected
risk exhibited by the system is equivalent to the probability of
classification error exhibited by the system.

We realize, however, that not all of the random points $\mathbf{x\sim}$
$p\left(  \mathbf{x};\omega_{1}\right)  $ and $\mathbf{x\sim}$ $p\left(
\mathbf{x};\omega_{2}\right)  $ generated by the probability density functions
$p\left(  \mathbf{x};\omega_{1}\right)  $ and $p\left(  \mathbf{x};\omega
_{2}\right)  $ contribute to the expected risk exhibited by the system.

To see this, take any given probability density functions $p\left(
\mathbf{x};\omega_{1}\right)  $ and $p\left(  \mathbf{x};\omega_{2}\right)  $
for two classes $\omega_{1}$ and $\omega_{2}$ of random vectors $\mathbf{x\in
}$ $%
\mathbb{R}
^{d}$ such that the probability density functions $p\left(  \mathbf{x}%
;\omega_{1}\right)  $ and $p\left(  \mathbf{x};\omega_{2}\right)  $ determine
either overlapping distributions of random points $\mathbf{x}$, wherein
$\bigcap\nolimits_{i=1}^{2}\omega_{i}\neq\emptyset$ and $\bigcap
\nolimits_{i=1}^{2}\mathbf{x}\omega_{i}\neq\emptyset$, or non-overlapping
distributions of random points $\mathbf{x}$, wherein $\bigcap\nolimits_{i=1}%
^{2}\omega_{i}=\emptyset$ and $\bigcap\nolimits_{i=1}^{2}\mathbf{x}\omega
_{i}=\emptyset$.

We realize that all of the random points $\mathbf{x\sim}$ $p\left(
\mathbf{x};\omega_{1}\right)  $ and $\mathbf{x\sim}$ $p\left(  \mathbf{x}%
;\omega_{2}\right)  $ that contribute to the expected risk exhibited by the
minimum risk binary classification system $\ln p\left(  \mathbf{x};\omega
_{1}\right)  -\ln p\left(  \mathbf{x};\omega_{2}\right)  \overset{\omega
_{1}}{\underset{\omega_{2}}{\gtrless}}0$ are located within either overlapping
regions or near tail regions of distributions determined by the probability
density functions $p\left(  \mathbf{x};\omega_{1}\right)  $ and $p\left(
\mathbf{x};\omega_{2}\right)  $.

We have named those random points that contribute to the expected risk
exhibited by a minimum risk binary classification system \textquotedblleft
extreme points\textquotedblright\ since the vector components of any given
extreme vector determine directions and locations for which a collection of
random points is most variable or spread out. We originally defined extreme
points in our working papers
\citep{Reeves2015resolving}
and
\citep{Reeves2018design}%
.

\subsection{Extreme Points}

Let $p\left(  \mathbf{x};\omega_{1}\right)  $ and $p\left(  \mathbf{x}%
;\omega_{2}\right)  $ be probability density functions of random vectors
$\mathbf{x\in}$ $%
\mathbb{R}
^{d}$ that determine either overlapping distributions of random points
$\mathbf{x}$, such that $\bigcap\nolimits_{i=1}^{2}\omega_{i}\neq\emptyset$
and $\bigcap\nolimits_{i=1}^{2}\mathbf{x}\omega_{i}\neq\emptyset$, or
non-overlapping distributions of random points $\mathbf{x}$, such that
$\bigcap\nolimits_{i=1}^{2}\omega_{i}=\emptyset$ and $\bigcap\nolimits_{i=1}%
^{2}\mathbf{x}\omega_{i}=\emptyset$, where any given random point
$\mathbf{x\sim}$ $p\left(  \mathbf{x};\omega_{1}\right)  $ or $\mathbf{x\sim}$
$p\left(  \mathbf{x};\omega_{2}\right)  $ is generated by either $p\left(
\mathbf{x};\omega_{1}\right)  $ or $p\left(  \mathbf{x};\omega_{2}\right)  $
and thereby belongs to either class $\omega_{1}$ or class $\omega_{2}$.

Now let those random points $\mathbf{x\sim}$ $p\left(  \mathbf{x};\omega
_{1}\right)  $ and $\mathbf{x\sim}$ $p\left(  \mathbf{x};\omega_{2}\right)  $
that are located within either overlapping regions or near tail regions of
distributions determined by $p\left(  \mathbf{x};\omega_{1}\right)  $ and
$p\left(  \mathbf{x};\omega_{2}\right)  $ contribute to the expected risk
exhibited by a minimum risk binary classification system $\ln p\left(
\mathbf{x};\omega_{1}\right)  -\ln p\left(  \mathbf{x};\omega_{2}\right)
\overset{\omega_{1}}{\underset{\omega_{2}}{\gtrless}}0$.

By Definition \ref{Locus of a Point Definition} and Axiom
\ref{Locus of a Point Axiom}, we can define certain geometrical and
statistical properties exhibited by a distinct random point $\mathbf{x}$ that
is located within either an overlapping region or near a tail region of
distributions determined by the probability density functions $p\left(
\mathbf{x};\omega_{1}\right)  $ and $p\left(  \mathbf{x};\omega_{2}\right)  $.
Definition \ref{Extreme Point Definition} expresses the notion of an extreme point.

\begin{definition}
\label{Extreme Point Definition}Any given random point $\mathbf{x\in}$ $%
\mathbb{R}
^{d}$ such that $\mathbf{x\sim}$ $p\left(  \mathbf{x};\omega_{1}\right)  $ or
$\mathbf{x\sim}$ $p\left(  \mathbf{x};\omega_{2}\right)  $, where certain
probability density functions $p\left(  \mathbf{x};\omega_{1}\right)  $ and
$p\left(  \mathbf{x};\omega_{2}\right)  $ determine overlapping distributions
of two classes $\omega_{1}$ and $\omega_{2}$ of random points $\mathbf{x}$, so
that $\bigcap\nolimits_{i=1}^{2}\omega_{i}\neq\emptyset$ and $\bigcap
\nolimits_{i=1}^{2}\mathbf{x}\omega_{i}\neq\emptyset$, is said to be an
extreme point if and only if the random point $\mathbf{x}$ is located within
an overlapping region of the distributions determined by $p\left(
\mathbf{x};\omega_{1}\right)  $ and $p\left(  \mathbf{x};\omega_{2}\right)  $,
at which point the vector components of the random vector $\mathbf{x}$%
\[%
\begin{pmatrix}
\left\Vert \mathbf{x}\right\Vert \cos\mathbb{\alpha}_{\mathbf{e}_{1}%
\mathbf{x}}, & \left\Vert \mathbf{x}\right\Vert \cos\mathbb{\alpha
}_{\mathbf{e}_{2}\mathbf{x}}, & \cdots, & \left\Vert \mathbf{x}\right\Vert
\cos\mathbb{\alpha}_{\mathbf{e}_{d}\mathbf{x}}%
\end{pmatrix}
^{T}%
\]
determine directions and locations for which a collection of random points
such that $\mathbf{x\sim}$ $p\left(  \mathbf{x};\omega_{1}\right)  $ and
$\mathbf{x\sim}$ $p\left(  \mathbf{x};\omega_{2}\right)  $ is most variable or
spread out.

Correspondingly, any given random point $\mathbf{x\in}$ $%
\mathbb{R}
^{d}$ such that $\mathbf{x\sim}$ $p\left(  \mathbf{x};\omega_{1}\right)  $ or
$\mathbf{x\sim}$ $p\left(  \mathbf{x};\omega_{2}\right)  $, where certain
probability density functions $p\left(  \mathbf{x};\omega_{1}\right)  $ and
$p\left(  \mathbf{x};\omega_{2}\right)  $ determine non-overlapping
distributions of two classes $\omega_{1}$ and $\omega_{2}$ of random points
$\mathbf{x}$, so that $\bigcap\nolimits_{i=1}^{2}\omega_{i}=\emptyset$ and
$\bigcap\nolimits_{i=1}^{2}\mathbf{x}\omega_{i}=\emptyset$, is said to be an
extreme point if and only if the random point $\mathbf{x}$ is located near a
tail region of the distributions determined by $p\left(  \mathbf{x};\omega
_{1}\right)  $ and $p\left(  \mathbf{x};\omega_{2}\right)  $, at which point
the vector components of the random vector $\mathbf{x}$%
\[%
\begin{pmatrix}
\left\Vert \mathbf{x}\right\Vert \cos\mathbb{\alpha}_{\mathbf{e}_{1}%
\mathbf{x}}, & \left\Vert \mathbf{x}\right\Vert \cos\mathbb{\alpha
}_{\mathbf{e}_{2}\mathbf{x}}, & \cdots, & \left\Vert \mathbf{x}\right\Vert
\cos\mathbb{\alpha}_{\mathbf{e}_{d}\mathbf{x}}%
\end{pmatrix}
^{T}%
\]
determine directions and locations for which a collection of random points
such that $\mathbf{x\sim}$ $p\left(  \mathbf{x};\omega_{1}\right)  $ and
$\mathbf{x\sim}$ $p\left(  \mathbf{x};\omega_{2}\right)  $ is most variable or
spread out.
\end{definition}

In this treatise, we express an extreme vector by $\mathbf{x}_{i\ast}$, such
that the locus of an extreme vector $\mathbf{x}_{i\ast}$ or a corresponding
extreme point $\mathbf{x}_{i\ast}$ is formally written as%
\[
\mathbf{x}_{i\ast}\mathbf{=}%
\begin{pmatrix}
\left\Vert \mathbf{x}_{i\ast}\right\Vert \cos\mathbb{\alpha}_{\mathbf{e}%
_{1}\mathbf{x}_{i\ast}}, & \left\Vert \mathbf{x}_{i\ast}\right\Vert
\cos\mathbb{\alpha}_{\mathbf{e}_{2}\mathbf{x}_{i\ast}}, & \cdots, & \left\Vert
\mathbf{x}_{i\ast}\right\Vert \cos\mathbb{\alpha}_{\mathbf{e}_{d}%
\mathbf{x}_{i\ast}}%
\end{pmatrix}
^{T}\text{.}%
\]

\subsection{Distribution Constraints on Decision Spaces}

We realize that the shape of the decision space $Z=Z_{1}\cup Z_{2}$ of any
given minimum risk binary classification system $\ln p\left(  \mathbf{x}%
;\omega_{1}\right)  -\ln p\left(  \mathbf{x};\omega_{2}\right)
\overset{\omega_{1}}{\underset{\omega_{2}}{\gtrless}}0$ is a function of
likely locations of extreme points $\mathbf{x}_{1_{\ast}}\mathbf{\sim}$
$p\left(  \mathbf{x};\omega_{1}\right)  $ and $\mathbf{x}_{2_{\ast}%
}\mathbf{\sim}$ $p\left(  \mathbf{x};\omega_{2}\right)  $, such that locations
of the extreme points $\mathbf{x}_{1_{\ast}}\mathbf{\in}$ $%
\mathbb{R}
^{d}$ and $\mathbf{x}_{2_{\ast}}\mathbf{\in}$ $%
\mathbb{R}
^{d}$ are distributed throughout the decision space $Z=Z_{1}\cup Z_{2}$ of the
system in such a manner that the extreme points $\mathbf{x}_{1_{\ast}}$ and
$\mathbf{x}_{2_{\ast}}$ are located in either overlapping distributions of two
classes $\omega_{1}$ and $\omega_{2}$ of random points $\mathbf{x\in}$ $%
\mathbb{R}
^{d}$, wherein $\bigcap\nolimits_{i=1}^{2}\omega_{i}\neq\emptyset$ and
$\bigcap\nolimits_{i=1}^{2}\mathbf{x}\omega_{i}\neq\emptyset$, or
non-overlapping distributions of two classes $\omega_{1}$ and $\omega_{2}$ of
random points $\mathbf{x\in}$ $%
\mathbb{R}
^{d}$, wherein $\bigcap\nolimits_{i=1}^{2}\omega_{i}=\emptyset$ and
$\bigcap\nolimits_{i=1}^{2}\mathbf{x}\omega_{i}=\emptyset$.

Axioms \ref{Overlapping Extreme Points Axiom} and
\ref{Non-overlapping Extreme Points Axiom} express these conditions.

\begin{axiom}
\label{Overlapping Extreme Points Axiom}Take any given probability density
functions $p\left(  \mathbf{x};\omega_{1}\right)  $ and $p\left(
\mathbf{x};\omega_{2}\right)  $ that determine overlapping distributions of
two classes $\omega_{1}$ and $\omega_{2}$ of random points $\mathbf{x\in}$ $%
\mathbb{R}
^{d}$, such that $\bigcap\nolimits_{i=1}^{2}\omega_{i}\neq\emptyset$ and
$\bigcap\nolimits_{i=1}^{2}\mathbf{x}\omega_{i}\neq\emptyset$, so that all of
the extreme points $\mathbf{x}_{1_{\ast}}\mathbf{\sim}$ $p\left(
\mathbf{x};\omega_{1}\right)  $ and $\mathbf{x}_{2_{\ast}}\mathbf{\sim}$
$p\left(  \mathbf{x};\omega_{2}\right)  $ are located within the regions of
distribution overlap.

The decision space $Z=Z_{1}\cup Z_{2}$ of the minimum risk binary
classification system $\ln p\left(  \mathbf{x};\omega_{1}\right)  -\ln
p\left(  \mathbf{x};\omega_{2}\right)  \overset{\omega_{1}}{\underset{\omega
_{2}}{\gtrless}}0$ is composed of two finite decision regions $Z_{1}$ and
$Z_{2}$---which may be contiguous or non-contiguous---that span the regions of
distribution overlap.

Thereby, locations of the extreme points $\mathbf{x}_{1_{i\ast}}$ and
$\mathbf{x}_{2_{i\ast}}$ are distributed throughout the decision space
$Z=Z_{1}\cup Z_{2}$ of the minimum risk binary classification system $\ln
p\left(  \mathbf{x};\omega_{1}\right)  -\ln p\left(  \mathbf{x};\omega
_{2}\right)  \overset{\omega_{1}}{\underset{\omega_{2}}{\gtrless}}0$ in a
symmetrically balanced manner, at which point right decisions made by the
system are related to likely locations of extreme points $\mathbf{x}%
_{1_{i\ast}}$ in the decision region $Z_{1}$ and likely locations of extreme
points $\mathbf{x}_{2_{i\ast}}$ in the decision region $Z_{2}$, whereas wrong
decisions made by the system are related to likely locations of extreme points
$\mathbf{x}_{2_{i\ast}}$ in the decision region $Z_{1}$ and likely locations
of extreme points $\mathbf{x}_{1_{i\ast}}$ in the decision region $Z_{2}$.
\end{axiom}

Figure $3$ illustrates how all of the extreme points $\mathbf{x}_{1_{i\ast}}$
and $\mathbf{x}_{2_{i\ast}}$ of overlapping distributions of random points
$\mathbf{x}$---that are determined by certain probability density functions
$p\left(  \mathbf{x};\omega_{1}\right)  $ and $p\left(  \mathbf{x};\omega
_{2}\right)  $---are located within regions of distribution overlap, wherein
locations of the extreme points are depicted by the solid indigo bar along the
horizontal axis.%

\begin{figure}[h]%
\centering
\includegraphics[
trim=0.000000in 0.000000in -0.095971in -0.377328in,
height=2.4085in,
width=5.5988in
]%
{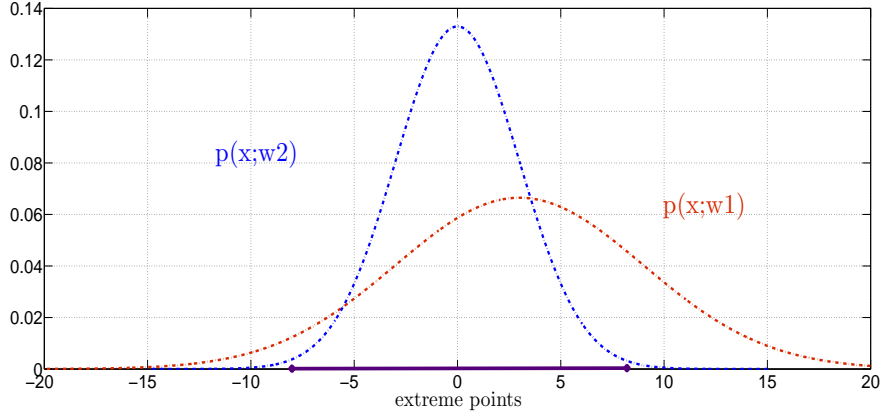}%
\caption{Given overlapping distributions of random points $\mathbf{x}$ that
are determined by certain probability density functions $p\left(
\mathbf{x};\omega_{1}\right)  $ and $p\left(  \mathbf{x};\omega_{2}\right)  $,
all of the extreme points $\mathbf{x}_{1_{\ast}}\mathbf{\sim}$ $p\left(
\mathbf{x};\omega_{1}\right)  $ and $\mathbf{x}_{2_{\ast}}\mathbf{\sim}$
$p\left(  \mathbf{x};\omega_{2}\right)  $ are located within regions of
distribution overlap, wherein locations of the extreme points $\mathbf{x}%
_{1_{i\ast}}$ and $\mathbf{x}_{2_{i\ast}}$ are depicted by the solid indigo
bar along the horizontal axis.}%
\end{figure}

\begin{axiom}
\label{Non-overlapping Extreme Points Axiom}Take any given probability density
functions $p\left(  \mathbf{x};\omega_{1}\right)  $ and $p\left(
\mathbf{x};\omega_{2}\right)  $ that determine non-overlapping distributions
of two classes $\omega_{1}$ and $\omega_{2}$ of random points $\mathbf{x\in}$
$%
\mathbb{R}
^{d}$, such that $\bigcap\nolimits_{i=1}^{2}\omega_{i}=\emptyset$ and
$\bigcap\nolimits_{i=1}^{2}\mathbf{x}\omega_{i}=\emptyset$, so that all of the
extreme points $\mathbf{x}_{1_{\ast}}\mathbf{\sim}$ $p\left(  \mathbf{x}%
;\omega_{1}\right)  $ and $\mathbf{x}_{2_{\ast}}\mathbf{\sim}$ $p\left(
\mathbf{x};\omega_{2}\right)  $ are located near the tail regions of the
non-overlapping distributions.

The decision space $Z=Z_{1}\cup Z_{2}$ of the minimum risk binary
classification system $\ln p\left(  \mathbf{x};\omega_{1}\right)  -\ln
p\left(  \mathbf{x};\omega_{2}\right)  $ is composed of two finite decision
regions $Z_{1}$ and $Z_{2}$---which are contiguous---that span the tail
regions located between the distributions.

Thereby, locations of the extreme points $\mathbf{x}_{1_{i\ast}}$ and
$\mathbf{x}_{2_{i\ast}}$ are distributed throughout the decision space
$Z=Z_{1}\cup Z_{2}$ of the minimum risk binary classification system $\ln
p\left(  \mathbf{x};\omega_{1}\right)  -\ln p\left(  \mathbf{x};\omega
_{2}\right)  \overset{\omega_{1}}{\underset{\omega_{2}}{\gtrless}}0$ in a
symmetrically balanced manner, at which point likely locations of the extreme
points $\mathbf{x}_{1_{i\ast}}$ and $\mathbf{x}_{2_{i\ast}}$ are only related
to right decisions made by the system, such that right decisions made by the
system are related to likely locations of extreme points $\mathbf{x}%
_{1_{i\ast}}$ in the decision region $Z_{1}$ and likely locations of extreme
points $\mathbf{x}_{2_{i\ast}}$ in the decision region $Z_{2}$.
\end{axiom}

Figure $4$ illustrates how all of the extreme points $\mathbf{x}_{1_{i\ast}}$
and $\mathbf{x}_{2_{i\ast}}$ of non-overlapping distributions of random points
$\mathbf{x}$---that are determined by certain probability density functions
$p\left(  \mathbf{x};\omega_{1}\right)  $ and $p\left(  \mathbf{x};\omega
_{2}\right)  $---are located within the tail regions of the non-overlapping
distributions, wherein locations of the extreme points are depicted by the
solid indigo bar along the horizontal axis.%
\begin{figure}[h]%
\centering
\includegraphics[
trim=0.000000in 0.000000in 0.013330in 0.000000in,
height=2.1024in,
width=5.5988in
]%
{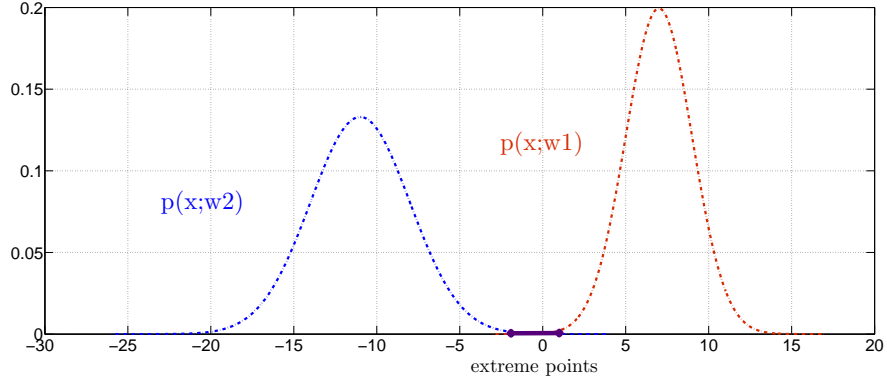}%
\caption{Given non-overlapping distributions of random points $\mathbf{x}$
that are determined by certain probability density functions $p\left(
\mathbf{x};\omega_{1}\right)  $ and $p\left(  \mathbf{x};\omega_{2}\right)  $,
all of the extreme points $\mathbf{x}_{1_{\ast}}\mathbf{\sim}$ $p\left(
\mathbf{x};\omega_{1}\right)  $ and $\mathbf{x}_{2_{\ast}}\mathbf{\sim}$
$p\left(  \mathbf{x};\omega_{2}\right)  $ are located within the tail regions
of the non-overlapping distributions, wherein locations of the extreme points
$\mathbf{x}_{1_{i\ast}}$ and $\mathbf{x}_{2_{i\ast}}$ are depicted by the
solid indigo bar along the horizontal axis.}%
\end{figure}

\subsection{Development of a General Statistical Model}

The conditions expressed by Axioms \ref{Overlapping Extreme Points Axiom} and
\ref{Non-overlapping Extreme Points Axiom}, along with the conditions
expressed by Axiom \ref{Rotation of Intrinsic Coordinate Axes Axiom}, which
demonstrates that we can find an equivalent form of a locus equation by
changing the positions of the coordinate axes of the intrinsic coordinate
system of the locus, motivate us to devise an \emph{equivalent form} of the
vector algebra locus equation of (\ref{Norm_Dec_Bound})%
\begin{align*}
&  \mathbf{x}^{T}\mathbf{\Sigma}_{1}^{-1}\mathbf{x}-2\mathbf{x}^{T}%
\mathbf{\Sigma}_{1}^{-1}\boldsymbol{\mu}_{1}-\mathbf{x}^{T}\mathbf{\Sigma}%
_{2}^{-1}\mathbf{x}+2\mathbf{x}^{T}\mathbf{\Sigma}_{2}^{-1}\boldsymbol{\mu
}_{2}\\
&  +\boldsymbol{\mu}_{1}^{T}\mathbf{\Sigma}_{1}^{-1}\boldsymbol{\mu}%
_{1}\mathbf{-}\boldsymbol{\mu}_{2}^{T}\mathbf{\Sigma}_{2}^{-1}\boldsymbol{\mu
}_{2}+\ln\left(  \left\vert \mathbf{\Sigma}_{2}\right\vert \right)
-\ln\left(  \left\vert \mathbf{\Sigma}_{1}\right\vert \right) \\
&  =0
\end{align*}
that is \textbf{determined} by the \textbf{mathematical structure of}---an
equivalent representation of the discriminant function and the intrinsic
coordinate system of the geometric locus of the decision boundary---of any
given minimum risk binary classification system, which is subject to random
vectors $\mathbf{x\sim}$ $p\left(  \mathbf{x};\omega_{1}\right)  $ and
$\mathbf{x\sim}$ $p\left(  \mathbf{x};\omega_{2}\right)  $ that are generated
by certain probability density functions $p\left(  \mathbf{x};\omega
_{1}\right)  $ and $p\left(  \mathbf{x};\omega_{2}\right)  $, \textbf{so that}:

An \emph{equivalent representation} of the discriminant function%
\[
d\left(  \mathbf{x}\right)  \triangleq\ln p\left(  \mathbf{x};\omega
_{1}\right)  -\ln p\left(  \mathbf{x};\omega_{2}\right)
\]
of an \emph{equivalent representation} of \emph{any} given minimum risk binary
classification system%
\[
\ln p\left(  \mathbf{x};\omega_{1}\right)  -\ln p\left(  \mathbf{x};\omega
_{2}\right)  \overset{\omega_{1}}{\underset{\omega_{2}}{\gtrless}}0
\]
determines \emph{likelihood values} and \emph{likely locations} of extreme
points $\mathbf{x}_{1_{\ast}}\mathbf{\sim}$ $p\left(  \mathbf{x};\omega
_{1}\right)  $ and $\mathbf{x}_{2_{\ast}}\mathbf{\sim}$ $p\left(
\mathbf{x};\omega_{2}\right)  $ located \emph{throughout} the decision space
$Z=Z_{1}\cup Z_{2}$ of the system---with respect to and in relation to---an
\emph{equivalent representation} of the intrinsic coordinate system%
\[
\ln p\left(  \mathbf{x};\omega_{1}\right)  -\ln p\left(  \mathbf{x};\omega
_{2}\right)
\]
that is an inherent part of an \emph{equivalent form }of the vector algebra
locus equation of the geometric locus of the decision boundary%
\[
\ln p\left(  \mathbf{x};\omega_{1}\right)  -\ln p\left(  \mathbf{x};\omega
_{2}\right)  =0
\]
of the system.

We use the reasoning about the formulae and the vector algebra locus equation
that are outlined above to develop a general locus formula for finding
discriminant functions of minimum risk binary classifications systems, so that
the general locus formula provides a general statistical model for a minimum
risk binary classification system.

\subsection{Development of a General Locus Formula}

The conditions expressed by Theorem \ref{Basis of Locus Formula Theorem},
Corollaries \ref{Form of Decision Boundary Corollary} -
\ref{Secondary Integral Equation Corollary}, Axioms
\ref{Overlapping Extreme Points Axiom} -
\ref{Non-overlapping Extreme Points Axiom}, and Axiom
\ref{Rotation of Intrinsic Coordinate Axes Axiom} motivate us to develop a
general locus formula for finding discriminant functions of minimum risk
binary classifications systems---that has the general form of a system of
fundamental locus equations of binary classification, subject to distinctive
geometrical and statistical conditions for a minimum risk binary
classification system in statistical equilibrium---that is satisfied by
extreme points whose coordinates are solutions of the locus equations.

\subsection{A\ Data-driven Theoretical Blueprint}

It will be seen that the general locus formula outlined above provides a
\textquotedblleft data-driven theoretical blueprint\textquotedblright\ for
resolving the inverse problem of the binary classification of random vectors,
so that the overall statistical structure and behavior and properties of a
minimum risk binary classification system are determined by transforming a
collection of observations into a data-driven mathematical model that
represents fundamental aspects of the system.

By way of motivation, we now identify how the discriminant function in
(\ref{Gaussian Rule}) determines conditional likelihood values and likely
locations of normal random vectors that are being classified. By way of
demonstration, formulae of random vectors that determine vector projections
and signed magnitudes are defined in Axiom
\ref{Statistical Structures of Hilbert Space Axiom}.

\subsection{Likelihood Values and Likely Locations}

Take any given normal random vector $\mathbf{x\sim}$ $p\left(  \mathbf{x}%
;\boldsymbol{\mu}_{1},\mathbf{\Sigma}_{1}\right)  $ or $\mathbf{x\sim}$
$p\left(  \mathbf{x};\boldsymbol{\mu}_{2},\mathbf{\Sigma}_{2}\right)  $ that
is being classified by the discriminant function in (\ref{Gaussian Rule})%
\begin{align*}
d\left(  \mathbf{x}\right)   &  =\mathbf{x}^{T}\mathbf{\Sigma}_{1}%
^{-1}\mathbf{x}-2\mathbf{x}^{T}\mathbf{\Sigma}_{1}^{-1}\boldsymbol{\mu}%
_{1}-\mathbf{x}^{T}\mathbf{\Sigma}_{2}^{-1}\mathbf{x}+2\mathbf{x}%
^{T}\mathbf{\Sigma}_{2}^{-1}\boldsymbol{\mu}_{2}\\
&  +\boldsymbol{\mu}_{1}^{T}\mathbf{\Sigma}_{1}^{-1}\boldsymbol{\mu}%
_{1}\mathbf{-}\boldsymbol{\mu}_{2}^{T}\mathbf{\Sigma}_{2}^{-1}\boldsymbol{\mu
}_{2}+\ln\left(  \left\vert \mathbf{\Sigma}_{2}\right\vert \right)
-\ln\left(  \left\vert \mathbf{\Sigma}_{1}\right\vert \right)
\end{align*}
such that the normal random vector $\mathbf{x\sim}$ $p\left(  \mathbf{x}%
;\boldsymbol{\mu}_{1},\mathbf{\Sigma}_{1}\right)  $ belongs to class
$\omega_{1}$, and the normal random vector $\mathbf{x\sim}$ $p\left(
\mathbf{x};\boldsymbol{\mu}_{2},\mathbf{\Sigma}_{2}\right)  $ belongs to class
$\omega_{2}$.

The discriminant function $d\left(  \mathbf{x}\right)  $ in
(\ref{Gaussian Rule}) determines the conditional likelihood value of the
normal random vector $\mathbf{x}$ by projecting the normal random vector
$\mathbf{x}$ onto the intrinsic vectors $\boldsymbol{s}_{1}\triangleq\left(
\mathbf{\Sigma}_{1}^{-1}\mathbf{x}-\mathbf{\Sigma}_{2}^{-1}\mathbf{x}\right)
$ and $\boldsymbol{s}_{2}\triangleq2\left(  \mathbf{\Sigma}_{1}^{-1}%
\boldsymbol{\mu}_{1}-\mathbf{\Sigma}_{2}^{-1}\boldsymbol{\mu}_{2}\right)  $,
and thereby recognizes the category $\omega_{1}$ or $\omega_{2}$ of the normal
random vector $\mathbf{x}$ from the sign of the statistical expression%
\begin{align}
d\left(  \mathbf{x}\right)   &  =\left\Vert \boldsymbol{s}_{1}\right\Vert
\left[  \left\Vert \mathbf{x}\right\Vert \cos\theta_{\boldsymbol{s}%
_{1}\mathbf{x}}\right]  -\left\Vert \boldsymbol{s}_{2}\right\Vert \left[
\left\Vert \mathbf{x}\right\Vert \cos\theta_{\boldsymbol{s}_{2}\mathbf{x}%
}\right] \tag{5.5}\label{Likelihood & Location}\\
&  +\left(  \boldsymbol{\mu}_{1}^{T}\mathbf{\Sigma}_{1}^{-1}\boldsymbol{\mu
}_{1}-\boldsymbol{\mu}_{2}^{T}\mathbf{\Sigma}_{2}^{-1}\boldsymbol{\mu}%
_{2}\right)  +\left(  \ln\left(  \left\vert \mathbf{\Sigma}_{2}\right\vert
\right)  -\ln\left(  \left\vert \mathbf{\Sigma}_{1}\right\vert \right)
\right)  \text{,}\nonumber
\end{align}
so that the value of $\operatorname{sign}\left(  d\left(  \mathbf{x}\right)
\right)  $ indicates the decision region $Z_{1}$ or $Z_{2}$ that the normal
random vector $\mathbf{x}$ is located within, at which point the signed
magnitudes of the vector projections of the normal random vector $\mathbf{x}$
onto the intrinsic vectors $\boldsymbol{s}_{1}$ and $\boldsymbol{s}_{2}$%
\[
\left\Vert \boldsymbol{s}_{1}\right\Vert \left[  \left\Vert \mathbf{x}%
\right\Vert \cos\theta_{\boldsymbol{s}_{1}\mathbf{x}}\right]  -\left\Vert
\boldsymbol{s}_{2}\right\Vert \left[  \left\Vert \mathbf{x}\right\Vert
\cos\theta_{\boldsymbol{s}_{2}\mathbf{x}}\right]
\]
constitute an \emph{implicit} random vector that determines the likely
location of the normal random vector $\mathbf{x}$ within the decision space
$Z=Z_{1}\cup Z_{2}$ of the minimum risk binary classification system $d\left(
\mathbf{x}\right)  \overset{\omega_{1}}{\underset{\omega_{2}}{\gtrless}}0$.

By the expression in (\ref{Likelihood & Location}), it follows the vector
projection of the normal random vector $\mathbf{x}$ onto the intrinsic vectors
$\boldsymbol{s}_{1}\triangleq\left(  \mathbf{\Sigma}_{1}^{-1}\mathbf{x}%
-\mathbf{\Sigma}_{2}^{-1}\mathbf{x}\right)  $ and $\boldsymbol{s}%
_{2}\triangleq2\left(  \mathbf{\Sigma}_{1}^{-1}\boldsymbol{\mu}_{1}%
-\mathbf{\Sigma}_{2}^{-1}\boldsymbol{\mu}_{2}\right)  $ determines a
conditional likelihood value \emph{and} a likely location for the normal
random vector $\mathbf{x}$---based on the vector difference between implicit
random vectors determined by $\left\Vert \boldsymbol{s}_{1}\right\Vert \left[
\left\Vert \mathbf{x}\right\Vert \cos\theta_{\boldsymbol{s}_{1}\mathbf{x}%
}\right]  $ and $\left\Vert \boldsymbol{s}_{2}\right\Vert \left[  \left\Vert
\mathbf{x}\right\Vert \cos\theta_{\boldsymbol{s}_{2}\mathbf{x}}\right]  $, at
which point the magnitudes and the directions of the implicit random vectors
are determined by signed magnitudes along the geometric loci of the intrinsic
vectors $\boldsymbol{s}_{1}$ and $\boldsymbol{s}_{2}$.

\subsubsection{Determination of Conditional Likelihood Values}

By the expression in (\ref{Likelihood & Location}), we realize that the
conditional likelihood value of any given normal random vector $\mathbf{x\sim
}$ $p\left(  \mathbf{x};\boldsymbol{\mu}_{1},\mathbf{\Sigma}_{1}\right)  $ or
$\mathbf{x\sim}$ $p\left(  \mathbf{x};\boldsymbol{\mu}_{2},\mathbf{\Sigma}%
_{2}\right)  $ is determined by distributions of the normal random vector
$\mathbf{x}$ that are conditional on distributions described by the intrinsic
vectors $\boldsymbol{s}_{1}\triangleq\left(  \mathbf{\Sigma}_{1}%
^{-1}\mathbf{x}-\mathbf{\Sigma}_{2}^{-1}\mathbf{x}\right)  $ and
$\boldsymbol{s}_{2}\triangleq2\left(  \mathbf{\Sigma}_{1}^{-1}\boldsymbol{\mu
}_{1}-\mathbf{\Sigma}_{2}^{-1}\boldsymbol{\mu}_{2}\right)  $, at which point
the intrinsic vectors $\left(  \mathbf{\Sigma}_{1}^{-1}\mathbf{x}%
-\mathbf{\Sigma}_{2}^{-1}\mathbf{x}\right)  $ and $2\left(  \mathbf{\Sigma
}_{1}^{-1}\boldsymbol{\mu}_{1}-\mathbf{\Sigma}_{2}^{-1}\boldsymbol{\mu}%
_{2}\right)  $ map covariance and distribution information for both categories
$\omega_{1}$ and $\omega_{2}$ of normal random vectors $\mathbf{x\sim}$
$p\left(  \mathbf{x};\boldsymbol{\mu}_{1},\mathbf{\Sigma}_{1}\right)  $ and
$\mathbf{x\sim}$ $p\left(  \mathbf{x};\boldsymbol{\mu}_{2},\mathbf{\Sigma}%
_{2}\right)  $ onto the normal random vector $\mathbf{x}$, where the intrinsic
vector $2\left(  \mathbf{\Sigma}_{1}^{-1}\boldsymbol{\mu}_{1}-\mathbf{\Sigma
}_{2}^{-1}\boldsymbol{\mu}_{2}\right)  $ determines a locus of average
risk---that is located on or near the decision boundary of the minimum risk
binary classification system $d\left(  \mathbf{x}\right)  \overset{\omega
_{1}}{\underset{\omega_{2}}{\gtrless}}0$.

\subsubsection{Determination of Likely Locations}

On the other hand, we realize that the likely location of any given normal
random vector $\mathbf{x\sim}$ $p\left(  \mathbf{x};\boldsymbol{\mu}%
_{1},\mathbf{\Sigma}_{1}\right)  $ or $\mathbf{x\sim}$ $p\left(
\mathbf{x};\boldsymbol{\mu}_{2},\mathbf{\Sigma}_{2}\right)  $ is determined by
the signed magnitudes of the normal random vector $\mathbf{x}$ along the
geometric loci of the intrinsic vectors $\boldsymbol{s}_{1}\triangleq\left(
\mathbf{\Sigma}_{1}^{-1}\mathbf{x}-\mathbf{\Sigma}_{2}^{-1}\mathbf{x}\right)
$ and $\boldsymbol{s}_{2}\triangleq2\left(  \mathbf{\Sigma}_{1}^{-1}%
\boldsymbol{\mu}_{1}-\mathbf{\Sigma}_{2}^{-1}\boldsymbol{\mu}_{2}\right)  $,
at which point the statistical expression $2\mathbf{x}^{T}\left(
\mathbf{\Sigma}_{1}^{-1}\boldsymbol{\mu}_{1}-\mathbf{\Sigma}_{2}%
^{-1}\boldsymbol{\mu}_{2}\right)  $ determines the distance that the normal
random vector $\mathbf{x}$ is located from a locus of average risk, and the
statistical expression $\mathbf{x}^{T}\left(  \mathbf{\Sigma}_{1}%
^{-1}\mathbf{x}-\mathbf{\Sigma}_{2}^{-1}\mathbf{x}\right)  $ determines a
likely location of the normal random vector $\mathbf{x}$ that is conditional
on how the locus of the normal random vector $\mathbf{x}$ is distributed along
the locus of the intrinsic vector $\boldsymbol{s}_{1}\triangleq\left(
\mathbf{\Sigma}_{1}^{-1}\mathbf{x}-\mathbf{\Sigma}_{2}^{-1}\mathbf{x}\right)
$.

Thereby, we realize that the likely location of any given normal random vector
$\mathbf{x\sim}$ $p\left(  \mathbf{x};\boldsymbol{\mu}_{1},\mathbf{\Sigma}%
_{1}\right)  $ or $\mathbf{x\sim}$ $p\left(  \mathbf{x};\boldsymbol{\mu}%
_{2},\mathbf{\Sigma}_{2}\right)  $---within the decision space $Z=Z_{1}\cup
Z_{2}$ of a minimum risk binary classification system---is gauged relative to
its position from the locus of the decision boundary of the system.

\subsubsection{Interdependence of Likelihoods and Likely Locations}

Since the discriminant function $d\left(  \mathbf{x}\right)  $ in
(\ref{Gaussian Rule}) uses the coordinates $\left\{  x_{i}\right\}  _{i=1}%
^{d}$ of a normal random vector $\mathbf{x}$ in the statistical expression%
\[
\mathbf{x}^{T}\left[  \left(  \mathbf{\Sigma}_{1}^{-1}\mathbf{x}%
-\mathbf{\Sigma}_{2}^{-1}\mathbf{x}\right)  -2\left(  \mathbf{\Sigma}_{1}%
^{-1}\boldsymbol{\mu}_{1}-\mathbf{\Sigma}_{2}^{-1}\boldsymbol{\mu}_{2}\right)
\right]
\]
to determine the position of the normal random vector $\mathbf{x}$ within the
decision space $Z=Z_{1}\cup Z_{2}$ of the minimum risk binary classification
system%
\begin{align*}
d\left(  \mathbf{x}\right)   &  \triangleq\mathbf{x}^{T}\mathbf{\Sigma}%
_{1}^{-1}\mathbf{x}-2\mathbf{x}^{T}\mathbf{\Sigma}_{1}^{-1}\boldsymbol{\mu
}_{1}+\boldsymbol{\mu}_{1}^{T}\mathbf{\Sigma}_{1}^{-1}\boldsymbol{\mu}_{1}%
-\ln\left(  \left\vert \mathbf{\Sigma}_{1}\right\vert \right) \\
&  -\mathbf{x}^{T}\mathbf{\Sigma}_{2}^{-1}\mathbf{x}+2\mathbf{x}%
^{T}\mathbf{\Sigma}_{2}^{-1}\boldsymbol{\mu}_{2}\mathbf{-}\boldsymbol{\mu}%
_{2}^{T}\mathbf{\Sigma}_{2}^{-1}\boldsymbol{\mu}_{2}+\ln\left(  \left\vert
\mathbf{\Sigma}_{2}\right\vert \right)  \overset{\omega_{1}}{\underset{\omega
_{2}}{\gtrless}}0\text{,}%
\end{align*}
it follows that the conditional likelihood value $d\left(  \mathbf{x}\right)
$ of the normal random vector $\mathbf{x}$ is determined by its \emph{likely
location} within a decision region $Z_{1}$ or $Z_{2}$ of the system $d\left(
\mathbf{x}\right)  \overset{\omega_{1}}{\underset{\omega_{2}}{\gtrless}}0$.

Given the above analysis---wherein we identified how the discriminant function
in (\ref{Gaussian Rule}) determines conditional likelihood values and likely
locations of normal random vectors that are being classified---we realize that
the algebraic vector expression%
\[
\mathbf{x}^{T}\mathbf{\Sigma}_{1}^{-1}\mathbf{x-x}^{T}\mathbf{\Sigma}_{2}%
^{-1}\mathbf{x}%
\]
determines the mathematical structure of both the discriminant function
\emph{and} the intrinsic coordinate system---of the geometric locus of the
decision boundary---of any given minimum risk binary classification that is
subject to multivariate normal data.

We are now in a position to lay the groundwork for identifying the geometrical
and statistical essence of an exclusive principal eigen-coordinate
system---which is the solution of an equivalent form of the vector algebra
locus equation of (\ref{Norm_Dec_Bound}).

\section{\label{Section 6}Exclusive Principal Eigen-coordinate Systems}

Most surprisingly, the graph of the vector algebra locus equation of
(\ref{Norm_Dec_Bound})%
\begin{align*}
d\left(  \mathbf{x}\right)   &  :\mathbf{x}^{T}\mathbf{\Sigma}_{1}%
^{-1}\mathbf{x}-2\mathbf{x}^{T}\mathbf{\Sigma}_{1}^{-1}\boldsymbol{\mu}%
_{1}+\boldsymbol{\mu}_{1}^{T}\mathbf{\Sigma}_{1}^{-1}\boldsymbol{\mu}_{1}%
-\ln\left(  \left\vert \mathbf{\Sigma}_{1}\right\vert \right) \\
&  -\mathbf{x}^{T}\mathbf{\Sigma}_{2}^{-1}\mathbf{x}+2\mathbf{x}%
^{T}\mathbf{\Sigma}_{2}^{-1}\boldsymbol{\mu}_{2}\mathbf{-}\boldsymbol{\mu}%
_{2}^{T}\mathbf{\Sigma}_{2}^{-1}\boldsymbol{\mu}_{2}+\ln\left(  \left\vert
\mathbf{\Sigma}_{2}\right\vert \right)  =0
\end{align*}
\emph{always represents} a distinctive conic section or quadratic
surface---that constitutes the \emph{geometric locus} of a \emph{decision
boundary}---that divides the decision space of the minimum risk binary
classification system in (\ref{Gaussian Rule}) into symmetrical decision
regions---so that the discriminant function of the system exhibits the minimum
probability of classification error.

We have discovered that the shape and the fundamental properties exhibited by
the geometric locus of any given decision boundary---that is represented by
the graph of the vector algebra locus equation of (\ref{Norm_Dec_Bound})---are
regulated in the following manner.

\subsection{Shape and Property Regulation of Decision Boundaries}

We have discovered that the \emph{shape} and the \emph{fundamental properties}
exhibited by the geometric \emph{locus} of the decision boundary of any given
minimum risk binary classification system, subject to multivariate normal
vectors $\mathbf{x\in}$ $%
\mathbb{R}
^{d}$ such that $\mathbf{x\sim}$ $p\left(  \mathbf{x};\boldsymbol{\mu}%
_{1},\mathbf{\Sigma}_{1}\right)  $ and $\mathbf{x\sim}$ $p\left(
\mathbf{x};\boldsymbol{\mu}_{2},\mathbf{\Sigma}_{2}\right)  $, are
\emph{regulated} by the structure and the fundamental properties exhibited by
a \emph{mixture} of geometrical and statistical \emph{components---}produced
by\emph{ a novel principal eigen-coordinate transform} of the algebraic vector
expressions $\mathbf{x}^{T}\mathbf{\Sigma}_{1}^{-1}\mathbf{x}$ and
$\mathbf{x}^{T}\mathbf{\Sigma}_{2}^{-1}\mathbf{x}$---that \emph{jointly
determine} the mathematical structure of the discriminant function \emph{and}
the intrinsic coordinate system of the geometric locus of the decision
boundary of the system.

Thereby, we have discovered that principal components of the transformed
algebraic vector expression $\mathbf{x}^{T}\mathbf{\Sigma}_{1}^{-1}%
\mathbf{x-x}^{T}\mathbf{\Sigma}_{2}^{-1}\mathbf{x}$ in the vector algebra
locus equation of (\ref{Norm_Dec_Bound}) are \emph{blended together} in such a
manner that the discriminant function and the intrinsic coordinate system in
(\ref{Gaussian Rule}) are \emph{dual components} of a minimum risk binary
classification system---that have different \emph{functions} and
\emph{properties}.

We have also discovered that a \emph{pair} of \emph{signed random quadratic
forms} jointly provide \emph{dual representation} of the discriminant function
and the intrinsic coordinate system---of the geometric locus of the decision
boundary---of any given minimum risk binary classification system in
(\ref{Gaussian Rule}).

\subsection{Random Quadratic Forms of Decision Systems}

We realize that the algebraic vector expression in (\ref{Gaussian Rule}) and
(\ref{Norm_Dec_Bound})%
\[
\mathbf{x}^{T}\mathbf{\Sigma}_{1}^{-1}\mathbf{x-x}^{T}\mathbf{\Sigma}_{2}%
^{-1}\mathbf{x}%
\]
constitutes a pair of \emph{signed} \emph{random quadratic forms }%
$\mathbf{x}^{T}\mathbf{\Sigma}_{1}^{-1}\mathbf{x}$ and $-\mathbf{x}%
^{T}\mathbf{\Sigma}_{2}^{-1}\mathbf{x}$ that jointly provide dual
representation of the discriminant function and the intrinsic coordinate
system---of the geometric locus of the decision boundary---of any given
minimum risk binary classification system, subject to multivariate normal
data, at which point the signed random quadratic forms $\mathbf{x}%
^{T}\mathbf{\Sigma}_{1}^{-1}\mathbf{x}$ and $-\mathbf{x}^{T}\mathbf{\Sigma
}_{2}^{-1}\mathbf{x}$ and corresponding normal random vectors $\mathbf{x\sim}$
$p\left(  \mathbf{x};\boldsymbol{\mu}_{1},\mathbf{\Sigma}_{1}\right)  $ and
$\mathbf{x\sim}$ $p\left(  \mathbf{x};\boldsymbol{\mu}_{2},\mathbf{\Sigma}%
_{2}\right)  $ are solutions of the vector algebra locus equation of
(\ref{Norm_Dec_Bound})%
\begin{align*}
d\left(  \mathbf{x}\right)   &  :\mathbf{x}^{T}\mathbf{\Sigma}_{1}%
^{-1}\mathbf{x}-2\mathbf{x}^{T}\mathbf{\Sigma}_{1}^{-1}\boldsymbol{\mu}%
_{1}+\boldsymbol{\mu}_{1}^{T}\mathbf{\Sigma}_{1}^{-1}\boldsymbol{\mu}_{1}%
-\ln\left(  \left\vert \mathbf{\Sigma}_{1}\right\vert \right) \\
&  -\mathbf{x}^{T}\mathbf{\Sigma}_{2}^{-1}\mathbf{x}+2\mathbf{x}%
^{T}\mathbf{\Sigma}_{2}^{-1}\boldsymbol{\mu}_{2}\mathbf{-}\boldsymbol{\mu}%
_{2}^{T}\mathbf{\Sigma}_{2}^{-1}\boldsymbol{\mu}_{2}+\ln\left(  \left\vert
\mathbf{\Sigma}_{2}\right\vert \right)  =0\text{.}%
\end{align*}

Thereby, given the conditions expressed by Axioms
\ref{Overlapping Extreme Points Axiom} and
\ref{Non-overlapping Extreme Points Axiom}, which demonstrate that the shape
of the decision space $Z=Z_{1}\cup Z_{2}$ of any given minimum risk binary
classification system $\ln p\left(  \mathbf{x};\omega_{1}\right)  -\ln
p\left(  \mathbf{x};\omega_{2}\right)  \overset{\omega_{1}}{\underset{\omega
_{2}}{\gtrless}}0$ is a function of likely locations of extreme points
$\mathbf{x}_{1_{\ast}}\mathbf{\sim}$ $p\left(  \mathbf{x};\omega_{1}\right)  $
and $\mathbf{x}_{2_{\ast}}\mathbf{\sim}$ $p\left(  \mathbf{x};\omega
_{2}\right)  $, along with the conditions expressed by Axiom
\ref{Rotation of Intrinsic Coordinate Axes Axiom}, which demonstrates that we
can find an equivalent form of a locus equation by changing the positions of
the coordinate axes of the intrinsic coordinate system of the locus, we are
motivated to determine how we can find an equivalent form of the vector
algebra locus equation of (\ref{Norm_Dec_Bound})---so that \emph{extreme
points} $\mathbf{x}_{1_{\ast}}\mathbf{\sim}p\left(  \mathbf{x};\boldsymbol{\mu
}_{1},\mathbf{\Sigma}_{1}\right)  $ and $\mathbf{x}_{2_{\ast}}\mathbf{\sim
}p\left(  \mathbf{x};\boldsymbol{\mu}_{2},\mathbf{\Sigma}_{2}\right)  $ have
coordinates that are \emph{solutions} of the \emph{equivalent form} of the
vector algebra locus equation---by discovering \emph{how} we can
\emph{transform} the positions of the \emph{basis} of the \emph{intrinsic
coordinate system}%
\[
\mathbf{x}^{T}\mathbf{\Sigma}_{1}^{-1}\mathbf{x-x}^{T}\mathbf{\Sigma}_{2}%
^{-1}\mathbf{x}\text{,}%
\]
such that random points $\mathbf{x}$ are generated according to $\mathbf{x\sim
}$ $p\left(  \mathbf{x};\boldsymbol{\mu}_{1},\mathbf{\Sigma}_{1}\right)  $ and
$\mathbf{x\sim}$ $p\left(  \mathbf{x};\boldsymbol{\mu}_{2},\mathbf{\Sigma}%
_{2}\right)  $, so that likelihood values \emph{and} likely locations of
extreme points $\mathbf{x}_{1_{\ast}}\mathbf{\sim}p\left(  \mathbf{x}%
;\boldsymbol{\mu}_{1},\mathbf{\Sigma}_{1}\right)  $ and $\mathbf{x}_{2_{\ast}%
}\mathbf{\sim}p\left(  \mathbf{x};\boldsymbol{\mu}_{2},\mathbf{\Sigma}%
_{2}\right)  $ \emph{determine} the \emph{positions} of the \emph{basis} of
the \emph{transformed} intrinsic coordinate system.

\subsection{Finding a Suitable Change of Coordinate System}

Lemma \ref{Equivalent Form of Algebraic Equation Lemma} is an important result
that clearly defines the algebraic and geometric essence of an exclusive
principal eigen-coordinate system that is the solution of an equivalent form
of the vector algebra locus equation of (\ref{Norm_Dec_Bound}).

\begin{lemma}
\label{Equivalent Form of Algebraic Equation Lemma}Let the locus of an
algebraic equation be any given conic section or quadratic surface. Then an
equivalent form of the algebraic equation exists that is determined by the
mathematical structure of the major intrinsic axis of the locus---which
coincides as the principal eigenaxis of the locus---so that the principal
eigenaxis is the exclusive coordinate axis of the locus; the principal
eigenaxis satisfies the locus in terms of its eigenenergy; and the uniform
property exhibited by all of the points that lie on the locus is the
eigenenergy exhibited by the principal eigenaxis of the locus.
\end{lemma}

\begin{proof}
Lemma \ref{Equivalent Form of Algebraic Equation Lemma} is a generalization of
conditions expressed by Theorems
\ref{Vector Algebra Equation of Linear Loci Theorem} -
\ref{Vector Algebra Equation of Spherical Loci Theorem}, whose proofs can be
found in our working paper
\citep{Reeves2018design}%
.
\end{proof}

Theorems \ref{Vector Algebra Equation of Linear Loci Theorem} -
\ref{Vector Algebra Equation of Spherical Loci Theorem} are presented below.
Each theorem expresses the form of a general vector algebra locus equation of
a class of conic sections and quadratic surfaces.

\subsection{An Exclusive and Distinctive Coordinate Axis}

Conic sections and quadratic surfaces are represented by algebraic equations,
such that any given conic section or quadratic surface is the graph of an
algebraic equation, so that the intrinsic coordinate system of any given locus
of points is either a Cartesian coordinate system in standard position or a
transformed Cartesian coordinate system---at which point each axis of a
Cartesian coordinate system is rotated in a consistent manner
\citep{Eisenhart1939,Hewson2009,Lay2006,Nichols1893,Tanner1898}%
.

We have devised general vector algebra locus equations for each class of conic
sections and quadratic surfaces, including lines, planes, and hyperplanes,
such that the form of each vector algebra locus equation is determined by the
principal eigenaxis of a locus, so that the principal eigenaxis is the
exclusive coordinate axis of the locus; the principal eigenaxis satisfies the
locus in terms of its eigenenergy; and the uniform property exhibited by all
of the points that lie on the locus is the eigenenergy exhibited by the
principal eigenaxis of the locus.

Theorem \ref{Vector Algebra Equation of Linear Loci Theorem} expresses the
form of a general vector algebra locus equation of lines, planes and
hyperplanes in $d$-dimensional Hilbert space.

\begin{theorem}
\label{Vector Algebra Equation of Linear Loci Theorem}Let a general vector
algebra locus equation of any given line, plane or hyperplane in
$d$-dimensional Hilbert space be given by%
\begin{equation}
\mathbf{x}^{T}\boldsymbol{\nu}=\left\Vert \boldsymbol{\nu}\right\Vert
^{2}\text{,} \tag{6.1}\label{Linear Locus}%
\end{equation}
where $\mathbf{x\in}$ $%
\mathbb{R}
^{d}$ is a point on the locus, $\boldsymbol{\nu}$ $\mathbf{\in}$ $%
\mathbb{R}
^{d}$ is the principal eigenaxis of the locus, and $\left\Vert \boldsymbol{\nu
}\right\Vert ^{2}$ is the eigenenergy exhibited by the principal eigenaxis
$\boldsymbol{\nu}$, at which point the principal eigenaxis $\boldsymbol{\nu}$
is an exclusive principal eigen-coordinate system of the locus.

Thereby, the geometric locus of any given line, plane or hyperplane in
$d$-dimensional Hilbert space is represented by the graph of a vector algebra
locus equation that has the form%
\[
\mathbf{x}^{T}\boldsymbol{\nu}=\left\Vert \boldsymbol{\nu}\right\Vert
^{2}\text{,}%
\]
so that the principal eigenaxis $\boldsymbol{\nu}$ is an exclusive principal
eigen-coordinate system of the geometric locus; the principal eigenaxis
$\boldsymbol{\nu}$ satisfies the geometric locus in terms of its eigenenergy
$\left\Vert \boldsymbol{\nu}\right\Vert ^{2}$; and the uniform property
exhibited by all of the points $\mathbf{x}$ that lie on the geometric locus is
the eigenenergy $\left\Vert \boldsymbol{\nu}\right\Vert ^{2}$ exhibited by the
principal eigenaxis $\boldsymbol{\nu}$ of the geometric locus.
\end{theorem}

Theorem \ref{Vector Algebra Equation of Linear Loci Theorem} is proved in our
working paper
\citep{Reeves2018design}%
.

Theorem \ref{Vector Algebra Equation of Quadratic Loci Theorem} expresses the
form of a general vector algebra locus equation of ellipses, hyperbolas and
parabolas in $d$-dimensional Hilbert space.

\begin{theorem}
\label{Vector Algebra Equation of Quadratic Loci Theorem}Let a general vector
algebra locus equation of any given ellipse, hyperbola or parabola in
$d$-dimensional Hilbert space be given by%
\begin{equation}
2\mathbf{x}^{T}\boldsymbol{\nu}-\left\Vert \mathbf{x}\right\Vert ^{2}+\left(
e^{2}\cos^{2}\theta\right)  \left\Vert \mathbf{x}\right\Vert ^{2}=\left\Vert
\boldsymbol{\nu}\right\Vert ^{2}\text{,} \tag{6.2}\label{Quadratic Locus}%
\end{equation}
where $\mathbf{x\in}$ $%
\mathbb{R}
^{d}$ is a point on the locus, $\boldsymbol{\nu}$ $\mathbf{\in}$ $%
\mathbb{R}
^{d}$ is the principal eigenaxis of the locus, $\theta$ is the angle between
$\mathbf{x}$ and $\boldsymbol{\nu}$ , $e$ is the eccentricity of the locus,
$\left\Vert \mathbf{x}\right\Vert ^{2}$ is the squared length of the vector
$\mathbf{x}$, and $\left\Vert \boldsymbol{\nu}\right\Vert ^{2}$ is the
eigenenergy exhibited by the principal eigenaxis $\boldsymbol{\nu}$, at which
point the principal eigenaxis $\boldsymbol{\nu}$ is an exclusive principal
eigen-coordinate system of the locus.

Thereby, the geometric locus of any given ellipse, hyperbola or parabola in
$d$-dimensional Hilbert space is represented by the graph of a vector algebra
locus equation that has the form%
\[
2\mathbf{x}^{T}\boldsymbol{\nu}-\left\Vert \mathbf{x}\right\Vert ^{2}+\left(
e^{2}\cos^{2}\theta\right)  \left\Vert \mathbf{x}\right\Vert ^{2}=\left\Vert
\boldsymbol{\nu}\right\Vert ^{2}\text{,}%
\]
so that the principal eigenaxis $\boldsymbol{\nu}$ is an exclusive principal
eigen-coordinate system of the geometric locus; the principal eigenaxis
$\boldsymbol{\nu}$ satisfies the geometric locus in terms of its eigenenergy
$\left\Vert \boldsymbol{\nu}\right\Vert ^{2}$; and the uniform property
exhibited by all of the points $\mathbf{x}$ that lie on the geometric locus is
the eigenenergy $\left\Vert \boldsymbol{\nu}\right\Vert ^{2}$ exhibited by the
principal eigenaxis $\boldsymbol{\nu}$ of the geometric locus.
\end{theorem}

Theorem \ref{Vector Algebra Equation of Quadratic Loci Theorem} is proved in
our working paper
\citep{Reeves2018design}%
.

Theorem \ref{Vector Algebra Equation of Spherical Loci Theorem} expresses the
form of a general vector algebra locus equation of circles and spheres in
$d$-dimensional Hilbert space.

\begin{theorem}
\label{Vector Algebra Equation of Spherical Loci Theorem}Let a general vector
algebra locus equation of any given circle or sphere in $d$-dimensional
Hilbert space be given by%
\begin{equation}
2\mathbf{x}^{T}\boldsymbol{\nu}-\left\Vert \mathbf{x}\right\Vert
^{2}+\left\Vert \mathbf{r}\right\Vert ^{2}=\left\Vert \boldsymbol{\nu
}\right\Vert ^{2}\text{,} \tag{6.3}\label{Spherical Locus}%
\end{equation}
where $\mathbf{x\in}$ $%
\mathbb{R}
^{d}$ is a point on the locus, $\boldsymbol{\nu}$ $\mathbf{\in}$ $%
\mathbb{R}
^{d}$ is the principal eigenaxis of the locus, $\left\Vert \mathbf{x}%
\right\Vert ^{2}$ is the squared length of the vector $\mathbf{x}$,
$\mathbf{r}$ is the radius of the locus, $\left\Vert \mathbf{r}\right\Vert
^{2}$ is the squared length of the radius $\mathbf{r}$, and $\left\Vert
\boldsymbol{\nu}\right\Vert ^{2}$ is the eigenenergy exhibited by the
principal eigenaxis $\boldsymbol{\nu}$, at which point the principal eigenaxis
$\boldsymbol{\nu}$ is an exclusive principal eigen-coordinate system of the locus.

Thereby, the geometric locus of any given circle or sphere in $d$-dimensional
Hilbert space is represented by the graph of a vector algebra locus equation
that has the form%
\[
2\mathbf{x}^{T}\boldsymbol{\nu}-\left\Vert \mathbf{x}\right\Vert
^{2}+\left\Vert \mathbf{r}\right\Vert ^{2}=\left\Vert \boldsymbol{\nu
}\right\Vert ^{2}\text{,}%
\]
so that the principal eigenaxis $\boldsymbol{\nu}$ is an exclusive principal
eigen-coordinate system of the geometric locus; the principal eigenaxis
$\boldsymbol{\nu}$ satisfies the geometric locus in terms of its eigenenergy
$\left\Vert \boldsymbol{\nu}\right\Vert ^{2}$; and the uniform property
exhibited by all of the points $\mathbf{x}$ that lie on the geometric locus is
the eigenenergy $\left\Vert \boldsymbol{\nu}\right\Vert ^{2}$ exhibited by the
principal eigenaxis $\boldsymbol{\nu}$ of the geometric locus.
\end{theorem}

Theorem \ref{Vector Algebra Equation of Spherical Loci Theorem} is proved in
our working paper
\citep{Reeves2018design}%
.

\subsection{Generatrices of Quadratic Curves and Surfaces}

Geometric figures can be defined in two ways: $(1)$ as a figure with certain
known properties; and $(2)$ as the path of a point which moves under known
conditions
\citep{Nichols1893,Tanner1898}%
. The path of a point which moves under known conditions is called a
\emph{generatrix}. Well-known generatrices include quadratic curves and
surfaces
\citep{Hilbert1952,thomas1995calculus}%
. We used the definition of a generatrix of a quadratic curve or surface
contained in (\ref{Quadratic Locus}) to devise the general vector algebra
locus equations and (\ref{Spherical Locus})
\citep{Reeves2018design}%
.

A generatrix is a point $P_{\mathbf{x}}$ which moves along a given path such
that the path generates a curve or surface. Three of the quadratic curves and
surfaces are traced by a point $P_{\mathbf{x}}$ which moves so that its
distance from a fixed point $P_{\mathbf{f}}$ always bears a constant ratio to
its distance from a fixed line, plane, or hyperplane $D$. Quadratic curves and
surfaces that are generated in this manner include $d$-dimensional parabolas,
hyperbolas and ellipses
\citep{thomas1995calculus,Zwillinger1996}%
. The geometric nature of this generatrix in $2$-dimensional Hilbert space can
be described as follows.

Take a fixed point $P_{\mathbf{f}}$ in $2$-dimensional Hilbert space, a line
$D$ not going through $P_{\mathbf{f}}$, and a positive real number $e$. The
set of points $P_{\mathbf{x}}$ such that the distance from $P_{\mathbf{x}}$ to
$P_{\mathbf{f}}$ is $e$ times the shortest distance from $P_{\mathbf{x}}$ to
$D$, where distance is measured along a perpendicular, is a locus of points
termed a conic section. For any given conic section, the point $P_{\mathbf{f}%
}$ is called the focus, the line $D$ is called the directrix, and the term $e$
is called the eccentricity. If $e<1$, the conic is an ellipse; if $e=1$, the
conic is a parabola; if $e>1$, the conic is an hyperbola. The definition of a
conic section is readily generalized to quadratic surfaces by taking a fixed
point $P_{\mathbf{f}}$ in $%
\mathbb{R}
^{d}$, a $\left(  d-1\right)  $-dimensional hyperplane not going through
$P_{\mathbf{f}}$, and a positive real number $e$
\citep{thomas1995calculus,Zwillinger1996}%
.

Accordingly, quantities that determine the size and shape of any given
quadratic curve or surface are its eccentricity $e$ and the distance of the
focus $P_{\mathbf{f}}$ from the directrix $D$.

We realize that the focus $P_{\mathbf{f}}$ of any given quadratic curve or
surface can be represented by the principal eigenaxis $\boldsymbol{\nu}$ of
the geometric locus of the quadratic curve or surface. Thereby, given Theorems
\ref{Vector Algebra Equation of Linear Loci Theorem} -
\ref{Vector Algebra Equation of Spherical Loci Theorem}, we recognize that the
distance of the focus $P_{\mathbf{f}}$ from the directrix $D$ of any given
quadratic curve or surface is determined by the length $\left\Vert
\boldsymbol{\nu}\right\Vert $ of the principal eigenaxis $\boldsymbol{\nu}$ of
the quadratic curve or surface.

The eccentricity of a quadratic curve or surface is a non-negative real number
that uniquely characterizes the shape and describes the general proportions of
the quadratic curve or surface
\citep{thomas1995calculus}%
. Thus, two quadratic curves or surfaces are similar if and only if they have
the same eccentricity. We have devised vector algebra locus equations that
determine the eccentricity for each class of conic sections and quadratic
surfaces, including lines, planes, and hyperplanes.

\subsection{Eccentricity of Ellipses, Hyperbolas and Parabolas}

We have demonstrated that the geometric locus of any given ellipse, hyperbola
or parabola in $d$-dimensional Hilbert space satisfies the vector algebra
locus equation%
\[
\left\Vert \mathbf{x-}\boldsymbol{\nu}\right\Vert =e\times\left\Vert
\mathbf{x}\right\Vert \cos\theta\text{,}%
\]
so that the scaled $e$ signed magnitude $\left\Vert \mathbf{x}\right\Vert
\cos\theta$ determined by the vector projection of any given point
$\mathbf{x}$ that lies on the locus---onto the principal eigenaxis
$\boldsymbol{\nu}$ of the locus---determines the distance between the point
$\mathbf{x}$ and the directrix $D$ of the locus, where $e$ is the eccentricity
of the locus
\citep{Reeves2018design}%
.

\subsection{The Law of Cosines for Vectors}

The relationship $\mathbf{x}^{T}\boldsymbol{y}=\left\Vert \mathbf{x}%
\right\Vert \left\Vert \boldsymbol{y}\right\Vert \cos\theta$ between two
vectors $\mathbf{x}$ and $\boldsymbol{y}$ in $d$-dimensional Hilbert space can
be derived by using the law of cosines
\citep{Lay2006}%
\[
\left\Vert \mathbf{x}-\boldsymbol{y}\right\Vert ^{2}=\left\Vert \mathbf{x}%
\right\Vert ^{2}+\left\Vert \boldsymbol{y}\right\Vert ^{2}-2\left\Vert
\mathbf{x}\right\Vert \left\Vert \boldsymbol{y}\right\Vert \cos\theta
\]
which reduces to%
\begin{align*}
\left\Vert \mathbf{x}\right\Vert \left\Vert \boldsymbol{y}\right\Vert
\cos\theta &  =x_{1}y_{1}+x_{2}y_{2}+\cdots+x_{d}y_{d}\\
&  =\mathbf{x}^{T}\boldsymbol{y}=\boldsymbol{y}^{T}\mathbf{x}\text{,}%
\end{align*}
at which point the distance $\left\Vert \mathbf{x-}\boldsymbol{y}\right\Vert $
between the vectors $\mathbf{x}$ and $\boldsymbol{y}$ is correlated with the
inner product relationship $\mathbf{x}^{T}\boldsymbol{y}=\boldsymbol{y}%
^{T}\mathbf{x}$.

\subsection{Eccentricity of Lines, Planes, and Hyperplanes}

Let $e$ denote the eccentricity of any given quadratic curve or surface in
$d$-dimensional Hilbert space. Now take the general vector algebra locus
equation of any given line, plane or hyperplane in $d$-dimensional Hilbert
space%
\[
\mathbf{x}^{T}\boldsymbol{\nu}=\left\Vert \boldsymbol{\nu}\right\Vert
^{2}\text{,}%
\]
where $\mathbf{x}$ is a point on the locus and $\boldsymbol{\nu}$ is the
principal eigenaxis of the locus, at which point the vector $\mathbf{x}$ and
the principal eigenaxis $\boldsymbol{\nu}$ satisfy the relation%
\[
\mathbf{x}^{T}\boldsymbol{\nu}=\left\Vert \mathbf{x}\right\Vert \left\Vert
\boldsymbol{\nu}\right\Vert \cos\theta\text{,}%
\]
where $\theta$ is the acute angle between $\boldsymbol{\nu}$ and $\mathbf{x}$.

It follows that all of the points $\mathbf{x}$ that lie on the locus of any
given line, plane or hyperplane---including the principal eigenaxis
$\boldsymbol{\nu}$ of the locus---satisfy the relation%
\[
\left\Vert \mathbf{x}-\boldsymbol{\nu}\right\Vert =\left\Vert \mathbf{x}%
\right\Vert \left\Vert \boldsymbol{\nu}\right\Vert \cos\theta\text{,}%
\]
such that the distance between any given point $\mathbf{x}$ on the locus and
the principal eigenaxis $\boldsymbol{\nu}$ of the locus satisfies the inner
product relation $\left\Vert \mathbf{x}\right\Vert \left\Vert \boldsymbol{\nu
}\right\Vert \cos\theta$.

Thereby, the eccentricity $e$ of any given line, plane or hyperplane in
$d$-dimensional Hilbert space is determined by the relation%
\begin{align*}
\left\Vert \mathbf{x-}\boldsymbol{\nu}\right\Vert  &  =e\times\left\Vert
\mathbf{x}\right\Vert \cos\theta\\
&  =\left\Vert \boldsymbol{\nu}\right\Vert \left\Vert \mathbf{x}\right\Vert
\cos\theta\text{,}%
\end{align*}
so that the eccentricity $e$ of the geometric locus of the line, plane or
hyperplane is determined by the length $\left\Vert \boldsymbol{\nu}\right\Vert
$ of the principal eigenaxis $\boldsymbol{\nu}$ of the locus.

\subsection{Eccentricity of Circles and Spheres}

A circle or $d$-dimensional sphere is considered a special case of a
$d$-dimensional ellipse, where the eccentricity $e\approx0$ in the limit
$e\rightarrow0$
\citep{Zwillinger1996}%
.

However, we realize that the eccentricity $e$ of any given circle or sphere
\emph{cannot be zero}. Indeed, if $e\approx0$, it follows that%
\begin{align*}
\left\Vert \mathbf{x-}\boldsymbol{\nu}\right\Vert  &  =e\times\left\Vert
\mathbf{x}\right\Vert \cos\theta\\
&  \approx0\text{,}%
\end{align*}
at which point the radius $\mathbf{r}$ of a circle or sphere is zero
$\mathbf{r}\approx0$, since if $\left\Vert \mathbf{x-}\boldsymbol{\nu
}\right\Vert \approx0$, it follows that $\left\Vert \mathbf{r}\right\Vert
\approx0$.

Instead, we realize that the eccentricity $e$ for any given circle or sphere
\emph{varies} with $\left\Vert \mathbf{x}\right\Vert $ \emph{and}
$\arccos\theta$%
\[
e=\frac{\left\Vert \mathbf{r}\right\Vert }{\left\Vert \mathbf{x}\right\Vert
}\arccos\theta\text{,}%
\]
since the length $\left\Vert \mathbf{r}\right\Vert $ of the radius
$\mathbf{r}$ of the circle or sphere is fixed.

We now turn our attention to the fundamental property exhibited by the locus
of any given quadratic curve or surface.

\subsection{The Fundamental Property of Eigenenergy}

Each and every one of the general vector algebra locus equations of
(\ref{Linear Locus}) - (\ref{Spherical Locus}) demonstrate that the
fundamental property exhibited by all of the points that lie on the geometric
locus---of any given quadratic curve or surface in $d$-dimensional Hilbert
space---is the eigenenergy exhibited by the principal eigenaxis of the locus,
such that: $\left(  1\right)  $ the principal eigenaxis of the locus of a
quadratic curve or surface provides an exclusive principal eigen-coordinate
system for the locus of the quadratic curve or surface, so that all of the
points that lie on the locus of the quadratic curve or surface exclusively
reference the principal eigenaxis and also satisfy the eigenenergy exhibited
by the principal eigenaxis; $\left(  2\right)  $ the principal eigenaxis
satisfies the locus of a quadratic curve or surface in terms of its
eigenenergy; and $\left(  3\right)  $ the vector components of the principal
eigenaxis determine the algebraic and geometric structure and the fundamental
property exhibited by the locus of a quadratic curve or surface.

\subsection{Characteristic Locus of a Quadratic Curve or Surface}

Given the conditions expressed by Theorems
\ref{Vector Algebra Equation of Linear Loci Theorem} -
\ref{Vector Algebra Equation of Spherical Loci Theorem}, we realize that the
principal eigenaxis of the geometric locus of any given quadratic curve or
surface in $d$-dimensional Hilbert space---including the principal eigenaxis
of the geometric locus of the decision boundary of any given minimum risk
binary classification system---is the \emph{characteristic locus} of the curve
or surface. Thereby, we realize that any given characteristic locus
\emph{belongs to} and \emph{pre-wires} \emph{important generalizations} that
determine the overall mathematical structure and the fundamental property
exhibited by the geometric locus of a certain quadratic curve or surface. We
previously coined the term \textquotedblleft eigenlocus\textquotedblright\ to
express this relationship.

\subsection{Eigenlocus of a Decision Boundary}

In previous working papers
\citep{Reeves2015resolving}
and
\citep{Reeves2018design}%
, we used the term \textquotedblleft eigenlocus\textquotedblright\ to refer to
the characteristic locus of the geometric locus of the decision boundary of a
minimum risk binary classification system, such that an eigenlocus of a
minimum risk binary classification system \emph{belongs to} and
\emph{pre-wires} \emph{important generalizations} that \emph{determine} the
overall statistical structure and behavior and fundamental properties of
\emph{the system}. In this treatise, we refer to an eigenlocus as
a\ \textquotedblleft geometric locus of a novel principal
eigenaxis.\textquotedblright

By Theorems \ref{Vector Algebra Equation of Linear Loci Theorem} -
\ref{Vector Algebra Equation of Spherical Loci Theorem}, we recognize that the
structure and the fundamental property exhibited by the characteristic
locus---namely the principal eigenaxis---of the geometric locus of \emph{any}
given quadratic curve or surface \emph{regulates} the structure and the
fundamental property exhibited by all of the points that lie on the locus of
the quadratic curve or surface, where the fundamental property is a certain
amount of eigenenergy exhibited by the locus of the principal eigenaxis.

Moreover, by Theorems \ref{Vector Algebra Equation of Linear Loci Theorem} -
\ref{Vector Algebra Equation of Spherical Loci Theorem}, we recognize that the
structure and the fundamental \emph{properties} exhibited by the
\emph{geometric locus} of a \emph{novel principal eigenaxis} of the decision
boundary---of \emph{any} given minimum risk binary classification
system---regulates the structure and the fundamental properties exhibited by
all of the points that lie on the geometric locus of the \emph{decision
boundary} of the system, where the fundamental properties are \emph{certain
amounts of eigenenergies and probabilities of risk }exhibited by the novel
principal eigenaxis---which is structured as a dual locus of likelihood
components and principal eigenaxis components.

So, \emph{how} do we \emph{find} the geometric locus of the \emph{novel
principal eigenaxis}---of a minimum risk binary classification system?

\subsection{Suitable Transformations of Coordinate Systems}

A large number of problems in science and engineering have been resolved by a
\emph{suitable change} of \emph{coordinate system}, at which point a given
problem has a \emph{natural solution}. For example, the Fourier transform
takes a signal and represents the signal in terms of frequencies of
waveforms---so that sine and cosine components make up the signal---which
makes it easy to examine or process certain frequencies of the signal
\citep{allen2004signal}%
.

For certain problems in machine learning and statistics, we consider the bias
and variance dilemma to be a matter of finding a \emph{suitable statistical
representation} of\emph{ }a given \emph{system}---which requires finding a
\emph{suitable statistical representation} for the transformed basis of
\emph{an intrinsic coordinate system} of the system.

It will be seen that the bias and variance dilemma---for the fundamental
problem of finding discriminant functions of minimum risk binary
classification systems, subject to certain random vectors---is resolved by
finding a natural solution that determines a \emph{suitable statistical
representation} for the \emph{transformed basis} of the intrinsic coordinate
system $\mathbf{x}^{T}\mathbf{\Sigma}_{1}^{-1}\mathbf{x-x}^{T}\mathbf{\Sigma
}_{2}^{-1}\mathbf{x}$ in the vector algebra locus equation of
(\ref{Norm_Dec_Bound})\textbf{\textbf{,}} so that the natural solution
transforms the vector algebra locus equation of (\ref{Norm_Dec_Bound}) by a
\emph{suitable transformation} of the \emph{basis} of the intrinsic coordinate
system $\mathbf{x}^{T}\mathbf{\Sigma}_{1}^{-1}\mathbf{x-x}^{T}\mathbf{\Sigma
}_{2}^{-1}\mathbf{x}$.

Thereby, it will be seen that the general problem of the binary classification
of random vectors is a \emph{statistical coordinate transform problem}, so
that the general problem is resolved by a suitable change of the basis of the
intrinsic coordinate system $\mathbf{x}^{T}\mathbf{\Sigma}_{1}^{-1}%
\mathbf{x-x}^{T}\mathbf{\Sigma}_{2}^{-1}\mathbf{x}$.

It will also be seen that the general problem of the binary classification of
random vectors is a \emph{system identification problem}, so that the overall
statistical structure and behavior and properties of a minimum risk binary
classification system are determined by transforming a collection of
observations into a data-driven mathematical model that \emph{represents}
fundamental \emph{aspects} of the \emph{system}.

Returning now to Axiom \ref{Rotation of Intrinsic Coordinate Axes Axiom},
recall that the \emph{form} of an algebraic equation of a locus can be
changed, along with the \emph{coordinates} of all of the points that lie on
the locus---by changing the positions of the coordinate axes of the intrinsic
coordinate system of the locus---which is an inherent part of the algebraic
equation of the locus. Indeed, \emph{any} given locus of a quadratic curve or
surface is \emph{independent} of the \emph{coordinate system} that is used to
describe it---including the locus of the decision boundary of a minimum risk
binary classification system.

We show that we can transform the vector algebra locus equation of
(\ref{Norm_Dec_Bound})%
\begin{align*}
d\left(  \mathbf{x}\right)   &  :\mathbf{x}^{T}\mathbf{\Sigma}_{1}%
^{-1}\mathbf{x}-2\mathbf{x}^{T}\mathbf{\Sigma}_{1}^{-1}\boldsymbol{\mu}%
_{1}+\boldsymbol{\mu}_{1}^{T}\mathbf{\Sigma}_{1}^{-1}\boldsymbol{\mu}_{1}%
-\ln\left(  \left\vert \mathbf{\Sigma}_{1}\right\vert \right) \\
&  -\mathbf{x}^{T}\mathbf{\Sigma}_{2}^{-1}\mathbf{x}+2\mathbf{x}%
^{T}\mathbf{\Sigma}_{2}^{-1}\boldsymbol{\mu}_{2}\mathbf{-}\boldsymbol{\mu}%
_{2}^{T}\mathbf{\Sigma}_{2}^{-1}\boldsymbol{\mu}_{2}+\ln\left(  \left\vert
\mathbf{\Sigma}_{2}\right\vert \right)  =0
\end{align*}
by a suitable change of the basis of the intrinsic coordinate system%
\[
\mathbf{x}^{T}\mathbf{\Sigma}_{1}^{-1}\mathbf{x-x}^{T}\mathbf{\Sigma}_{2}%
^{-1}\mathbf{x}\text{,}%
\]
so that likelihood values and likely locations of extreme points
$\mathbf{x}_{1_{\ast}}$ and $\mathbf{x}_{2_{\ast}}$ \emph{determine} the
\emph{positions} of the \emph{coordinates axes} of the \emph{transformed
basis} of the intrinsic coordinate system $\mathbf{x}^{T}\mathbf{\Sigma}%
_{1}^{-1}\mathbf{x-x}^{T}\mathbf{\Sigma}_{2}^{-1}\mathbf{x}$, at which point
the transformed basis has the form of a locus of signed and scaled extreme
vectors $\mathbf{x}_{1_{\ast}}\mathbf{\sim}p\left(  \mathbf{x};\boldsymbol{\mu
}_{1},\mathbf{\Sigma}_{1}\right)  $ and $\mathbf{x}_{2_{\ast}}\mathbf{\sim
}p\left(  \mathbf{x};\boldsymbol{\mu}_{2},\mathbf{\Sigma}_{2}\right)  $.

\emph{First and foremost}, however, we need to demonstrate \emph{how} to
\emph{represent} the \emph{solution} of the transformed basis, so that the
vector algebra locus equation of (\ref{Norm_Dec_Bound}) is \emph{transformed}
by a \emph{suitable} \emph{change} of the basis of the intrinsic coordinate
system $\mathbf{x}^{T}\mathbf{\Sigma}_{1}^{-1}\mathbf{x-x}^{T}\mathbf{\Sigma
}_{2}^{-1}\mathbf{x}$.

In the next two sections of our treatise, we consider how we might represent
the solution of the transformed basis of the intrinsic coordinate system
$\mathbf{x}^{T}\mathbf{\Sigma}_{1}^{-1}\mathbf{x-x}^{T}\mathbf{\Sigma}%
_{2}^{-1}\mathbf{x}$, so that the transformed basis is formed by a locus of
signed and scaled extreme vectors $\mathbf{x}_{1_{\ast}}\mathbf{\sim}p\left(
\mathbf{x};\boldsymbol{\mu}_{1},\mathbf{\Sigma}_{1}\right)  $ and
$\mathbf{x}_{2_{\ast}}\mathbf{\sim}p\left(  \mathbf{x};\boldsymbol{\mu}%
_{2},\mathbf{\Sigma}_{2}\right)  $.

\section{\label{Section 7}Novel Principal Eigen-coordinate Transforms}

In this section of our treatise, we demonstrate that a \emph{geometric locus}
of a \emph{principal eigenaxis} is the principal part of an equivalent
representation of \emph{any given} quadratic form, such that the principal
eigenaxis is an exclusive principal eigen-coordinate system of the geometric
locus of a certain quadratic curve or surface---that satisfies the locus of
the quadratic curve or surface in terms of its total allowed eigenenergy---so
that the equivalent representation of the quadratic form determines the total
allowed eigenenergy exhibited by all of the components of the principal
eigenaxis of the quadratic curve or surface.

We use these results to develop a novel principal eigen-coordinate transform
algorithm that we use to \emph{find} the geometric locus of the \emph{novel
principal eigenaxis}---of \emph{any} given minimum risk binary classification system.

\subsection{Equivalent Representations of Random Quadratic Forms}

Returning again to the vector algebra locus equation of (\ref{Norm_Dec_Bound}%
), recall that a pair of signed random \emph{quadratic forms} $\mathbf{x}%
^{T}\mathbf{\Sigma}_{1}^{-1}\mathbf{x}$ and $-\mathbf{x}^{T}\mathbf{\Sigma
}_{2}^{-1}\mathbf{x}$ jointly provide dual representation of the discriminant
function and the intrinsic coordinate system---of the geometric locus of the
decision boundary---of any given minimum risk binary classification system,
subject to multivariate normal data.

It will be seen that quadratic forms are inherent parts of vector algebra
locus equations of quadratic curves and surfaces.

It will be also seen that any given quadratic form that is the solution of a
vector algebra locus equation, wherein the graph of the vector algebra locus
equation represents a certain quadratic curve or surface, has an equivalent
representation that is related to the \emph{principal eigenvector} of the
symmetric matrix of the quadratic form, such that the \emph{principal
eigenvector} is symmetrically and equivalently related to the \emph{principal
eigenaxis} of the quadratic \emph{curve or surface}---so that the principal
eigenaxis satisfies the locus of the quadratic curve or surface in terms of
its total allowed eigenenergy.

\subsection{Eigendecompositions of Random Quadratic Forms}

Take any given inverted covariance matrices $\mathbf{\Sigma}_{1}^{-1}$ and
$\mathbf{\Sigma}_{2}^{-1}$ that are correlated with the covariance matrices
$\mathbf{\Sigma}_{1}$ and $\mathbf{\Sigma}_{2}$ in the vector algebra locus
equation of (\ref{Norm_Dec_Bound}), where all of the matrices $\mathbf{\Sigma
}_{1}^{-1}$, $\mathbf{\Sigma}_{2}^{-1}$, $\mathbf{\Sigma}_{1}$ and
$\mathbf{\Sigma}_{2}$ are \emph{symmetric matrices}.

We recognize that the inverted covariance matrices $\mathbf{\Sigma}_{1}^{-1}$
and $\mathbf{\Sigma}_{2}^{-1}$ in the vector algebra locus equation of
(\ref{Norm_Dec_Bound}) are \emph{symmetric} \emph{matrices} of random
\emph{quadratic forms} $\mathbf{x}^{T}\mathbf{\Sigma}_{1}^{-1}\mathbf{x}$ and
$\mathbf{x}^{T}\mathbf{\Sigma}_{2}^{-1}\mathbf{x}$.

We also recognize that the symmetric eigenvalue decomposition theorem
guarantees us that any given covariance matrices $\mathbf{\Sigma}_{1}$ and
$\mathbf{\Sigma}_{2}$---as well as the inverted covariance matrices
$\mathbf{\Sigma}_{1}^{-1}$ and $\mathbf{\Sigma}_{2}^{-1}$ in the vector
algebra locus equation of (\ref{Norm_Dec_Bound})---have an equivalent
representation that is determined by a similarity transformation. Accordingly,
the symmetric eigenvalue decomposition theorem guarantees us that for
\emph{any} given covariance matrix $\mathbf{\Sigma}$---we can find a basis of
eigenvectors that are real and orthogonal---that are part of an equivalent
representation of the covariance matrix $\mathbf{\Sigma}$. Correspondingly,
for \emph{any} given inverted covariance matrix $\mathbf{\Sigma}^{-1}$---we
can find a basis of eigenvectors that are real and orthogonal---that are part
of an equivalent representation of the inverted covariance matrix
$\mathbf{\Sigma}^{-1}$.

We now consider practical uses of the symmetric eigenvalue decomposition
theorem---also known as the spectral theorem.

\subsection{Practical Uses of the Spectral Theorem}

The \emph{spectral theorem} states that for any given symmetric matrix
$\mathbf{Q\in}$ $\Re^{N\times N}$, there are exactly $N$ (possibly not
distinct) eigenvalues, such that all of the eigenvalues are real. Furthermore,
the associated eigenvectors can be chosen so as to form an \emph{orthonormal
basis }%
\citep{strang1986introduction}%
.

Thereby, an equivalent representation of any given symmetric matrix
$\mathbf{Q\in}$ $\Re^{N\times N}$ is determined by the similarity
transformation%
\[
\mathbf{Q=V\Lambda V}^{-1}\text{,}%
\]
where $\mathbf{V}$ is a square $N\times N$ matrix whose $i$th column is a unit
orthogonal eigenvector $\widehat{\mathbf{v}}_{i}$ of $\mathbf{Q}$, and
$\mathbf{\Lambda}$ is diagonal matrix whose diagonal elements are the
corresponding eigenvalues $\mathbf{\Lambda}_{ii}=\lambda_{i}$ of $\mathbf{Q}$
\citep{strang1986introduction}%
.

Correspondingly, an equivalent representation of any given inverted symmetric
matrix $\mathbf{Q}^{-1}\mathbf{\in}$ $\Re^{N\times N}$ is determined by the
similarity transformation%
\[
\mathbf{Q}^{-1}\mathbf{=V\Lambda V}^{-1}\text{,}%
\]
where $\mathbf{V}$ is a square $N\times N$ matrix whose $i$th column is a unit
orthogonal eigenvector $\widehat{\mathbf{v}}_{i}$ of $\mathbf{Q}^{-1}$, and
$\mathbf{\Lambda}$ is diagonal matrix whose diagonal elements are the
corresponding eigenvalues $\mathbf{\Lambda}_{ii}=\lambda_{i}^{-1}$ of
$\mathbf{Q}^{-1}$.

What is more, any given quadratic form $\mathbf{x}^{T}\mathbf{Qx}$ or
$\mathbf{x}^{T}\mathbf{Q}^{-1}\mathbf{x}$ has an equivalent representation
that is guaranteed by the \emph{principal axes theorem}---which uses the
spectral theorem to determine the equivalent representation of the quadratic
form
\citep{Hewson2009,Lay2006}%
.

By way of discovery, we examine how quadratic forms are inherent parts of
vector algebra locus equations of quadratic curves and surfaces.

Moreover, for any given quadratic form that is the solution of a vector
algebra locus equation, wherein the graph of the vector algebra locus equation
represents a certain quadratic curve or surface, we use the \emph{principal
axes theorem} to show that the quadratic form can be transformed into an
\emph{equivalent representation} that is related to the \emph{principal
eigenvector} of the symmetric matrix of the quadratic form, such that the
\emph{principal eigenvector} of the symmetric matrix of the quadratic form is
\textbf{symmetrically} and \textbf{equivalently} related to the
\emph{principal eigenaxis} of the quadratic \emph{curve or surface}, so that
the principal eigenaxis satisfies the locus of the quadratic curve or surface
in terms of its total allowed eigenenergy, at which point the equivalent
representation of the quadratic form determines the total allowed eigenenergy
exhibited by all of the components of the principal eigenaxis of the quadratic
curve or surface.

We now consider how quadratic forms are inherent parts of vector algebra locus
equations of quadratic curves and surfaces.

\subsection{Quadratic Forms in Locus Equations}

It is well known that graphs of the general algebraic locus equation%
\[
ax^{2}+bx+cy^{2}+dy+exy=f
\]
represent conic sections or quadratic curves, such that---for any given conic
section or quadratic curve---the real variables $a$, $b$, $c$, $d$, $e$, and
$f$ satisfy certain fixed values, where $x$ is a scale factor for the standard
basis vector $\mathbf{e}_{1}=\left(  1,0\right)  $, and $y$ is a scale factor
for the standard basis vector $\mathbf{e}_{2}=\left(  0,1\right)  $
\citep{Eisenhart1939,Hewson2009, Nichols1893,Tanner1898}%
.

It has also been shown that graphs of equations that have the form%
\[
\mathbf{x}^{T}\mathbf{Qx}=c
\]
represent \emph{conic sections or quadratic curves}, such that $\mathbf{x}%
^{T}\mathbf{Qx}$ is a quadratic form, $\mathbf{Q}$ is a $2\times2$ symmetric
matrix $\mathbf{Q\in}$ $\Re^{2\times2}$, $c$ is a constant, and $\mathbf{x}$
is a vector $\mathbf{x\in}$ $%
\mathbb{R}
^{2}$ that is written as $\mathbf{x=}%
{\textstyle\sum\nolimits_{i=1}^{2}}
x_{i}\mathbf{e}_{i}$, where $x_{i}$ is a scale factor for a standard basis
vector $\mathbf{e}_{i}$ that belongs to the set $\left\{  \mathbf{e}%
_{1}=\left(  1,0\right)  ,\mathbf{e}_{2}=\left(  0,1\right)  \right\}  $
\citep{Hewson2009,Lay2006}%
.

Thereby, we recognize that any given equation $\mathbf{x}^{T}\mathbf{Qx}=c$,
such that $\mathbf{Q}$ is a symmetric matrix $\mathbf{Q\in}$ $\Re^{2\times2}$
of a certain quadratic form $\mathbf{x}^{T}\mathbf{Qx}$, $\mathbf{x}$ is a
vector $\mathbf{x\in}$ $%
\mathbb{R}
^{2}$, and $c$ is a certain constant, is a vector algebra locus equation that
represents the graph of a certain quadratic curve.

Correspondingly, it has been shown that graphs of equations that have the form%
\[
\mathbf{x}^{T}\mathbf{Qx}=c
\]
represent \emph{quadratic surfaces}, such that $\mathbf{x}^{T}\mathbf{Qx}$ is
a quadratic form, $\mathbf{Q}$ is an $N\times N$ symmetric matrix
$\mathbf{Q\in}$ $\Re^{N\times N}$, $c$ is a constant, and $\mathbf{x}$ is a
vector $\mathbf{x\in}$ $%
\mathbb{R}
^{N}$ that is written as $\mathbf{x=}%
{\textstyle\sum\nolimits_{i=1}^{N}}
x_{i}\mathbf{e}_{i}$, where $x_{i}$ is a scale factor for a standard basis
vector $\mathbf{e}_{i}$ that belongs to the set $\left\{  \mathbf{e}%
_{1}=\left(  1,0,\ldots,0\right)  ,\ldots,\mathbf{e}_{N}=\left(
0,0,\ldots,1\right)  \right\}  $
\citep{Hewson2009,Lay2006}%
.

Thereby, we recognize that any given equation $\mathbf{x}^{T}\mathbf{Qx}=c$,
such that $\mathbf{Q}$ is a symmetric matrix $\mathbf{Q\in}$ $\Re^{N\times N}$
of a certain quadratic form $\mathbf{x}^{T}\mathbf{Qx}$, $\mathbf{x}$ is a
vector $\mathbf{x\in}$ $%
\mathbb{R}
^{N}$, and $c$ is a certain constant, is a vector algebra locus equation that
represents the graph of a certain quadratic surface.

\subsection{New Significance of the Principal Axes Theorem}

We have discovered that the eigenvalues and the orthonormal eigenvectors of
any given symmetric matrix $\mathbf{Q\in}$ $\Re^{2\times2}$ or $\mathbf{Q\in}$
$\Re^{N\times N}$ of a corresponding quadratic form $\mathbf{x}^{T}%
\mathbf{Qx}$---that is the solution of a vector algebra locus equation
$\mathbf{x}^{T}\mathbf{Qx}=c$---have \emph{algebraic} and \emph{geometric
significance} in relation to an equivalent representation of the quadratic
form $\mathbf{x}^{T}\mathbf{Qx}$.

We have determined that the set of orthonormal eigenvectors of any given
symmetric matrix $\mathbf{Q\in}$ $\Re^{2\times2}$ or $\mathbf{Q\in}$
$\Re^{N\times N}$ of a quadratic form $\mathbf{x}^{T}\mathbf{Qx}$ are an
\emph{eigenvector basis} of the principal eigenvector $\mathbf{v}$ of the
symmetric matrix $\mathbf{Q\in}$ $\Re^{2\times2}$ or $\mathbf{Q\in}$
$\Re^{N\times N}$ of a quadratic form $\mathbf{v}^{T}\mathbf{Qv}$, so that the
eigenvalues of the symmetric matrix $\mathbf{Q\in}$ $\Re^{2\times2}$ or
$\mathbf{Q\in}$ $\Re^{N\times N}$ of the quadratic form $\mathbf{v}%
^{T}\mathbf{Qv}$ modulate the eigenenergies exhibited by the components of a
\emph{principal eigenaxis }$\boldsymbol{\nu}$ ---of a certain quadratic curve
or surface---that is symmetrically and equivalently \emph{related} to the
\emph{principal eigenvector} $\mathbf{v}$ of the symmetric matrix
$\mathbf{Q\in}$ $\Re^{2\times2}$ or $\mathbf{Q\in}$ $\Re^{N\times N}$ of the
quadratic form $\mathbf{v}^{T}\mathbf{Qv}$.

We use the principal axes theorem to show that any given \emph{quadratic form}
$\mathbf{x}^{T}\mathbf{Qx}$ that is the solution of a vector algebra locus
equation $\mathbf{x}^{T}\mathbf{Qx}=c$, such that the graph of the vector
algebra locus equation $\mathbf{x}^{T}\mathbf{Qx}=c$ represents a certain
quadratic curve or surface, can be \emph{transformed} into an \emph{equivalent
representation}---that is related to a symmetrical and equivalent
representation of the \emph{principal eigenvector} $\mathbf{v}$ of the
symmetric \emph{matrix} $\mathbf{Q}$ of a quadratic form $\mathbf{v}%
^{T}\mathbf{Qv}$---so that the equivalent representation of the quadratic form
$\mathbf{x}^{T}\mathbf{Qx}$ is determined by the total allowed eigenenergy
$\left\Vert \boldsymbol{\nu}\right\Vert ^{2}$ exhibited by the \emph{principal
eigenaxis} $\boldsymbol{\nu}$ of the quadratic curve or surface, such that the
\emph{total allowed eigenenergy} $\left\Vert \boldsymbol{\nu}\right\Vert ^{2}$
exhibited by the \emph{principal eigenaxis} $\boldsymbol{\nu}$ of the
quadratic curve or surface is \emph{regulated} by the \emph{eigenvalues} of
the symmetric matrix $\mathbf{Q}$ of the quadratic form $\mathbf{v}%
^{T}\mathbf{Qv}$.

\subsection{Similar Systems of Principal Eigen-coordinates}

We now show that the \emph{principal eigenvector }$\mathbf{v}$ of the
symmetric matrix $\mathbf{Q\in}$ $\Re^{2\times2}$ or $\mathbf{Q\in}$
$\Re^{N\times N}$ of any given quadratic form $\mathbf{x}^{T}\mathbf{Qx}$ that
is the solution of a vector algebra locus equation $\mathbf{x}^{T}%
\mathbf{Qx}=c$, such that the graph of the vector algebra locus equation
$\mathbf{x}^{T}\mathbf{Qx}=c$ represents a certain quadratic curve or surface,
is symmetrically and equivalently related to the \emph{principal eigenaxis}
$\boldsymbol{\nu}$ of the quadratic \emph{curve or surface}, so that the
equivalent representation of the quadratic form determines the total allowed
eigenenergy exhibited by the geometric locus of the principal eigenaxis of the
quadratic curve or surface.

Thereby, we show that the \emph{principal eigenvector }$\mathbf{v}$ of the
symmetric matrix $\mathbf{Q\in}$ $\Re^{2\times2}$ or $\mathbf{Q\in}$
$\Re^{N\times N}$ of any given constrained quadratic form $\mathbf{x}%
^{T}\mathbf{Qx}=c$, such that the graph of the vector algebra locus equation
$\mathbf{x}^{T}\mathbf{Qx}=c$ represents a certain quadratic curve or surface,
is symmetrically and equivalently related to the \emph{principal eigenaxis}
$\boldsymbol{\nu}$ of the quadratic curve or surface, so that the principal
eigenvector $\mathbf{v}$ of the symmetric matrix $\mathbf{Q}\in$ $\Re
^{2\times2}$ or $\mathbf{Q}\in$ $\Re^{N\times N}$ and the principal eigenaxis
$\boldsymbol{\nu}$ of the quadratic curve or surface are similar systems of
principal eigen-coordinates related to equivalent representations of quadratic
curves or surfaces.

We first devise similar systems of principal eigen-coordinates that are
related to equivalent representations of quadratic curves.

\subsection{Equivalent Representations of Quadratic Curves}

Take any given quadratic form $\mathbf{x}^{T}\mathbf{Qx}$ that satisfies an
equation that has the form $\mathbf{x}^{T}\mathbf{Qx}=c$, such that the graph
of the equation $\mathbf{x}^{T}\mathbf{Qx}=c$ represents a certain quadratic
curve, where $\mathbf{x=}%
{\textstyle\sum\nolimits_{i=1}^{2}}
x_{i}\mathbf{e}_{i}$ is a vector $\mathbf{x\in}$ $%
\mathbb{R}
^{2}$, $\mathbf{Q}$ is a symmetric matrix $\mathbf{Q\in}$ $\Re^{2\times2}$,
and $c$ is a certain constant.

Let $\boldsymbol{T}$ represent the transformation of a mathematical object.
Using conditions expressed by the principal axes theorem
\citep{Hewson2009,Lay2006}%
, the constrained quadratic form $\mathbf{x}^{T}\mathbf{Qx}=c$ can be
\emph{transformed} into an equivalent representation that is given by a novel
principal eigen-coordinate transform $\boldsymbol{T}\left[  \mathbf{x}%
^{T}\mathbf{Qx}=c\right]  $, so that%
\begin{align*}
\boldsymbol{T}\left[  \mathbf{x}^{T}\mathbf{Qx}=c\right]  \mathbf{\ }  &
=\ \left(
{\textstyle\sum\nolimits_{i=1}^{2}}
x_{i\ast}\widehat{\mathbf{v}}_{i}\right)  ^{T}\mathbf{Q}\left(
{\textstyle\sum\nolimits_{j=1}^{2}}
x_{j\ast}\widehat{\mathbf{v}}_{j}\right) \\
&  =\mathbf{v}^{T}\mathbf{Qv}\\
&  =\left(
{\textstyle\sum\nolimits_{i=1}^{2}}
x_{i\ast}\widehat{\mathbf{v}}_{i}\right)  ^{T}\left(
{\textstyle\sum\nolimits_{j=1}^{2}}
\lambda_{j}x_{j\ast}\widehat{\mathbf{v}}_{j}\right) \\
&  =\lambda_{1}x_{1\ast}^{2}\widehat{\mathbf{v}}_{1}^{2}+\lambda_{2}x_{2\ast
}^{2}\widehat{\mathbf{v}}_{2}^{2}\\
&  =%
{\textstyle\sum\nolimits_{i=1}^{2}}
\lambda_{i}\left\Vert x_{i\ast}\widehat{\mathbf{v}}_{i}\right\Vert
^{2}=c\text{,}%
\end{align*}
where $\lambda_{i}$ is an eigenvalue of the matrix $\mathbf{Q}$, and
$\widehat{\mathbf{v}}_{i}$ is a corresponding unit eigenvector of the matrix
$\mathbf{Q}$, wherein the vector $\mathbf{x=}%
{\textstyle\sum\nolimits_{i=1}^{2}}
x_{i}\mathbf{e}_{i}$ is \emph{transformed} into an \emph{eigenvector}
$\mathbf{v}=%
{\textstyle\sum\nolimits_{i=1}^{2}}
x_{i\ast}\widehat{\mathbf{v}}_{i}$ of the symmetric matrix $\mathbf{Q\in}$
$\Re^{2\times2}$, such that each component $x_{i\ast}\widehat{\mathbf{v}}_{i}$
of the eigenvector $\mathbf{v}$ is a principal axis of the eigenvector
$\mathbf{v}$, and the eigenenergy $\left\Vert \mathbf{v}\right\Vert ^{2}$
exhibited by the eigenvector $\mathbf{v}$ is given by $\left\Vert
\mathbf{v}\right\Vert ^{2}=%
{\textstyle\sum\nolimits_{i=1}^{2}}
\left\Vert x_{i\ast}\widehat{\mathbf{v}}_{i}\right\Vert ^{2}$.

We realize that the eigenvector $\mathbf{v}=x_{1\ast}\widehat{\mathbf{v}}%
_{1}+x_{2\ast}\widehat{\mathbf{v}}_{2}$ is the \emph{principal eigenvector} of
the symmetric matrix $\mathbf{Q\in}$ $\Re^{2\times2}$ of the\ quadratic form
$\mathbf{v}^{T}\mathbf{Qv}$.

Moreover, given conditions expressed by Theorems
\ref{Vector Algebra Equation of Linear Loci Theorem} -
\ref{Vector Algebra Equation of Spherical Loci Theorem}, we also realize that
the \emph{principal eigenvector} $\mathbf{v}=%
{\textstyle\sum\nolimits_{i=1}^{2}}
x_{i\ast}\widehat{\mathbf{v}}_{i}$ of the symmetric matrix $\mathbf{Q\in}$
$\Re^{2\times2}$ of the quadratic form $\mathbf{v}^{T}\mathbf{Qv}$ is
\emph{symmetrically and equivalently} related to an exclusive principal
eigen-coordinate system $\boldsymbol{\nu}$ $\boldsymbol{=}%
{\textstyle\sum\nolimits_{i=1}^{2}}
\sqrt{\lambda_{i}}x_{i\ast}\widehat{\mathbf{v}}_{i}$ of the geometric locus of
the quadratic curve---originally represented by the graph of $\mathbf{x}%
^{T}\mathbf{Qx}=c$, such that the eigenvalues $\lambda_{2}\leq\lambda_{1}$ of
the symmetric matrix $\mathbf{Q}$ of the quadratic form $\mathbf{v}%
^{T}\mathbf{Qv}$ determine scale factors $\sqrt{\lambda_{i}}$ for the
components $\sqrt{\lambda_{i}}x_{i\ast}\widehat{\mathbf{v}}_{i}$ of the
principal eigenaxis $\boldsymbol{\nu}$ $\boldsymbol{=}%
{\textstyle\sum\nolimits_{i=1}^{2}}
\sqrt{\lambda_{i}}x_{i\ast}\widehat{\mathbf{v}}_{i}$ of the geometric locus of
the quadratic curve, so that the principal eigenaxis $\boldsymbol{\nu}%
=\sqrt{\lambda_{1}}x_{1\ast}\widehat{\mathbf{v}}_{1}+\sqrt{\lambda_{2}%
}x_{2\ast}\widehat{\mathbf{v}}_{2}$ of the geometric locus of the quadratic
curve satisfies the geometric locus of the quadratic curve in terms of its
\emph{total allowed eigenenergy}%
\begin{align*}
\left\Vert \boldsymbol{\nu}\right\Vert ^{2}  &  =\lambda_{1}x_{1\ast}%
^{2}\widehat{\mathbf{v}}_{1}^{2}+\lambda_{2}x_{2\ast}^{2}\widehat{\mathbf{v}%
}_{2}^{2}\\
&  =\lambda_{1}\left\Vert x_{1\ast}\widehat{\mathbf{v}}_{1}\right\Vert
^{2}+\lambda_{2}\left\Vert x_{2\ast}\widehat{\mathbf{v}}_{2}\right\Vert ^{2}\\
&  =%
{\textstyle\sum\nolimits_{i=1}^{2}}
\lambda_{i}\left\Vert x_{i\ast}\widehat{\mathbf{v}}_{i}\right\Vert
^{2}\text{.}%
\end{align*}

Thereby, we realize that an equivalent representation of any given constrained
quadratic form $\mathbf{x}^{T}\mathbf{Qx}=c$, such that the graph of the
equation $\mathbf{x}^{T}\mathbf{Qx}=c$ represents a certain quadratic curve,
is given by a novel principal eigen-coordinate transform $\boldsymbol{T}%
\left[  \mathbf{x}^{T}\mathbf{Qx}=c\right]  $%
\begin{align*}
\boldsymbol{T}\left[  \mathbf{x}^{T}\mathbf{Qx}=c\right]   &  =\ \left(
{\textstyle\sum\nolimits_{i=1}^{2}}
x_{i\ast}\widehat{\mathbf{v}}_{i}\right)  ^{T}\mathbf{Q}\left(
{\textstyle\sum\nolimits_{j=1}^{2}}
x_{j\ast}\widehat{\mathbf{v}}_{j}\right) \\
&  =\mathbf{v}^{T}\mathbf{Qv}\\
&  =\left(
{\textstyle\sum\nolimits_{i=1}^{2}}
x_{i\ast}\widehat{\mathbf{v}}_{i}\right)  ^{T}\left(
{\textstyle\sum\nolimits_{j=1}^{2}}
\lambda_{j}x_{j\ast}\widehat{\mathbf{v}}_{j}\right) \\
&  =\lambda_{1}x_{1\ast}^{2}\widehat{\mathbf{v}}_{1}^{2}+\lambda_{2}x_{2\ast
}^{2}\widehat{\mathbf{v}}_{2}^{2}\\
&  =\lambda_{1}\left\Vert x_{1\ast}\widehat{\mathbf{v}}_{1}\right\Vert
^{2}+\lambda_{2}\left\Vert x_{2\ast}\widehat{\mathbf{v}}_{2}\right\Vert ^{2}\\
&  =%
{\textstyle\sum\nolimits_{i=1}^{2}}
\lambda_{i}\left\Vert x_{i\ast}\widehat{\mathbf{v}}_{i}\right\Vert ^{2}\\
&  =\left\Vert \boldsymbol{\nu}\right\Vert ^{2}=c\text{,}%
\end{align*}
wherein the vector $\mathbf{x=}%
{\textstyle\sum\nolimits_{i=1}^{2}}
x_{i}\mathbf{e}_{i}$ is transformed into the \emph{principal eigenvector}
$\mathbf{v}=%
{\textstyle\sum\nolimits_{i=1}^{2}}
x_{i\ast}\widehat{\mathbf{v}}_{i}$ of the symmetric matrix $\mathbf{Q\in}$
$\Re^{2\times2}$ of the quadratic form $\mathbf{v}^{T}\mathbf{Qv}$, so that
the transform $\boldsymbol{T}\left[  \mathbf{x}^{T}\mathbf{Qx}=c\right]  $
determines the\emph{ total allowed eigenenergy} $\left\Vert \boldsymbol{\nu
}\right\Vert ^{2}$ that is \emph{exhibited by} \emph{all} of the components
$\sqrt{\lambda_{i}}x_{i\ast}\widehat{\mathbf{v}}_{i}$ of an exclusive
principal eigen-coordinate system%
\[
\boldsymbol{\nu}=\sqrt{\lambda_{1}}x_{1\ast}\widehat{\mathbf{v}}_{1}%
+\sqrt{\lambda_{2}}x_{2\ast}\widehat{\mathbf{v}}_{2}%
\]
of the geometric locus of the quadratic curve, at which point the geometric
locus of the principal eigenaxis $\boldsymbol{\nu}$ satisfies the geometric
locus of the quadratic curve in terms of its total allowed eigenenergy%
\begin{align*}
\lambda_{1}x_{1\ast}^{2}\widehat{\mathbf{v}}_{1}^{2}+\lambda_{2}x_{2\ast}%
^{2}\widehat{\mathbf{v}}_{2}^{2}  &  =\left\Vert \boldsymbol{\nu}\right\Vert
^{2}\\
\lambda_{1}\left\Vert x_{1\ast}\widehat{\mathbf{v}}_{1}\right\Vert
^{2}+\lambda_{2}\left\Vert x_{2\ast}\widehat{\mathbf{v}}_{2}\right\Vert ^{2}
&  =\left\Vert \boldsymbol{\nu}\right\Vert ^{2}\text{,}%
\end{align*}
such that the eigenenergy $\lambda_{i}\left\Vert x_{i\ast}\widehat{\mathbf{v}%
}_{i}\right\Vert ^{2}$ exhibited by each component $\sqrt{\lambda_{i}}%
x_{i\ast}\widehat{\mathbf{v}}_{i}$ of the \emph{principal eigenaxis}
$\boldsymbol{\nu}$ of the quadratic curve is modulated by an eigenvalue
$\lambda_{i}$ of the symmetric matrix $\mathbf{Q\in}$ $\Re^{2\times2}$ of the
quadratic form $\mathbf{v}^{T}\mathbf{Qv}$.

Thus, we have discovered that the \emph{shape} and the \emph{fundamental
property}---exhibited by the geometric locus of any given quadratic curve that
is represented by a vector algebra locus equation that has the form%
\begin{align*}
\lambda_{1}x_{1\ast}^{2}\widehat{\mathbf{v}}_{1}^{2}+\lambda_{2}x_{2\ast}%
^{2}\widehat{\mathbf{v}}_{2}^{2}  &  =%
{\textstyle\sum\nolimits_{i=1}^{2}}
\lambda_{i}\left\Vert x_{i\ast}\widehat{\mathbf{v}}_{i}\right\Vert ^{2}\\
&  =\left\Vert \boldsymbol{\nu}\right\Vert ^{2}\text{,}%
\end{align*}
so that an exclusive principal eigen-coordinate system $\boldsymbol{\nu}%
=\sqrt{\lambda_{1}}x_{1\ast}\widehat{\mathbf{v}}_{1}+\sqrt{\lambda_{2}%
}x_{2\ast}\widehat{\mathbf{v}}_{2}$ of the quadratic curve satisfies the
geometric locus of the quadratic curve in terms of its total allowed
eigenenergy $\left\Vert \boldsymbol{\nu}\right\Vert ^{2}=%
{\textstyle\sum\nolimits_{i=1}^{2}}
\lambda_{i}\left\Vert x_{i\ast}\widehat{\mathbf{v}}_{i}\right\Vert ^{2}$, are
both \emph{determined} by the \emph{total allowed eigenenergy} $\left\Vert
\boldsymbol{\nu}\right\Vert ^{2}$ exhibited by the \emph{principal eigenaxis}
$\boldsymbol{\nu}$ of the geometric locus of the quadratic curve, such that
the eigenvalues $\lambda_{i}$ of the symmetric matrix $\mathbf{Q\in}$
$\Re^{2\times2}$ of the quadratic form $\mathbf{v}^{T}\mathbf{Qv}$ modulate
the total allowed eigenenergy $\left\Vert \boldsymbol{\nu}\right\Vert ^{2}=%
{\textstyle\sum\nolimits_{i=1}^{2}}
\lambda_{i}\left\Vert x_{i\ast}\widehat{\mathbf{v}}_{i}\right\Vert ^{2}$
exhibited by the principal eigenaxis $\boldsymbol{\nu}$ $\boldsymbol{=}%
{\textstyle\sum\nolimits_{i=1}^{2}}
\sqrt{\lambda_{i}}x_{i\ast}\widehat{\mathbf{v}}_{i}$ of the geometric locus of
the quadratic curve; and the uniform property exhibited by all of the points
that lie on the geometric locus of the quadratic curve is the total allowed
eigenenergy $\left\Vert \boldsymbol{\nu}\right\Vert ^{2}$ exhibited by the
principal eigenaxis $\boldsymbol{\nu}$ of the geometric locus of the quadratic curve.

Next, we devise similar systems of principal eigen-coordinates that are
related to equivalent representations of quadratic surfaces.

\subsection{Equivalent Representations of Quadratic Surfaces}

Take any given quadratic form $\mathbf{x}^{T}\mathbf{Qx}$ that satisfies an
equation that has the form $\mathbf{x}^{T}\mathbf{Qx}=c$, such that the graph
of the equation $\mathbf{x}^{T}\mathbf{Qx}=c$ represents a certain quadratic
surface, where $\mathbf{x=}%
{\textstyle\sum\nolimits_{i=1}^{N}}
x_{i}\widehat{\mathbf{v}}_{i}$ is a vector $\mathbf{x\in}$ $%
\mathbb{R}
^{N}$, $\mathbf{Q}$ is a symmetric matrix $\mathbf{Q\in}$ $\Re^{N\times N}$,
and $c$ is a certain constant.

Using conditions expressed by the principal axes theorem
\citep{Hewson2009,Lay2006}%
, the constrained quadratic form $\mathbf{x}^{T}\mathbf{Qx}=c$ can be
\emph{transformed} into an equivalent representation that is given by a novel
principal eigen-coordinate transform $\boldsymbol{T}\left[  \mathbf{x}%
^{T}\mathbf{Qx}=c\right]  $, so that%
\begin{align*}
\boldsymbol{T}\left[  \mathbf{x}^{T}\mathbf{Qx}=c\right]   &  =\left(
{\textstyle\sum\nolimits_{i=1}^{N}}
x_{i\ast}\widehat{\mathbf{v}}_{i}\right)  ^{T}\mathbf{Q}\left(
{\textstyle\sum\nolimits_{j=1}^{N}}
x_{j\ast}\widehat{\mathbf{v}}_{j}\right) \\
&  =\mathbf{v}^{T}\mathbf{Qv}\\
&  =\left(
{\textstyle\sum\nolimits_{i=1}^{N}}
x_{i\ast}\widehat{\mathbf{v}}_{i}\right)  ^{T}\left(
{\textstyle\sum\nolimits_{j=1}^{N}}
\lambda_{j}x_{j\ast}\widehat{\mathbf{v}}_{j}\right) \\
&  =\lambda_{1}x_{1\ast}^{2}\widehat{\mathbf{v}}_{1}^{2}+\mathbf{\ldots
+\ }\lambda_{N}x_{N\ast}^{2}\widehat{\mathbf{v}}_{N}^{2}\\
&  =%
{\textstyle\sum\nolimits_{i=1}^{N}}
\lambda_{i}\left\Vert x_{i\ast}\widehat{\mathbf{v}}_{i}\right\Vert
^{2}=c\text{,}%
\end{align*}
where $\lambda_{i}$ is an eigenvalue of the matrix $\mathbf{Q}$, and
$\widehat{\mathbf{v}}_{i}$ is a corresponding unit eigenvector of the matrix
$\mathbf{Q}$, wherein the vector $\mathbf{x=}%
{\textstyle\sum\nolimits_{i=1}^{N}}
x_{i}\mathbf{e}_{i}$ is \emph{transformed} into an \emph{eigenvector}
$\mathbf{v}=%
{\textstyle\sum\nolimits_{i=1}^{N}}
x_{i\ast}\widehat{\mathbf{v}}_{i}$ of the symmetric matrix $\mathbf{Q\in}$
$\Re^{N\times N}$, such that each component $x_{i\ast}\widehat{\mathbf{v}}%
_{i}$ of the eigenvector $\mathbf{v}$ is a principal axis of the eigenvector
$\mathbf{v}$, and the eigenenergy $\left\Vert \mathbf{v}\right\Vert ^{2}$
exhibited by the eigenvector $\mathbf{v}$ is given by $\left\Vert
\mathbf{v}\right\Vert ^{2}=%
{\textstyle\sum\nolimits_{i=1}^{N}}
\left\Vert x_{i\ast}\widehat{\mathbf{v}}_{i}\right\Vert ^{2}$.

We realize that the eigenvector $\mathbf{v}=x_{1\ast}\widehat{\mathbf{v}}%
_{1}+\mathbf{\ldots+}x_{N\ast}\widehat{\mathbf{v}}_{N}$ is the \emph{principal
eigenvector} of the symmetric matrix $\mathbf{Q\in}$ $\Re^{N\times N}$ of the
quadratic form $\mathbf{v}^{T}\mathbf{Qv}$.

Moreover, given conditions expressed by Theorems
\ref{Vector Algebra Equation of Linear Loci Theorem} -
\ref{Vector Algebra Equation of Spherical Loci Theorem}, we also realize that
the \emph{principal eigenvector} $\mathbf{v}=%
{\textstyle\sum\nolimits_{i=1}^{N}}
x_{i\ast}\widehat{\mathbf{v}}_{i}$ of the symmetric matrix $\mathbf{Q\in}$
$\Re^{N\times N}$ of the quadratic form $\mathbf{v}^{T}\mathbf{Qv}$ is
\emph{symmetrically and equivalently} related to an exclusive principal
eigen-coordinate system $\boldsymbol{\nu}$ $\boldsymbol{=}%
{\textstyle\sum\nolimits_{i=1}^{N}}
\sqrt{\lambda_{i}}x_{i\ast}\widehat{\mathbf{v}}_{i}$ of the geometric locus of
the quadratic surface---originally represented by the graph of $\mathbf{x}%
^{T}\mathbf{Qx}=c$, such that the eigenvalues $\lambda_{N}\leq\mathbf{\ldots
}\leq\lambda_{1}$ of the symmetric matrix $\mathbf{Q}$ of the quadratic form
$\mathbf{v}^{T}\mathbf{Qv}$ determine scale factors $\sqrt{\lambda_{i}}$ for
the components $\sqrt{\lambda_{i}}x_{i\ast}\widehat{\mathbf{v}}_{i}$ of the
principal eigenaxis $\boldsymbol{\nu}$ $\boldsymbol{=}%
{\textstyle\sum\nolimits_{i=1}^{2}}
\sqrt{\lambda_{i}}x_{i\ast}\widehat{\mathbf{v}}_{i}$ of the geometric locus of
the quadratic surface, so that the principal eigenaxis $\boldsymbol{\nu}%
=\sqrt{\lambda_{1}}x_{1\ast}\widehat{\mathbf{v}}_{1}+\mathbf{\ldots+\ }%
\sqrt{\lambda_{N}}x_{N\ast}\widehat{\mathbf{v}}_{N}$ of the quadratic surface
satisfies the quadratic surface in terms of its \emph{total allowed
eigenenergy}%
\begin{align*}
\left\Vert \boldsymbol{\nu}\right\Vert ^{2}  &  =\lambda_{1}x_{1\ast}%
^{2}\widehat{\mathbf{v}}_{1}^{2}+\mathbf{\ldots+}\lambda_{N}x_{N\ast}%
^{2}\widehat{\mathbf{v}}_{N}^{2}\\
&  =\lambda_{1}\left\Vert x_{1\ast}\widehat{\mathbf{v}}_{1}\right\Vert
^{2}+\mathbf{\ldots+\ }\lambda_{N}\left\Vert x_{N\ast}\widehat{\mathbf{v}}%
_{N}\right\Vert ^{2}\\
&  =%
{\textstyle\sum\nolimits_{i=1}^{N}}
\lambda_{i}\left\Vert x_{i\ast}\widehat{\mathbf{v}}_{i}\right\Vert
^{2}\text{.}%
\end{align*}

Thereby, we realize that an equivalent representation of any given constrained
quadratic form $\mathbf{x}^{T}\mathbf{Qx}=c$, such that the graph of the
equation $\mathbf{x}^{T}\mathbf{Qx}=c$ represents a certain quadratic surface,
is given by a novel principal eigen-coordinate transform $\boldsymbol{T}%
\left[  \mathbf{x}^{T}\mathbf{Qx}=c\right]  $%
\begin{align*}
\boldsymbol{T}\left[  \mathbf{x}^{T}\mathbf{Qx}=c\right]   &  =\left(
{\textstyle\sum\nolimits_{i=1}^{N}}
x_{i\ast}\widehat{\mathbf{v}}_{i}\right)  ^{T}\mathbf{Q}\left(
{\textstyle\sum\nolimits_{j=1}^{N}}
x_{j\ast}\widehat{\mathbf{v}}_{j}\right) \\
&  =\mathbf{v}^{T}\mathbf{Qv}\\
&  =\left(
{\textstyle\sum\nolimits_{i=1}^{N}}
x_{i\ast}\widehat{\mathbf{v}}_{i}\right)  ^{T}\left(
{\textstyle\sum\nolimits_{j=1}^{N}}
\lambda_{j}x_{j\ast}\widehat{\mathbf{v}}_{j}\right) \\
&  =\lambda_{1}x_{1\ast}^{2}\widehat{\mathbf{v}}_{1}^{2}+\mathbf{\ldots
+\ }\lambda_{N}x_{N\ast}^{2}\widehat{\mathbf{v}}_{N}^{2}\\
&  =\lambda_{1}\left\Vert x_{1\ast}\widehat{\mathbf{v}}_{1}\right\Vert
^{2}+\mathbf{\ldots+\ }\lambda_{N}\left\Vert x_{N\ast}\widehat{\mathbf{v}}%
_{N}\right\Vert ^{2}\\
&  =%
{\textstyle\sum\nolimits_{i=1}^{N}}
\lambda_{i}\left\Vert x_{i\ast}\widehat{\mathbf{v}}_{i}\right\Vert ^{2}\\
&  =\left\Vert \boldsymbol{\nu}\right\Vert ^{2}=c\text{,}%
\end{align*}
wherein the vector $\mathbf{x=}%
{\textstyle\sum\nolimits_{i=1}^{N}}
x_{i}\mathbf{e}_{i}$ is transformed into the \emph{principal eigenvector}
$\mathbf{v}=%
{\textstyle\sum\nolimits_{i=1}^{N}}
x_{i\ast}\widehat{\mathbf{v}}_{i}$ of the symmetric matrix $\mathbf{Q\in}$
$\Re^{N\times N}$ of the quadratic form $\mathbf{v}^{T}\mathbf{Qv}$, so that
the transform $\boldsymbol{T}\left[  \mathbf{x}^{T}\mathbf{Qx}=c\right]  $
determines the\emph{ total allowed eigenenergy} $\left\Vert \boldsymbol{\nu
}\right\Vert ^{2}$ that is \emph{exhibited by all} of the components
$\sqrt{\lambda_{i}}x_{i\ast}\widehat{\mathbf{v}}_{i}$ of an exclusive
principal eigen-coordinate system%
\[
\boldsymbol{\nu}=\sqrt{\lambda_{1}}x_{1\ast}\widehat{\mathbf{v}}%
_{1}+\mathbf{\ldots+\ }\sqrt{\lambda_{N}}x_{N\ast}\widehat{\mathbf{v}}_{N}%
\]
of the geometric locus of the quadratic surface, so that the geometric locus
of the principal eigenaxis $\boldsymbol{\nu}$ satisfies the geometric locus of
the quadratic surface in terms of its total allowed eigenenergy%
\begin{align*}
\lambda_{1}x_{1\ast}^{2}\widehat{\mathbf{v}}_{1}^{2}+\mathbf{\ldots+\ }%
\lambda_{N}x_{N\ast}^{2}\widehat{\mathbf{v}}_{N}^{2}  &  =\left\Vert
\boldsymbol{\nu}\right\Vert ^{2}\\
\lambda_{1}\left\Vert x_{1\ast}\widehat{\mathbf{v}}_{1}\right\Vert
^{2}+\mathbf{\ldots+\ }\lambda_{N}\left\Vert x_{N\ast}\widehat{\mathbf{v}}%
_{N}\right\Vert ^{2}  &  =\left\Vert \boldsymbol{\nu}\right\Vert ^{2}\text{,}%
\end{align*}
such that the eigenenergy $\lambda_{i}\left\Vert x_{i\ast}\widehat{\mathbf{v}%
}_{i}\right\Vert ^{2}$ exhibited by each component $\sqrt{\lambda_{i}}%
x_{i\ast}\widehat{\mathbf{v}}_{i}$ of the \emph{principal eigenaxis}
$\boldsymbol{\nu}$ of the quadratic surface is modulated by an eigenvalue
$\lambda_{i}$ of the symmetric matrix $\mathbf{Q\in}$ $\Re^{N\times N}$ of the
quadratic form $\mathbf{v}^{T}\mathbf{Qv}$.

Thus, we have discovered that the \emph{shape} and the \emph{fundamental
property}---exhibited by the geometric locus of any given quadratic surface
that is represented by a vector algebra locus equation that has the form%
\begin{align*}
\lambda_{1}x_{1\ast}^{2}\widehat{\mathbf{v}}_{1}^{2}+\mathbf{\ldots+\ }%
\lambda_{N}x_{N\ast}^{2}\widehat{\mathbf{v}}_{N}^{2}  &  =%
{\textstyle\sum\nolimits_{i=1}^{N}}
\lambda_{i}\left\Vert x_{i\ast}\widehat{\mathbf{v}}_{i}\right\Vert ^{2}\\
&  =\left\Vert \boldsymbol{\nu}\right\Vert ^{2}\text{,}%
\end{align*}
so that an exclusive principal eigen-coordinate system $\boldsymbol{\nu}%
=\sqrt{\lambda_{1}}x_{1\ast}\widehat{\mathbf{v}}_{1}+\mathbf{\ldots+\ }%
\sqrt{\lambda_{N}}x_{N\ast}\widehat{\mathbf{v}}_{N}$ of the quadratic surface
satisfies the geometric locus of the quadratic surface in terms of its total
allowed eigenenergy $\left\Vert \boldsymbol{\nu}\right\Vert ^{2}=%
{\textstyle\sum\nolimits_{i=1}^{N}}
\lambda_{i}\left\Vert x_{i\ast}\widehat{\mathbf{v}}_{i}\right\Vert ^{2}$, are
both \emph{determined} by the \emph{total allowed eigenenergy} $\left\Vert
\boldsymbol{\nu}\right\Vert ^{2}$ exhibited by the \emph{principal eigenaxis}
$\boldsymbol{\nu}$ of the geometric locus of the quadratic surface, such that
the eigenvalues $\lambda_{i}$ of the symmetric matrix $\mathbf{Q\in}$
$\Re^{N\times N}$ of the quadratic form $\mathbf{v}^{T}\mathbf{Qv}%
\boldsymbol{\ }$modulate the total allowed eigenenergy $\left\Vert
\boldsymbol{\nu}\right\Vert ^{2}=%
{\textstyle\sum\nolimits_{i=1}^{N}}
\lambda_{i}\left\Vert x_{i\ast}\widehat{\mathbf{v}}_{i}\right\Vert ^{2}$
exhibited by the principal eigenaxis $\boldsymbol{\nu=}%
{\textstyle\sum\nolimits_{i=1}^{N}}
\sqrt{\lambda_{i}}x_{i\ast}\widehat{\mathbf{v}}_{i}$ of the geometric locus of
the quadratic surface; and the uniform property exhibited by all of the points
that lie on the geometric locus of the quadratic curve is the total allowed
eigenenergy $\left\Vert \boldsymbol{\nu}\right\Vert ^{2}$ exhibited by the
principal eigenaxis $\boldsymbol{\nu}$ of the geometric locus of the quadratic surface.

These discoveries lead us to express an existence theorem which guarantees the
existence of an exclusive principal eigen-coordinate system---which is the
principal part of an equivalent representation of a certain quadratic
form---such that the exclusive principal eigen-coordinate system is the
solution of an equivalent form of the vector algebra locus equation of the
geometric locus of a certain quadratic curve or surface---so that the
principal eigenaxis of the geometric locus of the quadratic curve or surface
satisfies the geometric locus of the quadratic curve or surface in terms of
its \emph{total allowed eigenenergy}.

\subsection{A General Vector Algebra Locus Equation}

We have devised a general vector algebra locus equation for each class of
conic sections and quadratic surfaces, including lines, planes, and
hyperplanes, such that the form of the locus equation is determined by the
principal eigenaxis of the locus of a certain quadratic curve or surface, so
that $\left(  1\right)  $ the principal eigenaxis is the exclusive coordinate
axis of the locus of the quadratic curve or surface; $\left(  2\right)  $ the
principal eigenaxis satisfies the locus of the quadratic curve or surface in
terms of its total allowed eigenenergy; and $\left(  3\right)  $ the uniform
property exhibited by all of the points that lie on the locus of the quadratic
curve or surface is the total allowed eigenenergy exhibited by the principal
eigenaxis of the locus of the quadratic curve or surface.

\subsection{Equivalent Representations of Quadratic Forms}

The general vector algebra locus equation that is outlined above is determined
by a novel principal eigen-coordinate transform method, wherein an exclusive
principal eigen-coordinate system is the solution of an equivalent form of a
vector algebra locus equation that is satisfied by a correlated quadratic
form, at which point the graph of the vector algebra locus equation represents
a certain quadratic curve or surface, such that the \emph{exclusive principal
eigen-coordinate system} is the principal part of an \emph{equivalent
representation} of the \emph{quadratic form} in such a manner that the
exclusive principal eigen-coordinate system is the principal eigenaxis of the
locus of the quadratic curve or surface, so that the principal eigenaxis
satisfies the locus of the quadratic curve or surface in terms of its
\emph{total allowed eigenenergy}.

We express the novel principal eigen-coordinate transform method as an
existence theorem which guarantees the \emph{existence} of the \emph{principal
eigenaxis }of a \emph{certain quadratic curve or surface}---so that the
principal eigenaxis is the solution of an equivalent form of the vector
algebra locus equation of the quadratic curve or surface, such that the
original vector algebra locus equation is satisfied by a certain quadratic form.

\subsection{Existence Theorem of a Principal Eigenaxis}

Theorem \ref{Principal Eigen-coordinate System Theorem} is a significant
result regarding certain mathematical aspects of an exclusive principal
eigen-coordinate system of a certain quadratic curve or surface, so that the
exclusive principal eigen-coordinate system is the solution of an equivalent
form of a vector algebra locus equation of the quadratic curve or surface,
such that a certain quadratic form is the solution of the original vector
algebra locus equation of the quadratic curve or surface.

Theorem \ref{Principal Eigen-coordinate System Theorem} is motivated by
conditions expressed by Lemma
\ref{Equivalent Form of Algebraic Equation Lemma} and Theorems
\ref{Vector Algebra Equation of Linear Loci Theorem} -
\ref{Vector Algebra Equation of Spherical Loci Theorem}, along with conditions
expressed by the spectral theorem and the principal axes theorem.

Most importantly, Theorem \ref{Principal Eigen-coordinate System Theorem} is
an existence theorem that guarantees the existence of an exclusive principal
eigen-coordinate system---which is the principal part of an equivalent
representation of a correlated quadratic form---such that the exclusive
principal eigen-coordinate system is the solution of an equivalent form of the
vector algebra locus equation of the geometric locus of a certain quadratic
curve or surface---so that the principal eigenaxis of the geometric locus of
the quadratic curve or surface satisfies the geometric locus of the quadratic
curve or surface in terms of its \emph{total allowed eigenenergy}.

\begin{theorem}
\label{Principal Eigen-coordinate System Theorem}Take any given vector algebra
locus equation of a quadratic curve that has the form%
\[
\mathbf{x}^{T}\mathbf{Qx}=c\text{,}%
\]
such that $\mathbf{Q}$ is a $2\times2$ symmetric matrix $\mathbf{Q\in}$
$\Re^{2\times2}$ of a certain quadratic form $\mathbf{x}^{T}\mathbf{Qx}$, $c$
is a certain constant, and the vector $\mathbf{x}$ is written as $\mathbf{x=}%
{\textstyle\sum\nolimits_{i=1}^{2}}
x_{i}\mathbf{e}_{i}$, where $x_{i}$ is a scale factor for a standard basis
vector $\mathbf{e}_{i}$ that belongs to the set $\left\{  \mathbf{e}%
_{1}=\left(  1,0\right)  ,\mathbf{e}_{2}=\left(  0,1\right)  \right\}  $.

Let an equivalent form of the vector algebra locus equation $\mathbf{x}%
^{T}\mathbf{Qx}=c$ be generated by transforming the positions of the
coordinate axes of the quadratic curve into the axes of an exclusive principal
eigen-coordinate system, so that the principal eigenaxis of the quadratic
curve satisfies the geometric locus of the quadratic curve in terms of its
total allowed eigenenergy.

Thereby, let the principal eigenaxis of any given quadratic curve be generated
by the following principal eigen-coordinate transform method.

Take any given $2\times2$ symmetric matrix $\mathbf{Q\in}$ $\Re^{2\times2}$ of
a quadratic form $\mathbf{x}^{T}\mathbf{Qx}$ that is the solution of a vector
algebra locus equation $\mathbf{x}^{T}\mathbf{Qx}=c$, such that $c$ is a
certain constant, and $\mathbf{x=}%
{\textstyle\sum\nolimits_{i=1}^{2}}
x_{i}\mathbf{e}_{i}$ is a vector $\mathbf{x\in}$ $%
\mathbb{R}
^{2}$, so that the graph of the vector algebra locus equation $\mathbf{x}%
^{T}\mathbf{Qx}=c$ represents a certain quadratic curve.

Write the vector $\mathbf{x=}%
{\textstyle\sum\nolimits_{i=1}^{2}}
x_{i}\mathbf{e}_{i}$ in terms of a basis of unit eigenvectors $\left\{
\widehat{\mathbf{v}}_{1},\widehat{\mathbf{v}}_{2}\right\}  $ of the matrix
$\mathbf{Q}$, so that%
\begin{align*}
\mathbf{x}  &  =%
{\textstyle\sum\nolimits_{i=1}^{2}}
x_{i}\mathbf{e}_{i}\\
&  \equiv%
{\textstyle\sum\nolimits_{i=1}^{2}}
x_{i\ast}\widehat{\mathbf{v}}_{i}\text{,}%
\end{align*}
at which point the vector $\mathbf{x=}%
{\textstyle\sum\nolimits_{i=1}^{2}}
x_{i}\mathbf{e}_{i}$ is transformed into the principal eigenvector
$\mathbf{v}=%
{\textstyle\sum\nolimits_{i=1}^{2}}
x_{i\ast}\widehat{\mathbf{v}}_{i}$ of the symmetric matrix $\mathbf{Q}$.

Take the eigenvalues $\lambda_{2}\leq\lambda_{1}$ of the $2\times2$ symmetric
matrix $\mathbf{Q\in}$ $\Re^{2\times2}$ and let the principal eigenvector
$\mathbf{v}$ of the symmetric matrix $\mathbf{Q}$ of the quadratic form
$\mathbf{v}^{T}\mathbf{Qv}$ be symmetrically and equivalently related to the
principal eigenaxis $\boldsymbol{\nu}$ of the quadratic curve, so that the
eigenvalues $\lambda_{2}\leq\lambda_{1}$ of the symmetric matrix $\mathbf{Q}$
of the quadratic form $\mathbf{v}^{T}\mathbf{Qv}$ determine scale factors
$\sqrt{\lambda_{i}}$ for the components $\sqrt{\lambda_{i}}x_{i\ast
}\widehat{\mathbf{v}}_{i}$ of an exclusive principal eigen-coordinate system%
\begin{align*}
\boldsymbol{\nu}  &  =\sqrt{\lambda_{1}}x_{1\ast}\widehat{\mathbf{v}}%
_{1}+\sqrt{\lambda_{2}}x_{2\ast}\widehat{\mathbf{v}}_{2}\\
&  =%
{\textstyle\sum\nolimits_{i=1}^{2}}
\sqrt{\lambda_{i}}x_{i\ast}\widehat{\mathbf{v}}_{i}%
\end{align*}
of the quadratic curve.

It follows that the eigenvalues $\lambda_{2}\leq\lambda_{1}$ of the symmetric
matrix $\mathbf{Q}$ of the quadratic form $\mathbf{v}^{T}\mathbf{Qv}$ are
interconnected with the eigenenergies $\lambda_{i}\left\Vert x_{i\ast
}\widehat{\mathbf{v}}_{i}\right\Vert ^{2}$ exhibited by the components
$\sqrt{\lambda_{i}}x_{i\ast}\widehat{\mathbf{v}}_{i}$ of the principal
eigenaxis $\boldsymbol{\nu}$ of the quadratic curve, such that the principal
eigenaxis $\boldsymbol{\nu}$ is the solution of the vector algebra locus
equation%
\begin{align*}%
{\textstyle\sum\nolimits_{i=1}^{2}}
\left(  \sqrt{\lambda_{i}}x_{i\ast}\widehat{\mathbf{v}}_{i}^{T}\right)
\left(  \sqrt{\lambda_{i}}x_{i\ast}\widehat{\mathbf{v}}_{i}\right)   &
=\lambda_{1}x_{1\ast}^{2}\left\Vert \widehat{\mathbf{v}}_{1}\right\Vert
^{2}+\lambda_{2}x_{2\ast}^{2}\left\Vert \widehat{\mathbf{v}}_{2}\right\Vert
^{2}\\
&  =%
{\textstyle\sum\nolimits_{i=1}^{2}}
\lambda_{i}\left\Vert x_{i\ast}\widehat{\mathbf{v}}_{i}\right\Vert ^{2}\\
&  =\left\Vert \boldsymbol{\nu}\right\Vert ^{2}\equiv c\text{,}%
\end{align*}
wherein the constant $c$ in the locus equation $\mathbf{x}^{T}\mathbf{Qx}=c$,
where $\mathbf{x\equiv\ }%
{\textstyle\sum\nolimits_{i=1}^{2}}
x_{i\ast}\widehat{\mathbf{v}}_{i}$, is determined by the total allowed
eigenenergy $\left\Vert \boldsymbol{\nu}\right\Vert ^{2}$ exhibited by the
exclusive principal eigen-coordinate system $\boldsymbol{\nu}$, so that the
eigenenergy $\lambda_{i}\left\Vert x_{1\ast}\widehat{\mathbf{v}}%
_{i}\right\Vert ^{2}$ exhibited by each component $\sqrt{\lambda_{i}}x_{i\ast
}\widehat{\mathbf{v}}_{i}$ of the principal eigenaxis $\boldsymbol{\nu}$ of
the geometric locus of the quadratic curve is modulated by an eigenvalue
$\lambda_{i}$ of the symmetric matrix $\mathbf{Q\in}$ $\Re^{2\times2}$ of the
quadratic form $\mathbf{v}^{T}\mathbf{Qv}$.

Thereby, each component $x_{i}\mathbf{e}_{i}$ of the vector $\mathbf{x}$ is
transformed by an eigenvalue $\lambda_{i}$ and a unit eigenvector
$\widehat{\mathbf{v}}_{i}$ of the symmetric matrix $\mathbf{Q}$ of the
quadratic form $\mathbf{v}^{T}\mathbf{Qv}$, such that the exclusive principal
eigen-coordinate system $\boldsymbol{\nu}=\sqrt{\lambda_{1}}x_{1\ast
}\widehat{\mathbf{v}}_{1}+\sqrt{\lambda_{2}}x_{2\ast}\widehat{\mathbf{v}}_{2}$
is the solution of the equivalent form%
\[
\lambda_{1}\left\Vert x_{1\ast}\widehat{\mathbf{v}}_{1}\right\Vert
^{2}+\lambda_{2}\left\Vert x_{2\ast}\widehat{\mathbf{v}}_{2}\right\Vert
^{2}=\left\Vert \boldsymbol{\nu}\right\Vert ^{2}%
\]
of the vector algebra locus equation $\mathbf{x}^{T}\mathbf{Qx}=c$, where
$\mathbf{x\equiv}%
{\textstyle\sum\nolimits_{i=1}^{2}}
x_{i\ast}\widehat{\mathbf{v}}_{i}$, so that the geometric locus of the
principal eigenaxis $\boldsymbol{\nu=}%
{\textstyle\sum\nolimits_{i=1}^{2}}
\sqrt{\lambda_{i}}x_{i\ast}\widehat{\mathbf{v}}_{i}$ of the geometric locus of
the quadratic curve satisfies the geometric locus of the quadratic curve in
terms of its total allowed eigenenergy%
\[
\left\Vert \boldsymbol{\nu}\right\Vert ^{2}=%
{\textstyle\sum\nolimits_{i=1}^{2}}
\lambda_{i}\left\Vert x_{i\ast}\widehat{\mathbf{v}}_{i}\right\Vert
^{2}\text{,}%
\]
such that the eigenenergy $\lambda_{i}\left\Vert x_{i\ast}\widehat{\mathbf{v}%
}_{i}\right\Vert ^{2}$ exhibited by each component $\sqrt{\lambda_{i}}%
x_{i\ast}\widehat{\mathbf{v}}_{i}$ of the principal eigenaxis $\boldsymbol{\nu
=}%
{\textstyle\sum\nolimits_{i=1}^{2}}
\sqrt{\lambda_{i}}x_{i\ast}\widehat{\mathbf{v}}_{i}$ is modulated by an
eigenvalue $\lambda_{i}$ of the symmetric matrix $\mathbf{Q\in}$ $\Re
^{2\times2}$ of the quadratic form $\mathbf{v}^{T}\mathbf{Qv}$; and the
uniform property exhibited by all of the points that lie on the geometric
locus of the quadratic curve is the total allowed eigenenergy $\left\Vert
\boldsymbol{\nu}\right\Vert ^{2}=%
{\textstyle\sum\nolimits_{i=1}^{2}}
\lambda_{i}\left\Vert x_{i\ast}\widehat{\mathbf{v}}_{i}\right\Vert ^{2}$
exhibited by the principal eigenaxis $\boldsymbol{\nu}$ of the geometric locus
of the quadratic curve.

It follows that the exclusive principal eigen-coordinate system%
\[
\boldsymbol{\nu=}%
{\textstyle\sum\nolimits_{i=1}^{2}}
\sqrt{\lambda_{i}}x_{i\ast}\widehat{\mathbf{v}}_{i}%
\]
is the principal part of an equivalent representation of the quadratic form
$\mathbf{x}^{T}\mathbf{Qx}$ in the following manner%
\begin{align*}
\left(
{\textstyle\sum\nolimits_{i=1}^{2}}
x_{i\ast}\widehat{\mathbf{v}}_{i}\right)  ^{T}\mathbf{Q}\left(
{\textstyle\sum\nolimits_{j=1}^{2}}
x_{j\ast}\widehat{\mathbf{v}}_{j}\right)   &  =\mathbf{v}^{T}\mathbf{Qv}\\
&  =\left(
{\textstyle\sum\nolimits_{i=1}^{2}}
x_{i\ast}\widehat{\mathbf{v}}_{i}\right)  ^{T}\left(
{\textstyle\sum\nolimits_{j=1}^{2}}
\lambda_{j}x_{j\ast}\widehat{\mathbf{v}}_{j}\right) \\
&  =%
{\textstyle\sum\nolimits_{i=1}^{2}}
\left(  \sqrt{\lambda_{i}}x_{i\ast}\widehat{\mathbf{v}}_{i}^{T}\right)
\left(  \sqrt{\lambda_{i}}x_{i\ast}\widehat{\mathbf{v}}_{i}\right) \\
&  =%
{\textstyle\sum\nolimits_{i=1}^{2}}
\lambda_{i}\left\Vert x_{i\ast}\widehat{\mathbf{v}}_{i}\right\Vert ^{2}\\
&  =\left\Vert \boldsymbol{\nu}\right\Vert ^{2}\text{,}%
\end{align*}
wherein the vector $\mathbf{x=}%
{\textstyle\sum\nolimits_{i=1}^{2}}
x_{i}\mathbf{e}_{i}$ is transformed into the principal eigenvector
$\mathbf{v}=%
{\textstyle\sum\nolimits_{i=1}^{2}}
x_{i\ast}\widehat{\mathbf{v}}_{i}$ of the symmetric matrix $\mathbf{Q\in}$
$\Re^{2\times2}$ of the quadratic form $\mathbf{v}^{T}\mathbf{Qv}$, such that
the principal eigenvector $\mathbf{v}=%
{\textstyle\sum\nolimits_{i=1}^{2}}
x_{i\ast}\widehat{\mathbf{v}}_{i}$ of the symmetric matrix $\mathbf{Q}$ of the
quadratic form $\mathbf{v}^{T}\mathbf{Qv}$ is symmetrically and equivalently
related to the principal eigenaxis $\boldsymbol{\nu=}%
{\textstyle\sum\nolimits_{i=1}^{2}}
\sqrt{\lambda_{i}}x_{i\ast}\widehat{\mathbf{v}}_{i}$ of a certain quadratic
curve, so that the exclusive principal eigen-coordinate system
$\boldsymbol{\nu=}%
{\textstyle\sum\nolimits_{i=1}^{2}}
\sqrt{\lambda_{i}}x_{i\ast}\widehat{\mathbf{v}}_{i}$ of the quadratic curve
satisfies the geometric locus of the quadratic curve in terms of its in terms
of its total allowed eigenenergy $\left\Vert \boldsymbol{\nu}\right\Vert ^{2}=%
{\textstyle\sum\nolimits_{i=1}^{2}}
\lambda_{i}\left\Vert x_{i\ast}\widehat{\mathbf{v}}_{i}\right\Vert ^{2}$, such
that the eigenenergy $\lambda_{i}\left\Vert x_{i\ast}\widehat{\mathbf{v}}%
_{i}\right\Vert ^{2}$ exhibited by each component $\sqrt{\lambda_{i}}x_{i\ast
}\widehat{\mathbf{v}}_{i}$ of the principal eigenaxis $\boldsymbol{\nu}%
=\sqrt{\lambda_{1}}x_{1\ast}\widehat{\mathbf{v}}_{1}+\sqrt{\lambda_{2}%
}x_{2\ast}\widehat{\mathbf{v}}_{2}$ of the geometric locus of the quadratic
curve is modulated by an eigenvalue $\lambda_{i}$ of the symmetric matrix
$\mathbf{Q}$ of the quadratic form $\mathbf{v}^{T}\mathbf{Qv}$ in a manner
that regulates the total allowed eigenenergy $\left\Vert \boldsymbol{\nu
}\right\Vert ^{2}=%
{\textstyle\sum\nolimits_{i=1}^{2}}
\lambda_{i}\left\Vert x_{i\ast}\widehat{\mathbf{v}}_{i}\right\Vert ^{2}$
exhibited by the geometric locus of the principal eigenaxis $\boldsymbol{\nu=}%
{\textstyle\sum\nolimits_{i=1}^{2}}
\sqrt{\lambda_{i}}x_{i\ast}\widehat{\mathbf{v}}_{i}$, wherein the $2\times2$
symmetric matrix $\mathbf{Q\in}$ $\Re^{2\times2}$ has the simple diagonal form%
\[
\mathbf{Q}_{ij}=\left\{
\begin{array}
[c]{c}%
0\text{ if }i\neq j\\
\lambda_{i}\text{ if }i=j
\end{array}
\right.  \text{.}%
\]

Thereby, the shape and the fundamental property exhibited by the geometric
locus of any given quadratic curve that is represented by a vector algebra
locus equation that has the form%
\begin{align*}
\lambda_{1}x_{1\ast}^{2}\widehat{\mathbf{v}}_{1}^{2}+\lambda_{2}x_{2\ast}%
^{2}\widehat{\mathbf{v}}_{2}^{2}  &  =%
{\textstyle\sum\nolimits_{i=1}^{2}}
\lambda_{i}x_{i\ast}^{2}\left\Vert \widehat{\mathbf{v}}_{i}\right\Vert ^{2}\\
&  =%
{\textstyle\sum\nolimits_{i=1}^{2}}
\lambda_{i}\left\Vert x_{i\ast}\widehat{\mathbf{v}}_{i}\right\Vert ^{2}\\
&  =\left\Vert \boldsymbol{\nu}\right\Vert ^{2}\text{,}%
\end{align*}
so that an exclusive principal eigen-coordinate system%
\begin{align*}
\boldsymbol{\nu}  &  =\sqrt{\lambda_{1}}x_{1\ast}\widehat{\mathbf{v}}%
_{1}+\sqrt{\lambda_{2}}x_{2\ast}\widehat{\mathbf{v}}_{2}\\
&  =%
{\textstyle\sum\nolimits_{i=1}^{2}}
\sqrt{\lambda_{i}}x_{i\ast}\widehat{\mathbf{v}}_{i}%
\end{align*}
of the quadratic curve satisfies the geometric locus of the quadratic curve in
terms of its total allowed eigenenergy $\left\Vert \boldsymbol{\nu}\right\Vert
^{2}=%
{\textstyle\sum\nolimits_{i=1}^{2}}
\lambda_{i}\left\Vert x_{i\ast}\widehat{\mathbf{v}}_{i}\right\Vert ^{2}$, are
both determined by the total allowed eigenenergy $\left\Vert \boldsymbol{\nu
}\right\Vert ^{2}=%
{\textstyle\sum\nolimits_{i=1}^{2}}
\lambda_{i}\left\Vert x_{i\ast}\widehat{\mathbf{v}}_{i}\right\Vert ^{2}$
exhibited by the principal eigenaxis $\boldsymbol{\nu}$ $\boldsymbol{=}%
{\textstyle\sum\nolimits_{i=1}^{2}}
\sqrt{\lambda_{i}}x_{i\ast}\widehat{\mathbf{v}}_{i}$ of the geometric locus of
the quadratic curve, such that the eigenvalues $\lambda_{i}$ of a $2\times2$
symmetric matrix $\mathbf{Q\in}$ $\Re^{2\times2}$ of a quadratic form
$\mathbf{v}^{T}\mathbf{Qv}$ regulate the total allowed eigenenergy $\left\Vert
\boldsymbol{\nu}\right\Vert ^{2}$ exhibited by the principal eigenaxis
$\boldsymbol{\nu}$ of the geometric locus of the quadratic curve; and the
uniform property exhibited by all of the points that lie on the geometric
locus of the quadratic curve is the total allowed eigenenergy $\left\Vert
\boldsymbol{\nu}\right\Vert ^{2}$ exhibited by the principal eigenaxis
$\boldsymbol{\nu}$ of the geometric locus of the quadratic curve.

Correspondingly, take any given vector algebra locus equation of a quadratic
surface that has the form%
\[
\mathbf{x}^{T}\mathbf{Qx}=c\text{,}%
\]
such that $\mathbf{Q}$ is an $N\times N$ symmetric matrix $\mathbf{Q\in}$
$\Re^{N\times N}$ of a certain quadratic form $\mathbf{x}^{T}\mathbf{Qx}$, $c$
is a certain constant, and the vector $\mathbf{x}$ is written as $\mathbf{x=}%
{\textstyle\sum\nolimits_{i=1}^{N}}
x_{i}\mathbf{e}_{i}$, where $x_{i}$ is a scale factor for a standard basis
vector $\mathbf{e}_{i}$ that belongs to the set%
\[
\left\{  \mathbf{e}_{1}=\left(  1,0,\ldots,0\right)  ,\ldots,\mathbf{e}%
_{N}=\left(  0,0,\ldots,1\right)  \right\}  \text{.}%
\]

Let an equivalent form of the vector algebra locus equation $\mathbf{x}%
^{T}\mathbf{Qx}=c$ be generated by transforming the positions of the
coordinate axes of the quadratic surface into the axes of an exclusive
principal eigen-coordinate system, so that the principal eigenaxis of the
quadratic surface satisfies the geometric locus of the quadratic surface in
terms of its total allowed eigenenergy.

Thereby, let the principal eigenaxis of any given quadratic surface be
generated by the following principal eigen-coordinate transform method.

Take any given $N\times N$ symmetric matrix $\mathbf{Q\in}$ $\Re^{N\times N}$
of a quadratic form $\mathbf{x}^{T}\mathbf{Qx}$ that is the solution of a
vector algebra locus equation $\mathbf{x}^{T}\mathbf{Qx}=c$, such that $c$ is
a certain constant, and $\mathbf{x=}%
{\textstyle\sum\nolimits_{i=1}^{N}}
x_{i}\mathbf{e}_{i}$ is a vector $\mathbf{x\in}$ $%
\mathbb{R}
^{N}$, so that the graph of the vector algebra locus equation $\mathbf{x}%
^{T}\mathbf{Qx}=c$ represents a certain quadratic surface.

Write the vector $\mathbf{x=}%
{\textstyle\sum\nolimits_{i=1}^{N}}
x_{i}\mathbf{e}_{i}$ in terms of a basis of unit eigenvectors $\left\{
\widehat{\mathbf{v}}_{1},\ldots\widehat{\mathbf{v}}_{N}\right\}  $ of the
matrix $\mathbf{Q}$, so that%
\begin{align*}
\mathbf{x}  &  \mathbf{=}%
{\textstyle\sum\nolimits_{i=1}^{N}}
x_{i}\mathbf{e}_{i}\\
&  \mathbf{\equiv}%
{\textstyle\sum\nolimits_{i=1}^{N}}
x_{i\ast}\widehat{\mathbf{v}}_{i}\text{,}%
\end{align*}
at which point the vector $\mathbf{x=}%
{\textstyle\sum\nolimits_{i=1}^{N}}
x_{i}\mathbf{e}_{i}$ is transformed into the principal eigenvector
$\mathbf{v}=%
{\textstyle\sum\nolimits_{i=1}^{N}}
x_{i\ast}\widehat{\mathbf{v}}_{i}$ of the symmetric matrix $\mathbf{Q}$.

Take the eigenvalues $\lambda_{N}\leq\mathbf{\ldots}\leq\lambda_{1}$ of the
$N\times N$ symmetric matrix $\mathbf{Q\in}$ $\Re^{N\times N}$ and let the
principal eigenvector $\mathbf{v}$ of the symmetric matrix $\mathbf{Q}$ of the
quadratic form $\mathbf{v}^{T}\mathbf{Qv}$ be symmetrically and equivalently
related to the principal eigenaxis $\boldsymbol{\nu}$ of the quadratic
surface, so that the eigenvalues $\lambda_{N}\leq\mathbf{\ldots}\leq
\lambda_{1}$ of the symmetric matrix $\mathbf{Q}$ of the quadratic form
$\mathbf{v}^{T}\mathbf{Qv}$ determine scale factors $\sqrt{\lambda_{i}}$ for
the components $\sqrt{\lambda_{i}}x_{i\ast}\widehat{\mathbf{v}}_{i}$ of an
exclusive principal eigen-coordinate system%
\begin{align*}
\boldsymbol{\nu}  &  =\sqrt{\lambda_{1}}x_{1\ast}\widehat{\mathbf{v}}%
_{1}+\mathbf{\ldots+}\sqrt{\lambda_{N}}x_{N\ast}\widehat{\mathbf{v}}_{N}\\
&  =%
{\textstyle\sum\nolimits_{i=1}^{N}}
\sqrt{\lambda_{i}}x_{i\ast}\widehat{\mathbf{v}}_{i}%
\end{align*}
of the quadratic surface.

It follows that the eigenvalues $\lambda_{N}\leq\mathbf{\ldots}\leq\lambda
_{1}$ of the symmetric matrix $\mathbf{Q}$ of the quadratic form
$\mathbf{v}^{T}\mathbf{Qv}$ are interconnected with the eigenenergies
$\lambda_{i}\left\Vert x_{i\ast}\widehat{\mathbf{v}}_{i}\right\Vert ^{2}$
exhibited by the components $\sqrt{\lambda_{i}}x_{i\ast}\widehat{\mathbf{v}%
}_{i}$ of the principal eigenaxis $\boldsymbol{\nu}$ of the quadratic surface,
such that the principal eigenaxis $\boldsymbol{\nu}$ is the solution of the
vector algebra locus equation%
\begin{align*}%
{\textstyle\sum\nolimits_{i=1}^{N}}
\left(  \sqrt{\lambda_{i}}x_{i\ast}\widehat{\mathbf{v}}_{i}^{T}\right)
\left(  \sqrt{\lambda_{i}}x_{i\ast}\widehat{\mathbf{v}}_{i}\right)   &
=\lambda_{1}x_{1\ast}^{2}\widehat{\mathbf{v}}_{1}^{2}+\mathbf{\ldots+}%
\lambda_{N}x_{N\ast}^{2}\widehat{\mathbf{v}}_{N}^{2}\\
&  =%
{\textstyle\sum\nolimits_{i=1}^{N}}
\lambda_{i}\left\Vert x_{i\ast}\widehat{\mathbf{v}}_{i}\right\Vert ^{2}\\
&  =\left\Vert \boldsymbol{\nu}\right\Vert ^{2}\equiv c\text{,}%
\end{align*}
wherein the constant $c$ in the locus equation $\mathbf{x}^{T}\mathbf{Qx}=c$,
where $\mathbf{x\equiv}%
{\textstyle\sum\nolimits_{i=1}^{N}}
x_{i\ast}\widehat{\mathbf{v}}_{i}$, is determined by the total allowed
eigenenergy $\left\Vert \boldsymbol{\nu}\right\Vert ^{2}$ exhibited by the
exclusive principal eigen-coordinate system $\boldsymbol{\nu}$, so that the
eigenenergy $\lambda_{i}\left\Vert x_{i\ast}\widehat{\mathbf{v}}%
_{i}\right\Vert ^{2}$ exhibited by each component $\sqrt{\lambda_{i}}x_{i\ast
}\widehat{\mathbf{v}}_{i}$ of the principal eigenaxis $\boldsymbol{\nu}$ of
the geometric locus of the quadratic surface is modulated by an eigenvalue
$\lambda_{i}$ of the symmetric matrix $\mathbf{Q\in}$ $\Re^{N\times N}$ of the
quadratic form $\mathbf{v}^{T}\mathbf{Qv}$.

Thereby, each component $x_{i}\mathbf{e}_{i}$ of the vector $\mathbf{x}$ is
transformed by an eigenvalue $\lambda_{i}$ and a unit eigenvector
$\widehat{\mathbf{v}}_{i}$ of the symmetric matrix $\mathbf{Q}$ of the
quadratic form $\mathbf{v}^{T}\mathbf{Qv}$, such that the exclusive principal
eigen-coordinate system $\boldsymbol{\nu}=\sqrt{\lambda_{1}}x_{1\ast
}\widehat{\mathbf{v}}_{1}+\mathbf{\ldots+}\sqrt{\lambda_{N}}x_{N\ast
}\widehat{\mathbf{v}}_{N}$ is the solution of the equivalent form%
\[
\lambda_{1}x_{1\ast}^{2}\widehat{\mathbf{v}}_{1}^{2}+\mathbf{\ldots+}%
\lambda_{N}x_{N\ast}^{2}\widehat{\mathbf{v}}_{N}^{2}=\left\Vert
\boldsymbol{\nu}\right\Vert ^{2}%
\]
of the vector algebra locus equation $\mathbf{x}^{T}\mathbf{Qx}=c$, where
$\mathbf{x\equiv}%
{\textstyle\sum\nolimits_{i=1}^{N}}
x_{i\ast}\widehat{\mathbf{v}}_{i}$, so that the geometric locus of the
principal eigenaxis $\boldsymbol{\nu=}%
{\textstyle\sum\nolimits_{i=1}^{N}}
\sqrt{\lambda_{i}}x_{1\ast}\widehat{\mathbf{v}}_{i}$ of the geometric locus of
the quadratic surface satisfies the geometric locus of the quadratic surface
in terms of its total allowed eigenenergy%
\[
\left\Vert \boldsymbol{\nu}\right\Vert ^{2}=%
{\textstyle\sum\nolimits_{i=1}^{N}}
\lambda_{i}\left\Vert x_{i\ast}\widehat{\mathbf{v}}_{i}\right\Vert
^{2}\text{,}%
\]
such that the eigenenergy $\lambda_{i}\left\Vert x_{i\ast}\widehat{\mathbf{v}%
}_{i}\right\Vert ^{2}$ exhibited by each component $\sqrt{\lambda_{i}}%
x_{i\ast}\widehat{\mathbf{v}}_{i}$ of the principal eigenaxis $\boldsymbol{\nu
=}%
{\textstyle\sum\nolimits_{i=1}^{N}}
\sqrt{\lambda_{i}}x_{1\ast}\widehat{\mathbf{v}}_{i}$ of the quadratic surface
is modulated by an eigenvalue $\lambda_{i}$ of the symmetric matrix
$\mathbf{Q\in}$ $\Re^{N\times N}$ of the quadratic form $\mathbf{v}%
^{T}\mathbf{Qv}$; and the uniform property exhibited by all of the points that
lie on the geometric locus of the quadratic surface is the total allowed
eigenenergy $\left\Vert \boldsymbol{\nu}\right\Vert ^{2}=%
{\textstyle\sum\nolimits_{i=1}^{N}}
\lambda_{i}\left\Vert x_{i\ast}\widehat{\mathbf{v}}_{i}\right\Vert ^{2}$
exhibited by the principal eigenaxis $\boldsymbol{\nu}$ of the geometric locus
of the quadratic surface.

It follows that the exclusive principal eigen-coordinate system%
\[
\boldsymbol{\nu}=%
{\textstyle\sum\nolimits_{i=1}^{N}}
\sqrt{\lambda_{i}}x_{i\ast}\widehat{\mathbf{v}}_{i}%
\]
is the principal part of an equivalent representation of the quadratic form
$\mathbf{x}^{T}\mathbf{Qx}$ in the following manner%
\begin{align*}
\left(
{\textstyle\sum\nolimits_{i=1}^{N}}
x_{i\ast}\widehat{\mathbf{v}}_{i}\right)  ^{T}\mathbf{Q}\left(
{\textstyle\sum\nolimits_{j=1}^{N}}
x_{j\ast}\widehat{\mathbf{v}}_{j}\right)   &  =\mathbf{v}^{T}\mathbf{Qv}\\
&  =\left(
{\textstyle\sum\nolimits_{i=1}^{N}}
x_{i\ast}\widehat{\mathbf{v}}_{i}\right)  ^{T}\left(
{\textstyle\sum\nolimits_{j=1}^{N}}
\lambda_{j}x_{j\ast}\widehat{\mathbf{v}}_{j}\right) \\
&  =%
{\textstyle\sum\nolimits_{i=1}^{N}}
\left(  \sqrt{\lambda_{i}}x_{i\ast}\widehat{\mathbf{v}}_{i}\right)
^{T}\left(  \sqrt{\lambda_{i}}x_{i\ast}\widehat{\mathbf{v}}_{i}\right) \\
&  =%
{\textstyle\sum\nolimits_{i=1}^{N}}
\lambda_{i}x_{i\ast}^{2}\left\Vert \widehat{\mathbf{v}}_{i}\right\Vert ^{2}=%
{\textstyle\sum\nolimits_{i=1}^{N}}
\lambda_{i}\left\Vert x_{i\ast}\widehat{\mathbf{v}}_{i}\right\Vert ^{2}\\
&  =\left\Vert \boldsymbol{\nu}\right\Vert ^{2}\text{,}%
\end{align*}
wherein the vector $\mathbf{x=}%
{\textstyle\sum\nolimits_{i=1}^{N}}
x_{i}\mathbf{e}_{i}$ is transformed into the principal eigenvector
$\mathbf{v}=%
{\textstyle\sum\nolimits_{i=1}^{N}}
x_{i\ast}\widehat{\mathbf{v}}_{i}$ of the symmetric matrix $\mathbf{Q\in}$
$\Re^{N\times N}$ of the quadratic form $\mathbf{v}^{T}\mathbf{Qv}$, such that
the principal eigenvector $\mathbf{v}=%
{\textstyle\sum\nolimits_{i=1}^{N}}
x_{i\ast}\widehat{\mathbf{v}}_{i}$ of the symmetric matrix $\mathbf{Q}$ of the
quadratic form $\mathbf{v}^{T}\mathbf{Qv}$ is symmetrically and equivalently
related to the principal eigenaxis $\boldsymbol{\nu=}%
{\textstyle\sum\nolimits_{i=1}^{N}}
\sqrt{\lambda_{i}}x_{i\ast}\widehat{\mathbf{v}}_{i}$ of a certain quadratic
surface, so that the exclusive principal eigen-coordinate system
$\boldsymbol{\nu=}%
{\textstyle\sum\nolimits_{i=1}^{N}}
\sqrt{\lambda_{i}}x_{i\ast}\widehat{\mathbf{v}}_{i}$ of the quadratic surface
satisfies the geometric locus of the quadratic surface in terms of its in
terms of its total allowed eigenenergy $\left\Vert \boldsymbol{\nu}\right\Vert
^{2}=%
{\textstyle\sum\nolimits_{i=1}^{N}}
\lambda_{i}\left\Vert x_{i\ast}\widehat{\mathbf{v}}_{i}\right\Vert ^{2}$, such
that the eigenenergy $\lambda_{i}\left\Vert x_{i\ast}\widehat{\mathbf{v}}%
_{i}\right\Vert ^{2}$ exhibited by each component $\sqrt{\lambda_{i}}x_{i\ast
}\widehat{\mathbf{v}}_{i}$ of the principal eigenaxis $\boldsymbol{\nu}%
=\sqrt{\lambda_{1}}x_{1\ast}\widehat{\mathbf{v}}_{1}+\mathbf{\ldots+}%
\sqrt{\lambda_{N}}x_{N\ast}\widehat{\mathbf{v}}_{N}$ of the geometric locus of
the quadratic surface is modulated by an eigenvalue $\lambda_{i}$ of the
symmetric matrix $\mathbf{Q}$ of the quadratic form $\mathbf{v}^{T}%
\mathbf{Qv}$ in a manner that regulates the total allowed eigenenergy
$\left\Vert \boldsymbol{\nu}\right\Vert ^{2}=%
{\textstyle\sum\nolimits_{i=1}^{N}}
\lambda_{i}\left\Vert x_{i\ast}\widehat{\mathbf{v}}_{i}\right\Vert ^{2}$
exhibited by the geometric locus of the principal eigenaxis $\boldsymbol{\nu=}%
{\textstyle\sum\nolimits_{i=1}^{N}}
\sqrt{\lambda_{i}}x_{i\ast}\widehat{\mathbf{v}}_{i}$, wherein the $N\times N$
symmetric matrix $\mathbf{Q\in}$ $\Re^{N\times N}$ has the simple diagonal
form%
\[
\mathbf{Q}_{ij}=\left\{
\begin{array}
[c]{c}%
0\text{ if }i\neq j\\
\lambda_{i}\text{ if }i=j
\end{array}
\right.  \text{.}%
\]

Thereby, the shape and the fundamental property exhibited by the geometric
locus of any given quadratic surface that is represented by a vector algebra
locus equation that has the form%
\begin{align*}
\lambda_{1}x_{1\ast}^{2}\widehat{\mathbf{v}}_{1}^{2}+\mathbf{\ldots+}%
\lambda_{N}x_{N\ast}^{2}\widehat{\mathbf{v}}_{N}^{2}  &  =%
{\textstyle\sum\nolimits_{i=1}^{N}}
x_{i\ast}^{2}\lambda_{i}\left\Vert \widehat{\mathbf{v}}_{i}\right\Vert ^{2}\\
&  =%
{\textstyle\sum\nolimits_{i=1}^{N}}
\lambda_{i}\left\Vert x_{i\ast}\widehat{\mathbf{v}}_{i}\right\Vert ^{2}\\
&  =\left\Vert \boldsymbol{\nu}\right\Vert ^{2}\text{,}%
\end{align*}
so that an exclusive principal eigen-coordinate system%
\begin{align*}
\boldsymbol{\nu}  &  =\sqrt{\lambda_{1}}x_{1\ast}\widehat{\mathbf{v}}%
_{1}+\mathbf{\ldots+}\sqrt{\lambda_{N}}x_{N\ast}\widehat{\mathbf{v}}_{N}\\
&  =%
{\textstyle\sum\nolimits_{i=1}^{N}}
\sqrt{\lambda_{i}}x_{i\ast}\widehat{\mathbf{v}}_{i}%
\end{align*}
of the quadratic surface satisfies the geometric locus of the quadratic
surface in terms of its total allowed eigenenergy $\left\Vert \boldsymbol{\nu
}\right\Vert ^{2}=%
{\textstyle\sum\nolimits_{i=1}^{N}}
\lambda_{i}\left\Vert x_{i\ast}\widehat{\mathbf{v}}_{i}\right\Vert ^{2}$, are
both determined by the total allowed eigenenergy $\left\Vert \boldsymbol{\nu
}\right\Vert ^{2}=%
{\textstyle\sum\nolimits_{i=1}^{N}}
\lambda_{i}\left\Vert x_{i\ast}\widehat{\mathbf{v}}_{i}\right\Vert ^{2}$
exhibited by the principal eigenaxis $\boldsymbol{\nu}$ $\boldsymbol{=}%
{\textstyle\sum\nolimits_{i=1}^{N}}
\sqrt{\lambda_{i}}x_{i\ast}\widehat{\mathbf{v}}_{i}$ of the geometric locus of
the quadratic surface, such that the eigenvalues $\lambda_{i}$ of an $N\times
N$ symmetric matrix $\mathbf{Q\in}$ $\Re^{N\times N}$ of a quadratic form
$\mathbf{v}^{T}\mathbf{Qv}$ regulate the total allowed eigenenergy $\left\Vert
\boldsymbol{\nu}\right\Vert ^{2}$ exhibited by the principal eigenaxis
$\boldsymbol{\nu}$ of the geometric locus of the quadratic surface; and the
uniform property exhibited by all of the points that lie on the geometric
locus of the quadratic surface is the total allowed eigenenergy $\left\Vert
\boldsymbol{\nu}\right\Vert ^{2}$ exhibited by the principal eigenaxis
$\boldsymbol{\nu}$ of the geometric locus of the quadratic surface.
\end{theorem}

\begin{proof}
Take any given vector algebra locus equation $\mathbf{x}^{T}\mathbf{Qx}=c$ of
a quadratic surface that is satisfied by a quadratic form $\mathbf{x}%
^{T}\mathbf{Qx}$, such that an $N\times N$ symmetric matrix $\mathbf{Q\in}$
$\Re^{N\times N}$ is defined through the vector algebra locus equation
$\mathbf{x}^{T}\mathbf{Qx}=c$, wherein $c$ is a certain constant, and the
vector $\mathbf{x}$ is written as $\mathbf{x=}%
{\textstyle\sum\nolimits_{i=1}^{N}}
x_{i}\mathbf{e}_{i}$, at which point $x_{i}$ is a scale factor for a standard
basis vector $\mathbf{e}_{i}$ that belongs to the set%
\[
\left\{  \mathbf{e}_{1}=\left(  1,0,\ldots,0\right)  ,\ldots,\mathbf{e}%
_{N}=\left(  0,0,\ldots,1\right)  \right\}  \text{.}%
\]

Write the vector $\mathbf{x=}%
{\textstyle\sum\nolimits_{i=1}^{N}}
x_{i}\mathbf{e}_{i}$ in terms of a basis of unit eigenvectors $\left\{
\widehat{\mathbf{v}}_{1},\ldots\widehat{\mathbf{v}}_{N}\right\}  $ of the
matrix $\mathbf{Q}$, so that%
\begin{align*}
\mathbf{x}  &  \mathbf{=}%
{\textstyle\sum\nolimits_{i=1}^{N}}
x_{i}\mathbf{e}_{i}\\
&  \equiv%
{\textstyle\sum\nolimits_{i=1}^{N}}
x_{i\ast}\widehat{\mathbf{v}}_{i}\text{,}%
\end{align*}
at which point the vector $\mathbf{x=}%
{\textstyle\sum\nolimits_{i=1}^{N}}
x_{i}\mathbf{e}_{i}$ is transformed into the principal eigenvector
$\mathbf{v}=%
{\textstyle\sum\nolimits_{i=1}^{N}}
x_{i\ast}\widehat{\mathbf{v}}_{i}$ of the symmetric matrix $\mathbf{Q}$.

Take the eigenvalues $\lambda_{N}\leq\mathbf{\ldots}\leq\lambda_{1}$ of the
$N\times N$ symmetric matrix $\mathbf{Q\in}$ $\Re^{N\times N}$ and let the
principal eigenvector $\mathbf{v}=%
{\textstyle\sum\nolimits_{i=1}^{N}}
x_{i\ast}\widehat{\mathbf{v}}_{i}$ of the symmetric matrix $\mathbf{Q}$ of the
quadratic form $\mathbf{v}^{T}\mathbf{Qv}$ be symmetrically and equivalently
related to the principal eigenaxis $\boldsymbol{\nu}$ $\boldsymbol{=}%
{\textstyle\sum\nolimits_{i=1}^{N}}
\sqrt{\lambda_{i}}x_{i\ast}\widehat{\mathbf{v}}_{i}$ of the quadratic surface,
so that substitution of the expression $\mathbf{x\triangleq\ }%
{\textstyle\sum\nolimits_{i=1}^{N}}
x_{i\ast}\widehat{\mathbf{v}}_{i}$ into the quadratic form $\mathbf{x}%
^{T}\mathbf{Qx}$ in the vector algebra locus equation $\mathbf{x}%
^{T}\mathbf{Qx}=c$ produces the equivalent form of the vector algebra locus
equation $\mathbf{x}^{T}\mathbf{Qx}=c$%
\begin{align*}
\boldsymbol{T}\left[  \mathbf{x}^{T}\mathbf{Qx=c}\right]   &  =\left(
{\textstyle\sum\nolimits_{i=1}^{N}}
x_{i\ast}\widehat{\mathbf{v}}_{i}\right)  ^{T}\mathbf{Q}\left(
{\textstyle\sum\nolimits_{j=1}^{N}}
x_{j\ast}\widehat{\mathbf{v}}_{j}\right) \\
&  =\mathbf{v}^{T}\mathbf{Qv}\\
&  =%
{\textstyle\sum\nolimits_{i=1}^{N}}
{\textstyle\sum\nolimits_{j=1}^{N}}
x_{i\ast}x_{j\ast}\left(  \widehat{\mathbf{v}}_{i}^{T}\mathbf{Q}%
\widehat{\mathbf{v}}_{j}\right) \\
&  =%
{\textstyle\sum\nolimits_{i=1}^{N}}
{\textstyle\sum\nolimits_{j=1}^{N}}
x_{i\ast}x_{j\ast}\lambda_{j}\left(  \widehat{\mathbf{v}}_{i}^{T}%
\widehat{\mathbf{v}}_{j}\right) \\
&  =%
{\textstyle\sum\nolimits_{i=1}^{N}}
\left(  \sqrt{\lambda_{i}}x_{i\ast}\widehat{\mathbf{v}}_{i}^{T}\right)
\left(  \sqrt{\lambda_{i}}x_{i\ast}\widehat{\mathbf{v}}_{i}\right) \\
&  =%
{\textstyle\sum\nolimits_{i=1}^{N}}
\lambda_{i}x_{i\ast}^{2}\widehat{\mathbf{v}}_{i}^{T}\widehat{\mathbf{v}}_{i}=%
{\textstyle\sum\nolimits_{i=1}^{N}}
\lambda_{i}\left\Vert x_{i\ast}\widehat{\mathbf{v}}_{i}\right\Vert ^{2}\\
&  =\left\Vert \boldsymbol{\nu}\right\Vert ^{2}=c\text{,}%
\end{align*}
such that the principal eigenvector $\mathbf{v}=%
{\textstyle\sum\nolimits_{i=1}^{N}}
x_{i\ast}\widehat{\mathbf{v}}_{i}$ of the $N\times N$ symmetric matrix
$\mathbf{Q}$ of the quadratic form $\mathbf{v}^{T}\mathbf{Qv}$ is
symmetrically and equivalently related to the principal eigenaxis
$\boldsymbol{\nu}$ $\boldsymbol{=}%
{\textstyle\sum\nolimits_{i=1}^{N}}
\sqrt{\lambda_{i}}x_{i\ast}\widehat{\mathbf{v}}_{i}$ of the quadratic surface,
so that the geometric locus of the principal eigenaxis $\boldsymbol{\nu=}%
{\textstyle\sum\nolimits_{i=1}^{N}}
\sqrt{\lambda_{i}}x_{\ast}\widehat{\mathbf{v}}_{i}$ of the quadratic surface
satisfies the geometric locus of the quadratic surface in terms of its total
allowed eigenenergy%
\[
\left\Vert \boldsymbol{\nu}\right\Vert ^{2}=%
{\textstyle\sum\nolimits_{i=1}^{N}}
\lambda_{i}\left\Vert x_{\ast}\widehat{\mathbf{v}}_{i}\right\Vert ^{2}\text{,}%
\]
such that the eigenenergy $\lambda_{i}\left\Vert x_{\ast}\widehat{\mathbf{v}%
}_{i}\right\Vert ^{2}$ exhibited by each component $\sqrt{\lambda_{i}}%
x_{i\ast}\widehat{\mathbf{v}}_{i}$ of the principal eigenaxis $\boldsymbol{\nu
=}%
{\textstyle\sum\nolimits_{i=1}^{N}}
\sqrt{\lambda_{i}}x_{\ast}\widehat{\mathbf{v}}_{i}$ of the quadratic surface
is modulated by an eigenvalue $\lambda_{i}$ of the symmetric matrix
$\mathbf{Q}$ of the quadratic form $\mathbf{v}^{T}\mathbf{Qv}$; and the
uniform property exhibited by all of the points that lie on the locus of the
quadratic surface is the total allowed eigenenergy $\left\Vert \boldsymbol{\nu
}\right\Vert ^{2}$ exhibited by the principal eigenaxis $\boldsymbol{\nu}$ of
the geometric locus of the quadratic surface.

Thereby, the exclusive principal eigen-coordinate system%
\[
\boldsymbol{\nu}=%
{\textstyle\sum\nolimits_{i=1}^{N}}
\sqrt{\lambda_{i}}x_{i\ast}\widehat{\mathbf{v}}_{i}%
\]
is the principal part of an equivalent representation of the quadratic form
$\mathbf{x}^{T}\mathbf{Qx}$ in the following manner%
\begin{align*}
\left(
{\textstyle\sum\nolimits_{i=1}^{N}}
x_{i\ast}\widehat{\mathbf{v}}_{i}\right)  ^{T}\mathbf{Q}\left(
{\textstyle\sum\nolimits_{j=1}^{N}}
x_{j\ast}\widehat{\mathbf{v}}_{j}\right)   &  =\mathbf{v}^{T}\mathbf{Qv}\\
&  =\left(
{\textstyle\sum\nolimits_{i=1}^{N}}
x_{i\ast}\widehat{\mathbf{v}}_{i}\right)  ^{T}\left(
{\textstyle\sum\nolimits_{j=1}^{N}}
\lambda_{j}x_{j\ast}\widehat{\mathbf{v}}_{j}\right) \\
&  =%
{\textstyle\sum\nolimits_{i=1}^{N}}
\left(  \sqrt{\lambda_{i}}x_{i\ast}\widehat{\mathbf{v}}_{i}\right)
^{T}\left(  \sqrt{\lambda_{i}}x_{i\ast}\widehat{\mathbf{v}}_{i}\right) \\
&  =%
{\textstyle\sum\nolimits_{i=1}^{N}}
\lambda_{i}x_{i\ast}^{2}\left\Vert \widehat{\mathbf{v}}_{i}\right\Vert ^{2}=%
{\textstyle\sum\nolimits_{i=1}^{N}}
\lambda_{i}\left\Vert x_{i\ast}\widehat{\mathbf{v}}_{i}\right\Vert ^{2}\\
&  =\left\Vert \boldsymbol{\nu}\right\Vert ^{2}\text{,}%
\end{align*}
so that the principal eigenvector $\mathbf{v}=%
{\textstyle\sum\nolimits_{i=1}^{N}}
x_{i\ast}\widehat{\mathbf{v}}_{i}$ of the $N\times N$ symmetric matrix
$\mathbf{Q}$ of the quadratic form $\mathbf{v}^{T}\mathbf{Qv}$ is
symmetrically and equivalently related to the principal eigenaxis
$\boldsymbol{\nu=}%
{\textstyle\sum\nolimits_{i=1}^{N}}
\sqrt{\lambda_{i}}x_{i\ast}\widehat{\mathbf{v}}_{i}$ of a correlated quadratic
surface, so that the geometric locus of the principal eigenaxis
$\boldsymbol{\nu=}%
{\textstyle\sum\nolimits_{i=1}^{N}}
\sqrt{\lambda_{i}}x_{\ast}\widehat{\mathbf{v}}_{i}$ of the quadratic surface
satisfies the geometric locus of the quadratic surface in terms of its total
allowed eigenenergy%
\[
\left\Vert \boldsymbol{\nu}\right\Vert ^{2}=%
{\textstyle\sum\nolimits_{i=1}^{N}}
\lambda_{i}\left\Vert x_{\ast}\widehat{\mathbf{v}}_{i}\right\Vert ^{2}\text{,}%
\]
such that the eigenenergy $\lambda_{i}\left\Vert x_{i\ast}\widehat{\mathbf{v}%
}_{i}\right\Vert ^{2}$ exhibited by each component $\sqrt{\lambda_{i}}%
x_{i\ast}\widehat{\mathbf{v}}_{i}$ of the principal eigenaxis $\boldsymbol{\nu
}=\sqrt{\lambda_{1}}x_{1\ast}\widehat{\mathbf{v}}_{1}+\mathbf{\ldots+\ }%
\sqrt{\lambda_{N}}x_{N\ast}\widehat{\mathbf{v}}_{N}$ of the geometric locus of
the quadratic surface is modulated by an eigenvalue $\lambda_{i}$ of the
$N\times N$ symmetric matrix $\mathbf{Q}$ of the quadratic form $\mathbf{v}%
^{T}\mathbf{Qv}$ in a manner that regulates the total allowed eigenenergy
$\left\Vert \boldsymbol{\nu}\right\Vert ^{2}=%
{\textstyle\sum\nolimits_{i=1}^{N}}
\lambda_{i}\left\Vert x_{i\ast}\widehat{\mathbf{v}}_{i}\right\Vert ^{2}$
exhibited by the geometric locus of the principal eigenaxis $\boldsymbol{\nu=}%
{\textstyle\sum\nolimits_{i=1}^{N}}
\sqrt{\lambda_{i}}x_{i\ast}\widehat{\mathbf{v}}_{i}$, wherein the $N\times N$
symmetric matrix $\mathbf{Q\in}$ $\Re^{N\times N}$ has the simple diagonal
form%
\[
\mathbf{Q}_{ij}=\left\{
\begin{array}
[c]{c}%
0\text{ if }i\neq j\\
\lambda_{i}\text{ if }i=j
\end{array}
\right.  \text{.}%
\]

Thereby, the shape and the fundamental property exhibited by the geometric
locus of any given quadratic surface that is represented by a vector algebra
locus equation that has the form%
\begin{align*}
\lambda_{1}x_{1\ast}^{2}\widehat{\mathbf{v}}_{1}^{2}+\mathbf{\ldots+}%
\lambda_{N}x_{N\ast}^{2}\widehat{\mathbf{v}}_{N}^{2}  &  =%
{\textstyle\sum\nolimits_{i=1}^{N}}
\lambda_{i}x_{i\ast}^{2}\left\Vert \widehat{\mathbf{v}}_{i}\right\Vert ^{2}\\
&  =%
{\textstyle\sum\nolimits_{i=1}^{N}}
\lambda_{i}\left\Vert x_{i\ast}\widehat{\mathbf{v}}_{i}\right\Vert ^{2}\\
&  =\left\Vert \boldsymbol{\nu}\right\Vert ^{2}\text{,}%
\end{align*}
so that an exclusive principal eigen-coordinate system%
\begin{align*}
\boldsymbol{\nu}  &  =\sqrt{\lambda_{1}}x_{1\ast}\widehat{\mathbf{v}}%
_{1}+\mathbf{\ldots+}\sqrt{\lambda_{N}}x_{N\ast}\widehat{\mathbf{v}}_{N}\\
&  =%
{\textstyle\sum\nolimits_{i=1}^{N}}
\sqrt{\lambda_{i}}x_{i\ast}\widehat{\mathbf{v}}_{i}%
\end{align*}
of the quadratic surface satisfies the geometric locus of the quadratic
surface in terms of its total allowed eigenenergy $\left\Vert \boldsymbol{\nu
}\right\Vert ^{2}=%
{\textstyle\sum\nolimits_{i=1}^{N}}
\lambda_{i}\left\Vert x_{i\ast}\widehat{\mathbf{v}}_{i}\right\Vert ^{2}$, are
both determined by the total allowed eigenenergy $\left\Vert \boldsymbol{\nu
}\right\Vert ^{2}=%
{\textstyle\sum\nolimits_{i=1}^{N}}
\lambda_{i}\left\Vert x_{i\ast}\widehat{\mathbf{v}}_{i}\right\Vert ^{2}$
exhibited by the principal eigenaxis $\boldsymbol{\nu=}%
{\textstyle\sum\nolimits_{i=1}^{N}}
\sqrt{\lambda_{i}}x_{i\ast}\widehat{\mathbf{v}}_{i}$ of the geometric locus of
the quadratic surface, at which point the eigenvalues $\lambda_{i}$ of an
$N\times N$ symmetric matrix $\mathbf{Q\in}$ $\Re^{N\times N}$ of a quadratic
form $\mathbf{v}^{T}\mathbf{Qv}$ regulate the total allowed eigenenergy
$\left\Vert \boldsymbol{\nu}\right\Vert ^{2}=%
{\textstyle\sum\nolimits_{i=1}^{N}}
\lambda_{i}\left\Vert x_{i\ast}\widehat{\mathbf{v}}_{i}\right\Vert ^{2}$
exhibited by the principal eigenaxis $\boldsymbol{\nu}$ $\boldsymbol{=}%
{\textstyle\sum\nolimits_{i=1}^{N}}
\sqrt{\lambda_{i}}x_{i\ast}\widehat{\mathbf{v}}_{i}$ of the geometric locus of
the quadratic surface; and the uniform property exhibited by all of the points
that lie on the geometric locus of the quadratic surface is the total allowed
eigenenergy $\left\Vert \boldsymbol{\nu}\right\Vert ^{2}$ exhibited by the
principal eigenaxis $\boldsymbol{\nu}$ of the geometric locus of the quadratic surface.

Therefore, it is concluded the shape and the fundamental property exhibited by
the geometric locus of any given quadratic surface are both determined by an
exclusive principal eigen-coordinate system of the geometric locus of the
quadratic surface, such that the eigenvalues of a symmetric matrix of a
correlated quadratic form modulate the eigenenergies exhibited by the
components of the exclusive principal eigen-coordinate system---which is the
principal part of an equivalent representation of the quadratic form---at
which point the exclusive principal eigen-coordinate system is the principal
eigenaxis of the geometric locus of the quadratic surface, so that the
principal eigenaxis satisfies the geometric locus of the quadratic surface in
terms of its total allowed eigenenergy; and the uniform property exhibited by
all of the points that lie on the geometric locus of the quadratic surface is
the total allowed eigenenergy exhibited by the principal eigenaxis of the
geometric locus of the quadratic surface.
\end{proof}

The above proof clearly applies to any given vector algebra locus equation
$\mathbf{x}^{T}\mathbf{Qx}=c$ that is satisfied by a quadratic form
$\mathbf{x}^{T}\mathbf{Qx}$, such that a $2\times2$ symmetric matrix
$\mathbf{Q\in}$ $\Re^{2\times2}$ is defined through the vector algebra locus
equation $\mathbf{x}^{T}\mathbf{Qx}=c$, wherein $c$ is a certain constant, and
the vector $\mathbf{x}$ is written as $\mathbf{x=}%
{\textstyle\sum\nolimits_{i=1}^{2}}
x_{i}\mathbf{e}_{i}$, where $x_{i}$ is a scale factor for a standard basis
vector $\mathbf{e}_{i}$ that belongs to the set $\left\{  \mathbf{e}%
_{1}=\left(  1,0\right)  ,\mathbf{e}_{2}=\left(  0,1\right)  \right\}  $.

Thereby, it is concluded the shape and the fundamental property exhibited by
the geometric locus of any given quadratic curve are both determined by an
exclusive principal eigen-coordinate system of the locus of the quadratic
curve, such that the eigenvalues of a symmetric matrix of a correlated
quadratic form modulate the eigenenergies exhibited by the components of the
exclusive principal eigen-coordinate system---which is the principal part of
an equivalent representation of the quadratic form---at which point the
exclusive principal eigen-coordinate system is the principal eigenaxis of the
geometric locus of the quadratic curve, so that the principal eigenaxis
satisfies the geometric locus of the quadratic curve in terms of its total
allowed eigenenergy; and the uniform property exhibited by all of the points
that lie on the geometric locus of the quadratic curve is the total allowed
eigenenergy exhibited by the principal eigenaxis of the geometric locus of the
quadratic curve.

It is important that Theorem \ref{Principal Eigen-coordinate System Theorem}
is readily generalized in the manner expressed by Corollary
\ref{Principal Eigen-coordinate System Corollary} since the algebraic vector
expression%
\[
\mathbf{x}^{T}\mathbf{\Sigma}_{1}^{-1}\mathbf{x-x}^{T}\mathbf{\Sigma}_{2}%
^{-1}\mathbf{x}%
\]
determines the mathematical structure of both the discriminant function and
the intrinsic coordinate system---of the geometric locus of the decision
boundary---of any given minimum risk binary classification that is subject to
multivariate normal data.

\subsection{Existence Corollary of a Principal Eigenaxis}

Corollary \ref{Principal Eigen-coordinate System Corollary} guarantees the
existence of an exclusive principal eigen-coordinate system---which is the
principal part of an equivalent representation of a correlated quadratic form
$\mathbf{x}^{T}\mathbf{Q}^{-1}\mathbf{x}$---such that the exclusive principal
eigen-coordinate system is the solution of an equivalent form of the vector
algebra locus equation of the geometric locus of a certain quadratic curve or
surface---so that the principal eigenaxis of the geometric locus of the
quadratic curve or surface satisfies the geometric locus of the quadratic
curve or surface in terms of its \emph{total allowed eigenenergy}.

\begin{corollary}
\label{Principal Eigen-coordinate System Corollary}Take any given vector
algebra locus equation of a quadratic curve that has the form%
\[
\mathbf{x}^{T}\mathbf{Q}^{-1}\mathbf{x}=c\text{,}%
\]
such that $\mathbf{Q}^{-1}$ is a $2\times2$ symmetric matrix $\mathbf{Q}%
^{-1}\mathbf{\in}$ $\Re^{2\times2}$ of a certain quadratic form $\mathbf{x}%
^{T}\mathbf{Q}^{-1}\mathbf{x}$, $c$ is a certain constant, and the vector
$\mathbf{x}$ is written as $\mathbf{x=}%
{\textstyle\sum\nolimits_{i=1}^{2}}
x_{i}\mathbf{e}_{i}$, where $x_{i}$ is a scale factor for a standard basis
vector $\mathbf{e}_{i}$ that belongs to the set $\left\{  \mathbf{e}%
_{1}=\left(  1,0\right)  ,\mathbf{e}_{2}=\left(  0,1\right)  \right\}  $.

Let an equivalent form of the vector algebra locus equation $\mathbf{x}%
^{T}\mathbf{Q}^{-1}\mathbf{x}=c$ be generated by transforming the positions of
the coordinate axes of the quadratic curve into the axes of an exclusive
principal eigen-coordinate system, so that the principal eigenaxis of the
quadratic curve satisfies the geometric locus of the quadratic curve in terms
of its total allowed eigenenergy.

It follows that the exclusive principal eigen-coordinate system%
\[
\boldsymbol{\nu=}%
{\textstyle\sum\nolimits_{i=1}^{2}}
\sqrt{\lambda_{i}^{-1}}x_{i\ast}\widehat{\mathbf{v}}_{i}%
\]
is the principal part of an equivalent representation of the quadratic form
$\mathbf{x}^{T}\mathbf{Q}^{-1}\mathbf{x}$ in the following manner%
\begin{align*}
\left(
{\textstyle\sum\nolimits_{i=1}^{2}}
x_{i\ast}\widehat{\mathbf{v}}_{i}\right)  ^{T}\mathbf{Q}^{-1}\left(
{\textstyle\sum\nolimits_{j=1}^{2}}
x_{j\ast}\widehat{\mathbf{v}}_{j}\right)   &  =\mathbf{v}^{T}\mathbf{Q}%
^{-1}\mathbf{v}\\
&  =\left(
{\textstyle\sum\nolimits_{i=1}^{2}}
x_{i\ast}\widehat{\mathbf{v}}_{i}\right)  ^{T}\left(
{\textstyle\sum\nolimits_{j=1}^{2}}
\lambda_{j}^{-1}x_{j\ast}\widehat{\mathbf{v}}_{j}\right) \\
&  =%
{\textstyle\sum\nolimits_{i=1}^{2}}
\left(  \sqrt{\lambda_{i}^{-1}}x_{i\ast}\widehat{\mathbf{v}}_{i}^{T}\right)
\left(  \sqrt{\lambda_{i}^{-1}}x_{i\ast}\widehat{\mathbf{v}}_{i}\right) \\
&  =%
{\textstyle\sum\nolimits_{i=1}^{2}}
\lambda_{i}^{-1}\left\Vert x_{i\ast}\widehat{\mathbf{v}}_{i}\right\Vert ^{2}\\
&  =\left\Vert \boldsymbol{\nu}\right\Vert ^{2}\text{,}%
\end{align*}
wherein the vector $\mathbf{x=}%
{\textstyle\sum\nolimits_{i=1}^{2}}
x_{i}\mathbf{e}_{i}$ is transformed into the principal eigenvector
$\mathbf{v}=%
{\textstyle\sum\nolimits_{i=1}^{2}}
x_{i\ast}\widehat{\mathbf{v}}_{i}$ of the symmetric matrix $\mathbf{Q}%
^{-1}\mathbf{\in}$ $\Re^{2\times2}$ of the quadratic form $\mathbf{v}%
^{T}\mathbf{Q}^{-1}\mathbf{v}$, such that the principal eigenvector
$\mathbf{v}=%
{\textstyle\sum\nolimits_{i=1}^{2}}
x_{i\ast}\widehat{\mathbf{v}}_{i}$ of the symmetric matrix $\mathbf{Q}^{-1}$
of the quadratic form $\mathbf{v}^{T}\mathbf{Q}^{-1}\mathbf{v}$ is
symmetrically and equivalently related to the principal eigenaxis
$\boldsymbol{\nu=}%
{\textstyle\sum\nolimits_{i=1}^{2}}
\sqrt{\lambda_{i}^{-1}}x_{i\ast}\widehat{\mathbf{v}}_{i}$ of a certain
quadratic curve, so that the exclusive principal eigen-coordinate system
$\boldsymbol{\nu=}%
{\textstyle\sum\nolimits_{i=1}^{2}}
\sqrt{\lambda_{i}^{-1}}x_{i\ast}\widehat{\mathbf{v}}_{i}$ of the quadratic
curve satisfies the geometric locus of the quadratic curve in terms of its in
terms of its total allowed eigenenergy $\left\Vert \boldsymbol{\nu}\right\Vert
^{2}=%
{\textstyle\sum\nolimits_{i=1}^{2}}
\lambda_{i}^{-1}\left\Vert x_{i\ast}\widehat{\mathbf{v}}_{i}\right\Vert ^{2}$,
such that the eigenenergy $\lambda_{i}^{-1}\left\Vert x_{i\ast}%
\widehat{\mathbf{v}}_{i}\right\Vert ^{2}$ exhibited by each component
$\sqrt{\lambda_{i}^{-1}}x_{i\ast}\widehat{\mathbf{v}}_{i}$ of the principal
eigenaxis $\boldsymbol{\nu}=\sqrt{\lambda_{1}^{-1}}x_{1\ast}%
\widehat{\mathbf{v}}_{1}+\sqrt{\lambda_{2}^{-1}}x_{2\ast}\widehat{\mathbf{v}%
}_{2}$ of the geometric locus of the quadratic curve is modulated by an
eigenvalue $\lambda_{i}^{-1}$ of the symmetric matrix $\mathbf{Q}^{-1}$ of the
quadratic form $\mathbf{v}^{T}\mathbf{Q}^{-1}\mathbf{v}$ in a manner that
regulates the total allowed eigenenergy $\left\Vert \boldsymbol{\nu
}\right\Vert ^{2}=%
{\textstyle\sum\nolimits_{i=1}^{2}}
\lambda_{i}^{-1}\left\Vert x_{i\ast}\widehat{\mathbf{v}}_{i}\right\Vert ^{2}$
exhibited by the geometric locus of the principal eigenaxis $\boldsymbol{\nu=}%
{\textstyle\sum\nolimits_{i=1}^{2}}
\sqrt{\lambda_{i}^{-1}}x_{i\ast}\widehat{\mathbf{v}}_{i}$, wherein the
$2\times2$ symmetric matrix $\mathbf{Q}^{-1}\mathbf{\in}$ $\Re^{2\times2}$ has
the simple diagonal form%
\[
\mathbf{Q}_{ij}^{-1}=\left\{
\begin{array}
[c]{c}%
0\text{ if }i\neq j\\
\lambda_{i}^{-1}\text{ if }i=j
\end{array}
\right.  \text{.}%
\]

Thereby, the shape and the fundamental property exhibited by the geometric
locus of any given quadratic curve that is represented by a vector algebra
locus equation that has the form%
\begin{align*}
\lambda_{1}^{-1}x_{1\ast}^{2}\widehat{\mathbf{v}}_{1}^{2}+\lambda_{2}%
^{-1}x_{2\ast}^{2}\widehat{\mathbf{v}}_{2}^{2}  &  =%
{\textstyle\sum\nolimits_{i=1}^{2}}
\lambda_{i}^{-1}x_{i\ast}^{2}\left\Vert \widehat{\mathbf{v}}_{i}\right\Vert
^{2}\\
&  =%
{\textstyle\sum\nolimits_{i=1}^{2}}
\lambda_{i}^{-1}\left\Vert x_{i\ast}\widehat{\mathbf{v}}_{i}\right\Vert ^{2}\\
&  =\left\Vert \boldsymbol{\nu}\right\Vert ^{2}\text{,}%
\end{align*}
so that an exclusive principal eigen-coordinate system%
\begin{align*}
\boldsymbol{\nu}  &  =\sqrt{\lambda_{1}^{-1}}x_{1\ast}\widehat{\mathbf{v}}%
_{1}+\sqrt{\lambda_{2}^{-1}}x_{2\ast}\widehat{\mathbf{v}}_{2}\\
&  =%
{\textstyle\sum\nolimits_{i=1}^{2}}
\sqrt{\lambda_{i}^{-1}}x_{i\ast}\widehat{\mathbf{v}}_{i}%
\end{align*}
of the quadratic curve satisfies the geometric locus of the quadratic curve in
terms of its total allowed eigenenergy $\left\Vert \boldsymbol{\nu}\right\Vert
^{2}=%
{\textstyle\sum\nolimits_{i=1}^{2}}
\lambda_{i}^{-1}\left\Vert x_{i\ast}\widehat{\mathbf{v}}_{i}\right\Vert ^{2}$,
are both determined by the total allowed eigenenergy $\left\Vert
\boldsymbol{\nu}\right\Vert ^{2}=%
{\textstyle\sum\nolimits_{i=1}^{2}}
\lambda_{i}^{-1}\left\Vert x_{i\ast}\widehat{\mathbf{v}}_{i}\right\Vert ^{2}$
exhibited by the principal eigenaxis $\boldsymbol{\nu}$ $\boldsymbol{=}%
{\textstyle\sum\nolimits_{i=1}^{2}}
\sqrt{\lambda_{i}^{-1}}x_{i\ast}\widehat{\mathbf{v}}_{i}$ of the geometric
locus of the quadratic curve, such that the eigenvalues $\lambda_{i}^{-1}$ of
a $2\times2$ symmetric matrix $\mathbf{Q}^{-1}\mathbf{\in}$ $\Re^{2\times2}$
of a quadratic form $\mathbf{v}^{T}\mathbf{Q}^{-1}\mathbf{v}$ regulate the
total allowed eigenenergy $\left\Vert \boldsymbol{\nu}\right\Vert ^{2}$
exhibited by the principal eigenaxis $\boldsymbol{\nu}$ of the geometric locus
of the quadratic curve; and the uniform property exhibited by all of the
points that lie on the geometric locus of the quadratic curve is the total
allowed eigenenergy $\left\Vert \boldsymbol{\nu}\right\Vert ^{2}$ exhibited by
the principal eigenaxis $\boldsymbol{\nu}$ of the geometric locus of the
quadratic curve.

Correspondingly, take any given vector algebra locus equation of a quadratic
surface that has the form%
\[
\mathbf{x}^{T}\mathbf{Q}^{-1}\mathbf{x}=c\text{,}%
\]
such that $\mathbf{Q}^{-1}$ is an $N\times N$ symmetric matrix $\mathbf{Q}%
^{-1}\mathbf{\in}$ $\Re^{N\times N}$ of a certain quadratic form
$\mathbf{x}^{T}\mathbf{Q}^{-1}\mathbf{x}$, $c$ is a certain constant, and the
vector $\mathbf{x}$ is written as $\mathbf{x=}%
{\textstyle\sum\nolimits_{i=1}^{N}}
x_{i}\mathbf{e}_{i}$, where $x_{i}$ is a scale factor for a standard basis
vector $\mathbf{e}_{i}$ that belongs to the set$\left\{  \mathbf{e}%
_{1}=\left(  1,0,\ldots,0\right)  ,\ldots,\mathbf{e}_{N}=\left(
0,0,\ldots,1\right)  \right\}  $.

Let an equivalent form of the vector algebra locus equation $\mathbf{x}%
^{T}\mathbf{Q}^{-1}\mathbf{x}=c$ be generated by transforming the positions of
the coordinate axes of the quadratic surface into the axes of an exclusive
principal eigen-coordinate system, so that the principal eigenaxis of the
quadratic surface satisfies the geometric locus of the quadratic surface in
terms of its total allowed eigenenergy.

It follows that the exclusive principal eigen-coordinate system%
\[
\boldsymbol{\nu}=%
{\textstyle\sum\nolimits_{i=1}^{N}}
\sqrt{\lambda_{i}^{-1}}x_{i\ast}\widehat{\mathbf{v}}_{i}%
\]
is the principal part of an equivalent representation of the quadratic form
$\mathbf{x}^{T}\mathbf{Q}^{-1}\mathbf{x}$ in the following manner%
\begin{align*}
\left(
{\textstyle\sum\nolimits_{i=1}^{N}}
x_{i\ast}\widehat{\mathbf{v}}_{i}\right)  ^{T}\mathbf{Q}^{-1}\left(
{\textstyle\sum\nolimits_{j=1}^{N}}
x_{j\ast}\widehat{\mathbf{v}}_{j}\right)   &  =\mathbf{v}^{T}\mathbf{Q}%
^{-1}\mathbf{v}\\
&  =\left(
{\textstyle\sum\nolimits_{i=1}^{N}}
x_{i\ast}\widehat{\mathbf{v}}_{i}\right)  ^{T}\left(
{\textstyle\sum\nolimits_{j=1}^{N}}
\lambda_{j}^{-1}x_{j\ast}\widehat{\mathbf{v}}_{j}\right) \\
&  =%
{\textstyle\sum\nolimits_{i=1}^{N}}
\left(  \sqrt{\lambda_{i}^{-1}}x_{i\ast}\widehat{\mathbf{v}}_{i}\right)
^{T}\left(  \sqrt{\lambda_{i}^{-1}}x_{i\ast}\widehat{\mathbf{v}}_{i}\right) \\
&  =%
{\textstyle\sum\nolimits_{i=1}^{N}}
\lambda_{i}^{-1}x_{i\ast}^{2}\left\Vert \widehat{\mathbf{v}}_{i}\right\Vert
^{2}=%
{\textstyle\sum\nolimits_{i=1}^{N}}
\lambda_{i}^{-1}\left\Vert x_{i\ast}\widehat{\mathbf{v}}_{i}\right\Vert ^{2}\\
&  =\left\Vert \boldsymbol{\nu}\right\Vert ^{2}\text{,}%
\end{align*}
wherein the vector $\mathbf{x=}%
{\textstyle\sum\nolimits_{i=1}^{N}}
x_{i}\mathbf{e}_{i}$ is transformed into the principal eigenvector
$\mathbf{v}=%
{\textstyle\sum\nolimits_{i=1}^{N}}
x_{i\ast}\widehat{\mathbf{v}}_{i}$ of the symmetric matrix $\mathbf{Q}%
^{-1}\mathbf{\in}$ $\Re^{N\times N}$ of the quadratic form $\mathbf{v}%
^{T}\mathbf{Q}^{-1}\mathbf{v}$, such that the principal eigenvector
$\mathbf{v}=%
{\textstyle\sum\nolimits_{i=1}^{N}}
x_{i\ast}\widehat{\mathbf{v}}_{i}$ of the symmetric matrix $\mathbf{Q}^{-1}$
of the quadratic form $\mathbf{v}^{T}\mathbf{Q}^{-1}\mathbf{v}$ is
symmetrically and equivalently related to the principal eigenaxis
$\boldsymbol{\nu=}%
{\textstyle\sum\nolimits_{i=1}^{N}}
\sqrt{\lambda_{i}^{-1}}x_{i\ast}\widehat{\mathbf{v}}_{i}$ of a certain
quadratic surface, so that the exclusive principal eigen-coordinate system
$\boldsymbol{\nu=}%
{\textstyle\sum\nolimits_{i=1}^{N}}
\sqrt{\lambda_{i}^{-1}}x_{i\ast}\widehat{\mathbf{v}}_{i}$ of the quadratic
surface satisfies the geometric locus of the quadratic surface in terms of its
in terms of its total allowed eigenenergy $\left\Vert \boldsymbol{\nu
}\right\Vert ^{2}=%
{\textstyle\sum\nolimits_{i=1}^{N}}
\lambda_{i}^{-1}\left\Vert x_{i\ast}\widehat{\mathbf{v}}_{i}\right\Vert ^{2}$,
such that the eigenenergy $\lambda_{i}^{-1}\left\Vert x_{i\ast}%
\widehat{\mathbf{v}}_{i}\right\Vert ^{2}$ exhibited by each component
$\sqrt{\lambda_{i}^{-1}}x_{i\ast}\widehat{\mathbf{v}}_{i}$ of the principal
eigenaxis $\boldsymbol{\nu}=\sqrt{\lambda_{1}^{-1}}x_{1\ast}%
\widehat{\mathbf{v}}_{1}+\mathbf{\ldots+}\sqrt{\lambda_{N}^{-1}}x_{N\ast
}\widehat{\mathbf{v}}_{N}$ of the geometric locus of the quadratic surface is
modulated by an eigenvalue $\lambda_{i}^{-1}$ of the symmetric matrix
$\mathbf{Q}^{-1}$ of the quadratic form $\mathbf{v}^{T}\mathbf{Q}%
^{-1}\mathbf{v}$ in a manner that regulates the total allowed eigenenergy
$\left\Vert \boldsymbol{\nu}\right\Vert ^{2}=%
{\textstyle\sum\nolimits_{i=1}^{N}}
\lambda_{i}^{-1}\left\Vert x_{i\ast}\widehat{\mathbf{v}}_{i}\right\Vert ^{2}$
exhibited by the geometric locus of the principal eigenaxis $\boldsymbol{\nu=}%
{\textstyle\sum\nolimits_{i=1}^{N}}
\sqrt{\lambda_{i}^{-1}}x_{i\ast}\widehat{\mathbf{v}}_{i}$, wherein the
$N\times N$ symmetric matrix $\mathbf{Q}^{-1}\mathbf{\in}$ $\Re^{N\times N}$
has the simple diagonal form%
\[
\mathbf{Q}_{ij}^{-1}=\left\{
\begin{array}
[c]{c}%
0\text{ if }i\neq j\\
\lambda_{i}^{-1}\text{ if }i=j
\end{array}
\right.  \text{.}%
\]

Thereby, the shape and the fundamental property exhibited by the geometric
locus of any given quadratic surface that is represented by a vector algebra
locus equation that has the form%
\begin{align*}
\lambda_{1}^{-1}x_{1\ast}^{2}\widehat{\mathbf{v}}_{1}^{2}+\mathbf{\ldots
+}\lambda_{N}^{-1}x_{N\ast}^{2}\widehat{\mathbf{v}}_{N}^{2}  &  =%
{\textstyle\sum\nolimits_{i=1}^{N}}
x_{i\ast}^{2}\lambda_{i}^{-1}\left\Vert \widehat{\mathbf{v}}_{i}\right\Vert
^{2}\\
&  =%
{\textstyle\sum\nolimits_{i=1}^{N}}
\lambda_{i}^{-1}\left\Vert x_{i\ast}\widehat{\mathbf{v}}_{i}\right\Vert ^{2}\\
&  =\left\Vert \boldsymbol{\nu}\right\Vert ^{2}\text{,}%
\end{align*}
so that an exclusive principal eigen-coordinate system%
\begin{align*}
\boldsymbol{\nu}  &  =\sqrt{\lambda_{1}^{-1}}x_{1\ast}\widehat{\mathbf{v}}%
_{1}+\mathbf{\ldots+}\sqrt{\lambda_{N}^{-1}}x_{N\ast}\widehat{\mathbf{v}}%
_{N}\\
&  =%
{\textstyle\sum\nolimits_{i=1}^{N}}
\sqrt{\lambda_{i}^{-1}}x_{i\ast}\widehat{\mathbf{v}}_{i}%
\end{align*}
of the quadratic surface satisfies the geometric locus of the quadratic
surface in terms of its total allowed eigenenergy $\left\Vert \boldsymbol{\nu
}\right\Vert ^{2}=%
{\textstyle\sum\nolimits_{i=1}^{N}}
\lambda_{i}^{-1}\left\Vert x_{i\ast}\widehat{\mathbf{v}}_{i}\right\Vert ^{2}$,
are both determined by the total allowed eigenenergy $\left\Vert
\boldsymbol{\nu}\right\Vert ^{2}=%
{\textstyle\sum\nolimits_{i=1}^{N}}
\lambda_{i}^{-1}\left\Vert x_{i\ast}\widehat{\mathbf{v}}_{i}\right\Vert ^{2}$
exhibited by the principal eigenaxis $\boldsymbol{\nu}$ $\boldsymbol{=}%
{\textstyle\sum\nolimits_{i=1}^{N}}
\sqrt{\lambda_{i}^{-1}}x_{i\ast}\widehat{\mathbf{v}}_{i}$ of the geometric
locus of the quadratic surface, such that the eigenvalues $\lambda_{i}^{-1}$
of an $N\times N$ symmetric matrix $\mathbf{Q}^{-1}\mathbf{\in}$ $\Re^{N\times
N}$ of a quadratic form $\mathbf{v}^{T}\mathbf{Q}^{-1}\mathbf{v}$ regulate the
total allowed eigenenergy $\left\Vert \boldsymbol{\nu}\right\Vert ^{2}$
exhibited by the principal eigenaxis $\boldsymbol{\nu}$ of the geometric locus
of the quadratic surface; and the uniform property exhibited by all of the
points that lie on the geometric locus of the quadratic surface is the total
allowed eigenenergy $\left\Vert \boldsymbol{\nu}\right\Vert ^{2}$ exhibited by
the principal eigenaxis $\boldsymbol{\nu}$ of the geometric locus of the
quadratic surface.
\end{corollary}

\begin{proof}
Corollary \ref{Principal Eigen-coordinate System Corollary} is proved by using
Theorem \ref{Principal Eigen-coordinate System Theorem}, wherein the principal
eigenvector $\mathbf{v}=%
{\textstyle\sum\nolimits_{i=1}^{2}}
x_{i\ast}\widehat{\mathbf{v}}_{i}$ of the symmetric matrix $\mathbf{Q}^{-1}$
of the quadratic form $\mathbf{v}^{T}\mathbf{Q}^{-1}\mathbf{v}$ satisfies the
relation%
\[
\mathbf{v}^{T}\mathbf{Q}^{-1}\mathbf{v=}\left(
{\textstyle\sum\nolimits_{i=1}^{2}}
x_{i\ast}\widehat{\mathbf{v}}_{i}\right)  ^{T}\left(
{\textstyle\sum\nolimits_{j=1}^{2}}
\lambda_{j}^{-1}x_{j\ast}\widehat{\mathbf{v}}_{j}\right)
\]
and the principal eigenvector $\mathbf{v}=%
{\textstyle\sum\nolimits_{i=1}^{N}}
x_{i\ast}\widehat{\mathbf{v}}_{i}$ of the symmetric matrix $\mathbf{Q}^{-1}$
of the quadratic form $\mathbf{v}^{T}\mathbf{Q}^{-1}\mathbf{v}$ satisfies the
relation%
\[
\mathbf{v}^{T}\mathbf{Q}^{-1}\mathbf{v=}\left(
{\textstyle\sum\nolimits_{i=1}^{N}}
x_{i\ast}\widehat{\mathbf{v}}_{i}\right)  ^{T}\left(
{\textstyle\sum\nolimits_{j=1}^{N}}
\lambda_{j}^{-1}x_{j\ast}\widehat{\mathbf{v}}_{j}\right)  \text{.}%
\]

\end{proof}

In previous working papers
\citep{Reeves2015resolving}
and
\citep{Reeves2018design}%
, we noted that the shape of any given quadratic surface is completely
determined by the eigenvalues of a symmetric matrix associated with a
quadratic form.

However, the novel principal eigen-coordinate transform method expressed by
Theorem \ref{Principal Eigen-coordinate System Theorem} and Corollary
\ref{Principal Eigen-coordinate System Corollary} reveals that the shape and
the fundamental property exhibited by the geometric locus of any given
quadratic curve or surface are both determined by an exclusive principal
eigen-coordinate system, such that the eigenvalues of a symmetric matrix of a
transformed quadratic form modulate the eigenenergies exhibited by the
components of the exclusive principal eigen-coordinate system---which is the
principal part of an equivalent representation of the quadratic form---at
which point the exclusive principal eigen-coordinate system is the principal
eigenaxis of the geometric locus of the quadratic curve or surface, so that
the principal eigenaxis satisfies the geometric locus of the quadratic curve
or surface in terms of its total allowed eigenenergy; and the uniform property
exhibited by all of the points that lie on the geometric locus of the
quadratic curve or surface is the total allowed eigenenergy exhibited by the
principal eigenaxis of the geometric locus of the quadratic curve or surface.

Most importantly, the conditions expressed by Theorem
\ref{Principal Eigen-coordinate System Theorem} and Corollary
\ref{Principal Eigen-coordinate System Corollary} guarantee us that any given
quadratic form that is the solution of a vector algebra locus equation, such
that the graph of the vector algebra locus equation represents a certain
quadratic curve or surface, can be represented by an exclusive principal
eigen-coordinate system, so that the exclusive principal eigen-coordinate
system is the solution of an equivalent form of the vector algebra locus
equation of the quadratic curve or surface, such that the eigenenergies
exhibited by the components of the exclusive principal eigen-coordinate system
are modulated by the eigenvalues of the symmetric matrix of the
\emph{transformed} quadratic form, so that the principal eigenaxis of the
geometric locus of the quadratic curve or surface satisfies the geometric
locus of the quadratic curve or surface in terms of its \emph{total allowed
eigenenergy}.

Equally important, the conditions expressed by Corollary
\ref{Symmetrical and Equivalent Principal Eigenaxes} guarantee us of the
existence of a pair of exclusive principal eigen-coordinate systems that are
symmetrically and equivalently related to each other---such that a pair of
principal eigenaxes are principal parts of equivalent representations of
correlated quadratic forms---so that the pair of principal eigenaxes exhibit
symmetrical and equivalent total allowed eigenenergies.

\subsection{Symmetrical and Equivalent Principal Eigenaxes}

Corollary \ref{Symmetrical and Equivalent Principal Eigenaxes} guarantees the
existence of a pair of exclusive principal eigen-coordinate systems that are
symmetrically and equivalently related to each other---each of which is the
principal part of an equivalent representation of a correlated quadratic form
$\mathbf{x}^{T}\mathbf{Qx}$ or $\mathbf{x}^{T}\mathbf{Q}^{-1}\mathbf{x}%
$---such that each principal eigenaxis of a certain quadratic curve or surface
is the solution of an equivalent form of the vector algebra locus equation of
the quadratic curve or surface---so that the pair of principal eigenaxes
exhibit symmetrical and equivalent total allowed eigenenergies.

\begin{corollary}
\label{Symmetrical and Equivalent Principal Eigenaxes}\bigskip Take any given
pair of quadratic forms $\mathbf{x}^{T}\mathbf{Qx}$ and $\mathbf{x}%
^{T}\mathbf{Q}^{-1}\mathbf{x}$, such that the matrix $\mathbf{Q}%
^{-1}\mathbf{\in}$ $\Re^{2\times2}$ is the inverse of the matrix
$\mathbf{Q\in}$ $\Re^{2\times2}$, so that the elements of the symmetric
matrices $\mathbf{Q}$ and $\mathbf{Q}^{-1}\mathbf{\ }$contain similar information.

Let the quadratic form $\mathbf{x}^{T}\mathbf{Qx}$ be the solution of a vector
algebra locus equation of a certain quadratic curve that has the form%
\[
\mathbf{x}^{T}\mathbf{Qx}=c_{1}\text{,}%
\]
where $c_{1}$ is a certain constant.

In addition, let the quadratic form $\mathbf{x}^{T}\mathbf{Q}^{-1}\mathbf{x}$
be the solution of a vector algebra locus equation of a similar quadratic
curve that has the form%
\[
\mathbf{x}^{T}\mathbf{Q}^{-1}\mathbf{x}=c_{2}\text{,}%
\]
where $c_{2}$ is a certain constant.

Now, let an equivalent form of each vector algebra locus equation
$\mathbf{x}^{T}\mathbf{Qx}=c_{1}$ and $\mathbf{x}^{T}\mathbf{Q}^{-1}%
\mathbf{x}=c_{2}$ be generated by transforming the positions of the coordinate
axes of each quadratic curve into the axes of an exclusive principal
eigen-coordinate system, so that the principal eigenaxis of each quadratic
curve satisfies the geometric locus of the quadratic curve in terms of its
total allowed eigenenergy.

It follows that an exclusive principal eigen-coordinate system%
\[
\boldsymbol{\nu}_{1}=%
{\textstyle\sum\nolimits_{i=1}^{2}}
\sqrt{\lambda_{i}}x_{i\ast}\widehat{\mathbf{v}}_{i}%
\]
is the principal part of an equivalent representation of the quadratic form
$\mathbf{x}^{T}\mathbf{Qx}$ in the following manner%
\begin{align*}
\left(
{\textstyle\sum\nolimits_{i=1}^{2}}
x_{i\ast}\widehat{\mathbf{v}}_{i}\right)  ^{T}\mathbf{Q}\left(
{\textstyle\sum\nolimits_{j=1}^{2}}
x_{j\ast}\widehat{\mathbf{v}}_{j}\right)   &  =\mathbf{v}^{T}\mathbf{Qv}\\
&  =%
{\textstyle\sum\nolimits_{i=1}^{2}}
\left(  \sqrt{\lambda_{i}}x_{i\ast}\widehat{\mathbf{v}}_{i}^{T}\right)
\left(  \sqrt{\lambda_{i}}x_{i\ast}\widehat{\mathbf{v}}_{i}\right) \\
&  =%
{\textstyle\sum\nolimits_{i=1}^{2}}
\lambda_{i}\left\Vert x_{i\ast}\widehat{\mathbf{v}}_{i}\right\Vert ^{2}\\
&  =\left\Vert \boldsymbol{\nu}_{1}\right\Vert ^{2}\text{,}%
\end{align*}
wherein the vector $\mathbf{x=}%
{\textstyle\sum\nolimits_{i=1}^{2}}
x_{i}\mathbf{e}_{i}$ is transformed into the principal eigenvector
$\mathbf{v}=%
{\textstyle\sum\nolimits_{i=1}^{2}}
x_{i\ast}\widehat{\mathbf{v}}_{i}$ of the symmetric matrix $\mathbf{Q\in}$
$\Re^{2\times2}$ of the quadratic form $\mathbf{v}^{T}\mathbf{Qv}$, such that
the principal eigenvector $\mathbf{v}=%
{\textstyle\sum\nolimits_{i=1}^{2}}
x_{i\ast}\widehat{\mathbf{v}}_{i}$ of the symmetric matrix $\mathbf{Q}$ of the
quadratic form $\mathbf{v}^{T}\mathbf{Qv}$ is symmetrically and equivalently
related to the principal eigenaxis $\boldsymbol{\nu}_{1}=%
{\textstyle\sum\nolimits_{i=1}^{2}}
\sqrt{\lambda_{i}}x_{i\ast}\widehat{\mathbf{v}}_{i}$ of a certain quadratic
curve, so that the exclusive principal eigen-coordinate system
$\boldsymbol{\nu}_{1}=%
{\textstyle\sum\nolimits_{i=1}^{2}}
\sqrt{\lambda_{i}}x_{i\ast}\widehat{\mathbf{v}}_{i}$ satisfies the geometric
locus of the quadratic curve in terms of its total allowed eigenenergy
$\left\Vert \boldsymbol{\nu}_{1}\right\Vert ^{2}$, such that the eigenenergy
$\lambda_{i}\left\Vert x_{i\ast}\widehat{\mathbf{v}}_{i}\right\Vert ^{2}$
exhibited by each component $\sqrt{\lambda_{i}}x_{i\ast}\widehat{\mathbf{v}%
}_{i}$ of the principal eigenaxis $\boldsymbol{\nu}_{1}=\sqrt{\lambda_{1}%
}x_{1\ast}\widehat{\mathbf{v}}_{1}+\sqrt{\lambda_{2}}x_{2\ast}%
\widehat{\mathbf{v}}_{2}$ of the geometric locus of the quadratic curve is
modulated by an eigenvalue $\lambda_{i}$ of the symmetric matrix $\mathbf{Q}$
of the quadratic form $\mathbf{v}^{T}\mathbf{Qv}$ in a manner that regulates
the total allowed eigenenergy $\left\Vert \boldsymbol{\nu}_{1}\right\Vert
^{2}=%
{\textstyle\sum\nolimits_{i=1}^{2}}
\lambda_{i}\left\Vert x_{i\ast}\widehat{\mathbf{v}}_{i}\right\Vert ^{2}$
exhibited by the geometric locus of the principal eigenaxis $\boldsymbol{\nu
}_{1}=%
{\textstyle\sum\nolimits_{i=1}^{2}}
\sqrt{\lambda_{i}}x_{i\ast}\widehat{\mathbf{v}}_{i}$, wherein the $2\times2$
symmetric matrix $\mathbf{Q\in}$ $\Re^{2\times2}$ has the simple diagonal form%
\[
\mathbf{Q}_{ij}=\left\{
\begin{array}
[c]{c}%
0\text{ if }i\neq j\\
\lambda_{i}\text{ if }i=j
\end{array}
\right.  \text{.}%
\]

It also follows that an exclusive principal eigen-coordinate system%
\[
\boldsymbol{\nu}_{2}=%
{\textstyle\sum\nolimits_{i=1}^{2}}
\sqrt{\lambda_{i}^{-1}}x_{i\ast}\widehat{\mathbf{v}}_{i}%
\]
is the principal part of an equivalent representation of the quadratic form
$\mathbf{x}^{T}\mathbf{Q}^{-1}\mathbf{x}$ in the following manner%
\begin{align*}
\left(
{\textstyle\sum\nolimits_{i=1}^{2}}
x_{i\ast}\widehat{\mathbf{v}}_{i}\right)  ^{T}\mathbf{Q}^{-1}\left(
{\textstyle\sum\nolimits_{j=1}^{2}}
x_{j\ast}\widehat{\mathbf{v}}_{j}\right)   &  =\mathbf{v}^{T}\mathbf{Q}%
^{-1}\mathbf{v}\\
&  =%
{\textstyle\sum\nolimits_{i=1}^{2}}
\left(  \sqrt{\lambda_{i}^{-1}}x_{i\ast}\widehat{\mathbf{v}}_{i}^{T}\right)
\left(  \sqrt{\lambda_{i}^{-1}}x_{i\ast}\widehat{\mathbf{v}}_{i}\right) \\
&  =%
{\textstyle\sum\nolimits_{i=1}^{2}}
\lambda_{i}^{-1}\left\Vert x_{i\ast}\widehat{\mathbf{v}}_{i}\right\Vert ^{2}\\
&  =\left\Vert \boldsymbol{\nu}_{2}\right\Vert ^{2}\text{,}%
\end{align*}
wherein the vector $\mathbf{x=}%
{\textstyle\sum\nolimits_{i=1}^{2}}
x_{i}\mathbf{e}_{i}$ is transformed into the principal eigenvector
$\mathbf{v}=%
{\textstyle\sum\nolimits_{i=1}^{2}}
x_{i\ast}\widehat{\mathbf{v}}_{i}$ of the symmetric matrix $\mathbf{Q}%
^{-1}\mathbf{\in}$ $\Re^{2\times2}$ of the quadratic form $\mathbf{v}%
^{T}\mathbf{Q}^{-1}\mathbf{v}$, such that the principal eigenvector
$\mathbf{v}=%
{\textstyle\sum\nolimits_{i=1}^{2}}
x_{i\ast}\widehat{\mathbf{v}}_{i}$ of the symmetric matrix $\mathbf{Q}^{-1}$
of the quadratic form $\mathbf{v}^{T}\mathbf{Q}^{-1}\mathbf{v}$ is
symmetrically and equivalently related to the principal eigenaxis
$\boldsymbol{\nu}_{2}=%
{\textstyle\sum\nolimits_{i=1}^{2}}
\sqrt{\lambda_{i}^{-1}}x_{i\ast}\widehat{\mathbf{v}}_{i}$ of a certain
quadratic curve, so that the exclusive principal eigen-coordinate system
$\boldsymbol{\nu}_{1}=%
{\textstyle\sum\nolimits_{i=1}^{2}}
\sqrt{\lambda_{i}}x_{i\ast}\widehat{\mathbf{v}}_{i}$ satisfies the geometric
locus of the quadratic curve in terms of its total allowed eigenenergy
$\left\Vert \boldsymbol{\nu}_{1}\right\Vert ^{2}$, such that the eigenenergy
$\lambda_{i}^{-1}\left\Vert x_{i\ast}\widehat{\mathbf{v}}_{i}\right\Vert ^{2}$
exhibited by each component $\sqrt{\lambda_{i}^{-1}}x_{i\ast}%
\widehat{\mathbf{v}}_{i}$ of the principal eigenaxis $\boldsymbol{\nu}%
_{2}=\sqrt{\lambda_{1}^{-1}}x_{1\ast}\widehat{\mathbf{v}}_{1}+\sqrt
{\lambda_{2}^{-1}}x_{2\ast}\widehat{\mathbf{v}}_{2}$ of the geometric locus of
the quadratic curve is modulated by an eigenvalue $\lambda_{i}^{-1}$ of the
symmetric matrix $\mathbf{Q}^{-1}$ of the quadratic form $\mathbf{v}%
^{T}\mathbf{Q}^{-1}\mathbf{v}$ in a manner that regulates the total allowed
eigenenergy $\left\Vert \boldsymbol{\nu}_{2}\right\Vert ^{2}=%
{\textstyle\sum\nolimits_{i=1}^{2}}
\lambda_{i}^{-1}\left\Vert x_{i\ast}\widehat{\mathbf{v}}_{i}\right\Vert ^{2}$
exhibited by the geometric locus of the principal eigenaxis $\boldsymbol{\nu
}_{2}=%
{\textstyle\sum\nolimits_{i=1}^{2}}
\sqrt{\lambda_{i}^{-1}}x_{i\ast}\widehat{\mathbf{v}}_{i}$, wherein the
$2\times2$ symmetric matrix $\mathbf{Q}^{-1}\mathbf{\in}$ $\Re^{2\times2}$ has
the simple diagonal form
\[
\mathbf{Q}_{ij}^{-1}=\left\{
\begin{array}
[c]{c}%
0\text{ if }i\neq j\\
\lambda_{i}^{-1}\text{ if }i=j
\end{array}
\right.  \text{.}%
\]

Thereby, the principal eigenaxis $\boldsymbol{\nu}_{1}=%
{\textstyle\sum\nolimits_{i=1}^{2}}
\sqrt{\lambda_{i}}x_{i\ast}\widehat{\mathbf{v}}_{i}$ of a certain quadratic
curve is symmetrically and equivalently related to the principal eigenaxis
$\boldsymbol{\nu}_{2}=%
{\textstyle\sum\nolimits_{i=1}^{2}}
\sqrt{\lambda_{i}^{-1}}x_{i\ast}\widehat{\mathbf{v}}_{i}$ of a similar
quadratic curve, at which point both of the principal eigenaxes
$\boldsymbol{\nu}_{1}=%
{\textstyle\sum\nolimits_{i=1}^{2}}
\sqrt{\lambda_{i}}x_{i\ast}\widehat{\mathbf{v}}_{i}$ and $\boldsymbol{\nu}%
_{2}=%
{\textstyle\sum\nolimits_{i=1}^{2}}
\sqrt{\lambda_{i}^{-1}}x_{i\ast}\widehat{\mathbf{v}}_{i}$ of the quadratic
curves are symmetrically and equivalently related to the principal eigenvector
$\mathbf{v}=%
{\textstyle\sum\nolimits_{i=1}^{2}}
x_{i\ast}\widehat{\mathbf{v}}_{i}$ \ of the symmetric matrices $\mathbf{Q\in}$
$\Re^{2\times2}$ and $\mathbf{Q}^{-1}\mathbf{\in}$ $\Re^{2\times2}$ of the
quadratic forms $\mathbf{v}^{T}\mathbf{Qv}$ and $\mathbf{v}^{T}\mathbf{Q}%
^{-1}\mathbf{v}$, so that the total allowed eigenenergy $\left\Vert
\boldsymbol{\nu}_{1}\right\Vert ^{2}=%
{\textstyle\sum\nolimits_{i=1}^{2}}
\lambda_{i}\left\Vert x_{i\ast}\widehat{\mathbf{v}}_{i}\right\Vert ^{2}$
exhibited by the geometric locus of the principal eigenaxis $\boldsymbol{\nu
}_{1}$ is symmetrically and equivalently related to the total allowed
eigenenergy $\left\Vert \boldsymbol{\nu}_{2}\right\Vert ^{2}=%
{\textstyle\sum\nolimits_{i=1}^{2}}
\lambda_{i}^{-1}\left\Vert x_{i\ast}\widehat{\mathbf{v}}_{i}\right\Vert ^{2}$
exhibited by the geometric locus of the principal eigenaxis $\boldsymbol{\nu
}_{2}$, wherein%
\[%
{\textstyle\sum\nolimits_{i=1}^{2}}
\lambda_{i}\left\Vert x_{i\ast}\widehat{\mathbf{v}}_{i}\right\Vert ^{2}\equiv%
{\textstyle\sum\nolimits_{i=1}^{2}}
\lambda_{i}^{-1}\left\Vert x_{i\ast}\widehat{\mathbf{v}}_{i}\right\Vert ^{2}%
\]
since%
\[
\left\Vert \boldsymbol{\nu}_{1}\right\Vert ^{2}\equiv\left\Vert
\boldsymbol{\nu}_{2}\right\Vert ^{2}\text{.}%
\]

Corresponding, take any given quadratic forms $\mathbf{x}^{T}\mathbf{Qx}$ and
$\mathbf{x}^{T}\mathbf{Q}^{-1}\mathbf{x}$, such that the matrix $\mathbf{Q}%
^{-1}\mathbf{\in}$ $\Re^{N\times N}$ is the inverse of the matrix
$\mathbf{Q\in}$ $\Re^{N\times N}$, so that the elements of the symmetric
matrices $\mathbf{Q}$ and $\mathbf{Q}^{-1}\mathbf{\ }$contain similar information.

Let the quadratic form $\mathbf{x}^{T}\mathbf{Qx}$ be the solution of a vector
algebra locus equation of a certain quadratic surface that has the form%
\[
\mathbf{x}^{T}\mathbf{Qx}=c_{1}\text{,}%
\]
where $c_{1}$ is a certain constant.

In addition, let the quadratic form $\mathbf{x}^{T}\mathbf{Q}^{-1}\mathbf{x}$
be the solution of a vector algebra locus equation of a similar quadratic
surface that has the form%
\[
\mathbf{x}^{T}\mathbf{Q}^{-1}\mathbf{x}=c_{2}\text{,}%
\]
where $c_{2}$ is a certain constant.

Now, let an equivalent form of each vector algebra locus equation
$\mathbf{x}^{T}\mathbf{Qx}=c_{1}$ and $\mathbf{x}^{T}\mathbf{Q}^{-1}%
\mathbf{x}=c_{2}$ be generated by transforming the positions of the coordinate
axes of each quadratic surface into the axes of an exclusive principal
eigen-coordinate system, so that the principal eigenaxis of each quadratic
surface satisfies the geometric locus of the quadratic surface in terms of its
total allowed eigenenergy.

It follows that an exclusive principal eigen-coordinate system%
\[
\boldsymbol{\nu}_{1}=%
{\textstyle\sum\nolimits_{i=1}^{N}}
\sqrt{\lambda_{i}}x_{i\ast}\widehat{\mathbf{v}}_{i}%
\]
is the principal part of an equivalent representation of the quadratic form
$\mathbf{x}^{T}\mathbf{Qx}$ in the following manner%
\begin{align*}
\left(
{\textstyle\sum\nolimits_{i=1}^{N}}
x_{i\ast}\widehat{\mathbf{v}}_{i}\right)  ^{T}\mathbf{Q}\left(
{\textstyle\sum\nolimits_{j=1}^{N}}
x_{j\ast}\widehat{\mathbf{v}}_{j}\right)   &  =\mathbf{v}^{T}\mathbf{Qv}\\
&  =%
{\textstyle\sum\nolimits_{i=1}^{N}}
\left(  \sqrt{\lambda_{i}}x_{i\ast}\widehat{\mathbf{v}}_{i}^{T}\right)
\left(  \sqrt{\lambda_{i}}x_{i\ast}\widehat{\mathbf{v}}_{i}\right) \\
&  =%
{\textstyle\sum\nolimits_{i=1}^{N}}
\lambda_{i}\left\Vert x_{i\ast}\widehat{\mathbf{v}}_{i}\right\Vert ^{2}\\
&  =\left\Vert \boldsymbol{\nu}_{1}\right\Vert ^{2}\text{,}%
\end{align*}
wherein the vector $\mathbf{x=}%
{\textstyle\sum\nolimits_{i=1}^{N}}
x_{i}\mathbf{e}_{i}$ is transformed into the principal eigenvector
$\mathbf{v}=%
{\textstyle\sum\nolimits_{i=1}^{N}}
x_{i\ast}\widehat{\mathbf{v}}_{i}$ of the symmetric matrix $\mathbf{Q\in}$
$\Re^{N\times N}$ of the quadratic form $\mathbf{v}^{T}\mathbf{Qv}$, such that
the principal eigenvector $\mathbf{v}=%
{\textstyle\sum\nolimits_{i=1}^{N}}
x_{i\ast}\widehat{\mathbf{v}}_{i}$ of the symmetric matrix $\mathbf{Q}$ of the
quadratic form $\mathbf{v}^{T}\mathbf{Qv}$ is symmetrically and equivalently
related to the principal eigenaxis $\boldsymbol{\nu}_{1}=%
{\textstyle\sum\nolimits_{i=1}^{N}}
\sqrt{\lambda_{i}}x_{i\ast}\widehat{\mathbf{v}}_{i}$ of a certain quadratic
surface, so that the exclusive principal eigen-coordinate system
$\boldsymbol{\nu}_{1}=%
{\textstyle\sum\nolimits_{i=1}^{N}}
\sqrt{\lambda_{i}}x_{i\ast}\widehat{\mathbf{v}}_{i}$ satisfies the geometric
locus of the quadratic surface in terms of its total allowed eigenenergy
$\left\Vert \boldsymbol{\nu}_{1}\right\Vert ^{2}$, such that the eigenenergy
$\lambda_{i}\left\Vert x_{i\ast}\widehat{\mathbf{v}}_{i}\right\Vert ^{2}$
exhibited by each component $\sqrt{\lambda_{i}}x_{i\ast}\widehat{\mathbf{v}%
}_{i}$ of the principal eigenaxis $\boldsymbol{\nu}_{1}=%
{\textstyle\sum\nolimits_{i=1}^{N}}
\sqrt{\lambda_{i}}x_{i\ast}\widehat{\mathbf{v}}_{i}$ of the geometric locus of
the quadratic surface is modulated by an eigenvalue $\lambda_{i}$ of the
symmetric matrix $\mathbf{Q}$ of the quadratic form $\mathbf{v}^{T}%
\mathbf{Qv}$ in a manner that regulates the total allowed eigenenergy
$\left\Vert \boldsymbol{\nu}_{1}\right\Vert ^{2}=%
{\textstyle\sum\nolimits_{i=1}^{N}}
\lambda_{i}\left\Vert x_{i\ast}\widehat{\mathbf{v}}_{i}\right\Vert ^{2}$
exhibited by the geometric locus of the principal eigenaxis $\boldsymbol{\nu
}_{1}=%
{\textstyle\sum\nolimits_{i=1}^{N}}
\sqrt{\lambda_{i}}x_{i\ast}\widehat{\mathbf{v}}_{i}$, wherein the $N\times N$
symmetric matrix $\mathbf{Q\in}$ $\Re^{N\times N}$ has the simple diagonal
form%
\[
\mathbf{Q}_{ij}=\left\{
\begin{array}
[c]{c}%
0\text{ if }i\neq j\\
\lambda_{i}\text{ if }i=j
\end{array}
\right.  \text{.}%
\]

It also follows that an exclusive principal eigen-coordinate system%
\[
\boldsymbol{\nu}_{2}=%
{\textstyle\sum\nolimits_{i=1}^{N}}
\sqrt{\lambda_{i}^{-1}}x_{i\ast}\widehat{\mathbf{v}}_{i}%
\]
is the principal part of an equivalent representation of the quadratic form
$\mathbf{x}^{T}\mathbf{Q}^{-1}\mathbf{x}$ in the following manner%
\begin{align*}
\left(
{\textstyle\sum\nolimits_{i=1}^{N}}
x_{i\ast}\widehat{\mathbf{v}}_{i}\right)  ^{T}\mathbf{Q}^{-1}\left(
{\textstyle\sum\nolimits_{j=1}^{N}}
x_{j\ast}\widehat{\mathbf{v}}_{j}\right)   &  =\mathbf{v}^{T}\mathbf{Q}%
^{-1}\mathbf{v}\\
&  =%
{\textstyle\sum\nolimits_{i=1}^{N}}
\left(  \sqrt{\lambda_{i}^{-1}}x_{i\ast}\widehat{\mathbf{v}}_{i}^{T}\right)
\left(  \sqrt{\lambda_{i}^{-1}}x_{i\ast}\widehat{\mathbf{v}}_{i}\right) \\
&  =%
{\textstyle\sum\nolimits_{i=1}^{N}}
\lambda_{i}^{-1}\left\Vert x_{i\ast}\widehat{\mathbf{v}}_{i}\right\Vert ^{2}\\
&  =\left\Vert \boldsymbol{\nu}_{2}\right\Vert ^{2}\text{,}%
\end{align*}
wherein the vector $\mathbf{x=}%
{\textstyle\sum\nolimits_{i=1}^{N}}
x_{i}\mathbf{e}_{i}$ is transformed into the principal eigenvector
$\mathbf{v}=%
{\textstyle\sum\nolimits_{i=1}^{N}}
x_{i\ast}\widehat{\mathbf{v}}_{i}$ of the symmetric matrix $\mathbf{Q}%
^{-1}\mathbf{\in}$ $\Re^{N\times N}$ of the quadratic form $\mathbf{v}%
^{T}\mathbf{Q}^{-1}\mathbf{v}$, such that the principal eigenvector
$\mathbf{v}=%
{\textstyle\sum\nolimits_{i=1}^{N}}
x_{i\ast}\widehat{\mathbf{v}}_{i}$ of the symmetric matrix $\mathbf{Q}^{-1}$
of the quadratic form $\mathbf{v}^{T}\mathbf{Q}^{-1}\mathbf{v}$ is
symmetrically and equivalently related to the principal eigenaxis
$\boldsymbol{\nu}_{2}=%
{\textstyle\sum\nolimits_{i=1}^{N}}
\sqrt{\lambda_{i}^{-1}}x_{i\ast}\widehat{\mathbf{v}}_{i}$ of a certain
quadratic surface, so that the exclusive principal eigen-coordinate system
$\boldsymbol{\nu}_{2}=%
{\textstyle\sum\nolimits_{i=1}^{N}}
\sqrt{\lambda_{i}^{-1}}x_{i\ast}\widehat{\mathbf{v}}_{i}$ satisfies the
geometric locus of the quadratic surface in terms of its total allowed
eigenenergy $\left\Vert \boldsymbol{\nu}_{2}\right\Vert ^{2}$, such that the
eigenenergy $\lambda_{i}^{-1}\left\Vert x_{i\ast}\widehat{\mathbf{v}}%
_{i}\right\Vert ^{2}$ exhibited by each component $\sqrt{\lambda_{i}^{-1}%
}x_{i\ast}\widehat{\mathbf{v}}_{i}$ of the principal eigenaxis
$\boldsymbol{\nu}_{2}=%
{\textstyle\sum\nolimits_{i=1}^{N}}
\sqrt{\lambda_{i}^{-1}}x_{i\ast}\widehat{\mathbf{v}}_{i}$ of the geometric
locus of the quadratic surface is modulated by an eigenvalue $\lambda_{i}%
^{-1}$ of the symmetric matrix $\mathbf{Q}^{-1}$ of the quadratic form
$\mathbf{v}^{T}\mathbf{Q}^{-1}\mathbf{v}$ in a manner that regulates the total
allowed eigenenergy $\left\Vert \boldsymbol{\nu}_{2}\right\Vert ^{2}=%
{\textstyle\sum\nolimits_{i=1}^{N}}
\lambda_{i}^{-1}\left\Vert x_{i\ast}\widehat{\mathbf{v}}_{i}\right\Vert ^{2}$
exhibited by the geometric locus of the principal eigenaxis $\boldsymbol{\nu
}_{2}=%
{\textstyle\sum\nolimits_{i=1}^{2}}
\sqrt{\lambda_{i}^{-1}}x_{i\ast}\widehat{\mathbf{v}}_{i}$, wherein the
$N\times N$ symmetric matrix $\mathbf{Q}^{-1}\mathbf{\in}$ $\Re^{N\times N}$
has the simple diagonal form%
\[
\mathbf{Q}_{ij}^{-1}=\left\{
\begin{array}
[c]{c}%
0\text{ if }i\neq j\\
\lambda_{i}^{-1}\text{ if }i=j
\end{array}
\right.  \text{.}%
\]

Thereby, the principal eigenaxis $\boldsymbol{\nu}_{1}=%
{\textstyle\sum\nolimits_{i=1}^{N}}
\sqrt{\lambda_{i}}x_{i\ast}\widehat{\mathbf{v}}_{i}$ of a certain quadratic
surface is symmetrically and equivalently related to the principal eigenaxis
$\boldsymbol{\nu}_{2}=%
{\textstyle\sum\nolimits_{i=1}^{N}}
\sqrt{\lambda_{i}^{-1}}x_{i\ast}\widehat{\mathbf{v}}_{i}$ of a similar
quadratic surface, such that both of the principal eigenaxes $\boldsymbol{\nu
}_{1}=%
{\textstyle\sum\nolimits_{i=1}^{N}}
\sqrt{\lambda_{i}}x_{i\ast}\widehat{\mathbf{v}}_{i}$ and $\boldsymbol{\nu}%
_{2}=%
{\textstyle\sum\nolimits_{i=1}^{N}}
\sqrt{\lambda_{i}^{-1}}x_{i\ast}\widehat{\mathbf{v}}_{i}$ of the quadratic
surfaces are symmetrically and equivalently related to the principal
eigenvector $\mathbf{v}=%
{\textstyle\sum\nolimits_{i=1}^{N}}
x_{i\ast}\widehat{\mathbf{v}}_{i}$ \ of the symmetric matrices $\mathbf{Q\in}$
$\Re^{N\times N}$ and $\mathbf{Q}^{-1}\mathbf{\in}$ $\Re^{N\times N}$ of the
quadratic forms $\mathbf{v}^{T}\mathbf{Qv}$ and $\mathbf{v}^{T}\mathbf{Q}%
^{-1}\mathbf{v}$, so that the total allowed eigenenergy $\left\Vert
\boldsymbol{\nu}_{1}\right\Vert ^{2}=%
{\textstyle\sum\nolimits_{i=1}^{N}}
\lambda_{i}\left\Vert x_{i\ast}\widehat{\mathbf{v}}_{i}\right\Vert ^{2}$
exhibited by the geometric locus of the principal eigenaxis $\boldsymbol{\nu
}_{1}$ is symmetrically and equivalently related to the total allowed
eigenenergy $\left\Vert \boldsymbol{\nu}_{2}\right\Vert ^{2}=%
{\textstyle\sum\nolimits_{i=1}^{N}}
\lambda_{i}^{-1}\left\Vert x_{i\ast}\widehat{\mathbf{v}}_{i}\right\Vert ^{2}$
exhibited by the geometric locus of the principal eigenaxis $\boldsymbol{\nu
}_{2}$, wherein%
\[%
{\textstyle\sum\nolimits_{i=1}^{N}}
\lambda_{i}\left\Vert x_{i\ast}\widehat{\mathbf{v}}_{i}\right\Vert ^{2}\equiv%
{\textstyle\sum\nolimits_{i=1}^{N}}
\lambda_{i}^{-1}\left\Vert x_{i\ast}\widehat{\mathbf{v}}_{i}\right\Vert ^{2}%
\]
since%
\[
\left\Vert \boldsymbol{\nu}_{1}\right\Vert ^{2}\equiv\left\Vert
\boldsymbol{\nu}_{2}\right\Vert ^{2}\text{.}%
\]

\end{corollary}

\begin{proof}
Corollary \ref{Symmetrical and Equivalent Principal Eigenaxes} is proved by
using conditions expressed in Theorem
\ref{Principal Eigen-coordinate System Theorem} and Corollary
\ref{Principal Eigen-coordinate System Corollary}.
\end{proof}

It will be seen that the guarantees provided by Theorem
\ref{Principal Eigen-coordinate System Theorem} and Corollaries
\ref{Principal Eigen-coordinate System Corollary} -
\ref{Symmetrical and Equivalent Principal Eigenaxes} have \emph{far reaching
consequences} for resolving the inverse problem of the binary classification
of random vectors.

Thereby, it will be seen that finding the discriminant function of a minimum
risk binary classification system is a novel geometric locus problem---that
involves finding the geometric locus of the novel principal eigenaxis of the
system---which is structured as a dual locus of likelihood components and
principal eigenaxis components.

We now consider the algebraic and geometrical significance of reproducing
kernels---which are seen to be fundamental components of minimum risk binary
classification systems.

\section{\label{Section 8}Significance of Reproducing Kernels}

It is widely believed that reproducing kernels map any given collection of
feature vectors into a \emph{higher dimensional} feature space, so that
\emph{distances}---between all of the feature vectors---\emph{are increased}
in some manner.

Transforming a collection of feature vectors in this manner is said to make
the collection of feature vectors \textquotedblleft linearly
separable,\textquotedblright\ such that \emph{overlapping} distributions of
feature vectors are transformed into \emph{nonoverlapping} distributions of
feature vectors. The outcome of such mappings is known as the
\textquotedblleft kernel trick.\textquotedblright\ For example, support vector
learning machines use the kernel trick to map training data into higher
dimensional feature spaces, where separating hyperplanes can be found
\citep{Bennett2000,Boser1992,Cortes1995,Cristianini2000,Scholkopf2002}%
.

We realize that reproducing kernels for points map coordinates of vectors into
higher dimensional \emph{coordinate spaces}, so that the algebraic and
geometric structures of the \emph{point coordinates} of the vectors are
\emph{enlarged}. \emph{Distances} between the vectors, however, are \emph{not
increased}. So, why are reproducing kernels important? We now consider the
significance of reproducing kernels.

Let a Hilbert space $\mathcal{H}$ be a reproducing kernel Hilbert space (RKHS)
that is defined on vectors $\mathbf{x\in%
\mathbb{R}
}^{d}$, so that the Hilbert space $\mathcal{H}$ has a certain reproducing
kernel $k_{\mathbf{x}}\left(  \mathbf{s}\right)  $. Given $\mathcal{H}$, take
any given vector $\mathbf{x\in%
\mathbb{R}
}^{d}$. Then there exists a unique vector $k_{\mathbf{x}}\in$ $\mathcal{H}$
that is called the reproducing kernel for the point $\mathbf{x}$, where the
$2$-variable function $k_{\mathbf{x}}\left(  \mathbf{s}\right)  =K\left(
\mathbf{s,x}\right)  $ is called the reproducing kernel for $\mathcal{H}$
\citep{Aronszajn1950}%
.

We recognize any given reproducing kernel $K\left(  \mathbf{s,x}\right)  $ for
a Hilbert space $\mathcal{H}$ determines the algebraic structure of an inner
product relationship between any given vectors $k_{\mathbf{x}}\in$
$\mathcal{H}$ and $k_{\mathbf{s}}\in$ $\mathcal{H}$%
\begin{align*}
K\left(  \mathbf{s,x}\right)   &  =k_{\mathbf{x}}\left(  \mathbf{s}\right)
=\left\langle k_{\mathbf{x}}\left(  \mathbf{s}\right)  ,k_{\mathbf{s}}\left(
\mathbf{x}\right)  \right\rangle \\
&  =\left\langle K\left(  \mathbf{.,x}\right)  ,K\left(  \mathbf{.,s}\right)
\right\rangle =\left\langle K\left(  \mathbf{.,s}\right)  ,K\left(
\mathbf{.,x}\right)  \right\rangle
\end{align*}
in the RKHS $\mathcal{H}$, where $k_{\mathbf{x}}\in$ $\mathcal{H}$,
$k_{\mathbf{s}}\in$ $\mathcal{H}$, $\mathbf{x\in%
\mathbb{R}
}^{d}$, $\mathbf{s\in%
\mathbb{R}
}^{d}$ and $K\left(  \mathbf{x,s}\right)  =K\left(  \mathbf{s,x}\right)  $
\citep{Small1994}%
.

\subsection{Customized Inner Product Relationships}

We realize that reproducing kernels $k_{\mathbf{x}}\in$ $\mathcal{H}$ for
points $\mathbf{x}$ determine enhanced vectors $k_{\mathbf{x}}\mathbf{\ }$that
are the basis of \emph{customized inner product relationships} for machine
learning and data-driven modeling applications, such that any given inner
product relationship%
\[
K\left(  \mathbf{s,x}\right)  =k_{\mathbf{x}}\left(  \mathbf{s}\right)
=\left\langle k_{\mathbf{x}}\left(  \mathbf{s}\right)  ,k_{\mathbf{s}}\left(
\mathbf{x}\right)  \right\rangle
\]
between vectors $k_{\mathbf{x}}\in$ $\mathcal{H}$ and $k_{\mathbf{s}}\in$
$\mathcal{H}$ in any given RKHS $\mathcal{H}$ \emph{enlarges} the algebraic
and geometric structures of point coordinates of vectors $\mathbf{s\in%
\mathbb{R}
}^{d}$\ and $\mathbf{x\in%
\mathbb{R}
}^{d}$ in Hilbert space $\mathcal{H}$.

\subsection{Utility of Reproducing Kernels}

Practically speaking, reproducing kernels $k_{\mathbf{x}}\left(
\mathbf{s}\right)  =K\left(  \mathbf{s,x}\right)  $ replace straight line
segments of vectors $\mathbf{x}$ with curves, such that vectors $\mathbf{x}$
and corresponding points $\mathbf{x}$ contain first degree components $x_{i}$,
second degree components $x_{i}^{2}$, third degree\emph{\ }components
$x_{i}^{3}$, and up to $\emph{d}$\emph{\ }degree\emph{\ }components $x_{i}%
^{d}$, where the highest degree $d$ exhibited by the components in any given
vector $k_{\mathbf{x}}\in$ $\mathcal{H}$ is a function of the reproducing
kernel $K\left(  \mathbf{s,x}\right)  $.

Moreover, we have determined that certain types of reproducing kernels replace
vectors with second-order curves---which are formed by first and second degree
vector components---that are more or less sinuous and thereby preserve
topological properties of vectors in Hilbert space $\mathcal{H}$
\citep{Rapport1963,Reeves2018design}%
.

Accordingly, given the conditions expressed by Theorems
\ref{Vector Algebra Equation of Quadratic Loci Theorem} -
\ref{Vector Algebra Equation of Spherical Loci Theorem}, Theorem
\ref{Principal Eigen-coordinate System Theorem} and Corollary
\ref{Principal Eigen-coordinate System Corollary}, we realize that reproducing
kernels---that replace straight line segments of vectors with second-order
curves---are fundamental components of vector algebra locus equations of
quadratic curves and surfaces.

\subsection{Reproducing Kernels for Extreme Points}

Recall that the Gaussian discriminant function in (\ref{Gaussian Rule})
contains a pair of signed random quadratic forms\emph{\ }$\mathbf{x}%
^{T}\mathbf{\Sigma}_{1}^{-1}\mathbf{x}$ and $-\mathbf{x}^{T}\mathbf{\Sigma
}_{2}^{-1}\mathbf{x}$ that jointly provide dual representation of the
discriminant function and the intrinsic coordinate system of the decision
boundary of a minimum risk binary classification system, such that the dual
component $\mathbf{x}^{T}\mathbf{\Sigma}_{1}^{-1}\mathbf{x-x}^{T}%
\mathbf{\Sigma}_{2}^{-1}\mathbf{x}$ is the solution of the vector algebra
locus equation of (\ref{Norm_Dec_Bound}) that represents the decision boundary
of the system, so that any given decision boundary is a certain quadratic
curve or surface.

Given the geometrical and statistical structure of the dual component
$\mathbf{x}^{T}\mathbf{\Sigma}_{1}^{-1}\mathbf{x-x}^{T}\mathbf{\Sigma}%
_{2}^{-1}\mathbf{x}$\textbf{,} along with the conditions expressed by Theorems
\ref{Vector Algebra Equation of Quadratic Loci Theorem} -
\ref{Vector Algebra Equation of Spherical Loci Theorem}, it follows that the
form of the general vector algebra locus equation---of any given circle,
ellipse, parabola, hyperbola, hypersphere, hyperellipsoid, hyperparaboloid or
hyperhyperboloid---is determined by first degree \emph{and} second degree
vector components of vectors.

On the other hand, given conditions expressed by Theorem
\ref{Vector Algebra Equation of Linear Loci Theorem}, it follows that the form
of the vector algebra locus equation of any given line, plane or hyperplane is
determined by first degree vector components of vectors.

We also realize that the pair of signed random quadratic forms $\mathbf{x}%
^{T}\mathbf{\Sigma}_{1}^{-1}\mathbf{x}$ and $\mathbf{-x}^{T}\mathbf{\Sigma
}_{2}^{-1}\mathbf{x}$ in the vector algebra locus equation of
(\ref{Norm_Dec_Bound}) determines the geometrical and statistical structure of
an intrinsic coordinate system $\mathbf{x}^{T}\mathbf{\Sigma}_{1}%
^{-1}\mathbf{x-x}^{T}\mathbf{\Sigma}_{2}^{-1}\mathbf{x}$ of a \emph{nonlinear}
decision boundary---of \emph{any} given minimum risk binary classification
system that is subject to multivariate normal data, so that the geometric
locus of the nonlinear decision boundary has the form of a $d$-dimensional
circle, ellipse, parabola, hyperbola, hypersphere, hyperellipsoid,
hyperparaboloid or hyperhyperboloid.

Then again, we don't always know whether the geometric locus of a decision
boundary of minimum risk binary classification system has the form of a line,
plane or hyperplane.

Even so, the conditions expressed by Theorem
\ref{Principal Eigen-coordinate System Theorem} and Corollary
\ref{Principal Eigen-coordinate System Corollary} guarantee us that any given
quadratic form that is the solution of a vector algebra locus equation,
wherein the graph of the vector algebra locus equation represents a certain
line, plane or hyperplane, can be represented by an exclusive principal
eigen-coordinate system, such that the exclusive principal eigen-coordinate
system is the solution of an equivalent form of the vector algebra locus
equation of the line, plane or hyperplane in such a manner that the
eigenenergies exhibited by the components of the exclusive principal
eigen-coordinate system are modulated by the eigenvalues of the symmetric
matrix of the transformed quadratic form, at which point the principal
eigenaxis of the geometric locus of the line, plane or hyperplane satisfies
the geometric locus of the line, plane or hyperplane in terms of its total
allowed eigenenergy.

Thereby, we realize that \emph{all} of the components $\left\{  x_{1i\ast
}\right\}  _{i=1}^{d}$ and $\left\{  x_{2i\ast}\right\}  _{i=1}^{d}$ of any
given extreme vectors $\mathbf{x}_{1_{i\ast}}$ and $\mathbf{x}_{2_{i\ast}}$
that are solutions of an equivalent form of the locus equation of
(\ref{Norm_Dec_Bound}) \emph{need} to contain both first degree components
$x_{1i\ast}$ and $x_{2i\ast}$ \emph{and} second degree components $x_{1i\ast
}^{2}$ and $x_{2i\ast}^{2}$.

We have demonstrated that second-degree polynomial reproducing kernels
$k_{\mathbf{x}}\left(  \mathbf{s}\right)  =\left(  \mathbf{s}^{T}%
\mathbf{x}+1\right)  ^{2}$ and Gaussian reproducing kernels that have the form
$k_{\mathbf{x}}\left(  \mathbf{s}\right)  =\exp\left(  -\gamma\left\Vert
\mathbf{s}-\mathbf{x}\right\Vert ^{2}\right)  $, where $0.01\leq\gamma\leq
0.1$, provide this essential algebraic and geometric structure since: $\left(
1\right)  $ both types of reproducing kernels replace vectors $\mathbf{x\in%
\mathbb{R}
}^{d}$ in Hilbert space $\mathcal{H}$ with second-order curves---formed by
first $x_{i}$\ and second $x_{i}^{2}$ degree vector components---that are more
or less sinuous and thereby preserve topological properties of vectors in
Hilbert space $\mathcal{H}$
\citep{Rapport1963}%
; and $\left(  2\right)  $ both types of reproducing kernels implement inner
products of vectors in a Hilbert space $\mathcal{H}$ that is a reproducing
kernel Hilbert space.

Moreover, we have demonstrated that geometric loci of both linear and
quadratic decision boundaries are well-approximated by such second-order
curves in reproducing kernel Hilbert spaces
\citep{Reeves2018design}%
.

\subsection{Decision Boundaries in a RKHS}

Geometric loci of linear and quadratic decision boundaries are both
well-approximated by second-order curves in reproducing kernel Hilbert spaces,
where the reproducing kernel is a second-degree polynomial reproducing kernel
$k_{\mathbf{x}}\left(  \mathbf{s}\right)  =\left(  \mathbf{s}^{T}%
\mathbf{x}+1\right)  ^{2}$ or a Gaussian reproducing kernel that has the form
$k_{\mathbf{x}}\left(  \mathbf{s}\right)  =\exp\left(  -\gamma\left\Vert
\mathbf{s}-\mathbf{x}\right\Vert ^{2}\right)  $, where $0.01\leq\gamma\leq0.1$.

By way of demonstration, we now present examples of geometric loci of
quadratic and linear decision boundaries that have been approximated by a
second-order curve in a RKHS, where the reproducing kernel is a second-degree
polynomial reproducing kernel.

\subsubsection{Linear Decision Boundaries in a RKHS}

Take any two classes of random vectors that have similar covariance matrices.
By (\ref{Norm_Dec_Bound}), the discriminant function of the minimum risk
binary classification system is the solution of a vector algebra locus
equation that represents the geometric locus of a linear decision boundary.

Figure $5$ illustrates the geometric locus of a linear decision boundary of a
minimum risk binary classification system that has been estimated in a
reproducing kernel Hilbert space that has a second-degree polynomial
reproducing kernel, where the geometric locus of the linear decision boundary
of the system is bounded by the geometric loci of a pair of symmetrically
positioned linear decision borders. The linear decision boundary is black, the
pair of symmetrically positioned linear decision borders are blue and red, and
each extreme point is enclosed in a black circle.%
\begin{figure}[h]%
\centering
\includegraphics[
height=2.0669in,
width=5.5988in
]%
{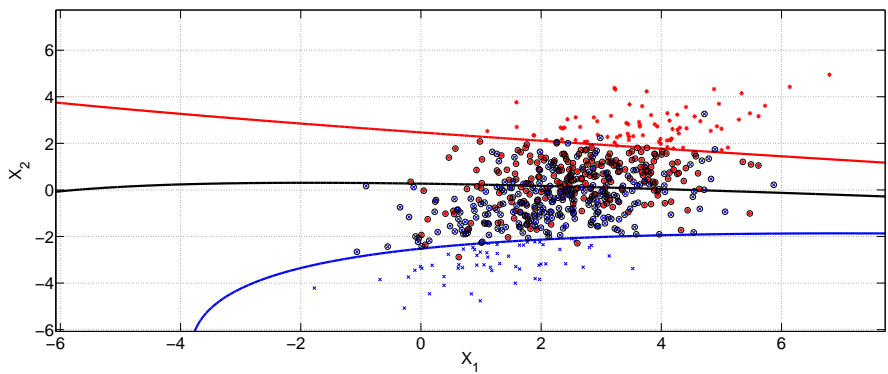}%
\caption{Illustration of geometric loci of a linear decision boundary and a
pair of symmetrically positioned linear decision borders of a minimum risk
binary classification system, all of which have been estimated in a
reproducing kernel Hilbert space that has a second-order polynomial
reproducing kernel.}%
\end{figure}

\subsubsection{Quadratic Decision Boundaries in a RKHS}

Take two classes of random vectors that have dissimilar covariance matrices,
such that the covariance matrices for class $\omega_{1}$ and class $\omega
_{2}$ are given by%
\[
\Sigma_{1}=\left[
\begin{array}
[c]{cc}%
1.5 & 0.75\\
0.75 & 2
\end{array}
\right]  \text{, \ \ }\Sigma_{2}=\left[
\begin{array}
[c]{cc}%
2 & 0.5\\
0.5 & 1.2
\end{array}
\right]  \text{,}%
\]
the mean vector for class $\omega_{1}$ is given by $M_{1}=%
\begin{pmatrix}
3, & 2
\end{pmatrix}
^{T}$, and the mean vector for class $\omega_{2}$ is given by $M_{2}=%
\begin{pmatrix}
2, & 2
\end{pmatrix}
^{T}$.

By (\ref{Norm_Dec_Bound}), the discriminant function of the minimum risk
binary classification system is the solution of a vector algebra locus
equation that represents the geometric locus of a hyperbolic decision
boundary. Figure $6$ illustrates the geometric locus of the hyperbolic
decision boundary of the minimum risk binary classification system---that has
been estimated in a reproducing kernel Hilbert space that has a second-degree
polynomial reproducing kernel, where the geometric locus of the hyperbolic
decision boundary of the system is bounded by the geometric loci of a pair of
symmetrically positioned hyperbolic decision borders. The hyperbolic decision
boundary is black, the pair of symmetrically positioned hyperbolic decision
borders are blue and red, and each extreme point is enclosed in a black
circle.%
\begin{figure}[h]%
\centering
\includegraphics[
height=2.0669in,
width=5.5979in
]%
{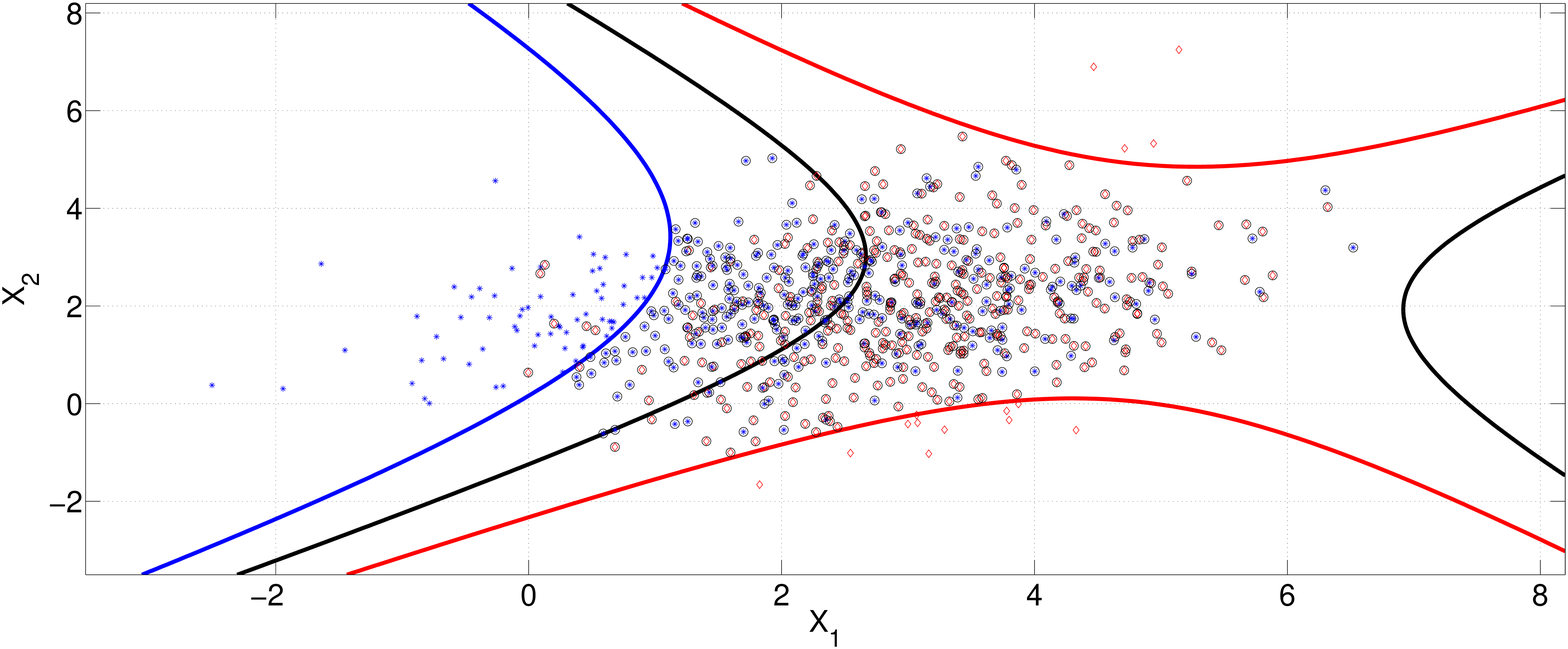}%
\caption{Illustration of geometric loci of a hyperbolic decision boundary and
a pair of symmetrically positioned hyperbolic decision borders of a minimum
risk binary classification system, all of which have been estimated in a
reproducing kernel Hilbert space that has a second-order polynomial
reproducing kernel.}%
\end{figure}

\subsection{Principal Eigenvectors of Joint Covariance Matrices}

In this part of our treatise, we turn our attention to principal eigenvectors
of \emph{joint} covariance matrices, so that the elements of any given joint
covariance matrix describe differences between joint variabilities of normal
random vectors $\mathbf{x\sim}p\left(  \mathbf{x};\boldsymbol{\mu}%
_{1},\mathbf{\Sigma}_{1}\right)  $ and $\mathbf{x\sim}p\left(  \mathbf{x}%
;\boldsymbol{\mu}_{2},\mathbf{\Sigma}_{2}\right)  $ that belong to a
collection of two categories $\omega_{1}$ and $\omega_{2}$ of normal random
vectors $\mathbf{x}$, at which point the magnitude and the direction of the
principal eigenvector---of the joint covariance matrix---are both functions of
differences between joint variabilities of normal \emph{extreme vectors}
$\mathbf{x}_{1_{i\ast}}\mathbf{\sim}p\left(  \mathbf{x};\boldsymbol{\mu}%
_{1},\mathbf{\Sigma}_{1}\right)  $ and $\mathbf{x}_{2_{i\ast}}\mathbf{\sim
}p\left(  \mathbf{x};\boldsymbol{\mu}_{2},\mathbf{\Sigma}_{2}\right)  $.

We begin by considering how we might use conditions stated in Theorem
\ref{Principal Eigen-coordinate System Theorem}, Corollary
\ref{Principal Eigen-coordinate System Corollary} and Corollary
\ref{Symmetrical and Equivalent Principal Eigenaxes} to \emph{transform} the
vector algebra locus equation of (\ref{Norm_Dec_Bound})%
\begin{align*}
d\left(  \mathbf{x}\right)   &  :\mathbf{x}^{T}\mathbf{\Sigma}_{1}%
^{-1}\mathbf{x}-2\mathbf{x}^{T}\mathbf{\Sigma}_{1}^{-1}\boldsymbol{\mu}%
_{1}+\boldsymbol{\mu}_{1}^{T}\mathbf{\Sigma}_{1}^{-1}\boldsymbol{\mu}_{1}%
-\ln\left(  \left\vert \mathbf{\Sigma}_{1}\right\vert \right) \\
&  -\mathbf{x}^{T}\mathbf{\Sigma}_{2}^{-1}\mathbf{x}+2\mathbf{x}%
^{T}\mathbf{\Sigma}_{2}^{-1}\boldsymbol{\mu}_{2}\mathbf{-}\boldsymbol{\mu}%
_{2}^{T}\mathbf{\Sigma}_{2}^{-1}\boldsymbol{\mu}_{2}+\ln\left(  \left\vert
\mathbf{\Sigma}_{2}\right\vert \right)  =0
\end{align*}
by a suitable change of the basis of the coordinate system $\mathbf{x}%
^{T}\mathbf{\Sigma}_{1}^{-1}\mathbf{x-x}^{T}\mathbf{\Sigma}_{2}^{-1}%
\mathbf{x}$, so that \emph{likelihood values} and \emph{likely locations} of
extreme points $\mathbf{x}_{1_{i\ast}}\mathbf{\ }$and $\mathbf{x}_{2_{i\ast}}$
\emph{determine} the \emph{positions} of the \emph{basis }of the transformed
intrinsic coordinate system $\mathbf{x}^{T}\mathbf{\Sigma}_{1}^{-1}%
\mathbf{x-x}^{T}\mathbf{\Sigma}_{2}^{-1}\mathbf{x}$, at which point the
transformed basis has the form of a locus of signed and scaled extreme vectors
$\mathbf{x}_{1_{\ast}}\mathbf{\sim}p\left(  \mathbf{x};\boldsymbol{\mu}%
_{1},\mathbf{\Sigma}_{1}\right)  $ and $\mathbf{x}_{2_{\ast}}\mathbf{\sim
}p\left(  \mathbf{x};\boldsymbol{\mu}_{2},\mathbf{\Sigma}_{2}\right)  $.

We start by examining principal eigenvectors of covariance matrices.

\subsubsection{Principal Eigenvectors of Covariance Matrices}

Let $\mathbf{v}_{1}$ be the principal eigenvector of the inverted covariance
matrix $\mathbf{\Sigma}_{1}^{-1}$ and the covariance matrix $\mathbf{\Sigma
}_{1}$ in the vector algebra locus equation of (\ref{Norm_Dec_Bound}), so that
the principal eigenvector $\mathbf{v}_{1}$ exhibits a magnitude and a
direction for which a class $\omega_{1}$ of normal extreme random vectors
$\mathbf{x}_{1_{i\ast}}\mathbf{\sim}p\left(  \mathbf{x};\boldsymbol{\mu}%
_{1},\mathbf{\Sigma}_{1}\right)  $ varies the most.

By Corollary \ref{Principal Eigen-coordinate System Corollary}, it follows
that a principal eigenaxis $\widetilde{\mathbf{v}}_{1}\mathbf{\ }$of a certain
quadratic curve or surface\ is the principal part of an equivalent
representation of the random quadratic form $\mathbf{v}_{1}^{T}\mathbf{\Sigma
}_{1}^{-1}\mathbf{v}_{1}$, such that the principal eigenaxis
$\widetilde{\mathbf{v}}_{1}$ is symmetrically and equivalently related to the
principal eigenvector $\mathbf{v}_{1}$ of the symmetric matrix $\mathbf{\Sigma
}_{1}^{-1}$ of the random quadratic form $\mathbf{v}_{1}^{T}\mathbf{\Sigma
}_{1}^{-1}\mathbf{v}_{1}$, so that the principal eigenaxis
$\widetilde{\mathbf{v}}_{1}$ satisfies the quadratic curve or surface in terms
of its total allowed eigenenergy $\left\Vert \widetilde{\mathbf{v}}%
_{1}\right\Vert ^{2}$, at which point the total allowed eigenenergy
$\left\Vert \widetilde{\mathbf{v}}_{1}\right\Vert ^{2}$ exhibited by the
principal eigenaxis $\widetilde{\mathbf{v}}_{1}$ is regulated by the
eigenvalues of the symmetric matrix $\mathbf{\Sigma}_{1}^{-1}$ of the random
quadratic form $\mathbf{v}_{1}^{T}\mathbf{\Sigma}_{1}^{-1}\mathbf{v}_{1}$,
wherein the principal eigenaxis $\widetilde{\mathbf{v}}_{1}\mathbf{\ }%
$exhibits a magnitude and a direction for which the class $\omega_{1}$ of
normal extreme random vectors $\mathbf{x}_{1_{i\ast}}\mathbf{\sim}p\left(
\mathbf{x};\boldsymbol{\mu}_{1},\mathbf{\Sigma}_{1}\right)  $ varies the most.

By Corollary \ref{Symmetrical and Equivalent Principal Eigenaxes}, it also
follows that a principal eigenaxis$\mathbf{\ }\overrightarrow{\mathbf{v}}_{1}$
of a certain quadratic curve or surface\ is the principal part of an
equivalent representation of the random quadratic form $\mathbf{v}_{1}%
^{T}\mathbf{\Sigma}_{1}\mathbf{v}_{1}$, such that the principal
eigenaxis\textbf{ }$\overrightarrow{\mathbf{v}}_{1}$ is symmetrically and
equivalently related to the principal eigenvector $\mathbf{v}_{1}$ of the
symmetric matrix $\mathbf{\Sigma}_{1}$ of the random quadratic form
$\mathbf{v}_{1}^{T}\mathbf{\Sigma}_{1}\mathbf{v}_{1}$ and is also
symmetrically and equivalently related to the principal eigenaxis
$\widetilde{\mathbf{v}}_{1}$ of a similar quadratic curve or surface, so that
the principal eigenaxis $\overrightarrow{\mathbf{v}}_{1}$ satisfies the
quadratic curve or surface in terms of its total allowed eigenenergy
$\left\Vert \overrightarrow{\mathbf{v}}_{1}\right\Vert ^{2}$, at which point
the total allowed eigenenergy $\left\Vert \overrightarrow{\mathbf{v}}%
_{1}\right\Vert ^{2}$ exhibited by the principal eigenaxis
$\overrightarrow{\mathbf{v}}_{1}$ is regulated by the eigenvalues of the
symmetric matrix $\mathbf{\Sigma}_{1}$ of the random quadratic form
$\mathbf{v}_{1}^{T}\mathbf{\Sigma}_{1}\mathbf{v}_{1}$, wherein the principal
eigenaxis $\overrightarrow{\mathbf{v}}_{1}\mathbf{\ }$exhibits a magnitude and
a direction for which the class $\omega_{1}$ of normal extreme random vectors
$\mathbf{x}_{1_{i\ast}}\mathbf{\sim}p\left(  \mathbf{x};\boldsymbol{\mu}%
_{1},\mathbf{\Sigma}_{1}\right)  $ varies the most.

Correspondingly, let $\mathbf{v}_{2}$ be the principal eigenvector of the
inverted covariance matrix $\mathbf{\Sigma}_{2}^{-1}$ and the covariance
matrix $\mathbf{\Sigma}_{2}$ in the vector algebra locus equation of
(\ref{Norm_Dec_Bound}), so that the principal eigenvector $\mathbf{v}_{2}$
exhibits a magnitude and a direction for which a class $\omega_{2}$ of normal
extreme random vectors $\mathbf{x}_{2_{i\ast}}\mathbf{\sim}p\left(
\mathbf{x};\boldsymbol{\mu}_{2},\mathbf{\Sigma}_{2}\right)  $ varies the most.

By Corollary \ref{Principal Eigen-coordinate System Corollary}, it follows
that a principal eigenaxis $\widetilde{\mathbf{v}}_{2}\mathbf{\ }$of a certain
quadratic curve or surface\ is the principal part of an equivalent
representation of the random quadratic form $\mathbf{v}_{2}^{T}\mathbf{\Sigma
}_{2}^{-1}\mathbf{v}_{2}$, such that the principal eigenaxis
$\widetilde{\mathbf{v}}_{2}$ is symmetrically and equivalently related to the
principal eigenvector $\mathbf{v}_{2}$ of the symmetric matrix $\mathbf{\Sigma
}_{2}^{-1}$ of the random quadratic form $\mathbf{v}_{2}^{T}\mathbf{\Sigma
}_{2}^{-1}\mathbf{v}_{2}$, so that the principal eigenaxis
$\widetilde{\mathbf{v}}_{2}$ satisfies the quadratic curve or surface in terms
of its total allowed eigenenergy$\left\Vert \widetilde{\mathbf{v}}%
_{2}\right\Vert ^{2}$, at which point the total allowed eigenenergy
$\left\Vert \widetilde{\mathbf{v}}_{2}\right\Vert ^{2}$ exhibited by the
principal eigenaxis $\widetilde{\mathbf{v}}_{2}$ is regulated by the
eigenvalues of the symmetric matrix $\mathbf{\Sigma}_{2}^{-1}$ of the random
quadratic form $\mathbf{v}_{2}^{T}\mathbf{\Sigma}_{2}^{-1}\mathbf{v}_{2}$,
wherein the principal eigenaxis $\widetilde{\mathbf{v}}_{2}\mathbf{\ }%
$exhibits a magnitude and a direction for which the class $\omega_{2}$ of
normal random vectors $\mathbf{x}_{2_{i\ast}}\mathbf{\sim}p\left(
\mathbf{x};\boldsymbol{\mu}_{2},\mathbf{\Sigma}_{2}\right)  $ varies the most.

By Corollary \ref{Symmetrical and Equivalent Principal Eigenaxes}, it also
follows that a principal eigenaxis$\mathbf{\ }\overrightarrow{\mathbf{v}}_{2}$
of a certain quadratic curve or surface\ is the principal part of an
equivalent representation of the random quadratic form $\mathbf{v}_{1}%
^{T}\mathbf{\Sigma}_{2}\mathbf{v}_{1}$, such that the principal
eigenaxis\textbf{ }$\overrightarrow{\mathbf{v}}_{2}$ is symmetrically and
equivalently related to the principal eigenvector $\mathbf{v}_{2}$ of the
symmetric matrix $\mathbf{\Sigma}_{2}$ of the random quadratic form
$\mathbf{v}_{1}^{T}\mathbf{\Sigma}_{2}\mathbf{v}_{1}$ and is also
symmetrically and equivalently related to the principal eigenaxis
$\widetilde{\mathbf{v}}_{2}$ of a similar quadratic curve or surface, so that
the principal eigenaxis $\overrightarrow{\mathbf{v}}_{2}$ satisfies the
quadratic curve or surface in terms of its total allowed eigenenergy
$\left\Vert \overrightarrow{\mathbf{v}}_{2}\right\Vert ^{2}$, at which point
the total allowed eigenenergy $\left\Vert \overrightarrow{\mathbf{v}}%
_{2}\right\Vert ^{2}$ exhibited by the principal eigenaxis
$\overrightarrow{\mathbf{v}}_{2}$ is regulated by the eigenvalues of the
symmetric matrix $\mathbf{\Sigma}_{2}$ of the random quadratic form
$\mathbf{v}_{1}^{T}\mathbf{\Sigma}_{2}\mathbf{v}_{1}$, wherein the principal
eigenaxis$\mathbf{\ }\overrightarrow{\mathbf{v}}_{2}\mathbf{\ }$exhibits a
magnitude and a direction for which the class $\omega_{2}$ of normal random
vectors $\mathbf{x}_{2_{i\ast}}\mathbf{\sim}p\left(  \mathbf{x}%
;\boldsymbol{\mu}_{2},\mathbf{\Sigma}_{2}\right)  $ varies the most.

We realize that we need to determine an \emph{equivalent representation}---for
random quadratic forms---that \emph{joins} both \emph{pairs} of random
quadratic forms: $\mathbf{v}_{1}^{T}\mathbf{\Sigma}_{1}^{-1}\mathbf{v}_{1}$
and $\mathbf{v}_{2}^{T}\mathbf{\Sigma}_{2}^{-1}\mathbf{v}_{2}$; and
$\mathbf{v}_{2}^{T}\mathbf{\Sigma}_{2}^{-1}\mathbf{v}_{2}$ and $\mathbf{v}%
_{1}^{T}\mathbf{\Sigma}_{2}\mathbf{v}_{1}$ in (\ref{Norm_Dec_Bound}). We now
turn our attention to principal eigenvectors of \emph{joint} covariance
matrices. We begin with the notion of joint covariance matrices.

\subsection{Joint Covariance Matrices}

Let $\mathbf{Q}$ denote a joint covariance matrix, and let $\mathbf{Q}^{-1}$
denote the inverted joint covariance matrix, so that the elements of
$\mathbf{Q}$ and $\mathbf{Q}^{-1}$ both describe differences between joint
variabilities of normal random vectors $\mathbf{x\sim}p\left(  \mathbf{x}%
;\boldsymbol{\mu}_{1},\mathbf{\Sigma}_{1}\right)  $ and $\mathbf{x\sim
}p\left(  \mathbf{x};\boldsymbol{\mu}_{2},\mathbf{\Sigma}_{2}\right)  $ that
belong to a collection of two categories $\omega_{1}$ and $\omega_{2}$ of
normal random vectors $\mathbf{x}$.

Accordingly, let $\mathbf{Q}$ denote an $N\times N$ joint covariance matrix
that is formed by $N$ labeled $\pm1$ reproducing kernels $k_{\mathbf{x}_{i}}$
for $N$ feature vectors $\mathbf{x\in}$ $%
\mathbb{R}
^{d}$%
\[
\mathbf{Q=\ }%
\begin{bmatrix}
\left\Vert k_{\mathbf{x}_{1}}\right\Vert \left\Vert k_{\mathbf{x}_{1}%
}\right\Vert \cos\theta_{k_{\mathbf{x_{1}}}k_{\mathbf{x}_{1}}} & \cdots &
-\left\Vert k_{\mathbf{x}_{1}}\right\Vert \left\Vert k_{\mathbf{x}_{N}%
}\right\Vert \cos\theta_{k_{\mathbf{x}_{1}}k_{\mathbf{x}_{N}}}\\
\vdots & \ddots & \vdots\\
-\left\Vert k_{\mathbf{x}_{N}}\right\Vert \left\Vert k_{\mathbf{x}_{1}%
}\right\Vert \cos\theta_{k_{\mathbf{x}_{N}}k_{\mathbf{x}_{1}}} & \cdots &
\left\Vert k_{\mathbf{x}_{N}}\right\Vert \left\Vert k_{\mathbf{x}_{N}%
}\right\Vert \cos\theta_{k_{\mathbf{x}_{N}}k_{\mathbf{x}_{N}}}%
\end{bmatrix}
\text{,}%
\]
so that $\mathbf{Q}$ is composed of $N\times N$ elements $y_{i}\left\Vert
k_{\mathbf{x}_{i}}\right\Vert y_{j}\left\Vert k_{\mathbf{x}_{j}}\right\Vert
\cos\theta_{k_{\mathbf{x}_{i}}k_{\mathbf{x}_{j}}}$, such that each element
$y_{i}\left\Vert k_{\mathbf{x}_{i}}\right\Vert y_{j}\left\Vert k_{\mathbf{x}%
_{j}}\right\Vert \cos\theta_{k_{\mathbf{x}_{i}}k_{\mathbf{x}_{j}}}$ of
$\mathbf{Q}$ where $y_{i}y_{j}=-1$ describes differences between joint
variabilities of feature vectors $k_{\mathbf{x}_{i}}$ and $k_{\mathbf{x}_{j}}$
that belong to different pattern classes $\omega_{1}$ and $\omega_{2}$, at
which point each element $\left\Vert k_{\mathbf{x}_{i}}\right\Vert \left\Vert
k_{\mathbf{x}_{j}}\right\Vert \cos\theta_{k_{\mathbf{x}_{i}}k_{\mathbf{x}_{j}%
}}$ of the joint covariance matrix $\mathbf{Q}$ is correlated with the
distance $\left\Vert k_{\mathbf{x}_{i}}-k_{\mathbf{x}_{j}}\right\Vert $
between the loci of certain feature vectors $k_{\mathbf{x}_{i}}$ and
$k_{\mathbf{x}_{j}}$.

\subsubsection{Principal Eigenvectors of Joint Covariance Matrices}

Take any given joint covariance matrix $\mathbf{Q}$ that is described above.
Let $\mathbf{v=v}_{1}+\mathbf{v}_{2}$ denote the principal eigenvector of the
joint covariance matrix $\mathbf{Q}$ and the inverted joint covariance matrix
$\mathbf{Q}^{-1}$, so that the principal eigenvector $\mathbf{v=v}%
_{1}+\mathbf{v}_{2}$ of $\mathbf{Q}$ and $\mathbf{Q}^{-1}$ exhibits a
magnitude and a direction for which both classes $\omega_{1}$ and $\omega_{2}$
of normal random vectors $\mathbf{x}_{1_{i\ast}}\mathbf{\sim}p\left(
\mathbf{x};\boldsymbol{\mu}_{1},\mathbf{\Sigma}_{1}\right)  $ and
$\mathbf{x}_{2_{i\ast}}\mathbf{\sim}p\left(  \mathbf{x};\boldsymbol{\mu}%
_{2},\mathbf{\Sigma}_{2}\right)  $ vary the most.

Let $\mathbf{v}^{T}\mathbf{Qv}$ and $\mathbf{v}^{T}\mathbf{Q}^{-1}\mathbf{v}$
be random quadratic forms that are solutions of vector algebra locus
equations, such that $\mathbf{v=v}_{1}+\mathbf{v}_{2}$ is the principal
eigenvector of both $\mathbf{Q}$ and $\mathbf{Q}^{-1}$, so that the magnitude
and the direction of the principal eigenvector $\mathbf{v=v}_{1}%
+\mathbf{v}_{2}$ of $\mathbf{Q}$ and $\mathbf{Q}^{-1}$ are both functions of
differences between joint variabilities of normal \emph{extreme vectors}
$\mathbf{x}_{1_{i\ast}}\mathbf{\sim}p\left(  \mathbf{x};\boldsymbol{\mu}%
_{1},\mathbf{\Sigma}_{1}\right)  $ and $\mathbf{x}_{2_{i\ast}}\mathbf{\sim
}p\left(  \mathbf{x};\boldsymbol{\mu}_{2},\mathbf{\Sigma}_{2}\right)  $.

Given Theorem \ref{Principal Eigen-coordinate System Theorem}, Corollary
\ref{Principal Eigen-coordinate System Corollary} and Corollary
\ref{Symmetrical and Equivalent Principal Eigenaxes}, it follows that a
principal eigenaxis $\widetilde{\mathbf{v}}\ \mathbf{=\ }\widetilde{\mathbf{v}%
}_{1}-\widetilde{\mathbf{v}}_{2}$ of a certain quadratic curve or surface is
the principal part of an equivalent representation of the pair of random
quadratic forms $\mathbf{v}^{T}\mathbf{Qv}$ and $\mathbf{v}^{T}\mathbf{Q}%
^{-1}\mathbf{v}$, such that the principal eigenaxis $\widetilde{\mathbf{v}%
}\ \mathbf{=\ }\widetilde{\mathbf{v}}_{1}-\widetilde{\mathbf{v}}_{2}$ is
symmetrically and equivalently related to the principal eigenvector
$\mathbf{v=v}_{1}+\mathbf{v}_{2}$ of the symmetric matrix $\mathbf{Q}$ of the
random quadratic form $\mathbf{v}^{T}\mathbf{Qv}$ and the symmetric matrix
$\mathbf{Q}^{-1}$ of the correlated random quadratic form $\mathbf{v}%
^{T}\mathbf{Q}^{-1}\mathbf{v}$, coupled with the principal eigenaxis
$\overrightarrow{\mathbf{v}}=\overrightarrow{\mathbf{v}}_{1}%
-\overrightarrow{\mathbf{v}}_{2}$ of a similar quadratic curve or surface, so
that the principal eigenaxis $\widetilde{\mathbf{v}}\ \mathbf{=\ }%
\widetilde{\mathbf{v}}_{1}-\widetilde{\mathbf{v}}_{2}$ satisfies the certain
quadratic curve or surface in terms of its total allowed eigenenergy
$\mathbf{\ }\left\Vert \widetilde{\mathbf{v}}_{1}-\widetilde{\mathbf{v}}%
_{2}\right\Vert \ ^{2}$, at which point the total allowed eigenenergy is
regulated by the eigenvalues of the symmetric matrices $\mathbf{Q}$ and
$\mathbf{Q}^{-1}$ of the pair of random quadratic forms $\mathbf{v}%
^{T}\mathbf{Qv}$ and $\mathbf{v}^{T}\mathbf{Q}^{-1}\mathbf{v}$, wherein the
magnitude and the direction of the principal eigenaxis $\widetilde{\mathbf{v}%
}\ \mathbf{=\ }\widetilde{\mathbf{v}}_{1}-\widetilde{\mathbf{v}}_{2}$ are both
functions of differences between joint variabilities of normal \emph{extreme
vectors} $\mathbf{x}_{1_{i\ast}}\mathbf{\sim}p\left(  \mathbf{x}%
;\boldsymbol{\mu}_{1},\mathbf{\Sigma}_{1}\right)  $ and $\mathbf{x}_{2_{i\ast
}}\mathbf{\sim}p\left(  \mathbf{x};\boldsymbol{\mu}_{2},\mathbf{\Sigma}%
_{2}\right)  $.

Given the above assumptions and notation, along with the guarantees provided
by Theorem \ref{Principal Eigen-coordinate System Theorem}, Corollary
\ref{Principal Eigen-coordinate System Corollary} and Corollary
\ref{Symmetrical and Equivalent Principal Eigenaxes}, we have discovered that
the basis of the intrinsic coordinate system%
\[
\mathbf{x}^{T}\mathbf{\Sigma}_{1}^{-1}\mathbf{x-x}^{T}\mathbf{\Sigma}_{2}%
^{-1}\mathbf{x}%
\]
in the vector algebra locus equation of (\ref{Norm_Dec_Bound}) has an
equivalent representation that is determined by an exclusive principal
eigen-coordinate system of a certain quadratic curve or surface---that is
structured as a locus of signed and scaled extreme points $\mathbf{x}%
_{1_{i\ast}}\mathbf{\sim}p\left(  \mathbf{x};\boldsymbol{\mu}_{1}%
,\mathbf{\Sigma}_{1}\right)  $ and $\mathbf{x}_{2_{i\ast}}\mathbf{\sim
}p\left(  \mathbf{x};\boldsymbol{\mu}_{2},\mathbf{\Sigma}_{2}\right)  $%
\begin{align*}
\boldsymbol{\rho}  &  =\sum\nolimits_{i=1}^{l_{1}}\psi_{1_{i_{\ast}}%
}k_{\mathbf{x}_{1_{i\ast}}}-\sum\nolimits_{i=1}^{l_{2}}\psi_{2_{i_{\ast}}%
}k_{\mathbf{x}_{2_{i\ast}}}\\
&  =\boldsymbol{\rho}_{1}-\boldsymbol{\rho}_{2}\text{,}%
\end{align*}
where $\psi_{1_{i_{\ast}}}$ and $\psi_{2_{i_{\ast}}}$ are scale factors, and
$k_{\mathbf{x}_{1_{i\ast}}}$ and $k_{\mathbf{x}_{2_{i\ast}}}$ are reproducing
kernels for extreme points $\mathbf{x}_{1_{i\ast}}$ and $\mathbf{x}_{2_{i\ast
}}$, such that the locus of signed and scaled extreme points $\boldsymbol{\rho
}=\boldsymbol{\rho}_{1}\mathbf{-}\boldsymbol{\rho}_{2}$ is the principal part
of an equivalent representation of a pair of random quadratic forms
$\mathbf{v}^{T}\mathbf{Qv}$ and $\mathbf{v}^{T}\mathbf{Q}^{-1}\mathbf{v}$%
\begin{align*}
\mathbf{v}^{T}\mathbf{Qv\ }  &  \mathbf{\equiv v}^{T}\mathbf{Q}^{-1}%
\mathbf{v}\\
&  \mathbf{\equiv\ }\mathbf{x}^{T}\mathbf{\Sigma}_{1}^{-1}\mathbf{x-x}%
^{T}\mathbf{\Sigma}_{2}^{-1}\mathbf{x}\text{\textbf{\textbf{,}}}%
\end{align*}
where $\mathbf{v=v}_{1}+\mathbf{v}_{2}$ is the principal eigenvector of a
joint covariance matrix $\mathbf{Q}$ and the inverted joint covariance matrix
$\mathbf{Q}^{-1}$, wherein the exclusive principal eigen-coordinate system
$\boldsymbol{\rho}=\boldsymbol{\rho}_{1}\mathbf{-}\boldsymbol{\rho}_{2}$ of
the quadratic curve or surface is symmetrically and equivalently related to
the principal eigenvector $\mathbf{v=v}_{1}+\mathbf{v}_{2}$ of the joint
covariance matrix $\mathbf{Q}$ and the inverted joint covariance matrix
$\mathbf{Q}^{-1}$, so that the exclusive principal eigen-coordinate system
$\boldsymbol{\rho}=\boldsymbol{\rho}_{1}\mathbf{-}\boldsymbol{\rho}_{2}$ is
the principal eigenaxis of the decision boundary of a minimum risk binary
classification system, at which point the geometric locus of the novel
principal eigenaxis $\boldsymbol{\rho}=\boldsymbol{\rho}_{1}\mathbf{-}%
\boldsymbol{\rho}_{2}$ satisfies the geometric locus of the decision boundary
in terms of a critical minimum eigenenergy $\left\Vert \boldsymbol{\rho}%
_{1}-\boldsymbol{\rho}_{2}\right\Vert ^{2}$, such that the total allowed
eigenenergy $\left\Vert \boldsymbol{\rho}_{1}-\boldsymbol{\rho}_{2}\right\Vert
^{2}$ exhibited by the geometric locus of the novel principal eigenaxis
$\boldsymbol{\rho}=\boldsymbol{\rho}_{1}\mathbf{-}\boldsymbol{\rho}_{2}$ is
regulated by the eigenvalues of the symmetric matrices $\mathbf{Q}$ and
$\mathbf{Q}^{-1}$ of the pair of random quadratic forms $\mathbf{v}%
^{T}\mathbf{Qv}$ and $\mathbf{v}^{T}\mathbf{Q}^{-1}\mathbf{v}$, wherein the
magnitude and the direction of the novel principal eigenaxis $\boldsymbol{\rho
}=\boldsymbol{\rho}_{1}\mathbf{-}\boldsymbol{\rho}_{2}$ are both functions of
differences between joint variabilities of extreme vectors $\mathbf{x}%
_{1_{i\ast}}\mathbf{\sim}p\left(  \mathbf{x};\boldsymbol{\mu}_{1}%
,\mathbf{\Sigma}_{1}\right)  $ and $\mathbf{x}_{2_{i\ast}}\mathbf{\sim
}p\left(  \mathbf{x};\boldsymbol{\mu}_{2},\mathbf{\Sigma}_{2}\right)
\mathbf{\ }$located within either overlapping regions or near tail regions of
distributions determined by the probability density functions $p\left(
\mathbf{x};\boldsymbol{\mu}_{1},\mathbf{\Sigma}_{1}\right)  $ and $p\left(
\mathbf{x};\boldsymbol{\mu}_{2},\mathbf{\Sigma}_{2}\right)  $.

The above discoveries are readily generalized in the following manner.

\subsection{Generalization of Discoveries}

The novel principal eigen-coordinate transform method expressed by Theorem
\ref{Principal Eigen-coordinate System Theorem}, Corollary
\ref{Principal Eigen-coordinate System Corollary} and Corollary
\ref{Symmetrical and Equivalent Principal Eigenaxes} substantiates the
following discoveries.

Let $\mathbf{Q}$ denote a joint covariance matrix and let $\mathbf{Q}^{-1}$
denote the inverted joint covariance matrix, so that the elements of
$\mathbf{Q}$ and $\mathbf{Q}^{-1}$ both describe differences between joint
variabilities of random vectors $\mathbf{x\sim}p\left(  \mathbf{x};\omega
_{1}\right)  $ and $\mathbf{x\sim}p\left(  \mathbf{x};\omega_{1}\right)  $
that belong to two classes $\omega_{1}$ and $\omega_{2}$, where $p\left(
\mathbf{x};\omega_{1}\right)  $ and $p\left(  \mathbf{x};\omega_{2}\right)  $
are certain probability density functions for the two classes $\omega_{1}$ and
$\omega_{2}$ of random vectors $\mathbf{x\in}$ $%
\mathbb{R}
^{d}$.

Now let $\mathbf{v}^{T}\mathbf{Qv}$ and $\mathbf{v}^{T}\mathbf{Q}%
^{-1}\mathbf{v}$ be correlated random quadratic forms, such that
$\mathbf{v=v}_{1}+\mathbf{v}_{2}$ is the principal eigenvector of both
$\mathbf{Q}$ and $\mathbf{Q}^{-1}$, so that the magnitude and the direction of
the principal eigenvector $\mathbf{v=v}_{1}-\mathbf{v}_{2}$ of $\mathbf{Q}$
and $\mathbf{Q}^{-1}$ are both functions of differences between joint
variabilities of extreme vectors $\mathbf{x}_{1_{i\ast}}\mathbf{\sim}p\left(
\mathbf{x};\omega_{1}\right)  $ and $\mathbf{x}_{2_{i\ast}}\mathbf{\sim
}p\left(  \mathbf{x};\omega_{1}\right)  $ that belong to the classes
$\omega_{1}$ and $\omega_{2}$ of random vectors $\mathbf{x\in}$ $%
\mathbb{R}
^{d}$.

Next, let the correlated random quadratic forms $\mathbf{v}^{T}\mathbf{Qv}$
and $\mathbf{v}^{T}\mathbf{Q}^{-1}\mathbf{v}$ be solutions of a system of
well-posed vector algebra locus equations, so that the pair of random
quadratic forms $\mathbf{v}^{T}\mathbf{Qv}$ and $\mathbf{v}^{T}\mathbf{Q}%
^{-1}\mathbf{v}$ have an \emph{equivalent representation}, such that the
principal part of the equivalent representation of the pair of random
quadratic forms $\mathbf{v}^{T}\mathbf{Qv}$ and $\mathbf{v}^{T}\mathbf{Q}%
^{-1}\mathbf{v}$ is an exclusive principal eigen-coordinate system---of the
geometric locus of a certain quadratic curve or surface---structured as a
locus of signed and scaled extreme points $\mathbf{x}_{1_{i\ast}}\mathbf{\sim
}p\left(  \mathbf{x};\omega_{1}\right)  $ and $\mathbf{x}_{2_{i\ast}%
}\mathbf{\sim}p\left(  \mathbf{x};\omega_{1}\right)  $%
\begin{align*}
\boldsymbol{\rho}  &  =\sum\nolimits_{i=1}^{l_{1}}\psi_{1_{i_{\ast}}%
}k_{\mathbf{x}_{1_{i\ast}}}-\sum\nolimits_{i=1}^{l_{2}}\psi_{2_{i_{\ast}}%
}k_{\mathbf{x}_{2_{i\ast}}}\\
&  =\boldsymbol{\rho}_{1}-\boldsymbol{\rho}_{2}\text{,}%
\end{align*}
where $\psi_{1_{i_{\ast}}}$ and $\psi_{2_{i_{\ast}}}$ are scale factors for
reproducing kernels $k_{\mathbf{x}_{1_{i\ast}}}$ and $k_{\mathbf{x}_{2_{i\ast
}}}$ of extreme points $\mathbf{x}_{1_{i\ast}}$ and $\mathbf{x}_{2_{i\ast}}$,
at which point the locus of signed and scaled extreme points $\boldsymbol{\rho
}=\sum\nolimits_{i=1}^{l_{1}}\psi_{1_{i_{\ast}}}k_{\mathbf{x}_{1_{i\ast}}%
}-\sum\nolimits_{i=1}^{l_{2}}\psi_{2_{i_{\ast}}}k_{\mathbf{x}_{2_{i\ast}}}$ is
a geometric locus of a novel principal eigenaxis $\boldsymbol{\rho
}=\boldsymbol{\rho}_{1}\mathbf{-}\boldsymbol{\rho}_{2}$ that is symmetrically
and equivalently related to the principal eigenvector $\mathbf{v=v}%
_{1}+\mathbf{v}_{2}$ of the joint covariance matrix $\mathbf{Q}$ and the
inverted joint covariance matrix $\mathbf{Q}^{-1}$, so that $\left(  1\right)
$ the geometric locus of the novel principal eigenaxis $\boldsymbol{\rho
}=\boldsymbol{\rho}_{1}\mathbf{-}\boldsymbol{\rho}_{2}$ is the principal
eigenaxis of the geometric locus of the decision boundary of a minimum risk
binary classification system, at which point the geometric locus of the novel
principal eigenaxis $\boldsymbol{\rho}=\boldsymbol{\rho}_{1}\mathbf{-}%
\boldsymbol{\rho}_{2}$ satisfies the geometric locus of the decision boundary
in terms of a critical minimum eigenenergy $\left\Vert \boldsymbol{\rho}%
_{1}-\boldsymbol{\rho}_{2}\right\Vert ^{2}$, wherein the total allowed
eigenenergy $\left\Vert \boldsymbol{\rho}_{1}-\boldsymbol{\rho}_{2}\right\Vert
^{2}$ exhibited by the geometric locus of the novel principal eigenaxis
$\boldsymbol{\rho}=\boldsymbol{\rho}_{1}\mathbf{-}\boldsymbol{\rho}_{2}$ is
regulated by the eigenvalues of the symmetric matrices $\mathbf{Q}$ and
$\mathbf{Q}^{-1}$ of the pair of random quadratic forms $\mathbf{v}%
^{T}\mathbf{Qv}$ and $\mathbf{v}^{T}\mathbf{Q}^{-1}\mathbf{v}$; and $\left(
2\right)  $ the uniform property exhibited by all of the points that lie on
the geometric locus of the decision boundary is the critical minimum
eigenenergy $\left\Vert \boldsymbol{\rho}_{1}-\boldsymbol{\rho}_{2}\right\Vert
^{2}$ exhibited by the principal eigenaxis $\boldsymbol{\rho}=\boldsymbol{\rho
}_{1}\mathbf{-}\boldsymbol{\rho}_{2}$ of the geometric locus of the decision
boundary, wherein the magnitude and the direction of the novel principal
eigenaxis $\boldsymbol{\rho}=\boldsymbol{\rho}_{1}\mathbf{-}\boldsymbol{\rho
}_{2}$ are both functions of differences between joint variabilities of
extreme vectors $\mathbf{x}_{1_{i\ast}}\mathbf{\sim}p\left(  \mathbf{x}%
;\omega_{1}\right)  $ and $\mathbf{x}_{2_{i\ast}}\mathbf{\sim}p\left(
\mathbf{x};\omega_{1}\right)  \mathbf{\ }$located within either overlapping
regions or near tail regions of distributions determined by the probability
density functions $p\left(  \mathbf{x};\omega_{1}\right)  $ and $p\left(
\mathbf{x};\omega_{2}\right)  $.

We have named the exclusive principal eigen-coordinate system
$\boldsymbol{\rho}=\boldsymbol{\rho}_{1}\mathbf{-}\boldsymbol{\rho}_{2}$ a
\textquotedblleft geometric locus of a novel principal
eigenaxis.\textquotedblright

Thereby, given the conditions expressed by Axiom
\ref{Rotation of Intrinsic Coordinate Axes Axiom}, Axioms
\ref{Overlapping Extreme Points Axiom} -
\ref{Non-overlapping Extreme Points Axiom}, Theorem
\ref{Principal Eigen-coordinate System Theorem} and Corollaries
\ref{Principal Eigen-coordinate System Corollary} -
\ref{Symmetrical and Equivalent Principal Eigenaxes}, along with the argument
and assumptions presented above, we are motivated to determine \emph{how} we
might \emph{find} an exclusive principal eigen-coordinate system that is
\emph{structured} as a locus of signed and scaled extreme points
$\psi_{1_{i_{\ast}}}k_{\mathbf{x}_{1_{i\ast}}}$ and $-\psi_{2_{i_{\ast}}%
}k_{\mathbf{x}_{2_{i\ast}}}$, so that the exclusive principal eigen-coordinate
system%
\begin{align*}
\boldsymbol{\rho}  &  =\sum\nolimits_{i=1}^{l_{1}}\psi_{1_{i_{\ast}}%
}k_{\mathbf{x}_{1_{i\ast}}}-\sum\nolimits_{i=1}^{l_{2}}\psi_{2_{i_{\ast}}%
}k_{\mathbf{x}_{2_{i\ast}}}\\
&  =\boldsymbol{\rho}_{1}-\boldsymbol{\rho}_{2}%
\end{align*}
is the principal \emph{part} of an \emph{equivalent representation} of a pair
of random quadratic forms $\mathbf{v}^{T}\mathbf{Qv}$ and $\mathbf{v}%
^{T}\mathbf{Q}^{-1}\mathbf{v}$ associated with a joint covariance matrix
$\mathbf{Q}$ and the inverted joint covariance matrix $\mathbf{Q}^{-1}$, where
$\mathbf{v=v}_{1}+\mathbf{v}_{2}$ is the principal eigenvector of $\mathbf{Q}$
and $\mathbf{Q}^{-1}$, where $\mathbf{v}^{T}\mathbf{Qv\equiv v}^{T}%
\mathbf{Q}^{-1}\mathbf{v}$, such that the exclusive principal eigen-coordinate
system $\boldsymbol{\rho}=\boldsymbol{\rho}_{1}\mathbf{-}\boldsymbol{\rho}%
_{2}$ is symmetrically and equivalently related to the \emph{principal
eigenvector} $\mathbf{v=v}_{1}+\mathbf{v}_{2}$ of the joint covariance matrix
$\mathbf{Q}$ and the inverted joint covariance matrix $\mathbf{Q}^{-1}$, so
that $\left(  1\right)  $ the exclusive principal eigen-coordinate system
$\boldsymbol{\rho}=\boldsymbol{\rho}_{1}\mathbf{-}\boldsymbol{\rho}_{2}$ is
the principal eigenaxis of the geometric locus of the decision boundary of a
minimum risk binary classification system, at which point $\left(  2\right)  $
the geometric locus of the novel principal eigenaxis $\boldsymbol{\rho
}=\boldsymbol{\rho}_{1}\mathbf{-}\boldsymbol{\rho}_{2}$ satisfies the
geometric locus of the decision boundary in terms of a critical minimum
eigenenergy $\left\Vert \boldsymbol{\rho}_{1}-\boldsymbol{\rho}_{2}\right\Vert
^{2}$, wherein the total allowed eigenenergy $\left\Vert \boldsymbol{\rho}%
_{1}-\boldsymbol{\rho}_{2}\right\Vert ^{2}$ exhibited by the geometric locus
of the novel principal eigenaxis $\boldsymbol{\rho}=\boldsymbol{\rho}%
_{1}\mathbf{-}\boldsymbol{\rho}_{2}$ is regulated by the eigenvalues of the
symmetric matrices $\mathbf{Q}$ and $\mathbf{Q}^{-1}$ of the pair of random
quadratic forms $\mathbf{v}^{T}\mathbf{Qv}$ and $\mathbf{v}^{T}\mathbf{Q}%
^{-1}\mathbf{v}$; and $\left(  3\right)  $ the uniform property exhibited by
all of the points that lie on the geometric locus of the decision boundary is
the critical minimum eigenenergy $\left\Vert \boldsymbol{\rho}_{1}%
-\boldsymbol{\rho}_{2}\right\Vert ^{2}$ exhibited by the geometric locus of
the novel principal eigenaxis $\boldsymbol{\rho}=\boldsymbol{\rho}%
_{1}\mathbf{-}\boldsymbol{\rho}_{2}$, wherein the magnitude and the direction
of the novel principal eigenaxis $\boldsymbol{\rho}=\boldsymbol{\rho}%
_{1}\mathbf{-}\boldsymbol{\rho}_{2}$ are both functions of differences between
joint variabilities of extreme vectors $\mathbf{x}_{1_{\ast}}\mathbf{\sim}$
$p\left(  \mathbf{x};\omega_{1}\right)  $ and $\mathbf{x}_{2_{\ast}%
}\mathbf{\sim}$ $p\left(  \mathbf{x};\omega_{2}\right)  $ located within
either overlapping regions or near tail regions of distributions determined by
certain probability density functions $p\left(  \mathbf{x};\omega_{1}\right)
$ and $p\left(  \mathbf{x};\omega_{2}\right)  $.

\subsection{An Eigenaxis of Symmetry that Spans Decision Spaces}

Figure $5$ and Figure $6$ both illustrate that a geometric locus of a novel
principal eigenaxis $\boldsymbol{\rho}=\boldsymbol{\rho}_{1}\mathbf{-}%
\boldsymbol{\rho}_{2}$ is the principal eigenaxis of the geometric locus of
the decision boundary of a minimum risk binary classification system, at which
point the geometric locus of the decision boundary of the system is bounded by
the geometric loci of a pair of symmetrically positioned decision borders.

It will be seen that the geometric locus of the novel principal eigenaxis
$\boldsymbol{\rho}=\boldsymbol{\rho}_{1}-\boldsymbol{\rho}_{2}$ of any given
minimum risk binary classification system completely determines the shape of
the decision space $Z=Z_{1}\cup Z_{2}$ of the system, wherein the geometric
locus of the novel principal eigenaxis $\boldsymbol{\rho}=\boldsymbol{\rho
}_{1}-\boldsymbol{\rho}_{2}$ represents an eigenaxis of symmetry that spans
the decision space $Z=Z_{1}\cup Z_{2}$ of the system.

We have determined that a geometric locus of a novel principal eigenaxis
$\boldsymbol{\rho}=\boldsymbol{\rho}_{1}\mathbf{-}\boldsymbol{\rho}_{2}%
$\ provides dual representation of the discriminant function, an
exclusive\emph{\ }and distinctive principal eigen-coordinate system of the
geometric locus of the decision boundary, and an eigenaxis of symmetry that
spans the decision space---of \emph{any} given minimum risk binary
classification system that is subject to two categories of random vectors
$\mathbf{x\in}$ $%
\mathbb{R}
^{d}$, such that $\mathbf{x\sim}$ $p\left(  \mathbf{x};\omega_{1}\right)  $
and $\mathbf{x\sim}$ $p\left(  \mathbf{x};\omega_{2}\right)  $, where
distributions of the random vectors $\mathbf{x}$ are determined by certain
probability density functions $p\left(  \mathbf{x};\omega_{1}\right)  $ and
$p\left(  \mathbf{x};\omega_{2}\right)  $.

Moreover, in the next part of our treatise, we reveal a constrained
optimization algorithm that \emph{finds} the geometric locus of the novel
principal eigenaxis of \emph{any} given minimum risk binary classification system.

We now define a geometric locus of a novel principal eigenaxis.

\subsection{Geometric Locus of a Novel Principal Eigenaxis}

Let $k_{\mathbf{x}_{1_{i\ast}}}\mathbf{\in}$ $%
\mathbb{R}
^{d}$ and $k_{\mathbf{x}_{2_{i\ast}}}\mathbf{\in}$ $%
\mathbb{R}
^{d}$ be reproducing kernels for extreme points $\mathbf{x}_{1_{i\ast}%
}\mathbf{\in}$ $%
\mathbb{R}
^{d}$ and $\mathbf{x}_{2_{i\ast}}\mathbf{\in}$ $%
\mathbb{R}
^{d}$ located within either overlapping regions or near tail regions of
distributions determined by certain probability density functions $p\left(
\mathbf{x};\omega_{1}\right)  $ and $p\left(  \mathbf{x};\omega_{2}\right)  $,
such that $\mathbf{x}_{1_{\ast}}\mathbf{\sim}$ $p\left(  \mathbf{x};\omega
_{1}\right)  $ and $\mathbf{x}_{2_{\ast}}\mathbf{\sim}$ $p\left(
\mathbf{x};\omega_{2}\right)  $, where a reproducing kernel $k_{\mathbf{x}%
}\left(  \mathbf{s}\right)  $ is recognized as a vector that has the form of
either $k_{\mathbf{x}}\left(  \mathbf{s}\right)  =\left(  \mathbf{s}%
^{T}\mathbf{x}+1\right)  ^{2}$ or $k_{\mathbf{x}}\left(  \mathbf{s}\right)
=\exp\left(  -\gamma\left\Vert \mathbf{s}-\mathbf{x}\right\Vert ^{2}\right)
$, wherein $0.01\leq\gamma\leq0.1$. Also, let $\psi_{1_{i_{\ast}}}$ and
$\psi_{2_{i_{\ast}}}$ be scale factors for $k_{\mathbf{x}_{1_{i\ast}}}$ and
$k_{\mathbf{x}_{2_{i\ast}}}$ respectively, and let $l_{1}$ and $l_{2}$ be
finite numbers.

The following definition expresses the idea of a geometric locus of a novel
principal eigenaxis.

\begin{definition}
\label{Geometric Locus of a Novel Principal Eigenaxis Definition}The
expression%
\begin{align*}
\boldsymbol{\rho}  &  =\sum\nolimits_{i=1}^{l_{1}}\psi_{1_{i_{\ast}}%
}k_{\mathbf{x}_{1_{i\ast}}}-\sum\nolimits_{i=1}^{l_{2}}\psi_{2_{i_{\ast}}%
}k_{\mathbf{x}_{2_{i\ast}}}\\
&  =\boldsymbol{\rho}_{1}-\boldsymbol{\rho}_{2}%
\end{align*}
is said to be a geometric locus of a novel principal eigenaxis, structured as
a dual locus of likelihood components and principal eigenaxis components
$\psi_{1_{i_{\ast}}}k_{\mathbf{x}_{1_{i\ast}}}$ and $\psi_{2_{i_{\ast}}%
}k_{\mathbf{x}_{2_{i\ast}}}$, if and only if the expression $\boldsymbol{\rho
}=\sum\nolimits_{i=1}^{l_{1}}\psi_{1_{i_{\ast}}}k_{\mathbf{x}_{1_{i\ast}}%
}-\sum\nolimits_{i=1}^{l_{2}}\psi_{2_{i_{\ast}}}k_{\mathbf{x}_{2_{i\ast}}}$
represents $\left(  1\right)  $ a discriminant function of a minimum risk
binary classification system that is subject to two categories $\omega_{1}$
and $\omega_{2}$ of random vectors $\mathbf{x\in}$ $%
\mathbb{R}
^{d}$, such that $\mathbf{x\sim}$ $p\left(  \mathbf{x};\omega_{1}\right)  $
and $\mathbf{x\sim}$ $p\left(  \mathbf{x};\omega_{2}\right)  $, where
distributions of the random vectors $\mathbf{x}$ are determined by certain
probability density functions $p\left(  \mathbf{x};\omega_{1}\right)  $ and
$p\left(  \mathbf{x};\omega_{2}\right)  $; $\left(  2\right)  $ an exclusive
principal eigen-coordinate system of the geometric locus of the decision
boundary of the system, so that all of the points that lie on the geometric
locus of the decision boundary exclusively reference the novel principal
eigenaxis $\boldsymbol{\rho}=\boldsymbol{\rho}_{1}\mathbf{-}\boldsymbol{\rho
}_{2}$; and $\left(  3\right)  $ an eigenaxis of symmetry that spans the
decision space of the system; at which point the discriminant function and the
exclusive principal eigen-coordinate system and the eigenaxis of symmetry are
jointly determined by a locus of signed and scaled extreme vectors
$\psi_{1_{i_{\ast}}}k_{\mathbf{x}_{1_{i\ast}}}$ and $-\psi_{2_{i_{\ast}}%
}k_{\mathbf{x}_{2_{i\ast}}}$, wherein each scale factor $\psi_{1_{i_{\ast}}}$
or $\psi_{2_{i_{\ast}}}$ has a value that determines a scaled extreme vector
$\psi_{1_{i_{\ast}}}k_{\mathbf{x}_{1_{i\ast}}}$ or $\psi_{2_{i_{\ast}}%
}k_{\mathbf{x}_{2_{i\ast}}}$, such that every scaled extreme vector
$\psi_{1_{i_{\ast}}}k_{\mathbf{x}_{1_{i\ast}}}$ and $\psi_{2_{i_{\ast}}%
}k_{\mathbf{x}_{2_{i\ast}}}$ represents a principal eigenaxis component on the
exclusive principal eigen-coordinate system $\boldsymbol{\rho}%
=\boldsymbol{\rho}_{1}-\boldsymbol{\rho}_{2}$ that determines a likely
location for a correlated extreme point $\mathbf{x}_{1_{i\ast}}$ or
$\mathbf{x}_{2_{i\ast}}$, along with a likelihood component that determines a
likelihood value for the correlated extreme point $\mathbf{x}_{1_{i\ast}}$ or
$\mathbf{x}_{2_{i\ast}}$.
\end{definition}

\subsection{Dual Locus of a Binary Classification System}

A geometric locus of a novel principal eigenaxis of a minimum risk binary
classification system is said to be the dual locus of the system---which we
define next.

\begin{definition}
\label{Dual Locus Definition}A geometric locus of a novel principal eigenaxis
of a minimum risk binary classification system that is subject to two
categories $\omega_{1}$ and $\omega_{2}$ of random vectors $\mathbf{x\in}$ $%
\mathbb{R}
^{d}$, such that $\mathbf{x\sim}$ $p\left(  \mathbf{x};\omega_{1}\right)  $
and $\mathbf{x\sim}$ $p\left(  \mathbf{x};\omega_{2}\right)  $, where
distributions of the random vectors $\mathbf{x}$ are determined by certain
probability density functions $p\left(  \mathbf{x};\omega_{1}\right)  $ and
$p\left(  \mathbf{x};\omega_{2}\right)  $, is said to be the dual locus of the
system if and only if the discriminant function of the system and the
exclusive principal eigen-coordinate system of the geometric locus of the
decision boundary of the system and the eigenaxis of symmetry that spans the
decision space of the system are jointly represented by the geometric locus of
the novel principal eigenaxis, at which point the discriminant function and
the exclusive principal eigen-coordinate system and the eigenaxis of symmetry
are dual components that exhibit distinctive properties and functionality.
\end{definition}

Theorem \ref{Geometric Locus of a Novel Principal Eigenaxis Theorem} expresses
the statistical structure and the functionality of a geometric locus of a
novel principal eigenaxis.

\begin{theorem}
\label{Geometric Locus of a Novel Principal Eigenaxis Theorem}Take the
discriminant function of any given minimum risk binary classification system
that is subject to random inputs $\mathbf{x\in}$ $%
\mathbb{R}
^{d}$ such that $\mathbf{x\sim}$ $p\left(  \mathbf{x};\omega_{1}\right)  $ and
$\mathbf{x\sim}$ $p\left(  \mathbf{x};\omega_{2}\right)  $, where $p\left(
\mathbf{x};\omega_{1}\right)  $ and $p\left(  \mathbf{x};\omega_{2}\right)  $
are certain probability density functions for two classes $\omega_{1}$ and
$\omega_{2}$ of random vectors $\mathbf{x\in}$ $%
\mathbb{R}
^{d}$.

The discriminant function is represented by a geometric locus of a novel
principal eigenaxis%
\begin{align}
\boldsymbol{\rho}  &  =\sum\nolimits_{i=1}^{l_{1}}\psi_{1_{i_{\ast}}%
}k_{\mathbf{x}_{1_{i\ast}}}-\sum\nolimits_{i=1}^{l_{2}}\psi_{2_{i_{\ast}}%
}k_{\mathbf{x}_{2_{i\ast}}}\tag{8.1}\label{Novel Principal Eigenaxis}\\
&  =\boldsymbol{\rho}_{1}-\boldsymbol{\rho}_{2}\nonumber
\end{align}
structured as a locus of signed and scaled extreme vectors $\psi_{1_{i_{\ast}%
}}k_{\mathbf{x}_{1_{i\ast}}}$ and $-\psi_{2_{i_{\ast}}}k_{\mathbf{x}%
_{2_{i\ast}}}$, so that a dual locus $\boldsymbol{\rho}=\boldsymbol{\rho}%
_{1}\mathbf{-}\boldsymbol{\rho}_{2}$ of likelihood components $\psi
_{1_{i_{\ast}}}k_{\mathbf{x}_{1_{i\ast}}}$ and $\psi_{2_{i_{\ast}}%
}k_{\mathbf{x}_{2_{i\ast}}}$ and principal eigenaxis components $\psi
_{1_{i_{\ast}}}k_{\mathbf{x}_{1_{i\ast}}}$ and $\psi_{2_{i_{\ast}}%
}k_{\mathbf{x}_{2_{i\ast}}}$ represents an exclusive principal
eigen-coordinate system of the geometric locus of the decision boundary of the
system, such that all of the points that lie on the geometric locus of the
decision boundary exclusively reference the novel principal eigenaxis
$\boldsymbol{\rho}=\boldsymbol{\rho}_{1}\mathbf{-}\boldsymbol{\rho}_{2}$, and
also represents an eigenaxis of symmetry $\boldsymbol{\rho}=\boldsymbol{\rho
}_{1}\mathbf{-}\boldsymbol{\rho}_{2}$ that spans the decision space of the
system, at which point each scale factor $\psi_{1_{i_{\ast}}}$ or
$\psi_{2_{i_{\ast}}}$ determines a scaled extreme vector $\psi_{1_{i_{\ast}}%
}k_{\mathbf{x}_{1_{i\ast}}}$ or $\psi_{2_{i_{\ast}}}k_{\mathbf{x}_{2_{i\ast}}%
}$ that represents a principal eigenaxis component that determines a likely
location for a correlated extreme point $\mathbf{x}_{1_{i\ast}}\mathbf{\sim}$
$p\left(  \mathbf{x};\omega_{1}\right)  $ or $\mathbf{x}_{2_{i\ast}%
}\mathbf{\sim}$ $p\left(  \mathbf{x};\omega_{2}\right)  $, along with a
likelihood component that determines a likelihood value for the correlated
extreme point $\mathbf{x}_{1_{i\ast}}$ or $\mathbf{x}_{2_{i\ast}}$, where the
reproducing kernel for each extreme point $k_{\mathbf{x}_{1_{i\ast}}}$ and
$k_{\mathbf{x}_{2_{i\ast}}}$ has the preferred form of either $k_{\mathbf{x}%
}\left(  \mathbf{s}\right)  =\left(  \mathbf{s}^{T}\mathbf{x}+1\right)  ^{2}$
or $k_{\mathbf{x}}\left(  \mathbf{s}\right)  =\exp\left(  -\gamma\left\Vert
\mathbf{s}-\mathbf{x}\right\Vert ^{2}\right)  $, wherein $0.01\leq\gamma
\leq0.1$.
\end{theorem}

\begin{proof}
We prove Theorem \ref{Geometric Locus of a Novel Principal Eigenaxis Theorem}
by a constructive proof that demonstrates how a well-posed constrained
optimization algorithm finds a geometric locus of a novel principal eigenaxis.
\end{proof}

\subsection{A\ Deep-rooted Locus Problem in Binary Classification}

We have discovered that the general locus formula that resolves the direct
problem---of the binary classification of random vectors---is the source of
deep-seated locus problems in binary classification that \emph{can only be
resolved} by a well-posed constrained optimization algorithm that
\emph{finds}\textbf{ }and thereby \emph{determines} the \emph{parameters} of
the \emph{general locus formula}.

We have also discovered that the constrained optimization algorithm that
resolves the inverse problem of the binary classification of random vectors
executes novel and elegant processes---that include a novel principal
eigen-coordinate transform algorithm---which represent the solution for
finding discriminant functions of minimum risk binary classification
systems---at which point the direct problem is transformed into a feasible one.

We now outline the process by which a novel principal eigen-coordinate
transform algorithm resolves what we consider to be a deep-rooted locus
problem in binary classification---that \emph{must} be \emph{resolved} to find
discriminant functions of minimum risk binary classifications systems. We
begin by motivating the theoretical and empirical basis behind the novel
principal eigen-coordinate transform algorithm---which we use to find
discriminant functions of minimum risk binary classifications systems.

\section{\label{Section 9}A\ Novel Eigen-coordinate Transform Algorithm}

We have previously demonstrated that a pair of signed random quadratic forms%
\[
\mathbf{x}^{T}\mathbf{\Sigma}_{1}^{-1}\mathbf{x-x}^{T}\mathbf{\Sigma}_{2}%
^{-1}\mathbf{x}%
\]
jointly provide dual representation of the discriminant function and the
intrinsic coordinate system of the geometric locus of the decision boundary of
any given minimum risk binary classification system that is subject to two
categories of multivariate normal vectors $\mathbf{x\in}$ $%
\mathbb{R}
^{d}$, such that $\mathbf{x\sim}p\left(  \mathbf{x};\boldsymbol{\mu}%
_{1},\mathbf{\Sigma}_{1}\right)  $ and $\mathbf{x\sim}p\left(  \mathbf{x}%
;\boldsymbol{\mu}_{2},\mathbf{\Sigma}_{2}\right)  $.

We have also demonstrated that the pair of signed random quadratic forms
$\mathbf{x}^{T}\mathbf{\Sigma}_{1}^{-1}\mathbf{x-x}^{T}\mathbf{\Sigma}%
_{2}^{-1}\mathbf{x}$ can be represented by a pair of random quadratic forms
$\mathbf{v}^{T}\mathbf{Qv}$ and $\mathbf{v}^{T}\mathbf{Q}^{-1}\mathbf{v}$,
such that%
\begin{align*}
\mathbf{v}^{T}\mathbf{Qv\ }  &  \mathbf{\equiv v}^{T}\mathbf{Q}^{-1}%
\mathbf{v}\\
&  \mathbf{\equiv\ }\mathbf{x}^{T}\mathbf{\Sigma}_{1}^{-1}\mathbf{x-x}%
^{T}\mathbf{\Sigma}_{2}^{-1}\mathbf{x}\text{\textbf{,}}%
\end{align*}
so that the elements of a joint covariance matrix $\mathbf{Q}$ and the
inverted joint covariance matrix $\mathbf{Q}^{-1}$ both describe differences
between joint variabilities of normal random vectors $\mathbf{x\sim}p\left(
\mathbf{x};\boldsymbol{\mu}_{1},\mathbf{\Sigma}_{1}\right)  $ and
$\mathbf{x\sim}p\left(  \mathbf{x};\boldsymbol{\mu}_{2},\mathbf{\Sigma}%
_{2}\right)  $ that belong to a collection of two categories $\omega_{1}$ and
$\omega_{2}$ of normal random vectors $\mathbf{x\in}$ $%
\mathbb{R}
^{d}$, such that the magnitude and the direction of the principal eigenvector
$\mathbf{v=v}_{1}+\mathbf{v}_{2}$ of the joint covariance matrix $\mathbf{Q}$
and the inverted joint covariance matrix $\mathbf{Q}^{-1}$ are both functions
of differences between joint variabilities of normal \emph{extreme vectors}
$\mathbf{x}_{1_{i\ast}}\mathbf{\sim}p\left(  \mathbf{x};\boldsymbol{\mu}%
_{1},\mathbf{\Sigma}_{1}\right)  $ and $\mathbf{x}_{2_{i\ast}}\mathbf{\sim
}p\left(  \mathbf{x};\boldsymbol{\mu}_{2},\mathbf{\Sigma}_{2}\right)  $.

Moreover, given the novel principal eigen-coordinate transform method
expressed by Theorem \ref{Principal Eigen-coordinate System Theorem} and
Corollary \ref{Principal Eigen-coordinate System Corollary}, we are guaranteed
the existence of an exclusive principal eigen-coordinate system---which is the
principal part of an equivalent representation of a certain quadratic
form---that is the solution of an equivalent form of the vector algebra locus
equation of the geometric locus of a certain quadratic curve or surface, such
that the exclusive principal eigen-coordinate system is the principal
eigenaxis of the geometric locus of the quadratic curve or surface, so that
the principal eigenaxis satisfies the geometric locus of the quadratic curve
or surface in terms of its total allowed eigenenergy, and the uniform property
exhibited by all of the points that lie on the geometric locus of the
quadratic curve or surface is the total allowed eigenenergy exhibited by the
principal eigenaxis of the geometric locus of the quadratic curve or surface.

Furthermore, conditions expressed by Theorems
\ref{Vector Algebra Equation of Linear Loci Theorem} -
\ref{Vector Algebra Equation of Spherical Loci Theorem} substantiate Theorem
\ref{Principal Eigen-coordinate System Theorem} and Corollary
\ref{Principal Eigen-coordinate System Corollary}---since Theorems
\ref{Vector Algebra Equation of Linear Loci Theorem} -
\ref{Vector Algebra Equation of Spherical Loci Theorem} guarantee the
existence of the general vector algebra locus equations of (\ref{Linear Locus}%
) - (\ref{Spherical Locus}), wherein the principal eigenaxis of the locus of
any given quadratic curve or surface provides an exclusive principal
eigen-coordinate system for the locus of the quadratic curve or surface, such
that all of the points that lie on the locus of the quadratic curve or surface
explicitly and exclusively reference the principal eigenaxis and also satisfy
the total allowed eigenenergy exhibited by the principal eigenaxis, at which
point the principal eigenaxis satisfies the locus of the quadratic curve or
surface in terms of its eigenenergy, and the uniform property exhibited by all
of the points that lie on the locus of the quadratic curve or surface is the
eigenenergy exhibited by the principal eigenaxis.

Therefore, let the pair of random quadratic forms $\mathbf{v}^{T}\mathbf{Qv}$
and $\mathbf{v}^{T}\mathbf{Q}^{-1}\mathbf{v}$ that jointly represent the pair
of signed random quadratic forms
\[
\mathbf{x}^{T}\mathbf{\Sigma}_{1}^{-1}\mathbf{x-x}^{T}\mathbf{\Sigma}_{2}%
^{-1}\mathbf{x}%
\]
be solutions of a system of well-posed vector algebra locus equations.

By Theorem \ref{Principal Eigen-coordinate System Theorem} and Corollaries
\ref{Principal Eigen-coordinate System Corollary} -
\ref{Symmetrical and Equivalent Principal Eigenaxes}, it follows that a
principal eigenaxis $\widetilde{\mathbf{v}}\ \mathbf{=\ }\widetilde{\mathbf{v}%
}_{1}-\widetilde{\mathbf{v}}_{2}$ of a certain quadratic curve or surface is
the principal part of an equivalent representation of the pair of random
quadratic forms $\mathbf{v}^{T}\mathbf{Qv}$ and $\mathbf{v}^{T}\mathbf{Q}%
^{-1}\mathbf{v}$, such that the principal eigenaxis $\widetilde{\mathbf{v}%
}\ \mathbf{=\ }\widetilde{\mathbf{v}}_{1}-\widetilde{\mathbf{v}}_{2}$ of the
quadratic curve or surface is symmetrically and equivalently related to the
principal eigenvector $\mathbf{v=v}_{1}+\mathbf{v}_{2}$ of the joint
covariance matrix $\mathbf{Q}$ of the random quadratic form $\mathbf{v}%
^{T}\mathbf{Qv}$ \emph{and} the inverted joint covariance matrix
$\mathbf{Q}^{-1}$ of the random quadratic form $\mathbf{v}^{T}\mathbf{Q}%
^{-1}\mathbf{v}$, so that the principal eigenaxis $\widetilde{\mathbf{v}%
}\ \mathbf{=\ }\widetilde{\mathbf{v}}_{1}-\widetilde{\mathbf{v}}_{2}$
satisfies the quadratic curve or surface in terms of its total allowed
eigenenergy $\left\Vert \widetilde{\mathbf{v}}_{1}-\widetilde{\mathbf{v}}%
_{2}\right\Vert ^{2}$---which is regulated by the eigenvalues of the symmetric
matrices $\mathbf{Q}$ and $\mathbf{Q}^{-1}$ of the random quadratic forms
$\mathbf{v}^{T}\mathbf{Qv}$ and $\mathbf{v}^{T}\mathbf{Q}^{-1}\mathbf{v}$,
wherein the magnitude and the direction of the principal eigenaxis
$\widetilde{\mathbf{v}}\ \mathbf{=\ }\widetilde{\mathbf{v}}_{1}%
-\widetilde{\mathbf{v}}_{2}$ are both functions of differences between joint
variabilities of normal \emph{extreme vectors} $\mathbf{x}_{1_{i\ast}%
}\mathbf{\sim}p\left(  \mathbf{x};\boldsymbol{\mu}_{1},\mathbf{\Sigma}%
_{1}\right)  $ and $\mathbf{x}_{2_{i\ast}}\mathbf{\sim}p\left(  \mathbf{x}%
;\boldsymbol{\mu}_{2},\mathbf{\Sigma}_{2}\right)  $.

We have generalized these findings by letting the elements of $\mathbf{Q}$ and
$\mathbf{Q}^{-1}$ both describe differences between joint variabilities of
random vectors $\mathbf{x\sim}p\left(  \mathbf{x};\omega_{1}\right)  $ and
$\mathbf{x\sim}p\left(  \mathbf{x};\omega_{1}\right)  $ that belong to a
collection of two categories $\omega_{1}$ and $\omega_{2}$ of random vectors
$\mathbf{x\in}$ $%
\mathbb{R}
^{d}$, where $p\left(  \mathbf{x};\omega_{1}\right)  $ and $p\left(
\mathbf{x};\omega_{2}\right)  $ are certain probability density functions for
the two classes $\omega_{1}$ and $\omega_{2}$ of random vectors $\mathbf{x}$.
Thereby, the magnitude and the direction of the principal eigenvector
$\mathbf{v=v}_{1}+\mathbf{v}_{2}$ of the joint covariance matrix $\mathbf{Q}$
and the inverted joint covariance matrix $\mathbf{Q}^{-1}$ are both functions
of differences between joint variabilities of extreme vectors $\mathbf{x}%
_{1_{i\ast}}\mathbf{\sim}p\left(  \mathbf{x};\omega_{1}\right)  $ and
$\mathbf{x}_{2_{i\ast}}\mathbf{\sim}p\left(  \mathbf{x};\omega_{1}\right)  $
that belong to the classes $\omega_{1}$ and $\omega_{2}$ of random vectors
$\mathbf{x\in}$ $%
\mathbb{R}
^{d}$.

Accordingly, let the pair of random quadratic forms $\mathbf{v}^{T}%
\mathbf{Qv}$ and $\mathbf{v}^{T}\mathbf{Q}^{-1}\mathbf{v}$ be solutions of a
system of well-posed vector algebra locus equations, such that $\mathbf{v=v}%
_{1}+\mathbf{v}_{2}$ is the principal eigenvector of both $\mathbf{Q}$ and
$\mathbf{Q}^{-1}$, so that the magnitude and the direction of the principal
eigenvector $\mathbf{v=v}_{1}+\mathbf{v}_{2}$ of $\mathbf{Q}$ and
$\mathbf{Q}^{-1}$ are both functions of differences between joint
variabilities of extreme vectors $\mathbf{x}_{1_{i\ast}}\mathbf{\sim}p\left(
\mathbf{x};\omega_{1}\right)  $ and $\mathbf{x}_{2_{i\ast}}\mathbf{\sim
}p\left(  \mathbf{x};\omega_{1}\right)  $ that belong to the classes
$\omega_{1}$ and $\omega_{2}$ of random vectors $\mathbf{x\in}$ $%
\mathbb{R}
^{d}$.

By Theorem \ref{Principal Eigen-coordinate System Theorem} and Corollaries
\ref{Principal Eigen-coordinate System Corollary} -
\ref{Symmetrical and Equivalent Principal Eigenaxes}, it follows that a
principal eigenaxis $\widetilde{\mathbf{v}}\ \mathbf{=\ }\widetilde{\mathbf{v}%
}_{1}-\widetilde{\mathbf{v}}_{2}$ of a certain quadratic curve or surface is
the principal part of an equivalent representation of the pair of random
quadratic forms $\mathbf{v}^{T}\mathbf{Qv}$ and $\mathbf{v}^{T}\mathbf{Q}%
^{-1}\mathbf{v}$, such that the principal eigenaxis $\widetilde{\mathbf{v}%
}\ \mathbf{=\ }\widetilde{\mathbf{v}}_{1}-\widetilde{\mathbf{v}}_{2}$ of the
quadratic curve or surface is symmetrically and equivalently related to the
principal eigenvector $\mathbf{v=v}_{1}+\mathbf{v}_{2}$ of the joint
covariance matrix $\mathbf{Q}$ of the random quadratic form $\mathbf{v}%
^{T}\mathbf{Qv}$ \emph{and} the inverted joint covariance matrix
$\mathbf{Q}^{-1}$ of the random quadratic form $\mathbf{v}^{T}\mathbf{Q}%
^{-1}\mathbf{v}$, so that the principal eigenaxis $\widetilde{\mathbf{v}%
}\ \mathbf{=\ }\widetilde{\mathbf{v}}_{1}-\widetilde{\mathbf{v}}_{2}$
satisfies the quadratic curve or surface in terms of its total allowed
eigenenergy $\left\Vert \widetilde{\mathbf{v}}_{1}-\widetilde{\mathbf{v}}%
_{2}\right\Vert ^{2}$---which is regulated by the eigenvalues of the symmetric
matrices $\mathbf{Q}$ and $\mathbf{Q}^{-1}$ of the random quadratic forms
$\mathbf{v}^{T}\mathbf{Qv}$ and $\mathbf{v}^{T}\mathbf{Q}^{-1}\mathbf{v}$,
wherein the magnitude and the direction of the principal eigenaxis
$\widetilde{\mathbf{v}}\ \mathbf{=\ }\widetilde{\mathbf{v}}_{1}%
-\widetilde{\mathbf{v}}_{2}$ are both functions of differences between joint
variabilities of extreme vectors $\mathbf{x}_{1_{i\ast}}\mathbf{\sim}p\left(
\mathbf{x};\omega_{1}\right)  $ and $\mathbf{x}_{2_{i\ast}}\mathbf{\sim
}p\left(  \mathbf{x};\omega_{1}\right)  $ that belong to the classes
$\omega_{1}$ and $\omega_{2}$ of random vectors $\mathbf{x\in}$ $%
\mathbb{R}
^{d}$.

Even more, Theorem
\ref{Geometric Locus of a Novel Principal Eigenaxis Theorem} promises us that
a geometric locus of a novel principal eigenaxis%
\begin{align*}
\boldsymbol{\rho}  &  =\sum\nolimits_{i=1}^{l_{1}}\psi_{1_{i_{\ast}}%
}k_{\mathbf{x}_{1_{i\ast}}}-\sum\nolimits_{i=1}^{l_{2}}\psi_{2_{i_{\ast}}%
}k_{\mathbf{x}_{2_{i\ast}}}\\
&  =\boldsymbol{\rho}_{1}-\boldsymbol{\rho}_{2}%
\end{align*}
provides dual representation of the discriminant function and an exclusive
principal coordinate eigen-coordinate system of the geometric locus of the
decision boundary of \emph{any} given minimum risk binary classification
system, along with an eigenaxis of symmetry that spans the decision space of
the given system, such that the geometric locus of the decision boundary of
the system is the geometric locus of a certain quadratic curve or surface.

Moreover, we have conducted numerous simulation studies and corresponding
analysis that substantiate Theorem
\ref{Geometric Locus of a Novel Principal Eigenaxis Theorem}, wherein a
well-posed constrained optimization algorithm finds the geometrical and
statistical components of a geometric locus of a novel principal eigenaxis
\citep{Reeves2018design}%
.

Finally, conditions expressed by Axiom
\ref{Rotation of Intrinsic Coordinate Axes Axiom}, Lemma
\ref{Equivalent Form of Algebraic Equation Lemma}, Theorem
\ref{Principal Eigen-coordinate System Theorem}, Corollaries
\ref{Principal Eigen-coordinate System Corollary} -
\ref{Symmetrical and Equivalent Principal Eigenaxes} and Theorem
\ref{Geometric Locus of a Novel Principal Eigenaxis Theorem} provide us with a
collective guarantee that we \textbf{can determine} an \emph{equivalent form}
of the vector algebra locus equation of (\ref{Norm_Dec_Bound})%
\begin{align*}
d\left(  \mathbf{x}\right)   &  :\mathbf{x}^{T}\mathbf{\Sigma}_{1}%
^{-1}\mathbf{x}-2\mathbf{x}^{T}\mathbf{\Sigma}_{1}^{-1}\boldsymbol{\mu}%
_{1}+\boldsymbol{\mu}_{1}^{T}\mathbf{\Sigma}_{1}^{-1}\boldsymbol{\mu}_{1}%
-\ln\left(  \left\vert \mathbf{\Sigma}_{1}\right\vert \right) \\
&  -\mathbf{x}^{T}\mathbf{\Sigma}_{2}^{-1}\mathbf{x}+2\mathbf{x}%
^{T}\mathbf{\Sigma}_{2}^{-1}\boldsymbol{\mu}_{2}\mathbf{-}\boldsymbol{\mu}%
_{2}^{T}\mathbf{\Sigma}_{2}^{-1}\boldsymbol{\mu}_{2}+\ln\left(  \left\vert
\mathbf{\Sigma}_{2}\right\vert \right)  =0
\end{align*}
\emph{if} we can determine \emph{how} to \emph{transform} the basis of the
intrinsic coordinate system $\mathbf{x}^{T}\mathbf{\Sigma}_{1}^{-1}%
\mathbf{x-x}^{T}\mathbf{\Sigma}_{2}^{-1}\mathbf{x}$, such that $\mathbf{x\sim
}p\left(  \mathbf{x};\boldsymbol{\mu}_{1},\mathbf{\Sigma}_{1}\right)  $ and
$\mathbf{x\sim}p\left(  \mathbf{x};\boldsymbol{\mu}_{2},\mathbf{\Sigma}%
_{2}\right)  $, \emph{into} a locus of signed and scaled extreme vectors
$\psi_{1_{i_{\ast}}}k_{\mathbf{x}_{1_{i\ast}}}$ and $-\psi_{2_{i_{\ast}}%
}k_{\mathbf{x}_{2_{i\ast}}}$%
\begin{align*}
\boldsymbol{\rho}  &  =\sum\nolimits_{i=1}^{l_{1}}\psi_{1_{i_{\ast}}%
}k_{\mathbf{x}_{1_{i\ast}}}-\sum\nolimits_{i=1}^{l_{2}}\psi_{2_{i_{\ast}}%
}k_{\mathbf{x}_{2_{i\ast}}}\\
&  =\boldsymbol{\rho}_{1}-\boldsymbol{\rho}_{2}\text{,}%
\end{align*}
so that likelihood values and likely locations of a collection of extreme
points $\mathbf{x}_{1_{\ast}}\mathbf{\sim}p\left(  \mathbf{x};\boldsymbol{\mu
}_{1},\mathbf{\Sigma}_{1}\right)  $ and $\mathbf{x}_{2_{\ast}}\mathbf{\sim
}p\left(  \mathbf{x};\boldsymbol{\mu}_{2},\mathbf{\Sigma}_{2}\right)  $
\emph{determine} the \emph{positions} of the \emph{basis} of the
\emph{transformed} intrinsic coordinate system $\mathbf{x}^{T}\mathbf{\Sigma
}_{1}^{-1}\mathbf{x-\mathbf{x}^{T}\mathbf{\Sigma}_{2}^{-1}\mathbf{x}}$, such
that a geometric locus of a novel principal eigenaxis $\boldsymbol{\rho}$
$=\boldsymbol{\rho}_{1}-\boldsymbol{\rho}_{2}$ is the solution of the
equivalent form of (\ref{Norm_Dec_Bound}), at which point the geometric locus
of the novel principal eigenaxis $\boldsymbol{\rho}=\sum\nolimits_{i=1}%
^{l_{1}}\psi_{1_{i_{\ast}}}k_{\mathbf{x}_{1_{i\ast}}}-\sum\nolimits_{i=1}%
^{l_{2}}\psi_{2_{i_{\ast}}}k_{\mathbf{x}_{2_{i\ast}}}$ is the principal part
of an equivalent representation of a pair of random quadratic forms
$\mathbf{v}^{T}\mathbf{Qv}$ and $\mathbf{v}^{T}\mathbf{Q}^{-1}\mathbf{v}$
associated with a joint covariance matrix $\mathbf{Q}$ and the inverted joint
covariance matrix $\mathbf{Q}^{-1}$, such that $\mathbf{v=v}_{1}%
+\mathbf{v}_{2}$ is the principal eigenvector of the joint covariance matrix
$\mathbf{Q}$ and the inverted joint covariance matrix $\mathbf{Q}^{-1}$.

Even so, \textbf{how} do we \textbf{implement} an \emph{algorithm} that is
subject to essential criterion in the novel principal eigen-coordinate
transform method expressed by Theorem
\ref{Principal Eigen-coordinate System Theorem} and Corollaries
\ref{Principal Eigen-coordinate System Corollary} -
\ref{Symmetrical and Equivalent Principal Eigenaxes}?

\subsection{A\ System of Well-posed Vector Algebra Locus Equations}

Given the guarantees expressed by Theorem
\ref{Principal Eigen-coordinate System Theorem} and Corollaries
\ref{Principal Eigen-coordinate System Corollary} -
\ref{Symmetrical and Equivalent Principal Eigenaxes}, take any given random
quadratic form $\mathbf{v}^{T}\mathbf{Qv}$, such that $\left(  1\right)  $
$\mathbf{v=v}_{1}+\mathbf{v}_{2}$ is the principal eigenvector of $\mathbf{Q}$
and $\mathbf{Q}^{-1}$; and $\left(  2\right)  $ the elements of $\mathbf{Q}$
and $\mathbf{Q}^{-1}$ both describe differences between joint variabilities of
random vectors $\mathbf{x\sim}p\left(  \mathbf{x};\omega_{1}\right)  $ and
$\mathbf{x\sim}p\left(  \mathbf{x};\omega_{1}\right)  $ that belong to two
classes $\omega_{1}$ and $\omega_{2}$, where $p\left(  \mathbf{x};\omega
_{1}\right)  $ and $p\left(  \mathbf{x};\omega_{2}\right)  $ are certain
probability density functions for the two classes $\omega_{1}$ and $\omega
_{2}$ of random vectors $\mathbf{x\in}$ $%
\mathbb{R}
^{d}$.

We have discovered a system of well-posed vector algebra locus equations that
are satisfied by the pair of random quadratic forms $\mathbf{v}^{T}%
\mathbf{Qv}$ and $\mathbf{v}^{T}\mathbf{Q}^{-1}\mathbf{v}$---wherein the
principal eigenvector $\mathbf{v=v}_{1}+\mathbf{v}_{2}$ of $\mathbf{Q}$ and
$\mathbf{Q}^{-1}$ is symmetrically and equivalently related to the principal
part of an equivalent representation of the random quadratic forms
$\mathbf{v}^{T}\mathbf{Qv}$ and $\mathbf{v}^{T}\mathbf{Q}^{-1}\mathbf{v}$,
such that an exclusive principal eigen-coordinate system---structured as a
locus of signed and scaled extreme vectors $\psi_{1_{i_{\ast}}}k_{\mathbf{x}%
_{1_{i\ast}}}$ and $-\psi_{2_{i_{\ast}}}k_{\mathbf{x}_{2_{i\ast}}}$%
\begin{align*}
\boldsymbol{\rho}  &  =\sum\nolimits_{i=1}^{l_{1}}\psi_{1_{i_{\ast}}%
}k_{\mathbf{x}_{1_{i\ast}}}-\sum\nolimits_{i=1}^{l_{2}}\psi_{2_{i_{\ast}}%
}k_{\mathbf{x}_{2_{i\ast}}}\\
&  =\boldsymbol{\rho}_{1}-\boldsymbol{\rho}_{2}%
\end{align*}
is the \emph{solution} of the system of well-posed vector algebra locus
equations, at which point the geometric locus of the novel principal eigenaxis
$\boldsymbol{\rho}$ $=\boldsymbol{\rho}_{1}-\boldsymbol{\rho}_{2}$ is
symmetrically and equivalently related to the principal eigenvector
$\mathbf{v=v}_{1}+\mathbf{v}_{2}$ of the joint covariance matrix $\mathbf{Q}$
and the inverted joint covariance matrix $\mathbf{Q}^{-1}$, such that the
geometric locus of the novel principal eigenaxis $\boldsymbol{\rho}$
$=\boldsymbol{\rho}_{1}-\boldsymbol{\rho}_{2}$ is the principal eigenaxis of
the decision boundary of a minimum risk binary classification system, so that
the novel principal eigenaxis $\boldsymbol{\rho}$ $=\boldsymbol{\rho}%
_{1}-\boldsymbol{\rho}_{2}$ satisfies the geometric locus of the decision
boundary of the minimum risk binary classification system in terms of a
critical minimum eigenenergy $\left\Vert \boldsymbol{\rho}_{1}%
-\boldsymbol{\rho}_{2}\right\Vert ^{2}$, such that the total allowed
eigenenergy $\left\Vert \boldsymbol{\rho}_{1}-\boldsymbol{\rho}_{2}\right\Vert
^{2}$ exhibited by the geometric locus of the novel principal eigenaxis
$\boldsymbol{\rho}$ $=\boldsymbol{\rho}_{1}-\boldsymbol{\rho}_{2}$ is
regulated by the eigenvalues of the symmetric matrices $\mathbf{Q}$ and
$\mathbf{Q}^{-1}$ of the pair of random quadratic forms $\mathbf{v}%
^{T}\mathbf{Qv}$ and $\mathbf{v}^{T}\mathbf{Q}^{-1}\mathbf{v}$.

Accordingly, we have discovered that the geometric locus of the novel
principal eigenaxis $\boldsymbol{\rho}$ $=\boldsymbol{\rho}_{1}%
-\boldsymbol{\rho}_{2}$ is subject to essential criterion in the novel
principal eigen-coordinate transform method expressed by Theorem
\ref{Principal Eigen-coordinate System Theorem} and Corollaries
\ref{Principal Eigen-coordinate System Corollary} -
\ref{Symmetrical and Equivalent Principal Eigenaxes}, along with conditions
expressed by Theorem
\ref{Geometric Locus of a Novel Principal Eigenaxis Theorem}, such that the
geometric locus of the novel principal eigenaxis $\boldsymbol{\rho}$
$=\boldsymbol{\rho}_{1}-\boldsymbol{\rho}_{2}$ is the principal eigenaxis of
the geometric locus of the decision boundary of a minimum risk binary
classification system, so that the geometric locus of the novel principal
eigenaxis $\boldsymbol{\rho}$ $=\boldsymbol{\rho}_{1}-\boldsymbol{\rho}_{2}$
satisfies the geometric locus of the decision boundary in terms of a critical
minimum eigenenergy $\left\Vert \boldsymbol{\rho}_{1}-\boldsymbol{\rho}%
_{2}\right\Vert _{\min_{c}}^{2}$ and a minimum expected risk $\mathfrak{R}%
_{\mathfrak{\min}}\left(  \left\Vert \boldsymbol{\rho}_{1}-\boldsymbol{\rho
}_{2}\right\Vert _{\min_{c}}^{2}\right)  $, at which point the geometric locus
of the novel principal eigenaxis $\boldsymbol{\rho}$ $=\boldsymbol{\rho}%
_{1}-\boldsymbol{\rho}_{2}$ provides dual representation of the discriminant
function, an exclusive principal eigen-coordinate system of the geometric
locus of the decision boundary, and an eigenaxis of symmetry that spans the
decision space of the minimum risk binary classification system.

Even so, we realize that we cannot expect to change the positions of the basis
of \emph{any} given intrinsic coordinate system $\mathbf{x}^{T}\mathbf{\Sigma
}_{1}^{-1}\mathbf{x-x}^{T}\mathbf{\Sigma}_{2}^{-1}\mathbf{x}$, such that
$\mathbf{x\sim}p\left(  \mathbf{x};\boldsymbol{\mu}_{1},\mathbf{\Sigma}%
_{1}\right)  $ and $\mathbf{x\sim}p\left(  \mathbf{x};\boldsymbol{\mu}%
_{2},\mathbf{\Sigma}_{2}\right)  $, in the vector algebra locus equation of
(\ref{Norm_Dec_Bound}) by \emph{simply} \emph{rotating} the coordinates
\emph{axes} of a Cartesian coordinate system.

\emph{So}, \textbf{how} do we \textbf{change} the \textbf{positions} of the
\textbf{basis} of \emph{any} given intrinsic coordinate system $\mathbf{x}%
^{T}\mathbf{\Sigma}_{1}^{-1}\mathbf{x-\mathbf{x}^{T}\mathbf{\Sigma}_{2}%
^{-1}\mathbf{x}}$, such that $\mathbf{x\sim}p\left(  \mathbf{x}%
;\boldsymbol{\mu}_{1},\mathbf{\Sigma}_{1}\right)  $ and $\mathbf{x\sim
}p\left(  \mathbf{x};\boldsymbol{\mu}_{2},\mathbf{\Sigma}_{2}\right)  $---to
which geometric loci of decision boundaries of minimum risk binary
classification systems in (\ref{Gaussian Rule}) are referenced---such that a
geometric locus of a novel principal eigenaxis $\boldsymbol{\rho}$
$=\boldsymbol{\rho}_{1}-\boldsymbol{\rho}_{2}$ is the principal part of an
equivalent representation of a pair of random quadratic forms $\mathbf{v}%
^{T}\mathbf{Qv}$ and $\mathbf{v}^{T}\mathbf{Q}^{-1}\mathbf{v}$ associated with
a joint covariance matrix $\mathbf{Q}$ and the inverted joint covariance
matrix $\mathbf{Q}^{-1}$, where $\mathbf{v=v}_{1}+\mathbf{v}_{2}$ is the
principal eigenvector of $\mathbf{Q}$ and $\mathbf{Q}^{-1}$, so that the
geometric locus of the novel principal eigenaxis $\boldsymbol{\rho}$
$=\boldsymbol{\rho}_{1}-\boldsymbol{\rho}_{2}$ provides \textbf{dual
representation} of the discriminant function, an exclusive principal
eigen-coordinate system of the geometric locus of the decision boundary, and
an eigenaxis of symmetry that spans the decision space---of \emph{any} given
minimum risk binary classification system?

\subsection{A\ Principal Eigen-coordinate Transform Algorithm}

We have discovered that changing the positions of the basis of any given
intrinsic coordinate system $\mathbf{x}^{T}\mathbf{\Sigma}_{1}^{-1}%
\mathbf{x-x}^{T}\mathbf{\Sigma}_{2}^{-1}\mathbf{x}$ in the vector algebra
locus equation of (\ref{Norm_Dec_Bound})---is accomplished by a well-posed
constrained optimization algorithm that executes a \emph{novel principal
eigen-coordinate transform}, wherein a geometric locus of a novel principal
eigenaxis%
\begin{align*}
\boldsymbol{\rho}  &  =\boldsymbol{\rho}_{1}-\boldsymbol{\rho}_{2}\\
&  =\sum\nolimits_{i=1}^{l_{1}}\psi_{1_{i_{\ast}}}k_{\mathbf{x}_{1_{i\ast}}%
}-\sum\nolimits_{i=1}^{l_{2}}\psi_{2_{i_{\ast}}}k_{\mathbf{x}_{2_{i\ast}}}%
\end{align*}
is the principal part of an equivalent representation of a pair of correlated
random quadratic forms $\mathbf{v}^{T}\mathbf{Qv}$ and $\mathbf{v}%
^{T}\mathbf{Q}^{-1}\mathbf{v}$ associated with a joint covariance matrix
$\mathbf{Q}$ and the inverted joint covariance matrix $\mathbf{Q}^{-1}$, such
that $\mathbf{v=v}_{1}+\mathbf{v}_{2}$ is the principal eigenvector of
$\mathbf{Q}$ and $\mathbf{Q}^{-1}$, at which point the geometric locus of the
novel principal eigenaxis $\boldsymbol{\rho}$ $=\boldsymbol{\rho}%
_{1}-\boldsymbol{\rho}_{2}$ provides dual representation of the discriminant
function, an exclusive principal eigen-coordinate system of the geometric
locus of the decision boundary, and an eigenaxis of symmetry that spans the
decision space---of \emph{any} given minimum risk binary classification
system, so that the geometric locus of the novel principal eigenaxis
$\boldsymbol{\rho}$ $=\boldsymbol{\rho}_{1}-\boldsymbol{\rho}_{2}$ satisfies
the geometric locus of the decision boundary in terms of a critical minimum
eigenenergy $\left\Vert \boldsymbol{\rho}_{1}-\boldsymbol{\rho}_{2}\right\Vert
_{\min_{c}}^{2}$ and a minimum expected risk $\mathfrak{R}_{\mathfrak{\min}%
}\left(  \left\Vert \boldsymbol{\rho}_{1}-\boldsymbol{\rho}_{2}\right\Vert
_{\min_{c}}^{2}\right)  $, such that the total allowed eigenenergy $\left\Vert
\boldsymbol{\rho}_{1}-\boldsymbol{\rho}_{2}\right\Vert _{\min_{c}}^{2}$ and
the expected risk $\mathfrak{R}_{\mathfrak{\min}}\left(  \left\Vert
\boldsymbol{\rho}_{1}-\boldsymbol{\rho}_{2}\right\Vert _{\min_{c}}^{2}\right)
$ exhibited by the geometric locus of the novel principal eigenaxis
$\boldsymbol{\rho}$ $=\boldsymbol{\rho}_{1}-\boldsymbol{\rho}_{2}$ are
regulated by the eigenvalues of the symmetric matrices $\mathbf{Q}$ and
$\mathbf{Q}^{-1}$ associated with the pair of random quadratic forms
$\mathbf{v}^{T}\mathbf{Qv}$ and $\mathbf{v}^{T}\mathbf{Q}^{-1}\mathbf{v}$.

We have determined that the algorithm finds the solution for a well-posed
inequality constrained optimization problem, known as the primal problem, such
that the geometric locus of a novel principal eigenaxis $\boldsymbol{\rho
}=\sum\nolimits_{i=1}^{l_{1}}\psi_{1_{i_{\ast}}}k_{\mathbf{x}_{1_{i\ast}}%
}-\sum\nolimits_{i=1}^{l_{2}}\psi_{2_{i_{\ast}}}k_{\mathbf{x}_{2_{i\ast}}}$ is
subject to certain constraints, by using Lagrange multipliers $\psi_{i}\geq0$
and a Lagrangian function, wherein the objective function of the novel
principal eigenaxis $\boldsymbol{\rho}$ $=\boldsymbol{\rho}_{1}%
-\boldsymbol{\rho}_{2}$ and its constraints are combined with each other.

Thereby, we have determined that the algorithm introduces a geometric locus of
a Wolfe dual novel principal eigenaxis%
\begin{align*}
\boldsymbol{\psi}  &  =\sum\nolimits_{i=1}^{l_{1}}\psi_{1i\ast}\frac
{k_{\mathbf{x}_{1i\ast}}}{\left\Vert k_{\mathbf{x}_{1i\ast}}\right\Vert }%
+\sum\nolimits_{i=1}^{l_{2}}\psi_{2i\ast}\frac{k_{\mathbf{x}_{2i\ast}}%
}{\left\Vert k_{\mathbf{x}_{2i\ast}}\right\Vert }\\
&  =\boldsymbol{\psi}_{1}+\boldsymbol{\psi}_{2}%
\end{align*}
that is \emph{symmetrically} and \emph{equivalently} \emph{related to} the
geometric locus of the primal novel principal eigenaxis%
\begin{align*}
\boldsymbol{\rho}  &  =\sum\nolimits_{i=1}^{l_{1}}\psi_{1_{i_{\ast}}%
}k_{\mathbf{x}_{1_{i\ast}}}-\sum\nolimits_{i=1}^{l_{2}}\psi_{2_{i_{\ast}}%
}k_{\mathbf{x}_{2_{i\ast}}}\\
&  =\boldsymbol{\rho}_{1}-\boldsymbol{\rho}_{2}%
\end{align*}
inside a vector space that we have named the \textquotedblleft Wolfe-dual
principal eigenspace,\textquotedblright\ and finds extrema for the
\emph{restriction} of the primal novel principal eigenaxis $\boldsymbol{\rho}$
$=\boldsymbol{\rho}_{1}-\boldsymbol{\rho}_{2}$ \emph{to} the Wolfe-dual
principal \emph{eigenspace}, such that the scale factors $\psi_{1i\ast}$ and
$\psi_{2i\ast}$ of the vector components of the novel principal eigenaxes
$\boldsymbol{\psi}=\boldsymbol{\psi}_{1}+\boldsymbol{\psi}_{2}$ and
$\boldsymbol{\rho}$ $=\boldsymbol{\rho}_{1}-\boldsymbol{\rho}_{2}$ are the
fundamental unknown parameters associated with the algorithm.

We have also determined that the algorithm finds the scale factors
$\psi_{1i\ast}$ and $\psi_{2i\ast}$ by solving the Wolfe dual problem%
\[
\max\Xi_{\boldsymbol{\psi}}\left(  \boldsymbol{\psi}\right)  =\mathbf{1}%
^{T}\boldsymbol{\psi}-\frac{\boldsymbol{\psi}^{T}\mathbf{Q}\boldsymbol{\psi}%
}{2}\text{,}%
\]
where $\mathbf{Q}$ is a joint covariance matrix of a collection of two
categories $\omega_{1}$ and $\omega_{2}$ of random vectors $\mathbf{x\sim}$
$p\left(  \mathbf{x};\omega_{1}\right)  $ and $\mathbf{x\sim}$ $p\left(
\mathbf{x};\omega_{2}\right)  $, such that the Wolfe dual novel principal
eigenaxis $\boldsymbol{\psi}$ is subject to the constraints $\boldsymbol{\psi
}^{T}\mathbf{y}=0$ and $\psi_{i}\geq0$, wherein the inequalities $\psi_{i}>0$
only hold for certain values of $\psi_{i}$, so that the Wolfe dual novel
principal eigenaxis $\boldsymbol{\psi}$ is symmetrically and equivalently
related to the principal eigenvector $\boldsymbol{\psi}_{\max}$ of the joint
covariance matrix $\mathbf{Q}$ and the inverted joint covariance matrix
$\mathbf{Q}^{-1}$ associated with the pair of random quadratic forms
$\boldsymbol{\psi}^{T}\mathbf{Q}\boldsymbol{\psi}$ and $\boldsymbol{\psi}%
^{T}\mathbf{Q}^{-1}\boldsymbol{\psi}$, \emph{and} is \emph{also} symmetrically
and equivalently related to the primal novel principal eigenaxis
$\boldsymbol{\rho}$.

In addition, we have determined that each scale factor $\psi_{1i\ast}$ and
$\psi_{2i\ast}$ determines a likelihood value and a likely location---both of
which are normalized relative to length---for a correlated extreme point
$\mathbf{x}_{1_{i\ast}}$ and $\mathbf{x}_{2_{i\ast}}$,\textbf{\ }so that each
scale factor $\psi_{1i\ast}$ and $\psi_{2i\ast}$ determines the magnitude
$\left\Vert \psi_{1_{i_{\ast}}}k_{\mathbf{x}_{1_{i\ast}}}\right\Vert $ and
$\left\Vert \psi_{2_{i_{\ast}}}k_{\mathbf{x}_{2_{i\ast}}}\right\Vert $ of a
correlated scaled extreme vector $\psi_{1_{i_{\ast}}}k_{\mathbf{x}_{1_{i\ast}%
}}$ and $\psi_{2_{i_{\ast}}}k_{\mathbf{x}_{2_{i\ast}}}$ that lies on the
geometric locus of the primal novel principal eigenaxis $\boldsymbol{\rho}$
$=\boldsymbol{\rho}_{1}-\boldsymbol{\rho}_{2}$, as well as a likelihood value
and a likely location for the correlated extreme vector $k_{\mathbf{x}%
_{1_{i\ast}}}$ and $k_{\mathbf{x}_{2_{i\ast}}}$.

Moreover, we have determined that the solution of the Wolfe dual problem is
\emph{implemented} by a Wolfe-dual \emph{eigenenergy functional} of a minimum
risk binary classification system%
\[
\max\Xi_{\boldsymbol{\psi}}\left(  \boldsymbol{\psi}\right)  =\mathbf{1}%
^{T}\boldsymbol{\psi}-\boldsymbol{\psi}^{T}\mathbf{Q}\boldsymbol{\psi
/}2\text{,}%
\]
wherein $\boldsymbol{\psi}^{T}\mathbf{y}=0$ and $\psi_{i}\geq0$, such that the
\emph{objective} of the Wolfe-dual eigenenergy functional is to \emph{find}
the principal eigenvector $\boldsymbol{\psi}_{\max}$ of the joint covariance
matrix $\mathbf{Q}$ and the inverted joint covariance matrix $\mathbf{Q}^{-1}$
associated with the pair of random quadratic forms $\boldsymbol{\psi}%
^{T}\mathbf{Q}\boldsymbol{\psi}$ and $\boldsymbol{\psi}^{T}\mathbf{Q}%
^{-1}\boldsymbol{\psi}$, so that the geometric locus of the Wolfe dual novel
principal eigenaxis $\boldsymbol{\psi}$ is symmetrically and equivalently
related to the principal eigenvector $\boldsymbol{\psi}_{\max}$ of the joint
covariance matrix $\mathbf{Q}$ associated with the random quadratic form
$\boldsymbol{\psi}^{T}\mathbf{Q}\boldsymbol{\psi}$ and the inverted joint
covariance matrix $\mathbf{Q}^{-1}$ associated with the random quadratic form
$\boldsymbol{\psi}^{T}\mathbf{Q}^{-1}\boldsymbol{\psi}$, and is also
symmetrically and equivalently related to the geometric locus of the primal
novel principal eigenaxis $\boldsymbol{\rho}$ in such a manner that the
geometric locus of the novel principal eigenaxis $\boldsymbol{\rho}$
$=\boldsymbol{\rho}_{1}-\boldsymbol{\rho}_{2}$ is the principal part of an
equivalent representation of the pair of random quadratic forms
$\boldsymbol{\psi}^{T}\mathbf{Q}\boldsymbol{\psi}$ and $\boldsymbol{\psi}%
^{T}\mathbf{Q}^{-1}\boldsymbol{\psi}$.

Thereby, the geometric locus of the novel principal eigenaxis
$\boldsymbol{\rho}$ $=\boldsymbol{\rho}_{1}-\boldsymbol{\rho}_{2}$ is the
principal eigenaxis of the geometric locus of the decision boundary of a
minimum risk binary classification system, so that the geometric locus of the
novel principal eigenaxis $\boldsymbol{\rho}$ $=\boldsymbol{\rho}%
_{1}-\boldsymbol{\rho}_{2}$ satisfies the geometric locus of the decision
boundary in terms of a critical minimum eigenenergy $\left\Vert
\boldsymbol{\rho}_{1}-\boldsymbol{\rho}_{2}\right\Vert _{\min_{c}}^{2}$ and a
minimum expected risk $\mathfrak{R}_{\mathfrak{\min}}\left(  \left\Vert
\boldsymbol{\rho}_{1}-\boldsymbol{\rho}_{2}\right\Vert _{\min_{c}}^{2}\right)
$, such that the total allowed eigenenergy $\left\Vert \boldsymbol{\rho}%
_{1}-\boldsymbol{\rho}_{2}\right\Vert _{\min_{c}}^{2}$ and the expected risk
$\mathfrak{R}_{\mathfrak{\min}}\left(  \left\Vert \boldsymbol{\rho}%
_{1}-\boldsymbol{\rho}_{2}\right\Vert _{\min_{c}}^{2}\right)  $ exhibited by
the geometric locus of the novel principal eigenaxis $\boldsymbol{\rho}$
$=\boldsymbol{\rho}_{1}-\boldsymbol{\rho}_{2}$ are regulated by the
eigenvalues of the symmetric matrices $\mathbf{Q}$ and $\mathbf{Q}^{-1}$
associated with the pair of random quadratic forms $\boldsymbol{\psi}%
^{T}\mathbf{Q}\boldsymbol{\psi}$ and $\boldsymbol{\psi}^{T}\mathbf{Q}%
^{-1}\boldsymbol{\psi}$, wherein the magnitude and the direction of the novel
principal eigenaxis $\boldsymbol{\rho}$ $=\boldsymbol{\rho}_{1}%
-\boldsymbol{\rho}_{2}$ are both functions of differences between joint
variabilities of extreme vectors $\mathbf{x}_{1_{i\ast}}\mathbf{\sim}p\left(
\mathbf{x};\omega_{1}\right)  $ and $\mathbf{x}_{2_{i\ast}}\mathbf{\sim
}p\left(  \mathbf{x};\omega_{1}\right)  $ that belong to two classes
$\omega_{1}$ and $\omega_{2}$ of random vectors $\mathbf{x\in}$ $%
\mathbb{R}
^{d}$.

\subsection{Minimization of a Vector-Valued Cost Function}

We have discovered that the Wolfe-dual eigenenergy functional of a minimum
risk binary classification system%
\[
\max\Xi_{\boldsymbol{\psi}}\left(  \boldsymbol{\psi}\right)  =\mathbf{1}%
^{T}\boldsymbol{\psi}-\boldsymbol{\psi}^{T}\mathbf{Q}\boldsymbol{\psi
/}2\text{,}%
\]
wherein $\boldsymbol{\psi}^{T}\mathbf{y}=0$ and $\psi_{i}\geq0$, uses a
\emph{vector-valued cost function} to find the principal eigenvector
$\boldsymbol{\psi}_{\max}$ of the joint covariance matrix $\mathbf{Q}$ and the
inverted joint covariance matrix $\mathbf{Q}^{-1}$ associated with the pair of
random quadratic forms $\boldsymbol{\psi}^{T}\mathbf{Q}\boldsymbol{\psi}$ and
$\boldsymbol{\psi}^{T}\mathbf{Q}^{-1}\boldsymbol{\psi}$, such that the
eigenenergy exhibited by both $\boldsymbol{\psi}_{\max}$ and $\boldsymbol{\rho
}$ is minimized in accordance with the eigenenergy condition%
\[
\lambda_{1}\left\Vert \boldsymbol{\psi}\right\Vert _{\min_{c}}^{2}%
=\boldsymbol{\psi}_{\max}^{T}\mathbf{Q}\boldsymbol{\psi}_{\max}\equiv
\left\Vert \boldsymbol{\rho}\right\Vert _{\min_{c}}^{2}\text{,}%
\]
at which point the random quadratic form $\boldsymbol{\psi}_{\max}%
^{T}\mathbf{Q}\boldsymbol{\psi}_{\max}$ is equivalently related to the
critical minimum eigenenergy $\left\Vert \boldsymbol{\rho}\right\Vert
_{\min_{c}}^{2}$ exhibited by the geometric locus of the primal novel
principal eigenaxis $\boldsymbol{\rho}$, so that the random quadratic form
$\boldsymbol{\psi}_{\max}^{T}\mathbf{Q}\boldsymbol{\psi}_{\max}$, plus the
total allowed eigenenergy $\left\Vert \boldsymbol{\rho}\right\Vert _{\min_{c}%
}^{2}$ and the expected risk $\mathfrak{R}_{\mathfrak{\min}}\left(  \left\Vert
\boldsymbol{\rho}\right\Vert _{\min_{c}}^{2}\right)  $ exhibited by the
geometric locus of the primal novel eigenaxis $\boldsymbol{\rho}$ jointly
reach their minimum values.

We have also discovered that the Wolfe-dual eigenenergy functional%
\[
\max\Xi_{\boldsymbol{\psi}}\left(  \boldsymbol{\psi}\right)  =\mathbf{1}%
^{T}\boldsymbol{\psi}-\boldsymbol{\psi}^{T}\mathbf{Q}\boldsymbol{\psi
/}2\text{,}%
\]
wherein $\boldsymbol{\psi}^{T}\mathbf{y}=0$ and $\psi_{i\ast}>0$, is
\emph{maximized} by the largest eigenvector $\boldsymbol{\psi}_{\max}$ of the
joint covariance matrix $\mathbf{Q}$%
\[
\mathbf{Q}\boldsymbol{\psi}_{\max}=\lambda_{1}\boldsymbol{\psi}_{\max}\text{,}%
\]
at which point the random quadratic form $\boldsymbol{\psi}_{\max}%
^{T}\mathbf{Q}\boldsymbol{\psi}_{\max}$ reaches its minimum value, so that the
total allowed eigenenergy $\left\Vert \boldsymbol{\rho}_{1}-\boldsymbol{\rho
}_{2}\right\Vert _{\min_{c}}^{2}$ and the expected risk $\mathfrak{R}%
_{\mathfrak{\min}}\left(  \left\Vert \boldsymbol{\rho}_{1}-\boldsymbol{\rho
}_{2}\right\Vert _{\min_{c}}^{2}\right)  $ exhibited by the geometric locus of
the primal novel principal eigenaxis $\boldsymbol{\rho}$ $=\boldsymbol{\rho
}_{1}-\boldsymbol{\rho}_{2}$ are \emph{jointly minimized}.

Thereby, we have discovered a novel principal eigen-coordinate transform
algorithm---that minimizes a vector-valued cost function---so that the total
allowed eigenenergy $\left\Vert \boldsymbol{\rho}_{1}-\boldsymbol{\rho}%
_{2}\right\Vert _{\min_{c}}^{2}$ and the correlated expected risk
$\mathfrak{R}_{\mathfrak{\min}}\left(  \left\Vert \boldsymbol{\rho}%
_{1}-\boldsymbol{\rho}_{2}\right\Vert _{\min_{c}}^{2}\right)  $ exhibited by a
minimum risk binary classification system are jointly minimized.

We have also discovered that the novel principal eigen-coordinate transform
algorithm finds scale factors $\psi_{1i\ast}$ and $\psi_{2i\ast}$ for the
components $\psi_{1i\ast}\frac{k_{\mathbf{x}_{1i\ast}}}{\left\Vert
k_{\mathbf{x}_{1i\ast}}\right\Vert }$ and $\psi_{2i\ast}\frac{k_{\mathbf{x}%
_{2i\ast}}}{\left\Vert k_{\mathbf{x}_{2i\ast}}\right\Vert }$ of the principal
eigenvector $\boldsymbol{\psi}_{\max}$ of a joint covariance matrix
$\mathbf{Q}$ and the inverted joint covariance matrix $\mathbf{Q}^{-1}%
$\textbf{, }so that the principal eigenvector $\boldsymbol{\psi}_{\max}$ is
symmetrically and equivalently related to the geometric locus of the Wolfe
dual novel principal eigenaxis $\boldsymbol{\psi}=\sum\nolimits_{i=1}^{l_{1}%
}\psi_{1i\ast}\frac{k_{\mathbf{x}_{1i\ast}}}{\left\Vert k_{\mathbf{x}_{1i\ast
}}\right\Vert }+\sum\nolimits_{i=1}^{l_{2}}\psi_{2i\ast}\frac{k_{\mathbf{x}%
_{2i\ast}}}{\left\Vert k_{\mathbf{x}_{2i\ast}}\right\Vert }$ of a minimum risk
binary classification system, \emph{and} is also symmetrically and
equivalently related to the geometric locus of the primal novel principal
eigenaxis $\boldsymbol{\rho}=\sum\nolimits_{i=1}^{l_{1}}\psi_{1_{i_{\ast}}%
}k_{\mathbf{x}_{1_{i\ast}}}-\sum\nolimits_{i=1}^{l_{2}}\psi_{2_{i_{\ast}}%
}k_{\mathbf{x}_{2_{i\ast}}}$ of the system, such that the scale factors
$\psi_{1i\ast}$ and $\psi_{2i\ast}$ determine critical minimum eigenenergies
$\left\Vert \psi_{1_{i\ast}}k_{\mathbf{x}_{1_{i\ast}}}\right\Vert _{\min_{c}%
}^{2}$ and $\left\Vert \psi_{2_{i\ast}}k_{\mathbf{x}_{2_{i\ast}}}\right\Vert
_{\min_{c}}^{2}$ exhibited by the principal eigenaxis components
$\psi_{1_{i_{\ast}}}k_{\mathbf{x}_{1_{i\ast}}}$ and $\psi_{2_{i\ast}%
}k_{\mathbf{x}_{2_{i\ast}}}$ that lie on the sides $\boldsymbol{\rho}_{1}$ and
$\boldsymbol{\rho}_{2}$ of the novel principal eigenaxis $\boldsymbol{\rho
}=\boldsymbol{\rho}_{1}-\boldsymbol{\rho}_{2}$ of the system%
\begin{align*}
\boldsymbol{\rho}  &  =\sum\nolimits_{i=1}^{l_{1}}\psi_{1_{i_{\ast}}%
}k_{\mathbf{x}_{1_{i\ast}}}-\sum\nolimits_{i=1}^{l_{2}}\psi_{2_{i_{\ast}}%
}k_{\mathbf{x}_{2_{i\ast}}}\\
&  =\boldsymbol{\rho}_{1}-\boldsymbol{\rho}_{2}\text{,}%
\end{align*}
so that eigenenergies $\left\Vert \psi_{1_{i\ast}}k_{\mathbf{x}_{1i\ast}%
}\right\Vert _{\min_{c}}^{2}$ and $\left\Vert \psi_{2_{i\ast}}k_{\mathbf{x}%
_{2i\ast}}\right\Vert _{\min_{c}}^{2}$ related to likely locations of
corresponding extreme points $\mathbf{x}_{1_{i\ast}}$ and $\mathbf{x}%
_{2_{i\ast}}$ determine \emph{costs} for expected counter risks of making
\emph{right decisions} or \emph{costs} for expected risks of making
\emph{wrong decisions}.

Accordingly, we have determined that the novel principal eigen-coordinate
transform algorithm \emph{finds} the extreme vectors $k_{\mathbf{x}_{1_{i\ast
}}}$ and $k_{\mathbf{x}_{2_{i\ast}}}$ and the scale factors $\psi_{1i\ast}$
and $\psi_{2i\ast}$ that determine the geometrical and statistical structure
of the geometric locus of a novel principal eigenaxis $\boldsymbol{\rho}%
=\sum\nolimits_{i=1}^{l_{1}}\psi_{1_{i_{\ast}}}k_{\mathbf{x}_{1_{i\ast}}}%
-\sum\nolimits_{i=1}^{l_{2}}\psi_{2_{i_{\ast}}}k_{\mathbf{x}_{2_{i\ast}}}$, so
that the geometric locus of the novel principal eigenaxis $\boldsymbol{\rho
}=\boldsymbol{\rho}_{1}-\boldsymbol{\rho}_{2}$ is the principal part of an
equivalent representation of a pair of random quadratic forms
$\boldsymbol{\psi}^{T}\mathbf{Q}\boldsymbol{\psi}$ and $\boldsymbol{\psi}%
^{T}\mathbf{Q}^{-1}\boldsymbol{\psi}$, at which point the random quadratic
forms $\boldsymbol{\psi}^{T}\mathbf{Q}\boldsymbol{\psi}$ and $\boldsymbol{\psi
}^{T}\mathbf{Q}^{-1}\boldsymbol{\psi}$ and the total allowed eigenenergy
$\left\Vert \boldsymbol{\rho}_{1}-\boldsymbol{\rho}_{2}\right\Vert _{\min_{c}%
}^{2}$ exhibited by the novel principal eigenaxis $\boldsymbol{\rho
}=\boldsymbol{\rho}_{1}-\boldsymbol{\rho}_{2}$ each reach their minimum value.

Correspondingly, we have determined that any given geometric locus of a novel
principal eigenaxis $\boldsymbol{\rho}=\boldsymbol{\rho}_{1}-\boldsymbol{\rho
}_{2}$ provides dual representation of the discriminant function, an exclusive
principal eigen-coordinate system of the geometric locus of the decision
boundary, and an eigenaxis of symmetry that spans the decision space---of a
certain minimum risk binary classification system---such that the geometric
locus of the novel principal eigenaxis $\boldsymbol{\rho}=\boldsymbol{\rho
}_{1}-\boldsymbol{\rho}_{2}$ is the principal eigenaxis of the geometric loci
of the decision boundary and a pair of symmetrically positioned decision
borders, so that the geometric locus of the novel principal eigenaxis
$\boldsymbol{\rho}=\boldsymbol{\rho}_{1}-\boldsymbol{\rho}_{2}$ satisfies the
geometric locus of the decision boundary in terms of a critical minimum
eigenenergy $\left\Vert \boldsymbol{\rho}_{1}-\boldsymbol{\rho}_{2}\right\Vert
_{\min_{c}}^{2}$ and a minimum expected risk $\mathfrak{R}_{\mathfrak{\min}%
}\left(  \left\Vert \boldsymbol{\rho}_{1}-\boldsymbol{\rho}_{2}\right\Vert
_{\min_{c}}^{2}\right)  $, such that the total allowed eigenenergy $\left\Vert
\boldsymbol{\rho}_{1}-\boldsymbol{\rho}_{2}\right\Vert _{\min_{c}}^{2}$ and
the expected risk $\mathfrak{R}_{\mathfrak{\min}}\left(  \left\Vert
\boldsymbol{\rho}_{1}-\boldsymbol{\rho}_{2}\right\Vert _{\min_{c}}^{2}\right)
$ exhibited by the geometric locus of the novel principal eigenaxis
$\boldsymbol{\rho}=\boldsymbol{\rho}_{1}-\boldsymbol{\rho}_{2}$ are regulated
by the eigenvalues of the symmetric matrices $\mathbf{Q}$ and $\mathbf{Q}%
^{-1}$ of a pair of random quadratic forms $\boldsymbol{\psi}^{T}%
\mathbf{Q}\boldsymbol{\psi}$ and $\boldsymbol{\psi}^{T}\mathbf{Q}%
^{-1}\boldsymbol{\psi}$, wherein the magnitude and the direction of the novel
principal eigenaxis $\boldsymbol{\rho}$ $=\boldsymbol{\rho}_{1}%
-\boldsymbol{\rho}_{2}$ are both functions of differences between joint
variabilities of extreme vectors $\mathbf{x}_{1_{i\ast}}\mathbf{\sim}p\left(
\mathbf{x};\omega_{1}\right)  $ and $\mathbf{x}_{2_{i\ast}}\mathbf{\sim
}p\left(  \mathbf{x};\omega_{1}\right)  $ that belong to two classes
$\omega_{1}$ and $\omega_{2}$ of random vectors $\mathbf{x\in}$ $%
\mathbb{R}
^{d}$.

We are now in a position to define fundamental locus equations of binary classification.

\section{\label{Section 10}Locus Equations of Binary Classification}

In this section, we develop vector algebra locus equations of binary
classification that express fundamental laws of binary classification---that
discriminant functions of minimum risk binary classification systems are
subject to. We begin by defining the remaining terms in the equivalent form of
the vector algebra locus equation of (\ref{Norm_Dec_Bound}).

\subsection{Locus of Average Risk}

Let any given random vector that has the form%
\begin{align*}
&  \frac{1}{l}\left(  \sum\nolimits_{i=1}^{l_{1}}k_{\mathbf{x}_{1_{i\ast}}%
}+\sum\nolimits_{i=1}^{l_{2}}k_{\mathbf{x}_{2_{i\ast}}}\right) \\
&  =\frac{1}{l}\sum\nolimits_{i=1}^{l}k_{\mathbf{x}_{i\ast}}\text{,}%
\end{align*}
wherein $l$ reproducing kernels $k_{\mathbf{x}_{i\ast}}$ of extreme points
$\mathbf{x}_{i\ast}\mathbf{\ }$belong to class $\omega_{1}$ and class
$\omega_{2}$, provide an equivalent representation of the random vector%
\[
2\left(  \mathbf{\Sigma}_{1}^{-1}\boldsymbol{\mu}_{1}-\mathbf{\Sigma}_{2}%
^{-1}\boldsymbol{\mu}_{2}\right)  \text{,}%
\]
so that the locus of each random vector $\frac{1}{l}\sum\nolimits_{i=1}%
^{l}k_{\mathbf{x}_{i\ast}}$ and $2\left(  \mathbf{\Sigma}_{1}^{-1}%
\boldsymbol{\mu}_{1}-\mathbf{\Sigma}_{2}^{-1}\boldsymbol{\mu}_{2}\right)  $
represents a \emph{locus of average risk} within the decision space
$Z=Z_{1}\cup Z_{2}$ of a minimum risk binary classification system, such that
the locus of average risk is located near the locus of the decision boundary
of the system.

Thereby, we realize that an equivalent representation of the vector%
\[
2\mathbf{x}^{T}\left(  \mathbf{\Sigma}_{1}^{-1}\boldsymbol{\mu}_{1}%
-\mathbf{\Sigma}_{2}^{-1}\boldsymbol{\mu}_{2}\right)
\]
is represented by a vector that determines the position of a random vector
$k_{\mathbf{s}}$ relative to a locus of average risk $\frac{1}{l}%
\sum\nolimits_{i=1}^{l}k_{\mathbf{x}_{i\ast}}$%
\begin{equation}
k_{\mathbf{s}}-\frac{1}{l}\sum\nolimits_{i=1}^{l}k_{\mathbf{x}_{i\ast}%
}\text{,} \tag{10.1}\label{Locus of Ave Risk}%
\end{equation}
such that the locus of average risk $\frac{1}{l}\sum\nolimits_{i=1}%
^{l}k_{\mathbf{x}_{i\ast}}$ is located on or near the locus of a linear
decision boundary or is centrally located and bounded by quadratic loci of a
quadratic decision boundary.

\subsection{Expected Likelihood of Observing Extreme Vectors}

An equivalent form of the vector algebra locus equation of
(\ref{Norm_Dec_Bound}) contains an equivalent representation of the
statistical expression%
\[
\left(  \boldsymbol{\mu}_{1}^{T}\mathbf{\Sigma}_{1}^{-1}\boldsymbol{\mu}%
_{1}-\boldsymbol{\mu}_{2}^{T}\mathbf{\Sigma}_{2}^{-1}\boldsymbol{\mu}%
_{2}\right)  +\left(  \ln\left(  \left\vert \mathbf{\Sigma}_{2}\right\vert
\right)  -\ln\left(  \left\vert \mathbf{\Sigma}_{1}\right\vert \right)
\right)  \text{,}%
\]
where the expression $\left(  \boldsymbol{\mu}_{1}^{T}\mathbf{\Sigma}_{1}%
^{-1}\boldsymbol{\mu}_{1}-\boldsymbol{\mu}_{2}^{T}\mathbf{\Sigma}_{2}%
^{-1}\boldsymbol{\mu}_{2}\right)  $ represents the difference between expected
likelihoods of observing normal random vectors that belong to two classes, and
the expression $\ln\left(  \left\vert \mathbf{\Sigma}_{2}\right\vert \right)
-\ln\left(  \left\vert \mathbf{\Sigma}_{1}\right\vert \right)  $ represents
the difference between expected distributions of normal random vectors that
belong to the two classes---which are related to expected likelihoods of
observing the normal random vectors.

Since we are going to use a constrained optimization algorithm to find
discriminant functions of minimum risk binary classification systems, we
realize that the statistic $\frac{1}{l}\sum\nolimits_{i=1}^{l}y_{i}$ is a
simple and effective way to represent the \emph{difference} between expected
likelihoods of observing $l$ extreme vectors $\left\{  k_{\mathbf{x}_{i\ast}%
}\right\}  _{i=1}^{l}$ that belong to two classes $\omega_{1}$ and $\omega
_{2}$ of random vectors $\mathbf{x\in}$ $%
\mathbb{R}
^{d}$ such that $\mathbf{x\sim}$ $p\left(  \mathbf{x};\omega_{1}\right)  $ and
$\mathbf{x\sim}$ $p\left(  \mathbf{x};\omega_{2}\right)  $%
\begin{equation}
\frac{1}{l}\sum\nolimits_{i=1}^{l}y_{i}\text{,} \tag{10.2}%
\label{Expected Likelihood}%
\end{equation}
where $l$ is the number of extreme vectors $k_{\mathbf{x}_{i\ast}}$ that
belong to the two classes $\omega_{1}$ and $\omega_{2}$, wherein $y_{i}=+1$ if
an extreme vector $k_{\mathbf{x}_{i\ast}}$ belongs to class $\omega_{1}$ and
$y_{i}=-1$ if an extreme vector $k_{\mathbf{x}_{i\ast}}$ belongs to class
$\omega_{2}$.

\subsection{Locus Equation of a Decision Boundary}

We are now in a position to define a vector algebra locus equation that
represents the geometric locus of the decision boundary of any given minimum
risk binary classification system that is subject to two categories of random
vectors $\mathbf{x\in}$ $%
\mathbb{R}
^{d}$. Corollary \ref{Locus Equation of Decision Boundary Corollary} expresses
an equivalent form of the vector algebra locus equation in
(\ref{Norm_Dec_Bound}) that represents the geometric locus of the decision
boundary of a minimum risk binary classification system, so that the
discriminant function of the system and the exclusive intrinsic
eigen-coordinate system of the geometric locus of the decision boundary of the
system are both represented by a geometric locus of a novel principal eigenaxis.

\begin{corollary}
\label{Locus Equation of Decision Boundary Corollary}Let%
\[
\left(  k_{\mathbf{s}}\boldsymbol{-}\frac{1}{l}\sum\nolimits_{i=1}%
^{l}k_{\mathbf{x}_{i\ast}}\right)  \left(  \boldsymbol{\rho}_{1}%
-\boldsymbol{\rho}_{2}\right)  +\frac{1}{l}\sum\nolimits_{i=1}^{l}%
y_{i}\overset{\omega_{1}}{\underset{\omega_{2}}{\gtrless}}0
\]
be any given minimum risk binary classification system that is subject to
random inputs $\mathbf{x\in}$ $%
\mathbb{R}
^{d}$ such that $\mathbf{x\sim}$ $p\left(  \mathbf{x};\omega_{1}\right)  $ and
$\mathbf{x\sim}$ $p\left(  \mathbf{x};\omega_{2}\right)  $, where $p\left(
\mathbf{x};\omega_{1}\right)  $ and $p\left(  \mathbf{x};\omega_{2}\right)  $
are certain probability density functions for two classes $\omega_{1}$ and
$\omega_{2}$ of random vectors $\mathbf{x\in}$ $%
\mathbb{R}
^{d}$, at which point the discriminant function of the system is represented
by a geometric locus of a novel principal eigenaxis $\boldsymbol{\rho
}=\boldsymbol{\rho}_{1}-\boldsymbol{\rho}_{2}$.

The geometric locus of the decision boundary of the system is represented by
the graph of a vector algebra locus equation that has the form%
\[
\left(  k_{\mathbf{s}}\boldsymbol{-}\frac{1}{l}\sum\nolimits_{i=1}%
^{l}k_{\mathbf{x}_{i\ast}}\right)  \left(  \boldsymbol{\rho}_{1}%
-\boldsymbol{\rho}_{2}\right)  +\frac{1}{l}\sum\nolimits_{i=1}^{l}%
y_{i}=0\text{,}%
\]
so that the geometric locus of the novel principal eigenaxis $\boldsymbol{\rho
}=\boldsymbol{\rho}_{1}-\boldsymbol{\rho}_{2}$ of the system is the solution
of the locus equation, at which point the geometric locus of the novel
principal eigenaxis $\boldsymbol{\rho}=\boldsymbol{\rho}_{1}-\boldsymbol{\rho
}_{2}$ provides dual representation of the discriminant function of the system
and an exclusive principal eigen-coordinate system of the geometric locus of
the decision boundary of the system, such that all of the points $\mathbf{s}$
that lie on the geometric locus of the decision boundary exclusively reference
the principal eigen-coordinate system $\boldsymbol{\rho}=\boldsymbol{\rho}%
_{1}-\boldsymbol{\rho}_{2}$, where the statistic $\frac{1}{l}\sum
\nolimits_{i=1}^{l}y_{i}:y_{i}=\pm1$ is an expected likelihood of observing
$l=l_{1}+l_{2}$ extreme vectors $k_{\mathbf{x}_{i\ast}}$, and the vector
difference of $k_{\mathbf{s}}\boldsymbol{-}\frac{1}{l}\sum\nolimits_{i=1}%
^{l}k_{\mathbf{x}_{i\ast}}$ determines the distance between the locus of a
random vector $\mathbf{s}$ and a locus of average risk $\frac{1}{l}%
\sum\nolimits_{i=1}^{l}k_{\mathbf{x}_{i\ast}}$ within the decision space
$Z=Z_{1}\cup Z_{2}$ of the system.
\end{corollary}

\begin{proof}
We prove Corollary \ref{Locus Equation of Decision Boundary Corollary} by a
constructive proof that demonstrates how a well-posed constrained optimization
algorithm resolves the inverse problem of the binary classification of random vectors.
\end{proof}

Returning now to the conditions expressed by Corollary
\ref{Partitioning of Decision Space Corollary} and Axioms
\ref{Overlapping Extreme Points Axiom} -
\ref{Non-overlapping Extreme Points Axiom}, along with the conditions
expressed by Corollaries \ref{Primary Integral Equation Corollary} -
\ref{Secondary Integral Equation Corollary}, recall that likely locations of
extreme points $\mathbf{x}_{1_{i\ast}}$ and $\mathbf{x}_{2_{i\ast}}$ are
spread throughout the decision space $Z=Z_{1}\cup Z_{2}$ of any given minimum
risk binary classification system in such a manner that likely locations of
the extreme points $\mathbf{x}_{1_{i\ast}}$ and $\mathbf{x}_{2_{i\ast}}$
determine right and wrong decisions made by the system. Properties of a
minimum risk binary classification system to make right and wrong decisions
are defined by the notions of counter and risk---which we originally defined
in our working paper
\citep{Reeves2018design}%
.

\subsection{Counter Risks of Making Right Decisions}

\begin{definition}
\label{Counter Risk Definition}The probability of finding an extreme point
$\mathbf{x}_{1_{i\ast}}$ inside the decision space $Z=Z_{1}\cup Z_{2}$ of a
minimum risk binary classification system is said to determine a region of
counter risk---where counter risk is the property of the system to make right
decisions---if and only if likely locations of the extreme point
$\mathbf{x}_{1_{i\ast}}$ are inside the decision region $Z_{1}$ of the system,
at which point the critical minimum eigenenergy $\left\Vert \psi_{1_{i\ast}%
}k_{\mathbf{x}_{1_{i\ast}}}\right\Vert _{\min_{c}}^{2}$ exhibited by the
principal eigenaxis component $\psi_{1_{i_{\ast}}}k_{\mathbf{x}_{1_{i\ast}}}$
on the side $\boldsymbol{\rho}_{1}$ of the geometric locus of the novel
principal eigenaxis $\boldsymbol{\rho}=\boldsymbol{\rho}_{1}-\boldsymbol{\rho
}_{2}$ of the system contributes to both the counter risk $\overline
{\mathfrak{R}}_{\mathfrak{\min}}\left(  Z_{1}|\boldsymbol{\rho}_{1}\right)  $
and the total allowed eigenenergy $\left\Vert Z_{1}|\boldsymbol{\rho}%
_{1}\right\Vert _{\min_{c}}^{2}$ that is given by the integral%
\begin{equation}
\overline{\mathfrak{R}}_{\mathfrak{\min}}\left(  Z_{1}|\boldsymbol{\rho}%
_{1}\right)  =\int_{Z_{1}}\boldsymbol{\rho}_{1}d\boldsymbol{\rho}_{1}%
\equiv\left\Vert Z_{1}|\boldsymbol{\rho}_{1}\right\Vert _{\min_{c}}^{2}\equiv
P\left(  Z_{1}|\boldsymbol{\rho}_{1}\right)  \text{,} \tag{10.3}%
\label{Counter Risk Integral Class 1}%
\end{equation}
over the decision region $Z_{1}$ of the system, where $P\left(  Z_{1}%
|\boldsymbol{\rho}_{1}\right)  $ is the conditional probability that extreme
points $\mathbf{x}_{1_{i\ast}}$ are located inside the decision region $Z_{1}%
$, and $\left\Vert Z_{1}|\boldsymbol{\rho}_{1}\right\Vert _{\min_{c}}^{2}$ is
the total allowed eigenenergy exhibited by all of the principal eigenaxis
components $\psi_{1_{i_{\ast}}}k_{\mathbf{x}_{1_{i\ast}}}$ on the side
$\boldsymbol{\rho}_{1}$ of the novel principal eigenaxis $\boldsymbol{\rho
}=\boldsymbol{\rho}_{1}-\boldsymbol{\rho}_{2}$.

Correspondingly, the probability of finding an extreme point $\mathbf{x}%
_{2_{i\ast}}$ inside the decision space $Z=Z_{1}\cup Z_{2}$ of a minimum risk
binary classification system is said to determine a region of counter risk if
and only if likely locations of the extreme point $\mathbf{x}_{2_{i\ast}}$ are
inside the decision region $Z_{2}$ of the system, at which point the critical
minimum eigenenergy $\left\Vert \psi_{2_{i\ast}}k_{\mathbf{x}_{2_{i\ast}}%
}\right\Vert _{\min_{c}}^{2}$ exhibited by the principal eigenaxis component
$\psi_{2_{i\ast}}k_{\mathbf{x}_{2_{i\ast}}}$ on the side $\boldsymbol{\rho
}_{2}$ of the geometric locus of the novel principal eigenaxis
$\boldsymbol{\rho}=\boldsymbol{\rho}_{1}-\boldsymbol{\rho}_{2}$ of the system
contributes to both the counter risk $\overline{\mathfrak{R}}_{\mathfrak{\min
}}\left(  Z_{2}|\boldsymbol{\rho}_{2}\right)  $ and the total allowed
eigenenergy $\left\Vert Z_{2}|\boldsymbol{\rho}_{2}\right\Vert _{\min_{c}}%
^{2}$ that is given by the integral%
\begin{equation}
\overline{\mathfrak{R}}_{\mathfrak{\min}}\left(  Z_{2}|\boldsymbol{\rho}%
_{2}\right)  =\int_{Z_{2}}\boldsymbol{\rho}_{2}d\boldsymbol{\rho}_{2}%
\equiv\left\Vert Z_{2}|\boldsymbol{\rho}_{2}\right\Vert _{\min_{c}}^{2}\equiv
P\left(  Z_{2}|\boldsymbol{\rho}_{2}\right)  \text{,} \tag{10.4}%
\label{Counter Risk Integral Class 2}%
\end{equation}
over the decision region $Z_{2}$ of the system, where $P\left(  Z_{2}%
|\boldsymbol{\rho}_{2}\right)  $ is the conditional probability that extreme
points $\mathbf{x}_{2_{i\ast}}$ are located inside the decision region $Z_{2}%
$, and $\left\Vert Z_{2}|\boldsymbol{\rho}_{2}\right\Vert _{\min_{c}}^{2}$ is
the total allowed eigenenergy exhibited by all of the principal eigenaxis
components $\psi_{2_{i\ast}}k_{\mathbf{x}_{2_{i\ast}}}$ on the side
$\boldsymbol{\rho}_{2}$ of the novel principal eigenaxis $\boldsymbol{\rho
}=\boldsymbol{\rho}_{1}-\boldsymbol{\rho}_{2}$.
\end{definition}

\subsection{Risks of Making Wrong Decisions}

\begin{definition}
\label{Risk Definition}The probability of finding an extreme point
$\mathbf{x}_{1_{i\ast}}$ inside the decision space $Z=Z_{1}\cup Z_{2}$ of a
minimum risk binary classification system is said to determine a region of
risk---where risk is the property of the system to make wrong decisions---if
and only if likely locations of the extreme point $\mathbf{x}_{1_{i\ast}}$ are
inside the decision region $Z_{2}$ of the system, at which point the critical
minimum eigenenergy $\left\Vert \psi_{1_{i\ast}}k_{\mathbf{x}_{1_{i\ast}}%
}\right\Vert _{\min_{c}}^{2}$ exhibited by the principal eigenaxis component
$\psi_{1_{i_{\ast}}}k_{\mathbf{x}_{1_{i\ast}}}$ on the side $\boldsymbol{\rho
}_{1}$ of the geometric locus of the novel principal eigenaxis
$\boldsymbol{\rho}=\boldsymbol{\rho}_{1}-\boldsymbol{\rho}_{2}$ of the system
contributes to both the risk $\mathfrak{R}_{\mathfrak{\min}}\left(
Z_{2}|\boldsymbol{\rho}_{1}\right)  $ and the total allowed eigenenergy
$\left\Vert Z_{2}|\boldsymbol{\rho}_{1}\right\Vert _{\min_{c}}^{2}$ that is
given by the integral%
\begin{equation}
\mathfrak{R}_{\mathfrak{\min}}\left(  Z_{2}|\boldsymbol{\rho}_{1}\right)
=\int_{Z_{2}}\boldsymbol{\rho}_{1}d\boldsymbol{\rho}_{1}\equiv\left\Vert
Z_{2}|\boldsymbol{\rho}_{1}\right\Vert _{\min_{c}}^{2}\equiv P\left(
Z_{2}|\boldsymbol{\rho}_{1}\right)  \text{,} \tag{10.5}%
\label{Risk Integral Class 2}%
\end{equation}
over the decision region $Z_{2}$ of the system, where $P\left(  Z_{2}%
|\boldsymbol{\rho}_{1}\right)  $ is the conditional probability that extreme
points $\mathbf{x}_{1_{i\ast}}$ are located inside the decision region $Z_{2}%
$, and $\left\Vert Z_{2}|\boldsymbol{\rho}_{1}\right\Vert _{\min_{c}}^{2}$ is
the total allowed eigenenergy exhibited by all of the principal eigenaxis
components $\psi_{1_{i_{\ast}}}k_{\mathbf{x}_{1_{i\ast}}}$ on the side
$\boldsymbol{\rho}_{1}$ of the novel principal eigenaxis $\boldsymbol{\rho
}=\boldsymbol{\rho}_{1}-\boldsymbol{\rho}_{2}$.

Correspondingly, the probability of finding an extreme point $\mathbf{x}%
_{2_{i\ast}}$ inside the decision space $Z=Z_{1}\cup Z_{2}$ of a minimum risk
binary classification system is said to determine a region of risk if and only
if likely locations of the extreme point $\mathbf{x}_{2_{i\ast}}$ are inside
the decision region $Z_{1}$ of the system, at which point the critical minimum
eigenenergy $\left\Vert \psi_{2_{i\ast}}k_{\mathbf{x}_{2_{i\ast}}}\right\Vert
_{\min_{c}}^{2}$ exhibited by the principal eigenaxis component $\psi
_{2_{i\ast}}k_{\mathbf{x}_{2_{i\ast}}}$ on the side $\boldsymbol{\rho}_{2}$ of
the geometric locus of the novel principal eigenaxis $\boldsymbol{\rho
}=\boldsymbol{\rho}_{1}-\boldsymbol{\rho}_{2}$ of the system contributes to
both the risk $\mathfrak{R}_{\mathfrak{\min}}\left(  Z_{1}|\boldsymbol{\rho
}_{2}\right)  $ and the total allowed eigenenergy $\left\Vert Z_{1}%
|\boldsymbol{\rho}_{2}\right\Vert _{\min_{c}}^{2}$ that is given by the
integral%
\begin{equation}
\mathfrak{R}_{\mathfrak{\min}}\left(  Z_{1}|\boldsymbol{\rho}_{2}\right)
=\int_{Z_{1}}\boldsymbol{\rho}_{2}d\boldsymbol{\rho}_{2}\equiv\left\Vert
Z_{1}|\boldsymbol{\rho}_{2}\right\Vert _{\min_{c}}^{2}\equiv P\left(
Z_{1}|\boldsymbol{\rho}_{2}\right)  \text{,} \tag{10.6}%
\label{Risk Integral Class 1}%
\end{equation}
over the decision region $Z_{1}$ of the system, where $P\left(  Z_{1}%
|\boldsymbol{\rho}_{2}\right)  $ is the conditional probability that extreme
points $\mathbf{x}_{2_{i\ast}}$ are located inside the decision region $Z_{1}%
$, and $\left\Vert Z_{1}|\boldsymbol{\rho}_{2}\right\Vert _{\min_{c}}^{2}$ is
the total allowed eigenenergy exhibited by all of the principal eigenaxis
components $\psi_{2_{i\ast}}k_{\mathbf{x}_{2_{i\ast}}}$ on the side
$\boldsymbol{\rho}_{2}$ of the novel principal eigenaxis $\boldsymbol{\rho
}=\boldsymbol{\rho}_{1}-\boldsymbol{\rho}_{2}$.
\end{definition}

\subsection{Vector Algebra Locus Equations of Decision Spaces}

Given the conditions expressed by Theorem
\ref{Principal Eigen-coordinate System Theorem}, Corollary
\ref{Principal Eigen-coordinate System Corollary} and Theorem
\ref{Geometric Locus of a Novel Principal Eigenaxis Theorem}, we realize that
the geometric locus of the novel principal eigenaxis $\boldsymbol{\rho
}=\boldsymbol{\rho}_{1}-\boldsymbol{\rho}_{2}$ of any given minimum risk
binary classification system $\left(  k_{\mathbf{s}}\boldsymbol{-}\frac{1}%
{l}\sum\nolimits_{i=1}^{l}k_{\mathbf{x}_{i\ast}}\right)  \left(
\boldsymbol{\rho}_{1}-\boldsymbol{\rho}_{2}\right)  +\frac{1}{l}%
\sum\nolimits_{i=1}^{l}y_{i}\overset{\omega_{1}}{\underset{\omega
_{2}}{\gtrless}}0$ completely determines the shape of the decision space
$Z=Z_{1}\cup Z_{2}$ of the system.

Thereby, given the properties of any given minimum risk binary classification
system to make right and wrong decisions---expressed by the notions of counter
and risk in Definitions \ref{Counter Risk Definition} and
\ref{Risk Definition}, along with Theorem
\ref{Principal Eigen-coordinate System Theorem}, Corollary
\ref{Principal Eigen-coordinate System Corollary} and Theorem
\ref{Geometric Locus of a Novel Principal Eigenaxis Theorem}, we realize that
any given geometric locus of a novel principal eigenaxis $\boldsymbol{\rho
}=\boldsymbol{\rho}_{1}-\boldsymbol{\rho}_{2}$ is the solution of each and
every one of the vector algebra locus equations expressed by Corollary
\ref{Locus Equations of Decision Space Corollary}.

\begin{corollary}
\label{Locus Equations of Decision Space Corollary}Let%
\[
\left(  k_{\mathbf{s}}\boldsymbol{-}\frac{1}{l}\sum\nolimits_{i=1}%
^{l}k_{\mathbf{x}_{i\ast}}\right)  \left(  \boldsymbol{\rho}_{1}%
-\boldsymbol{\rho}_{2}\right)  +\frac{1}{l}\sum\nolimits_{i=1}^{l}%
y_{i}\overset{\omega_{1}}{\underset{\omega_{2}}{\gtrless}}0
\]
be any given minimum risk binary classification system that is subject to
random inputs $\mathbf{x\in}$ $%
\mathbb{R}
^{d}$ such that $\mathbf{x\sim}$ $p\left(  \mathbf{x};\omega_{1}\right)  $ and
$\mathbf{x\sim}$ $p\left(  \mathbf{x};\omega_{2}\right)  $, where $p\left(
\mathbf{x};\omega_{1}\right)  $ and $p\left(  \mathbf{x};\omega_{2}\right)  $
are certain probability density functions for two classes $\omega_{1}$ and
$\omega_{2}$ of random vectors $\mathbf{x\in}$ $%
\mathbb{R}
^{d}$, at which point the discriminant function of the system is represented
by a geometric locus of a novel principal eigenaxis $\boldsymbol{\rho
}=\boldsymbol{\rho}_{1}-\boldsymbol{\rho}_{2}$.

Also, let $d\left(  \mathbf{s}\right)  =0$ denote the geometric locus of the
decision boundary of the system, let $d\left(  \mathbf{s}\right)  =+1$ denote
the geometric locus of the decision border of the decision region $Z_{1}$ of
the system, and let $d\left(  \mathbf{s}\right)  =-1$ denote the geometric
locus of decision border of the decision region $Z_{2}$ of the system, so that
the geometric loci of the decision boundary $d\left(  \mathbf{s}\right)  =0$
and the symmetrically positioned decision borders $d\left(  \mathbf{s}\right)
=+1$ and $d\left(  \mathbf{s}\right)  =-1$ partition the decision space
$Z=Z_{1}\cup Z_{2}$ of the minimum risk binary classification system in a
symmetrically balanced manner.

The geometric locus of the decision boundary of the minimum risk binary
classification system is represented by the graph of a vector algebra locus
equation that has the form%
\[
\left(  k_{\mathbf{s}}\boldsymbol{-}\frac{1}{l}\sum\nolimits_{i=1}%
^{l}k_{\mathbf{x}_{i\ast}}\right)  \left(  \boldsymbol{\rho}_{1}%
-\boldsymbol{\rho}_{2}\right)  +\frac{1}{l}\sum\nolimits_{i=1}^{l}%
y_{i}=0\text{,}%
\]
wherein the geometric locus of the novel principal eigenaxis $\boldsymbol{\rho
}=\boldsymbol{\rho}_{1}-\boldsymbol{\rho}_{2}$ of the system is the solution
of the locus equation, so that the geometric locus of the novel principal
eigenaxis $\boldsymbol{\rho}=\boldsymbol{\rho}_{1}-\boldsymbol{\rho}_{2}$
provides dual representation of the discriminant function of the system and an
exclusive principal eigen-coordinate system of the geometric locus of the
decision boundary of the system, at which point all of the points $\mathbf{s}$
that lie on the geometric locus of the decision boundary exclusively reference
the novel principal eigenaxis $\boldsymbol{\rho}=\boldsymbol{\rho}%
_{1}-\boldsymbol{\rho}_{2}$, and the geometric locus of the novel principal
eigenaxis $\boldsymbol{\rho}=\boldsymbol{\rho}_{1}-\boldsymbol{\rho}_{2}$
represents an eigenaxis of symmetry that spans the decision space $Z=Z_{1}\cup
Z_{2}$ of the minimum risk binary classification system $\left(
k_{\mathbf{s}}\boldsymbol{-}\frac{1}{l}\sum\nolimits_{i=1}^{l}k_{\mathbf{x}%
_{i\ast}}\right)  \left(  \boldsymbol{\rho}_{1}-\boldsymbol{\rho}_{2}\right)
+\frac{1}{l}\sum\nolimits_{i=1}^{l}y_{i}\overset{\omega_{1}}{\underset{\omega
_{2}}{\gtrless}}0$, so that the shape of the decision space $Z=Z_{1}\cup
Z_{2}$ is completely determined by the exclusive intrinsic eigen-coordinate
system $\boldsymbol{\rho}=\boldsymbol{\rho}_{1}-\boldsymbol{\rho}_{2}$ in the
following manner.

The geometric locus of the decision border of the decision region $Z_{1}$ of
the minimum risk binary classification system is represented by the graph of a
vector algebra locus equation that has the form%
\[
\left(  k_{\mathbf{s}}\boldsymbol{-}\frac{1}{l}\sum\nolimits_{i=1}%
^{l}k_{\mathbf{x}_{i\ast}}\right)  \left(  \boldsymbol{\rho}_{1}%
-\boldsymbol{\rho}_{2}\right)  +\frac{1}{l}\sum\nolimits_{i=1}^{l}%
y_{i}=+1\text{,}%
\]
wherein the geometric locus of the novel principal eigenaxis $\boldsymbol{\rho
}=\boldsymbol{\rho}_{1}-\boldsymbol{\rho}_{2}$ of the system is the solution
of the locus equation, so that all of the points $\mathbf{s}$ that lie on the
geometric locus of the decision border $d\left(  \mathbf{s}\right)  =+1$
exclusively reference the novel principal eigenaxis $\boldsymbol{\rho
}=\boldsymbol{\rho}_{1}-\boldsymbol{\rho}_{2}$, at which point the shape of
the decision region $Z_{1}$ of the minimum risk binary classification system
is completely determined by the exclusive intrinsic eigen-coordinate system
$\boldsymbol{\rho}=\boldsymbol{\rho}_{1}-\boldsymbol{\rho}_{2}$, such that the
shape of the decision region $Z_{1}$ is determined by the shapes of the
geometric loci of the decision border $d\left(  \mathbf{s}\right)  =+1$ and
the decision boundary $d\left(  \mathbf{s}\right)  =0$ of the system.

The geometric locus of the decision border of the decision region $Z_{2}$ of
the minimum risk binary classification system is represented by the graph of a
vector algebra locus equation that has the form%
\[
\left(  k_{\mathbf{s}}\boldsymbol{-}\frac{1}{l}\sum\nolimits_{i=1}%
^{l}k_{\mathbf{x}_{i\ast}}\right)  \left(  \boldsymbol{\rho}_{1}%
-\boldsymbol{\rho}_{2}\right)  +\frac{1}{l}\sum\nolimits_{i=1}^{l}%
y_{i}=-1\text{,}%
\]
wherein the geometric locus of the novel principal eigenaxis $\boldsymbol{\rho
}=\boldsymbol{\rho}_{1}-\boldsymbol{\rho}_{2}$ of the system is the solution
of the locus equation, so that all of the points $\mathbf{s}$ that lie on the
geometric locus of the decision border $d\left(  \mathbf{s}\right)  =-1$
exclusively reference the novel principal eigenaxis $\boldsymbol{\rho
}=\boldsymbol{\rho}_{1}-\boldsymbol{\rho}_{2}$, at which point the shape of
the decision region $Z_{2}$ of the minimum risk binary classification system
is completely determined by the exclusive intrinsic eigen-coordinate system
$\boldsymbol{\rho}=\boldsymbol{\rho}_{1}-\boldsymbol{\rho}_{2}$, such that the
shape of the decision region $Z_{2}$ is determined by the shapes of the
geometric loci of the decision border $d\left(  \mathbf{s}\right)  =-1$ and
the decision boundary $d\left(  \mathbf{s}\right)  =0$ of the system.
\end{corollary}

\begin{proof}
Corollary \ref{Locus Equations of Decision Space Corollary} is substantiated
by conditions expressed by Corollaries
\ref{Primary Integral Equation Corollary} -
\ref{Secondary Integral Equation Corollary}, Axioms
\ref{Overlapping Extreme Points Axiom} -
\ref{Non-overlapping Extreme Points Axiom}, Theorem
\ref{Principal Eigen-coordinate System Theorem} and Corollary
\ref{Principal Eigen-coordinate System Corollary}, and Theorem
\ref{Geometric Locus of a Novel Principal Eigenaxis Theorem}.

We prove Corollary \ref{Locus Equations of Decision Space Corollary} by a
constructive proof that demonstrates how a well-posed constrained optimization
algorithm resolves the inverse problem of the binary classification of random vectors.
\end{proof}

Returning now to the integral equation in (\ref{Primary Integral Equation})
expressed by Corollary \ref{Primary Integral Equation Corollary}, along with
the integral equation in (\ref{Secondary Integral Equation}) expressed by
Corollary \ref{Secondary Integral Equation Corollary}, we realize that the
total allowed eigenenergy and the expected risk exhibited by the geometric
locus of the novel principal eigenaxis $\boldsymbol{\rho}=\boldsymbol{\rho
}_{1}-\boldsymbol{\rho}_{2}$ of \textbf{any} given minimum risk binary
classification system are jointly minimized within the decision space
$Z=Z_{1}\cup Z_{2}$ of the system---including the situation of
\emph{completely overlapping distributions} of random points---such that
$100\%$ of the training data are extreme points and the error rate of the
system is $50\%$---which is the lowest possible error rate of the system.

Figure $7$ illustrates how a discriminant function of a minimum risk binary
classification system minimizes the expected risk of the system within the
decision space of the system---for completely overlapping distributions of
random points $\mathbf{x\in}$ $%
\mathbb{R}
^{2}$---such that the geometric locus of the novel principal eigenaxis of the
system is the solution of each and every one of the vector algebra locus
equations expressed by Corollary
\ref{Locus Equations of Decision Space Corollary}.

Thereby, the geometric locus of the novel principal eigenaxis represents an
eigenaxis of symmetry that spans the decision space of the minimum risk binary
classification system, so that the geometric loci of a hyperbolic decision
boundary and a pair of symmetrically positioned hyperbolic decision borders
symmetrically partition the decision space of the system. The decision
boundary is black, the decision borders are blue and red, and each extreme
point is enclosed in a black circle.%
\begin{figure}[h]%
\centering
\includegraphics[
height=2.5685in,
width=5.5988in
]%
{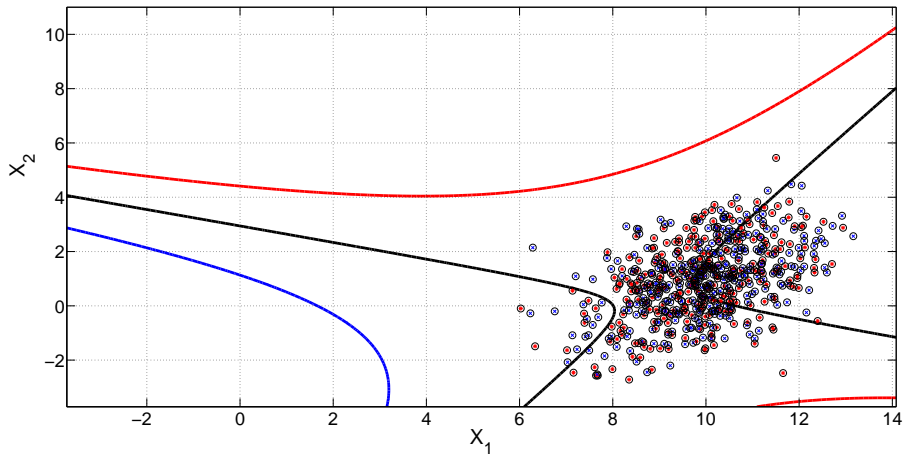}%
\caption{The constrained optimization algorithm that resolves the inverse
problem of the binary classification of random vectors finds discriminant
functions of minimum risk binary classification systems---for completely
overlapping distributions of random points---such that $100\%$ of the training
data are extreme points, and the error rate of any given system is
$50\%$---which is the lowest possible error rate of the system.}%
\end{figure}

\subsection{Joint Minimization of Eigenenergy and Risk}

Recall that a geometric locus of a novel principal eigenaxis provides dual
representation of the discriminant function, an exclusive principal
eigen-coordinate system of the geometric locus of the decision boundary, and
an eigenaxis of symmetry that spans the decision space---of any given minimum
risk binary classification system.

Given the conditions expressed by Corollaries
\ref{Primary Integral Equation Corollary} and
\ref{Secondary Integral Equation Corollary}, along with the conditions
expressed by Theorems \ref{Vector Algebra Equation of Linear Loci Theorem} -
\ref{Vector Algebra Equation of Spherical Loci Theorem}, Theorem
\ref{Principal Eigen-coordinate System Theorem} and Corollaries
\ref{Principal Eigen-coordinate System Corollary} -
\ref{Symmetrical and Equivalent Principal Eigenaxes}, we realize that any
given geometric locus of a novel principal eigenaxis exhibits a critical
minimum eigenenergy and a minimum expected risk in the manner that is
expressed by Corollary \ref{Minimization of Eigenenergy and Risk Corollary}.

\begin{corollary}
\label{Minimization of Eigenenergy and Risk Corollary}Let $\boldsymbol{\rho
}=\boldsymbol{\rho}_{1}-\boldsymbol{\rho}_{2}$ be\ the geometric locus of the
novel principal eigenaxis of any given minimum risk binary classification
system%
\[
\left(  k_{\mathbf{s}}\boldsymbol{-}\frac{1}{l}\sum\nolimits_{i=1}%
^{l}k_{\mathbf{x}_{i\ast}}\right)  \left(  \boldsymbol{\rho}_{1}%
-\boldsymbol{\rho}_{2}\right)  +\frac{1}{l}\sum\nolimits_{i=1}^{l}%
y_{i}\overset{\omega_{1}}{\underset{\omega_{2}}{\gtrless}}0
\]
that is subject to random inputs $\mathbf{x\in}$ $%
\mathbb{R}
^{d}$ such that $\mathbf{x\sim}$ $p\left(  \mathbf{x};\omega_{1}\right)  $ and
$\mathbf{x\sim}$ $p\left(  \mathbf{x};\omega_{2}\right)  $, where $p\left(
\mathbf{x};\omega_{1}\right)  $ and $p\left(  \mathbf{x};\omega_{2}\right)  $
are certain probability density functions for two classes $\omega_{1}$ and
$\omega_{2}$ of random vectors $\mathbf{x\in}$ $%
\mathbb{R}
^{d}$, at which point the discriminant function%
\[
d\left(  \mathbf{s}\right)  =\left(  k_{\mathbf{s}}\boldsymbol{-}\frac{1}%
{l}\sum\nolimits_{i=1}^{l}k_{\mathbf{x}_{i\ast}}\right)  \left(
\boldsymbol{\rho}_{1}-\boldsymbol{\rho}_{2}\right)  +\frac{1}{l}%
\sum\nolimits_{i=1}^{l}y_{i}%
\]
of the system is represented by the geometric locus of the novel principal
eigenaxis $\boldsymbol{\rho}=\boldsymbol{\rho}_{1}-\boldsymbol{\rho}_{2}$.

The geometric locus of the novel principal eigenaxis $\boldsymbol{\rho
}=\boldsymbol{\rho}_{1}-\boldsymbol{\rho}_{2}$ exhibits a total allowed
eigenenergy and an expected risk in such a manner that the dual locus
$\boldsymbol{\rho}=\boldsymbol{\rho}_{1}-\boldsymbol{\rho}_{2}$ of the
discriminant function of the minimum risk binary classification system is in
statistical equilibrium at the geometric locus of the decision boundary of the
system, at which point the geometric locus of the novel principal eigenaxis
$\boldsymbol{\rho}=\boldsymbol{\rho}_{1}-\boldsymbol{\rho}_{2}$ is an
eigenaxis of symmetry that satisfies the law of cosines in the symmetrically
balanced manner%
\begin{align}
\frac{1}{2}\left\Vert \boldsymbol{\rho}\right\Vert _{\min}^{2}  &  =\left\Vert
\boldsymbol{\rho}_{1}\right\Vert _{\min}^{2}-\left\Vert \boldsymbol{\rho}%
_{1}\right\Vert \left\Vert \boldsymbol{\rho}_{2}\right\Vert \cos
\theta_{\boldsymbol{\rho}_{1}\boldsymbol{\rho}_{2}} \tag{10.7}%
\label{Symmetrical Constraints}\\
&  =\left\Vert \boldsymbol{\rho}_{2}\right\Vert _{\min}^{2}-\left\Vert
\boldsymbol{\rho}_{2}\right\Vert \left\Vert \boldsymbol{\rho}_{1}\right\Vert
\cos\theta_{\boldsymbol{\rho}_{2}\boldsymbol{\rho}_{1}}\text{,}\nonumber
\end{align}
so that the dual locus $\boldsymbol{\rho}=\boldsymbol{\rho}_{1}%
-\boldsymbol{\rho}_{2}$ of the discriminant function satisfies the geometric
locus of the decision boundary in terms of a critical minimum eigenenergy and
a minimum expected risk, such that critical minimum eigenenergies related to
likely locations of extreme points $\mathbf{x}_{1_{i\ast}}\mathbf{\sim}$
$p\left(  \mathbf{x};\omega_{1}\right)  $ and $\mathbf{x}_{2_{i\ast}%
}\mathbf{\sim}$ $p\left(  \mathbf{x};\omega_{2}\right)  $ determine
conditional probabilities that the extreme points $\mathbf{x}_{1_{i\ast}}$ and
$\mathbf{x}_{2_{i\ast}}$ will be observed within localized areas of counter
risk and risk throughout the decision space $Z=Z_{1}\cup Z_{2}$ of the system,
such that the magnitude and the direction of the novel principal eigenaxis
$\boldsymbol{\rho}$ $=\boldsymbol{\rho}_{1}-\boldsymbol{\rho}_{2}$ are both
functions of differences between joint variabilities of extreme vectors
$\mathbf{x}_{1_{i\ast}}\mathbf{\sim}p\left(  \mathbf{x};\omega_{1}\right)  $
and $\mathbf{x}_{2_{i\ast}}\mathbf{\sim}p\left(  \mathbf{x};\omega_{1}\right)
$ that belong to the two classes $\omega_{1}$ and $\omega_{2}$ of random
vectors $\mathbf{x\in}$ $%
\mathbb{R}
^{d}$.
\end{corollary}

\begin{proof}
We prove Corollary \ref{Minimization of Eigenenergy and Risk Corollary} by a
constructive proof that demonstrates how a well-posed constrained optimization
algorithm resolves the inverse problem of the binary classification of random vectors.
\end{proof}

Returning again to the integral equations expressed by Corollaries
\ref{Primary Integral Equation Corollary} and
\ref{Secondary Integral Equation Corollary}, it follows that the discriminant
function of any given minimum risk binary classification system satisfies a
pair of fundamental integral equations of binary classification.

\subsection{Integral Equations of Binary Classification}

Corollary \ref{Primary Integral Equation of Binary Classification Corollary}
expresses the requirement that a discriminant function of a minimum risk
binary classification system is the solution of an integral
equation---corresponding to the integral equation in
(\ref{Primary Integral Equation}), over the decision space $Z=Z_{1}\cup Z_{2}$
of the system, so that the total allowed eigenenergy and the expected risk
exhibited by the system are regulated by the equilibrium requirement on the
dual locus of the discriminant function at the geometric locus of the decision
boundary of the system expressed by Corollary
\ref{Minimization of Eigenenergy and Risk Corollary}, at which point counter
risks\ and risks exhibited by the system are symmetrically balanced with each
other throughout the decision regions $Z_{1}$ and $Z_{2}$ of the system.

\begin{corollary}
\label{Primary Integral Equation of Binary Classification Corollary}Take the
discriminant function%
\[
d\left(  \mathbf{s}\right)  =\left(  k_{\mathbf{s}}\boldsymbol{-}\frac{1}%
{l}\sum\nolimits_{i=1}^{l}k_{\mathbf{x}_{i\ast}}\right)  \left(
\boldsymbol{\rho}_{1}-\boldsymbol{\rho}_{2}\right)  +\frac{1}{l}%
\sum\nolimits_{i=1}^{l}y_{i}%
\]
of any given minimum risk binary classification system%
\[
\left(  k_{\mathbf{s}}\boldsymbol{-}\frac{1}{l}\sum\nolimits_{i=1}%
^{l}k_{\mathbf{x}_{i\ast}}\right)  \left(  \boldsymbol{\rho}_{1}%
-\boldsymbol{\rho}_{2}\right)  +\frac{1}{l}\sum\nolimits_{i=1}^{l}%
y_{i}\overset{\omega_{1}}{\underset{\omega_{2}}{\gtrless}}0
\]
that is subject to random inputs $\mathbf{x\in}$ $%
\mathbb{R}
^{d}$ such that $\mathbf{x\sim}$ $p\left(  \mathbf{x};\omega_{1}\right)  $ and
$\mathbf{x\sim}$ $p\left(  \mathbf{x};\omega_{2}\right)  $, where $p\left(
\mathbf{x};\omega_{1}\right)  $ and $p\left(  \mathbf{x};\omega_{2}\right)  $
are certain probability density functions for two classes $\omega_{1}$ and
$\omega_{2}$ of random vectors $\mathbf{x\in}$ $%
\mathbb{R}
^{d}$, at which point the discriminant function of the system is represented
by the geometric locus of a novel principal eigenaxis $\boldsymbol{\rho
}=\boldsymbol{\rho}_{1}-\boldsymbol{\rho}_{2}$.

Given the integral equation in (\ref{Primary Integral Equation}) expressed by
Corollary \ref{Primary Integral Equation Corollary} and the conditions
expressed by (\ref{Symmetrical Constraints}) in Corollary
\ref{Minimization of Eigenenergy and Risk Corollary}, it follows that the
discriminant function is the solution of the integral equation%
\begin{align}
f_{1}\left(  d\left(  \mathbf{s}\right)  \right)   &  :\int_{Z_{1}%
}\boldsymbol{\rho}_{1}d\boldsymbol{\rho}_{1}+\int_{Z_{2}}\boldsymbol{\rho}%
_{1}d\boldsymbol{\rho}_{1}+C_{1} \tag{10.8}%
\label{Primary Locus Integral Equation}\\
&  =\int_{Z_{1}}\boldsymbol{\rho}_{2}d\boldsymbol{\rho}_{2}+\int_{Z_{2}%
}\boldsymbol{\rho}_{2}d\boldsymbol{\rho}_{2}+C_{2}\text{,}\nonumber
\end{align}
over the decision space $Z=Z_{1}\cup Z_{2}$ of the system, where $C_{1}$ and
$C_{2}$ are certain integration constants, so that the total allowed
eigenenergy $\left\Vert \boldsymbol{\rho}_{1}-\boldsymbol{\rho}_{2}\right\Vert
_{\min_{c}}^{2}$ and the expected risk $\mathfrak{R}_{\mathfrak{\min}}\left(
\left\Vert \boldsymbol{\rho}_{1}-\boldsymbol{\rho}_{2}\right\Vert _{\min_{c}%
}^{2}\right)  $ exhibited by the system are jointly regulated by the
equilibrium requirement on the dual locus $\boldsymbol{\rho}=\boldsymbol{\rho
}_{1}-\boldsymbol{\rho}_{2}$ of the discriminant function at the geometric
locus of the decision boundary of the system%
\begin{align*}
d\left(  \mathbf{s}\right)   &  :\left\Vert \boldsymbol{\rho}_{1}\right\Vert
_{\min_{c}}^{2}-\left\Vert \boldsymbol{\rho}_{1}\right\Vert \left\Vert
\boldsymbol{\rho}_{2}\right\Vert \cos\theta_{\boldsymbol{\rho}_{1}%
\boldsymbol{\rho}_{2}}\\
&  =\left\Vert \boldsymbol{\rho}_{2}\right\Vert _{\min_{c}}^{2}-\left\Vert
\boldsymbol{\rho}_{2}\right\Vert \left\Vert \boldsymbol{\rho}_{1}\right\Vert
\cos\theta_{\boldsymbol{\rho}_{2}\boldsymbol{\rho}_{1}}\\
&  =\frac{1}{2}\left\Vert \boldsymbol{\rho}_{1}-\boldsymbol{\rho}%
_{2}\right\Vert _{\min_{c}}^{2}\text{,}%
\end{align*}
at which point the dual locus $\boldsymbol{\rho}=\boldsymbol{\rho}%
_{1}-\boldsymbol{\rho}_{2}$ of the discriminant function satisfies the
geometric locus of the decision boundary in terms of a critical minimum
eigenenergy $\left\Vert \boldsymbol{\rho}\right\Vert _{\min_{c}}^{2}$ and a
minimum expected risk $\mathfrak{R}_{\mathfrak{\min}}\left(  \left\Vert
\boldsymbol{\rho}\right\Vert _{\min_{c}}^{2}\right)  $ in such a manner that
regions of counter risks of the system are symmetrically balanced with regions
of risks of the system.

Thereby, critical minimum eigenenergies $\left\Vert \psi_{1_{i\ast}%
}k_{\mathbf{x}_{1_{i\ast}}}\right\Vert _{\min_{c}}^{2}$ exhibited by principal
eigenaxis components $\psi_{1_{i_{\ast}}}k_{\mathbf{x}_{1_{i\ast}}}$ on side
$\boldsymbol{\rho}_{1}$ of the novel principal eigenaxis $\boldsymbol{\rho
}=\boldsymbol{\rho}_{1}-\boldsymbol{\rho}_{2}$---that determine probabilities
of finding extreme points $\mathbf{x}_{1_{i\ast}}$ located throughout the
decision space $Z=Z_{1}\cup Z_{2}$ of the system, are symmetrically balanced
with critical minimum eigenenergies $\left\Vert \psi_{2_{i_{\ast}}%
}k_{\mathbf{x}_{2_{i\ast}}}\right\Vert _{\min_{c}}^{2}$ exhibited by principal
eigenaxis components $\psi_{2_{i_{\ast}}}k_{\mathbf{x}_{2_{i\ast}}}$ on side
$\boldsymbol{\rho}_{2}$ of the novel principal eigenaxis $\boldsymbol{\rho
}=\boldsymbol{\rho}_{1}-\boldsymbol{\rho}_{2}$---that determine probabilities
of finding extreme points $\mathbf{x}_{2_{i\ast}}$ located throughout the
decision space $Z=Z_{1}\cup Z_{2}$ of the system.
\end{corollary}

\begin{proof}
We prove Corollary
\ref{Primary Integral Equation of Binary Classification Corollary} by a
constructive proof that demonstrates how a well-posed constrained optimization
algorithm resolves the inverse problem of the binary classification of random vectors.
\end{proof}

Corollary \ref{Secondary Integral Equation of Binary Classification Corollary}
expresses the requirement that a discriminant function of a minimum risk
binary classification system minimize an integral equation over the decision
regions $Z_{1}$ and $Z_{2}$ of the system, so that the total allowed
eigenenergy and the expected risk exhibited by the system are jointly
minimized within the decision space $Z=Z_{1}\cup Z_{2}$ of the system in such
a manner that the system satisfies a state of statistical equilibrium, at
which point regions of counter risks and risks of the system---located
throughout the decision region $Z_{1}$ of the system---are symmetrically
balanced with regions of counter risks and risks of the system---located
throughout the decision region $Z_{2}$ of the system. Thereby, the minimum
risk binary classification system satisfies a state of statistical equilibrium
so that the total allowed eigenenergy and the expected risk exhibited by the
system are jointly minimized within the decision space $Z=Z_{1}\cup Z_{2}$ of
the system, at which point the system exhibits the minimum probability of
classification error.

The integral equation expressed by Corollary
\ref{Secondary Integral Equation of Binary Classification Corollary} is
derived from the integral equation in (\ref{Primary Locus Integral Equation})
expressed by Corollary
\ref{Primary Integral Equation of Binary Classification Corollary}.

\begin{corollary}
\label{Secondary Integral Equation of Binary Classification Corollary}Take the
discriminant function%
\[
d\left(  \mathbf{s}\right)  =\left(  k_{\mathbf{s}}\boldsymbol{-}\frac{1}%
{l}\sum\nolimits_{i=1}^{l}k_{\mathbf{x}_{i\ast}}\right)  \left(
\boldsymbol{\rho}_{1}-\boldsymbol{\rho}_{2}\right)  +\frac{1}{l}%
\sum\nolimits_{i=1}^{l}y_{i}%
\]
of any given minimum risk binary classification system%
\[
\left(  k_{\mathbf{s}}\boldsymbol{-}\frac{1}{l}\sum\nolimits_{i=1}%
^{l}k_{\mathbf{x}_{i\ast}}\right)  \left(  \boldsymbol{\rho}_{1}%
-\boldsymbol{\rho}_{2}\right)  +\frac{1}{l}\sum\nolimits_{i=1}^{l}%
y_{i}\overset{\omega_{1}}{\underset{\omega_{2}}{\gtrless}}0
\]
that is subject to random inputs $\mathbf{x\in}$ $%
\mathbb{R}
^{d}$ such that $\mathbf{x\sim}$ $p\left(  \mathbf{x};\omega_{1}\right)  $ and
$\mathbf{x\sim}$ $p\left(  \mathbf{x};\omega_{2}\right)  $, where $p\left(
\mathbf{x};\omega_{1}\right)  $ and $p\left(  \mathbf{x};\omega_{2}\right)  $
are certain probability density functions for two classes $\omega_{1}$ and
$\omega_{2}$ of random vectors $\mathbf{x\in}$ $%
\mathbb{R}
^{d}$, at which point the discriminant function of the system is represented
by the geometric locus of a novel principal eigenaxis $\boldsymbol{\rho
}=\boldsymbol{\rho}_{1}-\boldsymbol{\rho}_{2}$.

Given the integral equation in (\ref{Secondary Integral Equation}) expressed
by Corollary \ref{Secondary Integral Equation Corollary} and the integral
equation in (\ref{Primary Locus Integral Equation}) expressed by Corollary
\ref{Primary Integral Equation of Binary Classification Corollary}, it follows
that the discriminant function minimizes the integral equation%
\begin{align}
f_{2}\left(  d\left(  \mathbf{s}\right)  \right)   &  :\int_{Z_{1}%
}\boldsymbol{\rho}_{1}d\boldsymbol{\rho}_{1}-\int_{Z_{1}}\boldsymbol{\rho}%
_{2}d\boldsymbol{\rho}_{2}+C_{1} \tag{10.9}%
\label{Secondary Locus Integral Equation}\\
&  =\int_{Z_{2}}\boldsymbol{\rho}_{2}d\boldsymbol{\rho}_{2}-\int_{Z_{2}%
}\boldsymbol{\rho}_{1}d\boldsymbol{\rho}_{1}+C_{2}\text{,}\nonumber
\end{align}
over the decision regions $Z_{1}$ and $Z_{2}$ of the system, where $C_{1}$ and
$C_{2}$ are certain integration constants, so that the system satisfies a
state of statistical equilibrium such that the total allowed eigenenergy
$\left\Vert \boldsymbol{\rho}_{1}-\boldsymbol{\rho}_{2}\right\Vert _{\min_{c}%
}^{2}$ and the expected risk $\mathfrak{R}_{\mathfrak{\min}}\left(  \left\Vert
\boldsymbol{\rho}_{1}-\boldsymbol{\rho}_{2}\right\Vert _{\min_{c}}^{2}\right)
$ exhibited by the system are jointly minimized within the decision space
$Z=Z_{1}\cup Z_{2}$ of the system in such a manner that critical minimum
eigenenergies $\left\Vert \psi_{1_{i\ast}}k_{\mathbf{x}_{1_{i\ast}}%
}\right\Vert _{\min_{c}}^{2}$ and $\left\Vert \psi_{2_{i_{\ast}}}%
k_{\mathbf{x}_{2_{i\ast}}}\right\Vert _{\min_{c}}^{2}$ exhibited by
corresponding principal eigenaxis components $\psi_{1_{i_{\ast}}}%
k_{\mathbf{x}_{1_{i\ast}}}$ and $\psi_{2_{i_{\ast}}}k_{\mathbf{x}_{2_{i\ast}}%
}$ that lie on side $\boldsymbol{\rho}_{1}$ and side $\boldsymbol{\rho}_{2}$
of the novel principal eigenaxis $\boldsymbol{\rho}=\boldsymbol{\rho}%
_{1}-\boldsymbol{\rho}_{2}$ are minimized throughout the decision regions
$Z_{1}$ and $Z_{2}$ of the system, at which point regions of counter risks and
risks of the system---located throughout the decision region $Z_{1}$ of the
system---are symmetrically balanced with regions of counter risks and risks of
the system---located throughout the decision region $Z_{2}$ of the system.

Thereby, the minimum risk binary classification system satisfies a state of
statistical equilibrium so that the total allowed eigenenergy and the expected
risk exhibited by the system are jointly minimized within the decision space
$Z=Z_{1}\cup Z_{2}$ of the system, at which point the system exhibits the
minimum probability of classification error for any given random vectors
$\mathbf{x\in}$ $%
\mathbb{R}
^{d}$ such that $\mathbf{x\sim}$ $p\left(  \mathbf{x};\omega_{1}\right)  $ and
$\mathbf{x\sim}$ $p\left(  \mathbf{x};\omega_{2}\right)  $.
\end{corollary}

\begin{proof}
We prove Corollary
\ref{Secondary Integral Equation of Binary Classification Corollary} by a
constructive proof that demonstrates how a well-posed constrained optimization
algorithm resolves the inverse problem of the binary classification of random vectors.
\end{proof}

We are now in a position to express the direct problem of the binary
classification of random vectors---according to the theoretical model that we
have developed. Theorem \ref{Direct Problem of Binary Classification Theorem}
is an existence theorem that expresses fundamental laws of binary
classification---that discriminant functions of minimum risk binary
classification systems are subject to---in terms of a general locus formula.

\section{\label{Section 11}The Direct Problem}

\begin{theorem}
\label{Direct Problem of Binary Classification Theorem}Let%
\begin{equation}
\left(  k_{\mathbf{s}}\boldsymbol{-}\frac{1}{l}\sum\nolimits_{i=1}%
^{l}k_{\mathbf{x}_{i\ast}}\right)  \left(  \boldsymbol{\rho}_{1}%
-\boldsymbol{\rho}_{2}\right)  +\frac{1}{l}\sum\nolimits_{i=1}^{l}%
y_{i}\overset{\omega_{1}}{\underset{\omega_{2}}{\gtrless}}0 \tag{11.1}%
\label{Theoretical System}%
\end{equation}
be any given minimum risk binary classification system that is subject to
random inputs $\mathbf{x\in}$ $%
\mathbb{R}
^{d}$ such that $\mathbf{x\sim}$ $p\left(  \mathbf{x};\omega_{1}\right)  $ and
$\mathbf{x\sim}$ $p\left(  \mathbf{x};\omega_{2}\right)  $, where $p\left(
\mathbf{x};\omega_{1}\right)  $ and $p\left(  \mathbf{x};\omega_{2}\right)  $
are certain probability density functions for two classes $\omega_{1}$ and
$\omega_{2}$ of random vectors $\mathbf{x\in}$ $%
\mathbb{R}
^{d}$, where $y_{i}=\pm1$ and $\omega_{1}$ or $\omega_{2}$ is the true
category, satisfying the following geometrical and statistical criteria:

$1$. The discriminant function%
\begin{equation}
d\left(  \mathbf{s}\right)  =\left(  k_{\mathbf{s}}\boldsymbol{-}\frac{1}%
{l}\sum\nolimits_{i=1}^{l}k_{\mathbf{x}_{i\ast}}\right)  \left(
\boldsymbol{\rho}_{1}-\boldsymbol{\rho}_{2}\right)  +\frac{1}{l}%
\sum\nolimits_{i=1}^{l}y_{i}\tag{11.2}\label{Decision Function T}%
\end{equation}
is represented by a geometric locus of a novel principal eigenaxis%
\begin{align}
\boldsymbol{\rho} &  =\sum\nolimits_{i=1}^{l_{1}}\psi_{1_{i_{\ast}}%
}k_{\mathbf{x}_{1_{i\ast}}}-\sum\nolimits_{i=1}^{l_{2}}\psi_{2_{i_{\ast}}%
}k_{\mathbf{x}_{2_{i\ast}}}\tag{11.3}\label{Novel Principal Eigenaxis T}\\
&  =\boldsymbol{\rho}_{1}-\boldsymbol{\rho}_{2}\nonumber
\end{align}
structured as a locus of signed and scaled extreme vectors $\psi_{1_{i_{\ast}%
}}k_{\mathbf{x}_{1_{i\ast}}}$ and $-\psi_{2_{i_{\ast}}}k_{\mathbf{x}%
_{2_{i\ast}}}$, so that a dual locus of likelihood components and principal
eigenaxis components $\psi_{1_{i_{\ast}}}k_{\mathbf{x}_{1_{i\ast}}}$ and
$\psi_{2_{i_{\ast}}}k_{\mathbf{x}_{2_{i\ast}}}$ represents an exclusive
principal eigen-coordinate system of the geometric locus of the decision
boundary of the system, and also represents an eigenaxis of symmetry that
spans the decision space of the system, such that each scale factor
$\psi_{1_{i_{\ast}}}$ or $\psi_{2_{i_{\ast}}}$ determines a scaled extreme
vector $\psi_{1_{i_{\ast}}}k_{\mathbf{x}_{1_{i\ast}}}$ or $\psi_{2_{i_{\ast}}%
}k_{\mathbf{x}_{2_{i\ast}}}$ that represents a principal eigenaxis component
that determines a likely location for a correlated extreme point
$\mathbf{x}_{1_{i\ast}}\mathbf{\sim}$ $p\left(  \mathbf{x};\omega_{1}\right)
$ or $\mathbf{x}_{2_{i\ast}}\mathbf{\sim}$ $p\left(  \mathbf{x};\omega
_{2}\right)  $, along with a likelihood component that determines a likelihood
value for the correlated extreme point $\mathbf{x}_{1_{i\ast}}$ or
$\mathbf{x}_{2_{i\ast}}$, where the reproducing kernel for each extreme point
$k_{\mathbf{x}_{1_{i\ast}}}$ and $k_{\mathbf{x}_{2_{i\ast}}}$ has the
preferred form of either $k_{\mathbf{x}}\left(  \mathbf{s}\right)  =\left(
\mathbf{s}^{T}\mathbf{x}+1\right)  ^{2}$ or $k_{\mathbf{x}}\left(
\mathbf{s}\right)  =\exp\left(  -\gamma\left\Vert \mathbf{s}-\mathbf{x}%
\right\Vert ^{2}\right)  $, wherein $0.01\leq\gamma\leq0.1$;

$2$. The geometric locus of the novel principal eigenaxis $\boldsymbol{\rho
}=\boldsymbol{\rho}_{1}-\boldsymbol{\rho}_{2}$ is the solution of the vector
algebra locus equation%
\begin{equation}
\left(  k_{\mathbf{s}}\boldsymbol{-}\frac{1}{l}\sum\nolimits_{i=1}%
^{l}k_{\mathbf{x}_{i\ast}}\right)  \left(  \boldsymbol{\rho}_{1}%
-\boldsymbol{\rho}_{2}\right)  +\frac{1}{l}\sum\nolimits_{i=1}^{l}y_{i}=0
\tag{11.4}\label{Equation of Decision Boundary}%
\end{equation}
that represents the geometric locus of the decision boundary of the system,
where the expression $\frac{1}{l}\sum\nolimits_{i=1}^{l}k_{\mathbf{x}_{i\ast}%
}$ represents a locus of average risk in the decision space $Z=Z_{1}\cup
Z_{2}$ of the system, and the statistic $\frac{1}{l}\sum\nolimits_{i=1}%
^{l}y_{i}:y_{i}=\pm1$ represents an expected likelihood of observing $l$
extreme vectors $\left\{  k_{\mathbf{x}_{i\ast}}\right\}  _{i=1}^{l}$ within
the decision space $Z=Z_{1}\cup Z_{2}$, so that all of the points $\mathbf{s}$
that lie on the geometric locus of the decision boundary exclusively reference
the novel principal eigenaxis $\boldsymbol{\rho}=\boldsymbol{\rho}%
_{1}-\boldsymbol{\rho}_{2}$, as well as the vector algebra locus equations%
\begin{equation}
\left(  k_{\mathbf{s}}\boldsymbol{-}\frac{1}{l}\sum\nolimits_{i=1}%
^{l}k_{\mathbf{x}_{i\ast}}\right)  \left(  \boldsymbol{\rho}_{1}%
-\boldsymbol{\rho}_{2}\right)  +\frac{1}{l}\sum\nolimits_{i=1}^{l}y_{i}=+1
\tag{11.5}\label{Equation of Border 1}%
\end{equation}
and%
\begin{equation}
\left(  k_{\mathbf{s}}\boldsymbol{-}\frac{1}{l}\sum\nolimits_{i=1}%
^{l}k_{\mathbf{x}_{i\ast}}\right)  \left(  \boldsymbol{\rho}_{1}%
-\boldsymbol{\rho}_{2}\right)  +\frac{1}{l}\sum\nolimits_{i=1}^{l}y_{i}=-1
\tag{11.6}\label{Equation of Border 2}%
\end{equation}
that represent the geometric loci of the decision borders of the corresponding
decision regions $Z_{1}$ and $Z_{2}$ of the system, so that all of the points
$\mathbf{s}$ that lie on the geometric loci of the decision borders
exclusively reference the novel principal eigenaxis $\boldsymbol{\rho
}=\boldsymbol{\rho}_{1}-\boldsymbol{\rho}_{2}$.

Thereby, the geometric locus of the novel principal eigenaxis
$\boldsymbol{\rho}=\boldsymbol{\rho}_{1}-\boldsymbol{\rho}_{2}$ represents an
eigenaxis of symmetry that spans the decision space $Z=Z_{1}\cup Z_{2}$ of the
minimum risk binary classification system%
\[
\left(  k_{\mathbf{s}}\boldsymbol{-}\frac{1}{l}\sum\nolimits_{i=1}%
^{l}k_{\mathbf{x}_{i\ast}}\right)  \left(  \boldsymbol{\rho}_{1}%
-\boldsymbol{\rho}_{2}\right)  +\frac{1}{l}\sum\nolimits_{i=1}^{l}%
y_{i}\overset{\omega_{1}}{\underset{\omega_{2}}{\gtrless}}0\text{,}%
\]
at which point the shape of the decision space $Z=Z_{1}\cup Z_{2}$ is
completely determined by the exclusive principal eigen-coordinate system
$\boldsymbol{\rho}=\boldsymbol{\rho}_{1}-\boldsymbol{\rho}_{2}$;

$3$. The discriminant function is the solution of the integral equation%
\begin{align}
f_{1}\left(  d\left(  \mathbf{s}\right)  \right)   &  :\int_{Z_{1}%
}\boldsymbol{\rho}_{1}d\boldsymbol{\rho}_{1}+\int_{Z_{2}}\boldsymbol{\rho}%
_{1}d\boldsymbol{\rho}_{1}+C_{1}\tag{11.7}\label{TIE1}\\
&  =\int_{Z_{1}}\boldsymbol{\rho}_{2}d\boldsymbol{\rho}_{2}+\int_{Z_{2}%
}\boldsymbol{\rho}_{2}d\boldsymbol{\rho}_{2}+C_{2}\text{,}\nonumber
\end{align}
over the decision space $Z=Z_{1}\cup Z_{2}$ of the minimum risk binary
classification system%
\[
\left(  k_{\mathbf{s}}\boldsymbol{-}\frac{1}{l}\sum\nolimits_{i=1}%
^{l}k_{\mathbf{x}_{i\ast}}\right)  \left(  \boldsymbol{\rho}_{1}%
-\boldsymbol{\rho}_{2}\right)  +\frac{1}{l}\sum\nolimits_{i=1}^{l}%
y_{i}\overset{\omega_{1}}{\underset{\omega_{2}}{\gtrless}}0\text{,}%
\]
where $C_{1}$ and $C_{2}$ are certain integration constants, so that the total
allowed eigenenergy $\left\Vert \boldsymbol{\rho}_{1}-\boldsymbol{\rho}%
_{2}\right\Vert _{\min_{c}}^{2}$ and the expected risk $\mathfrak{R}%
_{\mathfrak{\min}}\left(  \left\Vert \boldsymbol{\rho}_{1}-\boldsymbol{\rho
}_{2}\right\Vert _{\min_{c}}^{2}\right)  $ exhibited by the system are jointly
regulated by the equilibrium requirement on the dual locus $\boldsymbol{\rho
}=\boldsymbol{\rho}_{1}-\boldsymbol{\rho}_{2}$ of the discriminant function at
the geometric locus of the decision boundary of the system%
\begin{align*}
d\left(  \mathbf{s}\right)   &  :\left\Vert \boldsymbol{\rho}_{1}\right\Vert
_{\min_{c}}^{2}-\left\Vert \boldsymbol{\rho}_{1}\right\Vert \left\Vert
\boldsymbol{\rho}_{2}\right\Vert \cos\theta_{\boldsymbol{\rho}_{1}%
\boldsymbol{\rho}_{2}}\\
&  =\left\Vert \boldsymbol{\rho}_{2}\right\Vert _{\min_{c}}^{2}-\left\Vert
\boldsymbol{\rho}_{2}\right\Vert \left\Vert \boldsymbol{\rho}_{1}\right\Vert
\cos\theta_{\boldsymbol{\rho}_{2}\boldsymbol{\rho}_{1}}\\
&  =\frac{1}{2}\left\Vert \boldsymbol{\rho}_{1}-\boldsymbol{\rho}%
_{2}\right\Vert _{\min_{c}}^{2}\text{,}%
\end{align*}
at which point the dual locus $\boldsymbol{\rho}=\boldsymbol{\rho}%
_{1}-\boldsymbol{\rho}_{2}$ of the discriminant function satisfies the
geometric locus of the decision boundary in terms of a critical minimum
eigenenergy $\left\Vert \boldsymbol{\rho}\right\Vert _{\min_{c}}^{2}$ and a
minimum expected risk $\mathfrak{R}_{\mathfrak{\min}}\left(  \left\Vert
\boldsymbol{\rho}\right\Vert _{\min_{c}}^{2}\right)  $ in such a manner that
regions of counter risks of the system are symmetrically balanced with regions
of risks of the system, so that critical minimum eigenenergies $\left\Vert
\psi_{1_{i\ast}}k_{\mathbf{x}_{1_{i\ast}}}\right\Vert _{\min_{c}}^{2}$
exhibited by principal eigenaxis components $\psi_{1_{i_{\ast}}}%
k_{\mathbf{x}_{1_{i\ast}}}$ on side $\boldsymbol{\rho}_{1}$ of the novel
principal eigenaxis $\boldsymbol{\rho}=\boldsymbol{\rho}_{1}-\boldsymbol{\rho
}_{2}$---that determine probabilities of finding extreme points $\mathbf{x}%
_{1_{i\ast}}$ located throughout the decision space $Z=Z_{1}\cup Z_{2}$ of the
system, are symmetrically balanced with critical minimum eigenenergies
$\left\Vert \psi_{2_{i_{\ast}}}k_{\mathbf{x}_{2_{i\ast}}}\right\Vert
_{\min_{c}}^{2}$ exhibited by principal eigenaxis components $\psi
_{2_{i_{\ast}}}k_{\mathbf{x}_{2_{i\ast}}}$ on side $\boldsymbol{\rho}_{2}$ of
the novel principal eigenaxis $\boldsymbol{\rho}=\boldsymbol{\rho}%
_{1}-\boldsymbol{\rho}_{2}$---that determine probabilities of finding extreme
points $\mathbf{x}_{2_{i\ast}}$ located throughout the decision space
$Z=Z_{1}\cup Z_{2}$ of the system;

$4$. The discriminant function minimizes the integral equation%
\begin{align}
f_{2}\left(  d\left(  \mathbf{s}\right)  \right)   &  :\int_{Z_{1}%
}\boldsymbol{\rho}_{1}d\boldsymbol{\rho}_{1}-\int_{Z_{1}}\boldsymbol{\rho}%
_{2}d\boldsymbol{\rho}_{2}+C_{1}\tag{11.8}\label{TIE2}\\
&  =\int_{Z_{2}}\boldsymbol{\rho}_{2}d\boldsymbol{\rho}_{2}-\int_{Z_{2}%
}\boldsymbol{\rho}_{1}d\boldsymbol{\rho}_{1}+C_{2}\text{,}\nonumber
\end{align}
over the decision regions $Z_{1}$ and $Z_{2}$ of the minimum risk binary
classification system $\left(  k_{\mathbf{s}}\boldsymbol{-}\frac{1}{l}%
\sum\nolimits_{i=1}^{l}k_{\mathbf{x}_{i\ast}}\right)  \left(  \boldsymbol{\rho
}_{1}-\boldsymbol{\rho}_{2}\right)  +\frac{1}{l}\sum\nolimits_{i=1}^{l}%
y_{i}\overset{\omega_{1}}{\underset{\omega_{2}}{\gtrless}}0$, where $C_{1}$
and $C_{2}$ are certain integration constants, so that the system satisfies a
state of statistical equilibrium such that the total allowed eigenenergy
$\left\Vert \boldsymbol{\rho}_{1}-\boldsymbol{\rho}_{2}\right\Vert _{\min_{c}%
}^{2}$ and the expected risk $\mathfrak{R}_{\mathfrak{\min}}\left(  \left\Vert
\boldsymbol{\rho}_{1}-\boldsymbol{\rho}_{2}\right\Vert _{\min_{c}}^{2}\right)
$ exhibited by the system are jointly minimized within the decision space
$Z=Z_{1}\cup Z_{2}$ of the system, at which point critical minimum
eigenenergies $\left\Vert \psi_{1_{i\ast}}k_{\mathbf{x}_{1_{i\ast}}%
}\right\Vert _{\min_{c}}^{2}$ and $\left\Vert \psi_{2_{i_{\ast}}}%
k_{\mathbf{x}_{2_{i\ast}}}\right\Vert _{\min_{c}}^{2}$ exhibited by
corresponding principal eigenaxis components $\psi_{1_{i_{\ast}}}%
k_{\mathbf{x}_{1_{i\ast}}}$ and $\psi_{2_{i_{\ast}}}k_{\mathbf{x}_{2_{i\ast}}%
}$ on side $\boldsymbol{\rho}_{1}$ and side $\boldsymbol{\rho}_{2}$ of the
novel principal eigenaxis $\boldsymbol{\rho}=\boldsymbol{\rho}_{1}%
-\boldsymbol{\rho}_{2}$ are minimized throughout the decision regions $Z_{1}$
and $Z_{2}$ of the system, so that regions of counter risks and risks of the
system---located throughout the decision region $Z_{1}$ of the system---are
symmetrically balanced with regions of counter risks and risks of the
system---located throughout the decision region $Z_{2}$ of the system.

Thereby, the minimum risk binary classification system%
\[
\left(  k_{\mathbf{s}}\boldsymbol{-}\frac{1}{l}\sum\nolimits_{i=1}%
^{l}k_{\mathbf{x}_{i\ast}}\right)  \left(  \boldsymbol{\rho}_{1}%
-\boldsymbol{\rho}_{2}\right)  +\frac{1}{l}\sum\nolimits_{i=1}^{l}%
y_{i}\overset{\omega_{1}}{\underset{\omega_{2}}{\gtrless}}0
\]
satisfies a state of statistical equilibrium so that the total allowed
eigenenergy and the expected risk exhibited by the system are jointly
minimized within the decision space $Z=Z_{1}\cup Z_{2}$ of the system, at
which point the system exhibits the minimum probability of classification
error for any given random vectors $\mathbf{x\in}$ $%
\mathbb{R}
^{d}$ such that $\mathbf{x\sim}$ $p\left(  \mathbf{x};\omega_{1}\right)  $ and
$\mathbf{x\sim}$ $p\left(  \mathbf{x};\omega_{2}\right)  $;

$5$. The geometric locus of the novel principal eigenaxis $\boldsymbol{\rho
}=\boldsymbol{\rho}_{1}-\boldsymbol{\rho}_{2}$ satisfies the law of cosines in
the symmetrically balanced manner%
\begin{align}
\frac{1}{2}\left\Vert \boldsymbol{\rho}\right\Vert _{\min_{c}}^{2} &
=\left\Vert \boldsymbol{\rho}_{1}\right\Vert _{\min_{c}}^{2}-\left\Vert
\boldsymbol{\rho}_{1}\right\Vert \left\Vert \boldsymbol{\rho}_{2}\right\Vert
\cos\theta_{\boldsymbol{\rho}_{1}\boldsymbol{\rho}_{2}}\tag{11.9}%
\label{Law of Cosines T}\\
&  =\left\Vert \boldsymbol{\rho}_{2}\right\Vert _{\min_{c}}^{2}-\left\Vert
\boldsymbol{\rho}_{2}\right\Vert \left\Vert \boldsymbol{\rho}_{1}\right\Vert
\cos\theta_{\boldsymbol{\rho}_{2}\boldsymbol{\rho}_{1}}\text{,}\nonumber
\end{align}
where $\theta$ is the angle between $\boldsymbol{\rho}_{1}$ and
$\boldsymbol{\rho}_{2}$, so that the geometric locus of the novel principal
eigenaxis $\boldsymbol{\rho}=\boldsymbol{\rho}_{1}-\boldsymbol{\rho}_{2}$
represents an eigenaxis of symmetry that exhibits symmetrical dimensions and
densities, such that the magnitude and the direction of the novel principal
eigenaxis $\boldsymbol{\rho}$ $=\boldsymbol{\rho}_{1}-\boldsymbol{\rho}_{2}$
are both functions of differences between joint variabilities of extreme
vectors $k_{\mathbf{x}_{1_{i\ast}}}$ and $k_{\mathbf{x}_{2_{i\ast}}}$ that
belong to the two classes $\omega_{1}$ and $\omega_{2}$ of random vectors
$\mathbf{x\in}$ $%
\mathbb{R}
^{d}$, at which point the critical minimum eigenenergy $\left\Vert
\boldsymbol{\rho}_{1}\right\Vert _{\min_{c}}^{2}$ exhibited by side
$\boldsymbol{\rho}_{1}$ is symmetrically balanced with the critical minimum
eigenenergy $\left\Vert \boldsymbol{\rho}_{2}\right\Vert _{\min_{c}}^{2}$
exhibited by side $\boldsymbol{\rho}_{2}$%
\[
\left\Vert \boldsymbol{\rho}_{1}\right\Vert _{\min_{c}}^{2}=\left\Vert
\boldsymbol{\rho}_{2}\right\Vert _{\min_{c}}^{2}\text{,}%
\]
the length of side $\boldsymbol{\rho}_{1}$ equals the length of side
$\boldsymbol{\rho}_{2}$%
\[
\left\Vert \boldsymbol{\rho}_{1}\right\Vert =\left\Vert \boldsymbol{\rho}%
_{2}\right\Vert \text{,}%
\]
and counteracting and opposing forces and influences of the minimum risk
binary classification system%
\[
\left(  k_{\mathbf{s}}\boldsymbol{-}\frac{1}{l}\sum\nolimits_{i=1}%
^{l}k_{\mathbf{x}_{i\ast}}\right)  \left(  \boldsymbol{\rho}_{1}%
-\boldsymbol{\rho}_{2}\right)  +\frac{1}{l}\sum\nolimits_{i=1}^{l}%
y_{i}\overset{\omega_{1}}{\underset{\omega_{2}}{\gtrless}}0
\]
are symmetrically balanced with each other about the geometric center of the
locus of the novel principal eigenaxis $\boldsymbol{\rho}=\boldsymbol{\rho
}_{1}-\boldsymbol{\rho}_{2}$%
\begin{align*}
&  \left\Vert \boldsymbol{\rho}_{1}\right\Vert \left(  \sum\nolimits_{i=1}%
^{l_{1}}\operatorname{comp}_{\overrightarrow{\boldsymbol{\rho}_{1}}}\left(
\overrightarrow{\psi_{1_{i_{\ast}}}k_{\mathbf{x}_{1_{i\ast}}}}\right)
-\sum\nolimits_{i=1}^{l_{2}}\operatorname{comp}%
_{\overrightarrow{\boldsymbol{\rho}_{1}}}\left(  \overrightarrow{\psi
_{2_{i_{\ast}}}k_{\mathbf{x}_{2_{i\ast}}}}\right)  \right)  \\
&  =\left\Vert \boldsymbol{\rho}_{2}\right\Vert \left(  \sum\nolimits_{i=1}%
^{l_{2}}\operatorname{comp}_{\overrightarrow{\boldsymbol{\rho}_{2}}}\left(
\overrightarrow{\psi_{2_{i_{\ast}}}k_{\mathbf{x}_{2_{i\ast}}}}\right)
-\sum\nolimits_{i=1}^{l_{1}}\operatorname{comp}%
_{\overrightarrow{\boldsymbol{\rho}_{2}}}\left(  \overrightarrow{\psi
_{1_{i_{\ast}}}k_{\mathbf{x}_{1_{i\ast}}}}\right)  \right)  \text{,}%
\end{align*}
whereon the statistical fulcrum of the novel principal eigenaxis
$\boldsymbol{\rho}=\boldsymbol{\rho}_{1}-\boldsymbol{\rho}_{2}$ is located.

Thereby, counteracting and opposing components of critical minimum
eigenenergies related to likely locations of extreme points from class
$\omega_{1}$ and class $\omega_{2}$ that determine regions of counter risks
and risks of the system---along the dual locus of side $\boldsymbol{\rho}_{1}%
$---are symmetrically balanced with counteracting and opposing components of
critical minimum eigenenergies related to likely locations of extreme points
from class $\omega_{2}$ and class $\omega_{1}$ that determine regions of
counter risks and risks of the system---along the dual locus of side
$\boldsymbol{\rho}_{2}$;

$6$. The center of total allowed eigenenergy and expected risk of the minimum
risk binary classification system%
\[
\left(  k_{\mathbf{s}}\boldsymbol{-}\frac{1}{l}\sum\nolimits_{i=1}%
^{l}k_{\mathbf{x}_{i\ast}}\right)  \left(  \boldsymbol{\rho}_{1}%
-\boldsymbol{\rho}_{2}\right)  +\frac{1}{l}\sum\nolimits_{i=1}^{l}%
y_{i}\overset{\omega_{1}}{\underset{\omega_{2}}{\gtrless}}0
\]
is located at the geometric center of the locus of the novel principal
eigenaxis $\boldsymbol{\rho}=\boldsymbol{\rho}_{1}-\boldsymbol{\rho}_{2}$ of
the system, whereon the statistical fulcrum of the system is located;

Then the minimum risk binary classification system acts to jointly minimize
its eigenenergy and risk by locating a point of equilibrium, at which point
critical minimum eigenenergies exhibited by the system are symmetrically
concentrated in such a manner that the geometric locus of the novel principal
eigenaxis of the system is an eigenaxis of symmetry that exhibits symmetrical
dimensions and densities, so that the dual locus of the discriminant function
of the system is in statistical equilibrium at the geometric locus of the
decision boundary of the system, such that counteracting and opposing forces
and influences of the system are symmetrically balanced with each
other---about the geometric center of the locus of the novel principal
eigenaxis---whereon the statistical fulcrum of the system is located.

Thereby, the minimum risk binary classification system satisfies a state of
statistical equilibrium so that the total allowed eigenenergy and the expected
risk exhibited by the system are jointly minimized within the decision space
of the system, at which point the system exhibits the minimum probability of
classification error.
\end{theorem}

The general locus formula that resolves the direct problem of the binary
classification of random vectors---that is expressed by Theorem
\ref{Direct Problem of Binary Classification Theorem}---is readily generalized
to minimum risk multiclass classification systems.

\subsection{Minimum Risk Multiclass Classification Systems}

Corollary \ref{Multiclass Classification System Corollary} generalizes the
fundamental laws of binary classification expressed by Theorem
\ref{Direct Problem of Binary Classification Theorem} to minimum risk
multiclass classification systems.

\begin{corollary}
\label{Multiclass Classification System Corollary}Any given minimum risk
multiclass classification system that is subject to $M$ sources of random
vectors $\mathbf{x\in}$ $%
\mathbb{R}
^{d}$ is determined by $M$ ensembles of $M-1$ minimum risk binary
classification systems, such that each ensemble is determined by an
architecture wherein one class is compared with all of the other $M-1$
classes, so that every one of the $M-1$ minimum risk binary classification
systems in each and every one of the $M$ ensembles satisfies the geometrical
and statistical criteria expressed by Theorem
\ref{Direct Problem of Binary Classification Theorem}.

Thereby, the minimum risk multiclass classification system satisfies a state
of statistical equilibrium so that the total allowed eigenenergy and the
expected risk exhibited by the system are jointly minimized within the
decision space of the system, at which point the system exhibits the minimum
probability of classification error.
\end{corollary}

\begin{proof}
Corollary \ref{Multiclass Classification System Corollary} is proved by
Theorem \ref{Direct Problem of Binary Classification Theorem} and the
superposition principle
\citep{Lathi1998}%
---since any given minimum risk multiclass classification system is based on a
\textquotedblleft one versus all\textquotedblright\ architecture.
\end{proof}

\subsection{Fundamental Laws of Binary Classification}

Theorem \ref{Direct Problem of Binary Classification Theorem} expresses
fundamental laws of binary classification that discriminant functions of
minimum risk binary classification systems are subject to. These laws are
summarized below.

\subsubsection{The Law of Total Allowed Eigenenergy}

We have named the locus formula in (\ref{TIE1}) \textquotedblleft the law of
total allowed energy for minimum risk binary classification
systems.\textquotedblright\ The law of total allowed energy demonstrates that
the total allowed energy and the expected risk exhibited by any given minimum
risk binary classification system are jointly regulated by an equilibrium
requirement---on the dual locus of the discriminant function of the system at
the geometric locus of the decision boundary of the system---at which point
the dual locus of the discriminant function is an eigenaxis of symmetry that
spans the decision space of the system, so that the dual locus of the
discriminant function satisfies the geometric locus of the decision boundary
in terms of a critical minimum eigenenergy and a minimum expected risk.

\subsubsection{The Law of Statistical Equilibrium}

We have named the locus formula in (\ref{TIE2}) \textquotedblleft the law of
statistical equilibrium for minimum risk binary classification
systems.\textquotedblright\ The law of statistical equilibrium demonstrates
that any given minimum risk binary classification system acts to jointly
minimize its eigenenergy and risk by locating a point of equilibrium, at which
point the geometric locus of the novel principal eigenaxis of the system
represents an eigenaxis of symmetry that exhibits symmetrical dimensions and
densities, such  that critical minimum eigenenergies are minimized throughout
the decision space of the system, so that regions of counter risks and risks
of the system are symmetrically balanced with each other, at which point the
total allowed eigenenergy and the expected risk exhibited by the system are
jointly minimized within the decision space of the system---so that the system
exhibits the minimum probability of classification error.

\subsubsection{The Law of Symmetry}

We have named the locus formula in (\ref{Law of Cosines T}) \textquotedblleft
the law of symmetry for minimum risk binary classification
systems.\textquotedblright\ The law of symmetry demonstrates that the cost of
finding any given minimum risk binary classification system under uncertainty
is the critical minimum eigenenergy that is necessary for the system to
achieve a state of statistical equilibrium, at which point critical minimum
eigenenergies exhibited by the system are symmetrically concentrated in such a
manner that the geometric locus of the novel principal eigenaxis of the system
represents an eigenaxis of symmetry that exhibits symmetrical dimensions and
densities, so that counteracting and opposing forces and influences of the
system are symmetrically balanced with each other---about the geometric center
of the locus of the novel principal eigenaxis---whereon the statistical
fulcrum of the system is located.

In the next part of our treatise, we will prove the fundamental laws of binary
classification expressed by Theorem
\ref{Direct Problem of Binary Classification Theorem} by means of a
constructive proof that demonstrates how a certain constrained optimization
algorithm executes each and every one of the fundamental laws.

We now turn our attention to the constrained optimization algorithm that
resolves the inverse problem of the binary classification of random vectors.

\section{\label{Section 12}Finding a Novel Principal Eigenaxis}

In this part of our treatise, we use a well-posed constrained optimization
algorithm and a collection of labeled feature vectors to produce an example of
a minimum risk binary classification system that satisfies the geometrical and
statistical criteria expressed by Theorem
\ref{Direct Problem of Binary Classification Theorem}. We define a class of
discriminant functions of minimum risk binary classification systems---by an
objective function of an inequality constrained optimization problem---so that
a discriminant function of a minimum risk binary classification system is
represented by a geometric locus of a novel principal eigenaxis, such that the
constrained objective function of the geometric locus of the novel principal
eigenaxis is subject to the geometrical and statistical criteria expressed by
Theorem \ref{Direct Problem of Binary Classification Theorem}.

\subsection{Objective Function of a Novel Principal Eigenaxis}

Take any given collection of labeled feature vectors%
\[
\left(  \mathbf{x}_{1}\mathbf{,}y_{1}\right)  ,\ldots,\left(  \mathbf{x}%
_{N}\mathbf{,}y_{N}\right)  \in%
\mathbb{R}
^{d}\times Y,Y=\left\{  \pm1\right\}  \text{,}%
\]
where $N$ feature vectors $\mathbf{x}\in%
\mathbb{R}
^{d}$ are generated by certain probability density functions $p\left(
\mathbf{x};\omega_{1}\right)  $ and $p\left(  \mathbf{x};\omega_{2}\right)  $
that determine either overlapping distributions, such that $\bigcap
\nolimits_{i=1}^{2}\omega_{i}\neq\emptyset$ and $\bigcap\nolimits_{i=1}%
^{2}\mathbf{x}\omega_{i}\neq\emptyset$, or non-overlapping distributions, such
that $\bigcap\nolimits_{i=1}^{2}\omega_{i}=\emptyset$ and $\bigcap
\nolimits_{i=1}^{2}\mathbf{x}\omega_{i}=0$, of $d$-dimensional numerical
features $\mathbf{x\in}$ $%
\mathbb{R}
^{d}$.

We produce an example of a minimum risk binary classification system%
\[
\left(  k_{\mathbf{s}}\boldsymbol{-}\frac{1}{l}\sum\nolimits_{i=1}%
^{l}k_{\mathbf{x}_{i\ast}}\right)  \left(  \boldsymbol{\rho}_{1}%
-\boldsymbol{\rho}_{2}\right)  +\frac{1}{l}\sum\nolimits_{i=1}^{l}%
y_{i}\overset{\omega_{1}}{\underset{\omega_{2}}{\gtrless}}0
\]
that satisfies the geometrical and statistical criteria expressed by Theorem
\ref{Direct Problem of Binary Classification Theorem}, so that the
discriminant function of the system is represented by a geometric locus of a
novel principal eigenaxis%
\begin{align*}
\boldsymbol{\rho}  &  =\sum\nolimits_{i=1}^{l_{1}}\psi_{1_{i_{\ast}}%
}k_{\mathbf{x}_{1_{i\ast}}}-\sum\nolimits_{i=1}^{l_{2}}\psi_{2_{i_{\ast}}%
}k_{\mathbf{x}_{2_{i\ast}}}\\
&  =\boldsymbol{\rho}_{1}-\boldsymbol{\rho}_{2}%
\end{align*}
that satisfies the conditions of Theorem
\ref{Direct Problem of Binary Classification Theorem}, by using the collection
of labeled feature vectors and a well-posed constrained optimization algorithm
to determine the solution for the inequality constrained optimization
problem---known as the primal problem%
\begin{align}
\min\Psi\left(  \boldsymbol{\kappa}\right)   &  =\left\Vert \boldsymbol{\kappa
}\right\Vert ^{2}/2+C/2\sum\nolimits_{i=1}^{N}\xi_{i}^{2}\text{,}%
\tag{12.1}\label{Objective Function}\\
\text{s.t. }y_{i}\left(  k_{\mathbf{x}_{i}}\boldsymbol{\kappa}%
+\boldsymbol{\kappa}_{0}\right)   &  \geq1-\xi_{i},\ \ i=1,...,N\text{,}%
\nonumber
\end{align}
where $\boldsymbol{\kappa}\triangleq\boldsymbol{\kappa}_{1}{\large -}%
\boldsymbol{\kappa}_{2}$ is a geometric locus of a novel principal eigenaxis,
$k_{\mathbf{x}_{i}}$ is a reproducing kernel for the feature vector
$\mathbf{x}_{i}$, where the reproducing kernel $k_{\mathbf{x}}\left(
\mathbf{s}\right)  $ is either a Gaussian reproducing kernel $k_{\mathbf{x}%
}\left(  \mathbf{s}\right)  =\exp\left(  -\gamma\left\Vert \mathbf{s}%
-\mathbf{x}\right\Vert ^{2}\right)  :0.01\leq\gamma\leq0.1$, or a second-order
polynomial reproducing kernel $k_{\mathbf{x}}\left(  \mathbf{s}\right)
=\left(  \mathbf{s}^{T}\mathbf{x}+1\right)  ^{2}$, $\left\Vert
\boldsymbol{\kappa}\right\Vert ^{2}$ is the eigenenergy exhibited by the
geometric locus of the novel principal eigenaxis $\boldsymbol{\kappa}$,
$\boldsymbol{\kappa}_{0}$ is a functional of $\boldsymbol{\kappa}$, $C$ and
$\xi_{i}$ are regularization parameters for a joint covariance matrix, and
$y_{i}$ are class membership statistics, where $y_{i}=+1$ if $\mathbf{x}%
_{i}\in\omega_{1}$, and $y_{i}=-1$ if $\mathbf{x}_{i}\in\omega_{2}$.

\subsection{Objective of the Constrained Optimization Algorithm}

The objective of the constrained optimization algorithm that solves the primal
optimization problem in (\ref{Objective Function}) is to \emph{find} the
geometric locus of the novel principal eigenaxis $\boldsymbol{\kappa}$ that
minimizes the total allowed eigenenergy $\left\Vert \boldsymbol{\kappa
}\right\Vert _{\min_{c}}^{2}$ and the expected risk $\mathfrak{R}%
_{\mathfrak{\min}}\left(  \left\Vert \boldsymbol{\kappa}\right\Vert _{\min
_{c}}^{2}\right)  $ exhibited by the minimum risk binary classification system
$k_{\mathbf{s}}\boldsymbol{\kappa}+$ $\boldsymbol{\kappa}_{0}\overset{\omega
_{1}}{\underset{\omega_{2}}{\gtrless}}0$---within the decision space
$Z=Z_{1}\cup Z_{2}$ of the system---at which point the geometric locus of the
novel principal eigenaxis $\boldsymbol{\kappa}$ is subject to a critical
minimum eigenenergy constraint%
\[
\gamma\left(  \boldsymbol{\kappa}\right)  =\left\Vert \boldsymbol{\kappa
}\right\Vert _{\min_{c}}^{2}\text{,}%
\]
so that the system of $N$ inequalities%
\[
y_{i}\left(  k_{\mathbf{x}_{i}}\boldsymbol{\kappa}+\boldsymbol{\kappa}%
_{0}\right)  \geq1-\xi_{i},\ \ i=1,...,N
\]
is satisfied in the most suitable manner.

\subsection{The Primal Eigenenergy Functional}

The solution for the primal optimization problem in (\ref{Objective Function})
is found by using Lagrange multipliers $\psi_{i}\geq0$ and the \emph{primal
eigenenergy functional}%
\begin{align}
\Xi_{\boldsymbol{\kappa}}\left(  \boldsymbol{\kappa}\mathbf{,}%
\boldsymbol{\kappa}_{0},\psi_{i}\right)   &  =\left\Vert \boldsymbol{\kappa
}\right\Vert ^{2}/2+C/2\sum\nolimits_{i=1}^{N}\xi_{i}^{2} \tag{12.2}%
\label{Eigenenergy Functional}\\
&  -\sum\nolimits_{i=1}^{N}\psi_{i}\left\{  y_{i}\left(  k_{\mathbf{x}_{i}%
}\boldsymbol{\kappa}+\boldsymbol{\kappa}_{0}\right)  -1+\xi_{i}\right\}
\nonumber
\end{align}
of a minimum risk binary classification system $k_{\mathbf{s}}%
\boldsymbol{\kappa}+$ $\boldsymbol{\kappa}_{0}\overset{\omega_{1}%
}{\underset{\omega_{2}}{\gtrless}}0$, so that the objective function and its
constraints in (\ref{Objective Function}) are combined with each other, at
which point the primal eigenenergy functional in (\ref{Eigenenergy Functional}%
) is minimized with respect to the primal variables $\boldsymbol{\kappa}$ and
$\boldsymbol{\kappa}_{0}$ and is maximized with respect to the dual variables
$\psi_{i}$.

\subsection{The Wolfe-dual Principal Eigenspace}

The constrained optimization algorithm that resolves the inverse problem---of
the binary classification of random vectors---introduces a \emph{dual
eigenenergy functional} of a minimum risk binary classification system
$k_{\mathbf{s}}\boldsymbol{\kappa}+$ $\boldsymbol{\kappa}_{0}\overset{\omega
_{1}}{\underset{\omega_{2}}{\gtrless}}0$ inside a vector space that we have
named the \textquotedblleft Wolfe-dual principal eigenspace,\textquotedblright%
\ so that the Wolfe-dual novel principal eigenaxis $\boldsymbol{\psi}$ of the
system $k_{\mathbf{s}}\boldsymbol{\kappa}+$ $\boldsymbol{\kappa}%
_{0}\overset{\omega_{1}}{\underset{\omega_{2}}{\gtrless}}0$ is symmetrically
and equivalently\emph{\ }related to the primal novel principal eigenaxis
$\boldsymbol{\kappa}$ of the system $k_{\mathbf{s}}\boldsymbol{\kappa}+$
$\boldsymbol{\kappa}_{0}\overset{\omega_{1}}{\underset{\omega_{2}}{\gtrless}%
}0$, and finds \emph{extrema} for the \emph{restriction} of the geometric
locus of the novel principal eigenaxis $\boldsymbol{\kappa}$ to the Wolfe-dual
principal \emph{eigenspace}.

We determine the dual eigenenergy functional and the extrema by evaluating the
Karush-Kuhn-Tucker (KKT) conditions on the primal eigenenergy functional in
(\ref{Eigenenergy Functional}).

\subsection{The Karush-Kuhn-Tucker Conditions}

We use the Karush-Kuhn-Tucker theorem
\citep{Sundaram1996}
and the KKT conditions
\citep{Cristianini2000,Scholkopf2002}
on the primal eigenenergy functional in (\ref{Eigenenergy Functional}) to
determine geometrical and statistical conditions that the constrained
objective function of the geometric locus of the novel principal eigenaxis
$\boldsymbol{\kappa}$ \ is subject to---at which point the minimum risk binary
classification system $k_{\mathbf{s}}\boldsymbol{\kappa}+$ $\boldsymbol{\kappa
}_{0}\overset{\omega_{1}}{\underset{\omega_{2}}{\gtrless}}0$ satisfies the
geometrical and statistical criteria expressed by Theorem
\ref{Direct Problem of Binary Classification Theorem}.

Accordingly, we use the KKT conditions%
\begin{equation}
\boldsymbol{\kappa}-\sum\nolimits_{i=1}^{N}\psi_{i}y_{i}k_{\mathbf{x}_{i}%
}=0,\text{\ }i=1,...N\text{,} \tag{12.3}\label{KKT 1}%
\end{equation}%
\begin{equation}
\sum\nolimits_{i=1}^{N}\psi_{i}y_{i}=0,\text{ \ }i=1,...,N\text{,}
\tag{12.4}\label{KKT 2}%
\end{equation}%
\begin{equation}
C\sum\nolimits_{i=1}^{N}\xi_{i}-\sum\nolimits_{i=1}^{N}\psi_{i}=0\text{,}
\tag{12.5}\label{KKT 3}%
\end{equation}%
\begin{equation}
\psi_{i}\geq0,\text{ \ }i=1,...,N\text{,} \tag{12.6}\label{KKT 4}%
\end{equation}%
\begin{equation}
\psi_{i}\left[  y_{i}\left(  k_{\mathbf{x}_{i}}\boldsymbol{\kappa
}+\boldsymbol{\kappa}_{0}\right)  -1+\xi_{i}\right]  \geq0,\ i=1,...,N\text{,}
\tag{12.7}\label{KKT 5}%
\end{equation}
and the Karush-Kuhn-Tucker theorem to demonstrate that the primal novel
principal eigenaxis $\boldsymbol{\kappa}$ and the Wolfe-dual novel principal
eigenaxis $\boldsymbol{\psi}$ jointly satisfy an equivalent system of the
fundamental locus equations of binary classification---expressed by Theorem
\ref{Direct Problem of Binary Classification Theorem}.

Correspondingly, we demonstrate that the KKT conditions in (\ref{KKT 1}) -
(\ref{KKT 5}) ensure that the discriminant function $d\left(  \mathbf{s}%
\right)  =k_{\mathbf{s}}\boldsymbol{\kappa}+$ $\boldsymbol{\kappa}_{0}$ is the
solution of an equivalent system of the fundamental locus equations of binary
classification expressed by Theorem
\ref{Direct Problem of Binary Classification Theorem}, so that the
discriminant function satisfies the fundamental\emph{\ }statistical
laws---expressed by Theorem
\ref{Direct Problem of Binary Classification Theorem}---that the minimum risk
binary classification system $k_{\mathbf{s}}\boldsymbol{\kappa}+$
$\boldsymbol{\kappa}_{0}\overset{\omega_{1}}{\underset{\omega_{2}}{\gtrless}%
}0$ is subject to.

\subsection{The Fundamental Unknowns}

The fundamental unknowns associated with the primal optimization problem in
(\ref{Objective Function}) are the scale factors $\psi_{i}$ for $N$ scaled
unit feature vectors $\frac{k_{\mathbf{x}_{i}}}{\left\Vert k_{\mathbf{x}_{i}%
}\right\Vert }:$ $\left\{  \psi_{i}\frac{k_{\mathbf{x}_{i}}}{\left\Vert
k_{\mathbf{x}_{i}}\right\Vert }\right\}  _{i=1}^{N}$ that determine the
structure of the geometric locus of the Wolfe-dual novel principal eigenaxis
$\boldsymbol{\psi}$, so that the magnitude of each Wolfe-dual principal
eigenaxis component $\psi_{i}\frac{k_{\mathbf{x}_{i}}}{\left\Vert
k_{\mathbf{x}_{i}}\right\Vert }$ that lies on $\boldsymbol{\psi}$ has a
certain positive value $\psi_{i}>0$, and the direction of the principal
eigenaxis component is the direction of the feature vector $k_{\mathbf{x}_{i}%
}$.

It will be seen that the value of each active scale factor $\psi_{i}>0$
determines a likelihood value and a likely location---both of which are
normalized relative to length---for a correlated extreme point that is located
within either an overlapping region or near a tail region of distributions of
the $N$ feature vectors $\mathbf{x}$ in the collection of training data
$\left(  \mathbf{x}_{1}\mathbf{,}y_{1}\right)  ,\ldots,\left(  \mathbf{x}%
_{N}\mathbf{,}y_{N}\right)  \in%
\mathbb{R}
^{d}\times Y,Y=\left\{  \pm1\right\}  $.

It will also be seen that the value of each active scale factor $\psi_{i}>0$
determines the magnitude of a scaled extreme vector that lies on the geometric
locus of the primal novel principal eigenaxis $\boldsymbol{\kappa} $ ---as
well as a likelihood value and a likely location for the extreme vector.

Since the primal optimization problem in (\ref{Objective Function}) is a
convex optimization problem, it follows that the inequalities in (\ref{KKT 4})
and (\ref{KKT 5}) must only hold for certain values of the primal variables
$\boldsymbol{\kappa}$ and $\boldsymbol{\kappa}_{0}$ and the Wolfe-dual
variables $\psi_{i}$
\citep{Sundaram1996}%
. We demonstrate that the KKT conditions in (\ref{KKT 1}) - (\ref{KKT 5})
restrict the magnitudes and the eigenenergies exhibited by all of the
principal eigenaxis components on both $\boldsymbol{\psi}$ and
$\boldsymbol{\kappa}$ in such a manner that the minimum risk binary
classification system $k_{\mathbf{s}}\boldsymbol{\kappa}+$ $\boldsymbol{\kappa
}_{0}\overset{\omega_{1}}{\underset{\omega_{2}}{\gtrless}}0$ satisfies a state
of statistical equilibrium---so that the total allowed eigenenergy $\left\Vert
Z|\boldsymbol{\kappa}\right\Vert _{\min_{c}}^{2}$ and the expected risk
$\mathfrak{R}_{\mathfrak{\min}}\left(  Z|\left\Vert \boldsymbol{\kappa
}\right\Vert _{\min_{c}}^{2}\right)  $ exhibited by the system are jointly
minimized within the decision space $Z=Z_{1}\cup Z_{2}$ of the system.

\subsection{The Wolfe-dual Eigenenergy Functional}

Substituting the expressions for $\boldsymbol{\kappa}$ and $\psi_{i}$ in
(\ref{KKT 1}) and (\ref{KKT 2}) into the primal eigenenergy functional in
(\ref{Eigenenergy Functional}) and simplifying the resulting expression
determines the dual eigenenergy functional of a minimum risk binary
classification system $k_{\mathbf{s}}\boldsymbol{\kappa}+$ $\boldsymbol{\kappa
}_{0}\overset{\omega_{1}}{\underset{\omega_{2}}{\gtrless}}0$, also known as
the Wolfe dual problem%
\begin{equation}
\max\Xi_{\psi_{i}}\left(  \psi_{i}\right)  =\sum\nolimits_{i=1}^{N}\psi
_{i}-\sum\nolimits_{i,j=1}^{N}\psi_{i}\psi_{j}y_{i}y_{j}\frac{k_{\mathbf{x}%
_{i}}k_{\mathbf{x}_{j}}+\delta_{ij}/C}{2}\text{,} \tag{12.8}%
\label{Lagrangian Dual Prob}%
\end{equation}
where $\delta_{ij}$ is the Kronecker $\delta$ defined as unity for $i=j$ and
$0$ otherwise, at which point the scale factors $\psi_{i}$ for the $N$ scaled
unit feature vectors $\left\{  \psi_{i}\frac{k_{\mathbf{x}_{i}}}{\left\Vert
k_{\mathbf{x}_{i}}\right\Vert }\right\}  _{i=1}^{N}$ that determine the
structure of the geometric locus of the Wolfe-dual novel principal eigenaxis
$\boldsymbol{\psi}$ are subject to the constraints $\psi_{i}\geq0$ and
$\sum\nolimits_{i=1}^{N}y_{i}\psi_{i}=0$, where $y_{i}=\left\{  \pm1\right\}
$.

The dual eigenenergy functional in (\ref{Lagrangian Dual Prob}) can be written
in vector notation by letting $\mathbf{Q}\triangleq\epsilon\mathbf{I}%
+\widetilde{\mathbf{X}}\widetilde{\mathbf{X}}^{T}$, where $\epsilon\ll1$,
$\widetilde{\mathbf{X}}\triangleq\mathbf{D}_{y}\mathbf{X}$, $\mathbf{D}_{y}$
is an $N\times N$ diagonal matrix of class membership statistics $y_{i}$, and
the $N\times d$ matrix $\widetilde{\mathbf{X}}$ is a matrix of $N$ labeled
reproducing kernels for $N$ feature vectors $\mathbf{x\in}$ $%
\mathbb{R}
^{d}$%
\[
\widetilde{\mathbf{X}}=%
\begin{pmatrix}
y_{1}k_{\mathbf{x}_{1}}, & y_{2}k_{\mathbf{x}_{2}}, & \ldots, & y_{N}%
k_{\mathbf{x}_{N}}%
\end{pmatrix}
^{T}\text{.}%
\]
\qquad

Call the dual eigenenergy functional in (\ref{Lagrangian Dual Prob})
\textquotedblleft the Wolfe-dual eigenenergy functional of a minimum risk
binary classification system.\textquotedblright\ Accordingly, the matrix
version of the Wolfe-dual eigenenergy functional of a minimum risk binary
classification system $k_{\mathbf{s}}\boldsymbol{\kappa}+$ $\boldsymbol{\kappa
}_{0}\overset{\omega_{1}}{\underset{\omega_{2}}{\gtrless}}0$ is written as%
\begin{equation}
\max\Xi_{\boldsymbol{\psi}}\left(  \boldsymbol{\psi}\right)  =\mathbf{1}%
^{T}\boldsymbol{\psi}-\frac{\boldsymbol{\psi}^{T}\mathbf{Q}\boldsymbol{\psi}%
}{2}\text{,} \tag{12.9}\label{Matrix Version Wolfe Dual Problem}%
\end{equation}
at which point the structure and behavior and properties of the Wolfe-dual
novel principal eigenaxis $\boldsymbol{\psi}$ are symmetrically and
equivalently related to the structure and behavior and properties of the
primal novel principal eigenaxis $\boldsymbol{\kappa}$, such that the
geometric locus of the Wolfe-dual novel principal eigenaxis $\boldsymbol{\psi
}$ is subject to the constraints $\boldsymbol{\psi}^{T}\mathbf{y}=0$ and
$\psi_{i}\geq0$, where $y_{i}\in Y=\left\{  \pm1\right\}  $, such that the
inequalities $\psi_{i}>0$ only hold for certain values of $\psi_{i}$.

\subsection{Estimate of a Joint Covariance Matrix}

The symmetric matrix $\mathbf{Q}\triangleq\epsilon\mathbf{I}%
+\widetilde{\mathbf{X}}\widetilde{\mathbf{X}}^{T}$ of the random quadratic
form $\boldsymbol{\psi}^{T}\mathbf{Q}\boldsymbol{\psi}$ in the Wolfe-dual
eigenenergy functional of (\ref{Matrix Version Wolfe Dual Problem}) provides
an estimate of a joint covariance matrix, such that each element
$y_{i}\left\Vert k_{\mathbf{x}_{i}}\right\Vert y_{j}\left\Vert k_{\mathbf{x}%
_{j}}\right\Vert \cos\theta_{k_{\mathbf{x}_{i}}k_{\mathbf{x}_{j}}}$ of the
matrix $\mathbf{Q}$ provides an estimate of joint variabilities between
coordinates $\left\{  \left\Vert k_{\mathbf{x}_{i}}\right\Vert \cos
\mathbb{\alpha}_{\mathbf{e}_{i}k_{\mathbf{x}_{i}}}\right\}  _{i=1}^{d}$ and
$\left\{  \left\Vert k_{\mathbf{x}_{j}}\right\Vert \cos\mathbb{\alpha
}_{\mathbf{e}_{i}k_{\mathbf{x}_{j}}}\right\}  _{i=1}^{d}$ of certain feature
vectors $k_{\mathbf{x}_{i}}$ and $k_{\mathbf{x}_{j}}$, so that each element
$y_{i}y_{j}\left\Vert k_{\mathbf{x}_{i}}\right\Vert \left\Vert k_{\mathbf{x}%
_{j}}\right\Vert \cos\theta_{k_{\mathbf{x}_{i}}k_{\mathbf{x}_{j}}}$ of the
joint covariance matrix $\mathbf{Q}$ where $y_{i}y_{j}=-1$ describes
differences between joint variabilities of feature vectors $k_{\mathbf{x}_{i}%
}$ and $k_{\mathbf{x}_{j}}$ that belong to different pattern classes, at which
point each element $\left\Vert k_{\mathbf{x}_{i}}\right\Vert \left\Vert
k_{\mathbf{x}_{j}}\right\Vert \cos\theta_{k_{\mathbf{x}_{i}}k_{\mathbf{x}_{j}%
}}$ of the joint covariance matrix $\mathbf{Q}$ is correlated with the
distance $\left\Vert k_{\mathbf{x}_{i}}-k_{\mathbf{x}_{j}}\right\Vert $
between the loci of certain feature vectors $k_{\mathbf{x}_{i}}$ and
$k_{\mathbf{x}_{j}}$.

Recall that Theorem \ref{Principal Eigen-coordinate System Theorem} is an
existence theorem that guarantees the existence of an exclusive principal
eigen-coordinate system---which is the principal part of an equivalent
representation of a correlated quadratic form---such that the exclusive
principal eigen-coordinate system is the solution of an equivalent form of the
vector algebra locus equation of the geometric locus of a certain quadratic
curve or surface---at which point the principal eigenaxis of the geometric
locus of the quadratic curve or surface satisfies the geometric locus of the
quadratic curve or surface in terms of its \emph{total allowed eigenenergy}.

Moreover, the novel principal eigen-coordinate transform method expressed by
Theorem \ref{Principal Eigen-coordinate System Theorem} reveals that the shape
and the fundamental property exhibited by the geometric locus of the quadratic
curve or surface are both determined by the exclusive principal
eigen-coordinate system, such that the eigenvalues of the symmetric matrix of
the correlated quadratic form modulate the eigenenergies exhibited by the
components of the exclusive principal eigen-coordinate system---which is the
principal eigenaxis of the geometric locus of the quadratic curve or surface,
so that the principal eigenaxis satisfies the geometric locus of the quadratic
curve or surface in terms of its total allowed eigenenergy; and the uniform
property exhibited by all of the points that lie on the geometric locus of the
quadratic curve or surface is the total allowed eigenenergy exhibited by the
principal eigenaxis of the geometric locus of the quadratic curve or surface.

The objective of the Wolfe-dual eigenenergy functional in
(\ref{Matrix Version Wolfe Dual Problem}) is \emph{led} by the
\emph{guarantees} provided by Theorem
\ref{Principal Eigen-coordinate System Theorem}.

\subsection{Objective of the Wolfe-dual Eigenenergy Functional}

We realize that the Wolfe-dual eigenenergy functional in
(\ref{Matrix Version Wolfe Dual Problem}) constitutes a convex programming
problem, so that $\boldsymbol{\psi}$ and $\boldsymbol{\kappa}$ are both
subject to certain geometrical and statistical conditions---expressed by the
Karush-Kuhn-Tucker conditions in (\ref{KKT 1}) - (\ref{KKT 5})---inside the
Wolfe-dual principal eigenspace of $\boldsymbol{\psi}$ and $\boldsymbol{\kappa
}$. Moreover, the strong duality theorem
\citep{Fletcher2000,Luenberger1969,Luenberger2003,Nash1996}
provides us with a guarantee that the structure and behavior and properties
exhibited by the geometric locus of the Wolfe-dual novel principal eigenaxis
$\boldsymbol{\psi}$ are symmetrically and equivalently related to the
structure and behavior and properties exhibited by the geometric locus of the
primal novel principal eigenaxis $\boldsymbol{\kappa}$.

Accordingly, let $\psi_{i\ast}>0$ be an active scale factor associated with
the Wolfe-dual eigenenergy functional%
\[
\max\Xi_{\boldsymbol{\psi}}\left(  \boldsymbol{\psi}\right)  =\mathbf{1}%
^{T}\boldsymbol{\psi}-\frac{\boldsymbol{\psi}^{T}\mathbf{Q}\boldsymbol{\psi}%
}{2}\text{,}%
\]
at which point the geometric locus of the Wolfe-dual novel principal eigenaxis
$\boldsymbol{\psi}$ is subject to the constraints $\boldsymbol{\psi}%
^{T}\mathbf{y}=0$ and $\psi_{i\ast}>0$, and let $k_{\mathbf{x}_{i\ast}}$ be an
extreme vector that is used to construct the joint covariance matrix
$\mathbf{Q}$ of the random quadratic form $\boldsymbol{\psi}^{T}%
\mathbf{Q}\boldsymbol{\psi}$.

Also, let the geometric locus of the Wolfe-dual novel principal eigenaxis
$\boldsymbol{\psi}$ be symmetrically and equivalently related to the principal
eigenvector $\boldsymbol{\psi}_{\max}$ of the joint covariance matrix
$\mathbf{Q}$ and the inverted joint covariance matrix $\mathbf{Q}^{-1}$
associated with the pair of random quadratic forms $\boldsymbol{\psi}%
^{T}\mathbf{Q}\boldsymbol{\psi}$ and $\boldsymbol{\psi}^{T}\mathbf{Q}%
^{-1}\boldsymbol{\psi}$, so that the geometrical and statistical structure and
the statistical behavior and properties exhibited by both $\boldsymbol{\psi}$
and $\boldsymbol{\psi}_{\max}$ are determined by the values of the active
scale factors $\psi_{i\ast}>0$ for normalized extreme vectors $\frac
{k_{\mathbf{x}_{i\ast}}}{\left\Vert k_{\mathbf{x}_{i\ast}}\right\Vert }$.

Finally, let the Wolfe-dual novel principal eigenaxis $\boldsymbol{\psi}$ and
the primal novel principal eigenaxis $\boldsymbol{\kappa}$ \ be subject to the
Karush-Kuhn-Tucker conditions in (\ref{KKT 1}) - (\ref{KKT 5})---inside the
Wolfe-dual principal eigenspace of $\boldsymbol{\psi}$ and $\boldsymbol{\kappa
}$, so that the structure and behavior and properties exhibited by the
Wolfe-dual novel principal eigenaxis $\boldsymbol{\psi}$ are symmetrically and
equivalently related to the structure and behavior and properties exhibited by
the primal novel principal eigenaxis $\boldsymbol{\kappa}$.

Using the above notation and assumptions, along with the guarantees provided
by the strong duality theorem and the novel principal eigen-coordinate
transform method expressed by Theorem
\ref{Principal Eigen-coordinate System Theorem} and Corollaries
\ref{Principal Eigen-coordinate System Corollary} -
\ref{Symmetrical and Equivalent Principal Eigenaxes}, it follows that the
objective of the Wolfe-dual eigenenergy functional is to \emph{find} the
\emph{principal} \emph{eigenvector} $\boldsymbol{\psi}_{\max}$ of the joint
covariance matrix $\mathbf{Q}$ and the inverted joint covariance matrix
$\mathbf{Q}^{-1}$ associated with the pair of random quadratic forms
$\boldsymbol{\psi}^{T}\mathbf{Q}\boldsymbol{\psi}$ and $\boldsymbol{\psi}%
^{T}\mathbf{Q}^{-1}\boldsymbol{\psi}$, where the joint covariance matrix
$\mathbf{Q}$ is associated with the random quadratic form $\boldsymbol{\psi
}^{T}\mathbf{Q}\boldsymbol{\psi}$ in the Wolfe-dual eigenenergy functional
$\max\Xi_{\boldsymbol{\psi}}\left(  \boldsymbol{\psi}\right)  =\mathbf{1}%
^{T}\boldsymbol{\psi}-\frac{\boldsymbol{\psi}^{T}\mathbf{Q}\boldsymbol{\psi}%
}{2}$ of (\ref{Matrix Version Wolfe Dual Problem}), wherein $\boldsymbol{\psi
}^{T}\mathbf{y}=0$ and $\psi_{i\ast}>0$, and $y_{i}\in Y=\left\{
\pm1\right\}  $, so that the structure and behavior and properties of the
principal eigenvector $\boldsymbol{\psi}_{\max}$ of $\mathbf{Q}$ and
$\mathbf{Q}^{-1}$ are symmetrically and equivalently related to the structure
and behavior and properties of the primal novel principal eigenaxis
$\boldsymbol{\kappa}$ in such a manner that the geometric locus of the novel
principal eigenaxis $\boldsymbol{\kappa}$ is the principal part of an
equivalent representation of the pair of random quadratic forms
$\boldsymbol{\psi}^{T}\mathbf{Q}\boldsymbol{\psi}$ and $\boldsymbol{\psi}%
^{T}\mathbf{Q}^{-1}\boldsymbol{\psi}$, such that the geometric locus of the
novel principal eigenaxis $\boldsymbol{\kappa}$ is the principal eigenaxis of
the geometric locus of the decision boundary of a minimum risk binary
classification system, at which point the geometric locus of the novel
principal eigenaxis $\boldsymbol{\kappa}$ satisfies the geometric locus of the
decision boundary of the system in terms of a critical minimum eigenenergy
$\left\Vert \boldsymbol{\kappa}\right\Vert _{\min_{c}}^{2}$ and a minimum
expected risk $\mathfrak{R}_{\mathfrak{\min}}\left(  \left\Vert
\boldsymbol{\kappa}\right\Vert _{\min_{c}}^{2}\right)  $, such that the
uniform properties exhibited by all of the points that lie on the geometric
locus of the decision boundary are the critical minimum eigenenergy
$\left\Vert \boldsymbol{\kappa}\right\Vert _{\min_{c}}^{2}$ and the minimum
expected risk $\mathfrak{R}_{\mathfrak{\min}}\left(  \left\Vert
\boldsymbol{\kappa}\right\Vert _{\min_{c}}^{2}\right)  $ exhibited by the
geometric locus of the novel principal eigenaxis $\boldsymbol{\kappa}$.

It will be seen that the machine learning algorithm being examined
\emph{transforms} the random quadratic form $\boldsymbol{\psi}^{T}%
\mathbf{Q}\boldsymbol{\psi}$ in the Wolfe-dual eigenenergy functional $\max
\Xi\left(  \boldsymbol{\psi}\right)  =\mathbf{1}^{T}\boldsymbol{\psi
}-\boldsymbol{\psi}^{T}\mathbf{Q}\boldsymbol{\psi/}2$ of a minimum risk binary
classification system $k_{\mathbf{s}}\boldsymbol{\kappa}+\mathbf{\kappa}%
_{0}\overset{\omega_{1}}{\underset{\omega_{2}}{\gtrless}}0$---that is subject
to the constraints $\boldsymbol{\psi}^{T}\mathbf{y}=0$ and $\psi_{i\ast}%
>0$---\emph{into} a geometric locus of a novel principal eigenaxis
$\boldsymbol{\kappa}=\boldsymbol{\kappa}_{1}-\boldsymbol{\kappa}_{2}$ of the
system, so that the geometric locus of the novel principal eigenaxis
$\boldsymbol{\kappa}=\boldsymbol{\kappa}_{1}-\boldsymbol{\kappa}_{2}$
satisfies the geometric locus of the decision boundary $k_{\mathbf{s}%
}\boldsymbol{\kappa}+$ $\boldsymbol{\kappa}_{0}=0$ of the system in terms of a
critical minimum eigenenergy $\left\Vert \boldsymbol{\kappa}\right\Vert
_{\min_{c}}^{2}$ and a minimum expected risk $\mathfrak{R}_{\mathfrak{\min}%
}\left(  \left\Vert \boldsymbol{\kappa}\right\Vert _{\min_{c}}^{2}\right)  $,
such that the total allowed eigenenergy $\left\Vert \boldsymbol{\kappa
}\right\Vert _{\min_{c}}^{2}$ and the expected risk $\mathfrak{R}%
_{\mathfrak{\min}}\left(  \left\Vert \boldsymbol{\kappa}\right\Vert _{\min
_{c}}^{2}\right)  $ exhibited by the novel principal eigenaxis
$\boldsymbol{\kappa}=\boldsymbol{\kappa}_{1}-\boldsymbol{\kappa}_{2}$ are both
regulated by the \emph{total} \emph{value} of the scale factors $\psi_{1i\ast
}$ and $\psi_{2i\ast}$---for the components $\psi_{1i\ast}\frac{k_{\mathbf{x}%
_{1i\ast}}}{\left\Vert k_{\mathbf{x}_{1i\ast}}\right\Vert }$ and $\psi
_{2i\ast}\frac{k_{\mathbf{x}_{2i\ast}}}{\left\Vert k_{\mathbf{x}_{2i\ast}%
}\right\Vert }$ of the principal eigenvector $\boldsymbol{\psi}_{\max}$ of the
joint covariance matrix $\mathbf{Q}$ and the inverted joint covariance matrix
$\mathbf{Q}^{-1}$ associated with associated with the pair of random quadratic
forms $\boldsymbol{\psi}^{T}\mathbf{Q}\boldsymbol{\psi}$ and $\boldsymbol{\psi
}^{T}\mathbf{Q}^{-1}\boldsymbol{\psi}$, at which point values of the scale
factors $\psi_{1i\ast}$ and $\psi_{2i\ast}$ are functions of covariance and
distribution information for all of the extreme vectors $k_{\mathbf{x}%
_{1_{i\ast}}}$ and $k_{\mathbf{x}_{2_{i\ast}}}$---in a given collection of
extreme vectors---relative to the covariance and distribution information
represented by the eigenvalues $\lambda_{N}^{-1}\leq\mathbf{\ldots}\leq
\lambda_{1}^{-1}$ of the inverted joint covariance matrix $\mathbf{Q}^{-1}$
for a given collection $\left\{  \mathbf{x}_{i}\right\}  _{i=1}^{N}$ of
feature vectors $\mathbf{x}_{i}$.

We use the KKT condition in (\ref{KKT 5}) and the theorem of Karush, Kuhn, and
Tucker to demonstrate that the geometric locus of the novel principal
eigenaxis $\boldsymbol{\kappa}$ $=\boldsymbol{\kappa}_{1}-\boldsymbol{\kappa
}_{2}$ of any given minimum risk binary classification system $k_{\mathbf{s}%
}\boldsymbol{\kappa}+$ $\boldsymbol{\kappa}_{0}\overset{\omega_{1}%
}{\underset{\omega_{2}}{\gtrless}}0$ satisfies the geometric locus of the
decision boundary $k_{\mathbf{s}}\boldsymbol{\kappa}+$ $\boldsymbol{\kappa
}_{0}=0$ of the system in terms of its critical eigenenergy $\left\Vert
\boldsymbol{\kappa}\right\Vert _{\min_{c}}^{2}$ and its minimum expected risk
$\mathfrak{R}_{\mathfrak{\min}}\left(  \left\Vert \boldsymbol{\kappa
}\right\Vert _{\min_{c}}^{2}\right)  $, such that the total allowed
eigenenergy $\left\Vert \boldsymbol{\kappa}\right\Vert _{\min_{c}}^{2}$ and
the expected risk $\mathfrak{R}_{\mathfrak{\min}}\left(  \left\Vert
\boldsymbol{\kappa}\right\Vert _{\min_{c}}^{2}\right)  $ are regulated by the
\emph{total value} of the scale factors $\psi_{1i\ast}$ and $\psi_{2i\ast}$
for the components $\psi_{1i\ast}\frac{k_{\mathbf{x}_{1i\ast}}}{\left\Vert
k_{\mathbf{x}_{1i\ast}}\right\Vert }$ and $\psi_{2i\ast}\frac{k_{\mathbf{x}%
_{2i\ast}}}{\left\Vert k_{\mathbf{x}_{2i\ast}}\right\Vert }$ of the principal
eigenvector $\boldsymbol{\psi}_{\max}$ of the joint covariance matrix
$\mathbf{Q}$ and the inverted joint covariance matrix $\mathbf{Q}^{-1}$
associated with the pair of random quadratic forms $\boldsymbol{\psi}%
^{T}\mathbf{Q}\boldsymbol{\psi}$ and $\boldsymbol{\psi}^{T}\mathbf{Q}%
^{-1}\boldsymbol{\psi}$.

Correspondingly, we demonstrate that the elements $\left\Vert k_{\mathbf{x}%
_{i}}\right\Vert \left\Vert k_{\mathbf{x}_{j}}\right\Vert \cos\theta
_{k_{\mathbf{x}_{i}}k_{\mathbf{x}_{j}}}$ and the eigenvalues $\lambda_{N}%
\leq\mathbf{\ldots}\leq\lambda_{1}$ of the joint covariance matrix
$\mathbf{Q}$ of the random quadratic form $\boldsymbol{\psi}^{T}%
\mathbf{Q}\boldsymbol{\psi}$ in the Wolfe-dual eigenenergy functional $\max
\Xi_{\boldsymbol{\psi}}\left(  \boldsymbol{\psi}\right)  =\mathbf{1}%
^{T}\boldsymbol{\psi}-\frac{\boldsymbol{\psi}^{T}\mathbf{Q}\boldsymbol{\psi}%
}{2}$ of (\ref{Matrix Version Wolfe Dual Problem}), wherein $\boldsymbol{\psi
}^{T}\mathbf{y}=0$ and $\psi_{i\ast}>0$, along with the scale factors
$\psi_{1i\ast}$ and $\psi_{2i\ast}$ for the components $\psi_{1i\ast}%
\frac{k_{\mathbf{x}_{1i\ast}}}{\left\Vert k_{\mathbf{x}_{1i\ast}}\right\Vert
}$ and $\psi_{2i\ast}\frac{k_{\mathbf{x}_{2i\ast}}}{\left\Vert k_{\mathbf{x}%
_{2i\ast}}\right\Vert }$ of the principal eigenvector $\boldsymbol{\psi}%
_{\max}$ of $\mathbf{Q}$ and $\mathbf{Q}^{-1}$%
\[
\boldsymbol{\psi}_{\max}=\sum\nolimits_{i=1}^{l_{1}}\psi_{1i\ast}%
\frac{k_{\mathbf{x}_{1i\ast}}}{\left\Vert k_{\mathbf{x}_{1i\ast}}\right\Vert
}+\sum\nolimits_{i=1}^{l_{2}}\psi_{2i\ast}\frac{k_{\mathbf{x}_{2i\ast}}%
}{\left\Vert k_{\mathbf{x}_{2i\ast}}\right\Vert }%
\]
are statistically interconnected with the components $\psi_{1_{i_{\ast}}%
}k_{\mathbf{x}_{1_{i\ast}}}$ and $\psi_{2_{i_{\ast}}}k_{\mathbf{x}_{2_{i\ast}%
}}$ of the geometric locus of the novel principal eigenaxis%
\[
\boldsymbol{\kappa}=\sum\nolimits_{i=1}^{l_{1}}\psi_{1_{i_{\ast}}%
}k_{\mathbf{x}_{1_{i\ast}}}-\sum\nolimits_{i=1}^{l_{2}}\psi_{2_{i_{\ast}}%
}k_{\mathbf{x}_{2_{i\ast}}}%
\]
in such a manner that the geometric locus of the novel principal eigenaxis
$\boldsymbol{\kappa}$ $=\boldsymbol{\kappa}_{1}-\boldsymbol{\kappa}_{2}$ is
the principal part of an equivalent representation of the pair of random
quadratic forms $\boldsymbol{\psi}^{T}\mathbf{Q}\boldsymbol{\psi}$ and
$\boldsymbol{\psi}^{T}\mathbf{Q}^{-1}\boldsymbol{\psi}$, so that the geometric
locus of the novel principal eigenaxis $\boldsymbol{\kappa}$
$=\boldsymbol{\kappa}_{1}-\boldsymbol{\kappa}_{2}$ satisfies the geometric
locus of the decision boundary $k_{\mathbf{s}}\boldsymbol{\kappa}+$
$\boldsymbol{\kappa}_{0}=0$ of a minimum risk binary classification system
$k_{\mathbf{s}}\boldsymbol{\kappa}+$ $\boldsymbol{\kappa}_{0}\overset{\omega
_{1}}{\underset{\omega_{2}}{\gtrless}}0$ in terms of a critical minimum
eigenenergy $\left\Vert \boldsymbol{\kappa}\right\Vert _{\min_{c}}^{2}$ and a
minimum expected risk $\mathfrak{R}_{\mathfrak{\min}}\left(  \left\Vert
\boldsymbol{\kappa}\right\Vert _{\min_{c}}^{2}\right)  $ in the following
manner%
\begin{align*}
\left\Vert \boldsymbol{\kappa}\right\Vert _{\min_{c}}^{2}  &  =\sum
\nolimits_{i=1}^{l_{1}}\psi_{1i\ast}\left(  1-\xi_{i}\right)  +\sum
\nolimits_{i=1}^{l_{2}}\psi_{2i\ast}\left(  1-\xi_{i}\right) \\
&  =\boldsymbol{\psi}_{\max}-\left(  \sum\nolimits_{i=1}^{l_{1}}\xi_{i}%
\psi_{1i\ast}+\sum\nolimits_{i=1}^{l_{2}}\xi_{i}\psi_{2i\ast}\right)  \text{,}%
\end{align*}
at which point the total value of the scale factors $\psi_{1i\ast}$ and
$\psi_{2i\ast}$ for the components $\psi_{1i\ast}\frac{k_{\mathbf{x}_{1i\ast}%
}}{\left\Vert k_{\mathbf{x}_{1i\ast}}\right\Vert }$ and $\psi_{2i\ast}%
\frac{k_{\mathbf{x}_{2i\ast}}}{\left\Vert k_{\mathbf{x}_{2i\ast}}\right\Vert
}$ of the principal eigenvector $\boldsymbol{\psi}_{\max}$ of the joint
covariance matrix $\mathbf{Q}$ of the random quadratic form $\boldsymbol{\psi
}^{T}\mathbf{Q}\boldsymbol{\psi}$ and the inverted joint covariance matrix
$\mathbf{Q}^{-1}$ of the random quadratic form $\boldsymbol{\psi}%
^{T}\mathbf{Q}^{-1}\boldsymbol{\psi}$ regulates the total allowed eigenenergy
$\left\Vert \boldsymbol{\kappa}\right\Vert _{\min_{c}}^{2}$ and the expected
risk $\mathfrak{R}_{\mathfrak{\min}}\left(  \left\Vert \boldsymbol{\kappa
}\right\Vert _{\min_{c}}^{2}\right)  $ exhibited by the geometric locus of the
novel principal eigenaxis $\boldsymbol{\kappa}$, wherein the regularization
parameters $\xi_{i}=\xi\ll1$ determine negligible constraints.

\subsection{A Vector-valued Cost Function}

We demonstrate that the Wolfe-dual eigenenergy functional uses a
\emph{vector-valued cost function} to find the principal eigenvector
$\boldsymbol{\psi}_{\max}$ of the joint covariance matrix $\mathbf{Q}$ of the
random quadratic form $\boldsymbol{\psi}^{T}\mathbf{Q}\boldsymbol{\psi}$ in
(\ref{Matrix Version Wolfe Dual Problem}), so that the Wolfe-dual eigenenergy
functional%
\[
\max\Xi\left(  \boldsymbol{\psi}\right)  =\mathbf{1}^{T}\boldsymbol{\psi
}-\boldsymbol{\psi}^{T}\mathbf{Q}\boldsymbol{\psi/}2\text{,}%
\]
wherein $\boldsymbol{\psi}^{T}\mathbf{y}=0$ and $\psi_{i\ast}>0$, is
\emph{maximized} by the largest eigenvector $\boldsymbol{\psi}_{\max}$ of the
joint covariance matrix $\mathbf{Q}$%
\[
\mathbf{Q}\boldsymbol{\psi}_{\max}=\lambda_{1}\boldsymbol{\psi}_{\max}\text{,}%
\]
at which point the random quadratic form $\boldsymbol{\psi}_{\max}%
^{T}\mathbf{Q}\boldsymbol{\psi}_{\max}$, plus the total allowed eigenenergy
$\left\Vert \boldsymbol{\kappa}_{1}-\boldsymbol{\kappa}_{2}\right\Vert
_{\min_{c}}^{2}$ and the expected risk $\mathfrak{R}_{\mathfrak{\min}}\left(
\left\Vert \boldsymbol{\kappa}_{1}-\boldsymbol{\kappa}_{2}\right\Vert
_{\min_{c}}^{2}\right)  $ exhibited by the geometric locus of the novel
principal eigenaxis $\boldsymbol{\kappa}$ $=\boldsymbol{\kappa}_{1}%
-\boldsymbol{\kappa}_{2}$ \emph{jointly} reach their \emph{minimum} values.

\subsection{A Note on Scalar-valued Cost Functions}

Bayes' decision rule and support vector learning machines both \emph{use
scalar-valued cost functions} to \emph{find} decision functions or indicator
functions of binary classification systems.

By the analysis presented in this treatise, it will be seen that any given
statistical method or machine learning algorithm that employs
\emph{scalar-valued cost functions}---for the fundamental problem of finding
discriminant functions of minimum risk binary classification systems---grossly
\emph{oversimplifies} the \emph{complexity} of the problem.

We now devise vector expressions for the primal and the Wolfe-dual novel
principal eigenaxis.

\subsection{The Primal Novel Principal Eigenaxis}

By the KKT conditions in (\ref{KKT 1}) and (\ref{KKT 4}), it follows that the
geometrical and statistical structure of the geometric locus of the primal
novel principal eigenaxis $\boldsymbol{\kappa}$ is determined by the vector
expression%
\begin{equation}
\boldsymbol{\kappa}=\sum\nolimits_{i=1}^{N}y_{i}\psi_{i}k_{\mathbf{x}_{i}%
}\text{,} \tag{12.10}\label{Primal Novel Principal Eigenaxis}%
\end{equation}
where $\psi_{i}\geq0$, wherein $\psi_{i}>0$ by the KKT condition of
complementary slackness. Reproducing kernels $k_{\mathbf{x}_{i}}$ for feature
vectors $\mathbf{x}_{i}$ that are correlated with scaled unit reproducing
kernels for feature vectors $\psi_{i}\frac{k_{\mathbf{x}_{i}}}{\left\Vert
k_{\mathbf{x}_{i}}\right\Vert }$ that have non-zero magnitudes $\psi_{i}>0$
are called extreme vectors.

Let the scaled extreme vectors that belong to class $\omega_{1}$ and class
$\omega_{2}$ be denoted by $\psi_{1_{i\ast}}k_{\mathbf{x}_{1_{i\ast}}}$ and
$\psi_{2_{i\ast}}k_{\mathbf{x}_{2_{i\ast}}}$ respectively, so that
$\psi_{1_{i\ast}}$ is the scale factor for the extreme vector $k_{\mathbf{x}%
_{1_{i\ast}}}$ and $\psi_{2_{i\ast}}$ is the scale factor for the extreme
vector $k_{\mathbf{x}_{2_{i\ast}}}$. Let there be $l_{1}$ scaled extreme
vectors $\left\{  \psi_{1_{i\ast}}k_{\mathbf{x}_{1_{i\ast}}}\right\}
_{i=1}^{l_{1}}$ that belong to class $\omega_{1}$ and $l_{2}$ scaled extreme
vectors $\left\{  \psi_{2_{i\ast}}k_{\mathbf{x}_{2_{i\ast}}}\right\}
_{i=1}^{l_{2}}$ that belong to class $\omega_{2}$.

Using (\ref{Primal Novel Principal Eigenaxis}), class membership statistics
and the assumptions outlined above, it follows that the structure of the
geometric locus of the primal novel principal eigenaxis $\boldsymbol{\kappa}$
is determined by the vector difference of a pair of directed line segment
estimates%
\begin{align}
\boldsymbol{\kappa}  &  =\sum\nolimits_{i=1}^{l_{1}}\psi_{1_{i\ast}%
}k_{\mathbf{x}_{1_{i\ast}}}-\sum\nolimits_{i=1}^{l_{2}}\psi_{2_{i\ast}%
}k_{\mathbf{x}_{2_{i\ast}}}\tag{12.11}\label{Primal Dual Locus}\\
&  =\boldsymbol{\kappa}_{1}{\large -}\boldsymbol{\kappa}_{2}\text{,}\nonumber
\end{align}
at which point the primal novel principal eigenaxis $\boldsymbol{\kappa}$ is a
locus of principal eigenaxis components $\psi_{1_{i\ast}}k_{\mathbf{x}%
_{1_{i\ast}}}$ and $\psi_{2_{i\ast}}k_{\mathbf{x}_{2_{i\ast}}}$, where
$\boldsymbol{\kappa}_{1}$ and $\boldsymbol{\kappa}_{2}$ denote the sides of
$\boldsymbol{\kappa}$, such that the side $\boldsymbol{\kappa}_{1}$ is
determined by the vector expression $\boldsymbol{\kappa}_{1}=\sum
\nolimits_{i=1}^{l_{1}}\psi_{1_{i\ast}}k_{\mathbf{x}_{1_{i\ast}}}$, the side
$\boldsymbol{\kappa}_{2}$ is determined by the vector expression
$\boldsymbol{\kappa}_{2}=\sum\nolimits_{i=1}^{l_{2}}\psi_{2_{i\ast}%
}k_{\mathbf{x}_{2_{i\ast}}}$, and the geometric locus of the primal novel
eigenaxis $\boldsymbol{\kappa}$ is determined by the vector difference of side
$\boldsymbol{\kappa}_{1}$ and side $\boldsymbol{\kappa}_{2}$.

\subsection{The Wolfe-dual Novel Principal Eigenaxis}

Given the Wolfe-dual eigenenergy functional $\max\Xi_{\boldsymbol{\psi}%
}\left(  \boldsymbol{\psi}\right)  =\mathbf{1}^{T}\boldsymbol{\psi}%
-\frac{\boldsymbol{\psi}^{T}\mathbf{Q}\boldsymbol{\psi}}{2}$ in
(\ref{Matrix Version Wolfe Dual Problem}), wherein $\boldsymbol{\psi}%
^{T}\mathbf{y}=0$ and $\psi_{i\ast}>0$, and the guarantee provided by the
strong duality theorem
\citep{Fletcher2000,Luenberger1969,Luenberger2003,Nash1996}%
, it follows that the principal eigenaxis components $\psi_{1i\ast}%
\frac{k_{\mathbf{x}_{1i\ast}}}{\left\Vert k_{\mathbf{x}_{1i\ast}}\right\Vert
}$ and $\psi_{2i\ast}\frac{k_{\mathbf{x}_{2i\ast}}}{\left\Vert k_{\mathbf{x}%
_{2i\ast}}\right\Vert }$ on the geometric locus of the Wolfe-dual novel
principal eigenaxis $\boldsymbol{\psi}$ are symmetrically and equivalently
related to the principal eigenaxis components $\psi_{1_{i\ast}}k_{\mathbf{x}%
_{1_{i\ast}}}$ and $\psi_{2_{i\ast}}k_{\mathbf{x}_{2_{i\ast}}}$ on the
geometric locus of the primal novel principal eigenaxis $\boldsymbol{\kappa}$,
so that the geometrical and statistical structure of the geometric locus of
the Wolfe-dual novel principal eigenaxis $\boldsymbol{\psi}$ is determined by
the vector sum of a pair of directed line segment estimates%
\begin{align}
\boldsymbol{\psi}  &  =\sum\nolimits_{i=1}^{l_{1}}\psi_{1i\ast}\frac
{k_{\mathbf{x}_{1i\ast}}}{\left\Vert k_{\mathbf{x}_{1i\ast}}\right\Vert }%
+\sum\nolimits_{i=1}^{l_{2}}\psi_{2i\ast}\frac{k_{\mathbf{x}_{2i\ast}}%
}{\left\Vert k_{\mathbf{x}_{2i\ast}}\right\Vert } \tag{12.12}%
\label{Wolf-dual Dual Locus}\\
&  =\boldsymbol{\psi}_{1}+\boldsymbol{\psi}_{2}\text{,}\nonumber
\end{align}
at which point the Wolfe-dual novel principal eigenaxis $\boldsymbol{\psi}$
\ is a locus of principal eigenaxis components $\psi_{1i\ast}\frac
{k_{\mathbf{x}_{1i\ast}}}{\left\Vert k_{\mathbf{x}_{1i\ast}}\right\Vert }$ and
$\psi_{2i\ast}\frac{k_{\mathbf{x}_{2i\ast}}}{\left\Vert k_{\mathbf{x}_{2i\ast
}}\right\Vert }$, where $\boldsymbol{\psi}_{1}$ and $\boldsymbol{\psi}_{2}$
denote the sides of $\boldsymbol{\psi}$, such that the side $\boldsymbol{\psi
}_{1}$ is determined by the vector expression $\boldsymbol{\psi}_{1}%
=\sum\nolimits_{i=1}^{l_{1}}\psi_{1i\ast}\frac{k_{\mathbf{x}_{1i\ast}}%
}{\left\Vert k_{\mathbf{x}_{1i\ast}}\right\Vert }$, the side $\boldsymbol{\psi
}_{2}$ is determined by the vector expression $\boldsymbol{\psi}_{2}%
=\sum\nolimits_{i=1}^{l_{2}}\psi_{2i\ast}\frac{k_{\mathbf{x}_{2i\ast}}%
}{\left\Vert k_{\mathbf{x}_{2i\ast}}\right\Vert }$, and the geometric locus of
$\boldsymbol{\psi}$ is determined by the vector sum of side $\boldsymbol{\psi
}_{1}$ and side $\boldsymbol{\psi}_{2}$.

We now turn our attention to the eigenstructure content of the joint
covariance matrix $\mathbf{Q}$---of the random quadratic form
$\boldsymbol{\psi}^{T}\mathbf{Q}\boldsymbol{\psi}$---in the Wolfe-dual
eigenenergy functional $\max\Xi\left(  \boldsymbol{\psi}\right)
=\mathbf{1}^{T}\boldsymbol{\psi}-\boldsymbol{\psi}^{T}\mathbf{Q}%
\boldsymbol{\psi/}2$ in (\ref{Matrix Version Wolfe Dual Problem}).

\section{\label{Section 13}Eigenstructure Content of Data Matrices}

The machine learning algorithm being investigated is also solving a system
identification problem, such that the overall statistical structure and
behavior and properties of a minimum risk binary classification system are
determined by transforming a collection of training data into a data-driven
mathematical model that represents fundamental aspects of the system.

Indeed, we will demonstrate that the machine learning algorithm being
investigated \emph{statistically pre-wires the important generalizations }for
a minimum risk binary classification system $k_{\mathbf{s}}\boldsymbol{\kappa
}+$ $\boldsymbol{\kappa}_{0}\overset{\omega_{1}}{\underset{\omega
_{2}}{\gtrless}}0$ \emph{within the geometric locus of the novel principal
eigenaxis }$\boldsymbol{\kappa}=\boldsymbol{\kappa}_{1}-\boldsymbol{\kappa
}_{2}$ of the system.

We now demonstrate that any given minimum risk binary classification system
$k_{\mathbf{s}}\boldsymbol{\kappa}+$ $\boldsymbol{\kappa}_{0}\overset{\omega
_{1}}{\underset{\omega_{2}}{\gtrless}}0$ based on eigenstructure deficiencies
is \emph{ill-posed}---such that the geometric locus of the novel principal
eigenaxis $\boldsymbol{\kappa}$ of the system is not characteristic of the
system (is \emph{not unique}) and is \emph{unstable}---at which point the
total allowed \emph{eigenenergy} $\left\Vert \boldsymbol{\kappa}%
_{1}-\boldsymbol{\kappa}_{2}\right\Vert _{\min_{c}}^{2}$ exhibited by the
geometric locus of the novel principal eigenaxis $\boldsymbol{\kappa
}=\boldsymbol{\kappa}_{1}-\boldsymbol{\kappa}_{2}$ is \emph{maximized}.

Recall that solving a system identification problem involves solving an
inverse problem, such that a collection of training data are used to infer the
values of the parameters characterizing a given system. Thus, the machine
learning algorithm being investigated is also \emph{solving} a
\emph{well-posed} inverse problem---so that solutions obtained by the
algorithm are both \emph{unique} and \emph{stable}.

Given the novel principal eigen-coordinate transform method expressed by
Theorem \ref{Principal Eigen-coordinate System Theorem} and Corollary
\ref{Principal Eigen-coordinate System Corollary}, we know that the essential
information content of any given training data set---that is used to find a
geometric locus of a novel principal eigenaxis $\boldsymbol{\kappa}$---is
contained within the eigenstructures of the data set, such that all of the
individual pattern vectors `add up' to a complete and sufficient eigenstructure.

We also recognize that Bellman's \textquotedblleft curse of
dimensionality\textquotedblright\ is concerned with the fundamental problem of
parameter estimates that are based on insufficient eigenstructures
\citep{bellman1961adaptive}%
. We show that solutions obtained by the machine learning algorithm being
examined---which are based on eigenstructure deficiencies---are generally
ill-posed and ill-conditioned, and must be constrained in some manner.

\subsection{Insufficient Learning Capacity}

A training set of $N$ pattern vectors with dimension $d$ has at most $d$
non-zero eigenvalues, given that $N>d$. Accordingly, the Wolfe-dual principal
eigenspace that is associated with the Wolfe-dual eigenenergy functional
$\max\Xi\left(  \boldsymbol{\psi}\right)  =\mathbf{1}^{T}\boldsymbol{\psi
}-\boldsymbol{\psi}^{T}\mathbf{Q}\boldsymbol{\psi/}2$ in
(\ref{Matrix Version Wolfe Dual Problem}) is spanned by $d$ or fewer
eigenfunctions
\citep{Reeves2009,Reeves2011}%
. Thus, the machine learning algorithm being examined has insufficient
\textquotedblleft learning capacity\textquotedblright\ whenever $N>d$.

Eigenstructures are an inherent part of machine learning algorithms. Moreover,
machine learning solutions with eigenstructure deficiencies are generally
ill-posed and ill-conditioned, and must be constrained in some manner
\citep
{Engl2000,Groetsch1984,Groetsch1993,Hansen1998,Linz1979,Linz2003,Wahba1987}%
.

\subsection{Complete Eigenstructures}

In order for the machine learning algorithm being examined to find
discriminant functions of minimum risk binary classification systems, it is
vital that all of the $N$ feature vectors in any given collection of training
data%
\[
\left(  \mathbf{x}_{1}\mathbf{,}y_{1}\right)  ,\ldots,\left(  \mathbf{x}%
_{N}\mathbf{,}y_{N}\right)  \in%
\mathbb{R}
^{d}\times Y,Y=\left\{  \pm1\right\}
\]
\emph{boil down} to a \emph{complete eigenstructure} of the joint covariance
matrix $\mathbf{Q}$---of the random quadratic form $\boldsymbol{\psi}%
^{T}\mathbf{Q}\boldsymbol{\psi}$---in the Wolfe-dual eigenenergy functional
$\max\Xi\left(  \boldsymbol{\psi}\right)  =\mathbf{1}^{T}\boldsymbol{\psi
}-\boldsymbol{\psi}^{T}\mathbf{Q}\boldsymbol{\psi/}2$ in
(\ref{Matrix Version Wolfe Dual Problem}).

To see this, consider the Cayley--Hamilton theorem
\citep{Lathi1998,Meyer2000}%
, which states that the roots $p\left(  \lambda\right)  =0$ of the
characteristic polynomial $p\left(  \lambda\right)  $ of the joint covariance
matrix $\mathbf{Q}$%
\[
\det\left(
\begin{bmatrix}
\left\Vert k_{\mathbf{x}_{1}}\right\Vert \left\Vert k_{\mathbf{x}_{1}%
}\right\Vert \cos\theta_{k_{\mathbf{x_{1}}}k_{\mathbf{x}_{1}}}-\lambda_{1} &
\cdots & -\left\Vert k_{\mathbf{x}_{1}}\right\Vert \left\Vert k_{\mathbf{x}%
_{N}}\right\Vert \cos\theta_{k_{\mathbf{x}_{1}}k_{\mathbf{x}_{N}}}\\
\vdots & \ddots & \vdots\\
-\left\Vert k_{\mathbf{x}_{N}}\right\Vert \left\Vert k_{\mathbf{x}_{1}%
}\right\Vert \cos\theta_{k_{\mathbf{x}_{N}}k_{\mathbf{x}_{1}}} & \cdots &
\left\Vert k_{\mathbf{x}_{N}}\right\Vert \left\Vert k_{\mathbf{x}_{N}%
}\right\Vert \cos\theta_{k_{\mathbf{x}_{N}}k_{\mathbf{x}_{N}}}-\lambda_{N}%
\end{bmatrix}
\right)  =0
\]
are the eigenvalues $\lambda_{N}\leq\mathbf{\ldots}\leq\lambda_{1}$ of
$\mathbf{Q}$.

Since the roots $p\left(  \lambda\right)  =0$ of the characteristic polynomial
$p\left(  \lambda\right)  $ of the joint covariance matrix $\mathbf{Q}$ vary
continuously with its coefficients
\citep{Meyer2000}%
, it follows that the eigenvalues $\lambda_{N}\leq\mathbf{\ldots}\leq
\lambda_{1}$ of the joint covariance matrix $\mathbf{Q}$ vary continuously
with the elements $\left\Vert k_{\mathbf{x}_{i}}\right\Vert \left\Vert
k_{\mathbf{x}_{j}}\right\Vert \cos\theta_{k_{\mathbf{x_{i}}}k_{\mathbf{x}_{j}%
}}$ of $\mathbf{Q}$.

Thereby, the eigenvalues $\lambda_{N}\leq\mathbf{\ldots}\leq\lambda_{1}$ of
the joint covariance matrix $\mathbf{Q}$ of the random quadratic form
$\boldsymbol{\psi}^{T}\mathbf{Q}\boldsymbol{\psi}$ \emph{account for} joint
variabilities between coordinates $\left\{  \left\Vert k_{\mathbf{x}_{i}%
}\right\Vert \cos\mathbb{\alpha}_{\mathbf{e}_{i}k_{\mathbf{x}_{i}}}\right\}
_{i=1}^{d}$ and $\left\{  \left\Vert k_{\mathbf{x}_{j}}\right\Vert
\cos\mathbb{\alpha}_{\mathbf{e}_{i}k_{\mathbf{x}_{j}}}\right\}  _{i=1}^{d}$ of
feature vectors $k_{\mathbf{x}_{i}}$ and $k_{\mathbf{x}_{j}}$ used to
construct $\mathbf{Q}$, such that each element $y_{i}\left\Vert k_{\mathbf{x}%
_{i}}\right\Vert y_{j}\left\Vert k_{\mathbf{x}_{j}}\right\Vert \cos
\theta_{k_{\mathbf{x}_{i}}k_{\mathbf{x}_{j}}}$ of the joint covariance matrix
$\mathbf{Q}$ is correlated with the distance $\left\Vert k_{\mathbf{x}_{i}%
}-k_{\mathbf{x}_{j}}\right\Vert $ between the loci of certain feature vectors
$k_{\mathbf{x}_{i}}$ and $k_{\mathbf{x}_{j}}$.

We will demonstrate that it is crucial that all of the individual feature
vectors `add up' to a complete eigenstructure of the joint covariance matrix
$\mathbf{Q}$ of the random quadratic form $\boldsymbol{\psi}^{T}%
\mathbf{Q}\boldsymbol{\psi}$ \emph{and} the inverted joint covariance matrix
$\mathbf{Q}^{-1}$ of the random quadratic form $\boldsymbol{\psi}%
^{T}\mathbf{Q}^{-1}\boldsymbol{\psi}$, so that all of the individual feature
vectors `speak for themselves' in such a manner that joint variabilities
between all of the feature vectors are `accounted for.'

It will be seen that the machine learning algorithm being examined uses the
condensed eigenstructures of the joint covariance matrix $\mathbf{Q}$ of the
random quadratic form $\boldsymbol{\psi}^{T}\mathbf{Q}\boldsymbol{\psi}$ and
the inverted joint covariance matrix $\mathbf{Q}^{-1}$ of the random quadratic
form $\boldsymbol{\psi}^{T}\mathbf{Q}^{-1}\boldsymbol{\psi}$ to \emph{locate}
a point of equilibrium%
\[
\boldsymbol{\kappa}=\sum\nolimits_{i=1}^{l_{1}}\psi_{1_{i_{\ast}}%
}k_{\mathbf{x}_{1_{i\ast}}}-\sum\nolimits_{i=1}^{l_{2}}\psi_{2_{i_{\ast}}%
}k_{\mathbf{x}_{2_{i\ast}}}\text{,}%
\]
at which point all of the critical minimums eigenenergies $\left\Vert
\psi_{1_{i\ast}}k_{\mathbf{x}_{1i\ast}}\right\Vert _{\min_{c}}^{2}$ and
$\left\Vert \psi_{1_{i\ast}}k_{\mathbf{x}_{1i\ast}}\right\Vert _{\min_{c}}%
^{2}$ exhibited by a minimum risk binary classification system $k_{\mathbf{s}%
}\boldsymbol{\kappa}+$ $\boldsymbol{\kappa}_{0}\overset{\omega_{1}%
}{\underset{\omega_{2}}{\gtrless}}0$ are symmetrically concentrated in such a
manner that opposing and counteracting forces and influences of the system are
symmetrically balanced with each other---about the geometric center of the
locus of the novel principal eigenaxis $\boldsymbol{\kappa}$---whereon the
statistical fulcrum of the system is located.

Thereby, it will be seen that any given minimum risk binary classification
system $k_{\mathbf{s}}\boldsymbol{\kappa}+$ $\boldsymbol{\kappa}%
_{0}\overset{\omega_{1}}{\underset{\omega_{2}}{\gtrless}}0$ satisfies a state
of statistical equilibrium so that the total allowed eigenenergy $\left\Vert
\boldsymbol{\kappa}\right\Vert _{\min_{c}}^{2}$ and the expected risk
$\mathfrak{R}_{\mathfrak{\min}}\left(  \left\Vert \boldsymbol{\kappa
}\right\Vert _{\min_{c}}^{2}\right)  $ exhibited by the system are jointly
minimized within the decision space of the system, at which point the system
exhibits the minimum probability of classification error.

\subsection{Why Complete Eigenstructures Are Essential}

Let the Wolfe-dual novel principal eigenaxis $\boldsymbol{\psi}$ be
symmetrically and equivalently related to the principal eigenvector
$\boldsymbol{\psi}_{\max}$ of the joint covariance matrix $\mathbf{Q}$ of the
random quadratic form $\boldsymbol{\psi}^{T}\mathbf{Q}\boldsymbol{\psi}$ in
the Wolfe-dual eigenenergy functional $\max\Xi\left(  \boldsymbol{\psi
}\right)  =\mathbf{1}^{T}\boldsymbol{\psi}-\boldsymbol{\psi}^{T}%
\mathbf{Q}\boldsymbol{\psi/}2$ in (\ref{Matrix Version Wolfe Dual Problem}),
so that the primal novel principal eigenaxis $\boldsymbol{\kappa}$ is
symmetrically and equivalently related to the Wolfe-dual novel principal
eigenaxis $\boldsymbol{\psi}$ in such a manner that the primal novel principal
eigenaxis $\boldsymbol{\kappa}$ is the principal part of an equivalent
representation of the pair of random quadratic forms $\boldsymbol{\psi}%
^{T}\mathbf{Q}\boldsymbol{\psi}$ and $\boldsymbol{\psi}^{T}\mathbf{Q}%
^{-1}\boldsymbol{\psi}$.

By Theorem \ref{Principal Eigen-coordinate System Theorem} and Corollary
\ref{Principal Eigen-coordinate System Corollary}, we know that an equivalent
representation of the pair of random quadratic forms $\boldsymbol{\psi}%
^{T}\mathbf{Q}\boldsymbol{\psi}$ and $\boldsymbol{\psi}^{T}\mathbf{Q}%
^{-1}\boldsymbol{\psi}$ is based on complete eigenstructures of $\mathbf{Q}$
and $\mathbf{Q}^{-1}$, at which point the joint covariance matrix $\mathbf{Q}$
of the random quadratic form $\boldsymbol{\psi}^{T}\mathbf{Q}\boldsymbol{\psi
}$ and the inverted joint covariance matrix $\mathbf{Q}^{-1}$ of the random
quadratic form $\boldsymbol{\psi}^{T}\mathbf{Q}^{-1}\boldsymbol{\psi}$ both
have full rank.

We will demonstrate that finding an equivalent representation of the random
quadratic form $\boldsymbol{\psi}^{T}\mathbf{Q}\boldsymbol{\psi}$ requires
finding the values of the active scale factors $\psi_{i\ast}>0$---which
requires finding the inverse $\mathbf{Q}^{-1}$ of the joint covariance matrix
$\mathbf{Q}$---which is an ill-posed inverse problem if $\mathbf{Q}$ has low
rank
\citep
{Engl2000,Groetsch1984,Groetsch1993,Hansen1998,Linz1979,Linz2003,Wahba1987}%
.

We have previously demonstrated that the values of the active scale factors
$\psi_{i\ast}>0$ and the resulting locations of the principal eigenaxis
components on $\boldsymbol{\psi}$ and $\boldsymbol{\kappa}$ are considerably
affected by the rank and the eigenstructure of the joint covariance matrix
$\mathbf{Q}$
\citep{Reeves2018design}%
. For example, a low rank joint covariance matrix $\mathbf{Q}$ determines a
binary classification system $k_{\mathbf{s}}\boldsymbol{\kappa}+$
$\boldsymbol{\kappa}_{0}\overset{\omega_{1}}{\underset{\omega_{2}}{\gtrless}%
}0$ that exhibits poor generalization behavior, so that the decision space
$Z=Z_{1}\cup Z_{2}$ of the system is partitioned in an unbalanced manner, at
which point \emph{all} of the training \emph{data} are \emph{extreme points}.

Thereby, the geometric locus of the novel principal eigenaxis
$\boldsymbol{\kappa}$ of the binary classification system $k_{\mathbf{s}%
}\boldsymbol{\kappa}+$ $\boldsymbol{\kappa}_{0}\overset{\omega_{1}%
}{\underset{\omega_{2}}{\gtrless}}0$ is not characteristic of the system (is
not unique) and is unstable---such that the geometric locus of the novel
principal eigenaxis $\boldsymbol{\kappa}$ of the system does not represent an
eigenaxis of symmetry that spans the decision space of the system, at which
point the \emph{total allowed eigenenergy} $\left\Vert \boldsymbol{\kappa}%
_{1}-\boldsymbol{\kappa}_{2}\right\Vert _{\min_{c}}^{2}$ exhibited by the
novel principal eigenaxis $\boldsymbol{\kappa}$ is \emph{maximized}.

Any given joint covariance matrix $\mathbf{Q}$ has low rank whenever $d<N$ for
a collection of $N$ feature vectors of dimension $d$. We have resolved these
eigenstructure deficiencies by the regularization method that is described below.

\subsection{A Regularization Method for Complete Eigenstructures}

The regularized structure of the joint covariance matrix $\mathbf{Q}$ in
(\ref{Matrix Version Wolfe Dual Problem}), wherein $\mathbf{Q}\triangleq
\epsilon\mathbf{I}+\widetilde{\mathbf{X}}\widetilde{\mathbf{X}}^{T}$ and
$\epsilon\ll1$, ensures that $\mathbf{Q}$ has full rank and a complete
eigenvector set $\boldsymbol{\psi=}%
{\textstyle\sum\nolimits_{i=1}^{N}}
\psi_{i}\mathbf{v}_{i}$ that spans the parameter space of the unknown scale
factors $\psi_{i}$, so that $\mathbf{Q}$ has a complete eigenstructure. The
regularization constant $C$ in (\ref{Objective Function}) is related to the
regularization parameter $\epsilon$ in (\ref{Objective Function}) by $\frac
{1}{C}$.

These findings have been published in
\citep{Reeves2011}
and
\citep{Reeves2009}%
, where the joint covariance matrix in
(\ref{Matrix Version Wolfe Dual Problem}) is a Gram matrix, and are readily
extended to Kernel Gram matrices. Simulation studies published in
\citep{Reeves2011}
and
\citep{Reeves2009}
demonstrate that the decision space of a binary classification system is
partitioned in an unbalanced manner if a Gram matrix has low rank, where the
decision boundary is a linear decision boundary.

Thus, it follows that any given geometric locus of a novel principal eigenaxis
that is based on an incomplete eigenstructure---of a joint covariance
matrix---is not characteristic of a minimum risk binary classification system
(is not unique) and is unstable, so that the geometric locus of the novel
principal eigenaxis $\boldsymbol{\kappa}$ of the system does not represent an
eigenaxis of symmetry that spans the decision space of the system, at which
point the the total allowed eigenenergy exhibited by the novel principal
eigenaxis is maximized.

\subsection{Values of Regularization Parameters}

Given $N$ feature vectors of dimension $d$, where $d<N$, all of the
regularization parameters $\left\{  \xi_{i}\right\}  _{i=1}^{N}$ in
(\ref{Objective Function}) and all of its derivatives are set equal to a very
small value $\xi_{i}=\xi\ll1$, e.g. $\xi_{i}=\xi=0.02$, wherein $C=\frac
{1}{\xi}$.

Otherwise, given $N$ feature vectors of dimension $d$, where $N<d$, all of the
regularization parameters $\left\{  \xi_{i}\right\}  _{i=1}^{N}$ in
(\ref{Objective Function}) and all of its derivatives are set equal to the
value of zero $\xi_{i}=\xi=0$, wherein $C=\infty$.

\subsection{Ill-posed Binary Classification Systems}

Given conditions expressed by Theorem
\ref{Principal Eigen-coordinate System Theorem}, Corollaries
\ref{Principal Eigen-coordinate System Corollary} -
\ref{Symmetrical and Equivalent Principal Eigenaxes} and Theorem
\ref{Direct Problem of Binary Classification Theorem}, we realize that the
shape of the decision space of any given minimum risk binary classification
system $k_{\mathbf{s}}\boldsymbol{\kappa}+$ $\boldsymbol{\kappa}%
_{0}\overset{\omega_{1}}{\underset{\omega_{2}}{\gtrless}}0$ is completely
determined by the geometric locus of the novel principal eigenaxis
$\boldsymbol{\kappa}$ of the system, such that the novel principal eigenaxis
$\boldsymbol{\kappa}$ is the \emph{principal part} of an \emph{equivalent
representation} of the pair of random quadratic forms $\boldsymbol{\psi}%
^{T}\mathbf{Q}\boldsymbol{\psi}$ and $\boldsymbol{\psi}^{T}\mathbf{Q}%
^{-1}\boldsymbol{\psi}$---that is based on the \emph{eigenstructures} of the
joint covariance matrix $\mathbf{Q}$ of the random quadratic form
$\boldsymbol{\psi}^{T}\mathbf{Q}\boldsymbol{\psi}$ and the inverted joint
covariance matrix $\mathbf{Q}^{-1}$ of the random quadratic form
$\boldsymbol{\psi}^{T}\mathbf{Q}^{-1}\boldsymbol{\psi}$---so that the
geometric locus of the novel principal eigenaxis $\boldsymbol{\kappa}$
represents an eigenaxis of symmetry that spans the decision space of the
system, at which point the eigenenergy exhibited by the principal eigenvector
$\boldsymbol{\psi}_{\max}$ of $\mathbf{Q}$ and $\mathbf{Q}^{-1}$, along with
the eigenenergy exhibited by the novel principal eigenaxis $\boldsymbol{\kappa
}$, are jointly minimized.

Indeed, we now demonstrate that an insufficient eigenstructure of a joint
covariance matrix $\mathbf{Q}$ of a random quadratic form $\boldsymbol{\psi
}^{T}\mathbf{Q}\boldsymbol{\psi}$---in the Wolfe-dual eigenenergy functional
in (\ref{Matrix Version Wolfe Dual Problem})---of any given minimum risk
binary classification system $k_{\mathbf{s}}\boldsymbol{\kappa}+$
$\boldsymbol{\kappa}_{0}\overset{\omega_{1}}{\underset{\omega_{2}}{\gtrless}%
}0$ determines an irregularly shaped decision space of the binary
classification system, at which point the geometric locus of the novel
principal eigenaxis $\boldsymbol{\kappa}$ of the system does not represent an
eigenaxis of symmetry that spans the decision space of the system.

Thereby, we demonstrate that any given minimum risk binary classification
system $k_{\mathbf{s}}\boldsymbol{\kappa}+$ $\boldsymbol{\kappa}%
_{0}\overset{\omega_{1}}{\underset{\omega_{2}}{\gtrless}}0$ based on
eigenstructure deficiencies is \emph{ill-posed}---so that the geometric locus
of the novel principal eigenaxis $\boldsymbol{\kappa}$ of the system is
\emph{not unique} and is \emph{unstable}---such that the geometric locus of
the novel principal eigenaxis $\boldsymbol{\kappa}$ does not represent an
eigenaxis of symmetry that spans the decision space of the system, at which
point the total allowed \emph{eigenenergy} $\left\Vert \boldsymbol{\kappa}%
_{1}-\boldsymbol{\kappa}_{2}\right\Vert _{\min_{c}}^{2}$ exhibited by the
geometric locus of the novel principal eigenaxis $\boldsymbol{\kappa
}=\boldsymbol{\kappa}_{1}-\boldsymbol{\kappa}_{2}$ is \emph{maximized}.

By way of demonstration, we now present regularization examples for full rank
and low rank Gram matrices, as well as full rank and low rank Polynomial
Kernel Gram matrices---that illustrate the generalization performance of the
machine learning algorithm being examined---for two binary classification systems.

\subsection{Regularization Example One}

Consider the minimum risk binary classification system $k_{\mathbf{s}%
}\boldsymbol{\kappa}+$ $\boldsymbol{\kappa}_{0}\overset{\omega_{1}%
}{\underset{\omega_{2}}{\gtrless}}0$ for two classes of random vectors that
have similar covariance matrices, such that the covariance matrices for class
$\omega_{1}$ and class $\omega_{2}$ are both given by%
\[
\Sigma_{1}=\Sigma_{2}=\left[
\begin{array}
[c]{cc}%
0.95 & 0.45\\
0.45 & 0.35
\end{array}
\right]  \text{,}%
\]
the mean vector for class $\omega_{1}$ is given by $M_{1}=%
\begin{pmatrix}
3, & 0.25
\end{pmatrix}
^{T}$ and the mean vector for class $\omega_{2}$ is given by $M_{2}=%
\begin{pmatrix}
3, & -0.25
\end{pmatrix}
^{T}$, wherein the probability density functions of the two classes of random
vectors determine overlapping distributions of the random vectors.

We now consider the generalization performance of the machine learning
algorithm being examined---for the binary classification system outlined
above---for full rank and low rank Gram matrices, as well as full rank and low
rank Polynomial Kernel Gram matrices.

\paragraph{A Full Rank Gram Matrix}

Since the covariance matrices are similar, we can use a linear kernel in the
constrained optimization algorithm that resolves the inverse problem of binary
classification, such that the $N\times d$ matrix $\widetilde{\mathbf{X}}$ in
(\ref{Lagrangian Dual Prob}) is a matrix of $N$ labeled feature vectors
$\widetilde{\mathbf{X}}=%
\begin{pmatrix}
y_{1}\mathbf{x}_{1}, & y_{2}\mathbf{x}_{2}, & \ldots, & y_{N}\mathbf{x}_{N}%
\end{pmatrix}
^{T}$. Now suppose that we let all of the regularization parameters $\left\{
\xi_{i}\right\}  _{i=1}^{N}$ in (\ref{Objective Function}) and all of its
derivatives be equal to $\xi_{i}=\xi=0.02$, wherein $C=50$.

Figure $8$ illustrates that the constrained optimization algorithm being
examined finds a stable and characteristic geometric locus of a novel
principal eigenaxis $\boldsymbol{\tau}=\boldsymbol{\tau_{1}-\tau}_{2}$, such
that the geometric locus of the novel principal eigenaxis $\boldsymbol{\tau
}=\boldsymbol{\tau_{1}-\tau}_{2}$ represents an eigenaxis of symmetry that
spans the decision space $Z=Z_{1}\cup Z_{2}$ of the minimum risk binary
classification system $\mathbf{x}^{T}\boldsymbol{\tau}+\boldsymbol{\tau}%
_{0}\overset{\omega_{1}}{\underset{\omega_{2}}{\gtrless}}0$, at which point
the total allowed eigenenergy $\left\Vert \boldsymbol{\tau_{1}-\tau}%
_{2}\right\Vert _{\min_{c}}^{2}$ exhibited by the novel principal eigenaxis
$\boldsymbol{\tau}$ is minimized, so that $\boldsymbol{\tau}$ is the solution
of vector algebra locus equations that represent the geometric loci of a
linear decision boundary $d\left(  \mathbf{x}\right)  =0$ and a pair of
symmetrically positioned linear decision borders $d\left(  \mathbf{x}\right)
=+1$ and $d\left(  \mathbf{x}\right)  =-1$ that jointly partition the decision
space $Z=Z_{1}\cup Z_{2}$ of the minimum risk binary classification system
$\mathbf{x}^{T}\boldsymbol{\tau}+\boldsymbol{\tau}_{0}\overset{\omega
_{1}}{\underset{\omega_{2}}{\gtrless}}0$ into symmetrical decision regions
$Z_{1}$ and $Z_{2}$, where $80\%$ of the training data used to construct the
joint covariance matrix $\mathbf{Q}$ are extreme points.

The linear decision boundary $d\left(  \mathbf{x}\right)  =0$ is black, the
linear decision border $d\left(  \mathbf{x}\right)  =+1$ is red, the linear
decision border $d\left(  \mathbf{x}\right)  =-1$ is blue, and all of the
extreme points $\mathbf{x}_{1_{i\ast}}\mathbf{\sim}$ $p\left(  \mathbf{x}%
;\omega_{1}\right)  $ and $\mathbf{x}_{2_{i\ast}}\mathbf{\sim}$ $p\left(
\mathbf{x};\omega_{2}\right)  $ are enclosed in black circles. The error rate
of the minimum risk binary classification system $\mathbf{x}^{T}%
\boldsymbol{\tau}+\boldsymbol{\tau}_{0}\overset{\omega_{1}}{\underset{\omega
_{2}}{\gtrless}}0$ is $24\%$.%
\begin{figure}[h]%
\centering
\includegraphics[
height=2.0755in,
width=5.5988in
]%
{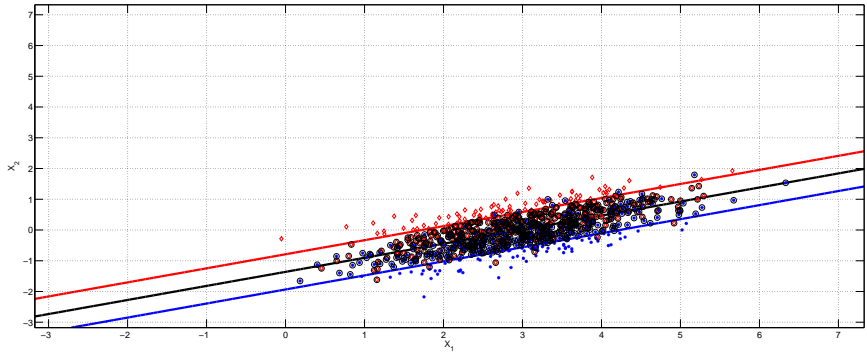}%
\caption{Given a stable and characteristic geometric locus of a novel
principal eigenaxis $\boldsymbol{\tau}=\boldsymbol{\tau}_{1}-\boldsymbol{\tau
}_{2}$ that is based on complete eigenstructures of a joint covariance matrix
$\mathbf{Q}$ and the inverted joint covariance matrix $\mathbf{Q}^{-1}$, the
decision space $Z=Z_{1}\cup Z_{2}$ of the minimum risk binary classification
system $\mathbf{x}^{T}\boldsymbol{\tau}+\boldsymbol{\tau}_{0}%
\protect\overset{\omega_{1}}{\protect\underset{\omega_{2}}{\gtrless}}0$ is
partitioned in a symmetrically balanced manner, at which point all of the
training data add up to sufficient eigenstructures of $\mathbf{Q}$ and
$\mathbf{Q}^{-1}$, such that $80\%$ of the training data are extreme points
$\mathbf{x}_{1_{i\ast}}\mathbf{\sim}$ $p\left(  \mathbf{x};\omega_{1}\right)
$ and $\mathbf{x}_{2_{i\ast}}\mathbf{\sim}$ $p\left(  \mathbf{x};\omega
_{2}\right)  $.}%
\end{figure}

\paragraph{A Low Rank Gram Matrix}

Next, suppose that we let all of the regularization parameters $\left\{
\xi_{i}\right\}  _{i=1}^{N}$ in (\ref{Objective Function}) and all of its
derivatives be equal to $\xi_{i}=\xi=0$, wherein $C=\inf$.

Figure $9$ illustrates that the constrained optimization algorithm being
examined finds an unstable and uncharacteristic geometric locus of a novel
principal eigenaxis $\boldsymbol{\tau}=\boldsymbol{\tau}_{1}-\boldsymbol{\tau
}_{2}$, such that the geometric locus of the novel principal eigenaxis
$\boldsymbol{\tau}=\boldsymbol{\tau_{1}-\tau}_{2}$ does not represent an
eigenaxis of symmetry that spans the decision space $Z=Z_{1}\cup Z_{2}$ of the
binary classification system $\mathbf{x}^{T}\boldsymbol{\tau}+\boldsymbol{\tau
}_{0}\overset{\omega_{1}}{\underset{\omega_{2}}{\gtrless}}0$, at which point
the total allowed eigenenergy $\left\Vert \boldsymbol{\tau}_{1}%
-\boldsymbol{\tau}_{2}\right\Vert ^{2}$ exhibited by the novel principal
eigenaxis $\boldsymbol{\tau}$ is maximized, so that $\boldsymbol{\tau}$ is the
solution of vector algebra locus equations that partition the decision space
$Z=Z_{1}\cup Z_{2}$ of the binary classification system $\mathbf{x}%
^{T}\boldsymbol{\tau}+\boldsymbol{\tau}_{0}\overset{\omega_{1}%
}{\underset{\omega_{2}}{\gtrless}}0$ in an irregular manner, such that the
geometric locus of the linear decision border $d\left(  \mathbf{x}\right)
=-1$ irregularly partitions a collection of feature vectors, where $100\%$ of
the training data used to construct the joint covariance matrix $\mathbf{Q}$
are extreme points.

The linear decision border $d\left(  \mathbf{x}\right)  =-1$ is blue, and all
of the extreme points $\mathbf{x}_{1_{i\ast}}\mathbf{\sim}$ $p\left(
\mathbf{x};\omega_{1}\right)  $ and $\mathbf{x}_{2_{i\ast}}\mathbf{\sim}$
$p\left(  \mathbf{x};\omega_{2}\right)  $ are enclosed in black circles. The
error rate of the binary classification system $\mathbf{x}^{T}\boldsymbol{\tau
}+\boldsymbol{\tau}_{0}\overset{\omega_{1}}{\underset{\omega_{2}}{\gtrless}}0$
is $34\%$.%
\begin{figure}[h]%
\centering
\includegraphics[
height=2.0755in,
width=5.5988in
]%
{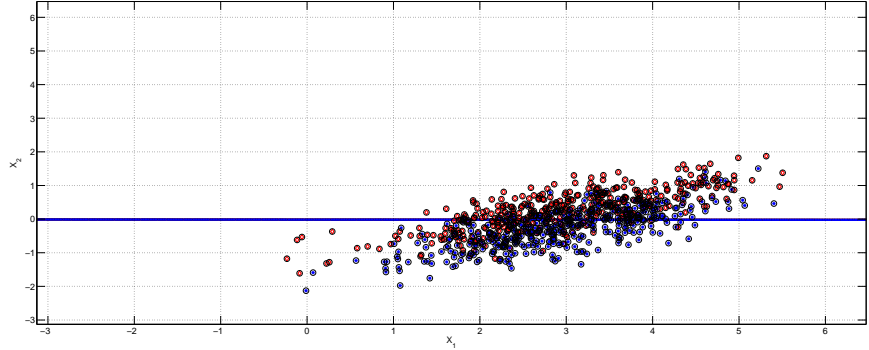}%
\caption{Given an unstable and uncharacteristic geometric locus of a novel
principal eigenaxis $\boldsymbol{\tau}=\boldsymbol{\tau}_{1}-\boldsymbol{\tau
}_{2}$ that is based on an incomplete eigenstructure of a joint covariance
matrix $\mathbf{Q}$ and the inverted joint covariance matrix $\mathbf{Q}^{-1}%
$, the decision space $Z=Z_{1}\cup Z_{2}$ of the binary classification system
$\mathbf{x}^{T}\boldsymbol{\tau}+\boldsymbol{\tau}_{0}\protect\overset{\omega
_{1}}{\protect\underset{\omega_{2}}{\gtrless}}0$ is partitioned in an
irregular and unbalanced manner, at which point all of the training data boil
down to insufficient eigenstructures of $\mathbf{Q}$ and $\mathbf{Q}^{-1}$,
such that $100\%$ of the training data are extreme points $\mathbf{x}%
_{1_{i\ast}}\mathbf{\sim}$ $p\left(  \mathbf{x};\omega_{1}\right)  $ and
$\mathbf{x}_{2_{i\ast}}\mathbf{\sim}$ $p\left(  \mathbf{x};\omega_{2}\right)
$.}%
\end{figure}

\paragraph{A Full Rank Polynomial Kernel Gram Matrix}

In this example, we use a second-order polynomial reproducing kernel
$k_{\mathbf{x}}\left(  \mathbf{s}\right)  =\left(  \mathbf{s}^{T}%
\mathbf{x}+1\right)  ^{2}$, and we let all of the regularization parameters
$\left\{  \xi_{i}\right\}  _{i=1}^{N}$ in (\ref{Objective Function}) and all
of its derivatives be equal to $\xi_{i}=\xi=0.02$, wherein $C=50$.

Figure $10$ illustrates that the constrained optimization algorithm being
examined finds a stable and characteristic geometric locus of a novel
principal eigenaxis $\boldsymbol{\kappa}=\boldsymbol{\kappa}_{1}%
-\boldsymbol{\kappa}_{2}$, such that the geometric locus of the novel
principal eigenaxis $\boldsymbol{\kappa}=\boldsymbol{\kappa}_{1}%
-\boldsymbol{\kappa}_{2}$ represents an eigenaxis of symmetry that spans the
decision space $Z=Z_{1}\cup Z_{2}$ of the minimum risk binary classification
system $k_{\mathbf{s}}\boldsymbol{\kappa}+\boldsymbol{\kappa}_{0}%
\overset{\omega_{1}}{\underset{\omega_{2}}{\gtrless}}0$, at which point the
total allowed eigenenergy $\left\Vert \boldsymbol{\kappa}_{1}%
-\boldsymbol{\kappa}_{2}\right\Vert _{\min_{c}}^{2}$ exhibited by the novel
principal eigenaxis $\boldsymbol{\kappa}$ is minimized, so that
$\boldsymbol{\kappa}$ is the solution of vector algebra locus equations that
represent the geometric loci of a nearly-linear decision boundary $d\left(
\mathbf{s}\right)  =0$ and a pair of symmetrically positioned nearly-linear
decision borders $d\left(  \mathbf{s}\right)  =+1$ and $d\left(
\mathbf{s}\right)  =-1$ that jointly partition the decision space $Z=Z_{1}\cup
Z_{2}$ of the minimum risk binary classification system $k_{\mathbf{s}%
}\boldsymbol{\kappa}+\boldsymbol{\kappa}_{0}\overset{\omega_{1}%
}{\underset{\omega_{2}}{\gtrless}}0$ into symmetrical decision regions $Z_{1}$
and $Z_{2}$, where $78\%$ of the training data used to construct the joint
covariance matrix $\mathbf{Q}$ are extreme points.

The nearly-linear decision boundary $d\left(  \mathbf{s}\right)  =0$ is black,
the nearly-linear decision border $d\left(  \mathbf{s}\right)  =+1$ is red,
the nearly-linear decision border $d\left(  \mathbf{s}\right)  =-1$ is blue,
and all of the extreme points $\mathbf{x}_{1_{i\ast}}\mathbf{\sim}$ $p\left(
\mathbf{x};\omega_{1}\right)  $ and $\mathbf{x}_{2_{i\ast}}\mathbf{\sim}$
$p\left(  \mathbf{x};\omega_{2}\right)  $ are enclosed in black circles. The
error rate of the minimum risk binary classification system $k_{\mathbf{s}%
}\boldsymbol{\kappa}+\boldsymbol{\kappa}_{0}\overset{\omega_{1}%
}{\underset{\omega_{2}}{\gtrless}}0$ is $24\%$.%

\begin{figure}[h]%
\centering
\includegraphics[
height=2.0842in,
width=5.5988in
]%
{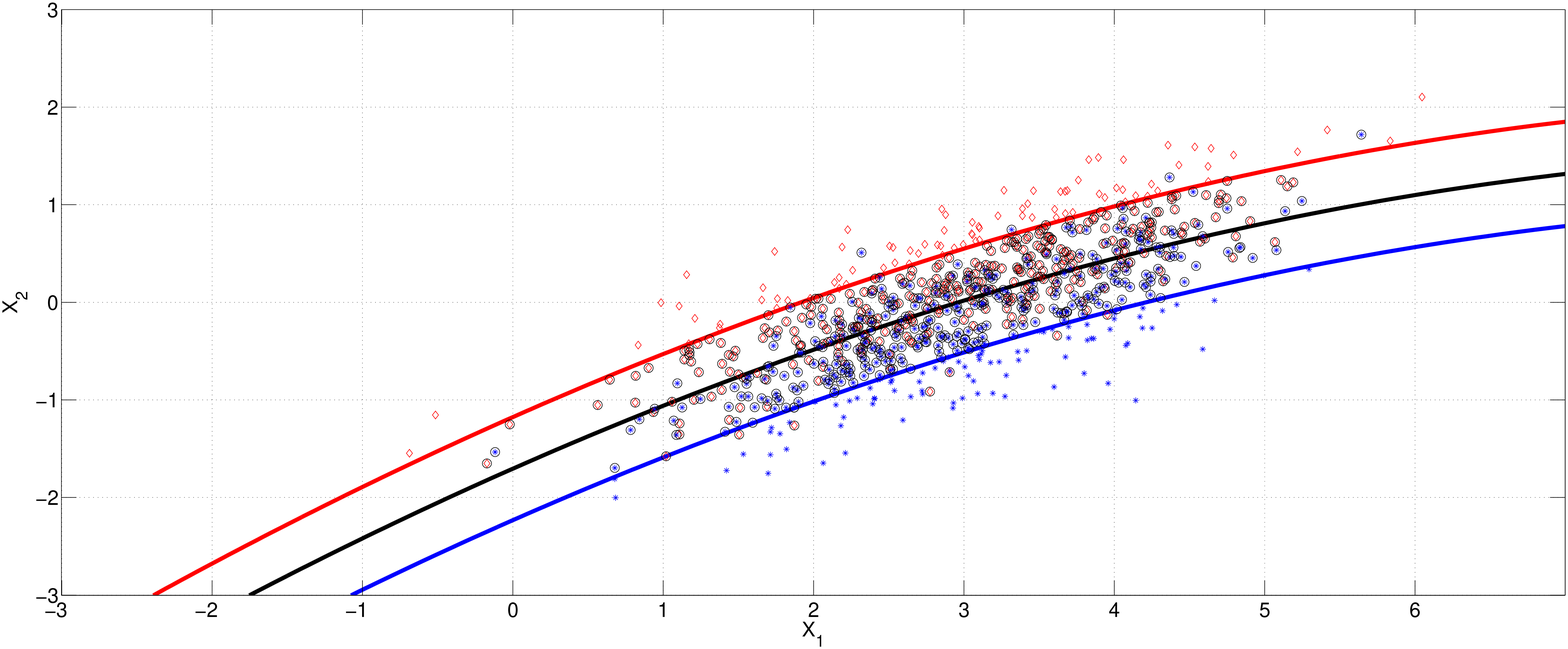}%
\caption{Given a stable and characteristic geometric locus of a novel
principal eigenaxis $\boldsymbol{\kappa}=\boldsymbol{\kappa}_{1}%
-\boldsymbol{\kappa}_{2}$ that is based on a complete eigenstructure of a
joint covariance matrix $\mathbf{Q}$ and the inverted joint covariance matrix
$\mathbf{Q}^{-1}$, the decision space $Z=Z_{1}\cup Z_{2}$ of the minimum risk
binary classification system $k_{\mathbf{s}}\boldsymbol{\kappa}%
+\boldsymbol{\kappa}_{0}\protect\overset{\omega_{1}}{\protect\underset{\omega
_{2}}{\gtrless}}0$ is partitioned in a symmetrically balanced manner, at ehich
point all of the training data add up to sufficient eigenstructures of
$\mathbf{Q}$ and $\mathbf{Q}^{-1}$, such that only $78\%$ of the training data
are extreme points $\mathbf{x}_{1_{i\ast}}\mathbf{\sim}$ $p\left(
\mathbf{x};\omega_{1}\right)  $ and $\mathbf{x}_{2_{i\ast}}\mathbf{\sim}$
$p\left(  \mathbf{x};\omega_{2}\right)  $.}%
\end{figure}

\paragraph{A Low Rank Polynomial Kernel Gram Matrix}

Finally, we use a second-order polynomial reproducing kernel $k_{\mathbf{x}%
}\left(  \mathbf{s}\right)  =\left(  \mathbf{s}^{T}\mathbf{x}+1\right)  ^{2}$,
and we let all of the regularization parameters $\left\{  \xi_{i}\right\}
_{i=1}^{N}$ in (\ref{Objective Function}) and all of its derivatives be equal
to $\xi_{i}=\xi=0$, wherein $C=\inf$.

Figure $11$ illustrates that the constrained optimization algorithm being
examined finds an unstable and uncharacteristic geometric locus of a novel
principal eigenaxis $\boldsymbol{\kappa}=\boldsymbol{\kappa}_{1}%
-\boldsymbol{\kappa}_{2}$, such that the geometric locus of a novel principal
eigenaxis $\boldsymbol{\kappa}=\boldsymbol{\kappa}_{1}-\boldsymbol{\kappa}%
_{2}$ does not represent an eigenaxis of symmetry that spans the decision
space $Z=Z_{1}\cup Z_{2}$ of the binary classification system $k_{\mathbf{s}%
}\boldsymbol{\kappa}+\boldsymbol{\kappa}_{0}\overset{\omega_{1}%
}{\underset{\omega_{2}}{\gtrless}}0$, at which point the total allowed
eigenenergy $\left\Vert \boldsymbol{\kappa}_{1}-\boldsymbol{\kappa}%
_{2}\right\Vert ^{2}$ exhibited by the novel principal eigenaxis
$\boldsymbol{\kappa}$ is maximized, so that $\boldsymbol{\kappa}$ is the
solution of vector algebra locus equations that partition the decision space
$Z=Z_{1}\cup Z_{2}$ of the binary classification system $k_{\mathbf{s}%
}\boldsymbol{\kappa}+\boldsymbol{\kappa}_{0}\overset{\omega_{1}%
}{\underset{\omega_{2}}{\gtrless}}0$ in an irregular manner, such that the
geometric locus of a hyperbolic decision border $d\left(  \mathbf{x}\right)
=-1$ partitions a collection of feature vectors in an unbalanced manner, where
$100\%$ of the training data used to construct the joint covariance matrix
$\mathbf{Q}$ are extreme points.

The hyperbolic decision border $d\left(  \mathbf{s}\right)  =-1$ is blue, and
all of the extreme points $\mathbf{x}_{1_{i\ast}}\mathbf{\sim}$ $p\left(
\mathbf{x};\omega_{1}\right)  $ and $\mathbf{x}_{2_{i\ast}}\mathbf{\sim}$
$p\left(  \mathbf{x};\omega_{2}\right)  $ are enclosed in black circles. The
error rate of the binary classification system $k_{\mathbf{s}}%
\boldsymbol{\kappa}+\boldsymbol{\kappa}_{0}\overset{\omega_{1}%
}{\underset{\omega_{2}}{\gtrless}}0$ is $36\%$.%
\begin{figure}[h]%
\centering
\includegraphics[
height=2.9283in,
width=5.5988in
]%
{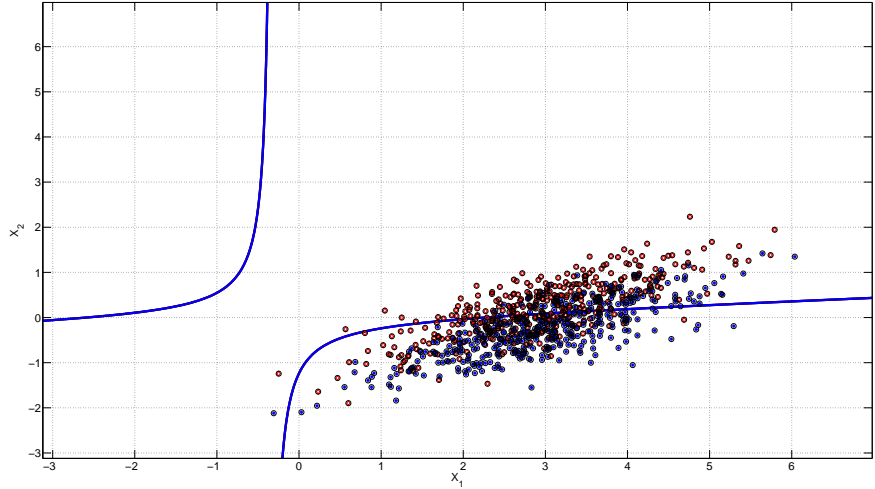}%
\caption{Given an unstable and uncharacteristic geometric locus of a novel
principal eigenaxis $\boldsymbol{\kappa}=\boldsymbol{\kappa}_{1}%
-\boldsymbol{\kappa}_{2}$ that is based on an incomplete eigenstructure of a
joint covariance matrix $\mathbf{Q}$ and the inverted joint covariance matrix
$\mathbf{Q}^{-1}$, the decision space $Z=Z_{1}\cup Z_{2}$ of the binary
classification system $k_{\mathbf{s}}\boldsymbol{\kappa}+\boldsymbol{\kappa
}_{0}\protect\overset{\omega_{1}}{\protect\underset{\omega_{2}}{\gtrless}}0$
is partitioned in an irregular manner, at which point all of the training data
boil down to insufficient eigenstructures of $\mathbf{Q}$ and $\mathbf{Q}%
^{-1}$, such that $100\%$ of the training data are extreme points
$\mathbf{x}_{1_{i\ast}}\mathbf{\sim}$ $p\left(  \mathbf{x};\omega_{1}\right)
$ and $\mathbf{x}_{2_{i\ast}}\mathbf{\sim}$ $p\left(  \mathbf{x};\omega
_{2}\right)  $.}%
\end{figure}

\subsection{Regularization Example Two}

Consider the minimum risk binary classification system $k_{\mathbf{s}%
}\boldsymbol{\kappa}+\boldsymbol{\kappa}_{0}\overset{\omega_{1}%
}{\underset{\omega_{2}}{\gtrless}}0$ for two classes of random vectors that
have similar covariance matrices, such that the covariance matrices for class
$\omega_{1}$ and class $\omega_{2}$ are both given by%
\[
\Sigma_{1}=\Sigma_{2}=\left[
\begin{array}
[c]{cc}%
0.65 & 0.25\\
0.25 & 0.45
\end{array}
\right]  \text{,}%
\]
the mean vector for class $\omega_{1}$ is given by $M_{1}=%
\begin{pmatrix}
1, & 13
\end{pmatrix}
^{T}$ and the mean vector for class $\omega_{2}$ is given by $M_{2}=%
\begin{pmatrix}
6, & 22
\end{pmatrix}
^{T}$, wherein the probability density functions of the two classes of random
vectors determine non-overlapping distributions of the random vectors.

We now consider the generalization performance of the machine learning
algorithm being examined---for the binary classification system outlined
above---for full rank and low rank Gram matrices, as well as full rank and low
rank Polynomial Kernel Gram matrices.

\paragraph{A Full Rank Gram Matrix}

Once more, since the covariance matrices are similar, we can use a linear
kernel in the constrained optimization algorithm that resolves the inverse
problem of binary classification, such that the $N\times d$ matrix
$\widetilde{\mathbf{X}}$ in (\ref{Lagrangian Dual Prob}) is a matrix of $N$
labeled feature vectors $\widetilde{\mathbf{X}}=%
\begin{pmatrix}
y_{1}\mathbf{x}_{1}, & y_{2}\mathbf{x}_{2}, & \ldots, & y_{N}\mathbf{x}_{N}%
\end{pmatrix}
^{T}$. Now suppose that we let all of the regularization parameters $\left\{
\xi_{i}\right\}  _{i=1}^{N}$ in (\ref{Objective Function}) and all of its
derivatives be equal to $\xi_{i}=\xi=0.02$, wherein $C=50$.

Figure $12$ illustrates that the constrained optimization algorithm being
examined finds a stable and characteristic geometric locus of a novel
principal eigenaxis $\boldsymbol{\tau}=\boldsymbol{\tau}_{1}-\boldsymbol{\tau
}_{2}$, such that the geometric locus of the novel principal eigenaxis
$\boldsymbol{\tau}=\boldsymbol{\tau_{1}-\tau}_{2}$ represents an eigenaxis of
symmetry that spans the decision space $Z=Z_{1}\cup Z_{2}$ of the minimum risk
binary classification system $\mathbf{x}^{T}\boldsymbol{\tau}+\boldsymbol{\tau
}_{0}\overset{\omega_{1}}{\underset{\omega_{2}}{\gtrless}}0$, at which point
the total allowed eigenenergy $\left\Vert \boldsymbol{\tau}_{1}%
-\boldsymbol{\tau}_{2}\right\Vert _{\min_{c}}^{2}$ exhibited by the novel
principal eigenaxis $\boldsymbol{\tau}$ is minimized, so that
$\boldsymbol{\tau}$ is the solution of vector algebra locus equations that
represent the geometric loci of a linear decision boundary $d\left(
\mathbf{x}\right)  =0$ and a pair of symmetrically positioned linear decision
borders $d\left(  \mathbf{x}\right)  =+1$ and $d\left(  \mathbf{x}\right)
=-1$ that jointly partition the decision space $Z=Z_{1}\cup Z_{2}$ of the
minimum risk binary classification system $\mathbf{x}^{T}\boldsymbol{\tau
}+\boldsymbol{\tau}_{0}\overset{\omega_{1}}{\underset{\omega_{2}}{\gtrless}}0$
into symmetrical decision regions $Z_{1}$ and $Z_{2}$, where only $0.3\%$ of
the training data used to construct the joint covariance matrix $\mathbf{Q}$
are extreme points.

The linear decision boundary $d\left(  \mathbf{x}\right)  =0$ is black, the
linear decision border $d\left(  \mathbf{x}\right)  =+1$ is red, the linear
decision border $d\left(  \mathbf{x}\right)  =-1$ is blue, and all of the
extreme points $\mathbf{x}_{1_{i\ast}}\mathbf{\sim}$ $p\left(  \mathbf{x}%
;\omega_{1}\right)  $ and $\mathbf{x}_{2_{i\ast}}\mathbf{\sim}$ $p\left(
\mathbf{x};\omega_{2}\right)  $ are enclosed in black circles. The error rate
of the minimum risk binary classification system $\mathbf{x}^{T}%
\boldsymbol{\tau}+\boldsymbol{\tau}_{0}\overset{\omega_{1}}{\underset{\omega
_{2}}{\gtrless}}0$ is $0\%$.%
\begin{figure}[h]%
\centering
\includegraphics[
height=2.1361in,
width=5.5988in
]%
{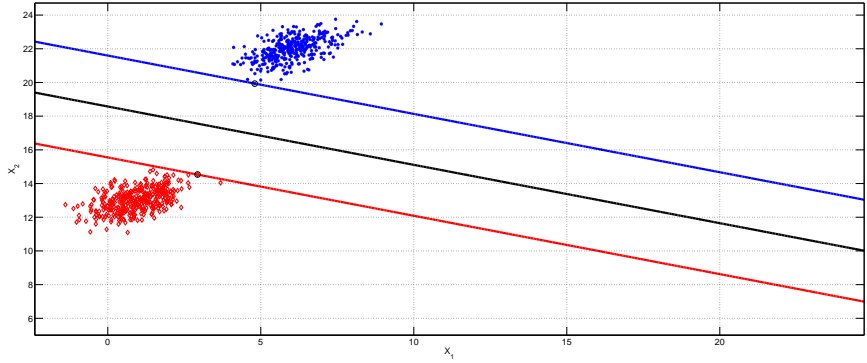}%
\caption{Given a stable and characteristic geometric locus of a novel
principal eigenaxis $\boldsymbol{\tau}=\boldsymbol{\tau}_{1}-\boldsymbol{\tau
}_{2}$ that is based on a complete eigenstructure of a joint covariance matrix
$\mathbf{Q}$ and the inverted joint covariance matrix $\mathbf{Q}^{-1}$, the
decision space $Z=Z_{1}\cup Z_{2}$ of the minimum risk binary classification
system $\mathbf{x}^{T}\boldsymbol{\tau}+\boldsymbol{\tau}_{0}%
\protect\overset{\omega_{1}}{\protect\underset{\omega_{2}}{\gtrless}}0$ is
partitioned in a symmetrically balanced manner, at which point all of the
training data add up to sufficient eigenstructures of $\mathbf{Q}$ and
$\mathbf{Q}^{-1}$, such that only $0.3\%$ of the training data are extreme
points $\mathbf{x}_{1_{i\ast}}\mathbf{\sim}$ $p\left(  \mathbf{x};\omega
_{1}\right)  $ and $\mathbf{x}_{2_{i\ast}}\mathbf{\sim}$ $p\left(
\mathbf{x};\omega_{2}\right)  $.}%
\end{figure}

\paragraph{A Low Rank Gram Matrix}

Next, suppose that we let all of the regularization parameters $\left\{
\xi_{i}\right\}  _{i=1}^{N}$ in (\ref{Objective Function}) and all of its
derivatives be equal to $\xi_{i}=\xi=0$, wherein $C=\inf$.

Figure $13$ illustrates that the constrained optimization algorithm being
examined finds an unstable and uncharacteristic geometric locus of a novel
principal eigenaxis $\boldsymbol{\tau}=\boldsymbol{\tau}_{1}-\boldsymbol{\tau
}_{2}$, such that the geometric locus of the novel principal eigenaxis
$\boldsymbol{\tau}=\boldsymbol{\tau_{1}-\tau}_{2}$ does not represent an
eigenaxis of symmetry that spans the decision space $Z=Z_{1}\cup Z_{2}$ of the
binary classification system $\mathbf{x}^{T}\boldsymbol{\tau}+\boldsymbol{\tau
}_{0}\overset{\omega_{1}}{\underset{\omega_{2}}{\gtrless}}0$, at which point
the total allowed eigenenergy $\left\Vert \boldsymbol{\tau}_{1}%
-\boldsymbol{\tau}_{2}\right\Vert ^{2}$ exhibited by the novel principal
eigenaxis $\boldsymbol{\tau}$ is maximized, so that $\boldsymbol{\tau}$ is the
solution of vector algebra locus equations that partition the decision space
$Z=Z_{1}\cup Z_{2}$ of the binary classification system $\mathbf{x}%
^{T}\boldsymbol{\tau}+\boldsymbol{\tau}_{0}\overset{\omega_{1}%
}{\underset{\omega_{2}}{\gtrless}}0$ in an irregular manner, such that the
geometric locus of the linear decision border $d\left(  \mathbf{x}\right)
=-1$ partitions a collection of feature vectors where $100\%$ of the training
data used to construct the joint covariance matrix $\mathbf{Q}$ are extreme points.

The linear decision border $d\left(  \mathbf{x}\right)  =-1$ is blue, and all
of the extreme points $\mathbf{x}_{1_{i\ast}}\mathbf{\sim}$ $p\left(
\mathbf{x};\omega_{1}\right)  $ and $\mathbf{x}_{2_{i\ast}}\mathbf{\sim}$
$p\left(  \mathbf{x};\omega_{2}\right)  $ are enclosed in black circles.
Moreover, even though \emph{all} of the training data are extreme points, the
error rate of the binary classification system $\mathbf{x}^{T}\boldsymbol{\tau
}+\boldsymbol{\tau}_{0}\overset{\omega_{1}}{\underset{\omega_{2}}{\gtrless}}0$
is $0\%$.%
\begin{figure}[h]%
\centering
\includegraphics[
height=2.0868in,
width=5.5988in
]%
{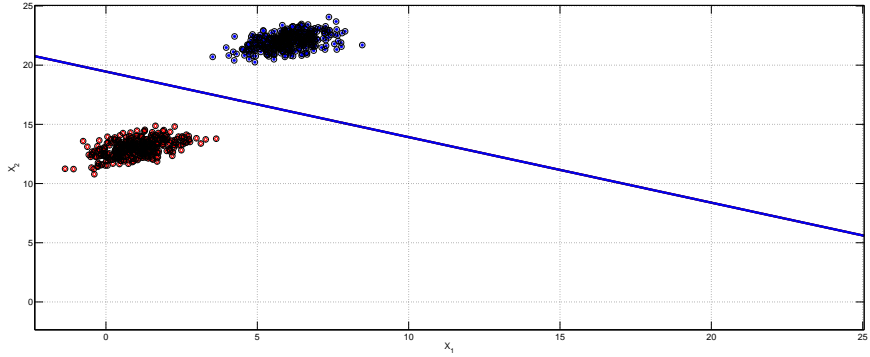}%
\caption{Given an unstable and uncharacteristic geometric locus of a novel
principal eigenaxis $\boldsymbol{\tau}=\boldsymbol{\tau}_{1}-\boldsymbol{\tau
}_{2}$ that is based on an incomplete eigenstructure of a joint covariance
matrix $\mathbf{Q}$ and the inverted joint covariance matrix $\mathbf{Q}^{-1}%
$, the decision space $Z=Z_{1}\cup Z_{2}$ of the binary classification system
$\mathbf{x}^{T}\boldsymbol{\tau}+\boldsymbol{\tau}_{0}\protect\overset{\omega
_{1}}{\protect\underset{\omega_{2}}{\gtrless}}0$ is partitioned in an
irregular manner, at which point all of the training data boil down to
insufficient eigenstructures of $\mathbf{Q}$ and $\mathbf{Q}^{-1}$, such that
$100\%$ of the training data are extreme points $\mathbf{x}_{1_{i\ast}%
}\mathbf{\sim}$ $p\left(  \mathbf{x};\omega_{1}\right)  $ and $\mathbf{x}%
_{2_{i\ast}}\mathbf{\sim}$ $p\left(  \mathbf{x};\omega_{2}\right)  $.}%
\end{figure}

\paragraph{A Full Rank Polynomial Kernel Gram Matrix}

In this example, we use a second-order polynomial reproducing kernel
$k_{\mathbf{x}}\left(  \mathbf{s}\right)  =\left(  \mathbf{s}^{T}%
\mathbf{x}+1\right)  ^{2}$, and we let all of the regularization parameters
$\left\{  \xi_{i}\right\}  _{i=1}^{N}$ in (\ref{Objective Function}) and all
of its derivatives be equal to $\xi_{i}=\xi=0.02$, wherein $C=50$.

Figure $14$ illustrates that the constrained optimization algorithm being
examined finds a stable and characteristic geometric locus of a novel
principal eigenaxis $\boldsymbol{\kappa}=\boldsymbol{\kappa}_{1}%
-\boldsymbol{\kappa}_{2}$, such that the geometric locus of the novel
principal eigenaxis $\boldsymbol{\kappa}=\boldsymbol{\kappa}_{1}%
-\boldsymbol{\kappa}_{2}$ represents an eigenaxis of symmetry that spans the
decision space $Z=Z_{1}\cup Z_{2}$ of the minimum risk binary classification
system $k_{\mathbf{s}}\boldsymbol{\kappa}+\boldsymbol{\kappa}_{0}%
\overset{\omega_{1}}{\underset{\omega_{2}}{\gtrless}}0$, at which point the
total allowed eigenenergy $\left\Vert \boldsymbol{\kappa}_{1}%
-\boldsymbol{\kappa}_{2}\right\Vert _{\min_{c}}^{2}$ exhibited by the novel
principal eigenaxis $\boldsymbol{\kappa}$ is minimized, so that
$\boldsymbol{\kappa}$ is the solution of vector algebra locus equations that
represent the geometric loci of a nearly-linear decision boundary $d\left(
\mathbf{s}\right)  =0$ and a pair of symmetrically positioned nearly-linear
decision borders $d\left(  \mathbf{s}\right)  =+1$ and $d\left(
\mathbf{s}\right)  =-1$ that jointly partition the decision space $Z=Z_{1}\cup
Z_{2}$ of the minimum risk binary classification system $k_{\mathbf{s}%
}\boldsymbol{\kappa}+\boldsymbol{\kappa}_{0}\overset{\omega_{1}%
}{\underset{\omega_{2}}{\gtrless}}0$ into symmetrical decision regions $Z_{1}$
and $Z_{2}$, where $0.3\%$ of the training data used to construct the joint
covariance matrix $\mathbf{Q}$ are extreme points.

The nearly-linear decision boundary $d\left(  \mathbf{s}\right)  =0$ is black,
the nearly-linear decision border $d\left(  \mathbf{s}\right)  =+1$ is red,
the nearly-linear decision border $d\left(  \mathbf{s}\right)  =-1$ is blue,
and all of the extreme points $\mathbf{x}_{1_{i\ast}}\mathbf{\sim}$ $p\left(
\mathbf{x};\omega_{1}\right)  $ and $\mathbf{x}_{2_{i\ast}}\mathbf{\sim}$
$p\left(  \mathbf{x};\omega_{2}\right)  $ are enclosed in black circles. The
error rate of the minimum risk binary classification system $k_{\mathbf{s}%
}\boldsymbol{\kappa}+\boldsymbol{\kappa}_{0}\overset{\omega_{1}%
}{\underset{\omega_{2}}{\gtrless}}0$ is $0\%$.%
\begin{figure}[h]%
\centering
\includegraphics[
height=2.0868in,
width=5.5988in
]%
{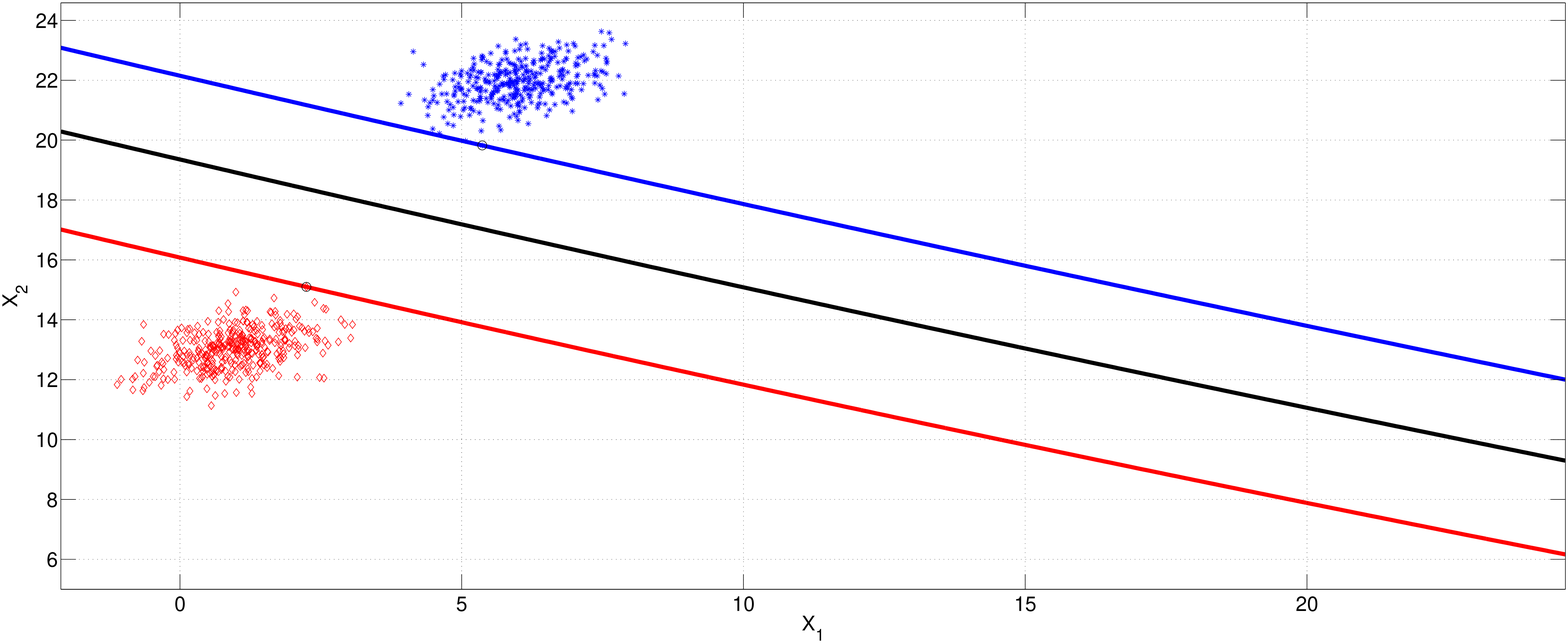}%
\caption{Given a stable and characteristic geometric locus of a novel
principal eigenaxis $\boldsymbol{\kappa}=\boldsymbol{\kappa}_{1}%
-\boldsymbol{\kappa}_{2}$ that is based on a complete eigenstructure of a
joint covariance matrix $\mathbf{Q}$ and the inverted joint covariance matrix
$\mathbf{Q}^{-1}$, the decision space $Z=Z_{1}\cup Z_{2}$ of the minimum risk
binary classification system $k_{\mathbf{s}}\boldsymbol{\kappa}%
+\boldsymbol{\kappa}_{0}\protect\overset{\omega_{1}}{\protect\underset{\omega
_{2}}{\gtrless}}0$ is partitioned in a symmetrically balanced manner, at which
point all of the training data add up to sufficient eigenstructures of
$\mathbf{Q}$ and $\mathbf{Q}^{-1}$, such that only $0.3\%$ of the training
data are extreme points $\mathbf{x}_{1_{i\ast}}\mathbf{\sim}$ $p\left(
\mathbf{x};\omega_{1}\right)  $ and $\mathbf{x}_{2_{i\ast}}\mathbf{\sim}$
$p\left(  \mathbf{x};\omega_{2}\right)  $.}%
\end{figure}

\paragraph{A Low Rank Polynomial Kernel Gram Matrix}

Finally, we use a second-order polynomial reproducing kernel $k_{\mathbf{x}%
}\left(  \mathbf{s}\right)  =\left(  \mathbf{s}^{T}\mathbf{x}+1\right)  ^{2}$,
and we let all of the regularization parameters $\left\{  \xi_{i}\right\}
_{i=1}^{N}$ in (\ref{Objective Function}) and all of its derivatives be equal
to $\xi_{i}=\xi=0$, wherein $C=\inf$.

Figure $15$ illustrates that the constrained optimization algorithm being
examined finds an unstable and uncharacteristic geometric locus of a novel
principal eigenaxis $\boldsymbol{\kappa}=\boldsymbol{\kappa}_{1}%
-\boldsymbol{\kappa}_{2}$, such that the geometric locus of the novel
principal eigenaxis $\boldsymbol{\kappa}=\boldsymbol{\kappa}_{1}%
-\boldsymbol{\kappa}_{2}$ does not represent an eigenaxis of symmetry that
spans the decision space $Z=Z_{1}\cup Z_{2}$ of the binary classification
system $k_{\mathbf{s}}\boldsymbol{\kappa}+\boldsymbol{\kappa}_{0}%
\overset{\omega_{1}}{\underset{\omega_{2}}{\gtrless}}0$, at which point the
total allowed eigenenergy $\left\Vert \boldsymbol{\kappa}_{1}%
-\boldsymbol{\kappa}_{2}\right\Vert ^{2}$ exhibited by the novel principal
eigenaxis $\boldsymbol{\kappa}$ is maximized, so that $\boldsymbol{\kappa}$ is
the solution of vector algebra locus equations that partition the decision
space $Z=Z_{1}\cup Z_{2}$ of the binary classification system $k_{\mathbf{s}%
}\boldsymbol{\kappa}+\boldsymbol{\kappa}_{0}\overset{\omega_{1}%
}{\underset{\omega_{2}}{\gtrless}}0$ in an irregular manner, such that the
geometric locus of a hyperbolic decision border $d\left(  \mathbf{x}\right)
=-1$ partitions a collection of feature vectors, where $100\%$ of the training
data used to construct the joint covariance matrix $\mathbf{Q}$ are extreme points.

The hyperbolic decision border $d\left(  \mathbf{s}\right)  =-1$ is blue, and
all of the extreme points $\mathbf{x}_{1_{i\ast}}\mathbf{\sim}$ $p\left(
\mathbf{x};\omega_{1}\right)  $ and $\mathbf{x}_{2_{i\ast}}\mathbf{\sim}$
$p\left(  \mathbf{x};\omega_{2}\right)  $ are enclosed in black circles. Yet
again, even though all of the training data are extreme points, the error rate
of the binary classification system $k_{\mathbf{s}}\boldsymbol{\kappa
}+\boldsymbol{\kappa}_{0}\overset{\omega_{1}}{\underset{\omega_{2}}{\gtrless}%
}0$ is $0\%$.%
\begin{figure}[h]%
\centering
\includegraphics[
height=2.0868in,
width=5.5988in
]%
{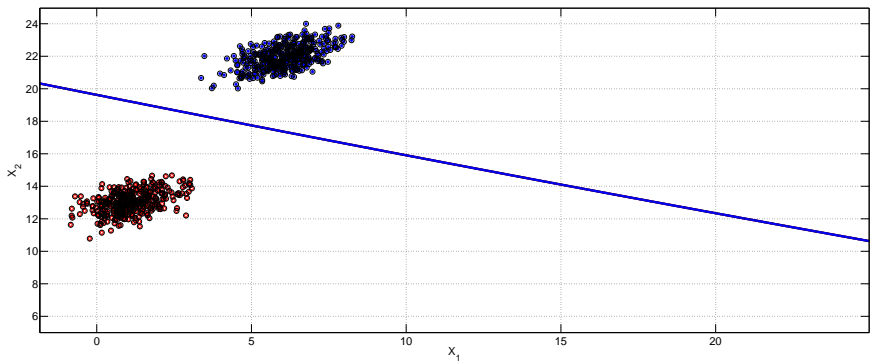}%
\caption{Given an unstable and uncharacteristic geometric locus of a novel
principal eigenaxis $\boldsymbol{\kappa}=\boldsymbol{\kappa}_{1}%
-\boldsymbol{\kappa}_{2}$ that is based on an incomplete eigenstructure of a
joint covariance matrix $\mathbf{Q}$ and the inverted joint covariance matrix
$\mathbf{Q}^{-1}$, the decision space $Z=Z_{1}\cup Z_{2}$ of the binary
classification system $k_{\mathbf{s}}\boldsymbol{\kappa}+\boldsymbol{\kappa
}_{0}\protect\overset{\omega_{1}}{\protect\underset{\omega_{2}}{\gtrless}}0$
is partitioned in an irregular manner, at which point all of the training data
boil down to insufficient eigenstructures of $\mathbf{Q}$ and $\mathbf{Q}%
^{-1}$, such that $100\%$ of the training data are extreme points
$\mathbf{x}_{1_{i\ast}}\mathbf{\sim}$ $p\left(  \mathbf{x};\omega_{1}\right)
$ and $\mathbf{x}_{2_{i\ast}}\mathbf{\sim}$ $p\left(  \mathbf{x};\omega
_{2}\right)  $.}%
\end{figure}

We now turn our attention to statistical relations between feature
vectors---inside reproducing kernel Hilbert spaces---that determine pointwise
covariance statistics, joint covariance statistics\ and conditional
distributions for individual feature vectors.

\section{\label{Section 14}Statistical Relations Inside Hilbert Spaces}

We have discovered that the overall structure and behavior and properties of
any given minimum risk binary classification system $k_{\mathbf{s}%
}\boldsymbol{\kappa}+\boldsymbol{\kappa}_{0}\overset{\omega_{1}%
}{\underset{\omega_{2}}{\gtrless}}0$---which is found by the constrained
optimization algorithm being examined---is determined by elegant statistical
relations and complex statistical interconnections between all of the
principal eigenaxis components and likelihood components that lie on both
sides of the novel principal eigenaxes $\boldsymbol{\psi}$ and
$\boldsymbol{\kappa}$---in accordance with algebraic and geometrical and
statistical conditions expressed by certain KKT conditions in (\ref{KKT 1}) -
(\ref{KKT 5}) that $\boldsymbol{\psi}$ and $\boldsymbol{\kappa}$ are subject to.

We identify certain statistical relations and interconnections between all of
the dual components that lie on both sides of $\boldsymbol{\psi}$ and
$\boldsymbol{\kappa}$ by using statistical relations that we have
devised---based on statistical relations and interconnections between random
vectors inside Hilbert spaces and reproducing kernel Hilbert spaces---that
determine pointwise covariance statistics, joint covariance statistics\ and
conditional distributions for individual random points. The original versions
of these statistical relations can be found in our working papers
\citep{Reeves2015resolving}
and
\citep{Reeves2018design}%
.

Axiom \ref{Statistical Structures of Hilbert Space Axiom} expresses
statistical relations between random vectors in Hilbert spaces and reproducing
kernel Hilbert spaces that determine pointwise covariance statistics, joint
covariance statistics\ and conditional distributions for individual random points.

\begin{axiom}
\label{Statistical Structures of Hilbert Space Axiom}Take any given random
vectors $\mathbf{x\in}$ $%
\mathbb{R}
^{d}$ and $\mathbf{y\in}$ $%
\mathbb{R}
^{d}$ in Hilbert space that are generated by certain probability density
functions $\mathbf{x\sim}$ $p\left(  \mathbf{x}\right)  $ and $\mathbf{y\sim}$
$p\left(  \mathbf{y}\right)  $, such that all of the $d$ random point
coordinates $\left\{  \left\Vert \mathbf{x}\right\Vert \cos\mathbb{\alpha
}_{\mathbf{e}_{i}\mathbf{x}}\right\}  _{i=1}^{d}$ on the locus of $\mathbf{x}$
and all of the $d$ random point coordinates $\left\{  \left\Vert
\mathbf{y}\right\Vert \cos\mathbb{\alpha}_{\mathbf{e}_{i}\mathbf{y}}\right\}
_{i=1}^{d}$ on the locus of $\mathbf{y}$ are random variables that have
expected values and covariances.

The random vectors $\mathbf{x}$ and $\mathbf{y}$ satisfy the law of cosines%
\[
\left\Vert \mathbf{x-y}\right\Vert ^{2}=\left\Vert \mathbf{x}\right\Vert
^{2}+\left\Vert \mathbf{y}\right\Vert ^{2}-2\left\Vert \mathbf{x}\right\Vert
\left\Vert \mathbf{y}\right\Vert \cos\theta
\]
which reduces to the following relation between the random vectors
$\mathbf{x}$ and $\mathbf{y}$%
\begin{align*}
\left\Vert \mathbf{x}\right\Vert \left\Vert \mathbf{y}\right\Vert \cos\theta
&  =x_{1}y_{1}+x_{2}y_{2}+\cdots+x_{d}y_{d}\\
&  =\mathbf{x}^{T}\mathbf{y=y}^{T}\mathbf{x}\text{,}%
\end{align*}
at which point the relation $\left\Vert \mathbf{x}\right\Vert \left\Vert
\mathbf{y}\right\Vert \cos\theta$ between the random vectors $\mathbf{x}$ and
$\mathbf{y}$ is correlated with the distance $\left\Vert \mathbf{x-y}%
\right\Vert $ between the loci of the random points $\mathbf{x}$ and
$\mathbf{y}$.

Thereby, the relation $\left\Vert \mathbf{x}\right\Vert \left\Vert
\mathbf{y}\right\Vert \cos\theta$ between the random vectors $\mathbf{x}$ and
$\mathbf{y}$ determines a pointwise covariance statistic that represents joint
variations between all of the $d$ random point coordinates on the locus of
$\mathbf{x}$%
\[
\mathbf{x=}%
\begin{pmatrix}
\left\Vert \mathbf{x}\right\Vert \cos\mathbb{\alpha}_{\mathbf{e}_{1}%
\mathbf{x}}, & \left\Vert \mathbf{x}\right\Vert \cos\mathbb{\alpha
}_{\mathbf{e}_{2}\mathbf{x}}, & \cdots, & \left\Vert \mathbf{x}\right\Vert
\cos\mathbb{\alpha}_{\mathbf{e}_{d}\mathbf{x}}%
\end{pmatrix}
^{T}%
\]
and all of the $d$ random point coordinates on the locus of $\mathbf{y}$%
\[
\mathbf{y=}%
\begin{pmatrix}
\left\Vert \mathbf{y}\right\Vert \cos\mathbb{\alpha}_{\mathbf{e}_{1}%
\mathbf{y}}, & \left\Vert \mathbf{y}\right\Vert \cos\mathbb{\alpha
}_{\mathbf{e}_{2}\mathbf{y}}, & \cdots, & \left\Vert \mathbf{y}\right\Vert
\cos\mathbb{\alpha}_{\mathbf{e}_{d}\mathbf{y}}%
\end{pmatrix}
^{T}\text{,}%
\]
so that the covariance of the random variables $\left\Vert \mathbf{x}%
\right\Vert \cos\alpha_{\mathbf{e}_{i}\mathbf{x}}$ and $\left\Vert
\mathbf{y}\right\Vert \cos\alpha_{\mathbf{e}_{i}\mathbf{y}}$ on the loci of
$\mathbf{x}$ and $\mathbf{y}$ is a function of the distance%
\[
\left\vert \left\Vert \mathbf{x}\right\Vert \cos\mathbb{\alpha}_{\mathbf{e}%
_{i}\mathbf{x}}-\left\Vert \mathbf{y}\right\Vert \cos\mathbb{\alpha
}_{\mathbf{e}_{i}\mathbf{y}}\right\vert
\]
between the random point coordinates $\left\Vert \mathbf{x}\right\Vert
\cos\mathbb{\alpha}_{\mathbf{e}_{i}\mathbf{x}}$ and $\left\Vert \mathbf{y}%
\right\Vert \cos\mathbb{\alpha}_{\mathbf{e}_{i}\mathbf{y}}$ on the orthonormal
coordinate axis $\mathbf{e}_{i}$.

Furthermore, the relation $\left\Vert \mathbf{x}\right\Vert \left\Vert
\mathbf{y}\right\Vert \cos\theta$ between the random vectors $\mathbf{x}$ and
$\mathbf{y}$ represents the length $\left\Vert \mathbf{x}\right\Vert $ of the
random vector $\mathbf{x}$ times the signed magnitude $\left\Vert
\mathbf{y}\right\Vert \cos\theta$ of the vector projection of the random
vector $\mathbf{y}$ onto the random vector $\mathbf{x}$%
\begin{align}
\mathbf{x}^{T}\mathbf{y}  &  \mathbf{=}\left\Vert \mathbf{x}\right\Vert
\times\left[  \left\Vert \mathbf{y}\right\Vert \cos\theta\right]
\tag{14.1}\label{Vector Projection}\\
&  \triangleq\left\Vert \mathbf{x}\right\Vert \operatorname{comp}%
_{\overrightarrow{\mathbf{x}}}\left(  \overrightarrow{\mathbf{y}}\right)
\text{,}\nonumber
\end{align}
such that the signed magnitude $\left\Vert \mathbf{y}\right\Vert \cos\theta
$---expressed by $\operatorname{comp}_{\overrightarrow{\mathbf{x}}}\left(
\overrightarrow{\mathbf{y}}\right)  $---is a random variable that is negative
if $\pi/2<\theta\leq\pi$, at which point the algebraic and geometric
relationship $\operatorname{comp}_{\overrightarrow{\mathbf{x}}}\left(
\overrightarrow{\mathbf{y}}\right)  $ between the locus of $\mathbf{x}$ and
the locus of $\mathbf{y}$ determines an implicit random vector, such that the
magnitude and the direction of the implicit random vector is determined by the
signed magnitude $\left\Vert \mathbf{y}\right\Vert \cos\theta$ along the locus
of the random vector $\mathbf{x}$.

Moreover, the signed magnitude $\operatorname{comp}%
_{\overrightarrow{\mathbf{x}}}\left(  \overrightarrow{\mathbf{y}}\right)  $ of
the vector projection of the random vector $\mathbf{y}$ onto the random vector
$\mathbf{x}$ represents a distribution of the random vector $\mathbf{y}$ that
is conditional on the relationship $\left\Vert \mathbf{x}\right\Vert
\times\left[  \left\Vert \mathbf{y}\right\Vert \cos\theta\right]  $, such that
the signed magnitude $\operatorname{comp}_{\overrightarrow{\mathbf{x}}}\left(
\overrightarrow{\mathbf{y}}\right)  $ represents a certain portion $\left\Vert
\mathbf{y}\right\Vert \cos\theta$ of the random vector $\mathbf{y}$ that is
distributed along the locus of the random vector $\mathbf{x}$, at which point
the distribution of the random vector $\mathbf{y}$ is either positive or negative.

Now let $k_{\mathbf{x}}\mathbf{\ }$and $k_{\mathbf{y}}$ be any given random
vectors in a reproducing kernel Hilbert space, such that a reproducing kernel
in Hilbert space is a certain reproducing kernel $k_{\mathbf{x}}\left(
\mathbf{s}\right)  $ that is defined on random vectors $\mathbf{x\in%
\mathbb{R}
}^{d}$, so that $k_{\mathbf{x}}$ and $k_{\mathbf{y}}$ are reproducing kernels
for the random points $\mathbf{x\in}$ $%
\mathbb{R}
^{d}$ and $\mathbf{y\in}$ $%
\mathbb{R}
^{d}$, at which point all of the $d$ random point coordinates $\left\{
\left\Vert k_{\mathbf{x}}\right\Vert \cos\mathbb{\alpha}_{\mathbf{e}%
_{i}k_{\mathbf{x}}}\right\}  _{i=1}^{d}$ on the locus of $k_{\mathbf{x}}$ and
all of the $d$ random point coordinates $\left\{  \left\Vert k_{\mathbf{y}%
}\right\Vert \cos\mathbb{\alpha}_{\mathbf{e}_{i}k_{\mathbf{y}}}\right\}
_{i=1}^{d}$ on the locus of $k_{\mathbf{y}}$ are random variables that have
expected values and covariances.

Since the random vectors $k_{\mathbf{x}}$ and $k_{\mathbf{y}}$ satisfy the law
of cosines%
\[
\left\Vert k_{\mathbf{x}}\mathbf{-}k_{\mathbf{y}}\right\Vert ^{2}=\left\Vert
k_{\mathbf{x}}\right\Vert ^{2}+\left\Vert k_{\mathbf{y}}\right\Vert
^{2}-2\left\Vert k_{\mathbf{x}}\right\Vert \left\Vert k_{\mathbf{y}%
}\right\Vert \cos\theta\text{,}%
\]
it follows that the relation $\left\Vert k_{\mathbf{x}}\right\Vert \left\Vert
k_{\mathbf{y}}\right\Vert \cos\theta_{k_{\mathbf{x}}k_{\mathbf{y}}}$ between
the reproducing kernels for the random points $\mathbf{x\in}$ $%
\mathbb{R}
^{d}$ and $\mathbf{y\in}$ $%
\mathbb{R}
^{d}$ determines a pointwise covariance statistic for the random points
$k_{\mathbf{x}}$ and $k_{\mathbf{y}}$, along with a distribution of the random
vector $k_{\mathbf{y}}$ that is conditional on the relationship $\left\Vert
k_{\mathbf{x}}\right\Vert \times\left[  \left\Vert k_{\mathbf{y}}\right\Vert
\cos\theta\right]  $, such that the signed magnitude $\operatorname{comp}%
_{\overrightarrow{k_{\mathbf{x}}}}\left(  \overrightarrow{k_{\mathbf{y}}%
}\right)  $ represents a certain portion $\left\Vert k_{\mathbf{y}}\right\Vert
\cos\theta$ of the random vector $k_{\mathbf{y}}$ that is distributed along
the locus of the random vector $k_{\mathbf{x}}$, at which point the
distribution of the random vector $k_{\mathbf{y}}$ is either positive or negative.

Therefore, take the $N\times N$ kernel matrix $\mathbf{Q}$ for any given
collection of random vectors $\left\{  k_{\mathbf{x}_{j}}\right\}  _{j=1}^{N}$%
\[
\mathbf{Q}=%
\begin{pmatrix}
\left\Vert k_{\mathbf{x}_{1}}\right\Vert \left\Vert k_{\mathbf{x}_{1}%
}\right\Vert \cos\theta_{k_{\mathbf{x_{1}}}k_{\mathbf{x}_{1}}} & \cdots &
\left\Vert k_{\mathbf{x}_{1}}\right\Vert \left\Vert k_{\mathbf{x}_{N}%
}\right\Vert \cos\theta_{k_{\mathbf{x}_{1}}k_{\mathbf{x}_{N}}}\\
\vdots & \ddots & \vdots\\
\left\Vert k_{\mathbf{x}_{N}}\right\Vert \left\Vert k_{\mathbf{x}_{1}%
}\right\Vert \cos\theta_{k_{\mathbf{x}_{N}}k_{\mathbf{x}_{1}}} & \cdots &
\left\Vert k_{\mathbf{x}_{N}}\right\Vert \left\Vert k_{\mathbf{x}_{N}%
}\right\Vert \cos\theta_{k_{\mathbf{x}_{N}}k_{\mathbf{x}_{N}}}%
\end{pmatrix}
\text{,}%
\]
such that each element $\mathbf{Q}\left(  i,j\right)  $ of $\mathbf{Q}$
represents joint variations between all of the $d$ random point coordinates on
the loci of two random vectors $k_{\mathbf{x}_{i}}$ and $k_{\mathbf{x}_{j}}$
by the relation $\left\Vert k_{\mathbf{x}_{i}}\right\Vert \left\Vert
k_{\mathbf{x}_{j}}\right\Vert \cos\theta_{k_{\mathbf{x}_{i}}k_{\mathbf{x}_{j}%
}}$, so that the signed magnitude $\operatorname{comp}%
_{\overrightarrow{k_{\mathbf{x}_{i}}}}\left(  \overrightarrow{k_{\mathbf{x}%
_{j}}}\right)  $ of the random vector $k_{\mathbf{x}_{j}}$ along the locus of
the random vector $k_{\mathbf{x}_{i}}$ determines a certain portion
$\left\Vert k_{\mathbf{x}_{j}}\right\Vert \cos\theta_{k_{\mathbf{x}_{i}%
}k_{\mathbf{x}_{j}}}$ of the random vector $k_{\mathbf{x}_{j}}$ that is
distributed along the locus of the random vector $k_{\mathbf{x}_{i}}$, at
which point the distribution of the random vector $k_{\mathbf{x}_{j}}$ is
either positive or negative.

Thereby, the kernel matrix $\mathbf{Q}$ determines an estimate of a covariance
matrix, such that each row $i$ of $\mathbf{Q}$ represents joint variations
between all of the $d$ random point coordinates on the locus of a random
vector $k_{\mathbf{x}_{i}}$ and all of the $d$ random point coordinates of
each of the random vectors $k_{\mathbf{x}_{j}}$ in the collection $\left\{
k_{\mathbf{x}_{j}}\right\}  _{j=1}^{N}$, along with a distribution of the
random vector $k_{\mathbf{x}_{i}}$ that is conditional on relations
$\left\Vert k_{\mathbf{x}_{i}}\right\Vert \left\Vert k_{\mathbf{x}_{j}%
}\right\Vert \cos\theta_{k_{\mathbf{x}_{i}}k_{\mathbf{x}_{j}}}$ between the
random vector $k_{\mathbf{x}_{i}}$ and all of the random vectors
$k_{\mathbf{x}_{j}}$ in the collection $\left\{  k_{\mathbf{x}_{j}}\right\}
_{j=1}^{N}$.

Next, take any given row $i$ of the $N\times N$ kernel matrix $\mathbf{Q}$. A
joint covariance statistic for the random vector $k_{\mathbf{x}_{i}}$ is given
by the expression%
\begin{align}
\widehat{\operatorname*{cov}}\left(  k_{\mathbf{x}_{i}}\right)   &
=\sum\nolimits_{j=1}^{N}\left\Vert k_{\mathbf{x}_{i}}\right\Vert \left\Vert
k_{\mathbf{x}_{j}}\right\Vert \cos\theta_{k_{\mathbf{x}_{i}}k_{\mathbf{x}_{j}%
}}\tag{14.2}\label{Joint Covariance Statistic}\\
&  =\left\Vert k_{\mathbf{x}_{i}}\right\Vert \sum\nolimits_{j=1}^{N}\left\Vert
k_{\mathbf{x}_{j}}\right\Vert \cos\theta_{k_{\mathbf{x}_{i}}k_{\mathbf{x}_{j}%
}}\nonumber\\
&  =\left\Vert k_{\mathbf{x}_{i}}\right\Vert \sum\nolimits_{j=1}%
^{N}\operatorname{comp}_{\overrightarrow{k_{\mathbf{x}_{i}}}}\left(
\overrightarrow{k_{\mathbf{x}_{j}}}\right) \nonumber\\
&  =\left\Vert k_{\mathbf{x}_{i}}\right\Vert \operatorname{comp}%
_{\overrightarrow{k_{\mathbf{x}_{i}}}}\left(  \overrightarrow{\sum
\nolimits_{j=1}^{N}k_{\mathbf{x}_{j}}}\right)  \text{,}\nonumber
\end{align}
based on the $N$ relations $\left\Vert k_{\mathbf{x}_{i}}\right\Vert
\left\Vert k_{\mathbf{x}_{j}}\right\Vert \cos\theta_{k_{\mathbf{x}_{i}%
}k_{\mathbf{x}_{j}}}$ between the random vector $k_{\mathbf{x}_{i}}$ and the
$N$ random vectors $k_{\mathbf{x}_{j}}$ in the collection $\left\{
k_{\mathbf{x}_{j}}\right\}  _{j=1}^{N}$, such that the relation $\left\Vert
k_{\mathbf{x}_{i}}\right\Vert \left\Vert k_{\mathbf{x}_{j}}\right\Vert
\cos\theta_{k_{\mathbf{x}_{i}}k_{\mathbf{x}_{j}}}$ between the random vector
$k_{\mathbf{x}_{i}}$ and any given random vector $k_{\mathbf{x}_{j}}$ in the
collection represents joint variations between all of the $d$ point
coordinates on the loci of the random vectors $k_{\mathbf{x}_{i}}$ and
$k_{\mathbf{x}_{j}}$, and the signed magnitude $\operatorname{comp}%
_{\overrightarrow{k_{\mathbf{x}_{i}}}}\left(  \overrightarrow{k_{\mathbf{x}%
_{j}}}\right)  $ of the random vector $k_{\mathbf{x}_{j}}$ along the locus of
the random vector $k_{\mathbf{x}_{i}}$ determines a certain portion
$\left\Vert k_{\mathbf{x}_{j}}\right\Vert \cos\theta_{k_{\mathbf{x}_{i}%
}k_{\mathbf{x}_{j}}}$ of the random vector $k_{\mathbf{x}_{j}}$ that is
distributed along the locus of the random vector $k_{\mathbf{x}_{i}}$, at
which point the distribution of the random vector $k_{\mathbf{x}_{j}}$ is
either positive or negative.

Accordingly, the expression $\left\Vert k_{\mathbf{x}_{i}}\right\Vert
\sum\nolimits_{j=1}^{N}\operatorname{comp}_{\overrightarrow{k_{\mathbf{x}_{i}%
}}}\left(  \overrightarrow{k_{\mathbf{x}_{j}}}\right)  $ determines a
condensed signed magnitude---along the locus of the random vector
$k_{\mathbf{x}_{i}}$---that represents a joint covariance
$\widehat{\operatorname*{cov}}\left(  k_{\mathbf{x}_{i}}\right)  $ of the
random vector $k_{\mathbf{x}_{i}}$ that is conditional on the relations
$\sum\nolimits_{j=1}^{N}\left\Vert k_{\mathbf{x}_{i}}\right\Vert \left\Vert
k_{\mathbf{x}_{j}}\right\Vert \cos\theta_{k_{\mathbf{x}_{i}}k_{\mathbf{x}_{j}%
}}$, along with an expected location of the random vector $k_{\mathbf{x}_{i}}$
that is conditional on how the loci of the $N$ vectors $\left\{
k_{\mathbf{x}_{j}}\right\}  _{i=1}^{N}$ are distributed along the locus of the
random vector $k_{\mathbf{x}_{i}}$.

The condensed signed magnitude $\left\Vert k_{\mathbf{x}_{i}}\right\Vert
\sum\nolimits_{j=1}^{N}\operatorname{comp}_{\overrightarrow{k_{\mathbf{x}_{i}%
}}}\left(  \overrightarrow{k_{\mathbf{x}_{j}}}\right)  $ along the locus of
the random vector $k_{\mathbf{x}_{i}}$ can be written in terms of signed
magnitudes $\left\Vert k_{\mathbf{x}_{i}}\right\Vert \cos\theta_{k_{\mathbf{x}%
_{j}}k_{\mathbf{x}_{i}}}$ along the loci of the $N$ random vectors
$k_{\mathbf{x}_{j}}$%
\begin{align}
\widehat{\operatorname*{cov}}\left(  k_{\mathbf{x}_{i}}\right)   &
=\sum\nolimits_{j=1}^{N}\left\Vert k_{\mathbf{x}_{j}}\right\Vert \left[
\left\Vert k_{\mathbf{x}_{i}}\right\Vert \cos\theta_{k_{\mathbf{x}_{j}%
}k_{\mathbf{x}_{i}}}\right] \tag{14.3}\label{Joint Covariance 2}\\
&  =\sum\nolimits_{j=1}^{N}\left\Vert k_{\mathbf{x}_{j}}\right\Vert
\operatorname{comp}_{\overrightarrow{k_{\mathbf{x}_{j}}}}\left(
\overrightarrow{k_{\mathbf{x}_{i}}}\right)  \text{,}\nonumber
\end{align}
so that the expected location and the joint covariance
$\widehat{\operatorname*{cov}}\left(  k_{\mathbf{x}_{i}}\right)  $ of the
random vector $k_{\mathbf{x}_{i}}$\ are both determined by how the locus of
the random vector $k_{\mathbf{x}_{i}}$ is distributed along the loci of the
$N$ random vectors $\left\{  k_{\mathbf{x}_{j}}\right\}  _{i=1}^{N}$.
\end{axiom}

We now turn our attention to the core of the machine learning algorithm that
finds discriminant functions of minimum risk binary classification systems.

\section{\label{Section 15}Inside the Wolfe-dual Principal Eigenspace}

In this section of our treatise, we reveal elegant statistical relations and
deep-seated statistical interconnections between all of the principal
eigenaxis components and likelihood components that lie on both sides of the
primal novel principal eigenaxis $\boldsymbol{\kappa}=\boldsymbol{\kappa}%
_{1}-\boldsymbol{\kappa}_{2}$ in (\ref{Primal Dual Locus})%
\begin{align*}
\boldsymbol{\kappa}  &  =\sum\nolimits_{i=1}^{l_{1}}\psi_{1_{i\ast}%
}k_{\mathbf{x}_{1_{i\ast}}}-\sum\nolimits_{i=1}^{l_{2}}\psi_{2_{i\ast}%
}k_{\mathbf{x}_{2_{i\ast}}}\\
&  =\boldsymbol{\kappa}_{1}{\large -}\boldsymbol{\kappa}_{2}%
\end{align*}
and the Wolfe-dual novel principal eigenaxis $\boldsymbol{\psi}%
=\boldsymbol{\psi}_{1}+\boldsymbol{\psi}_{2}$ in (\ref{Wolf-dual Dual Locus})%
\begin{align*}
\boldsymbol{\psi}  &  =\sum\nolimits_{i=1}^{l_{1}}\psi_{1i\ast}\frac
{k_{\mathbf{x}_{1i\ast}}}{\left\Vert k_{\mathbf{x}_{1i\ast}}\right\Vert }%
+\sum\nolimits_{i=1}^{l_{2}}\psi_{2i\ast}\frac{k_{\mathbf{x}_{2i\ast}}%
}{\left\Vert k_{\mathbf{x}_{2i\ast}}\right\Vert }\\
&  =\boldsymbol{\psi}_{1}+\boldsymbol{\psi}_{2}%
\end{align*}
of the minimum risk binary classification system $k_{\mathbf{s}}%
\boldsymbol{\kappa}+\boldsymbol{\kappa}_{0}\overset{\omega_{1}%
}{\underset{\omega_{2}}{\gtrless}}0$, such that critical interconnections
between intrinsic components of the system determine an exclusive principal
eigen-coordinate system $\boldsymbol{\kappa}=\boldsymbol{\kappa}%
_{1}-\boldsymbol{\kappa}_{2}$ that is the principal part of an equivalent
representation of the pair of random quadratic forms $\boldsymbol{\psi}%
^{T}\mathbf{Q}\boldsymbol{\psi}$ and $\boldsymbol{\psi}^{T}\mathbf{Q}%
^{-1}\boldsymbol{\psi}$, so that the structure and behavior and properties of
the geometric locus of the primal novel principal eigenaxis
$\boldsymbol{\kappa}=\boldsymbol{\kappa}_{1}-\boldsymbol{\kappa}_{2}$ are
symmetrically and equivalently related to the structure and behavior and
properties of the geometric locus of the Wolfe-dual novel principal eigenaxis
$\boldsymbol{\psi}=\boldsymbol{\psi}_{1}+\boldsymbol{\psi}_{2}$, at which
point the Wolfe-dual novel principal eigenaxis $\boldsymbol{\psi}$ is
symmetrically and equivalently related to the largest eigenvector
$\boldsymbol{\psi}_{\max}$ of the joint covariance matrix $\mathbf{Q}$ and the
inverted joint covariance matrix $\mathbf{Q}^{-1}$ associated with the pair of
random quadratic forms $\boldsymbol{\psi}^{T}\mathbf{Q}\boldsymbol{\psi}$ and
$\boldsymbol{\psi}^{T}\mathbf{Q}^{-1}\boldsymbol{\psi}$.

The Wolfe-dual eigenenergy functional in
(\ref{Matrix Version Wolfe Dual Problem}) is the core of the machine learning
algorithm that finds discriminant functions of minimum risk binary
classification systems.

In Section \ref{Section 20}, it will be seen that the machine learning
algorithm being examined \emph{transforms} the random quadratic form
$\boldsymbol{\psi}^{T}\mathbf{Q}\boldsymbol{\psi}$ in the Wolfe-dual
eigenenergy functional $\max\Xi\left(  \boldsymbol{\psi}\right)
=\mathbf{1}^{T}\boldsymbol{\psi}-\boldsymbol{\psi}^{T}\mathbf{Q}%
\boldsymbol{\psi/}2$ of a minimum risk binary classification system
$k_{\mathbf{s}}\boldsymbol{\kappa}+\mathbf{\kappa}_{0}\overset{\omega
_{1}}{\underset{\omega_{2}}{\gtrless}}0$---which is subject to the constraints
$\boldsymbol{\psi}^{T}\mathbf{y}=0$ and $\psi_{i\ast}>0$---\emph{into} a
geometric locus of a novel principal eigenaxis $\boldsymbol{\kappa
}=\boldsymbol{\kappa}_{1}-\boldsymbol{\kappa}_{2}$ of the system, so that the
geometric locus of the novel principal eigenaxis $\boldsymbol{\kappa
}=\boldsymbol{\kappa}_{1}-\boldsymbol{\kappa}_{2}$ satisfies the geometric
locus of the decision boundary $k_{\mathbf{s}}\boldsymbol{\kappa}+$
$\boldsymbol{\kappa}_{0}=0$ of the system in terms of a critical minimum
eigenenergy $\left\Vert \boldsymbol{\kappa}\right\Vert _{\min_{c}}^{2}$ and a
minimum expected risk $\mathfrak{R}_{\mathfrak{\min}}\left(  \left\Vert
\boldsymbol{\kappa}\right\Vert _{\min_{c}}^{2}\right)  $, at which point the
total allowed eigenenergy $\left\Vert \boldsymbol{\kappa}\right\Vert
_{\min_{c}}^{2}$ and the expected risk $\mathfrak{R}_{\mathfrak{\min}}\left(
\left\Vert \boldsymbol{\kappa}\right\Vert _{\min_{c}}^{2}\right)  $ exhibited
by the novel principal eigenaxis $\boldsymbol{\kappa}=\boldsymbol{\kappa}%
_{1}-\boldsymbol{\kappa}_{2}$ are both regulated by the total value of the
scale factors $\psi_{1i\ast}$ and $\psi_{2i\ast}$ for the components
$\psi_{1i\ast}\frac{k_{\mathbf{x}_{1i\ast}}}{\left\Vert k_{\mathbf{x}_{1i\ast
}}\right\Vert }$ and $\psi_{2i\ast}\frac{k_{\mathbf{x}_{2i\ast}}}{\left\Vert
k_{\mathbf{x}_{2i\ast}}\right\Vert }$ of the principal eigenvector
$\boldsymbol{\psi}_{\max}$ of the joint covariance matrix $\mathbf{Q}$ and the
inverted joint covariance matrix $\mathbf{Q}^{-1}$ associated with the pair of
random quadratic forms $\boldsymbol{\psi}^{T}\mathbf{Q}\boldsymbol{\psi}$ and
$\boldsymbol{\psi}^{T}\mathbf{Q}^{-1}\boldsymbol{\psi}$.

We now identify the essence of the Wolfe-dual eigenenergy functional $\max
\Xi\left(  \boldsymbol{\psi}\right)  =\mathbf{1}^{T}\boldsymbol{\psi
}-\boldsymbol{\psi}^{T}\mathbf{Q}\boldsymbol{\psi/}2$, which is subject to the
constraints $\boldsymbol{\psi}^{T}\mathbf{y}=0$ and $\psi_{i\ast}>0$, where
$y_{i}=\left\{  \pm1\right\}  $, so that the structure and behavior and
properties of the Wolfe-dual novel principal eigenaxis $\boldsymbol{\psi
}=\boldsymbol{\psi}_{1}+\boldsymbol{\psi}_{2}$ are symmetrically and
equivalently related to the structure and behavior and properties of the
primal novel principal eigenaxis $\boldsymbol{\kappa}=\boldsymbol{\kappa}%
_{1}-\boldsymbol{\kappa}_{2}$.

\subsection{Symmetrical and Equivalent Eigenenergies}

Consider again the critical minimum eigenenergy constraint $\gamma\left(
\boldsymbol{\kappa}\right)  =\left\Vert \boldsymbol{\kappa}\right\Vert
_{\min_{c}}^{2}$ on the geometric locus of the primal novel principal
eigenaxis $\boldsymbol{\kappa}$ in (\ref{Objective Function}), so that the
expected risk $\mathfrak{R}_{\mathfrak{\min}}\left(  \left\Vert
\boldsymbol{\kappa}\right\Vert _{\min_{c}}^{2}\right)  $ and the total allowed
eigenenergy $\left\Vert \boldsymbol{\kappa}\right\Vert _{\min_{c}}^{2}$
exhibited by a minimum risk binary classification system $k_{\mathbf{s}%
}\boldsymbol{\kappa}+\mathbf{\kappa}_{0}\overset{\omega_{1}}{\underset{\omega
_{2}}{\gtrless}}0$ are jointly minimized within the decision space
$Z=Z_{1}\cup Z_{2}$ of the system.

By the strong duality theorem
\citep{Fletcher2000,Luenberger1969,Luenberger2003,Nash1996}%
, we realize that the geometric locus of the Wolfe-dual novel principal
eigenaxis $\boldsymbol{\psi}$ is subject to a critical minimum eigenenergy
constraint that is symmetrically and equivalently related to the critical
minimum eigenenergy constraint on the geometric locus of the primal novel
principal eigenaxis $\boldsymbol{\kappa}$---inside the Wolfe-dual principal
eigenspace of $\boldsymbol{\psi}$ and $\boldsymbol{\kappa}$, so that the
Wolfe-dual novel principal eigenaxis $\boldsymbol{\psi}$ and the primal novel
principal eigenaxis $\boldsymbol{\kappa}$ exhibit symmetrical and equivalent
eigenenergies $\left\Vert Z|\boldsymbol{\psi}\right\Vert _{\min_{c}}^{2}%
\equiv\left\Vert Z|\boldsymbol{\kappa}\right\Vert _{\min_{c}}^{2}$.

Thereby, the geometric locus of the Wolfe-dual novel principal eigenaxis
$\boldsymbol{\psi}$ is subject to a critical minimum eigenenergy constraint%
\begin{equation}
\lambda_{1}\left\Vert \boldsymbol{\psi}\right\Vert _{\min_{c}}^{2}%
=\boldsymbol{\psi}_{\max}^{T}\mathbf{Q}\boldsymbol{\psi}_{\max}\equiv
\left\Vert \boldsymbol{\kappa}\right\Vert _{\min_{c}}^{2} \tag{15.1}%
\label{Dual Eigenenergy Constraint}%
\end{equation}
at which point the random quadratic form $\boldsymbol{\psi}_{\max}%
^{T}\mathbf{Q}\boldsymbol{\psi}_{\max}$ is symmetrically and equivalently
related to the critical minimum eigenenergy $\left\Vert \boldsymbol{\kappa
}\right\Vert _{\min_{c}}^{2}$ exhibited by the geometric locus of the primal
novel principal eigenaxis $\boldsymbol{\kappa}$, so that the random quadratic
form $\boldsymbol{\psi}_{\max}^{T}\mathbf{Q}\boldsymbol{\psi}_{\max}$, plus
the total allowed eigenenergy $\left\Vert \boldsymbol{\kappa}\right\Vert
_{\min_{c}}^{2}$ and the expected risk $\mathfrak{R}_{\mathfrak{\min}}\left(
\left\Vert \boldsymbol{\kappa}\right\Vert _{\min_{c}}^{2}\right)  $ exhibited
by the geometric locus of the primal novel eigenaxis $\boldsymbol{\kappa}$
jointly reach their minimum values.

Given the eigenenergy constraint on $\boldsymbol{\psi}$ in
(\ref{Dual Eigenenergy Constraint}), it follows that the Wolfe-dual
eigenenergy functional%
\[
\max\Xi\left(  \boldsymbol{\psi}\right)  =\mathbf{1}^{T}\boldsymbol{\psi
}-\boldsymbol{\psi}^{T}\mathbf{Q}\boldsymbol{\psi/}2\text{,}%
\]
such that $\boldsymbol{\psi}^{T}\mathbf{y}=0$ and $\psi_{i\ast}>0$, is
\emph{maximized} by the largest eigenvector $\boldsymbol{\psi}_{\max}$ of the
joint covariance matrix $\mathbf{Q}$%
\begin{equation}
\mathbf{Q}\boldsymbol{\psi}_{\max}=\lambda_{1}\boldsymbol{\psi}_{\max}\text{,}
\tag{15.2}\label{Principal Eigenvector Constraint}%
\end{equation}
at which point the random quadratic form $\boldsymbol{\psi}_{\max}%
^{T}\mathbf{Q}\boldsymbol{\psi}_{\max}$ reaches it \emph{minimum} value, so
that the total allowed eigenenergy $\left\Vert \boldsymbol{\kappa
}=\boldsymbol{\kappa}_{1}-\boldsymbol{\kappa}_{2}\right\Vert _{\min_{c}}^{2}$
and the expected risk $\mathfrak{R}_{\mathfrak{\min}}\left(  \left\Vert
\boldsymbol{\kappa}=\boldsymbol{\kappa}_{1}-\boldsymbol{\kappa}_{2}\right\Vert
_{\min_{c}}^{2}\right)  $ exhibited by the geometric locus of the primal novel
principal eigenaxis $\boldsymbol{\kappa}=\boldsymbol{\kappa}_{1}%
-\boldsymbol{\kappa}_{2}$ are jointly minimized.

Thereby, all of the principal eigenaxis components $\psi_{1i\ast}%
\frac{k_{\mathbf{x}_{1i\ast}}}{\left\Vert k_{\mathbf{x}_{1i\ast}}\right\Vert
}$ and $\psi_{2i\ast}\frac{k_{\mathbf{x}_{2i\ast}}}{\left\Vert k_{\mathbf{x}%
_{2i\ast}}\right\Vert }$ on the Wolfe-dual novel principal eigenaxis
$\boldsymbol{\psi}=\boldsymbol{\psi}_{1}+\boldsymbol{\psi}_{2}$, along with
all of the principal eigenaxis components $\psi_{1_{i\ast}}k_{\mathbf{x}%
_{1_{i\ast}}}$ and $\psi_{2_{i\ast}}k_{\mathbf{x}_{2_{i\ast}}}$ on the primal
novel principal eigenaxis $\boldsymbol{\kappa}=\boldsymbol{\kappa}%
_{1}-\boldsymbol{\kappa}_{2}$, are subject to minimum length constraints---in
accordance with the critical minimum eigenenergy constraints on both
$\boldsymbol{\psi}$ and $\boldsymbol{\kappa}$.

Given conditions expressed by Theorem
\ref{Principal Eigen-coordinate System Theorem} and the expression for the
principal eigenvector $\boldsymbol{\psi}_{\max}$ of the joint covariance
matrix $\mathbf{Q}$ in (\ref{Wolf-dual Dual Locus}), it follows that the
random quadratic form $\boldsymbol{\psi}_{\max}^{T}\mathbf{Q}\boldsymbol{\psi
}_{\max}$ has an equivalent representation that is given by%
\begin{equation}
\boldsymbol{\psi}_{\max}^{T}\mathbf{Q}\boldsymbol{\psi}_{\max}\boldsymbol{=}%
\sum\nolimits_{i=1}^{l_{1}}\lambda_{1i}\left\Vert \psi_{1i\ast}\frac
{k_{\mathbf{x}_{1i\ast}}}{\left\Vert k_{\mathbf{x}_{1i\ast}}\right\Vert
}\right\Vert ^{2}+\sum\nolimits_{i=1}^{l_{2}}\lambda_{2i}\left\Vert
\psi_{2i\ast}\frac{k_{\mathbf{x}_{2i\ast}}}{\left\Vert k_{\mathbf{x}_{2i\ast}%
}\right\Vert }\right\Vert ^{2}\text{,} \tag{15.3}%
\label{Equiv Rep Random Quadratic Form}%
\end{equation}
where $\lambda_{1i}$ and $\lambda_{2i}$ are eigenvalues of the joint
covariance matrix $\mathbf{Q}$ of the random quadratic form $\boldsymbol{\psi
}_{\max}^{T}\mathbf{Q}\boldsymbol{\psi}_{\max}$, $\psi_{1i\ast}$ and
$\psi_{2i\ast}$ are scale factors for principal eigenaxis components
$\psi_{1i\ast}\frac{k_{\mathbf{x}_{1i\ast}}}{\left\Vert k_{\mathbf{x}_{1i\ast
}}\right\Vert }$ and $\psi_{2i\ast}\frac{k_{\mathbf{x}_{2i\ast}}}{\left\Vert
k_{\mathbf{x}_{2i\ast}}\right\Vert }$ on the Wolfe-dual novel principal
eigenaxis $\boldsymbol{\psi}=\boldsymbol{\psi}_{1}+\boldsymbol{\psi}_{2}$, and
$\left\Vert \psi_{1i\ast}\frac{k_{\mathbf{x}_{1i\ast}}}{\left\Vert
k_{\mathbf{x}_{1i\ast}}\right\Vert }\right\Vert ^{2}$ and $\left\Vert
\psi_{2i\ast}\frac{k_{\mathbf{x}_{2i\ast}}}{\left\Vert k_{\mathbf{x}_{2i\ast}%
}\right\Vert }\right\Vert $ are eigenenergies exhibited by corresponding
principal eigenaxis components $\psi_{1i\ast}\frac{k_{\mathbf{x}_{1i\ast}}%
}{\left\Vert k_{\mathbf{x}_{1i\ast}}\right\Vert }$ and $\psi_{2i\ast}%
\frac{k_{\mathbf{x}_{2i\ast}}}{\left\Vert k_{\mathbf{x}_{2i\ast}}\right\Vert
}$ on $\boldsymbol{\psi}=\boldsymbol{\psi}_{1}+\boldsymbol{\psi}_{2}$.

Given the equivalent representation for the random quadratic form in
(\ref{Equiv Rep Random Quadratic Form}), along with the eigenenergy constraint
on $\boldsymbol{\psi}$ in (\ref{Dual Eigenenergy Constraint})
\[
\lambda_{1}\left\Vert \boldsymbol{\psi}\right\Vert _{\min_{c}}^{2}%
=\boldsymbol{\psi}_{\max}^{T}\mathbf{Q}\boldsymbol{\psi}_{\max}\equiv
\left\Vert \boldsymbol{\kappa}\right\Vert _{\min_{c}}^{2}\text{,}%
\]
it follows that the total allowed eigenenergy $\lambda_{1}\left\Vert
\boldsymbol{\psi}\right\Vert _{\min_{c}}^{2}$ exhibited by the Wolfe-dual
novel principal eigenaxis $\boldsymbol{\psi}$ and the total allowed
eigenenergy $\left\Vert \boldsymbol{\kappa}\right\Vert _{\min_{c}}^{2}$
exhibited by the primal novel principal eigenaxis $\boldsymbol{\kappa}$ \ are
both regulated by eigenvalues $\lambda_{1i}$ and $\lambda_{2i}$ of the joint
covariance matrix $\mathbf{Q}$ of the random quadratic form $\boldsymbol{\psi
}^{T}\mathbf{Q}\boldsymbol{\psi}$, where eigenvalues $\lambda_{i}$ of the
joint covariance matrix $\mathbf{Q}$ associated with scale factors $\psi_{i}$
for which $\psi_{i}=0$ also regulate the total allowed eigenenergies
$\lambda_{1}\left\Vert \boldsymbol{\psi}\right\Vert _{\min_{c}}^{2}$ and
$\left\Vert \boldsymbol{\kappa}\right\Vert _{\min_{c}}^{2}$ exhibited by
$\boldsymbol{\psi}$ and $\boldsymbol{\kappa}$.

We now begin identifying how the constrained geometric locus of the Wolfe-dual
novel principal eigenaxis%
\[
\boldsymbol{\psi}=\sum\nolimits_{i=1}^{l_{1}}\psi_{1i\ast}\frac{k_{\mathbf{x}%
_{1i\ast}}}{\left\Vert k_{\mathbf{x}_{1i\ast}}\right\Vert }+\sum
\nolimits_{i=1}^{l_{2}}\psi_{2i\ast}\frac{k_{\mathbf{x}_{2i\ast}}}{\left\Vert
k_{\mathbf{x}_{2i\ast}}\right\Vert }%
\]
is symmetrically and equivalently related to the constrained geometric locus
of the primal novel principal eigenaxis%
\[
\boldsymbol{\kappa}=\sum\nolimits_{i=1}^{l_{1}}\psi_{1_{i\ast}}k_{\mathbf{x}%
_{1_{i\ast}}}-\sum\nolimits_{i=1}^{l_{2}}\psi_{2_{i\ast}}k_{\mathbf{x}%
_{2_{i\ast}}}%
\]
inside the Wolfe-dual principal eigenspace of $\boldsymbol{\psi}$ and
$\boldsymbol{\kappa}$, where $l_{1}+l_{2}=$ $l$.

\subsection{Statistical Pre-wiring of Important Generalizations}

In this section of our treatise, we examine the vector algebra locus equation
$\boldsymbol{\psi}=\lambda_{1}^{-1}\boldsymbol{\psi}_{\max}^{T}\mathbf{Q}$, so
that the geometric locus of the Wolfe-dual novel principal eigenaxis
$\boldsymbol{\psi}$ is related to the scaled principal eigenvector
$\lambda_{1}^{-1}\boldsymbol{\psi}_{\max}$ \emph{acting} on the joint
covariance matrix $\mathbf{Q}$ of the random quadratic form $\boldsymbol{\psi
}^{T}\mathbf{Q}\boldsymbol{\psi}$.

Thereby, we demonstrate that certain eigenfunction actions on the joint
covariance matrix $\mathbf{Q}$ of the random quadratic form $\boldsymbol{\psi
}^{T}\mathbf{Q}\boldsymbol{\psi}$ \emph{statistically} \emph{pre-wire the
important generalizations} for a minimum risk binary classification system
$k_{\mathbf{s}}\boldsymbol{\kappa}+\mathbf{\kappa}_{0}\overset{\omega
_{1}}{\underset{\omega_{2}}{\gtrless}}0$---such that likely locations and
likelihood values for each and every one of the extreme points $\mathbf{x}%
_{1_{i\ast}}\mathbf{\sim}$ $p\left(  \mathbf{x};\omega_{1}\right)  $ and
$\mathbf{x}_{2_{i\ast}}\mathbf{\sim}$ $p\left(  \mathbf{x};\omega_{2}\right)
$ are \emph{statistically pre-wired within the geometric locus of the novel
principal eigenaxis }$\boldsymbol{\kappa}=\boldsymbol{\kappa}_{1}%
-\boldsymbol{\kappa}_{2}$ of the system---with \emph{respect to} and \emph{in
relation to} each and every one of the principal eigenaxis components
$\psi_{1_{i\ast}}k_{\mathbf{x}_{1_{i\ast}}}$ and $\psi_{2_{i\ast}%
}k_{\mathbf{x}_{2_{i\ast}}}$ and the correlated likelihood components
$\psi_{1_{i\ast}}k_{\mathbf{x}_{1_{i\ast}}}$ and $\psi_{2_{i\ast}%
}k_{\mathbf{x}_{2_{i\ast}}}$ that lie on the geometric locus of the novel
principal eigenaxis $\boldsymbol{\kappa}=\boldsymbol{\kappa}_{1}%
-\boldsymbol{\kappa}_{2}$.

\subsection{Eigenfunction Actions on Joint Covariance Matrices}

Using the eigenvector relation in (\ref{Principal Eigenvector Constraint}), it
follows that the geometric locus of a Wolfe-dual novel principal eigenaxis
$\boldsymbol{\psi}$ satisfies the vector algebra locus equation%
\begin{align}
\boldsymbol{\psi}  &  \triangleq\left(
\begin{array}
[c]{c}%
\psi_{1}\\
\psi_{2}\\
\vdots\\
\psi_{N}%
\end{array}
\right)  =\frac{\psi_{1}}{\lambda_{1}}%
\begin{pmatrix}
\left\Vert k_{\mathbf{x}_{1}}\right\Vert \left\Vert k_{\mathbf{x}_{1}%
}\right\Vert \cos\theta_{k_{\mathbf{x_{1}}}k_{\mathbf{x}_{1}}}\\
\left\Vert k_{\mathbf{x}_{2}}\right\Vert \left\Vert k_{\mathbf{x}_{1}%
}\right\Vert \cos\theta_{k_{\mathbf{x}_{2}}k_{\mathbf{x}_{1}}}\\
\vdots\\
-\left\Vert k_{\mathbf{x}_{N}}\right\Vert \left\Vert k_{\mathbf{x}_{1}%
}\right\Vert \cos\theta_{k_{\mathbf{x}_{N}}k_{\mathbf{x}_{1}}}%
\end{pmatrix}
+\cdots\tag{15.4}\label{Eig Eq}\\
&  +\cdots\frac{\psi_{N}}{\lambda_{1}}%
\begin{pmatrix}
-\left\Vert k_{\mathbf{x}_{1}}\right\Vert \left\Vert k_{\mathbf{x}_{N}%
}\right\Vert \cos\theta_{k_{\mathbf{x}_{1}}k_{\mathbf{x}_{N}}}\\
-\left\Vert k_{\mathbf{x}_{2}}\right\Vert \left\Vert k_{\mathbf{x}_{N}%
}\right\Vert \cos\theta_{k_{\mathbf{x}_{2}}k_{\mathbf{x}_{N}}}\\
\vdots\\
\left\Vert k_{\mathbf{x}_{N}}\right\Vert \left\Vert k_{\mathbf{x}_{N}%
}\right\Vert \cos\theta_{k_{\mathbf{x}_{N}}k_{\mathbf{x}_{N}}}%
\end{pmatrix}
\text{,}\nonumber
\end{align}
so that the geometric locus of the Wolfe-dual novel principal eigenaxis
$\boldsymbol{\psi}$ is related to the scaled principal eigenvector
$\lambda_{1}^{-1}\boldsymbol{\psi}_{\max}$ \emph{acting} on the joint
covariance matrix $\mathbf{Q}$ of the random quadratic form $\boldsymbol{\psi
}^{T}\mathbf{Q}\boldsymbol{\psi}$, at which point each scale factor $\psi_{i}$
for a principal axis of the principal eigenvector $\boldsymbol{\psi}_{\max}$
is correlated with scalar projections $\left\Vert k_{\mathbf{x}_{j}%
}\right\Vert \cos\theta_{k_{\mathbf{x_{i}}}k_{\mathbf{x}_{j}}}$ of a feature
vector $k_{\mathbf{x}_{j}}$ onto a collection of $N$ signed $\pm1$ feature
vectors $k_{\mathbf{x}_{i}}$.

It will be seen that the vector algebra locus equation $\boldsymbol{\psi
}=\lambda_{1}^{-1}\boldsymbol{\psi}^{T}\mathbf{Q}$ in (\ref{Eig Eq})
implements a vector-valued cost function, such that the eigenenergy exhibited
by both $\boldsymbol{\psi}$ and $\boldsymbol{\kappa}$ is minimized in
accordance with the eigenenergy condition $\boldsymbol{\psi}_{\max}%
^{T}\mathbf{Q}\boldsymbol{\psi}_{\max}=\lambda_{\boldsymbol{1}}\left\Vert
\boldsymbol{\psi}\right\Vert _{\min_{c}}^{2}\equiv\left\Vert
\boldsymbol{\kappa}\right\Vert _{\min_{c}}^{2}$, so that all of the principal
eigenaxis components $\psi_{1i\ast}\frac{k_{\mathbf{x}_{1i\ast}}}{\left\Vert
k_{\mathbf{x}_{1i\ast}}\right\Vert }$ and $\psi_{2i\ast}\frac{k_{\mathbf{x}%
_{2i\ast}}}{\left\Vert k_{\mathbf{x}_{2i\ast}}\right\Vert }$ on
$\boldsymbol{\psi}$ and all of the principal eigenaxis components
$\psi_{1_{i\ast}}k_{\mathbf{x}_{1_{i\ast}}}$ and $\psi_{2_{i\ast}%
}k_{\mathbf{x}_{2_{i\ast}}}$ on $\boldsymbol{\kappa}$ are subject to minimum
length constraints, at which point the random quadratic form $\boldsymbol{\psi
}_{\max}^{T}\mathbf{Q}\boldsymbol{\psi}_{\max}$ and the total allowed
eigenenergy $\left\Vert \boldsymbol{\kappa}\right\Vert _{\min_{c}}^{2}$
exhibited by the novel principal eigenaxis $\boldsymbol{\kappa}$ both reach
their minimum value.

We have previously demonstrated that the Wolfe-dual novel principal eigenaxis
$\boldsymbol{\psi}$ satisfies the Lagrangian relation%
\begin{equation}
\boldsymbol{\psi}=\mathbf{Q}^{-1}\left(  \mathbf{1+\lambda}\right)
+\lambda_{0}\mathbf{Q}^{-1}\boldsymbol{y}\text{,} \tag{15.5}%
\label{Inverse Covariance Matrix}%
\end{equation}
where $\mathbf{\lambda}$ and $\lambda_{0}$ denote additional Lagrange
multipliers for the Lagrangian of the Wolfe-dual eigenenergy functional in
(\ref{Matrix Version Wolfe Dual Problem})
\citep{Reeves2011,Reeves2009}%
.

By the Lagrangian relation in (\ref{Inverse Covariance Matrix}), it follows
that finding an equivalent representation of the random quadratic form
$\boldsymbol{\psi}^{T}\mathbf{Q}\boldsymbol{\psi}$ in
(\ref{Matrix Version Wolfe Dual Problem}) requires finding the values of the
active scale factors $\psi_{i\ast}>0$ that are associated with the inverted
$\mathbf{Q}^{-1}$ joint covariance matrix of the random quadratic form
$\boldsymbol{\psi}^{T}\mathbf{Q}^{-1}\boldsymbol{\psi}$.

It also follows that the geometric locus of the Wolfe-dual novel principal
eigenaxis $\boldsymbol{\psi}$ is determined by covariance and distribution
information for extreme vectors $k_{\mathbf{x}_{1_{i\ast}}}$ and
$k_{\mathbf{x}_{2_{i\ast}}}$ that is contained within the inverted joint
covariance matrix $\mathbf{Q}^{-1}$ of the random quadratic form
$\boldsymbol{\psi}^{T}\mathbf{Q}^{-1}\boldsymbol{\psi}$, such that the
covariance and distribution information is conditional on relations between
each extreme vector $k_{\mathbf{x}_{1_{i\ast}}}$ or $k_{\mathbf{x}_{2_{i\ast}%
}}$ and all of the feature vectors $k_{\mathbf{x}_{i}}$ in the collection
$\left\{  k_{\mathbf{x}_{i}}\right\}  _{i=1}^{N}$ of training data.

Moreover, by Axiom \ref{Statistical Structures of Hilbert Space Axiom} and the
Lagrangian relation in (\ref{Inverse Covariance Matrix}), it follows that each
active scale factor $\psi_{1i\ast}>0$ and $\psi_{2i\ast}>0$ is related to the
inverted joint covariance matrix $\mathbf{Q}^{-1}$ so that each principal
eigenaxis component $\psi_{1i\ast}\frac{k_{\mathbf{x}_{1i\ast}}}{\left\Vert
k_{\mathbf{x}_{1i\ast}}\right\Vert }$ and $\psi_{2i\ast}\frac{k_{\mathbf{x}%
_{2i\ast}}}{\left\Vert k_{\mathbf{x}_{2i\ast}}\right\Vert }$ on
$\boldsymbol{\psi}$ is a function of covariance and distribution information
for a correlated extreme vector $k_{\mathbf{x}_{1_{i\ast}}}$ and
$k_{\mathbf{x}_{2_{i\ast}}}$, at which point the information is contained
within the inverted joint covariance matrix $\mathbf{Q}^{-1}$ of the random
quadratic form $\boldsymbol{\psi}^{T}\mathbf{Q}^{-1}\boldsymbol{\psi}$---and
is conditional on relations between an extreme vector $k_{\mathbf{x}%
_{1_{i\ast}}}$ or $k_{\mathbf{x}_{2_{i\ast}}}$ and all of the feature vectors
$k_{\mathbf{x}_{i}}$ in the collection $\left\{  k_{\mathbf{x}_{i}}\right\}
_{i=1}^{N}$ of training data.

Correspondingly, given that the eigenvalues $\lambda_{N}^{-1}\leq
\mathbf{\ldots}\leq\lambda_{1}^{-1}$ of the inverted joint covariance matrix
$\mathbf{Q}^{-1}$ in (\ref{Inverse Covariance Matrix}) vary continuously with
the elements of the inverted joint covariance matrix $\mathbf{Q}^{-1}$, it
follows that each active scale factor $\psi_{1i\ast}>0$ and $\psi_{2i\ast}>0$
is a function of the eigenvalues of the inverted joint covariance matrix
$\mathbf{Q}^{-1}$, such that each principal eigenaxis component $\psi_{1i\ast
}\frac{k_{\mathbf{x}_{1i\ast}}}{\left\Vert k_{\mathbf{x}_{1i\ast}}\right\Vert
}$ and $\psi_{2i\ast}\frac{k_{\mathbf{x}_{2i\ast}}}{\left\Vert k_{\mathbf{x}%
_{2i\ast}}\right\Vert }$ on $\boldsymbol{\psi}$ is a function of covariance
and distribution information for a correlated extreme vector $k_{\mathbf{x}%
_{1_{i\ast}}}$ and $k_{\mathbf{x}_{2_{i\ast}}}$ that is represented by the
eigenvalues $\lambda_{N}^{-1}\leq\mathbf{\ldots}\leq\lambda_{1}^{-1}$ of the
inverted joint covariance matrix $\mathbf{Q}^{-1}$.

Even more, using Axiom \ref{Statistical Structures of Hilbert Space Axiom},
the Lagrangian relation in (\ref{Inverse Covariance Matrix}) and the
expression for the primal novel principal eigenaxis $\boldsymbol{\kappa}$ in
(\ref{Primal Dual Locus})%
\[
\boldsymbol{\kappa}=\sum\nolimits_{i=1}^{l_{1}}\psi_{1_{i\ast}}k_{\mathbf{x}%
_{1_{i\ast}}}-\sum\nolimits_{i=1}^{l_{2}}\psi_{2_{i\ast}}k_{\mathbf{x}%
_{2_{i\ast}}}\text{,}%
\]
it follows that each scale factor $\psi_{1i\ast}$ and $\psi_{2i\ast}$ maps
covariance and distribution information for a correlated extreme vector
$k_{\mathbf{x}_{1_{i\ast}}}$ and $k_{\mathbf{x}_{2_{i\ast}}}$ onto the extreme
vector $k_{\mathbf{x}_{1_{i\ast}}}$ and $k_{\mathbf{x}_{2_{i\ast}}}$, where
the covariance and distribution information is represented by the eigenvalues
$\lambda_{N}^{-1}\leq\mathbf{\ldots}\leq\lambda_{1}^{-1}$ of the inverted
joint covariance matrix $\mathbf{Q}^{-1}$.

Thereby, each principal eigenaxis component $\psi_{1i\ast}k_{\mathbf{x}%
_{1i\ast}}$ and $\psi_{2i\ast}k_{\mathbf{x}_{2i\ast}}$ on the primal novel
principal eigenaxis $\boldsymbol{\kappa}$ is a function of covariance and
distribution information--- for a correlated extreme vector $k_{\mathbf{x}%
_{1_{i\ast}}}$ and $k_{\mathbf{x}_{2_{i\ast}}}$---that is contained within the
inverted joint covariance matrix $\mathbf{Q}^{-1}$, such that the covariance
and distribution information for any given extreme vector $k_{\mathbf{x}%
_{1_{i\ast}}}$ or $k_{\mathbf{x}_{2_{i\ast}}}$ is represented by the
eigenvalues $\lambda_{N}^{-1}\leq\mathbf{\ldots}\leq\lambda_{1}^{-1}$ of the
inverted joint covariance matrix $\mathbf{Q}^{-1}$ and is conditional on the
entire collection $\left\{  k_{\mathbf{x}_{i}}\right\}  _{i=1}^{N}$ of feature
vectors $k_{\mathbf{x}_{i}}$.

What is more, given the interrelations between (\ref{Primal Dual Locus}),
(\ref{Eig Eq}) and (\ref{Inverse Covariance Matrix}), it follows that each
scale factor $\psi_{1i\ast}$ is a function---of \emph{all} of the covariance
and distribution \emph{information}---for \emph{all} of the \emph{extreme
vectors} $k_{\mathbf{x}_{1_{i\ast}}}$ and $k_{\mathbf{x}_{2_{i\ast}}}$ since
each scale factor $\psi_{1i\ast}$ satisfies the locus equation%
\[
\psi_{1i\ast}=\lambda_{1}^{-1}\left(  \sum\nolimits_{j=1}^{l_{1}}%
\psi_{1_{j\ast}}k_{\mathbf{x}_{1_{j\ast}}}-\sum\nolimits_{j=1}^{l_{2}}%
\psi_{2_{j\ast}}k_{\mathbf{x}_{2_{j\ast}}}\right)  k_{\mathbf{x}_{1_{i\ast}}%
}\text{,}%
\]
such that $k_{\mathbf{x}_{1_{j\ast}}}\equiv k_{\mathbf{x}_{1_{i\ast}}}$ and
$k_{\mathbf{x}_{2_{j\ast}}}\equiv k_{\mathbf{x}_{2_{i\ast}}}$, so that the
scale factors $\psi_{1_{j\ast}}$ and $\psi_{2_{j\ast}}$ map covariance and
distribution information for correlated extreme vectors $k_{\mathbf{x}%
_{1_{i\ast}}}$ and $k_{\mathbf{x}_{2_{i\ast}}}$ onto the extreme vectors
$k_{\mathbf{x}_{1_{j\ast}}}$ and $k_{\mathbf{x}_{2_{j\ast}}}$, at which point
the eigenvalue scaled dual locus of $\boldsymbol{\kappa}$%
\[
\lambda_{1}^{-1}\left(  \sum\nolimits_{j=1}^{l_{1}}\psi_{1_{j\ast}%
}k_{\mathbf{x}_{1_{j\ast}}}-\sum\nolimits_{j=1}^{l_{2}}\psi_{2_{j\ast}%
}k_{\mathbf{x}_{2_{j\ast}}}\right)
\]
contains \emph{all} of the covariance and distribution information for
\emph{all} of the extreme vectors $k_{\mathbf{x}_{1_{i\ast}}}$ and
$k_{\mathbf{x}_{2_{i\ast}}}$---such that the covariance and distribution
information for any given extreme vector $k_{\mathbf{x}_{1_{i\ast}}}$ or
$k_{\mathbf{x}_{2_{i\ast}}}$ is represented by the eigenvalues $\lambda
_{N}^{-1}\leq\mathbf{\ldots}\leq\lambda_{1}^{-1}$ of the inverted joint
covariance matrix $\mathbf{Q}^{-1}$---for a given collection $\left\{
\mathbf{x}_{i}\right\}  _{i=1}^{N}$ of feature vectors $\mathbf{x}_{i}$.

Correspondingly, each scale factor $\psi_{2i\ast}$ is a function---of
\emph{all} of the covariance and distribution \emph{information}---for
\emph{all} of the \emph{extreme vectors} $k_{\mathbf{x}_{1_{i\ast}}}$ and
$k_{\mathbf{x}_{2_{i\ast}}}$ since each scale factor $\psi_{2i\ast}$ satisfies
the locus equation%
\[
\psi_{2i\ast}=\lambda_{1}^{-1}\left(  \sum\nolimits_{j=1}^{l_{2}}%
\psi_{2_{j\ast}}k_{\mathbf{x}_{2_{j\ast}}}-\sum\nolimits_{j=1}^{l_{1}}%
\psi_{1_{j\ast}}k_{\mathbf{x}_{1_{j\ast}}}\right)  k_{\mathbf{x}_{2_{i\ast}}%
}\text{,}%
\]
such that $k_{\mathbf{x}_{1_{j\ast}}}\equiv k_{\mathbf{x}_{1_{i\ast}}}$ and
$k_{\mathbf{x}_{2_{j\ast}}}\equiv k_{\mathbf{x}_{2_{i\ast}}}$, so that the
scale factors $\psi_{2_{j\ast}}$ and $\psi_{1_{j\ast}}$ map covariance and
distribution information for correlated extreme vectors $k_{\mathbf{x}%
_{2_{i\ast}}}$ and $k_{\mathbf{x}_{1_{i\ast}}}$ onto the extreme vectors
$k_{\mathbf{x}_{2_{j\ast}}}$ and $k_{\mathbf{x}_{1_{j\ast}}}$, at which point
the eigenvalue scaled dual locus of $-\boldsymbol{\kappa}$%
\[
\lambda_{1}^{-1}\left(  \sum\nolimits_{j=1}^{l_{2}}\psi_{2_{j\ast}%
}k_{\mathbf{x}_{2_{j\ast}}}-\sum\nolimits_{j=1}^{l_{1}}\psi_{1_{j\ast}%
}k_{\mathbf{x}_{1_{j\ast}}}\right)
\]
contains \emph{all} of the covariance and distribution information for
\emph{all} of the extreme vectors $k_{\mathbf{x}_{1_{i\ast}}}$ and
$k_{\mathbf{x}_{2_{i\ast}}}$---such that the covariance and distribution
information for any given extreme vector $k_{\mathbf{x}_{1_{i\ast}}}$ or
$k_{\mathbf{x}_{2_{i\ast}}}$ is represented by the eigenvalues $\lambda
_{N}^{-1}\leq\mathbf{\ldots}\leq\lambda_{1}^{-1}$ of the inverted joint
covariance matrix $\mathbf{Q}^{-1}$---for a given collection $\left\{
\mathbf{x}_{i}\right\}  _{i=1}^{N}$ of feature vectors $\mathbf{x}_{i}$.

\subsection{Generating the Scale Factors}

We now use (\ref{Joint Covariance Statistic}), (\ref{Eig Eq}) and
(\ref{Inverse Covariance Matrix}) to devise vector algebra locus equations
that determine values of the scale factors for the Wolfe-dual principal
eigenaxis components that lie on $\boldsymbol{\psi}$. Let there be
$l_{1}+l_{2}=l$ scale factors.

It will be seen that likely locations and likelihood values of extreme vectors
$\left\{  k_{\mathbf{x}_{1_{i\ast}}}\right\}  _{i=1}^{l_{1}}$ and $\left\{
k_{\mathbf{x}_{2_{i\ast}}}\right\}  _{i=1}^{l_{2}}$ are \emph{statistically}
\textquotedblleft\emph{pre-wired}\textquotedblright\ within the components
$\psi_{1i\ast}\frac{k_{\mathbf{x}_{1i\ast}}}{\left\Vert k_{\mathbf{x}_{1i\ast
}}\right\Vert }$ and $\psi_{2i\ast}\frac{k_{\mathbf{x}_{2i\ast}}}{\left\Vert
k_{\mathbf{x}_{2i\ast}}\right\Vert }$ of the principal eigenvector
$\boldsymbol{\psi}_{\max}$ of the joint covariance matrix $\mathbf{Q}$ and the
inverted joint covariance matrix $\mathbf{Q}^{-1}$ associated with the pair of
random quadratic forms $\boldsymbol{\psi}^{T}\mathbf{Q}\boldsymbol{\psi}$ and
$\boldsymbol{\psi}^{T}\mathbf{Q}^{-1}\boldsymbol{\psi}$.

\subsubsection{Scale Factors for Class $\omega_{1}$}

Let $i=1:l_{1}$, such that each extreme vector $k_{\mathbf{x}_{1i\ast}}$ that
belongs to class $\omega_{1}$ is correlated with a Wolfe-dual principal
eigenaxis component $\psi_{1i\ast}\frac{k_{\mathbf{x}_{1i\ast}}}{\left\Vert
k_{\mathbf{x}_{1i\ast}}\right\Vert }$ that lies on side $\boldsymbol{\psi}%
_{1}$ of the principal eigenvector $\boldsymbol{\psi}_{\max}$. Now take the
extreme vector $k_{\mathbf{x}_{1i\ast}}$ that is correlated with the
Wolfe-dual principal eigenaxis component $\psi_{1i\ast}\frac{k_{\mathbf{x}%
_{1i\ast}}}{\left\Vert k_{\mathbf{x}_{1i\ast}}\right\Vert }$.

Using (\ref{Joint Covariance Statistic}), (\ref{Eig Eq}) and
(\ref{Inverse Covariance Matrix}), it follows that the scale factor
$\psi_{1i\ast}$ for the principal eigenaxis component $\psi_{1i\ast}%
\frac{k_{\mathbf{x}_{1i\ast}}}{\left\Vert k_{\mathbf{x}_{1i\ast}}\right\Vert
}$ that lies on side $\boldsymbol{\psi}_{1}$ is determined by the vector
algebra locus equation%
\begin{align}
\psi_{1i\ast}  &  =\lambda_{1}^{-1}\left\Vert k_{\mathbf{x}_{1_{i\ast}}%
}\right\Vert \sum\nolimits_{j=1}^{l_{1}}\psi_{1_{j\ast}}\left\Vert
k_{\mathbf{x}_{1_{j\ast}}}\right\Vert \cos\theta_{k_{\mathbf{x}_{1_{i\ast}}%
}k_{\mathbf{x}_{1_{j\ast}}}}\tag{15.6}\label{Wolf-dual Comp 1}\\
&  -\lambda_{1}^{-1}\left\Vert k_{\mathbf{x}_{1_{i\ast}}}\right\Vert
\sum\nolimits_{j=1}^{l_{2}}\psi_{2_{j\ast}}\left\Vert k_{\mathbf{x}_{2_{j\ast
}}}\right\Vert \cos\theta_{k_{\mathbf{x}_{1_{i\ast}}}k_{\mathbf{x}_{2_{j\ast}%
}}}\text{,}\nonumber
\end{align}
such that $k_{\mathbf{x}_{1_{j\ast}}}\equiv k_{\mathbf{x}_{1_{i\ast}}}$ and
$k_{\mathbf{x}_{2_{j\ast}}}\equiv k_{\mathbf{x}_{2_{i\ast}}}$, so that the
scale factors $\psi_{1_{j\ast}}$ and $\psi_{2_{j\ast}}$ map covariance and
distribution information for correlated extreme vectors $k_{\mathbf{x}%
_{1_{i\ast}}}$ and $k_{\mathbf{x}_{2_{i\ast}}}$ onto the extreme vectors
$k_{\mathbf{x}_{1_{j\ast}}}$ and $k_{\mathbf{x}_{2_{j\ast}}}$ in such a manner
that the eigenvalue scaled dual locus of $\boldsymbol{\kappa}$%
\[
\lambda_{1}^{-1}\left(  \sum\nolimits_{j=1}^{l_{1}}\psi_{1_{j\ast}%
}k_{\mathbf{x}_{1_{j\ast}}}-\sum\nolimits_{j=1}^{l_{2}}\psi_{2_{j\ast}%
}k_{\mathbf{x}_{2_{j\ast}}}\right)
\]
is mapped onto the extreme vector $k_{\mathbf{x}_{1_{i\ast}}}$ in the
following manner%
\[
\psi_{1i\ast}=\lambda_{1}^{-1}\left(  \sum\nolimits_{j=1}^{l_{1}}%
\psi_{1_{j\ast}}k_{\mathbf{x}_{1_{j\ast}}}-\sum\nolimits_{j=1}^{l_{2}}%
\psi_{2_{j\ast}}k_{\mathbf{x}_{2_{j\ast}}}\right)  k_{\mathbf{x}_{1_{i\ast}}%
}\text{,}%
\]
so that the scaled geometric locus of the novel principal eigenaxis
$\lambda_{1}^{-1}\boldsymbol{\kappa}$ contains all of the covariance and
distribution information for all of the extreme vectors $k_{\mathbf{x}%
_{1_{i\ast}}}$ and $k_{\mathbf{x}_{2_{i\ast}}}$---relative to the eigenvalues
$\lambda_{N}^{-1}\leq\mathbf{\ldots}\leq\lambda_{1}^{-1}$ of the inverted
joint covariance matrix $\mathbf{Q}^{-1}$ for a given collection $\left\{
\mathbf{x}_{i}\right\}  _{i=1}^{N}$ of feature vectors $\mathbf{x}_{i}$ ---at
which point the likelihood component $\psi_{1i\ast}\frac{k_{\mathbf{x}%
_{1i\ast}}}{\left\Vert k_{\mathbf{x}_{1i\ast}}\right\Vert }$ on side
$\boldsymbol{\psi}_{1}$ contains equivalent covariance and distribution
information for the extreme point $\mathbf{x}_{1_{i\ast}}$ that is normalized
relative to length%
\[
\psi_{1i\ast}\frac{k_{\mathbf{x}_{1i\ast}}}{\left\Vert k_{\mathbf{x}_{1i\ast}%
}\right\Vert }\equiv\lambda_{1}^{-1}\left(  \sum\nolimits_{j=1}^{l_{1}}%
\psi_{1_{j\ast}}k_{\mathbf{x}_{1_{j\ast}}}-\sum\nolimits_{j=1}^{l_{2}}%
\psi_{2_{j\ast}}k_{\mathbf{x}_{2_{j\ast}}}\right)  k_{\mathbf{x}_{1_{i\ast}}%
}\text{.}%
\]

Thereby, the likely location of the extreme point $\mathbf{x}_{1_{i\ast}}$
within the decision space $Z=Z_{1}\cup Z_{2}$ of the minimum risk binary
classification system $k_{\mathbf{s}}\boldsymbol{\kappa}+\boldsymbol{\kappa
}_{0}\overset{\omega_{1}}{\underset{\omega_{2}}{\gtrless}}0$ is determined by
the geometric locus of the principal eigenaxis component $\psi_{1i\ast
}k_{\mathbf{x}_{1i\ast}}$, whereas the likelihood value of the extreme point
$\mathbf{x}_{1_{i\ast}}$ is determined by the statistical contents of the
correlated likelihood component $\psi_{1i\ast}k_{\mathbf{x}_{1i\ast}}$.

\subsubsection{Scale Factors for Class $\omega_{2}$}

Let $i=1:l_{2}$, such that each extreme vector $k_{\mathbf{x}_{2i\ast}}$ that
belongs to class $\omega_{2}$ is correlated with a Wolfe principal eigenaxis
component $\psi_{2i\ast}\frac{k_{\mathbf{x}_{2i\ast}}}{\left\Vert
k_{\mathbf{x}_{2i\ast}}\right\Vert }$ that lies on side $\boldsymbol{\psi}%
_{2}$ of the principal eigenvector $\boldsymbol{\psi}_{\max}$. Now take the
extreme vector $k_{\mathbf{x}_{2i\ast}}$ that is correlated with the
Wolfe-dual principal eigenaxis component $\psi_{2i\ast}\frac{k_{\mathbf{x}%
_{2i\ast}}}{\left\Vert k_{\mathbf{x}_{2i\ast}}\right\Vert }$.

Using (\ref{Joint Covariance Statistic}), (\ref{Eig Eq}) and
(\ref{Inverse Covariance Matrix}), it follows that the scale factor
$\psi_{2i\ast}$ for the principal eigenaxis component $\psi_{2i\ast}%
\frac{k_{\mathbf{x}_{2i\ast}}}{\left\Vert k_{\mathbf{x}_{2i\ast}}\right\Vert
}$ that lies on side $\boldsymbol{\psi}_{2}$ is determined by the vector
algebra locus equation%
\begin{align}
\psi_{2i\ast}  &  =\lambda_{1}^{-1}\left\Vert k_{\mathbf{x}_{2_{i\ast}}%
}\right\Vert \sum\nolimits_{j=1}^{l_{2}}\psi_{2_{j\ast}}\left\Vert
k_{\mathbf{x}_{2_{j\ast}}}\right\Vert \cos\theta_{k_{\mathbf{x}_{2_{i\ast}}%
}k_{\mathbf{x}_{2_{j\ast}}}}\tag{15.7}\label{Wolf-dual Comp 2}\\
&  -\lambda_{1}^{-1}\left\Vert k_{\mathbf{x}_{2_{i\ast}}}\right\Vert
\sum\nolimits_{j=1}^{l_{1}}\psi_{1_{j\ast}}\left\Vert k_{\mathbf{x}_{1_{j\ast
}}}\right\Vert \cos\theta_{k_{\mathbf{x}_{2_{i\ast}}}k_{\mathbf{x}_{1_{j\ast}%
}}}\text{,}\nonumber
\end{align}
such that $k_{\mathbf{x}_{1_{j\ast}}}\equiv k_{\mathbf{x}_{1_{i\ast}}}$ and
$k_{\mathbf{x}_{2_{j\ast}}}\equiv k_{\mathbf{x}_{2_{i\ast}}}$, so that the
scale factors $\psi_{2_{j\ast}}$ and $\psi_{1_{j\ast}}$ map covariance and
distribution information for correlated extreme vectors $k_{\mathbf{x}%
_{1_{i\ast}}}$ and $k_{\mathbf{x}_{2_{i\ast}}}$ onto the extreme vectors
$k_{\mathbf{x}_{2_{j\ast}}}$ and $k_{\mathbf{x}_{1_{j\ast}}}$ in such a manner
that the eigenvalue scaled dual locus of $-\boldsymbol{\kappa}$%
\[
\lambda_{1}^{-1}\left(  \sum\nolimits_{j=1}^{l_{2}}\psi_{2_{j\ast}%
}k_{\mathbf{x}_{2_{j\ast}}}-\sum\nolimits_{j=1}^{l_{1}}\psi_{1_{j\ast}%
}k_{\mathbf{x}_{1_{j\ast}}}\right)
\]
is mapped onto the extreme vector $k_{\mathbf{x}_{2_{i\ast}}}$ in the
following manner%
\[
\psi_{2i\ast}=\lambda_{1}^{-1}\left(  \sum\nolimits_{j=1}^{l_{2}}%
\psi_{2_{j\ast}}k_{\mathbf{x}_{2_{j\ast}}}-\sum\nolimits_{j=1}^{l_{1}}%
\psi_{1_{j\ast}}k_{\mathbf{x}_{1_{j\ast}}}\right)  k_{\mathbf{x}_{2_{i\ast}}%
}\text{,}%
\]
so that the signed and scaled geometric locus of the novel principal eigenaxis
$-\lambda_{1}^{-1}\boldsymbol{\kappa}$ contains all of the covariance and
distribution information for all of the extreme vectors $k_{\mathbf{x}%
_{1_{i\ast}}}$ and $k_{\mathbf{x}_{2_{i\ast}}}$---relative to the eigenvalues
$\lambda_{N}^{-1}\leq\mathbf{\ldots}\leq\lambda_{1}^{-1}$ of the inverted
joint covariance matrix $\mathbf{Q}^{-1}$ for a given collection $\left\{
\mathbf{x}_{i}\right\}  _{i=1}^{N}$ of feature vectors $\mathbf{x}_{i}$ ---at
which point the likelihood component $\psi_{2i\ast}\frac{k_{\mathbf{x}%
_{2i\ast}}}{\left\Vert k_{\mathbf{x}_{2i\ast}}\right\Vert }$ on side
$\boldsymbol{\psi}_{2}$ contains equivalent covariance and distribution
information for the extreme point $\mathbf{x}_{2_{i\ast}}$ that is normalized
relative to length%
\[
\psi_{2i\ast}\frac{k_{\mathbf{x}_{2i\ast}}}{\left\Vert k_{\mathbf{x}_{2i\ast}%
}\right\Vert }\equiv\lambda_{1}^{-1}\left(  \sum\nolimits_{j=1}^{l_{2}}%
\psi_{2_{j\ast}}k_{\mathbf{x}_{2_{j\ast}}}-\sum\nolimits_{j=1}^{l_{1}}%
\psi_{1_{j\ast}}k_{\mathbf{x}_{1_{j\ast}}}\right)  k_{\mathbf{x}_{2_{i\ast}}%
}\text{.}%
\]

Thereby, the likely location of the extreme point $\mathbf{x}_{2_{i\ast}}$
within the decision space $Z=Z_{1}\cup Z_{2}$ of the minimum risk binary
classification system $k_{\mathbf{s}}\boldsymbol{\kappa}+\boldsymbol{\kappa
}_{0}\overset{\omega_{1}}{\underset{\omega_{2}}{\gtrless}}0$ is determined by
the geometric locus of the principal eigenaxis component $\psi_{2i\ast
}k_{\mathbf{x}_{2_{i\ast}}}$, whereas the likelihood value of the extreme
point $\mathbf{x}_{2_{i\ast}}$ is determined by the statistical contents of
the correlated likelihood component $\psi_{2i\ast}k_{\mathbf{x}_{2_{i\ast}}}$.

Given the locus equations in (\ref{Wolf-dual Comp 1}) and
(\ref{Wolf-dual Comp 2}), we realize that likely locations and likelihood
values of extreme vectors $\left\{  k_{\mathbf{x}_{1_{i\ast}}}\right\}
_{i=1}^{l_{1}}$ and $\left\{  k_{\mathbf{x}_{2_{i\ast}}}\right\}
_{i=1}^{l_{2}}$ are \emph{statistically} \textquotedblleft\emph{pre-wired}%
\textquotedblright\ within the components $\psi_{1i\ast}\frac{k_{\mathbf{x}%
_{1i\ast}}}{\left\Vert k_{\mathbf{x}_{1i\ast}}\right\Vert }$ and $\psi
_{2i\ast}\frac{k_{\mathbf{x}_{2i\ast}}}{\left\Vert k_{\mathbf{x}_{2i\ast}%
}\right\Vert }$ of the principal eigenvector $\boldsymbol{\psi}_{\max}$ of the
joint covariance matrix $\mathbf{Q}$ and the inverted joint covariance matrix
$\mathbf{Q}^{-1}$ associated with the pair of random quadratic forms
$\boldsymbol{\psi}^{T}\mathbf{Q}\boldsymbol{\psi}$ and $\boldsymbol{\psi}%
^{T}\mathbf{Q}^{-1}\boldsymbol{\psi}$.

\subsection{Regulation of Total Allowed Eigenenergy and Risk}

Given the expressions satisfied by the scale factors $\psi_{1i\ast}$ and
$\psi_{2i\ast}$ for the dual components $\psi_{1i\ast}k_{\mathbf{x}_{1i\ast}}$
and $\psi_{2i\ast}k_{\mathbf{x}_{2_{i\ast}}}$ of the geometric locus of the
primal novel principal eigenaxis $\boldsymbol{\kappa}=\boldsymbol{\kappa}%
_{1}-\boldsymbol{\kappa}_{2}$ in (\ref{Wolf-dual Comp 1}) and
(\ref{Wolf-dual Comp 2}), it follows that the eigenvalues $\lambda_{N}%
^{-1}\leq\mathbf{\ldots}\leq\lambda_{1}^{-1}$ of the inverted joint covariance
matrix $\mathbf{Q}^{-1}$ of the random quadratic form $\boldsymbol{\psi}%
^{T}\mathbf{Q}^{-1}\boldsymbol{\psi}$ regulate the total allowed eigenenergies
$\left\Vert \psi_{1_{i\ast}}k_{\mathbf{x}_{1i\ast}}\right\Vert _{\min_{c}}%
^{2}$ and $\left\Vert \psi_{2_{i\ast}}k_{\mathbf{x}_{2i\ast}}\right\Vert
_{\min_{c}}^{2}$ exhibited by the principal eigenaxis components $\psi
_{1i\ast}k_{\mathbf{x}_{1i\ast}}$ and $\psi_{2i\ast}k_{\mathbf{x}_{2_{i\ast}}%
}$ on the novel principal eigenaxis $\boldsymbol{\kappa}=\boldsymbol{\kappa
}_{1}-\boldsymbol{\kappa}_{2}$, along with the expected risks or counter risks
$\mathfrak{R}_{\mathfrak{\min}}\left(  \left\Vert \psi_{1_{i\ast}%
}k_{\mathbf{x}_{1i\ast}}\right\Vert _{\min_{c}}^{2}\right)  $ and
$\mathfrak{R}_{\mathfrak{\min}}\left(  \left\Vert \psi_{2_{i\ast}%
}k_{\mathbf{x}_{2i\ast}}\right\Vert _{\min_{c}}^{2}\right)  $ exhibited by the
likelihood components $\psi_{1i\ast}k_{\mathbf{x}_{1i\ast}}$ and $\psi
_{2i\ast}k_{\mathbf{x}_{2_{i\ast}}}$ on the novel principal eigenaxis
$\boldsymbol{\kappa}=\boldsymbol{\kappa}_{1}-\boldsymbol{\kappa}_{2}$.

Thereby, the total allowed eigenenergy $\left\Vert \boldsymbol{\kappa}%
_{1}-\boldsymbol{\kappa}_{2}\right\Vert _{\min_{c}}^{2}$ and the expected risk
$\mathfrak{R}_{\mathfrak{\min}}\left(  \left\Vert \boldsymbol{\kappa}%
_{1}-\boldsymbol{\kappa}_{2}\right\Vert _{\min_{c}}^{2}\right)  $ exhibited by
the geometric locus of the novel principal eigenaxis \ $\boldsymbol{\kappa
}=\boldsymbol{\kappa}_{1}-\boldsymbol{\kappa}_{2}$ are regulated by the
eigenvalues $\lambda_{N}^{-1}\leq\mathbf{\ldots}\leq\lambda_{1}^{-1}$ of the
inverted joint covariance matrix $\mathbf{Q}^{-1}$ of the random quadratic
form $\boldsymbol{\psi}^{T}\mathbf{Q}^{-1}\boldsymbol{\psi}$ .

We now devise normalized conditional density estimates for extreme points.

\subsection{Normalized Conditional Density Estimates}

Given (\ref{Joint Covariance 2}) and (\ref{Wolf-dual Comp 1}), it follows that
the geometric locus of each principal eigenaxis component $\psi_{1i\ast}%
\frac{k_{\mathbf{x}_{1i\ast}}}{\left\Vert k_{\mathbf{x}_{1i\ast}}\right\Vert
}$ that lies on side $\boldsymbol{\psi}_{1}$ of the principal eigenvector
$\boldsymbol{\psi}_{\max}$ is determined by how the first and second order
vector components of a correlated extreme vector $k_{\mathbf{x}_{1_{i_{\ast}}%
}}$ are symmetrically distributed along the eigenvalue scaled loci of $l$
signed and scaled extreme vectors $\left\{  \lambda_{1}^{-1}\psi_{1_{j\ast}%
}k_{\mathbf{x}_{1_{j\ast}}}\right\}  _{j=1}^{l_{1}}$ and $\left\{
-\lambda_{1}^{-1}\psi_{2_{j\ast}}k_{\mathbf{x}_{2_{j\ast}}}\right\}
_{j=1}^{l_{2}}$, such that each signed and scaled extreme vector
$\psi_{1_{j\ast}}k_{\mathbf{x}_{1_{j\ast}}}$ and $-\psi_{2_{j\ast}%
}k_{\mathbf{x}_{2_{j\ast}}}$ represents a symmetrically balanced distribution
of the $l$ scaled extreme vectors $\left\{  \psi_{_{k\ast}}k_{\mathbf{x}%
_{k\ast}}\right\}  _{k=1}^{l}$ that is conditional on the entire collection
$\left\{  k_{\mathbf{x}_{i}}\right\}  _{i=1}^{N}$ feature vectors
$k_{\mathbf{x}_{i}}$.

Therefore, each principal eigenaxis component $\psi_{1i\ast}\frac
{k_{\mathbf{x}_{1i\ast}}}{\left\Vert k_{\mathbf{x}_{1i\ast}}\right\Vert }$
that lies on side $\boldsymbol{\psi}_{1}$ constitutes a normalized conditional
density estimate for a correlated extreme point $\mathbf{x}_{1_{i\ast}}%
$---represented by a distribution of first and second degree coordinates of
the extreme point $\mathbf{x}_{1_{i\ast}}$---that is determined by the locus
equation in (\ref{Wolf-dual Comp 1}).

Thereby, each likelihood component $\psi_{1i\ast}\frac{k_{\mathbf{x}_{1i\ast}%
}}{\left\Vert k_{\mathbf{x}_{1i\ast}}\right\Vert }$ on that lies on side
$\boldsymbol{\psi}_{1}$ represents the distribution of a correlated extreme
point $\mathbf{x}_{1_{i\ast}}$ within the decision space $Z=Z_{1}\cup Z_{2}$
of the minimum risk binary classification system $k_{\mathbf{s}}%
\boldsymbol{\kappa}+\boldsymbol{\kappa}_{0}\overset{\omega_{1}%
}{\underset{\omega_{2}}{\gtrless}}0$, such that the likelihood component
$\psi_{1i\ast}\frac{k_{\mathbf{x}_{1i\ast}}}{\left\Vert k_{\mathbf{x}_{1i\ast
}}\right\Vert }$ determines a conditional density estimate and a conditional
likelihood value---both of which are normalized relative to length---for the
extreme point $\mathbf{x}_{1_{i\ast}}$.

Correspondingly, given (\ref{Joint Covariance 2}) and (\ref{Wolf-dual Comp 2}%
), it follows that the geometric locus of each principal eigenaxis component
$\psi_{2i\ast}\frac{k_{\mathbf{x}_{2i\ast}}}{\left\Vert k_{\mathbf{x}_{2i\ast
}}\right\Vert }$ that lies on side $\boldsymbol{\psi}_{2}$ of the principal
eigenvector $\boldsymbol{\psi}_{\max}$ constitutes a normalized conditional
density estimate for a correlated extreme point $\mathbf{x}_{2_{i\ast}}%
$---represented by a distribution of first and second degree coordinates of
the extreme point $\mathbf{x}_{2_{i\ast}}$---that is determined by the locus
equation in (\ref{Wolf-dual Comp 2}).

Thereby, each likelihood component $\psi_{2i\ast}\frac{k_{\mathbf{x}_{2i\ast}%
}}{\left\Vert k_{\mathbf{x}_{2i\ast}}\right\Vert }$ that lies on side
$\boldsymbol{\psi}_{2}$ represents the distribution of a correlated extreme
point $\mathbf{x}_{2_{i\ast}}$ within the decision space $Z=Z_{1}\cup Z_{2}$
of the minimum risk binary classification system $k_{\mathbf{s}}%
\boldsymbol{\kappa}+\boldsymbol{\kappa}_{0}\overset{\omega_{1}%
}{\underset{\omega_{2}}{\gtrless}}0$, such that the likelihood component
$\psi_{2i\ast}\frac{k_{\mathbf{x}_{2i\ast}}}{\left\Vert k_{\mathbf{x}_{2i\ast
}}\right\Vert }$ determines a conditional density estimate and a conditional
likelihood value---both of which are normalized relative to length---for the
extreme point $\mathbf{x}_{2_{i\ast}}$.

We are now in a position to devise conditional probability density functions
for class $\omega_{1}$ and class $\omega_{2}$.

\subsection{Conditional Probability Density Functions}

Given that each principal component $\psi_{1i\ast}\frac{k_{\mathbf{x}_{1i\ast
}}}{\left\Vert k_{\mathbf{x}_{1i\ast}}\right\Vert }$ that lies on side
$\boldsymbol{\psi}_{1}$ also represents a likelihood component $\psi_{1i\ast
}\frac{k_{\mathbf{x}_{1i\ast}}}{\left\Vert k_{\mathbf{x}_{1i\ast}}\right\Vert
}$ that determines a normalized conditional density estimate and a normalized
conditional likelihood value for a correlated extreme point $\mathbf{x}%
_{1_{i\ast}}$, it follows each principal eigenaxis component $\psi_{1_{i\ast}%
}k_{\mathbf{x}_{1_{i\ast}}}$ that lies on side $\boldsymbol{\kappa}_{1}$ also
represents a likelihood component $\psi_{1_{i\ast}}k_{\mathbf{x}_{1_{i\ast}}}$
that determines a conditional density estimate and a conditional likelihood
value for a correlated extreme point $\mathbf{x}_{1_{i\ast}}$---both of which
are conditional on all of the likelihood components $\psi_{1_{i\ast}%
}k_{\mathbf{x}_{1_{i\ast}}}$ and $-\psi_{2_{i\ast}}k_{\mathbf{x}_{2_{i\ast}}}$
that lie on the novel principal eigenaxis $\boldsymbol{\kappa}$.

Correspondingly, given that each principal component $\psi_{2i\ast}%
\frac{k_{\mathbf{x}_{2i\ast}}}{\left\Vert k_{\mathbf{x}_{2i\ast}}\right\Vert
}$ that lies on side $\boldsymbol{\psi}_{2}$ also represents a likelihood
component $\psi_{2i\ast}\frac{k_{\mathbf{x}_{2i\ast}}}{\left\Vert
k_{\mathbf{x}_{2i\ast}}\right\Vert }$ that determines a normalized conditional
density estimate and a normalized conditional likelihood value for a
correlated extreme point $\mathbf{x}_{2_{i\ast}}$, it follows that each
principal eigenaxis component $\psi_{2_{i\ast}}k_{\mathbf{x}_{2_{i\ast}}}$
that lies on side $\boldsymbol{\kappa}_{2}$ also represents a likelihood
component $\psi_{2_{i\ast}}k_{\mathbf{x}_{2_{i\ast}}}$ that determines a
conditional density estimate and a conditional likelihood value for a
correlated extreme point $\mathbf{x}_{2_{i\ast}}$---both of which are
conditional on all of the likelihood components $\psi_{2_{i\ast}}%
k_{\mathbf{x}_{2_{i\ast}}}$ and $-\psi_{1_{i\ast}}k_{\mathbf{x}_{1_{i\ast}}}$
that lie on the signed novel principal eigenaxis $-\boldsymbol{\kappa}$.

Thereby, it follows that principal eigenaxis components $\psi_{1_{i\ast}%
}k_{\mathbf{x}_{1_{i\ast}}}$ and $\psi_{2_{i\ast}}k_{\mathbf{x}_{2_{i\ast}}}$
and correlated likelihood components $\psi_{1_{i\ast}}k_{\mathbf{x}_{1_{i\ast
}}}$ and $\psi_{2_{i\ast}}k_{\mathbf{x}_{2_{i\ast}}}$ are identically and
symmetrically distributed over both sides $\boldsymbol{\kappa}_{1}$ and
$\boldsymbol{\kappa}_{2}$ of the geometric locus of the primal novel principal
eigenaxis%
\begin{align*}
\boldsymbol{\kappa}  &  =\sum\nolimits_{i=1}^{l_{1}}\psi_{1_{i\ast}%
}k_{\mathbf{x}_{1_{i\ast}}}-\sum\nolimits_{i=1}^{l_{2}}\psi_{2_{i\ast}%
}k_{\mathbf{x}_{2_{i\ast}}}\\
&  =\boldsymbol{\kappa}_{1}{\large -}\boldsymbol{\kappa}_{2}\text{,}%
\end{align*}
so that the dual locus of $\boldsymbol{\kappa}_{1}=\sum\nolimits_{i=1}^{l_{1}%
}\psi_{1_{i\ast}}k_{\mathbf{x}_{1_{i\ast}}}$ is a conditional probability
density function $p\left(  \mathbf{x}_{1_{i\ast}}|\boldsymbol{\kappa}%
_{1}\right)  $ that determines distributions of extreme points $\mathbf{x}%
_{1_{i\ast}}$ that are conditional on the likelihood components $\psi
_{1_{i\ast}}k_{\mathbf{x}_{1_{i\ast}}}$and $-\psi_{2_{i\ast}}k_{\mathbf{x}%
_{2_{i\ast}}}$ on $\boldsymbol{\kappa}$, and the dual locus of
$\boldsymbol{\kappa}_{2}=\sum\nolimits_{i=1}^{l_{2}}\psi_{2_{i\ast}%
}k_{\mathbf{x}_{2_{i\ast}}}$ is a conditional probability density function
$p\left(  \mathbf{x}_{2_{i\ast}}|\boldsymbol{\kappa}_{2}\right)  $ that
determines distributions of extreme points $\mathbf{x}_{2_{i\ast}}$ that are
conditional on the likelihood components $-\psi_{1_{i\ast}}k_{\mathbf{x}%
_{1_{i\ast}}}$and $\psi_{2_{i\ast}}k_{\mathbf{x}_{2_{i\ast}}}$ on
$-\boldsymbol{\kappa}$.

We now devise conditional probability functions for class $\omega_{1}$ and
class $\omega_{2}$.

\subsection{Conditional Probability Function for Class $\omega_{1}$}

Let the side $\boldsymbol{\kappa}_{1}$ of the geometric locus of the novel
principal eigenaxis $\boldsymbol{\kappa}=\boldsymbol{\kappa}_{1}%
{\large -}\boldsymbol{\kappa}_{2}$ be a conditional probability density
function $p\left(  \mathbf{x}_{1_{i\ast}}|\boldsymbol{\kappa}_{1}\right)  $
for class $\omega_{1}$, such that feature vectors $\mathbf{x}_{1_{i}}\in%
\mathbb{R}
^{d}$ that belong to class $\omega_{1}$ are generated by a certain probability
density function $\mathbf{\mathbf{x}_{1_{i}}\sim}$ $p\left(  \mathbf{x}%
;\omega_{1}\right)  $.

Then the integral of the conditional probability density function $p\left(
\mathbf{x}_{1_{i\ast}}|\boldsymbol{\kappa}_{1}\right)  $ for class $\omega
_{1}$%
\begin{align*}
P\left(  \mathbf{x}_{1_{i\ast}}|\boldsymbol{\kappa}_{1}\right)   &  =\int%
_{Z}\left(  \sum\nolimits_{i=1}^{l_{1}}\psi_{1_{i\ast}}k_{\mathbf{x}%
_{1_{i\ast}}}\right)  d\boldsymbol{\kappa}_{1}=\int_{Z}p\left(  k_{\mathbf{x}%
_{1i\ast}}|\boldsymbol{\kappa}_{1}\right)  d\boldsymbol{\kappa}_{1}\\
&  =\int_{Z}\boldsymbol{\kappa}_{1}d\boldsymbol{\kappa}_{1}=\int_{Z_{1}%
}\boldsymbol{\kappa}_{1}d\boldsymbol{\kappa}_{1}+\int_{Z_{2}}%
\boldsymbol{\kappa}_{1}d\boldsymbol{\kappa}_{1}\\
&  =\frac{1}{2}\left\Vert \boldsymbol{\kappa}_{1}\right\Vert ^{2}+C=\left\Vert
\boldsymbol{\kappa}_{1}\right\Vert ^{2}+C_{1}\text{,}%
\end{align*}
over the decision space $Z$ of the minimum risk binary classification system
$k_{\mathbf{s}}\boldsymbol{\kappa}+\mathbf{\kappa}_{0}\overset{\omega
_{1}}{\underset{\omega_{2}}{\gtrless}}0$, determines the conditional
probability $P\left(  \mathbf{x}_{1_{i\ast}}|\boldsymbol{\kappa}_{1}\right)  $
of observing a set $\left\{  \mathbf{x}_{1_{i\ast}}\right\}  _{i=1}^{l_{1}}$
of $l_{1}$ extreme points $\mathbf{x}_{1_{i\ast}}$ located within localized
areas of the decision space $Z=Z_{1}\cup Z_{2}$ of the system, so that the
probability of finding an extreme point $\mathbf{x}_{1_{i\ast}}$ within the
decision region $Z_{1}$ determines a region of counter risk, and the
probability of finding an extreme point $\mathbf{x}_{1_{i\ast}}$ within the
decision region $Z_{2}$ determines a region of risk.

\subsubsection{Costs for Right and Wrong Decisions}

Given the integral of the conditional probability density function $p\left(
\mathbf{x}_{1_{i\ast}}|\boldsymbol{\kappa}_{1}\right)  $ for class $\omega
_{1}$, it follows that eigenenergies $\left\Vert \psi_{1_{i\ast}}%
k_{\mathbf{x}_{1i\ast}}\right\Vert _{\min_{c}}^{2}$ related to likely
locations of extreme points $\mathbf{x}_{1_{i\ast}}$ within the decision
region $Z_{1}$ determine \emph{costs} for expected counter risks of making
\emph{right decisions}, whereas eigenenergies $\left\Vert \psi_{1_{i\ast}%
}k_{\mathbf{x}_{1i\ast}}\right\Vert _{\min_{c}}^{2}$ related to likely
locations of extreme points $\mathbf{x}_{1_{i\ast}}$ within the decision
region $Z_{2}$ determine \emph{costs} for expected risks of making \emph{wrong
decisions}.

Therefore, the conditional probability function $P\left(  \mathbf{x}%
_{1_{i\ast}}|\boldsymbol{\kappa}_{1}\right)  $ for class $\omega_{1}$ is given
by the integral%
\begin{align}
P\left(  \mathbf{x}_{1_{i\ast}}|\boldsymbol{\kappa}_{1}\right)   &  =\int%
_{Z}\boldsymbol{\kappa}_{1}d\boldsymbol{\kappa}_{1}=\int_{Z_{1}}%
\boldsymbol{\kappa}_{1}d\boldsymbol{\kappa}_{1}+\int_{Z_{2}}\boldsymbol{\kappa
}_{1}d\boldsymbol{\kappa}_{1}\tag{15.8}\label{Cond-Prob-Function Class 1}\\
&  =\left\Vert \boldsymbol{\kappa}_{1}\right\Vert _{\min_{c}}^{2}%
+C_{1}\text{,}\nonumber
\end{align}
over the decision space $Z=Z_{1}\cup Z_{2}$ of the minimum risk binary
classification system $k_{\mathbf{s}}\boldsymbol{\kappa}+\boldsymbol{\kappa
}_{0}\overset{\omega_{1}}{\underset{\omega_{2}}{\gtrless}}0$, so that the
integral has a solution in terms of the total allowed eigenenergy $\left\Vert
\boldsymbol{\kappa}_{1}\right\Vert _{\min_{c}}^{2}$ exhibited by
$\boldsymbol{\kappa}_{1}$ and a certain integration constant $C_{1}$.

\subsection{Conditional Probability Function for Class $\omega_{2}$}

Let the side $\boldsymbol{\kappa}_{2}$ of the geometric locus of the novel
principal eigenaxis $\boldsymbol{\kappa}=\boldsymbol{\kappa}_{1}%
{\large -}\boldsymbol{\kappa}_{2}$ be a conditional probability density
function $p\left(  \mathbf{x}_{2_{i\ast}}|\boldsymbol{\kappa}_{2}\right)  $
for class $\omega_{2}$, such that feature vectors $\mathbf{x}_{2_{i}}\in%
\mathbb{R}
^{d}$ that belong to class $\omega_{2}$ are generated by a certain probability
density function $\mathbf{\mathbf{x}_{2_{i}}\sim}$ $p\left(  \mathbf{x}%
;\omega_{2}\right)  $.

Then the integral of the conditional probability density function $p\left(
\mathbf{x}_{2_{i\ast}}|\boldsymbol{\kappa}_{2}\right)  $ for class $\omega
_{2}$%
\begin{align*}
P\left(  \mathbf{x}_{2_{i\ast}}|\boldsymbol{\kappa}_{2}\right)   &  =\int%
_{Z}\left(  \sum\nolimits_{i=1}^{l_{2}}\psi_{2_{i\ast}}k_{\mathbf{x}%
_{2_{i\ast}}}\right)  d\boldsymbol{\kappa}_{2}=\int_{Z}p\left(  k_{\mathbf{x}%
_{2i\ast}}|\boldsymbol{\kappa}_{2}\right)  d\boldsymbol{\kappa}_{2}\\
&  =\int_{Z}\boldsymbol{\kappa}_{2}d\boldsymbol{\kappa}_{2}=\int_{Z_{1}%
}\boldsymbol{\kappa}_{2}d\boldsymbol{\kappa}_{2}+\int_{Z_{2}}%
\boldsymbol{\kappa}_{2}d\boldsymbol{\kappa}_{2}\\
&  =\frac{1}{2}\left\Vert \boldsymbol{\kappa}_{2}\right\Vert ^{2}+C=\left\Vert
\boldsymbol{\kappa}_{2}\right\Vert ^{2}+C_{2}\text{,}%
\end{align*}
over the decision space $Z$ of the minimum risk binary classification system
$k_{\mathbf{s}}\boldsymbol{\kappa}+\mathbf{\kappa}_{0}\overset{\omega
_{1}}{\underset{\omega_{2}}{\gtrless}}0$, determines the conditional
probability $P\left(  \mathbf{x}_{2_{i\ast}}|\boldsymbol{\kappa}_{2}\right)  $
of observing a set $\left\{  \mathbf{x}_{2_{i\ast}}\right\}  _{i=1}^{l_{2}}$
of $l_{2}$ extreme points $\mathbf{x}_{2_{i\ast}}$ located within localized
areas of the decision space $Z=Z_{1}\cup Z_{2}$ of the system, so that the
probability of finding an extreme point $\mathbf{x}_{2_{i\ast}}$ within the
decision region $Z_{1}$ determines a region of risk, and the probability of
finding an extreme point $\mathbf{x}_{2_{i\ast}}$ within the decision region
$Z_{2}$ determines a region of counter risk.

\subsubsection{Costs for Right and Wrong Decisions}

Given the integral of the conditional probability density function $p\left(
\mathbf{x}_{2_{i\ast}}|\boldsymbol{\kappa}_{2}\right)  $ for class $\omega
_{2}$, it follows that eigenenergies $\left\Vert \psi_{2_{i\ast}}%
k_{\mathbf{x}_{2i\ast}}\right\Vert _{\min_{c}}^{2}$ related to likely
locations of extreme points $\mathbf{x}_{2_{i\ast}}$ within the decision
region $Z_{1}$ determine \emph{costs} for expected risks of making \emph{wrong
decisions}, whereas eigenenergies $\left\Vert \psi_{2_{i\ast}}k_{\mathbf{x}%
_{2i\ast}}\right\Vert _{\min_{c}}^{2}$ related to likely locations of extreme
points $\mathbf{x}_{2_{i\ast}}$ within the decision region $Z_{2}$ determine
\emph{costs} for expected counter risks of making \emph{right decisions}.

Therefore, the conditional probability function $P\left(  \mathbf{x}%
_{2_{i\ast}}|\boldsymbol{\kappa}_{2}\right)  $ for class $\omega_{2}$ is given
by the integral%
\begin{align}
P\left(  \mathbf{x}_{2_{i\ast}}|\boldsymbol{\kappa}_{2}\right)   &  =\int%
_{Z}\boldsymbol{\kappa}_{2}d\boldsymbol{\kappa}_{2}=\int_{Z_{1}}%
\boldsymbol{\kappa}_{2}d\boldsymbol{\kappa}_{2}+\int_{Z_{2}}\boldsymbol{\kappa
}_{2}d\boldsymbol{\kappa}_{2}\tag{15.9}\label{Cond-Prob-Function Class 2}\\
&  =\left\Vert \boldsymbol{\kappa}_{2}\right\Vert _{\min_{c}}^{2}%
+C_{2}\text{,}\nonumber
\end{align}
over the decision space $Z=Z_{1}\cup Z_{2}$ of the minimum risk binary
classification system $k_{\mathbf{s}}\boldsymbol{\kappa}+\boldsymbol{\kappa
}_{0}\overset{\omega_{1}}{\underset{\omega_{2}}{\gtrless}}0$, so that the
integral has a solution in terms of the total allowed eigenenergy $\left\Vert
\boldsymbol{\kappa}_{2}\right\Vert _{\min_{c}}^{2}$ exhibited by
$\boldsymbol{\kappa}_{2}$ and a certain integration constant $C_{2}$.

In the next section of our treatise, we demonstrate how the geometric locus of
the novel principal eigenaxis $\boldsymbol{\kappa=\kappa}_{1}{\large -}%
\boldsymbol{\kappa}_{2}$ of any given minimum risk binary classification
system $k_{\mathbf{s}}\boldsymbol{\kappa}+\boldsymbol{\kappa}_{0}%
\overset{\omega_{1}}{\underset{\omega_{2}}{\gtrless}}0$ represents an
exclusive principal eigen-coordinate system and an eigenaxis of symmetry that
spans the decision space $Z=Z_{1}\cup Z_{2}$ of the system, so that all of the
points that lie on the geometric loci of the decision boundary $k_{\mathbf{s}%
}\boldsymbol{\kappa}+\mathbf{\kappa}_{0}=0$ and a pair of symmetrically
positioned decision borders $k_{\mathbf{s}}\boldsymbol{\kappa}%
+\boldsymbol{\kappa}_{0}=+1$ and $k_{\mathbf{s}}\boldsymbol{\kappa
}+\boldsymbol{\kappa}_{0}=-1$ of the minimum risk binary classification system
$k_{\mathbf{s}}\boldsymbol{\kappa}+\boldsymbol{\kappa}_{0}\overset{\omega
_{1}}{\underset{\omega_{2}}{\gtrless}}0$ exclusively reference the geometric
locus of the novel principal eigenaxis $\boldsymbol{\kappa=\kappa}%
_{1}{\large -}\boldsymbol{\kappa}_{2}$, at which point the geometric locus of
the decision boundary $k_{\mathbf{s}}\boldsymbol{\kappa}+\boldsymbol{\kappa
}_{0}=0$ partitions the decision space $Z=Z_{1}\cup Z_{2}$ into symmetrical
decision regions $Z_{1}$ and $Z_{2}$ that are bounded by the geometric loci of
the pair of symmetrically positioned decision borders $k_{\mathbf{s}%
}\boldsymbol{\kappa}+\boldsymbol{\kappa}_{0}=+1$ and $k_{\mathbf{s}%
}\boldsymbol{\kappa}+\boldsymbol{\kappa}_{0}=-1$.

\section{\label{Section 16}Geometric Partitioning of Decision Spaces}

The capacity of a geometric locus of a novel principal eigenaxis
$\boldsymbol{\kappa=\kappa}_{1}{\large -}\boldsymbol{\kappa}_{2}$ to partition
the decision space of any given minimum risk binary classification system
$k_{\mathbf{s}}\boldsymbol{\kappa}+\boldsymbol{\kappa}_{0}\overset{\omega
_{1}}{\underset{\omega_{2}}{\gtrless}}0$---for any given collection $\left\{
\mathbf{x}_{i}\right\}  _{i=1}^{N}$ of $N$ feature vectors $\mathbf{x}_{i}%
$---is determined by the KKT condition in (\ref{KKT 5}) and the KKT condition
of complementary slackness
\citep{Sundaram1996}%
, so that a vector algebra locus equation is not active
\[
y_{i}\left(  k_{\mathbf{x}_{i}}\boldsymbol{\kappa}+\boldsymbol{\kappa}%
_{0}\right)  -1+\xi_{i}>0
\]
if a corresponding constraint is not active
\[
\psi_{i}=0\text{.}%
\]

\subsection{Capacity of a Novel Principal Eigenaxis}

It will be seen that the \emph{capacity} of the geometric locus of any given
novel principal eigenaxis $\boldsymbol{\kappa=\kappa}_{1}{\large -}%
\boldsymbol{\kappa}_{2}$ is based on elegant statistical relations and
deep-seated statistical interconnections between the geometric loci of a
primal novel principal eigenaxis $\boldsymbol{\kappa}=\boldsymbol{\kappa}%
_{1}-\boldsymbol{\kappa}_{2}$ and a Wolfe-dual novel principal eigenaxis
$\boldsymbol{\psi}=\boldsymbol{\psi}_{1}+\boldsymbol{\psi}_{2}$ of a minimum
risk binary classification system $k_{\mathbf{s}}\boldsymbol{\kappa
}+\boldsymbol{\kappa}_{0}\overset{\omega_{1}}{\underset{\omega_{2}}{\gtrless}%
}0$, such that the novel principal eigenaxis $\boldsymbol{\kappa=\kappa}%
_{1}{\large -}\boldsymbol{\kappa}_{2}$ is the solution of vector algebra locus
equations that represent the geometric loci of a decision boundary and a pair
of symmetrically positioned decision borders, at which point the geometric
locus of the novel principal eigenaxis $\boldsymbol{\kappa}=\boldsymbol{\kappa
}_{1}-\boldsymbol{\kappa}_{2}$ is the principal eigenaxis of the geometric
loci of the decision boundary and the pair of symmetrically positioned
decision borders, so that the novel principal eigenaxis $\boldsymbol{\kappa
}=\boldsymbol{\kappa}_{1}-\boldsymbol{\kappa}_{2}$ is an eigenaxis of symmetry
that spans the decision space of the system $k_{\mathbf{s}}\boldsymbol{\kappa
}+\boldsymbol{\kappa}_{0}\overset{\omega_{1}}{\underset{\omega_{2}}{\gtrless}%
}0$.

\subsection{KKT Conditions and Constraints}

Let there be $l$ active constraints, where $l=l_{1}+l_{2}$, so that $l$ locus
equations $y_{i}\left(  k_{\mathbf{x}_{i\ast}}\boldsymbol{\kappa
}+\boldsymbol{\kappa}_{0}\right)  -1+\xi_{i}=0,\ i=1,...,l$ are active, at
which point $l$ extreme vectors $k_{\mathbf{x}_{i\ast}}$ from class
$\omega_{1}$ and class $\omega_{2}$ are correlated with $l$ scale factors
$\psi_{i\ast}$ that have certain positive values $\psi_{i\ast}>0$.

It follows that the KKT condition in (\ref{KKT 5}) and the KKT condition of
complementary slackness
\citep{Sundaram1996}
determine the following system of vector algebra locus equations%
\[
y_{i}\left(  k_{\mathbf{x}_{i\ast}}\boldsymbol{\kappa}+\boldsymbol{\kappa}%
_{0}\right)  -1+\xi_{i}=0,\ i=1,...,l\text{,}%
\]
which are satisfied by a geometric locus of a novel principal eigenaxis
$\boldsymbol{\kappa}$ of a minimum risk binary classification system
$k_{\mathbf{s}}\boldsymbol{\kappa}+\mathbf{\kappa}_{0}\overset{\omega
_{1}}{\underset{\omega_{2}}{\gtrless}}0$, so that the discriminant function%
\begin{equation}
d\left(  \mathbf{s}\right)  =k_{\mathbf{s}}\boldsymbol{\kappa}%
+\boldsymbol{\kappa}_{0} \tag{16.1}\label{Discriminant F1}%
\end{equation}
of the minimum risk binary classification system $k_{\mathbf{s}}%
\boldsymbol{\kappa}+\boldsymbol{\kappa}_{0}\overset{\omega_{1}%
}{\underset{\omega_{2}}{\gtrless}}0$ is the solution of the vector algebra
locus equations%
\[
d\left(  \mathbf{s}\right)  =0\text{, \ }d\left(  \mathbf{s}\right)
=+1\text{, and \ }d\left(  \mathbf{s}\right)  =-1\text{,}%
\]
where $d\left(  \mathbf{s}\right)  =0$ denotes the geometric locus of a
quadratic or nearly linear decision boundary that symmetrically partitions the
decision space $Z=Z_{1}\cup Z_{2}$ of the minimum risk binary classification
system $k_{\mathbf{s}}\boldsymbol{\kappa}+\boldsymbol{\kappa}_{0}%
\overset{\omega_{1}}{\underset{\omega_{2}}{\gtrless}}0$ into symmetrical
decision regions $Z_{1}$ and $Z_{2}$ that are bounded by the geometric loci of
a pair of symmetrically positioned decision borders $d\left(  \mathbf{s}%
\right)  =+1$ and $d\left(  \mathbf{s}\right)  =-1$, where $d\left(
\mathbf{s}\right)  =+1$ denotes the geometric locus of the decision border for
the decision region $Z_{1}$, and $d\left(  \mathbf{s}\right)  =-1$ denotes the
geometric locus of the decision border for the decision region $Z_{2}$.

The KKT condition in (\ref{KKT 5}) and the KKT condition of complementary
slackness also determine the following system of vector algebra locus
equations%
\[
y_{i}\left(  k_{\mathbf{x}_{i\ast}}\boldsymbol{\kappa}+\boldsymbol{\kappa}%
_{0}\right)  -1+\xi_{i}=0,\ i=1,...,l\text{,}%
\]
which are satisfied by both $\boldsymbol{\kappa}_{0}$ and $\boldsymbol{\kappa
}$, so that $\boldsymbol{\kappa}_{0}$ is related to $\boldsymbol{\kappa}$ in
the following manner%
\begin{equation}
\boldsymbol{\kappa}_{0}=\frac{1}{l}\sum\nolimits_{i=1}^{l}y_{i}\left(
1-\xi_{i}\right)  -\left(  \frac{1}{l}\sum\nolimits_{i=1}^{l}k_{\mathbf{x}%
_{i\ast}}\right)  \boldsymbol{\kappa}\text{.} \tag{16.2}%
\label{Funct Primal Dual Locus}%
\end{equation}

\subsection{The Discriminant Function}

Using the expression for the discriminant function in (\ref{Discriminant F1})
and the equation for $\boldsymbol{\kappa}_{0}$\ in
(\ref{Funct Primal Dual Locus}), the discriminant function is rewritten as%
\begin{equation}
d\left(  \mathbf{s}\right)  =\left(  k_{\mathbf{s}}\boldsymbol{-}\frac{1}%
{l}\sum\nolimits_{i=1}^{l}k_{\mathbf{x}_{i\ast}}\right)  \boldsymbol{\kappa
}+\frac{1}{l}\sum\nolimits_{i=1}^{l}y_{i}\left(  1-\xi_{i}\right)  \text{,}
\tag{16.3}\label{Discriminant F2}%
\end{equation}
so that the discriminant function is represented by the geometric locus of the
novel principal eigenaxis $\boldsymbol{\kappa}$.

\subsection{Vector Algebra Locus Equations of Decision Spaces}

We now devise vector algebra locus equations that represent the geometric loci
of a decision boundary and a pair of symmetrically positioned decision borders
that jointly partition the decision space $Z=Z_{1}\cup Z_{2}$ of the minimum
risk binary classification system $k_{\mathbf{s}}\boldsymbol{\kappa
}+\boldsymbol{\kappa}_{0}\overset{\omega_{1}}{\underset{\omega_{2}}{\gtrless}%
}0$ into symmetrical decision regions $Z_{1}$ and $Z_{2}$ that jointly
delineate the decision space of the system.

\subsubsection{Quadratic or Nearly Linear Decision Boundaries}

Substituting the expression for the discriminant function in
(\ref{Discriminant F2}) into the vector algebra locus equation $d\left(
\mathbf{s}\right)  =0$ determines a vector algebra locus equation%
\begin{equation}
\left(  k_{\mathbf{s}}\boldsymbol{-}\frac{1}{l}\sum\nolimits_{i=1}%
^{l}k_{\mathbf{x}_{i\ast}}\right)  \boldsymbol{\kappa}+\frac{1}{l}%
\sum\nolimits_{i=1}^{l}y_{i}\left(  1-\xi_{i}\right)  =0 \tag{16.4}%
\label{Decision Boundary}%
\end{equation}
that represents the geometric locus of a quadratic or nearly linear decision
boundary, so that the discriminant function $\boldsymbol{\kappa}$ is the
solution of \ref{Decision Boundary}, at which point all of the points
$\mathbf{s}$ that lie on the geometric locus of the decision boundary
$d\left(  \mathbf{s}\right)  =0$ exclusively reference the novel principal
eigenaxis $\boldsymbol{\kappa}$.

\subsubsection{Quadratic or Nearly Linear Decision Borders}

Substituting the expression for the discriminant function in
(\ref{Discriminant F2}) into the vector algebra locus equation $d\left(
\mathbf{s}\right)  =+1$ determines a vector algebra locus equation%
\begin{equation}
\left(  k_{\mathbf{s}}\boldsymbol{-}\frac{1}{l}\sum\nolimits_{i=1}%
^{l}k_{\mathbf{x}_{i\ast}}\right)  \boldsymbol{\kappa}+\frac{1}{l}%
\sum\nolimits_{i=1}^{l}y_{i}\left(  1-\xi_{i}\right)  =+1 \tag{16.5}%
\label{Decision Border 1}%
\end{equation}
that represents the geometric locus of a quadratic or nearly linear decision
border, so that the discriminant function $\boldsymbol{\kappa}$ is the
solution of \ref{Decision Border 1}, at which point all of the points
$\mathbf{s}$ that lie on the geometric locus of the decision border $d\left(
\mathbf{s}\right)  =+1$ exclusively reference the principal eigenaxis
$\boldsymbol{\kappa}$.

Substituting the expression for the discriminant function in
(\ref{Discriminant F2}) into the vector algebra locus equation $d\left(
\mathbf{s}\right)  =-1$ determines a vector algebra locus equation%
\begin{equation}
\left(  k_{\mathbf{s}}\boldsymbol{-}\frac{1}{l}\sum\nolimits_{i=1}%
^{l}k_{\mathbf{x}_{i\ast}}\right)  \boldsymbol{\kappa}+\frac{1}{l}%
\sum\nolimits_{i=1}^{l}y_{i}\left(  1-\xi_{i}\right)  =-1 \tag{16.6}%
\label{Decision Border 2}%
\end{equation}
that represents the geometric locus of a quadratic or nearly linear decision
border, so that the discriminant function $\boldsymbol{\kappa}$ is the
solution of \ref{Decision Border 2}, at which point all of the points
$\mathbf{s}$ that lie on the geometric locus of the decision border $d\left(
\mathbf{s}\right)  =-1$ exclusively reference the novel principal eigenaxis
$\boldsymbol{\kappa}$.

\subsection{Symmetrical Decision Regions}

By (\ref{Decision Boundary}) - (\ref{Decision Border 2}), it follows that the
discriminant function%
\[
d\left(  \mathbf{s}\right)  :\left(  k_{\mathbf{s}}\boldsymbol{-}\frac{1}%
{l}\sum\nolimits_{i=1}^{l}k_{\mathbf{x}_{i\ast}}\right)  \boldsymbol{\kappa
}+\frac{1}{l}\sum\nolimits_{i=1}^{l}y_{i}\left(  1-\xi_{i}\right)
\]
is the solution of the vector algebra locus equations of
(\ref{Decision Boundary}) - (\ref{Decision Border 2}), so that graphs of the
vector algebra locus equations of (\ref{Decision Boundary}) -
(\ref{Decision Border 2}) represent the geometric loci of a quadratic or
nearly linear decision boundary $d\left(  \mathbf{s}\right)  =0$ and a pair of
symmetrically positioned decision borders $d\left(  \mathbf{s}\right)  =+1$
and $d\left(  \mathbf{s}\right)  =-1$ that jointly partition the decision
space $Z=Z_{1}\cup Z_{2}$ of the minimum risk binary classification system
$k_{\mathbf{s}}\boldsymbol{\kappa}+\boldsymbol{\kappa}_{0}\overset{\omega
_{1}}{\underset{\omega_{2}}{\gtrless}}0$ into symmetrical decision regions
$Z_{1}$ and $Z_{2}$---wherein $Z_{1}\simeq Z_{2}$---that cover the decision
space $Z=Z_{1}\cup Z_{2}$ in a symmetrically balanced manner, at which point
balanced portions of the extreme points $\mathbf{x}_{1_{i\ast}}$ and
$\mathbf{x}_{2_{i\ast}}$ from class $\omega_{1}$ and class $\omega_{2}$ have
locations throughout the decision regions $Z_{1}$ and $Z_{2}$ that account for
right and wrong decisions of the minimum risk binary classification system
$k_{\mathbf{s}}\boldsymbol{\kappa}+\boldsymbol{\kappa}_{0}\overset{\omega
_{1}}{\underset{\omega_{2}}{\gtrless}}0$.

\subsection{Eigenaxis of Symmetry}

By the vector algebra locus equations of (\ref{Decision Boundary}) -
(\ref{Decision Border 2}), it follows that the geometric locus of the novel
principal eigenaxis $\boldsymbol{\kappa}=\boldsymbol{\kappa}_{1}%
-\boldsymbol{\kappa}_{2}$ represents an eigenaxis of symmetry that spans the
decision space $Z=Z_{1}\cup Z_{2}$ of the minimum risk binary classification
system $k_{\mathbf{s}}\boldsymbol{\kappa}+\boldsymbol{\kappa}_{0}%
\overset{\omega_{1}}{\underset{\omega_{2}}{\gtrless}}0$, so that the decision
regions $Z_{1}$ and $Z_{2}$ are symmetrically partitioned by the geometric
locus of a quadratic or nearly linear decision boundary $d\left(
\mathbf{s}\right)  =0$ that is represented by the graph of the vector algebra
locus equation of (\ref{Decision Boundary}), and the span of the decision
regions $Z_{1}$ and $Z_{2}$ is regulated by the vector algebra locus equations
of (\ref{Decision Border 1}) and (\ref{Decision Border 2}) that represent the
geometric loci of a pair of symmetrically positioned decision borders
$d\left(  \mathbf{s}\right)  =+1$ and $d\left(  \mathbf{s}\right)  =-1$, at
which point the geometric loci of the decision borders $d\left(
\mathbf{s}\right)  =+1$ and $d\left(  \mathbf{s}\right)  =-1$ jointly
delineate the decision space $Z=Z_{1}\cup Z_{2}$ of the system $k_{\mathbf{s}%
}\boldsymbol{\kappa}+$ $\boldsymbol{\kappa}_{0}\overset{\omega_{1}%
}{\underset{\omega_{2}}{\gtrless}}0$.

We have determined that all of the symmetrical balancing feats outlined
above---which are exhibited by the geometric locus of a novel principal
eigenaxis $\boldsymbol{\kappa}=\boldsymbol{\kappa}_{1}-\boldsymbol{\kappa}%
_{2}$---are facilitated by certain balancing feats within the Wolfe dual
principal eigenspace of $\boldsymbol{\psi}$ and $\boldsymbol{\kappa}$.

\section{\label{Section 17}Balancing Feats in Principal Eigenspace}

The machine learning algorithm that is currently being examined executes
surprising statistical balancing feats---between all of the principal
eigenaxis components and likelihood components that lie on both sides of the
geometric loci of the novel principal eigenaxes $\boldsymbol{\psi}$ and
$\boldsymbol{\kappa}$---within the Wolfe-dual principal eigenspace of
$\boldsymbol{\psi}$ and $\boldsymbol{\kappa}$.

We now reveal statistical balancing feats---that are coincident with a minimum
risk binary classification system $k_{\mathbf{s}}\boldsymbol{\kappa}+$
$\boldsymbol{\kappa}_{0}\overset{\omega_{1}}{\underset{\omega_{2}}{\gtrless}%
}0$ acting to jointly minimize its eigenenergy and risk---by identifying
equilibrium requirements on the dual loci of $\boldsymbol{\psi}$ and
$\boldsymbol{\kappa}$---inside the Wolfe-dual principal eigenspace of
$\boldsymbol{\psi}$ and $\boldsymbol{\kappa}$. We originally identified these
balancing feats in our working paper
\citep{Reeves2018design}%
.

We demonstrate that these statistical balancing feats are coincident with a
minimum risk binary classification system $k_{\mathbf{s}}\boldsymbol{\kappa}+$
$\boldsymbol{\kappa}_{0}\overset{\omega_{1}}{\underset{\omega_{2}}{\gtrless}%
}0$ acting to jointly minimize its eigenenergy and risk, so that the system
locates a point of equilibrium---at which point the dual locus of the
discriminant function of the system is in statistical equilibrium---at the
geometric locus of the decision boundary of the system.

Moreover, for any given collection $\left\{  \mathbf{x}_{1i\ast}\right\}
_{i=1}^{l_{1}}$ and $\left\{  \mathbf{x}_{2i\ast}\right\}  _{i=1}^{l_{2}}$ of
extreme points $\mathbf{x}_{1i\ast}$ and $\mathbf{x}_{2i\ast}$ that have been
generated by certain probability density functions $p\left(  \mathbf{x}%
;\omega_{1}\right)  $ and $p\left(  \mathbf{x};\omega_{2}\right)  $ for two
classes $\omega_{1}$ and $\omega_{2}$ of random vectors $\mathbf{x\in}$ $%
\mathbb{R}
^{d}$, it will be seen that each and every one of these statistical balancing
feats is \emph{enabled} by the manner in which likely locations and likelihood
values for each and every one of the extreme points $\mathbf{x}_{1_{i\ast}}$
and $\mathbf{x}_{2_{i\ast}}\mathbf{\ }$within the decision space $Z=Z_{1}\cup
Z_{2}$ of a minimum risk binary classification system $k_{\mathbf{s}%
}\boldsymbol{\kappa}+$ $\boldsymbol{\kappa}_{0}\overset{\omega_{1}%
}{\underset{\omega_{2}}{\gtrless}}0$ are \emph{statistically}
\textquotedblleft\emph{pre-wired}\textquotedblright\ within the geometric
\emph{locus} of the novel \emph{principal eigenaxis} $\boldsymbol{\kappa
}=\boldsymbol{\kappa}_{1}-\boldsymbol{\kappa}_{2}$ of the system.

We begin by identifying equilibrium requirements on the components
$\psi_{1i\ast}\frac{k_{\mathbf{x}_{1i\ast}}}{\left\Vert k_{\mathbf{x}_{1i\ast
}}\right\Vert }$ and $\psi_{2i\ast}\frac{k_{\mathbf{x}_{2i\ast}}}{\left\Vert
k_{\mathbf{x}_{2i\ast}}\right\Vert }$ of the principal eigenvector
$\boldsymbol{\psi}_{\max}$ of the joint covariance matrix $\mathbf{Q}$ and the
inverted joint covariance matrix $\mathbf{Q}^{-1}$ associated with the pair of
random quadratic forms $\boldsymbol{\psi}^{T}\mathbf{Q}\boldsymbol{\psi}$ and
$\boldsymbol{\psi}^{T}\mathbf{Q}^{-1}\boldsymbol{\psi}$.

\subsection{The Wolfe-dual Equilibrium Point}

By the KKT condition in (\ref{KKT 2}), it follows that the vector algebra
locus equation of the Wolf-dual equilibrium point%
\begin{equation}
\sum\nolimits_{i=1}^{l_{1}}\psi_{1i\ast}\frac{k_{\mathbf{x}_{1i\ast}}%
}{\left\Vert k_{\mathbf{x}_{1i\ast}}\right\Vert }-\sum\nolimits_{i=1}^{l_{2}%
}\psi_{2i\ast}\frac{k_{\mathbf{x}_{2i\ast}}}{\left\Vert k_{\mathbf{x}_{2i\ast
}}\right\Vert }=0 \tag{17.1}\label{Wolfe-dual EQU Point}%
\end{equation}
of the minimum risk binary classification system $k_{\mathbf{s}}%
\boldsymbol{\kappa}+$ $\boldsymbol{\kappa}_{0}\overset{\omega_{1}%
}{\underset{\omega_{2}}{\gtrless}}0$ is determined by the following
equilibrium requirement%
\[
\left(  y_{i}=1\right)  \sum\nolimits_{i=1}^{l_{1}}\psi_{1_{i\ast}}+\left(
y_{i}=-1\right)  \sum\nolimits_{i=1}^{l_{2}}\psi_{2_{i\ast}}=0
\]
on the scale factors $\psi_{1_{i\ast}}$ and $\psi_{2_{i\ast}}$ for the
components $\psi_{1i\ast}\frac{k_{\mathbf{x}_{1i\ast}}}{\left\Vert
k_{\mathbf{x}_{1i\ast}}\right\Vert }$ and $\psi_{2i\ast}\frac{k_{\mathbf{x}%
_{2i\ast}}}{\left\Vert k_{\mathbf{x}_{2i\ast}}\right\Vert }$ of the principal
eigenvector $\boldsymbol{\psi}_{\max}$ of the joint covariance matrix
$\mathbf{Q}$ and the inverted joint covariance matrix $\mathbf{Q}^{-1}$
associated with the pair of random quadratic forms $\boldsymbol{\psi}%
^{T}\mathbf{Q}\boldsymbol{\psi}$ and $\boldsymbol{\psi}^{T}\mathbf{Q}%
^{-1}\boldsymbol{\psi}$, so that the geometric locus of the Wolfe-dual novel
principal eigenaxis $\boldsymbol{\psi}$ satisfies a state of statistical
equilibrium%
\begin{equation}
\sum\nolimits_{i=1}^{l_{1}}\psi_{1i\ast}\frac{k_{\mathbf{x}_{1i\ast}}%
}{\left\Vert k_{\mathbf{x}_{1i\ast}}\right\Vert }=\sum\nolimits_{i=1}^{l_{2}%
}\psi_{2i\ast}\frac{k_{\mathbf{x}_{2i\ast}}}{\left\Vert k_{\mathbf{x}_{2i\ast
}}\right\Vert }\text{,} \tag{17.2}\label{Wolfe-dual EQU State}%
\end{equation}
at which point counteracting and opposing forces and influences of the system
are symmetrically balanced with each other.

Returning to the locus equations in (\ref{Wolf-dual Comp 1}) and
(\ref{Wolf-dual Comp 2}), recall that likely locations and likelihood values
of extreme vectors $\left\{  k_{\mathbf{x}_{1_{i\ast}}}\right\}  _{i=1}%
^{l_{1}}$ and $\left\{  k_{\mathbf{x}_{2_{i\ast}}}\right\}  _{i=1}^{l_{2}}$
are \emph{statistically} \textquotedblleft\emph{pre-wired}\textquotedblright%
\ within the components $\psi_{1i\ast}\frac{k_{\mathbf{x}_{1i\ast}}%
}{\left\Vert k_{\mathbf{x}_{1i\ast}}\right\Vert }$ and $\psi_{2i\ast}%
\frac{k_{\mathbf{x}_{2i\ast}}}{\left\Vert k_{\mathbf{x}_{2i\ast}}\right\Vert
}$ of the principal eigenvector $\boldsymbol{\psi}_{\max}$.

By (\ref{Wolf-dual Dual Locus}) and (\ref{Wolfe-dual EQU State}), it follows
that the geometric locus of the Wolfe-dual novel principal eigenaxis
$\boldsymbol{\psi}=\boldsymbol{\psi}_{1}+\boldsymbol{\psi}_{2}$ exhibits
symmetrical dimensions and densities, so that the critical minimum eigenenergy
$\left\Vert \boldsymbol{\psi}_{1}\right\Vert _{\min_{c}}^{2}$ exhibited by all
of the principal eigenaxis components $\psi_{1i\ast}\frac{k_{\mathbf{x}%
_{1i\ast}}}{\left\Vert k_{\mathbf{x}_{1i\ast}}\right\Vert }$ on side
$\boldsymbol{\psi}_{1}$ is symmetrically balanced with the critical minimum
eigenenergy $\left\Vert \boldsymbol{\psi}_{2}\right\Vert _{\min_{c}}^{2}$
exhibited by all of the principal eigenaxis components $\psi_{2i\ast}%
\frac{k_{\mathbf{x}_{2i\ast}}}{\left\Vert k_{\mathbf{x}_{2i\ast}}\right\Vert
}$ on side $\boldsymbol{\psi}_{2}$%
\[
\left\Vert \boldsymbol{\psi}_{1}\right\Vert _{\min_{c}}^{2}=\left\Vert
\boldsymbol{\psi}_{2}\right\Vert _{\min_{c}}^{2}\text{,}%
\]
the length of side $\boldsymbol{\psi}_{1}$ equals the length of side
$\boldsymbol{\psi}_{2}$%
\[
\left\Vert \boldsymbol{\psi}_{1}\right\Vert =\left\Vert \boldsymbol{\psi}%
_{2}\right\Vert \text{,}%
\]
and counteracting and opposing forces and influences of the minimum risk
binary classification system $k_{\mathbf{s}}\boldsymbol{\kappa}+$
$\boldsymbol{\kappa}_{0}\overset{\omega_{1}}{\underset{\omega_{2}}{\gtrless}%
}0$ are symmetrically balanced with each other about the geometric center of
the Wolfe-dual novel principal eigenaxis $\boldsymbol{\psi}$%
\begin{align*}
&  \left\Vert \boldsymbol{\psi}_{1}\right\Vert \left(  \sum\nolimits_{i=1}%
^{l_{1}}\operatorname{comp}_{\overrightarrow{\boldsymbol{\psi}_{1}}}\left(
\overrightarrow{\psi_{1i\ast}\frac{k_{\mathbf{x}_{1i\ast}}}{\left\Vert
k_{\mathbf{x}_{1i\ast}}\right\Vert }}\right)  -\sum\nolimits_{i=1}^{l_{2}%
}\operatorname{comp}_{\overrightarrow{\boldsymbol{\psi}_{1}}}\left(
\overrightarrow{\psi_{2i\ast}\frac{k_{\mathbf{x}_{2i\ast}}}{\left\Vert
k_{\mathbf{x}_{2i\ast}}\right\Vert }}\right)  \right) \\
&  =\left\Vert \boldsymbol{\psi}_{2}\right\Vert \left(  \sum\nolimits_{i=1}%
^{l_{2}}\operatorname{comp}_{\overrightarrow{\boldsymbol{\psi}_{2}}}\left(
\overrightarrow{\psi_{2i\ast}\frac{k_{\mathbf{x}_{2i\ast}}}{\left\Vert
k_{\mathbf{x}_{2i\ast}}\right\Vert }}\right)  -\sum\nolimits_{i=1}^{l_{1}%
}\operatorname{comp}_{\overrightarrow{\boldsymbol{\psi}_{2}}}\left(
\overrightarrow{\psi_{1i\ast}\frac{k_{\mathbf{x}_{1i\ast}}}{\left\Vert
k_{\mathbf{x}_{1i\ast}}\right\Vert }}\right)  \right)  \text{,}%
\end{align*}
whereon the statistical fulcrum of the Wolfe-dual novel principal eigenaxis
$\boldsymbol{\psi}=\boldsymbol{\psi}_{1}+\boldsymbol{\psi}_{2}$ is located.

Thereby, counteracting and opposing components of critical minimum
eigenenergies related to likely locations of extreme points from class
$\omega_{1}$ and class $\omega_{2}$ that determine regions of counter risks
and risks of the system---along the dual locus of side $\boldsymbol{\psi}_{1}%
$---are symmetrically balanced with counteracting and opposing components of
critical minimum eigenenergies related to likely locations of extreme points
from class $\omega_{2}$ and class $\omega_{1}$ that determine regions of
counter risks and risks of the system---along the dual locus of side
$\boldsymbol{\psi}_{2}$.

It will be seen that the geometric locus of the primal novel principal
eigenaxis $\boldsymbol{\kappa}=\sum\nolimits_{i=1}^{l_{1}}\psi_{1_{i\ast}%
}k_{\mathbf{x}_{1_{i\ast}}}-\sum\nolimits_{i=1}^{l_{2}}\psi_{2_{i\ast}%
}k_{\mathbf{x}_{2_{i\ast}}}$ routinely achieves an equivalent statistical
balancing feat.

\subsection{An Elegant Statistical Balancing Feat}

The constrained optimization algorithm that is being examined finds the right
mix of principal eigenaxis components on both $\boldsymbol{\psi}$ and
$\boldsymbol{\kappa}$ by executing a surprisingly complex and elegant
statistical balancing feat---inside the Wolfe-dual principal eigenspace of
$\boldsymbol{\psi}$ and $\boldsymbol{\kappa}$:

Using (\ref{Wolf-dual Comp 1}), it follows that the summed scale factors
$\sum\nolimits_{i=1}^{l_{1}}\psi_{1i\ast}$ for the principal eigenaxis
components on side $\boldsymbol{\psi}_{1}$ of the principal eigenvector
$\boldsymbol{\psi}_{\max}$ satisfy the vector algebra locus equation%
\begin{equation}
\sum\nolimits_{i=1}^{l_{1}}\psi_{1i\ast}=\lambda_{1}^{-1}\sum\nolimits_{i=1}%
^{l_{1}}k_{\mathbf{x}_{1_{i\ast}}}\left(  \sum\nolimits_{j=1}^{l_{1}}%
\psi_{1_{j\ast}}k_{\mathbf{x}_{1_{j\ast}}}-\sum\nolimits_{j=1}^{l_{2}}%
\psi_{2_{j\ast}}k_{\mathbf{x}_{2_{j\ast}}}\right)  \text{.} \tag{17.3}%
\label{ID 1 Wolf-dual}%
\end{equation}

Using (\ref{Wolf-dual Comp 2}), it follows that the summed scale factors
$\sum\nolimits_{i=1}^{l_{2}}\psi_{2i\ast}$ for the principal eigenaxis
components on side $\boldsymbol{\psi}_{2}$ of the principal eigenvector
$\boldsymbol{\psi}_{\max}$ satisfy the vector algebra locus equation%
\begin{equation}
\sum\nolimits_{i=1}^{l_{2}}\psi_{2i\ast}=\lambda_{1}^{-1}\sum\nolimits_{i=1}%
^{l_{2}}k_{\mathbf{x}_{2_{i\ast}}}\left(  \sum\nolimits_{j=1}^{l_{2}}%
\psi_{2_{j\ast}}k_{\mathbf{x}_{2_{j\ast}}}-\sum\nolimits_{j=1}^{l_{1}}%
\psi_{1_{j\ast}}k_{\mathbf{x}_{1_{j\ast}}}\right)  \text{.} \tag{17.4}%
\label{ID 2 Wolf-dual}%
\end{equation}

Using the equilibrium equation that is satisfied by the geometric locus of the
Wolfe-dual novel principal eigenaxis $\boldsymbol{\psi}$ in
(\ref{Wolfe-dual EQU State}), it follows that the expression contained in the
right-hand side of (\ref{ID 1 Wolf-dual}) equals the expression contained in
the right-hand side of (\ref{ID 2 Wolf-dual})%
\begin{align*}
&  \lambda_{1}^{-1}\sum\nolimits_{i=1}^{l_{1}}k_{\mathbf{x}_{1_{i\ast}}%
}\left(  \sum\nolimits_{j=1}^{l_{1}}\psi_{1_{j\ast}}k_{\mathbf{x}_{1_{j\ast}}%
}-\sum\nolimits_{j=1}^{l_{2}}\psi_{2_{j\ast}}k_{\mathbf{x}_{2_{j\ast}}}\right)
\\
&  =\lambda_{1}^{-1}\sum\nolimits_{i=1}^{l_{2}}k_{\mathbf{x}_{2_{i\ast}}%
}\left(  \sum\nolimits_{j=1}^{l_{2}}\psi_{2_{j\ast}}k_{\mathbf{x}_{2_{j\ast}}%
}-\sum\nolimits_{j=1}^{l_{1}}\psi_{1_{j\ast}}k_{\mathbf{x}_{1_{j\ast}}%
}\right)  \text{,}%
\end{align*}
so that the geometric locus of the novel principal eigenaxis
$\boldsymbol{\kappa=\kappa}_{1}\mathbf{-}\boldsymbol{\kappa}_{2}$ satisfies
the vector algebra locus equation%
\begin{equation}
\sum\nolimits_{i=1}^{l_{1}}k_{\mathbf{x}_{1_{i\ast}}}\left(
\boldsymbol{\kappa}_{1}\mathbf{-}\boldsymbol{\kappa}_{2}\right)
=\sum\nolimits_{i=1}^{l_{2}}k_{\mathbf{x}_{2_{i\ast}}}\left(
\boldsymbol{\kappa}_{2}\mathbf{-}\boldsymbol{\kappa}_{1}\right)  \text{,}
\tag{17.5}\label{Primal EQ State 1}%
\end{equation}
at which point extreme vectors $k_{\mathbf{x}_{1_{i\ast}}}$ and $k_{\mathbf{x}%
_{2_{i\ast}}}$from class $\omega_{1}$ and class $\omega_{2}$ are distributed
over side $\boldsymbol{\kappa}_{1}$ and side $\boldsymbol{\kappa}_{2}$ of the
primal novel principal eigenaxis $\boldsymbol{\kappa}$ in a symmetrically
balanced manner, so that likely locations and likelihood values of each and
every one of the extreme vectors $\left\{  k_{\mathbf{x}_{1_{i\ast}}}\right\}
_{i=1}^{l_{1}}$ and $\left\{  k_{\mathbf{x}_{2_{i\ast}}}\right\}
_{i=1}^{l_{2}}$---relative to each and every one of the principal eigenaxis
components and likelihood components that lie on the geometric loci of
$\boldsymbol{\kappa}_{1}$ and $\boldsymbol{\kappa}_{2}$---are in
\emph{statistical equilibrium} with each other.

Moreover, by the vector algebra locus equation of (\ref{Primal EQ State 1}),
we realize that likely locations and likelihood values of each and every one
of the extreme vectors $\left\{  k_{\mathbf{x}_{1_{i\ast}}}\right\}
_{i=1}^{l_{1}}$ and $\left\{  k_{\mathbf{x}_{2_{i\ast}}}\right\}
_{i=1}^{l_{2}}$ are \emph{statistically} \textquotedblleft\emph{pre-wired}%
\textquotedblright\ within the geometric locus of the novel principal
eigenaxis $\boldsymbol{\kappa}=\boldsymbol{\kappa}_{1}{\large -}%
\boldsymbol{\kappa}_{2}$---relative to each and every one of the principal
eigenaxis components and likelihood components $\psi_{1_{i\ast}}%
k_{\mathbf{x}_{1_{i\ast}}}$and $\psi_{2_{i\ast}}k_{\mathbf{x}_{2_{i\ast}}}$
that lies on the geometric loci of side $\boldsymbol{\kappa}_{1}$ and side
$\boldsymbol{\kappa}_{2}$.

\subsection{The Primal Equilibrium Point}

Using the equilibrium equation that is satisfied by the primal novel principal
eigenaxis $\boldsymbol{\kappa}=\boldsymbol{\kappa}_{1}-\boldsymbol{\kappa}%
_{2}$ in (\ref{Primal EQ State 1}), it follows that the vector algebra locus
equation%
\begin{equation}
\left(  \sum\nolimits_{i=1}^{l_{1}}k_{\mathbf{x}_{1_{i\ast}}}+\sum
\nolimits_{i=1}^{l_{2}}k_{\mathbf{x}_{2_{i\ast}}}\right)  \left(
\boldsymbol{\kappa}_{1}-\boldsymbol{\kappa}_{2}\right)  =0 \tag{17.6}%
\label{Primal Equilibrium Point}%
\end{equation}
determines the primal equilibrium point of a minimum risk binary
classification system, so that likely locations and likelihood values of each
and every one of the extreme vectors $\left\{  k_{\mathbf{x}_{1_{i\ast}}%
}\right\}  _{i=1}^{l_{1}}$ and $\left\{  k_{\mathbf{x}_{2_{i\ast}}}\right\}
_{i=1}^{l_{2}}$---relative to each and every one of the principal eigenaxis
components and likelihood components $\psi_{1_{i\ast}}k_{\mathbf{x}_{1_{i\ast
}}}$and $\psi_{2_{i\ast}}k_{\mathbf{x}_{2_{i\ast}}}$ that lie on the geometric
loci of side $\boldsymbol{\kappa}_{1}$ and side $\boldsymbol{\kappa}_{2}$ of
the primal novel principal eigenaxis $\boldsymbol{\kappa}=\boldsymbol{\kappa
}_{1}-\boldsymbol{\kappa}_{2}$---are in statistical equilibrium with each
other%
\[
\left(  \sum\nolimits_{i=1}^{l_{1}}k_{\mathbf{x}_{1_{i\ast}}}+\sum
\nolimits_{i=1}^{l_{2}}k_{\mathbf{x}_{2_{i\ast}}}\right)  \boldsymbol{\kappa
}_{1}=\left(  \sum\nolimits_{i=1}^{l_{1}}k_{\mathbf{x}_{1_{i\ast}}}%
+\sum\nolimits_{i=1}^{l_{2}}k_{\mathbf{x}_{2_{i\ast}}}\right)
\boldsymbol{\kappa}_{2}\text{,}%
\]
at which point right and wrong decisions made by the minimum risk binary
classification system $k_{\mathbf{s}}\boldsymbol{\kappa}+$ $\boldsymbol{\kappa
}_{0}\overset{\omega_{1}}{\underset{\omega_{2}}{\gtrless}}0$ are symmetrically
balanced with each other.

By the equilibrium equation in (\ref{Discriminant Function in Equilibrium})
expressed by Corollary \ref{Equilibrium Requirement Corollary} and the vector
algebra locus equation of (\ref{Primal Equilibrium Point}), we realize that a
geometric locus of a novel principal eigenaxis $\boldsymbol{\kappa
}=\boldsymbol{\kappa}_{1}-\boldsymbol{\kappa}_{2}$ satisfies an
\emph{equilibrium requirement} for a \emph{discriminant function} of a minimum
risk binary classification system---that is satisfied \emph{at} the geometric
locus of the \emph{decision boundary} of the system.

The vector algebra locus equation of (\ref{Primal Equilibrium Point}) also
illustrates how likely locations of extreme points $\mathbf{x}_{1i\ast}$ and
$\mathbf{x}_{2i\ast}$ are symmetrically balanced with each other---with
respect to and in relation to---an exclusive principal eigen-coordinate system
and an eigenaxis of symmetry that spans the decision space of a minimum risk
binary classification system $k_{\mathbf{s}}\boldsymbol{\kappa}+$
$\boldsymbol{\kappa}_{0}\overset{\omega_{1}}{\underset{\omega_{2}}{\gtrless}%
}0$, each of which are represented by a geometric locus of a novel principal
eigenaxis $\boldsymbol{\kappa}=\boldsymbol{\kappa}_{1}-\boldsymbol{\kappa}%
_{2}$, so that extreme points $\mathbf{x}_{1i\ast}$ and $\mathbf{x}_{2i\ast}$
from class $\omega_{1}$ and class $\omega_{2}$ are symmetrically distributed
throughout the decision space of the minimum risk binary classification system.

\section{\label{Section 18}Dual Capacities of Novel Principal Eigenaxes}

So far, we have demonstrated that any given discriminant function of a minimum
risk binary classification system exhibits certain dual capacities, such that
the discriminant function of the system, the exclusive principal
eigen-coordinate system of the geometric locus of the decision boundary of the
system, and the eigenaxis of symmetry that spans the decision space of the
system---are each represented by a geometric locus of a novel principal
eigenaxis---which has the structure of a dual locus of likelihood components
and principal eigenaxis components.

We now summarize the essence of these dual capacities.

\subsection{Dual Locus of a Decision Space}

Take any given collection $\left\{  \mathbf{x}_{1i\ast}\right\}  _{i=1}%
^{l_{1}}$ and $\left\{  \mathbf{x}_{2i\ast}\right\}  _{i=1}^{l_{2}}$ of
extreme points $\mathbf{x}_{1i\ast}$ and $\mathbf{x}_{2i\ast}$ that are
generated by any given probability density functions $p\left(  \mathbf{x}%
;\omega_{1}\right)  $ and $p\left(  \mathbf{x};\omega_{2}\right)  $ for two
classes $\omega_{1}$ and $\omega_{2}$ of random vectors $\mathbf{x\in}$ $%
\mathbb{R}
^{d}$, such that likely locations of the collection $\left\{  \mathbf{x}%
_{1i\ast}\right\}  _{i=1}^{l_{1}}$ and $\left\{  \mathbf{x}_{2i\ast}\right\}
_{i=1}^{l_{2}}$ of the extreme points $\mathbf{x}_{1i\ast}$ and $\mathbf{x}%
_{2i\ast}$ effectively cover the decision space $Z=Z_{1}\cup Z_{2}$ of a
minimum risk binary classification system $k_{\mathbf{s}}\boldsymbol{\kappa}+$
$\boldsymbol{\kappa}_{0}\overset{\omega_{1}}{\underset{\omega_{2}}{\gtrless}%
}0$.

Returning to the locus equations in (\ref{Wolf-dual Comp 1}) and
(\ref{Wolf-dual Comp 2}), along with the vector algebra locus equation of
(\ref{Primal EQ State 1}), recall that likelihood values and likely locations
of all of the extreme points $\left\{  \mathbf{x}_{1i\ast}\right\}
_{i=1}^{l_{1}}$ and $\left\{  \mathbf{x}_{2i\ast}\right\}  _{i=1}^{l_{2}}$ are
\emph{statistically} \emph{\textquotedblleft pre-wired\textquotedblright%
}\ within the geometric locus of a novel principal eigenaxis
$\boldsymbol{\kappa}=\boldsymbol{\kappa}_{1}{\large -}\boldsymbol{\kappa}_{2}%
$---relative to each and every likelihood component $\psi_{1_{i\ast}%
}k_{\mathbf{x}_{1_{i\ast}}}$and $\psi_{2_{i\ast}}k_{\mathbf{x}_{2_{i\ast}}}$
and principal eigenaxis component $\psi_{1_{i\ast}}k_{\mathbf{x}_{1_{i\ast}}}%
$and $\psi_{2_{i\ast}}k_{\mathbf{x}_{2_{i\ast}}}$ that lies on the dual locus
$\boldsymbol{\kappa}=\boldsymbol{\kappa}_{1}{\large -}\boldsymbol{\kappa}_{2}$
of the discriminant function, the exclusive intrinsic eigen-coordinate system
$\boldsymbol{\kappa}=\boldsymbol{\kappa}_{1}{\large -}\boldsymbol{\kappa}_{2}%
$, and the eigenaxis of symmetry $\boldsymbol{\kappa}=\boldsymbol{\kappa}%
_{1}{\large -}\boldsymbol{\kappa}_{2}$ that spans the decision space
$Z=Z_{1}\cup Z_{2}$ of the minimum risk binary classification system
$k_{\mathbf{s}}\boldsymbol{\kappa}+$ $\boldsymbol{\kappa}_{0}\overset{\omega
_{1}}{\underset{\omega_{2}}{\gtrless}}0$.

Thereby, it follows that a geometric locus of a novel principal eigenaxis
$\boldsymbol{\kappa}=\boldsymbol{\kappa}_{1}{\large -}\boldsymbol{\kappa}_{2}$
represents a dual locus of the decision space $Z=Z_{1}\cup Z_{2}$ of any given
minimum risk binary classification system $k_{\mathbf{s}}\boldsymbol{\kappa}+$
$\boldsymbol{\kappa}_{0}\overset{\omega_{1}}{\underset{\omega_{2}}{\gtrless}%
}0$, such that likelihood values and likely locations of each extreme point
$\mathbf{x}_{1i\ast}$ and $\mathbf{x}_{2i\ast}$ from class $\omega_{1}$ and
class $\omega_{2}$ are \emph{statistically} \emph{pre-wired} within the dual
locus $\boldsymbol{\kappa}=\boldsymbol{\kappa}_{1}{\large -}\boldsymbol{\kappa
}_{2}$ of the system---relative to likelihood values and likely locations of
all of the extreme points $\left\{  \mathbf{x}_{1i\ast}\right\}  _{i=1}%
^{l_{1}}$ and $\left\{  \mathbf{x}_{2i\ast}\right\}  _{i=1}^{l_{2}}$ from
class $\omega_{1}$ and class $\omega_{2}$.

\subsection{Dual Capacities of a\ Novel Principal Eigenaxis}

We have demonstrated that a geometric locus of a novel principal eigenaxis
$\boldsymbol{\kappa}=\boldsymbol{\kappa}_{1}-\boldsymbol{\kappa}_{2}$
determines a dual locus of the entire decision space $Z=Z_{1}\cup Z_{2}$ of
any given minimum risk binary classification system $k_{\mathbf{s}%
}\boldsymbol{\kappa}+$ $\boldsymbol{\kappa}_{0}\overset{\omega_{1}%
}{\underset{\omega_{2}}{\gtrless}}0$ in such a manner that the geometric locus
of the novel principal eigenaxis $\boldsymbol{\kappa}=\boldsymbol{\kappa}%
_{1}-\boldsymbol{\kappa}_{2}$ is an exclusive principal eigen-coordinate
system and an eigenaxis of symmetry that spans the decision space of the
system, such that the dual locus $\boldsymbol{\kappa}=\boldsymbol{\kappa}%
_{1}-\boldsymbol{\kappa}_{2}$ contains all of the covariance and distribution
information for all of the extreme points $\mathbf{x}_{1_{i\ast}}\mathbf{\ }%
$and $\mathbf{x}_{2_{i\ast}}$---relative to the covariance and distribution
information for a given collection $\left\{  \mathbf{x}_{i}\right\}
_{i=1}^{N}$ of feature vectors $\mathbf{x}_{i}$ ---so that the geometric locus
of the novel principal eigenaxis $\boldsymbol{\kappa}=\boldsymbol{\kappa}%
_{1}-\boldsymbol{\kappa}_{2}$ determines likely locations and likelihood
values for each and every one of the extreme points $\mathbf{x}_{1_{i\ast}}$
and $\mathbf{x}_{2_{i\ast}}\mathbf{\ }$within the decision space $Z=Z_{1}\cup
Z_{2}$ of the system $k_{\mathbf{s}}\boldsymbol{\kappa}+$ $\boldsymbol{\kappa
}_{0}\overset{\omega_{1}}{\underset{\omega_{2}}{\gtrless}}0$, at which point
all of the extreme points $\mathbf{x}_{1_{i\ast}}\mathbf{\ }$and
$\mathbf{x}_{2_{i\ast}}$ effectively \emph{cover} the decision space
$Z=Z_{1}\cup Z_{2}$ of the system.

Correspondingly, we have demonstrated that the geometric locus of the novel
principal eigenaxis $\boldsymbol{\kappa}=\boldsymbol{\kappa}_{1}%
-\boldsymbol{\kappa}_{2}$ of any given minimum risk binary classification
system $k_{\mathbf{s}}\boldsymbol{\kappa}+$ $\boldsymbol{\kappa}%
_{0}\overset{\omega_{1}}{\underset{\omega_{2}}{\gtrless}}0$ exhibits certain
dual capacities---such that likely locations and likelihood values for each
and every one of the extreme points $\mathbf{x}_{1_{i\ast}}$ and
$\mathbf{x}_{2_{i\ast}}\mathbf{\ }$within the decision space $Z=Z_{1}\cup
Z_{2}$ of the system $k_{\mathbf{s}}\boldsymbol{\kappa}+$ $\boldsymbol{\kappa
}_{0}\overset{\omega_{1}}{\underset{\omega_{2}}{\gtrless}}0$ are
\emph{statistically} \emph{pre-wired}\ within the geometric locus of the novel
principal eigenaxis $\boldsymbol{\kappa}=\boldsymbol{\kappa}_{1}%
-\boldsymbol{\kappa}_{2}$ of the system $k_{\mathbf{s}}\boldsymbol{\kappa}+$
$\boldsymbol{\kappa}_{0}\overset{\omega_{1}}{\underset{\omega_{2}}{\gtrless}%
}0$, at which point the geometric locus of the novel principal eigenaxis
$\boldsymbol{\kappa}=\boldsymbol{\kappa}_{1}-\boldsymbol{\kappa}_{2}$
represents an exclusive principal eigen-coordinate system of the geometric
loci of the decision boundary $k_{\mathbf{s}}\boldsymbol{\kappa}+$
$\boldsymbol{\kappa}_{0}=0$ and a pair of symmetrically positioned decision
borders $k_{\mathbf{s}}\boldsymbol{\kappa}+\boldsymbol{\kappa}_{0}=+1$ and
$k_{\mathbf{s}}\boldsymbol{\kappa}+\boldsymbol{\kappa}_{0}=-1$ of the system
$k_{\mathbf{s}}\boldsymbol{\kappa}+$ $\boldsymbol{\kappa}_{0}\overset{\omega
_{1}}{\underset{\omega_{2}}{\gtrless}}0$, and also represents an eigenaxis of
symmetry that spans the decision space $Z=Z_{1}\cup Z_{2}$ of the system
$k_{\mathbf{s}}\boldsymbol{\kappa}+$ $\boldsymbol{\kappa}_{0}\overset{\omega
_{1}}{\underset{\omega_{2}}{\gtrless}}0$.

\subsection{Simulation Examples}

By way of demonstration, we now present simulation examples for three binary
classification systems---each of which illustrates the dual capacities
exhibited by the geometric locus of a novel principal eigenaxis
$\boldsymbol{\kappa}=\boldsymbol{\kappa}_{1}-\boldsymbol{\kappa}_{2}$ of a
minimum risk binary classification system $k_{\mathbf{s}}\boldsymbol{\kappa}+$
$\boldsymbol{\kappa}_{0}\overset{\omega_{1}}{\underset{\omega_{2}}{\gtrless}%
}0$.

\subsubsection{Simulation Example One}

Consider a minimum risk binary classification system $k_{\mathbf{s}%
}\boldsymbol{\kappa}+$ $\boldsymbol{\kappa}_{0}\overset{\omega_{1}%
}{\underset{\omega_{2}}{\gtrless}}0$ for two classes of random vectors that
have dissimilar covariance matrices, such that the covariance matrices for
class $\omega_{1}$ and class $\omega_{2}$ are given by%
\[
\Sigma_{1}=\left[
\begin{array}
[c]{cc}%
25 & 0\\
0 & 2
\end{array}
\right]  \text{, \ \ }\Sigma_{2}=\left[
\begin{array}
[c]{cc}%
2 & 0\\
0 & 25
\end{array}
\right]  \text{,}%
\]
the mean vector for class $\omega_{1}$ is given by $M_{1}=%
\begin{pmatrix}
3, & 1
\end{pmatrix}
^{T}$ and the mean vector for class $\omega_{2}$ is given by $M_{2}=%
\begin{pmatrix}
3, & -1
\end{pmatrix}
^{T}$. The error rate of the minimum risk binary classification system
$k_{\mathbf{s}}\boldsymbol{\kappa}+$ $\boldsymbol{\kappa}_{0}\overset{\omega
_{1}}{\underset{\omega_{2}}{\gtrless}}0$ is $16.9\%$.

\subsubsection{Illustrations of Dual Capacities}

Figure $16$ illustrates how the geometric locus of a novel principal eigenaxis
$\boldsymbol{\kappa}=\boldsymbol{\kappa}_{1}-\boldsymbol{\kappa}_{2}$ jointly
represents an exclusive principal eigen-coordinate system and an eigenaxis of
symmetry that spans the decision space $Z=Z_{1}\cup Z_{2}$ of the minimum risk
binary classification system $k_{\mathbf{s}}\boldsymbol{\kappa}+$
$\boldsymbol{\kappa}_{0}\overset{\omega_{1}}{\underset{\omega_{2}}{\gtrless}%
}0$, wherein the novel principal eigenaxis $\boldsymbol{\kappa}%
=\boldsymbol{\kappa}_{1}-\boldsymbol{\kappa}_{2}$ is the solution of vector
algebra locus equations that represent the geometric loci of a hyperbolic
decision boundary $d\left(  \mathbf{s}\right)  =0$ and a pair of symmetrically
positioned hyperbolic decision borders $d\left(  \mathbf{s}\right)  =+1$ and
$d\left(  \mathbf{s}\right)  =-1$---that jointly partition the decision space
$Z=Z_{1}\cup Z_{2}$ of the minimum risk binary classification system
$k_{\mathbf{s}}\boldsymbol{\kappa}+$ $\boldsymbol{\kappa}_{0}\overset{\omega
_{1}}{\underset{\omega_{2}}{\gtrless}}0$ into symmetrical decision regions
$Z_{1}$ and $Z_{2}$.

The hyperbolic decision boundary $d\left(  \mathbf{s}\right)  =0$ is black,
the hyperbolic decision border $d\left(  \mathbf{s}\right)  =+1$ is red, the
hyperbolic decision border $d\left(  \mathbf{s}\right)  =-1$ is blue, and all
of the extreme points $\mathbf{x}_{1_{i\ast}}$ and $\mathbf{x}_{2_{i\ast}}$
are enclosed in black circles.%

\begin{figure}[h]%
\centering
\includegraphics[
height=2.5849in,
width=5.047in
]%
{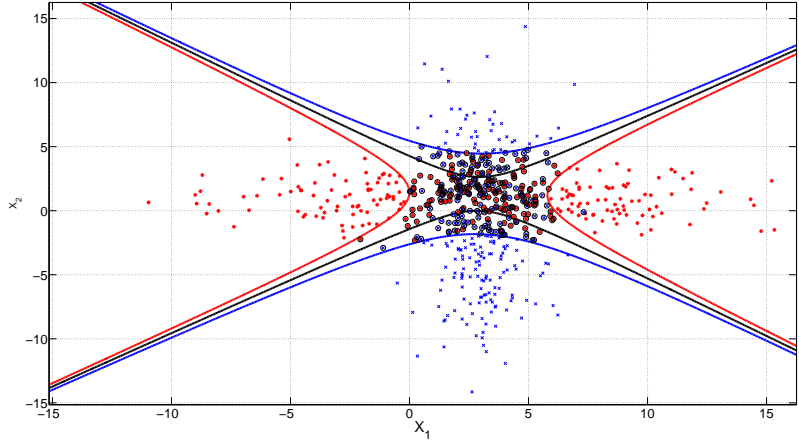}%
\caption{The geometric locus of a novel principal eigenaxis
$\boldsymbol{\kappa}=\boldsymbol{\kappa}_{1}-\boldsymbol{\kappa}_{2}$ jointly
represents an exclusive principal eigen-coordinate system and an eigenaxis of
symmetry that spans the decision space $Z=Z_{1}\cup Z_{2}$ of a minimum risk
binary classification system $k_{\mathbf{s}}\boldsymbol{\kappa}%
+\boldsymbol{\kappa}_{0}\protect\overset{\omega_{1}}{\protect\underset{\omega
_{2}}{\gtrless}}0$, so that all of the points $\mathbf{s}$ that lie on the
geometric loci of a hyperbolic decision boundary and a pair of symmetrically
positioned hyperbolic decision borders exclusively reference the novel
principal eigenaxis $\boldsymbol{\kappa}=\boldsymbol{\kappa}_{1}%
-\boldsymbol{\kappa}_{2}$.}%
\end{figure}

Alternatively, Figure $17$ illustrates how the geometric locus of the novel
principal eigenaxis $\boldsymbol{\kappa}=\boldsymbol{\kappa}_{1}%
-\boldsymbol{\kappa}_{2}$ jointly represents a discriminant function, an
exclusive principal eigen-coordinate system and an eigenaxis of symmetry that
spans the decision space $Z=Z_{1}\cup Z_{2}$ of the minimum risk binary
classification system $k_{\mathbf{s}}\boldsymbol{\kappa}+$ $\boldsymbol{\kappa
}_{0}\overset{\omega_{1}}{\underset{\omega_{2}}{\gtrless}}0$, so that the
novel principal eigenaxis $\boldsymbol{\kappa}=\boldsymbol{\kappa}%
_{1}-\boldsymbol{\kappa}_{2}$ determines likely locations and likelihood
values for all of the extreme points $\mathbf{x}_{1_{i\ast}}$ and
$\mathbf{x}_{2_{i\ast}}$ within the decision space $Z=Z_{1}\cup Z_{2}$ of the
minimum risk binary classification system $k_{\mathbf{s}}\boldsymbol{\kappa}+$
$\boldsymbol{\kappa}_{0}\overset{\omega_{1}}{\underset{\omega_{2}}{\gtrless}%
}0$, relative to the geometric loci of a hyperbolic decision boundary and a
pair of symmetrically positioned hyperbolic decision borders.
\begin{figure}[h]%
\centering
\includegraphics[
height=1.9553in,
width=5.047in
]%
{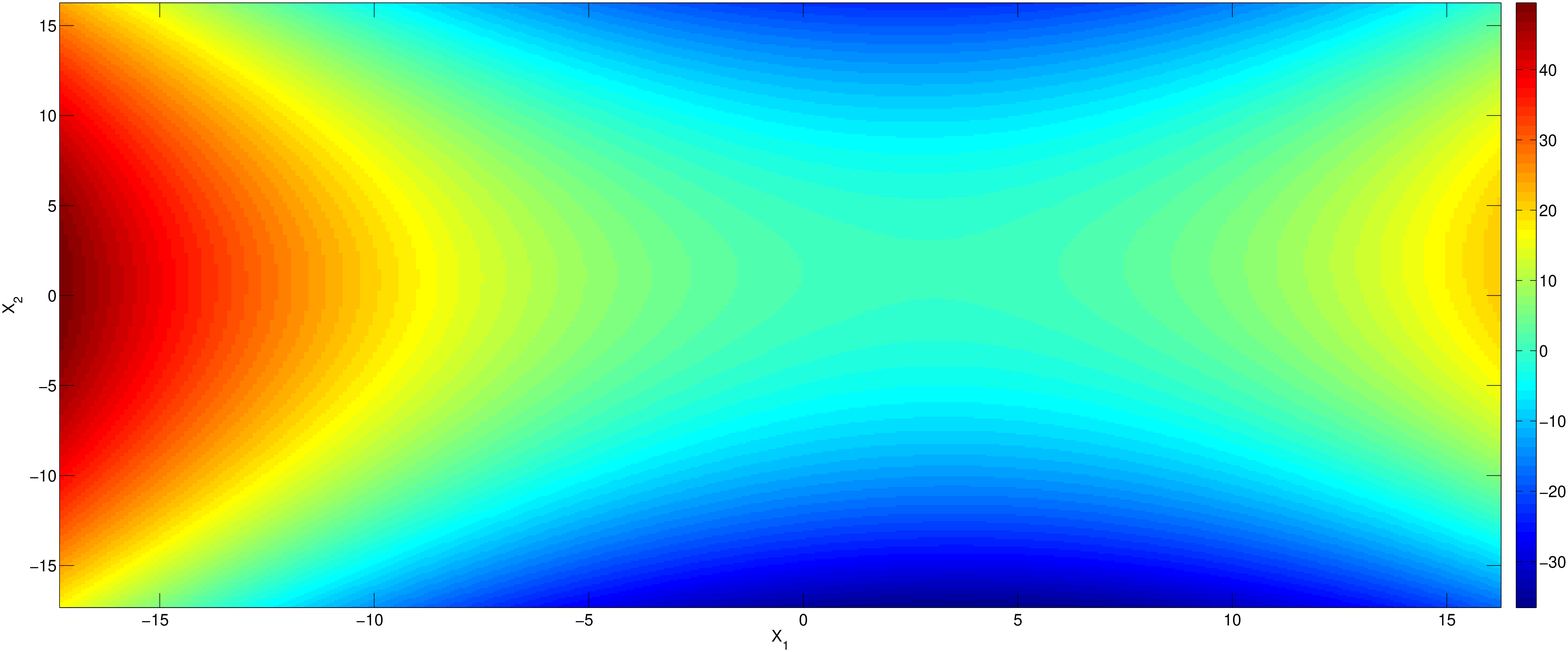}%
\caption{The geometric locus of a novel principal eigenaxis
$\boldsymbol{\kappa}=\boldsymbol{\kappa}_{1}-\boldsymbol{\kappa}_{2}$
determines likely locations and likelihood values for all of the extreme
points $\mathbf{x}_{1_{i\ast}}$ and $\mathbf{x}_{2_{i\ast}}$ within the
decision space $Z=Z_{1}\cup Z_{2}$ of a minimum risk binary classification
system $k_{\mathbf{s}}\boldsymbol{\kappa}+\boldsymbol{\kappa}_{0}%
\protect\overset{\omega_{1}}{\protect\underset{\omega_{2}}{\gtrless}}0$,
relative to the geometric loci of a hyperbolic decision boundary and a pair of
symmetrically positioned hyperbolic decision borders.}%
\end{figure}

\subsubsection{Simulation Example Two}

We now consider a minimum risk binary classification system $k_{\mathbf{s}%
}\boldsymbol{\kappa}+$ $\boldsymbol{\kappa}_{0}\overset{\omega_{1}%
}{\underset{\omega_{2}}{\gtrless}}0$ for two classes of random vectors that
have dissimilar covariance matrices, such that the covariance matrices for
class $\omega_{1}$ and class $\omega_{2}$ are given by%
\[
\Sigma_{1}=\left[
\begin{array}
[c]{cc}%
0.75 & 0.25\\
0.25 & 0.45
\end{array}
\right]  \text{, \ \ }\Sigma_{2}=\left[
\begin{array}
[c]{cc}%
2 & 0.5\\
0.5 & 1.2
\end{array}
\right]  \text{,}%
\]
the mean vector for class $\omega_{1}$ is given by $M_{1}=%
\begin{pmatrix}
2, & 7
\end{pmatrix}
^{T}$ and the mean vector for class $\omega_{2}$ is given by $M_{2}=%
\begin{pmatrix}
4, & 7
\end{pmatrix}
^{T}$. The error rate of the minimum risk binary classification system
$k_{\mathbf{s}}\boldsymbol{\kappa}+$ $\boldsymbol{\kappa}_{0}\overset{\omega
_{1}}{\underset{\omega_{2}}{\gtrless}}0$ is $14.92\%$.

\subsubsection{Illustrations of Dual Capacities}

Figure $18$ illustrates how the geometric locus of a novel principal eigenaxis
$\boldsymbol{\kappa}=\boldsymbol{\kappa}_{1}-\boldsymbol{\kappa}_{2}$ jointly
represents an exclusive principal eigen-coordinate system and an eigenaxis of
symmetry that spans the decision space $Z=Z_{1}\cup Z_{2}$ of the minimum risk
binary classification system $k_{\mathbf{s}}\boldsymbol{\kappa}+$
$\boldsymbol{\kappa}_{0}\overset{\omega_{1}}{\underset{\omega_{2}}{\gtrless}%
}0$, wherein the novel principal eigenaxis $\boldsymbol{\kappa}%
=\boldsymbol{\kappa}_{1}-\boldsymbol{\kappa}_{2}$ is the solution of vector
algebra locus equations that represent the geometric loci of an elliptical
decision boundary $d\left(  \mathbf{s}\right)  =0$ and a pair of symmetrically
positioned elliptical decision borders $d\left(  \mathbf{s}\right)  =+1$ and
$d\left(  \mathbf{s}\right)  =-1$---that jointly partition the decision space
$Z=Z_{1}\cup Z_{2}$ of the minimum risk binary classification system
$k_{\mathbf{s}}\boldsymbol{\kappa}+$ $\boldsymbol{\kappa}_{0}\overset{\omega
_{1}}{\underset{\omega_{2}}{\gtrless}}0$ into symmetrical decision regions
$Z_{1}$ and $Z_{2}$.

The elliptical decision boundary $d\left(  \mathbf{s}\right)  =0$ is black,
the elliptical decision border $d\left(  \mathbf{s}\right)  =+1$ is red, the
elliptical decision border $d\left(  \mathbf{s}\right)  =-1$ is blue, and all
of the extreme points $\mathbf{x}_{1_{i\ast}}$ and $\mathbf{x}_{2_{i\ast}}$
are enclosed in black circles.%
\begin{figure}[ptb]%
\centering
\includegraphics[
height=2.0522in,
width=5.047in
]%
{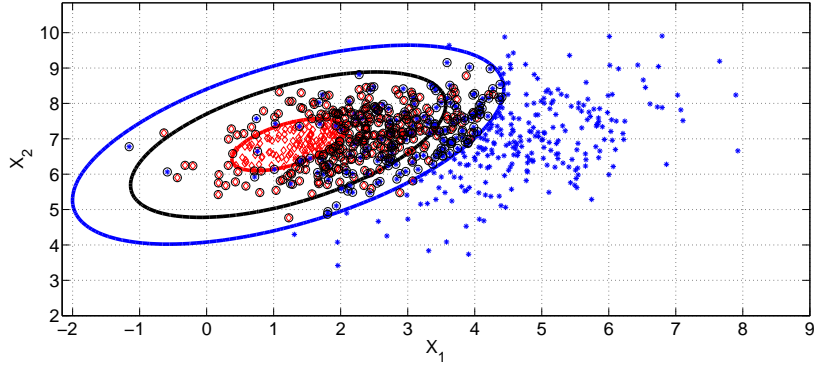}%
\caption{The geometric locus of a novel principal eigenaxis
$\boldsymbol{\kappa}=\boldsymbol{\kappa}_{1}-\boldsymbol{\kappa}_{2}$ jointly
represents an exclusive principal eigen-coordinate system and an eigenaxis of
symmetry that spans the decision space $Z=Z_{1}\cup Z_{2}$ of a minimum risk
binary classification system $k_{\mathbf{s}}\boldsymbol{\kappa}%
+\boldsymbol{\kappa}_{0}\protect\overset{\omega_{1}}{\protect\underset{\omega
_{2}}{\gtrless}}0$, so that all of the points $\mathbf{s}$ that lie on the
geometric loci of an elliptical decision boundary and a pair of symmetrically
positioned elliptical decision borders exclusively reference the novel
principal eigenaxis $\boldsymbol{\kappa}=\boldsymbol{\kappa}_{1}%
-\boldsymbol{\kappa}_{2}$.}%
\end{figure}

Alternatively, Figure $19$ illustrates how the geometric locus of the novel
principal eigenaxis $\boldsymbol{\kappa}=\boldsymbol{\kappa}_{1}%
-\boldsymbol{\kappa}_{2}$ jointly represents a discriminant function, an
exclusive principal eigen-coordinate system and an eigenaxis of symmetry that
spans the decision space $Z=Z_{1}\cup Z_{2}$ of the minimum risk binary
classification system $k_{\mathbf{s}}\boldsymbol{\kappa}+$ $\boldsymbol{\kappa
}_{0}\overset{\omega_{1}}{\underset{\omega_{2}}{\gtrless}}0$, so that the
novel principal eigenaxis $\boldsymbol{\kappa}=\boldsymbol{\kappa}%
_{1}-\boldsymbol{\kappa}_{2}$ determines likely locations and likelihood
values for all of the extreme points $\mathbf{x}_{1_{i\ast}}$ and
$\mathbf{x}_{2_{i\ast}}$ within the decision space $Z=Z_{1}\cup Z_{2}$ of the
minimum risk binary classification system $k_{\mathbf{s}}\boldsymbol{\kappa}+$
$\boldsymbol{\kappa}_{0}\overset{\omega_{1}}{\underset{\omega_{2}}{\gtrless}%
}0$, relative to the geometric loci of an elliptical decision boundary and a
pair of symmetrically positioned elliptical decision borders.
\begin{figure}[ptb]%
\centering
\includegraphics[
height=1.9476in,
width=5.047in
]%
{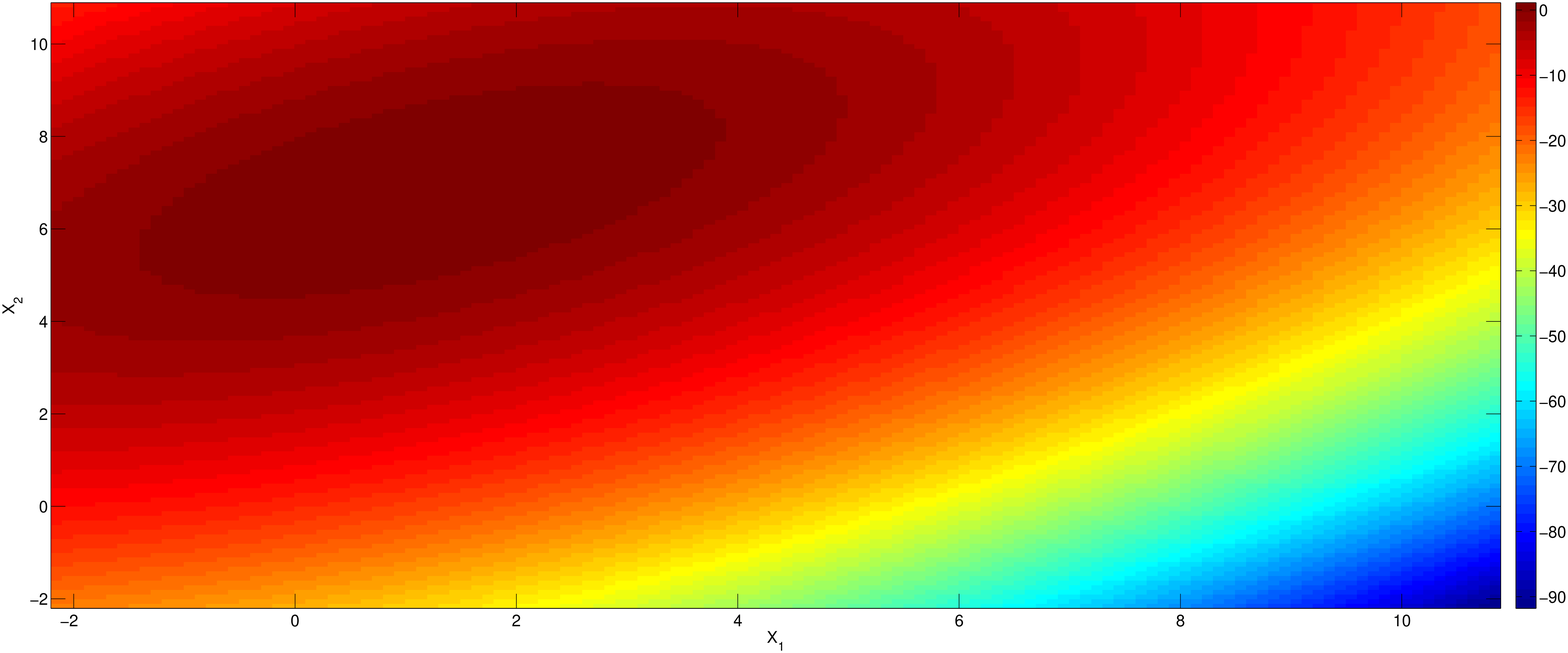}%
\caption{The geometric locus of a novel principal eigenaxis
$\boldsymbol{\kappa}=\boldsymbol{\kappa}_{1}-\boldsymbol{\kappa}_{2}$
determines likely locations and likelihood values for all of the extreme
points $\mathbf{x}_{1_{i\ast}}\mathbf{\ }$and $\mathbf{x}_{2_{i\ast}%
}\mathbf{\ }$ within the decision space $Z=Z_{1}\cup Z_{2}$ of a minimum risk
binary classification system $k_{\mathbf{s}}\boldsymbol{\kappa}%
+\boldsymbol{\kappa}_{0}\protect\overset{\omega_{1}}{\protect\underset{\omega
_{2}}{\gtrless}}0$, relative to the geometric loci of an elliptical decision
boundary and a pair of symmetrically positioned elliptical decision borders.}%
\end{figure}

\subsubsection{Simulation Example Three}

Finally, we consider a minimum risk binary classification system
$k_{\mathbf{s}}\boldsymbol{\kappa}+$ $\boldsymbol{\kappa}_{0}\overset{\omega
_{1}}{\underset{\omega_{2}}{\gtrless}}0$ for two classes of random vectors
that have dissimilar covariance matrices, such that the covariance matrices
for class $\omega_{1}$ and class $\omega_{2}$ are given by%
\[
\Sigma_{1}=\left[
\begin{array}
[c]{cc}%
0.5 & 0\\
0 & 2
\end{array}
\right]  \text{, \ \ }\Sigma_{2}=\left[
\begin{array}
[c]{cc}%
2 & 0\\
0 & 2
\end{array}
\right]  \text{,}%
\]
the mean vector for class $\omega_{1}$ is given by $M_{1}=%
\begin{pmatrix}
3, & 1
\end{pmatrix}
^{T}$ and the mean vector for class $\omega_{2}$ is given by $M_{2}=%
\begin{pmatrix}
3, & -1
\end{pmatrix}
^{T}$. The error rate of the minimum risk binary classification system
$k_{\mathbf{s}}\boldsymbol{\kappa}+$ $\boldsymbol{\kappa}_{0}\overset{\omega
_{1}}{\underset{\omega_{2}}{\gtrless}}0$ is $20\%$.

\subsubsection{Illustrations of Dual Capacities}

Figure $20$ illustrates how the geometric locus of a novel principal eigenaxis
$\boldsymbol{\kappa}=\boldsymbol{\kappa}_{1}-\boldsymbol{\kappa}_{2}$ jointly
represents an exclusive principal eigen-coordinate system and an eigenaxis of
symmetry that spans the decision space $Z=Z_{1}\cup Z_{2}$ of the minimum risk
binary classification system $k_{\mathbf{s}}\boldsymbol{\kappa}+$
$\boldsymbol{\kappa}_{0}\overset{\omega_{1}}{\underset{\omega_{2}}{\gtrless}%
}0$, wherein the novel principal eigenaxis $\boldsymbol{\kappa}%
=\boldsymbol{\kappa}_{1}-\boldsymbol{\kappa}_{2}$ is the solution of vector
algebra locus equations that represent the geometric loci of a parabolic
decision boundary $d\left(  \mathbf{s}\right)  =0$ and a pair of symmetrically
positioned parabolic decision borders $d\left(  \mathbf{s}\right)  =+1$ and
$d\left(  \mathbf{s}\right)  =-1$---that jointly partition the decision space
$Z=Z_{1}\cup Z_{2}$ of the minimum risk binary classification system
$k_{\mathbf{s}}\boldsymbol{\kappa}+$ $\boldsymbol{\kappa}_{0}\overset{\omega
_{1}}{\underset{\omega_{2}}{\gtrless}}0$ into symmetrical decision regions
$Z_{1}$ and $Z_{2}$.

The parabolic decision boundary $d\left(  \mathbf{s}\right)  =0$ is black, the
parabolic decision border $d\left(  \mathbf{s}\right)  =+1$ is red, the
parabolic decision border $d\left(  \mathbf{s}\right)  =-1$ is blue, and all
of the extreme points $\mathbf{x}_{1_{i\ast}}$ and $\mathbf{x}_{2_{i\ast}}$
are enclosed in black circles.%
\begin{figure}[ptb]%
\centering
\includegraphics[
height=1.9735in,
width=5.047in
]%
{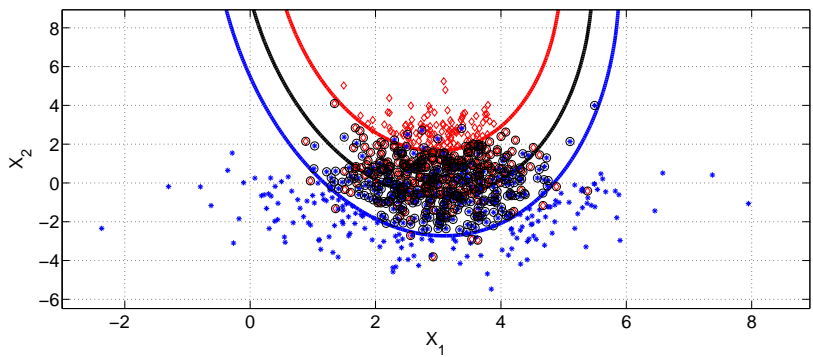}%
\caption{The geometric locus of a novel principal eigenaxis
$\boldsymbol{\kappa}=\boldsymbol{\kappa}_{1}-\boldsymbol{\kappa}_{2}$ jointly
represents an exclusive principal eigen-coordinate system and an eigenaxis of
symmetry that spans the decision space $Z=Z_{1}\cup Z_{2}$ of a minimum risk
binary classification system $k_{\mathbf{s}}\boldsymbol{\kappa}%
+\boldsymbol{\kappa}_{0}\protect\overset{\omega_{1}}{\protect\underset{\omega
_{2}}{\gtrless}}0$, so that all of the points $\mathbf{s}$ that lie on the
geometric loci of a parabolic decision boundary and a pair of symmetrically
positioned parabolic decision borders exclusively reference the novel
principal eigenaxis $\boldsymbol{\kappa}=\boldsymbol{\kappa}_{1}%
-\boldsymbol{\kappa}_{2}$.}%
\end{figure}

Alternatively, Figure $21$ illustrates how the geometric locus of the novel
principal eigenaxis $\boldsymbol{\kappa}=\boldsymbol{\kappa}_{1}%
-\boldsymbol{\kappa}_{2}$ jointly represents a discriminant function, an
exclusive principal eigen-coordinate system and an eigenaxis of symmetry that
spans the decision space $Z=Z_{1}\cup Z_{2}$ of the minimum risk binary
classification system $k_{\mathbf{s}}\boldsymbol{\kappa}+$ $\boldsymbol{\kappa
}_{0}\overset{\omega_{1}}{\underset{\omega_{2}}{\gtrless}}0$, so that the
novel principal eigenaxis $\boldsymbol{\kappa}=\boldsymbol{\kappa}%
_{1}-\boldsymbol{\kappa}_{2}$ determines likely locations and likelihood
values for all of the extreme points $\mathbf{x}_{1_{i\ast}}$ and
$\mathbf{x}_{2_{i\ast}}$ within the decision space $Z=Z_{1}\cup Z_{2}$ of the
minimum risk binary classification system $k_{\mathbf{s}}\boldsymbol{\kappa}+$
$\boldsymbol{\kappa}_{0}\overset{\omega_{1}}{\underset{\omega_{2}}{\gtrless}%
}0$, relative to the geometric loci of a parabolic decision boundary and a
pair of symmetrically positioned parabolic decision borders.
\begin{figure}[ptb]%
\centering
\includegraphics[
trim=0.000000in 0.000000in 0.002964in 0.000000in,
height=1.9484in,
width=5.047in
]%
{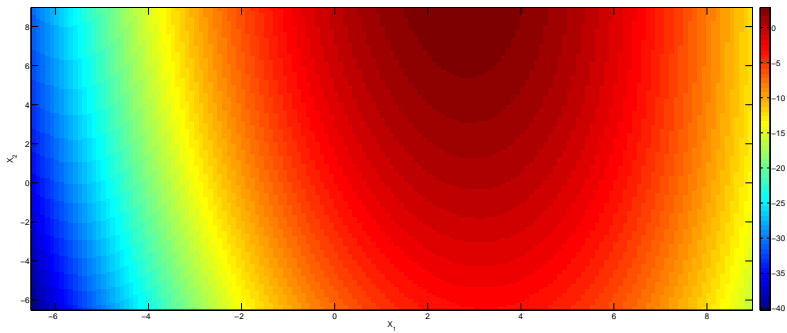}%
\caption{The geometric locus of a novel principal eigenaxis
$\boldsymbol{\kappa}=\boldsymbol{\kappa}_{1}-\boldsymbol{\kappa}_{2}$
determines likely locations and likelihood values for all of the extreme
points $\mathbf{x}_{1_{i\ast}}\mathbf{\sim}$ $p\left(  \mathbf{x};\omega
_{1}\right)  $ and $\mathbf{x}_{2_{i\ast}}\mathbf{\sim}$ $p\left(
\mathbf{x};\omega_{2}\right)  $ within the decision space $Z=Z_{1}\cup Z_{2}$
of a minimum risk binary classification system $k_{\mathbf{s}}%
\boldsymbol{\kappa}+\boldsymbol{\kappa}_{0}\protect\overset{\omega
_{1}}{\protect\underset{\omega_{2}}{\gtrless}}0$, relative to the geometric
loci of a parabolic decision boundary and a pair of symmetrically positioned
parabolic decision borders.}%
\end{figure}

\subsection{Dual Capacities and Generalization Behavior}

The simulation examples that were presented in Section \ref{Section 17}
illustrate that the geometric locus of the novel principal eigenaxis of any
given minimum risk binary classification system---which is structured as a
dual locus of likelihood components and principal eigenaxis
components---\emph{jointly represents} $\left(  1\right)  $ the discriminant
function of the system; $\left(  2\right)  $ an exclusive principal
eigen-coordinate system of the decision boundary of the system; and $\left(
3\right)  $ an eigenaxis of symmetry that spans the decision space of the
system. The simulation examples also illustrate that the \emph{statistical
structure} of a discriminant function is essential for its
\emph{functionality}---which includes the ability of the discriminant function
to \emph{generalize}.

We are now in a position to demonstrate how the discriminate function of any
given minimum risk binary classification system \emph{extrapolates and thereby
generalizes in a very nontrivial manner}.

\section{\label{Section 19}How Discriminant Functions Extrapolate}

Discriminant functions of minimum risk binary classification systems
extrapolate, and thereby generalize in a very nontrivial manner---because the
\emph{important generalizations} for a minimum risk binary classification
system are \emph{statistically} \emph{pre-wired} within the geometric
\emph{locus} of the \emph{novel principal eigenaxis }of the\emph{ system}---by
means a system of fundamental locus equations of binary classification,
subject to distinctive geometrical and statistical conditions for a minimum
risk binary classification system in statistical equilibrium---that is
satisfied by certain random vectors.

\subsection{Extrapolation by Machine Learning Algorithms}%

\citet{Geman1992}
considered difficult machine learning tasks to be problems of extrapolation,
rather than interpolation since training data will never \textquotedblleft
cover\textquotedblright\ the space of all possible inputs. Moreover, any given
learning machine that \textquotedblleft extrapolates,\textquotedblright\ also
generalizes in a very nontrivial sense
\citep{Geman1992}%
.%

\citet{Geman1992}
also noted that: \textquotedblleft Unfortunately, the most interesting
problems tend to be problems of extrapolation, that is, nontrivial
generalization. It would appear, then, that the only way to avoid having to
densely cover the input space with training examples---which is unfeasible in
practice---is to \emph{pre-wire} the important
generalizations.\textquotedblright

We agree with
\citet{Geman1992}%
. Even so, we realize that a fundamental problem still remains: How do we to
\emph{define} the \emph{extrapolation problem} for a given learning machine?
Equally important, how do we pre-wire the important generalizations for a
given extrapolation problem? Likewise, what does it mean to \emph{pre-wire}
the important generalizations \emph{within a given learning machine}?

\subsection{Pre-wiring of Important Generalizations}

We have demonstrated by analyses and simulation studies that the important
generalizations for a minimum risk binary classification system are
statistically \textquotedblleft\emph{pre-wired\textquotedblright}
\emph{within} the geometric \emph{locus} of the novel \emph{principal
eigenaxis} of the system by means a system of fundamental locus equations of
binary classification---subject to distinctive geometrical and statistical
conditions for a minimum risk binary classification system in statistical
equilibrium---that is satisfied by certain random vectors called extreme vectors.

Indeed, we have demonstrated that likelihood values and likely locations of
any given collection of extreme points---are \emph{statistically
}\textquotedblleft\emph{pre-wired\textquotedblright} within a dual locus of
likelihood components and principal eigenaxis components of a minimum risk
binary classification system---so that the given \emph{collection} of extreme
points \emph{covers} the decision \emph{space} of the system.

Moreover, we have demonstrated that the geometric locus of the novel principal
eigenaxis of any given minimum risk binary classification system---is
structured as a dual locus of likelihood components and principal eigenaxis
components---so that the novel principal eigenaxis \emph{jointly represents}
the discriminant function of the system, an exclusive principal
eigen-coordinate system of the decision boundary of the system, and an
eigenaxis of symmetry that spans the decision space of the system.

We now demonstrate \emph{how} the discriminant function $d\left(
\mathbf{s}\right)  =k_{\mathbf{s}}\boldsymbol{\kappa}+$ $\boldsymbol{\kappa
}_{0}$ of any given minimum risk binary classification system $k_{\mathbf{s}%
}\boldsymbol{\kappa}+$ $\boldsymbol{\kappa}_{0}\overset{\omega_{1}%
}{\underset{\omega_{2}}{\gtrless}}0$ extrapolates and thereby generalizes in a
significant manner.

Thereby, it will be seen that the \emph{statistical structure} of a
discriminant function is essential for its \emph{functionality}---which
includes the ability of the discriminant function to \emph{generalize}.

Substitute the expressions for $\boldsymbol{\kappa}$ and $\boldsymbol{\kappa
}_{0}$ in (\ref{Primal Dual Locus}) and (\ref{Funct Primal Dual Locus}) into
the expression for the discriminant function in (\ref{Discriminant F2}), so
that the discriminant function is rewritten as%
\begin{align}
d\left(  \mathbf{s}\right)   &  =\left(  k_{\mathbf{s}}\boldsymbol{-}\frac
{1}{l}\sum\nolimits_{i=1}^{l}k_{\mathbf{x}_{i\ast}}\right)  \boldsymbol{\kappa
}_{1}-\left(  k_{\mathbf{s}}\boldsymbol{-}\frac{1}{l}\sum\nolimits_{i=1}%
^{l}k_{\mathbf{x}_{i\ast}}\right)  \boldsymbol{\kappa}_{2} \tag{19.1}%
\label{Discriminant F3}\\
&  +\frac{1}{l}\sum\nolimits_{i=1}^{l}y_{i}\left(  1-\xi_{i}\right)
\text{,}\nonumber
\end{align}
such that the discriminant function is represented by a geometric locus of a
novel principal eigenaxis%
\begin{align*}
\boldsymbol{\kappa}  &  =\sum\nolimits_{i=1}^{l_{1}}\psi_{1_{i\ast}%
}k_{\mathbf{x}_{1_{i\ast}}}-\sum\nolimits_{i=1}^{l_{2}}\psi_{2_{i\ast}%
}k_{\mathbf{x}_{2_{i\ast}}}\\
&  =\boldsymbol{\kappa}_{1}{\large -}\boldsymbol{\kappa}_{2}\text{,}%
\end{align*}
structured as a locus of signed and scaled extreme vectors $\psi_{1_{i_{\ast}%
}}k_{\mathbf{x}_{1_{i\ast}}}$ and $-\psi_{2_{i_{\ast}}}k_{\mathbf{x}%
_{2_{i\ast}}}$, at which point the dual locus of likelihood components
$\psi_{1_{i_{\ast}}}k_{\mathbf{x}_{1_{i\ast}}}$ and $\psi_{2_{i_{\ast}}%
}k_{\mathbf{x}_{2_{i\ast}}}$ and principal eigenaxis components $\psi
_{1_{i_{\ast}}}k_{\mathbf{x}_{1_{i\ast}}}$ and $\psi_{2_{i_{\ast}}%
}k_{\mathbf{x}_{2_{i\ast}}}$ represents an exclusive principal
eigen-coordinate system $\boldsymbol{\kappa}=\boldsymbol{\kappa}_{1}%
{\large -}\boldsymbol{\kappa}_{2}$ of the geometric locus of the decision
boundary $k_{\mathbf{s}}\boldsymbol{\kappa}+$ $\boldsymbol{\kappa}_{0}=0$ of
the minimum risk binary classification system $k_{\mathbf{s}}%
\boldsymbol{\kappa}+$ $\boldsymbol{\kappa}_{0}\overset{\omega_{1}%
}{\underset{\omega_{2}}{\gtrless}}0$, and also represents an eigenaxis of
symmetry $\boldsymbol{\kappa}=\boldsymbol{\kappa}_{1}{\large -}%
\boldsymbol{\kappa}_{2}$ that spans the decision space $Z=Z_{1}\cup Z_{2}$ of
the system, such that each principal eigenaxis component $\psi_{1_{i_{\ast}}%
}k_{\mathbf{x}_{1_{i\ast}}}$ or $\psi_{2_{i_{\ast}}}k_{\mathbf{x}_{2_{i\ast}}%
}$ on $\boldsymbol{\kappa}=\boldsymbol{\kappa}_{1}{\large -}\boldsymbol{\kappa
}_{2}$ determines a likely location for a correlated extreme point
$\mathbf{x}_{1_{i\ast}}$ or $\mathbf{x}_{2_{i\ast}}$, and each likelihood
component $\psi_{1_{i_{\ast}}}k_{\mathbf{x}_{1_{i\ast}}}$ or $\psi
_{2_{i_{\ast}}}k_{\mathbf{x}_{2_{i\ast}}}$ on $\boldsymbol{\kappa
}=\boldsymbol{\kappa}_{1}{\large -}\boldsymbol{\kappa}_{2}$ determines a
likelihood value for the correlated extreme point $\mathbf{x}_{1_{i\ast}}$ or
$\mathbf{x}_{2_{i\ast}}$, wherein the expression $\frac{1}{l}\sum
\nolimits_{i=1}^{l}y_{i}\left(  1-\xi_{i}\right)  $ represents an expected
likelihood of observing $l$ extreme vectors $\left\{  k_{\mathbf{x}_{i\ast}%
}\right\}  _{i=1}^{l}$ that belong to class $\omega_{1}$ and class $\omega
_{2}$, and the vector $k_{\mathbf{s}}\boldsymbol{-}\frac{1}{l}\sum
\nolimits_{i=1}^{l}k_{\mathbf{x}_{i\ast}}$ gauges the position of the locus of
a random vector $\mathbf{s}$ being classified relative to its position from
the locus of average risk $\frac{1}{l}\sum\nolimits_{i=1}^{l}k_{\mathbf{x}%
_{i\ast}}$ within the decision space $Z=Z_{1}\cup Z_{2}$ of the system
$k_{\mathbf{s}}\boldsymbol{\kappa}+$ $\boldsymbol{\kappa}_{0}\overset{\omega
_{1}}{\underset{\omega_{2}}{\gtrless}}0$.

Let the discriminant function in (\ref{Discriminant F3}) be determined by
using the novel principal eigen-coordinate algorithm that is being
examined---to transform a collection $\left\{  \mathbf{x}_{i}\right\}
_{i=1}^{N}$ of $N$ labeled feature vectors $\mathbf{x}_{i}$ into a geometric
locus of a novel principal eigenaxis $\boldsymbol{\kappa}=\boldsymbol{\kappa
}_{1}{\large -}\boldsymbol{\kappa}_{2}$. Also, let $\mathbf{s}$ denote a
feature vector that either belongs to or is related to the collection of $N$
feature vectors.

Now take the discriminant function in (\ref{Discriminant F3}) along with any
given feature vector $\mathbf{s}$ that either belongs to or is related to the
collection of $N$ feature vectors.

The discriminant function in (\ref{Discriminant F3}) determines the likely
location of the feature vector $\mathbf{s}$ within the decision space
$Z=Z_{1}\cup Z_{2}$ of the minimum risk binary classification system%
\[
\left(  k_{\mathbf{s}}\boldsymbol{-}\frac{1}{l}\sum\nolimits_{i=1}%
^{l}k_{\mathbf{x}_{i\ast}}\right)  \left(  \boldsymbol{\kappa}_{1}%
-\boldsymbol{\kappa}_{2}\right)  +\frac{1}{l}\sum\nolimits_{i=1}^{l}%
y_{i}\left(  1-\xi_{i}\right)  \overset{\omega_{1}}{\underset{\omega
_{2}}{\gtrless}}0
\]
by projecting the vector difference of $k_{\mathbf{s}}\boldsymbol{-}\frac
{1}{l}\sum\nolimits_{i=1}^{l}k_{\mathbf{x}_{i\ast}}$ onto the geometric locus
of the novel principal eigenaxis $\boldsymbol{\kappa}=\boldsymbol{\kappa}%
_{1}-\boldsymbol{\kappa}_{2}$, and thereby recognizes the category $\omega
_{1}$ or $\omega_{2}$ of the feature vector $\mathbf{s}$ from the sign of the
expression%
\begin{equation}
d\left(  \mathbf{s}\right)  \triangleq\left\Vert \boldsymbol{\kappa}%
_{1}{\large -}\boldsymbol{\kappa}_{2}\right\Vert \left[  \left\Vert
k_{\mathbf{s}}\boldsymbol{-}\frac{1}{l}\sum\nolimits_{i=1}^{l}k_{\mathbf{x}%
_{i\ast}}\right\Vert \cos\theta\right]  +\frac{1}{l}\sum\nolimits_{i=1}%
^{l}y_{i}\left(  1-\xi_{i}\right)  \text{,} \tag{18.2}%
\label{Likelihood & Location Functional}%
\end{equation}
wherein $\operatorname{sign}\left(  d\left(  \mathbf{x}\right)  \right)  $
indicates the decision region $Z_{1}$ or $Z_{2}$ that the feature vector
$\mathbf{s}$ is located within, so that the signed magnitude of the vector
projection of the feature vector $\mathbf{s}$ onto the exclusive principal
eigen-coordinate system $\boldsymbol{\kappa}=\boldsymbol{\kappa}%
_{1}-\boldsymbol{\kappa}_{2}$ and the corresponding eigenaxis of symmetry
$\boldsymbol{\kappa}=\boldsymbol{\kappa}_{1}-\boldsymbol{\kappa}_{2}$---that
spans the decision space $Z=Z_{1}\cup Z_{2}$ of the minimum risk binary
classification system $k_{\mathbf{s}}\boldsymbol{\kappa}+$ $\boldsymbol{\kappa
}_{0}\overset{\omega_{1}}{\underset{\omega_{2}}{\gtrless}}0$---determines the
likely location of the feature vector $\mathbf{s}$ within the decision space
$Z=Z_{1}\cup Z_{2}$ of the system, at which point the vector $k_{\mathbf{s}%
}\boldsymbol{-}\frac{1}{l}\sum\nolimits_{i=1}^{l}k_{\mathbf{x}_{i\ast}}$
gauges the position of the locus of the feature vector $\mathbf{s}$ relative
to its position from the locus of average risk $\frac{1}{l}\sum\nolimits_{i=1}%
^{l}k_{\mathbf{x}_{i\ast}}$ within the decision space $Z=Z_{1}\cup Z_{2}$ of
the system, where the locus of average risk $\frac{1}{l}\sum\nolimits_{i=1}%
^{l}k_{\mathbf{x}_{i\ast}}$ is located on or near the locus of a linear
decision boundary or is centrally located and bounded by quadratic loci of a
quadratic decision boundary.

By (\ref{Likelihood & Location Functional}), it follows that the vector
projection of the vector $k_{\mathbf{s}}\boldsymbol{-}\frac{1}{l}%
\sum\nolimits_{i=1}^{l}k_{\mathbf{x}_{i\ast}}$ onto the geometric locus of the
novel principal eigenaxis $\boldsymbol{\kappa}=\boldsymbol{\kappa}%
_{1}-\boldsymbol{\kappa}_{2}$ determines a conditional likelihood value for
the feature vector $\mathbf{s}$ that is conditional on distributions---of all
of the extreme points $\mathbf{x}_{1_{i\ast}}$ and $\mathbf{x}_{2_{i\ast}}%
$---determined by the statistical content of $\boldsymbol{\kappa}_{1}$ and
$\boldsymbol{\kappa}_{2}$, along with a likely location of the feature vector
$\mathbf{s}$ that is determined by signed magnitudes of $k_{\mathbf{s}}$ and
$\boldsymbol{-}\frac{1}{l}\sum\nolimits_{i=1}^{l}k_{\mathbf{x}_{i\ast}}$ along
the geometric locus of the exclusive principal eigen-coordinate system
$\boldsymbol{\kappa}=\boldsymbol{\kappa}_{1}-\boldsymbol{\kappa}_{2}$ and the
corresponding eigenaxis of symmetry $\boldsymbol{\kappa}=\boldsymbol{\kappa
}_{1}-\boldsymbol{\kappa}_{2}$, so that the dual locus of the discriminant
function%
\[
\left(  k_{\mathbf{s}}\boldsymbol{-}\frac{1}{l}\sum\nolimits_{i=1}%
^{l}k_{\mathbf{x}_{i\ast}}\right)  \left(  \boldsymbol{\kappa}_{1}%
-\boldsymbol{\kappa}_{2}\right)
\]
determines an expected location and a likelihood value of the feature vector
$\mathbf{s}$ that is conditional on how the feature vector $\mathbf{s}$ is
distributed over the locus of the exclusive principal eigen-coordinate system
$\boldsymbol{\kappa}=\boldsymbol{\kappa}_{1}-\boldsymbol{\kappa}_{2}$ and the
corresponding eigenaxis of symmetry $\boldsymbol{\kappa}=\boldsymbol{\kappa
}_{1}-\boldsymbol{\kappa}_{2}$ that spans the decision space $Z=Z_{1}\cup
Z_{2}$ of the minimum risk binary classification system $k_{\mathbf{s}%
}\boldsymbol{\kappa}+$ $\boldsymbol{\kappa}_{0}\overset{\omega_{1}%
}{\underset{\omega_{2}}{\gtrless}}0$, at which point the vector $k_{\mathbf{s}%
}\boldsymbol{-}\frac{1}{l}\sum\nolimits_{i=1}^{l}$ gauges the position of the
locus of the feature vector $\mathbf{s}$ relative to its position from the
locus of average risk $\frac{1}{l}\sum\nolimits_{i=1}^{l}k_{\mathbf{x}_{i\ast
}}$ within the decision space $Z=Z_{1}\cup Z_{2}$, whereas the expression
$\frac{1}{l}\sum\nolimits_{i=1}^{l}y_{i}\left(  1-\xi_{i}\right)  $ represents
an expected likelihood of observing $l$ extreme vectors $\left\{
k_{\mathbf{x}_{i\ast}}\right\}  _{i=1}^{l}$ that belong to class $\omega_{1}$
and class $\omega_{2}$.

Alternatively, we now consider how the discriminate function of any given
minimum risk binary classification system \emph{generalizes in a nontrivial
manner} and thereby \emph{extrapolates}.

\subsection{How Discriminant Functions Generalize}

Let
\begin{align*}
d\left(  \mathbf{s}\right)   &  =k_{\mathbf{s}}\boldsymbol{\kappa
}+\boldsymbol{\kappa}_{0}\\
&  =\left(  k_{\mathbf{s}}\boldsymbol{-}\frac{1}{l}\sum\nolimits_{i=1}%
^{l}k_{\mathbf{x}_{i\ast}}\right)  \left(  \boldsymbol{\kappa}_{1}%
-\boldsymbol{\kappa}_{2}\right)  +\frac{1}{l}\sum\nolimits_{i=1}^{l}%
y_{i}\left(  1-\xi_{i}\right) \\
&  =\left(  k_{\mathbf{s}}\boldsymbol{-}\frac{1}{l}\sum\nolimits_{i=1}%
^{l}k_{\mathbf{x}_{i\ast}}\right)  \left[  \sum\nolimits_{i=1}^{l_{1}}%
\psi_{1_{i\ast}}k_{\mathbf{x}_{1_{i\ast}}}-\sum\nolimits_{i=1}^{l_{2}}%
\psi_{2_{i\ast}}k_{\mathbf{x}_{2_{i\ast}}}\right] \\
&  +\frac{1}{l}\sum\nolimits_{i=1}^{l}y_{i}\left(  1-\xi_{i}\right)
\end{align*}
be the discriminant function of any given minimum risk binary classification
system%
\begin{align*}
&  k_{\mathbf{s}}\boldsymbol{\kappa}+\boldsymbol{\kappa}_{0}\overset{\omega
_{1}}{\underset{\omega_{2}}{\gtrless}}0\\
=  &  \left(  k_{\mathbf{s}}\boldsymbol{-}\frac{1}{l}\sum\nolimits_{i=1}%
^{l}k_{\mathbf{x}_{i\ast}}\right)  \left(  \boldsymbol{\kappa}_{1}%
-\boldsymbol{\kappa}_{2}\right)  +\frac{1}{l}\sum\nolimits_{i=1}^{l}%
y_{i}\left(  1-\xi_{i}\right)  \overset{\omega_{1}}{\underset{\omega
_{2}}{\gtrless}}0
\end{align*}
that is subject to random inputs $\mathbf{x\in}$ $%
\mathbb{R}
^{d}$ such that $\mathbf{x\sim}$ $p\left(  \mathbf{x};\omega_{1}\right)  $ and
$\mathbf{x\sim}$ $p\left(  \mathbf{x};\omega_{2}\right)  $, where $p\left(
\mathbf{x};\omega_{1}\right)  $ and $p\left(  \mathbf{x};\omega_{2}\right)  $
are certain probability density functions for two classes $\omega_{1}$ and
$\omega_{2}$ of random vectors $\mathbf{x\in}$ $%
\mathbb{R}
^{d}$.

The discriminant function $d\left(  \mathbf{s}\right)  =k_{\mathbf{s}%
}\boldsymbol{\kappa}+$ $\boldsymbol{\kappa}_{0}$ generalizes and thereby
extrapolates by means of an exclusive principal eigen-coordinate system
$\boldsymbol{\kappa}=\boldsymbol{\kappa}_{1}-\boldsymbol{\kappa}_{2}$ and a
corresponding eigenaxis of symmetry $\boldsymbol{\kappa}=\boldsymbol{\kappa
}_{1}-\boldsymbol{\kappa}_{2}$ that constitutes a dual locus
$\boldsymbol{\kappa}=\sum\nolimits_{i=1}^{l_{1}}\psi_{1_{i\ast}}%
k_{\mathbf{x}_{1_{i\ast}}}-\sum\nolimits_{i=1}^{l_{2}}\psi_{2_{i\ast}%
}k_{\mathbf{x}_{2_{i\ast}}}$ of the entire decision space $Z=Z_{1}\cup Z_{2}$
of the system $k_{\mathbf{s}}\boldsymbol{\kappa}+$ $\boldsymbol{\kappa}%
_{0}\overset{\omega_{1}}{\underset{\omega_{2}}{\gtrless}}0$, so that a
geometric locus of a novel principal eigenaxis $\boldsymbol{\kappa
}=\boldsymbol{\kappa}_{1}-\boldsymbol{\kappa}_{2}\boldsymbol{\ }$contains all
of the covariance and distribution information for all of the extreme points
$\mathbf{x}_{1_{i\ast}}\mathbf{\sim}$ $p\left(  \mathbf{x};\omega_{1}\right)
$ and $\mathbf{x}_{2_{i\ast}}\mathbf{\sim}$ $p\left(  \mathbf{x};\omega
_{2}\right)  $---relative to the covariance and distribution information for a
given collection $\left\{  \mathbf{x}_{i}\right\}  _{i=1}^{N}$ of feature
vectors $\mathbf{x}_{i}$, such that each principal eigenaxis component
$\psi_{1_{i_{\ast}}}k_{\mathbf{x}_{1_{i\ast}}}$ or $\psi_{2_{i_{\ast}}%
}k_{\mathbf{x}_{2_{i\ast}}}$ on $\boldsymbol{\kappa}=\boldsymbol{\kappa}%
_{1}{\large -}\boldsymbol{\kappa}_{2}$ determines a likely location for a
correlated extreme point $\mathbf{x}_{1_{i\ast}}$ or $\mathbf{x}_{2_{i\ast}}$,
and each likelihood component $\psi_{1_{i_{\ast}}}k_{\mathbf{x}_{1_{i\ast}}}$
or $\psi_{2_{i_{\ast}}}k_{\mathbf{x}_{2_{i\ast}}}$ on $\boldsymbol{\kappa
}=\boldsymbol{\kappa}_{1}{\large -}\boldsymbol{\kappa}_{2}$ determines a
likelihood value for the correlated extreme point $\mathbf{x}_{1_{i\ast}}$ or
$\mathbf{x}_{2_{i\ast}}$, at which point the vector $k_{\mathbf{s}%
}\boldsymbol{-}\frac{1}{l}\sum\nolimits_{i=1}^{l}$ gauges the position of the
locus of a feature vector $\mathbf{s}$ relative to its position from the locus
of average risk $\frac{1}{l}\sum\nolimits_{i=1}^{l}k_{\mathbf{x}_{i\ast}}$
within the decision space $Z=Z_{1}\cup Z_{2}$ of the system $k_{\mathbf{s}%
}\boldsymbol{\kappa}+$ $\boldsymbol{\kappa}_{0}\overset{\omega_{1}%
}{\underset{\omega_{2}}{\gtrless}}0$, and the expression $\frac{1}{l}%
\sum\nolimits_{i=1}^{l}y_{i}\left(  1-\xi_{i}\right)  $ represents an expected
likelihood of observing $l$ extreme vectors $\left\{  k_{\mathbf{x}_{i\ast}%
}\right\}  _{i=1}^{l}$.

Thereby, the discriminant function
\[
d\left(  \mathbf{s}\right)  =\left(  k_{\mathbf{s}}\boldsymbol{-}\frac{1}%
{l}\sum\nolimits_{i=1}^{l}k_{\mathbf{x}_{i\ast}}\right)  \left(
\boldsymbol{\kappa}_{1}-\boldsymbol{\kappa}_{2}\right)  +\frac{1}{l}%
\sum\nolimits_{i=1}^{l}y_{i}\left(  1-\xi_{i}\right)
\]
determines conditional likelihood values and likely locations of any given
feature vectors $\mathbf{s}$ that either belong to or are related to the given
collection $\left\{  \mathbf{x}_{i}\right\}  _{i=1}^{N}$ of feature vectors
$\mathbf{x}_{i}$ that have been used to determine the discriminant function,
such that the dual locus of the discriminant function%
\[
\left(  k_{\mathbf{s}}\boldsymbol{-}\frac{1}{l}\sum\nolimits_{i=1}%
^{l}k_{\mathbf{x}_{i\ast}}\right)  \left(  \boldsymbol{\kappa}_{1}%
-\boldsymbol{\kappa}_{2}\right)
\]
determines an expected location and a likelihood value of a feature vector
$\mathbf{s}$ that is conditional on how the feature vector $\mathbf{s}$ is
distributed over the locus of the exclusive principal eigen-coordinate system
$\boldsymbol{\kappa}=\boldsymbol{\kappa}_{1}-\boldsymbol{\kappa}_{2}$ and the
corresponding eigenaxis of symmetry $\boldsymbol{\kappa}=\boldsymbol{\kappa
}_{1}-\boldsymbol{\kappa}_{2}$ that spans the decision space $Z=Z_{1}\cup
Z_{2}$ of the minimum risk binary classification system $k_{\mathbf{s}%
}\boldsymbol{\kappa}+$ $\boldsymbol{\kappa}_{0}\overset{\omega_{1}%
}{\underset{\omega_{2}}{\gtrless}}0$, at which point the vector $k_{\mathbf{s}%
}\boldsymbol{-}\frac{1}{l}\sum\nolimits_{i=1}^{l}$ gauges the position of the
locus of the feature vector $\mathbf{s}$ relative to its position from the
locus of average risk $\frac{1}{l}\sum\nolimits_{i=1}^{l}k_{\mathbf{x}_{i\ast
}}$, and the expression $\frac{1}{l}\sum\nolimits_{i=1}^{l}y_{i}\left(
1-\xi_{i}\right)  $ represents an expected likelihood of observing $l$ extreme
vectors $\left\{  k_{\mathbf{x}_{i\ast}}\right\}  _{i=1}^{l}$ within the
decision space $Z=Z_{1}\cup Z_{2}$.

Accordingly, we conclude that the \emph{statistical structure} of the
discriminant function of any given minimum risk binary classification system
is essential for its \emph{functionality}---which includes the ability of the
discriminant function to \emph{generalize in a significant manner}.

We now turn our attention to the action taken by a minimum risk binary
classification system to jointly minimize its eigenenergy and risk. We show
that any given minimum risk binary classification system acts to jointly
minimize its eigenenergy and risk by locating a point of equilibrium, at which
point critical minimum eigenenergies exhibited by the system are symmetrically
concentrated in such a manner that the dual locus of the discriminant function
of the system is \emph{in} statistical equilibrium---\emph{at} the geometric
locus of the decision boundary of the system, whereon the statistical fulcrum
of the system is located.

In the next section of our treatise, we demonstrate how any given minimum risk
binary classification system achieves this feat.

\section{\label{Section 20}Joint Minimization of Eigenenergy and Risk}

We now demonstrate how the dual locus $\boldsymbol{\kappa}=\boldsymbol{\kappa
}_{1}-\boldsymbol{\kappa}_{2}$ of the discriminant function $d\left(
\mathbf{s}\right)  =$ $k_{\mathbf{s}}\boldsymbol{\kappa}+$ $\boldsymbol{\kappa
}_{0}$ of any given minimum risk binary classification system $k_{\mathbf{s}%
}\boldsymbol{\kappa}+$ $\boldsymbol{\kappa}_{0}\overset{\omega_{1}%
}{\underset{\omega_{2}}{\gtrless}}0$ is in statistical equilibrium---at the
geometric locus of the decision boundary $k_{\mathbf{s}}\boldsymbol{\kappa}+$
$\boldsymbol{\kappa}_{0}=0$ of the system---at which point the dual locus
$\boldsymbol{\kappa}=\boldsymbol{\kappa}_{1}-\boldsymbol{\kappa}_{2}$ of the
discriminant function satisfies the geometric locus of the decision boundary
in terms of the critical minimum eigenenergy $\left\Vert \boldsymbol{\kappa
}\right\Vert _{\min_{c}}^{2}$ and the minimum expected risk $\mathfrak{R}%
_{\mathfrak{\min}}\left(  \left\Vert \boldsymbol{\kappa}\right\Vert _{\min
_{c}}^{2}\right)  $ that is exhibited by the geometric locus of the novel
principal eigenaxis $\boldsymbol{\kappa}=\boldsymbol{\kappa}_{1}%
-\boldsymbol{\kappa}_{2}$ of the system.

The equilibrium requirement on the dual locus of the discriminant function at
the geometric locus of the decision boundary of a minimum risk binary
classification system is regulated by the KKT condition in (\ref{KKT 5}) and
the theorem of Karush, Kuhn, and Tucker
\citep{Sundaram1996}%
.

Let there be $l$ active scale factors $\psi_{i\ast}>0$ and $l$ extreme vectors
$k_{\mathbf{x}_{i\ast}}$, so that each extreme vector $k_{\mathbf{x}_{i\ast}}$
is scaled by a correlated scale factor $\psi_{i\ast}$. Also, let there be
$l_{1}$ scaled extreme vectors $\left\{  \psi_{1_{i\ast}}k_{\mathbf{x}%
_{1_{i\ast}}}\right\}  _{i=1}^{l_{1}}$ that belong to class $\omega_{1}$ and
$l_{2}$ scaled extreme vectors $\left\{  \psi_{2_{i\ast}}k_{\mathbf{x}%
_{2_{i\ast}}}\right\}  _{i=1}^{l_{2}}$ that belong to class $\omega_{2}$.

By the KKT condition in (\ref{KKT 5}) and the theorem of Karush, Kuhn, and
Tucker, it follows that the geometric locus of the novel principal eigenaxis
$\boldsymbol{\kappa}$, the $l$ scale factors $\psi_{i\ast}$, the $l$ extreme
vectors $k_{\mathbf{x}_{i\ast}}$ and $\boldsymbol{\kappa}_{0}$ satisfy the
following system of $l$ vector algebra locus equations%

\begin{equation}
\psi_{i\ast}\left[  y_{i}\left(  k_{\mathbf{x}_{i\ast}}\boldsymbol{\kappa
}+\boldsymbol{\kappa}_{0}\right)  -1+\xi_{i}\right]  =0,\ i=1,...,l
\tag{20.1}\label{KKT EigE Cond}%
\end{equation}
inside the Wolfe-dual principal eigenspace of the minimum risk binary
classification system $k_{\mathbf{s}}\boldsymbol{\kappa}+$ $\boldsymbol{\kappa
}_{0}\overset{\omega_{1}}{\underset{\omega_{2}}{\gtrless}}0$, where either
$\xi_{i}=\xi=0$ or $\xi_{i}=\xi\ll1$, e.g. $\xi_{i}=\xi=0.02$.

Now take the $l_{1}$ scaled extreme vectors $\left\{  \psi_{1i\ast
}k_{\mathbf{x}_{1_{i\ast}}}\right\}  _{i=1}^{l_{1}}$ that belong to class
$\omega_{1}$. Using the KKT condition in (\ref{KKT EigE Cond}) and letting
$y_{i}=+1$, it follows that the total allowed eigenenergy and the minimum
expected risk exhibited by side $\boldsymbol{\kappa}_{1}$ of the geometric
locus of the novel principal eigenaxis $\boldsymbol{\kappa}=\boldsymbol{\kappa
}_{1}-\boldsymbol{\kappa}_{2}$ are both determined by the vector algebra locus
equation%
\begin{align}
\left\Vert \boldsymbol{\kappa}_{1}\right\Vert _{\min_{c}}^{2}-\left\Vert
\boldsymbol{\kappa}_{1}\right\Vert \left\Vert \boldsymbol{\kappa}%
_{2}\right\Vert \cos\theta_{\boldsymbol{\kappa}_{1}\boldsymbol{\kappa}_{2}}
&  =\sum\nolimits_{i=1}^{l_{1}}\psi_{1_{i_{\ast}}}\left(  1-\xi_{i}%
-\boldsymbol{\kappa}_{0}\right) \tag{20.2}\label{EigE of Class One}\\
&  =\boldsymbol{\psi}_{1}\left(  1-\xi_{i}-\boldsymbol{\kappa}_{0}\right)
\text{,}\nonumber
\end{align}
so that the dual locus $\boldsymbol{\kappa}=\boldsymbol{\kappa}_{1}%
-\boldsymbol{\kappa}_{2}$ of the discriminant function $d\left(
\mathbf{s}\right)  =$ $k_{\mathbf{s}}\boldsymbol{\kappa}+$ $\boldsymbol{\kappa
}_{0}$ satisfies the geometric locus of the decision border $k_{\mathbf{s}%
}\boldsymbol{\kappa}+$ $\boldsymbol{\kappa}_{0}=+1$ in terms of the critical
minimum eigenenergy $\left\Vert \boldsymbol{\kappa}_{1}\right\Vert _{\min_{c}%
}^{2}$ and the minimum expected risk $\mathfrak{R}_{\mathfrak{\min}}\left(
\left\Vert \boldsymbol{\kappa}_{1}\right\Vert _{\min_{c}}^{2}\right)  $
exhibited by side $\boldsymbol{\kappa}_{1}$ of the geometric locus of the
novel principal eigenaxis $\boldsymbol{\kappa}=\boldsymbol{\kappa}%
_{1}-\boldsymbol{\kappa}_{2}$, at which point side $\boldsymbol{\kappa}_{1}$
and side $\boldsymbol{\psi}_{1}$ are symmetrically and equivalently related to
each other inside the Wolfe-dual principal eigenspace.

Next, take the $l_{2}$ scaled extreme vectors $\left\{  \psi_{2i\ast
}k_{\mathbf{x}_{2_{i\ast}}}\right\}  _{i=1}^{l_{2}}$ that belong to class
$\omega_{2}$. Using the KKT condition in (\ref{KKT EigE Cond}) and letting
$y_{i}=-1$, it follows that the total allowed eigenenergy and the minimum
expected risk exhibited by side $\boldsymbol{\kappa}_{2}$ of the geometric
locus of the novel principal eigenaxis $\boldsymbol{\kappa}=\boldsymbol{\kappa
}_{1}-\boldsymbol{\kappa}_{2}$ are both determined by the vector algebra locus
equation%
\begin{align}
\left\Vert \boldsymbol{\kappa}_{2}\right\Vert _{\min_{c}}^{2}-\left\Vert
\boldsymbol{\kappa}_{2}\right\Vert \left\Vert \boldsymbol{\kappa}%
_{1}\right\Vert \cos\theta_{\boldsymbol{\kappa}_{2}\boldsymbol{\kappa}_{1}}
&  =\sum\nolimits_{i=1}^{l_{2}}\psi_{2_{i_{\ast}}}\left(  1-\xi_{i}%
+\boldsymbol{\kappa}_{0}\right) \tag{20.3}\label{EigE of Class Two}\\
&  =\boldsymbol{\psi}_{2}\left(  1-\xi_{i}+\boldsymbol{\kappa}_{0}\right)
\text{,}\nonumber
\end{align}
so that the dual locus $\boldsymbol{\kappa}=\boldsymbol{\kappa}_{1}%
-\boldsymbol{\kappa}_{2}$ of the discriminant function $d\left(
\mathbf{s}\right)  =$ $k_{\mathbf{s}}\boldsymbol{\kappa}+$ $\boldsymbol{\kappa
}_{0}$ satisfies the geometric locus of the decision border $k_{\mathbf{s}%
}\boldsymbol{\kappa}+$ $\boldsymbol{\kappa}_{0}=-1$ in terms of the critical
minimum eigenenergy $\left\Vert \boldsymbol{\kappa}_{2}\right\Vert _{\min_{c}%
}^{2}$ and the minimum expected risk $\mathfrak{R}_{\mathfrak{\min}}\left(
\left\Vert \boldsymbol{\kappa}_{2}\right\Vert _{\min_{c}}^{2}\right)  $
exhibited by side $\boldsymbol{\kappa}_{2}$ of the geometric locus of the
novel principal eigenaxis $\boldsymbol{\kappa}=\boldsymbol{\kappa}%
_{1}-\boldsymbol{\kappa}_{2}$, at which point side $\boldsymbol{\kappa}_{2}$
and side $\boldsymbol{\psi}_{2}$ are symmetrically and equivalently related to
each other inside the Wolfe-dual principal eigenspace.

Summation over the complete system of vector algebra locus equations that are
satisfied by side $\boldsymbol{\kappa}_{1}$%
\[
\left(  \sum\nolimits_{i=1}^{l_{1}}\psi_{1_{i_{\ast}}}k_{\mathbf{x}%
_{1_{i_{\ast}}}}\right)  \boldsymbol{\kappa}=\sum\nolimits_{i=1}^{l_{1}}%
\psi_{1_{i_{\ast}}}\left(  1-\xi_{i}-\boldsymbol{\kappa}_{0}\right)
\]
and by side $\boldsymbol{\kappa}_{2}$%
\[
\left(  -\sum\nolimits_{i=1}^{l_{2}}\psi_{2_{i_{\ast}}}k_{\mathbf{x}%
_{2_{i\ast}}}\right)  \boldsymbol{\kappa}=\sum\nolimits_{i=1}^{l_{2}}%
\psi_{2_{i_{\ast}}}\left(  1-\xi_{i}+\boldsymbol{\kappa}_{0}\right)  \text{,}%
\]
and using the equilibrium constraint on the geometric locus of the Wolfe-dual
novel principal eigenaxis $\boldsymbol{\psi}$ in (\ref{Wolfe-dual EQU State})%
\[
\sum\nolimits_{i=1}^{l_{1}}\psi_{1i\ast}\frac{k_{\mathbf{x}_{1i\ast}}%
}{\left\Vert k_{\mathbf{x}_{1i\ast}}\right\Vert }=\sum\nolimits_{i=1}^{l_{2}%
}\psi_{2i\ast}\frac{k_{\mathbf{x}_{2i\ast}}}{\left\Vert k_{\mathbf{x}_{2i\ast
}}\right\Vert }\text{,}%
\]
produces the vector algebra locus equation that determines the total allowed
eigenenergy $\left\Vert \boldsymbol{\kappa}\right\Vert _{\min_{c}}^{2}$ and
the expected risk $\mathfrak{R}_{\mathfrak{\min}}\left(  \left\Vert
\boldsymbol{\kappa}\right\Vert _{\min_{c}}^{2}\right)  $ exhibited by the
geometric locus of the novel principal eigenaxis $\boldsymbol{\kappa
}=\boldsymbol{\kappa}_{1}-\boldsymbol{\kappa}_{2}$%
\begin{align}
\left(  \boldsymbol{\kappa}_{1}-\boldsymbol{\kappa}_{2}\right)
\boldsymbol{\kappa}  &  =\sum\nolimits_{i=1}^{l_{1}}\psi_{1i\ast}\left(
1-\xi_{i}-\boldsymbol{\kappa}_{0}\right)  +\sum\nolimits_{i=1}^{l_{2}}%
\psi_{2_{i_{\ast}}}\left(  1-\xi_{i}+\boldsymbol{\kappa}_{0}\right)
\tag{20.4}\label{EigE of System}\\
&  =\sum\nolimits_{i=1}^{l}\psi_{i_{\ast}}\left(  1-\xi_{i}\right)
=\boldsymbol{\psi}\left(  1-\xi_{i}\right)  \text{,}\nonumber
\end{align}
so that the dual locus $\boldsymbol{\kappa}=\boldsymbol{\kappa}_{1}%
-\boldsymbol{\kappa}_{2}$ of the discriminant function $d\left(
\mathbf{s}\right)  =$ $k_{\mathbf{s}}\boldsymbol{\kappa}+$ $\boldsymbol{\kappa
}_{0}$ satisfies the geometric locus of the decision boundary $k_{\mathbf{s}%
}\boldsymbol{\kappa}+$ $\boldsymbol{\kappa}_{0}=0$ in terms of the critical
minimum eigenenergy $\left\Vert \boldsymbol{\kappa}\right\Vert _{\min_{c}}%
^{2}$ and the minimum expected risk $\mathfrak{R}_{\mathfrak{\min}}\left(
\left\Vert \boldsymbol{\kappa}\right\Vert _{\min_{c}}^{2}\right)  $ exhibited
by the geometric locus of the novel principal eigenaxis $\boldsymbol{\kappa}$,
at which point the total allowed eigenenergy $\left\Vert \boldsymbol{\kappa
}\right\Vert _{\min_{c}}^{2}$ and the expected risk $\mathfrak{R}%
_{\mathfrak{\min}}\left(  \left\Vert \boldsymbol{\kappa}\right\Vert _{\min
_{c}}^{2}\right)  $ exhibited by the geometric locus of the novel principal
eigenaxis $\boldsymbol{\kappa}=\boldsymbol{\kappa}_{1}-\boldsymbol{\kappa}%
_{2}$ are both regulated by the total value of the summed scale factors
$\psi_{1i\ast}$ and $\psi_{2i\ast}$ for the components $\psi_{1i\ast}%
\frac{k_{\mathbf{x}_{1i\ast}}}{\left\Vert k_{\mathbf{x}_{1i\ast}}\right\Vert
}$ and $\psi_{2i\ast}\frac{k_{\mathbf{x}_{2i\ast}}}{\left\Vert k_{\mathbf{x}%
_{2i\ast}}\right\Vert }$ of the principal eigenvector $\boldsymbol{\psi}%
_{\max}$ of the joint covariance matrix $\mathbf{Q}$ and the inverted joint
covariance matrix $\mathbf{Q}^{-1}$ associated with the pair of random
quadratic forms $\boldsymbol{\psi}^{T}\mathbf{Q}\boldsymbol{\psi}$ and
$\boldsymbol{\psi}^{T}\mathbf{Q}^{-1}\boldsymbol{\psi}$.

By (\ref{EigE of System}), it follows that the total allowed eigenenergy
$\left\Vert \boldsymbol{\kappa}\right\Vert _{\min_{c}}^{2}$ and the expected
risk $\mathfrak{R}_{\mathfrak{\min}}\left(  \left\Vert \boldsymbol{\kappa
}\right\Vert _{\min_{c}}^{2}\right)  $ exhibited by the geometric locus of the
novel principal eigenaxis $\boldsymbol{\kappa}=\boldsymbol{\kappa}%
_{1}-\boldsymbol{\kappa}_{2}$ are both regulated by the total value of the
summed scale factors $\sum\nolimits_{i=1}^{l}\psi_{i_{\ast}}$ for the
components $\psi_{i\ast}\frac{k_{\mathbf{x}_{i\ast}}}{\left\Vert
k_{\mathbf{x}_{i\ast}}\right\Vert }$ of the principal eigenvector
$\boldsymbol{\psi}_{\max}$
\[
\left\Vert \boldsymbol{\kappa}\right\Vert _{\min_{c}}^{2}=\sum\nolimits_{i=1}%
^{l}\psi_{i_{\ast}}\left(  1-\xi_{i}\right)  =\sum\nolimits_{i=1}^{l}%
\psi_{i_{\ast}}-\sum\nolimits_{i=1}^{l}\psi_{i_{\ast}}\xi_{i}\text{,}%
\]
at which point the novel principal eigenaxes $\boldsymbol{\kappa}$ and
$\boldsymbol{\psi}$ are symmetrically and equivalently related to each other
inside the Wolfe-dual principal eigenspace, wherein the regularization
parameters $\xi_{i}=\xi\ll1$ are seen to determine negligible constraints.

We are now in a position to demonstrate how the dual locus $\boldsymbol{\kappa
}=\boldsymbol{\kappa}_{1}-\boldsymbol{\kappa}_{2}$ of the discriminant
function $d\left(  \mathbf{s}\right)  =$ $k_{\mathbf{s}}\boldsymbol{\kappa}+$
$\boldsymbol{\kappa}_{0}$ of any given minimum risk binary classification
system $k_{\mathbf{s}}\boldsymbol{\kappa}+$ $\boldsymbol{\kappa}%
_{0}\overset{\omega_{1}}{\underset{\omega_{2}}{\gtrless}}0$ is in statistical
equilibrium---at the geometric locus of the decision boundary $k_{\mathbf{s}%
}\boldsymbol{\kappa}+$ $\boldsymbol{\kappa}_{0}=0$ of the system---so that the
geometric locus of the novel principal eigenaxis $\boldsymbol{\kappa
}=\boldsymbol{\kappa}_{1}-\boldsymbol{\kappa}_{2}$ of the system exhibits
symmetrical dimensions and densities---at which point the total allowed
eigenenergy $\left\Vert \boldsymbol{\kappa}_{1}-\boldsymbol{\kappa}%
_{2}\right\Vert _{\min_{c}}^{2}$ and the expected risk $\mathfrak{R}%
_{\mathfrak{\min}}\left(  \left\Vert \boldsymbol{\kappa}_{1}%
-\boldsymbol{\kappa}_{2}\right\Vert _{\min_{c}}^{2}\right)  $ exhibited by the
geometric locus of the novel principal eigenaxis $\boldsymbol{\kappa
}=\boldsymbol{\kappa}_{1}-\boldsymbol{\kappa}_{2}$ are jointly minimized
within the decision space $Z=Z_{1}\cup Z_{2}$ of the system $k_{\mathbf{s}%
}\boldsymbol{\kappa}+$ $\boldsymbol{\kappa}_{0}\overset{\omega_{1}%
}{\underset{\omega_{2}}{\gtrless}}0$. We begin by considering the locus
equations in (\ref{EigE of Class One}) - (\ref{EigE of System}).

The vector algebra locus equations in (\ref{EigE of Class One}) -
(\ref{EigE of System}) indicate that the minimum risk binary classification
system $k_{\mathbf{s}}\boldsymbol{\kappa}+$ $\boldsymbol{\kappa}%
_{0}\overset{\omega_{1}}{\underset{\omega_{2}}{\gtrless}}0$ satisfies the law
of total allowed eigenenergy for minimum risk binary classification systems
expressed by the integral equation in (\ref{TIE1}). We now demonstrate that
the discriminant function $d\left(  \mathbf{s}\right)  =$ $k_{\mathbf{s}%
}\boldsymbol{\kappa}+$ $\boldsymbol{\kappa}_{0}$ is the solution of an
integral equation that determines a data-driven version of the general form of
the integral equation in (\ref{TIE1}).

\subsection{Balancing Acts at the Decision Boundary}

Take the discriminant function $d\left(  \mathbf{s}\right)  =$ $k_{\mathbf{s}%
}\boldsymbol{\kappa}+$ $\boldsymbol{\kappa}_{0}$ of any given minimum risk
binary classification system $k_{\mathbf{s}}\boldsymbol{\kappa}+$
$\boldsymbol{\kappa}_{0}\overset{\omega_{1}}{\underset{\omega_{2}}{\gtrless}%
}0$ that is represented by a geometric locus of a novel principal eigenaxis
$\boldsymbol{\kappa}=\boldsymbol{\kappa}_{1}-\boldsymbol{\kappa}_{2}$.

We have used (\ref{Primal Dual Locus}), (\ref{Funct Primal Dual Locus}),
(\ref{Decision Boundary}) - (\ref{Decision Border 2}),
(\ref{Wolfe-dual EQU Point}) and (\ref{EigE of Class One}) -
(\ref{EigE of System}) to demonstrate that the geometric locus of the novel
principal eigenaxis $\boldsymbol{\kappa}=\boldsymbol{\kappa}_{1}%
-\boldsymbol{\kappa}_{2}$ satisfies the vector algebra locus equations%
\begin{equation}
\left\Vert \boldsymbol{\kappa}_{1}\right\Vert _{\min_{c}}^{2}-\left\Vert
\boldsymbol{\kappa}_{1}\right\Vert \left\Vert \boldsymbol{\kappa}%
_{2}\right\Vert \cos\theta_{\boldsymbol{\kappa}_{1}\boldsymbol{\kappa}_{2}%
}+\delta\left(  y\right)  \sum\nolimits_{i=1}^{l_{1}}\psi_{1_{i\ast}}=\frac
{1}{2}\left\Vert \boldsymbol{\kappa}\right\Vert _{\min_{c}}^{2}\text{,}
\tag{20.5}\label{EqualizerEq 1}%
\end{equation}
and%
\begin{equation}
\left\Vert \boldsymbol{\kappa}_{2}\right\Vert _{\min_{c}}^{2}-\left\Vert
\boldsymbol{\kappa}_{2}\right\Vert \left\Vert \boldsymbol{\kappa}%
_{1}\right\Vert \cos\theta_{\boldsymbol{\kappa}_{2}\boldsymbol{\kappa}_{1}%
}-\delta\left(  y\right)  \sum\nolimits_{i=1}^{l_{2}}\psi_{2_{i\ast}}=\frac
{1}{2}\left\Vert \boldsymbol{\kappa}\right\Vert _{\min_{c}}^{2}\text{,}
\tag{20.6}\label{EqualizerEq 2}%
\end{equation}
over the decision space $Z=Z_{1}\cup Z_{2}$ of the minimum risk binary
classification system $k_{\mathbf{s}}\boldsymbol{\kappa}+$ $\boldsymbol{\kappa
}_{0}\overset{\omega_{1}}{\underset{\omega_{2}}{\gtrless}}0$, where the
expressions $\delta\left(  y\right)  \sum\nolimits_{i=1}^{l_{1}}\psi
_{1_{i\ast}}$ and $-\delta\left(  y\right)  \sum\nolimits_{i=1}^{l_{2}}%
\psi_{2_{i\ast}}$ are equalizer statistics, such that $\delta\left(  y\right)
\triangleq\frac{1}{l}\sum\nolimits_{i=1}^{l}y_{i}\left(  1-\xi_{i}\right)  $,
where $y_{i}\in Y=\left\{  \pm1\right\}  $
\citep{Reeves2018design}%
.

We have used the vector algebra locus equations in (\ref{ID 1 Wolf-dual}) and
(\ref{ID 2 Wolf-dual}), along with the vector algebra locus equations in
(\ref{EqualizerEq 1}) and (\ref{EqualizerEq 2}) to demonstrate that the
geometric locus of the novel principal eigenaxis $\boldsymbol{\kappa
}=\boldsymbol{\kappa}_{1}-\boldsymbol{\kappa}_{2}$ also satisfies the vector
algebra locus equation%
\begin{align}
&  \left\Vert \boldsymbol{\kappa}_{1}\right\Vert _{\min_{c}}^{2}-\left\Vert
\boldsymbol{\kappa}_{1}\right\Vert \left\Vert \boldsymbol{\kappa}%
_{2}\right\Vert \cos\theta_{\boldsymbol{\kappa}_{1}\boldsymbol{\kappa}_{2}%
}+\delta\left(  y\right)  \lambda_{\max_{\boldsymbol{\psi}}}^{-1}%
\sum\nolimits_{i=1}^{l_{1}}k_{\mathbf{x}_{1_{i\ast}}}\boldsymbol{\kappa
}\tag{20.7}\label{Cond ID DiscrimF}\\
&  =\left\Vert \boldsymbol{\kappa}_{2}\right\Vert _{\min_{c}}^{2}-\left\Vert
\boldsymbol{\kappa}_{2}\right\Vert \left\Vert \boldsymbol{\kappa}%
_{1}\right\Vert \cos\theta_{\boldsymbol{\kappa}_{2}\boldsymbol{\kappa}_{1}%
}-\delta\left(  y\right)  \lambda_{\max_{\boldsymbol{\psi}}}^{-1}%
\sum\nolimits_{i=1}^{l_{2}}k_{\mathbf{x}_{2_{i\ast}}}\boldsymbol{\kappa
}\nonumber\\
&  =\frac{1}{2}\left\Vert \boldsymbol{\kappa}\right\Vert _{\min_{c}}^{2}%
=\frac{1}{2}\mathfrak{R}_{\mathfrak{\min}}\left(  \left\Vert
\boldsymbol{\kappa}\right\Vert _{\min_{c}}^{2}\right)  \text{,}\nonumber
\end{align}
at which point the expressions that enter into the left-hand side and the
right-hand side of (\ref{Cond ID DiscrimF}) both satisfy half the total
allowed eigenenergy and half the expected risk exhibited by the minimum risk
binary classification system $k_{\mathbf{s}}\boldsymbol{\kappa}+$
$\boldsymbol{\kappa}_{0}\overset{\omega_{1}}{\underset{\omega_{2}}{\gtrless}%
}0$ within the decision space $Z=Z_{1}\cup Z_{2}$ of the system
\citep{Reeves2018design}%
.

\subsection{The Applied Law of Cosines}

Using the vector algebra locus equation in (\ref{Primal EQ State 1})%
\[
\sum\nolimits_{i=1}^{l_{1}}k_{\mathbf{x}_{1_{i\ast}}}\left(
\boldsymbol{\kappa}_{1}\mathbf{-}\boldsymbol{\kappa}_{2}\right)
=\sum\nolimits_{i=1}^{l_{2}}k_{\mathbf{x}_{2_{i\ast}}}\left(
\boldsymbol{\kappa}_{2}\mathbf{-}\boldsymbol{\kappa}_{1}\right)  \text{,}%
\]
wherein extreme vectors $k_{\mathbf{x}_{1_{i\ast}}}$ and $k_{\mathbf{x}%
_{2_{i\ast}}}$from class $\omega_{1}$ and class $\omega_{2}$ are distributed
over side $\boldsymbol{\kappa}_{1}$ and side $\boldsymbol{\kappa}_{2}$ of the
geometric locus of the novel principal eigenaxis $\boldsymbol{\kappa
}=\boldsymbol{\kappa}_{1}-\boldsymbol{\kappa}_{2}$ in a symmetrically balanced
manner, along with the vector algebra locus equation in
(\ref{Cond ID DiscrimF}), it follows that the geometric locus of the novel
principal eigenaxis $\boldsymbol{\kappa}=\boldsymbol{\kappa}_{1}%
-\boldsymbol{\kappa}_{2}$ satisfies the law of cosines in the symmetrically
balanced manner%
\begin{align*}
\frac{1}{2}\left\Vert \boldsymbol{\kappa}\right\Vert _{\min_{c}}^{2}  &
=\left\Vert \boldsymbol{\kappa}_{1}\right\Vert _{\min_{c}}^{2}-\left\Vert
\boldsymbol{\kappa}_{1}\right\Vert \left\Vert \boldsymbol{\kappa}%
_{2}\right\Vert \cos\theta_{\boldsymbol{\kappa}_{1}\boldsymbol{\kappa}_{2}}\\
&  =\left\Vert \boldsymbol{\kappa}_{2}\right\Vert _{\min_{c}}^{2}-\left\Vert
\boldsymbol{\kappa}_{2}\right\Vert \left\Vert \boldsymbol{\kappa}%
_{1}\right\Vert \cos\theta_{\boldsymbol{\kappa}_{2}\boldsymbol{\kappa}_{1}%
}\text{,}%
\end{align*}
where $\theta$ is the angle between side $\boldsymbol{\kappa}_{1}$ and side
$\boldsymbol{\kappa}_{2}$ of the geometric locus of the novel principal
eigenaxis $\boldsymbol{\kappa}=\boldsymbol{\kappa}_{1}-\boldsymbol{\kappa}%
_{2}$, at which point the magnitude and the direction of the novel principal
eigenaxis $\boldsymbol{\kappa}=\boldsymbol{\kappa}_{1}-\boldsymbol{\kappa}%
_{2}$ are both functions of differences between joint variabilities of extreme
vectors $\mathbf{x}_{1_{i\ast}}\mathbf{\sim}p\left(  \mathbf{x};\omega
_{1}\right)  $ and $\mathbf{x}_{2_{i\ast}}\mathbf{\sim}p\left(  \mathbf{x}%
;\omega_{1}\right)  $.

The vector algebra locus equations in (\ref{Primal EQ State 1}) and
(\ref{Cond ID DiscrimF}) demonstrate how the constrained optimization
algorithm that resolves the inverse problem of binary classification explains
and executes the law of symmetry for minimum risk binary classification systems.

Returning now to the integral in (\ref{Cond-Prob-Function Class 1}) that
determines the conditional probability $P\left(  \mathbf{x}_{1_{i\ast}%
}|\boldsymbol{\kappa}_{1}\right)  $ for class $\omega_{1}$, along with the
integral in (\ref{Cond-Prob-Function Class 2}) that determines the conditional
probability $P\left(  \mathbf{x}_{2_{i\ast}}|\boldsymbol{\kappa}_{2}\right)  $
for class $\omega_{2}$, it follows that the value for the integration constant
$C_{1}$ in (\ref{Cond-Prob-Function Class 1}) is%
\[
C_{1}=-\left\Vert \boldsymbol{\kappa}_{1}\right\Vert \left\Vert
\boldsymbol{\kappa}_{2}\right\Vert \cos\theta_{\boldsymbol{\kappa}%
_{1}\boldsymbol{\kappa}_{2}}\text{,}%
\]
and the value for the integration constant $C_{2}$ in
(\ref{Cond-Prob-Function Class 2}) is%
\[
C_{2}=-\left\Vert \boldsymbol{\kappa}_{2}\right\Vert \left\Vert
\boldsymbol{\kappa}_{1}\right\Vert \cos\theta_{\boldsymbol{\kappa}%
_{2}\boldsymbol{\kappa}_{1}}\text{.}%
\]

Substituting the value for $C_{1}$ into the expression for the integral in
(\ref{Cond-Prob-Function Class 1}), and using
(\ref{Cond-Prob-Function Class 1}) and (\ref{Cond ID DiscrimF}), it follows
that the conditional risk $\mathfrak{R}_{\mathfrak{\min}}\left(
\mathbf{x}_{1_{i\ast}}|\boldsymbol{\kappa}_{1}\right)  $ for class $\omega
_{1}$ is given by the integral%
\begin{align}
P\left(  \mathbf{x}_{1_{i\ast}}|\boldsymbol{\kappa}_{1}\right)   &  =\int%
_{Z}\boldsymbol{\kappa}_{1}d\boldsymbol{\kappa}_{1}=\left\Vert
\boldsymbol{\kappa}_{1}\right\Vert _{\min_{c}}^{2}-\left\Vert
\boldsymbol{\kappa}_{1}\right\Vert \left\Vert \boldsymbol{\kappa}%
_{2}\right\Vert \cos\theta_{\boldsymbol{\kappa}_{1}\boldsymbol{\kappa}_{2}%
}\tag{20.8}\label{Int Cond Prob 1}\\
&  +\delta\left(  y\right)  \lambda_{\max_{\boldsymbol{\psi}}}^{-1}%
\sum\nolimits_{i=1}^{l_{1}}k_{\mathbf{x}_{1_{i\ast}}}\left(
\boldsymbol{\kappa}_{1}-\boldsymbol{\kappa}_{2}\right) \nonumber\\
&  =\frac{1}{2}\left\Vert \boldsymbol{\kappa}_{1}-\boldsymbol{\kappa}%
_{2}\right\Vert _{\min_{c}}^{2}=\frac{1}{2}\mathfrak{R}_{\mathfrak{\min}%
}\left(  \left\Vert \boldsymbol{\kappa}_{1}-\boldsymbol{\kappa}_{2}\right\Vert
_{\min_{c}}^{2}\right)  \text{,}\nonumber
\end{align}
over the decision space $Z=Z_{1}\cup Z_{2}$ of the minimum risk binary
classification system $k_{\mathbf{s}}\boldsymbol{\kappa}+$ $\boldsymbol{\kappa
}_{0}\overset{\omega_{1}}{\underset{\omega_{2}}{\gtrless}}0$, so that the
expression for the integral in (\ref{Cond-Prob-Function Class 2}) determines
the conditional probability $P\left(  k_{\mathbf{x}_{1_{i\ast}}}%
|\boldsymbol{\kappa}_{1}\right)  $ of observing a set $\left\{  \mathbf{x}%
_{1_{i\ast}}\right\}  _{i=1}^{l_{1}}$ of $l_{1}$ extreme points $\mathbf{x}%
_{1_{i\ast}}$ within localized areas of the decision space $Z=Z_{1}\cup Z_{2}%
$, at which point the conditional probability $P\left(  k_{\mathbf{x}%
_{1_{i\ast}}}|\boldsymbol{\kappa}_{1}\right)  $ for class $\omega_{1}$ is
equal to half the total allowed eigenenergy $\frac{1}{2}\left\Vert
\boldsymbol{\kappa}_{1}-\boldsymbol{\kappa}_{2}\right\Vert _{\min_{c}}^{2}$
and half the expected risk $\frac{1}{2}R_{\mathfrak{\min}}\left(  \left\Vert
\boldsymbol{\kappa}_{1}-\boldsymbol{\kappa}_{2}\right\Vert _{\min_{c}}%
^{2}\right)  $ that is exhibited by the geometric locus of the novel principal
eigenaxis $\boldsymbol{\kappa}=\boldsymbol{\kappa}_{1}-\boldsymbol{\kappa}%
_{2}$.

Substituting the value for $C_{2}$ into the expression for the integral in
(\ref{Cond-Prob-Function Class 2}), and using
(\ref{Cond-Prob-Function Class 2}) and (\ref{Cond ID DiscrimF}), it follows
that the conditional risk $\mathfrak{R}_{\mathfrak{\min}}\left(
\mathbf{x}_{2_{i\ast}}|\boldsymbol{\kappa}_{2}\right)  $ for class $\omega
_{2}$ is given by the integral%
\begin{align}
P\left(  \mathbf{x}_{2_{i\ast}}|\boldsymbol{\kappa}_{2}\right)   &  =\int%
_{Z}\boldsymbol{\kappa}_{2}d\boldsymbol{\kappa}_{2}=\left\Vert
\boldsymbol{\kappa}_{2}\right\Vert _{\min_{c}}^{2}-\left\Vert
\boldsymbol{\kappa}_{2}\right\Vert \left\Vert \boldsymbol{\kappa}%
_{1}\right\Vert \cos\theta_{\boldsymbol{\kappa}_{2}\boldsymbol{\kappa}_{1}%
}\tag{20.9}\label{Int Cond Prob 2}\\
&  +\delta\left(  y\right)  \lambda_{\max_{\boldsymbol{\psi}}}^{-1}%
\sum\nolimits_{i=1}^{l_{2}}k_{\mathbf{x}_{2_{i\ast}}}\left(
\boldsymbol{\kappa}_{2}-\boldsymbol{\kappa}_{1}\right) \nonumber\\
&  =\frac{1}{2}\left\Vert \boldsymbol{\kappa}_{1}-\boldsymbol{\kappa}%
_{2}\right\Vert _{\min_{c}}^{2}=\frac{1}{2}\mathfrak{R}_{\mathfrak{\min}%
}\left(  \left\Vert \boldsymbol{\kappa}_{1}-\boldsymbol{\kappa}_{2}\right\Vert
_{\min_{c}}^{2}\right)  \text{,}\nonumber
\end{align}
over the decision space $Z=Z_{1}\cup Z_{2}$ of the minimum risk binary
classification system $k_{\mathbf{s}}\boldsymbol{\kappa}+$ $\boldsymbol{\kappa
}_{0}\overset{\omega_{1}}{\underset{\omega_{2}}{\gtrless}}0$, so that the
expression for the integral in (\ref{Int Cond Prob 2}) determines the
conditional probability $P\left(  \mathbf{x}_{2_{i\ast}}|\boldsymbol{\kappa
}_{2}\right)  $ of observing a set $\left\{  \mathbf{x}_{2_{i\ast}}\right\}
_{i=1}^{l_{2}}$ of $l_{2}$ extreme points $\mathbf{x}_{2_{i\ast}}$ within
localized areas of the decision space $Z=Z_{1}\cup Z_{2}$, at which point the
conditional probability $P\left(  \mathbf{x}_{2_{i\ast}}|\boldsymbol{\kappa
}_{2}\right)  $ for class $\omega_{2}$ is equal to half the total allowed
eigenenergy $\frac{1}{2}\left\Vert \boldsymbol{\kappa}_{1}-\boldsymbol{\kappa
}_{2}\right\Vert _{\min_{c}}^{2}$ and half the expected risk $\frac{1}%
{2}\mathfrak{R}_{\mathfrak{\min}}\left(  \left\Vert \boldsymbol{\kappa}%
_{1}-\boldsymbol{\kappa}_{2}\right\Vert _{\min_{c}}^{2}\right)  $ that is
exhibited by the geometric locus of the novel principal eigenaxis
$\boldsymbol{\kappa=\kappa}_{1}-\boldsymbol{\kappa}_{2}$.

\subsection{The Applied Law of Total Allowed Eigenenergy}

Using the expressions for the integrals in (\ref{Int Cond Prob 1}) and
(\ref{Int Cond Prob 2}), it follows that the discriminant function is the
solution of the integral equation%
\begin{align}
f_{1}\left(  d\left(  \mathbf{s}\right)  \right)   &  :\int_{Z_{1}%
}\boldsymbol{\kappa}_{1}d\boldsymbol{\kappa}_{1}+\int_{Z_{2}}%
\boldsymbol{\kappa}_{1}d\boldsymbol{\kappa}_{1}+\delta\left(  y\right)
\lambda_{\max_{\boldsymbol{\psi}}}^{-1}\sum\nolimits_{i=1}^{l_{1}%
}k_{\mathbf{x}_{1_{i\ast}}}\left(  \boldsymbol{\kappa}_{1}-\boldsymbol{\kappa
}_{2}\right) \tag{20.10}\label{AIE1}\\
&  =\int_{Z_{1}}\boldsymbol{\kappa}_{2}d\boldsymbol{\kappa}_{2}+\int_{Z_{2}%
}\boldsymbol{\kappa}_{2}d\boldsymbol{\kappa}_{2}+\delta\left(  y\right)
\lambda_{\max_{\boldsymbol{\psi}}}^{-1}\sum\nolimits_{i=1}^{l_{2}%
}k_{\mathbf{x}_{2_{i\ast}}}\left(  \boldsymbol{\kappa}_{2}-\boldsymbol{\kappa
}_{1}\right)  \text{,}\nonumber
\end{align}
over the decision space $Z=Z_{1}\cup Z_{2}$ of the minimum risk binary
classification system $k_{\mathbf{s}}\boldsymbol{\kappa}+$ $\boldsymbol{\kappa
}_{0}\overset{\omega_{1}}{\underset{\omega_{2}}{\gtrless}}0$, so that the
total allowed eigenenergy $\left\Vert \boldsymbol{\kappa}_{1}%
-\boldsymbol{\kappa}_{2}\right\Vert _{\min_{c}}^{2}$ and the expected risk
$\mathfrak{R}_{\mathfrak{\min}}\left(  \left\Vert \boldsymbol{\kappa}%
_{1}-\boldsymbol{\kappa}_{2}\right\Vert _{\min_{c}}^{2}\right)  $ exhibited by
the system are jointly regulated by the equilibrium requirement on the dual
locus $\boldsymbol{\kappa=\kappa}_{1}-\boldsymbol{\kappa}_{2}$ of the
discriminant function at the geometric locus of decision boundary of the
system%
\begin{align*}
d\left(  \mathbf{s}\right)   &  :\left\Vert \boldsymbol{\kappa}_{1}\right\Vert
_{\min_{c}}^{2}-\left\Vert \boldsymbol{\kappa}_{1}\right\Vert \left\Vert
\boldsymbol{\kappa}_{2}\right\Vert \cos\theta_{\boldsymbol{\kappa}%
_{1}\boldsymbol{\kappa}_{2}}\\
&  =\left\Vert \boldsymbol{\kappa}_{2}\right\Vert _{\min_{c}}^{2}-\left\Vert
\boldsymbol{\kappa}_{2}\right\Vert \left\Vert \boldsymbol{\kappa}%
_{1}\right\Vert \cos\theta_{\boldsymbol{\kappa}_{2}\boldsymbol{\kappa}_{1}}\\
&  =\frac{1}{2}\left\Vert \boldsymbol{\kappa}_{1}-\boldsymbol{\kappa}%
_{2}\right\Vert _{\min_{c}}^{2}\text{,}%
\end{align*}
at which point the dual locus $\boldsymbol{\kappa=\kappa}_{1}%
-\boldsymbol{\kappa}_{2}$ of the discriminant function is an eigenaxis of
symmetry that satisfies the geometric locus of the decision boundary in terms
of a critical minimum eigenenergy $\left\Vert \boldsymbol{\kappa}\right\Vert
_{\min_{c}}^{2}$ and a minimum expected risk $\mathfrak{R}_{\mathfrak{\min}%
}\left(  \left\Vert \boldsymbol{\kappa}\right\Vert _{\min_{c}}^{2}\right)  $
in such a manner that regions of counter risks of the system are symmetrically
balanced with regions of risks of the system, so that critical minimum
eigenenergies $\left\Vert \psi_{1_{i\ast}}k_{\mathbf{x}_{1_{i\ast}}%
}\right\Vert _{\min_{c}}^{2}$ exhibited by principal eigenaxis components
$\psi_{1_{i_{\ast}}}k_{\mathbf{x}_{1_{i\ast}}}$ on side $\boldsymbol{\kappa
}_{1}$ that determine probabilities of finding extreme points $\mathbf{x}%
_{1_{i\ast}}$ located throughout the decision space $Z=Z_{1}\cup Z_{2}$ of the
system---are symmetrically balanced with critical minimum eigenenergies
$\left\Vert \psi_{2_{i_{\ast}}}k_{\mathbf{x}_{2_{i\ast}}}\right\Vert
_{\min_{c}}^{2}$ exhibited by principal eigenaxis components $\psi
_{2_{i_{\ast}}}k_{\mathbf{x}_{2_{i\ast}}}$ on side $\boldsymbol{\kappa}_{2}$
that determine probabilities of finding extreme points $\mathbf{x}_{2_{i\ast}%
}$ located throughout the decision space $Z=Z_{1}\cup Z_{2}$ of the system.

\subsection{The Applied Law of Symmetry}

Given (\ref{Primal EQ State 1}) and (\ref{Cond ID DiscrimF}), along with
(\ref{AIE1}), it follows that the geometric locus of the novel principal
eigenaxis $\boldsymbol{\kappa}=\boldsymbol{\kappa}_{1}-\boldsymbol{\kappa}%
_{2}$ satisfies the law of cosines in the symmetrically balanced manner%
\begin{align}
\frac{1}{2}\left\Vert \boldsymbol{\kappa}\right\Vert _{\min_{c}}^{2}  &
=\left\Vert \boldsymbol{\kappa}_{1}\right\Vert _{\min_{c}}^{2}-\left\Vert
\boldsymbol{\kappa}_{1}\right\Vert \left\Vert \boldsymbol{\kappa}%
_{2}\right\Vert \cos\theta_{\boldsymbol{\kappa}_{1}\boldsymbol{\kappa}_{2}%
}\tag{20.11}\label{Applied Law of Cosines}\\
&  =\left\Vert \boldsymbol{\kappa}_{2}\right\Vert _{\min_{c}}^{2}-\left\Vert
\boldsymbol{\kappa}_{2}\right\Vert \left\Vert \boldsymbol{\kappa}%
_{1}\right\Vert \cos\theta_{\boldsymbol{\kappa}_{2}\boldsymbol{\kappa}_{1}%
}\text{,}\nonumber
\end{align}
where $\theta$ is the angle between side $\boldsymbol{\kappa}_{1}$ and side
$\boldsymbol{\kappa}_{2}$ of the geometric locus of the novel principal
eigenaxis $\boldsymbol{\kappa}=\boldsymbol{\kappa}_{1}-\boldsymbol{\kappa}%
_{2}$, at which point the geometric locus of the novel principal eigenaxis
$\boldsymbol{\rho}=\boldsymbol{\rho}_{1}-\boldsymbol{\rho}_{2}$ is an
eigenaxis of symmetry that exhibits symmetrical dimensions and densities, so
that counteracting and opposing forces and influences of the minimum risk
binary classification system $k_{\mathbf{s}}\boldsymbol{\kappa}+$
$\boldsymbol{\kappa}_{0}\overset{\omega_{1}}{\underset{\omega_{2}}{\gtrless}%
}0$ are symmetrically balanced with each other---about the geometric center of
the locus of the novel principal eigenaxis---whereon the statistical fulcrum
of the system is located.

By (\ref{Applied Law of Cosines}), it follow that the minimum risk binary
classification system $k_{\mathbf{s}}\boldsymbol{\kappa}+$ $\boldsymbol{\kappa
}_{0}\overset{\omega_{1}}{\underset{\omega_{2}}{\gtrless}}0$ achieves a state
of statistical equilibrium, so that the geometric locus of the novel principal
eigenaxis $\boldsymbol{\kappa}=\boldsymbol{\kappa}_{1}-\boldsymbol{\kappa}%
_{2}$ of the system exhibits symmetrical dimensions and densities, at which
point critical minimum eigenenergies exhibited by the system are symmetrically
concentrated in such a manner that counteracting and opposing forces and
influences of the system are symmetrically balanced with each other---about
the geometric center of the locus of the novel principal eigenaxis---whereon
the statistical fulcrum of the system is located.

\subsection{The Applied Law of Statistical Equilibrium}

Since the discriminant function is the solution of the integral equation of
(\ref{AIE1}), it follows that the discriminant function minimizes the integral
equation%
\begin{align}
f_{2}\left(  d\left(  \mathbf{s}\right)  \right)   &  :\int_{Z_{1}%
}\boldsymbol{\kappa}_{1}d\boldsymbol{\kappa}_{1}-\int_{Z_{1}}%
\boldsymbol{\kappa}_{2}d\boldsymbol{\kappa}_{2}+\delta\left(  y\right)
\lambda_{1}^{-1}\sum\nolimits_{i=1}^{l_{1}}k_{\mathbf{x}_{1_{i\ast}}}\left(
\boldsymbol{\kappa}_{1}-\boldsymbol{\kappa}_{2}\right)  \tag{20.12}%
\label{AIE2}\\
&  =\int_{Z_{2}}\boldsymbol{\kappa}_{2}d\boldsymbol{\kappa}_{2}-\int_{Z_{2}%
}\boldsymbol{\kappa}_{1}d\boldsymbol{\kappa}_{1}+\delta\left(  y\right)
\lambda_{1}^{-1}\sum\nolimits_{i=1}^{l_{2}}k_{\mathbf{x}_{2_{i\ast}}}\left(
\boldsymbol{\kappa}_{2}-\boldsymbol{\kappa}_{1}\right)  \text{,}\nonumber
\end{align}
over the decision regions $Z_{1}$ and $Z_{2}$ of the minimum risk binary
classification system $k_{\mathbf{s}}\boldsymbol{\kappa}+$ $\boldsymbol{\kappa
}_{0}\overset{\omega_{1}}{\underset{\omega_{2}}{\gtrless}}0$, so that the
system satisfies a state of statistical equilibrium wherein the total allowed
eigenenergy $\left\Vert \boldsymbol{\kappa}_{1}-\boldsymbol{\kappa}%
_{2}\right\Vert _{\min_{c}}^{2}$ and the expected risk $\mathfrak{R}%
_{\mathfrak{\min}}\left(  \left\Vert \boldsymbol{\kappa}_{1}%
-\boldsymbol{\kappa}_{2}\right\Vert _{\min_{c}}^{2}\right)  $ exhibited by the
system are jointly minimized within the decision space $Z=Z_{1}\cup Z_{2}$ of
the system in such a manner that critical minimum eigenenergies $\left\Vert
\psi_{1_{i\ast}}k_{\mathbf{x}_{1_{i\ast}}}\right\Vert _{\min_{c}}^{2}$ and
$\left\Vert \psi_{2_{i_{\ast}}}k_{\mathbf{x}_{2_{i\ast}}}\right\Vert
_{\min_{c}}^{2}$ exhibited by corresponding principal eigenaxis components
$\psi_{1_{i_{\ast}}}k_{\mathbf{x}_{1_{i\ast}}}$ and $\psi_{2_{i_{\ast}}%
}k_{\mathbf{x}_{2_{i\ast}}}$ on side $\boldsymbol{\kappa}_{1}$ and side
$\boldsymbol{\kappa}_{2}$ of the novel principal eigenaxis $\boldsymbol{\kappa
}=\boldsymbol{\kappa}_{1}-\boldsymbol{\kappa}_{2}$ are minimized throughout
the decision regions $Z_{1}$ and $Z_{2}$ of the system, at which point regions
of counter risks and risks of the system located throughout the decision
region $Z_{1}$ are symmetrically balanced with regions of counter risks and
risks of the system located throughout the decision region $Z_{2}$.

Thereby, the minimum risk binary classification system $k_{\mathbf{s}%
}\boldsymbol{\kappa}+$ $\boldsymbol{\kappa}_{0}\overset{\omega_{1}%
}{\underset{\omega_{2}}{\gtrless}}0$ satisfies a state of statistical
equilibrium so that the total allowed eigenenergy and the expected risk
exhibited by the system are jointly minimized within the decision space
$Z=Z_{1}\cup Z_{2}$ of the system, at which point the system exhibits the
minimum probability of classification error for any given feature vectors
$\mathbf{x\in}$ $%
\mathbb{R}
^{d}$ such that $\mathbf{x\sim}$ $p\left(  \mathbf{x};\omega_{1}\right)  $ and
$\mathbf{x\sim}$ $p\left(  \mathbf{x};\omega_{2}\right)  $.

\subsection{Elegant Relations and Deep-seated Interconnections}

We previously noted that the overall structure and behavior and properties of
any given system are intimately intertwined. At this point in our treatise, we
make the following observation.

We realize that the generalization behavior---and all of the surprising
statistical balancing feats---exhibited by the discriminant of any given
minimum risk binary classification system%
\[
\left(  k_{\mathbf{s}}\boldsymbol{-}\frac{1}{l}\sum\nolimits_{i=1}%
^{l}k_{\mathbf{x}_{i\ast}}\right)  \left(  \boldsymbol{\kappa}_{1}%
-\boldsymbol{\kappa}_{2}\right)  +\frac{1}{l}\sum\nolimits_{i=1}^{l}%
y_{i}\left(  1-\xi_{i}\right)  \overset{\omega_{1}}{\underset{\omega
_{2}}{\gtrless}}0
\]
are both enabled by elegant statistical relations and deep-seated statistical
interconnections between each and every one of the principal eigenaxis
components and likelihood components that lie on both sides of the primal
novel principal eigenaxis $\boldsymbol{\kappa}=\boldsymbol{\kappa}%
_{1}-\boldsymbol{\kappa}_{2}$%
\begin{align*}
\boldsymbol{\kappa}  &  =\sum\nolimits_{i=1}^{l_{1}}\psi_{1_{i\ast}%
}k_{\mathbf{x}_{1_{i\ast}}}-\sum\nolimits_{i=1}^{l_{2}}\psi_{2_{i\ast}%
}k_{\mathbf{x}_{2_{i\ast}}}\\
&  =\boldsymbol{\kappa}_{1}{\large -}\boldsymbol{\kappa}_{2}%
\end{align*}
and the Wolfe-dual novel principal eigenaxis $\boldsymbol{\psi}%
=\boldsymbol{\psi}_{1}+\boldsymbol{\psi}_{2}$%
\begin{align*}
\boldsymbol{\psi}  &  =\sum\nolimits_{i=1}^{l_{1}}\psi_{1i\ast}\frac
{k_{\mathbf{x}_{1i\ast}}}{\left\Vert k_{\mathbf{x}_{1i\ast}}\right\Vert }%
+\sum\nolimits_{i=1}^{l_{2}}\psi_{2i\ast}\frac{k_{\mathbf{x}_{2i\ast}}%
}{\left\Vert k_{\mathbf{x}_{2i\ast}}\right\Vert }\\
&  =\boldsymbol{\psi}_{1}+\boldsymbol{\psi}_{2}%
\end{align*}
of the system, at which point critical \emph{interconnections}---between all
of the \emph{intrinsic components} of the minimum risk binary classification
system---are \emph{blended} into a \emph{cohesive set} of essential components
\emph{by a general locus formula}, so that the overall structure and behavior
and properties of the system are intimately intertwined.

Therefore, take any given primal novel principal eigenaxis $\boldsymbol{\kappa
}=\boldsymbol{\kappa}_{1}-\boldsymbol{\kappa}_{2}$ and Wolfe-dual novel
principal eigenaxis $\boldsymbol{\psi}=\boldsymbol{\psi}_{1}+\boldsymbol{\psi
}_{2}$ of a minimum risk binary classification system%
\[
\left(  k_{\mathbf{s}}\boldsymbol{-}\frac{1}{l}\sum\nolimits_{i=1}%
^{l}k_{\mathbf{x}_{i\ast}}\right)  \left(  \boldsymbol{\kappa}_{1}%
-\boldsymbol{\kappa}_{2}\right)  +\frac{1}{l}\sum\nolimits_{i=1}^{l}%
y_{i}\left(  1-\xi_{i}\right)  \overset{\omega_{1}}{\underset{\omega
_{2}}{\gtrless}}0\text{,}%
\]
where scale factors $\psi_{1i\ast}$ and $\psi_{2i\ast}$ for the Wolfe-dual
novel principal eigenaxis $\boldsymbol{\psi}=\boldsymbol{\psi}_{1}%
+\boldsymbol{\psi}_{2}$ of the system determine scale factors $\psi_{1i\ast}$
and $\psi_{2i\ast}$ for the primal novel principal eigenaxis
$\boldsymbol{\kappa}=\boldsymbol{\kappa}_{1}-\boldsymbol{\kappa}_{2}$ of the system.

We now identify critical interconnections---between the intrinsic components
of a minimum risk binary classification system---that determine the
\emph{statistical structure} and the \emph{functionality} of the discriminant
function of the system, so that the discriminant function of the minimum risk
binary classification system generalizes and thereby extrapolates in a
significant manner.

We have coined the term \textquotedblleft principal
eigenstructures\textquotedblright\ to express these significant relations.

\section{\label{Section 21}Principal Eigenstructures}

Take the geometric locus of the novel principal eigenaxis $\boldsymbol{\kappa
=\kappa}_{1}{\large -}\boldsymbol{\kappa}_{2}$ of any given minimum risk
binary classification system $k_{\mathbf{s}}\boldsymbol{\kappa}+$
$\boldsymbol{\kappa}_{0}\overset{\omega_{1}}{\underset{\omega_{2}}{\gtrless}%
}0$ that has been determined by the machine learning algorithm being examined
in this treatise, so that the geometric locus of the novel principal eigenaxis
$\boldsymbol{\kappa=\kappa}_{1}{\large -}\boldsymbol{\kappa}_{2}$ represents
the discriminant function of the system, the exclusive principal
eigen-coordinate system of the geometric locus of the decision boundary of the
system, and an eigenaxis of symmetry that spans the decision space of the
system---at which point the discriminant function and the exclusive principal
eigen-coordinate system and the eigenaxis of symmetry are \emph{dual
components} that have different functions and properties.

Theorem \ref{Principal Eigenstructures Theorem} expresses how the mathematical
\emph{structure} and \emph{behavior} and \emph{properties} exhibited by
$\left(  1\right)  $ the dual locus $\boldsymbol{\kappa}=\boldsymbol{\kappa
}_{1}-\boldsymbol{\kappa}_{2}$ of the discriminant function of the system
$k_{\mathbf{s}}\boldsymbol{\kappa}+$ $\boldsymbol{\kappa}_{0}\overset{\omega
_{1}}{\underset{\omega_{2}}{\gtrless}}0$; $\left(  2\right)  $ the exclusive
principal eigen-coordinate system $\boldsymbol{\kappa}=\boldsymbol{\kappa}%
_{1}-\boldsymbol{\kappa}_{2}$ of the geometric locus of the decision boundary
$k_{\mathbf{s}}\boldsymbol{\kappa}+$ $\boldsymbol{\kappa}_{0}=0$ of the system
$k_{\mathbf{s}}\boldsymbol{\kappa}+$ $\boldsymbol{\kappa}_{0}\overset{\omega
_{1}}{\underset{\omega_{2}}{\gtrless}}0$; and $\left(  3\right)  $ the
eigenaxis of symmetry $\boldsymbol{\kappa}=\boldsymbol{\kappa}_{1}%
-\boldsymbol{\kappa}_{2}$ that spans the decision space $Z=Z_{1}\cup Z_{2}$ of
the system $k_{\mathbf{s}}\boldsymbol{\kappa}+$ $\boldsymbol{\kappa}%
_{0}\overset{\omega_{1}}{\underset{\omega_{2}}{\gtrless}}0$ are jointly
determined by elegant statistical relations and complex statistical
interconnections between the geometric loci of the primal novel principal
eigenaxis $\boldsymbol{\kappa}=\boldsymbol{\kappa}_{1}-\boldsymbol{\kappa}%
_{2}$ and the Wolfe-dual novel principal eigenaxis $\boldsymbol{\psi
}=\boldsymbol{\psi}_{1}+\boldsymbol{\psi}_{2}$ of the system $k_{\mathbf{s}%
}\boldsymbol{\kappa}+$ $\boldsymbol{\kappa}_{0}\overset{\omega_{1}%
}{\underset{\omega_{2}}{\gtrless}}0$, wherein the dual loci of
$\boldsymbol{\kappa}=\boldsymbol{\kappa}_{1}-\boldsymbol{\kappa}_{2}$ and
$\boldsymbol{\psi}=\boldsymbol{\psi}_{1}+\boldsymbol{\psi}_{2}$ are both
subject to deep-seated statistical interconnections between the elements and
the eigenvalues of a joint covariance matrix $\mathbf{Q}$ and the inverted
joint covariance matrix $\mathbf{Q}^{-1}$ associated with a pair of random
quadratic forms $\boldsymbol{\psi}^{T}\mathbf{Q}\boldsymbol{\psi}$ and
$\boldsymbol{\psi}^{T}\mathbf{Q}^{-1}\boldsymbol{\psi}$, so that the exclusive
principal eigen-coordinate system $\boldsymbol{\kappa}=\boldsymbol{\kappa}%
_{1}-\boldsymbol{\kappa}_{2}$ is the principal part of an equivalent
representation of the pair of random quadratic forms $\boldsymbol{\psi}%
^{T}\mathbf{Q}\boldsymbol{\psi}$ and $\boldsymbol{\psi}^{T}\mathbf{Q}%
^{-1}\boldsymbol{\psi}$.

Thereby, the geometric locus of the novel principal eigenaxis
$\boldsymbol{\kappa}=\boldsymbol{\kappa}_{1}-\boldsymbol{\kappa}_{2}$ is the
principal eigenaxis of the geometric locus of the decision boundary of the
minimum risk binary classification system $k_{\mathbf{s}}\boldsymbol{\kappa}+$
$\boldsymbol{\kappa}_{0}\overset{\omega_{1}}{\underset{\omega_{2}}{\gtrless}%
}0$, such that $\left(  1\right)  $ the novel principal eigenaxis
$\boldsymbol{\kappa}=\boldsymbol{\kappa}_{1}-\boldsymbol{\kappa}_{2}$
represents an eigenaxis of symmetry that contains all of the covariance and
distribution information for all of the extreme vectors $k_{\mathbf{x}%
_{1_{i\ast}}}$ and $k_{\mathbf{x}_{2_{i\ast}}}$---relative to the covariance
and distribution information for a given collection $\left\{  \mathbf{x}%
_{i}\right\}  _{i=1}^{N}$ of feature vectors $\mathbf{x}_{i}$; $\left(
2\right)  $ the geometric locus of the novel principal eigenaxis
$\boldsymbol{\kappa}=\boldsymbol{\kappa}_{1}-\boldsymbol{\kappa}_{2}$
satisfies the geometric locus of the decision boundary in terms of a critical
minimum eigenenergy $\left\Vert \boldsymbol{\kappa}\right\Vert _{\min_{c}}%
^{2}$ and a minimum expected risk $\mathfrak{R}_{\mathfrak{\min}}\left(
\left\Vert \boldsymbol{\kappa}\right\Vert _{\min_{c}}^{2}\right)  $; and
$\left(  3\right)  $ the uniform properties exhibited by all of the points
that lie on the geometric locus of the decision boundary are the critical
minimum eigenenergy $\left\Vert \boldsymbol{\kappa}_{1}-\boldsymbol{\kappa
}_{2}\right\Vert _{\min_{c}}^{2}$ and the minimum expected risk $\mathfrak{R}%
_{\mathfrak{\min}}\left(  \left\Vert \boldsymbol{\kappa}_{1}%
-\boldsymbol{\kappa}_{2}\right\Vert _{\min_{c}}^{2}\right)  $ exhibited by the
geometric locus of the novel principal eigenaxis $\boldsymbol{\kappa
}=\boldsymbol{\kappa}_{1}-\boldsymbol{\kappa}_{2}$.

It follows that the shapes and the fundamental properties exhibited by the
geometric loci of the decision boundary and the pair of symmetrically
positioned decision borders of any given minimum risk binary classification
system are completely determined by the geometric locus of the novel principal
eigenaxis $\boldsymbol{\kappa}=\boldsymbol{\kappa}_{1}-\boldsymbol{\kappa}%
_{2}$ of the system, since the geometric locus of the novel principal
eigenaxis $\boldsymbol{\kappa}=\boldsymbol{\kappa}_{1}-\boldsymbol{\kappa}%
_{2}$ is the principal eigenaxis of the geometric loci of the decision
boundary and the pair of symmetrically positioned decision borders of the
decision regions $Z_{1}$ and $Z_{2}$ of the system, at which point the novel
principal eigenaxis $\boldsymbol{\kappa}=\boldsymbol{\kappa}_{1}%
-\boldsymbol{\kappa}_{2}$ represents an eigenaxis of symmetry that satisfies
the locus of the decision boundary in terms of a critical minimum eigenenergy
$\left\Vert \boldsymbol{\kappa}_{1}-\boldsymbol{\kappa}_{2}\right\Vert
_{\min_{c}}^{2}$ and a minimum expected risk $\mathfrak{R}_{\mathfrak{\min}%
}\left(  \left\Vert \boldsymbol{\kappa}_{1}-\boldsymbol{\kappa}_{2}\right\Vert
_{\min_{c}}^{2}\right)  $.

Theorem \ref{Principal Eigenstructures Theorem} is substantiated by the
guarantees expressed by the novel principal eigen-coordinate transform method
of Theorem \ref{Principal Eigen-coordinate System Theorem} and Corollary
\ref{Principal Eigen-coordinate System Corollary}, wherein an exclusive
principal eigen-coordinate system is the principal part of an equivalent
representation of a certain quadratic form that is the solution of a vector
algebra locus equation, so that the exclusive principal eigen-coordinate
system is the principal eigenaxis of the geometric locus of a certain
quadratic curve or surface, at which point the principal eigenaxis satisfies
the geometric locus of the quadratic curve or surface in terms of its total
allowed eigenenergy---which is regulated by the eigenvalues of the symmetric
matrix of the quadratic form.

\begin{theorem}
\label{Principal Eigenstructures Theorem}Take any given $N\times N$ joint
covariance matrix $\mathbf{Q}$ of a random quadratic form $\boldsymbol{\psi
}^{T}\mathbf{Q}\boldsymbol{\psi}$ in the Wolfe-dual eigenenergy functional
$\max\Xi_{\boldsymbol{\psi}}\left(  \boldsymbol{\psi}\right)  =\mathbf{1}%
^{T}\boldsymbol{\psi}-\boldsymbol{\psi}^{T}\mathbf{Q}\boldsymbol{\psi/}2$ of a
minimum risk binary classification system%
\[
\left(  k_{\mathbf{s}}\boldsymbol{-}\frac{1}{l}\sum\nolimits_{i=1}%
^{l}k_{\mathbf{x}_{i\ast}}\right)  \left(  \boldsymbol{\kappa}_{1}%
-\boldsymbol{\kappa}_{2}\right)  +\frac{1}{l}\sum\nolimits_{i=1}^{l}%
y_{i}\left(  1-\xi_{i}\right)  \overset{\omega_{1}}{\underset{\omega
_{2}}{\gtrless}}0
\]
that is subject to random inputs $\mathbf{x\in}$ $%
\mathbb{R}
^{d}$ such that $\mathbf{x\sim}$ $p\left(  \mathbf{x};\omega_{1}\right)  $ and
$\mathbf{x\sim}$ $p\left(  \mathbf{x};\omega_{2}\right)  $, where either
$\xi_{i}=\xi=0$ or $\xi_{i}=\xi\ll1$, $y_{i}=\pm1$, and $p\left(
\mathbf{x};\omega_{1}\right)  $ and $p\left(  \mathbf{x};\omega_{2}\right)  $
are certain probability density functions for two classes $\omega_{1}$ and
$\omega_{2}$ of random vectors $\mathbf{x\in}$ $%
\mathbb{R}
^{d}$, such that the eigenenergy functional%
\[
\max\Xi_{\boldsymbol{\psi}}\left(  \boldsymbol{\psi}\right)  =\mathbf{1}%
^{T}\boldsymbol{\psi}-\boldsymbol{\psi}^{T}\mathbf{Q}\boldsymbol{\psi/}2
\]
is subject to the constraints $\sum\nolimits_{i=1}^{N}y_{i}\psi_{i}=0$ and
$\psi_{i\ast}>0$, where $y_{i}=\left\{  \pm1\right\}  $, so that the geometric
locus of a Wolfe-dual novel principal eigenaxis $\boldsymbol{\psi
}=\boldsymbol{\psi}_{1}+\boldsymbol{\psi}_{2}$ is symmetrically and
equivalently related to the principal eigenvector $\boldsymbol{\psi}_{\max}$
of the joint covariance matrix $\mathbf{Q}$ and the inverted joint covariance
matrix $\mathbf{Q}^{-1}$ associated with a pair of random quadratic forms
$\boldsymbol{\psi}^{T}\mathbf{Q}\boldsymbol{\psi}$ and $\boldsymbol{\psi}%
^{T}\mathbf{Q}^{-1}\boldsymbol{\psi}$.

Let the statistical structure and functionality exhibited by the geometric
locus of the novel principal eigenaxis $\boldsymbol{\kappa}=\boldsymbol{\kappa
}_{1}-\boldsymbol{\kappa}_{2}$ of the minimum risk binary classification
system $k_{\mathbf{s}}\boldsymbol{\kappa}+$ $\boldsymbol{\kappa}%
_{0}\overset{\omega_{1}}{\underset{\omega_{2}}{\gtrless}}0$ be determined by
the following statistical relations and deep-seated statistical
interconnections between certain intrinsic components of the system, so that
the discriminant function%
\[
d\left(  \mathbf{s}\right)  =\left(  k_{\mathbf{s}}\boldsymbol{-}\frac{1}%
{l}\sum\nolimits_{i=1}^{l}k_{\mathbf{x}_{i\ast}}\right)  \left(
\boldsymbol{\kappa}_{1}-\boldsymbol{\kappa}_{2}\right)  +\frac{1}{l}%
\sum\nolimits_{i=1}^{l}y_{i}\left(  1-\xi_{i}\right)
\]
of the system generalizes and thereby extrapolates in a significant manner, at
which point the structure and behavior and properties exhibited by the
geometric locus of the Wolfe-dual novel principal eigenaxis%
\[
\boldsymbol{\psi}=\sum\nolimits_{i=1}^{l_{1}}\psi_{1i\ast}\frac{k_{\mathbf{x}%
_{1i\ast}}}{\left\Vert k_{\mathbf{x}_{1i\ast}}\right\Vert }+\sum
\nolimits_{i=1}^{l_{2}}\psi_{2i\ast}\frac{k_{\mathbf{x}_{2i\ast}}}{\left\Vert
k_{\mathbf{x}_{2i\ast}}\right\Vert }%
\]
are symmetrically and equivalently related to the structure and behavior and
properties exhibited by the geometric locus of the primal novel principal
eigenaxis%
\[
\boldsymbol{\kappa}=\sum\nolimits_{i=1}^{l_{1}}\psi_{1_{i\ast}}k_{\mathbf{x}%
_{1_{i\ast}}}-\sum\nolimits_{i=1}^{l_{2}}\psi_{2_{i\ast}}k_{\mathbf{x}%
_{2_{i\ast}}}%
\]
of the minimum risk binary classification system.

The eigenvalues $\lambda_{N}\leq\mathbf{\ldots}\leq\lambda_{1}$ of the joint
covariance matrix $\mathbf{Q}$%
\[
\det\left(
\begin{bmatrix}
\left\Vert k_{\mathbf{x}_{1}}\right\Vert \left\Vert k_{\mathbf{x}_{1}%
}\right\Vert \cos\theta_{k_{\mathbf{x_{1}}}k_{\mathbf{x}_{1}}}-\lambda_{1} &
\cdots & -\left\Vert k_{\mathbf{x}_{1}}\right\Vert \left\Vert k_{\mathbf{x}%
_{N}}\right\Vert \cos\theta_{k_{\mathbf{x}_{1}}k_{\mathbf{x}_{N}}}\\
\vdots & \ddots & \vdots\\
-\left\Vert k_{\mathbf{x}_{N}}\right\Vert \left\Vert k_{\mathbf{x}_{1}%
}\right\Vert \cos\theta_{k_{\mathbf{x}_{N}}k_{\mathbf{x}_{1}}} & \cdots &
\left\Vert k_{\mathbf{x}_{N}}\right\Vert \left\Vert k_{\mathbf{x}_{N}%
}\right\Vert \cos\theta_{k_{\mathbf{x}_{N}}k_{\mathbf{x}_{N}}}-\lambda_{N}%
\end{bmatrix}
\right)  =0
\]
vary continuously with the elements $\left\Vert k_{\mathbf{x}_{i}}\right\Vert
\left\Vert k_{\mathbf{x}_{j}}\right\Vert \cos\theta_{k_{\mathbf{x_{i}}%
}k_{\mathbf{x}_{j}}}$ of $\mathbf{Q}$ since the roots $p\left(  \lambda
\right)  =0$ of the characteristic polynomial $p\left(  \lambda\right)  $ of
$\mathbf{Q}$ vary continuously with its coefficients.

Thereby, the eigenvalues $\lambda_{N}\leq\mathbf{\ldots}\leq\lambda_{1}$ of
the joint covariance matrix $\mathbf{Q}$ represent joint variabilities between
all of the feature vectors $k_{\mathbf{x}_{i}}$ and $k_{\mathbf{x}_{j}}$ used
to construct $\mathbf{Q}$, so that each element $y_{i}\left\Vert
k_{\mathbf{x}_{i}}\right\Vert y_{j}\left\Vert k_{\mathbf{x}_{j}}\right\Vert
\cos\theta_{k_{\mathbf{x}_{i}}k_{\mathbf{x}_{j}}}$ of $\mathbf{Q}$ where
$y_{i}y_{j}=-1$ describes differences between joint variabilities of feature
vectors $k_{\mathbf{x}_{i}}$ and $k_{\mathbf{x}_{j}}$ that belong to different
pattern classes, at which point each element $\left\Vert k_{\mathbf{x}_{i}%
}\right\Vert \left\Vert k_{\mathbf{x}_{j}}\right\Vert \cos\theta
_{k_{\mathbf{x}_{i}}k_{\mathbf{x}_{j}}}$ of the joint covariance matrix
$\mathbf{Q}$ is correlated with the distance $\left\Vert k_{\mathbf{x}_{i}%
}-k_{\mathbf{x}_{j}}\right\Vert $ between the loci of certain feature vectors
$k_{\mathbf{x}_{i}}$ and $k_{\mathbf{x}_{j}}$.

Now let the geometric locus of the Wolfe-dual novel principal eigenaxis
$\boldsymbol{\psi}$ be subject to a critical minimum eigenenergy constraint%
\[
\lambda_{1}\left\Vert \boldsymbol{\psi}\right\Vert _{\min_{c}}^{2}%
=\boldsymbol{\psi}_{\max}^{T}\mathbf{Q}\boldsymbol{\psi}_{\max}\equiv
\left\Vert \boldsymbol{\kappa}\right\Vert _{\min_{c}}^{2}%
\]
that is symmetrically and equivalently related to the critical minimum
eigenenergy constraint $\left\Vert \boldsymbol{\kappa}\right\Vert _{\min_{c}%
}^{2}$ on the geometric locus of the primal novel principal eigenaxis
$\boldsymbol{\kappa}$, so that the random quadratic form $\boldsymbol{\psi
}_{\max}^{T}\mathbf{Q}\boldsymbol{\psi}_{\max}$ is symmetrically and
equivalently related to the critical minimum eigenenergy $\left\Vert
\boldsymbol{\kappa}\right\Vert _{\min_{c}}^{2}$ exhibited by the geometric
locus of the primal novel principal eigenaxis $\boldsymbol{\kappa}$, at which
point the random quadratic form $\boldsymbol{\psi}_{\max}^{T}\mathbf{Q}%
\boldsymbol{\psi}_{\max}$, plus the total allowed eigenenergy $\left\Vert
\boldsymbol{\kappa}\right\Vert _{\min_{c}}^{2}$ and the expected risk
$\mathfrak{R}_{\mathfrak{\min}}\left(  \left\Vert \boldsymbol{\kappa
}\right\Vert _{\min_{c}}^{2}\right)  $ exhibited by the geometric locus of the
primal novel eigenaxis $\boldsymbol{\kappa}$ jointly reach their minimum values.

Correspondingly, let the Wolfe-dual eigenenergy functional%
\[
\max\Xi\left(  \boldsymbol{\psi}\right)  =\mathbf{1}^{T}\boldsymbol{\psi
}-\boldsymbol{\psi}^{T}\mathbf{Q}\boldsymbol{\psi/}2\text{,}%
\]
such that $\boldsymbol{\psi}^{T}\mathbf{y}=0$ and $\psi_{i\ast}>0$, be
maximized by the largest eigenvector $\boldsymbol{\psi}_{\max}$ of the joint
covariance matrix $\mathbf{Q}$ of the random quadratic form $\boldsymbol{\psi
}^{T}\boldsymbol{Q\psi}$%
\[
\boldsymbol{Q\psi}_{\max}=\lambda_{1}\boldsymbol{\psi}_{\max}\text{,}%
\]
so that the random quadratic form $\boldsymbol{\psi}^{T}\mathbf{Q}%
\boldsymbol{\psi}$ reaches its minimum value, at which point the total allowed
eigenenergy $\left\Vert \boldsymbol{\kappa}\right\Vert _{\min_{c}}^{2}$ and
the expected risk $\mathfrak{R}_{\mathfrak{\min}}\left(  \left\Vert
\boldsymbol{\kappa}\right\Vert _{\min_{c}}^{2}\right)  $ exhibited by the
geometric locus of the primal novel principal eigenaxis $\boldsymbol{\kappa}$
are jointly minimized.

Next, let the Wolfe-dual novel principal eigenaxis $\boldsymbol{\psi}$ and the
primal novel principal eigenaxis $\boldsymbol{\kappa}$ be solutions of the
vector algebra locus equation%
\begin{align*}
\boldsymbol{\psi}  &  =\lambda_{1}^{-1}\boldsymbol{\psi}^{T}\mathbf{Q}\\
&  =\lambda_{1}\boldsymbol{\psi}_{\max}^{T}\mathbf{Q}\text{,}%
\end{align*}
such that the geometric locus of the Wolfe-dual novel principal eigenaxis
$\boldsymbol{\psi}$ is related to the scaled principal eigenvector
$\lambda_{1}^{-1}\boldsymbol{\psi}_{\max}$ of the joint covariance matrix
$\mathbf{Q}$ of the random quadratic form $\boldsymbol{\psi}^{T}%
\mathbf{Q}\boldsymbol{\psi}$ acting on the joint covariance matrix
$\mathbf{Q}$, so that the sides $\boldsymbol{\kappa}_{1}$ and
$\boldsymbol{\kappa}_{2}$ of the geometric locus of the novel principal
eigenaxis $\boldsymbol{\kappa}=\boldsymbol{\kappa}_{1}-\boldsymbol{\kappa}%
_{2}$ and the scale factors $\psi_{1i\ast}$ and $\psi_{2i\ast}$ for the
components $\psi_{1i\ast}\frac{k_{\mathbf{x}_{1i\ast}}}{\left\Vert
k_{\mathbf{x}_{1i\ast}}\right\Vert }$ and $\psi_{2i\ast}\frac{k_{\mathbf{x}%
_{2i\ast}}}{\left\Vert k_{\mathbf{x}_{2i\ast}}\right\Vert }$ of the principal
eigenvector $\boldsymbol{\psi}_{\max}$ of the joint covariance matrix
$\mathbf{Q}$ of the random quadratic form $\boldsymbol{\psi}^{T}%
\mathbf{Q}\boldsymbol{\psi}$ are solutions of the system of vector algebra
locus equations%
\begin{align*}
\sum\nolimits_{i=1}^{l_{1}}\psi_{1i\ast}  &  =\lambda_{1}^{-1}\sum
\nolimits_{i=1}^{l_{1}}k_{\mathbf{x}_{1_{i\ast}}}\left(  \boldsymbol{\kappa
}_{1}-\boldsymbol{\kappa}_{2}\right) \\
&  =\lambda_{1}^{-1}\sum\nolimits_{i=1}^{l_{1}}k_{\mathbf{x}_{1_{i\ast}}%
}\left(  \sum\nolimits_{j=1}^{l_{1}}\psi_{1_{j\ast}}k_{\mathbf{x}_{1_{j\ast}}%
}-\sum\nolimits_{j=1}^{l_{2}}\psi_{2_{j\ast}}k_{\mathbf{x}_{2_{j\ast}}%
}\right)  \text{,}%
\end{align*}
and%
\begin{align*}
\sum\nolimits_{i=1}^{l_{2}}\psi_{2i\ast}  &  =\lambda_{1}^{-1}\sum
\nolimits_{i=1}^{l_{2}}k_{\mathbf{x}_{2_{i\ast}}}\left(  \boldsymbol{\kappa
}_{2}-\boldsymbol{\kappa}_{1}\right) \\
&  =\lambda_{1}^{-1}\sum\nolimits_{i=1}^{l_{2}}k_{\mathbf{x}_{2_{i\ast}}%
}\left(  \sum\nolimits_{j=1}^{l_{2}}\psi_{2_{j\ast}}k_{\mathbf{x}_{2_{j\ast}}%
}-\sum\nolimits_{j=1}^{l_{1}}\psi_{1_{j\ast}}k_{\mathbf{x}_{1_{j\ast}}%
}\right)  \text{,}%
\end{align*}
at which point each and every one of the components $\psi_{1i\ast}%
\frac{k_{\mathbf{x}_{1i\ast}}}{\left\Vert k_{\mathbf{x}_{1i\ast}}\right\Vert
}$ and $\psi_{2i\ast}\frac{k_{\mathbf{x}_{2i\ast}}}{\left\Vert k_{\mathbf{x}%
_{2i\ast}}\right\Vert }$ of the principal eigenvector $\boldsymbol{\psi}%
_{\max}$ of the joint covariance matrix $\mathbf{Q}$ of the random quadratic
form $\boldsymbol{\psi}^{T}\boldsymbol{Q\psi}$, along with each and every one
of the components $\psi_{1_{i\ast}}k_{\mathbf{x}_{1_{i\ast}}}$and
$\psi_{2_{i\ast}}k_{\mathbf{x}_{2_{i\ast}}}$of the primal novel principal
eigenaxis $\boldsymbol{\kappa}=\boldsymbol{\kappa}_{1}-\boldsymbol{\kappa}%
_{2}$, are subject to deep-seated statistical interconnections with each
other, so that likely locations and likelihood values of extreme vectors
$k_{\mathbf{x}_{1_{i\ast}}}$ and $k_{\mathbf{x}_{2_{i\ast}}}$ are
statistically pre-wired\ within the components $\psi_{1i\ast}\frac
{k_{\mathbf{x}_{1i\ast}}}{\left\Vert k_{\mathbf{x}_{1i\ast}}\right\Vert }$ and
$\psi_{2i\ast}\frac{k_{\mathbf{x}_{2i\ast}}}{\left\Vert k_{\mathbf{x}_{2i\ast
}}\right\Vert }$ of the principal eigenvector $\boldsymbol{\psi}_{\max}$ of
the joint covariance matrix $\mathbf{Q}$ and the inverted joint covariance
matrix $\mathbf{Q}^{-1}$ associated with the pair of random quadratic forms
$\boldsymbol{\psi}^{T}\mathbf{Q}\boldsymbol{\psi}$ and $\boldsymbol{\psi}%
^{T}\mathbf{Q}^{-1}\boldsymbol{\psi}$, as well as the components
$\psi_{1_{i_{\ast}}}k_{\mathbf{x}_{1_{i\ast}}}$ and $\psi_{2_{i_{\ast}}%
}k_{\mathbf{x}_{2_{i\ast}}}$ of the geometric locus of the novel principal
eigenaxis $\boldsymbol{\kappa}=\boldsymbol{\kappa}_{1}-\boldsymbol{\kappa}%
_{2}$ of the minimum risk binary classification system $k_{\mathbf{s}%
}\boldsymbol{\kappa}+$ $\boldsymbol{\kappa}_{0}\overset{\omega_{1}%
}{\underset{\omega_{2}}{\gtrless}}0$, such that each and every one of the
statistical interconnections is regulated by the eigenvalues $\lambda_{N}%
\leq\mathbf{\ldots}\leq\lambda_{1}$ and $\lambda_{N}^{-1}\leq\mathbf{\ldots
}\leq\lambda_{1}^{-1}$ of the joint covariance matrix $\mathbf{Q}$ and the
inverted joint covariance matrix $\mathbf{Q}^{-1}$ associated with the pair of
random quadratic forms $\boldsymbol{\psi}^{T}\mathbf{Q}\boldsymbol{\psi}$ and
$\boldsymbol{\psi}^{T}\mathbf{Q}^{-1}\boldsymbol{\psi}$.

Furthermore, let the geometric locus of the novel principal eigenaxis
$\boldsymbol{\kappa}=\boldsymbol{\kappa}_{1}-\boldsymbol{\kappa}_{2}$ be the
solution of the vector algebra locus equation that represents the geometric
locus of the decision boundary of the minimum risk binary classification
system%
\[
\left(  k_{\mathbf{s}}\boldsymbol{-}\frac{1}{l}\sum\nolimits_{i=1}%
^{l}k_{\mathbf{x}_{i\ast}}\right)  \boldsymbol{\kappa}+\frac{1}{l}%
\sum\nolimits_{i=1}^{l}y_{i}\left(  1-\xi_{i}\right)  =0\text{,}%
\]
along with the vector algebra locus equation that represents the geometric
locus of the decision border of the decision region $Z_{1}$ of the system%
\[
\left(  k_{\mathbf{s}}\boldsymbol{-}\frac{1}{l}\sum\nolimits_{i=1}%
^{l}k_{\mathbf{x}_{i\ast}}\right)  \left(  \boldsymbol{\kappa}_{1}%
-\boldsymbol{\kappa}_{2}\right)  +\frac{1}{l}\sum\nolimits_{i=1}^{l}%
y_{i}\left(  1-\xi_{i}\right)  =+1\text{,}%
\]
and the vector algebra locus equation that represents the geometric locus of
the decision border of the decision region $Z_{2}$ of the system%
\[
\left(  k_{\mathbf{s}}\boldsymbol{-}\frac{1}{l}\sum\nolimits_{i=1}%
^{l}k_{\mathbf{x}_{i\ast}}\right)  \left(  \boldsymbol{\kappa}_{1}%
-\boldsymbol{\kappa}_{2}\right)  +\frac{1}{l}\sum\nolimits_{i=1}^{l}%
y_{i}\left(  1-\xi_{i}\right)  =-1\text{,}%
\]
so that each and every one of the points $\mathbf{s}$ that lies on the
geometric loci of the decision boundary and the pair of symmetrically
positioned decision borders of the decision regions $Z_{1}$ and $Z_{2}$ of the
minimum risk binary classification system exclusively reference the geometric
locus of the novel principal eigenaxis $\boldsymbol{\kappa}=\boldsymbol{\kappa
}_{1}-\boldsymbol{\kappa}_{2}$, at which point the geometric locus of the
novel principal eigenaxis $\boldsymbol{\kappa}=\boldsymbol{\kappa}%
_{1}-\boldsymbol{\kappa}_{2}$ represents an eigenaxis of symmetry that spans
the decision space $Z=Z_{1}\cup Z_{2}$ of the system.

It follows that the geometric locus of the novel principal eigenaxis
$\boldsymbol{\kappa}=\boldsymbol{\kappa}_{1}-\boldsymbol{\kappa}%
_{2}\boldsymbol{\ }$is an exclusive principal eigen-coordinate system that
contains all of the covariance and distribution information---for all of the
extreme vectors $k_{\mathbf{x}_{1_{i\ast}}}$ and $k_{\mathbf{x}_{2_{i\ast}}}%
$---relative to the covariance and distribution information for a given
collection $\left\{  \mathbf{x}_{i}\right\}  _{i=1}^{N}$ of feature vectors
$\mathbf{x}_{i}$, so that each principal eigenaxis component $\psi
_{1_{i_{\ast}}}k_{\mathbf{x}_{1_{i\ast}}}$ and $\psi_{2_{i_{\ast}}%
}k_{\mathbf{x}_{2_{i\ast}}}$ that lies on the novel principal eigenaxis
$\boldsymbol{\kappa}=\boldsymbol{\kappa}_{1}{\large -}\boldsymbol{\kappa}_{2}$
determines a likely location for a correlated extreme point $\mathbf{x}%
_{1_{i\ast}}\mathbf{\sim}$ $p\left(  \mathbf{x};\omega_{1}\right)  $ and
$\mathbf{x}_{2_{i\ast}}\mathbf{\sim}$ $p\left(  \mathbf{x};\omega_{2}\right)
$, and each likelihood component $\psi_{1_{i_{\ast}}}k_{\mathbf{x}_{1_{i\ast}%
}}$ and $\psi_{2_{i_{\ast}}}k_{\mathbf{x}_{2_{i\ast}}}$ that lies on the novel
principal eigenaxis $\boldsymbol{\kappa}=\boldsymbol{\kappa}_{1}%
{\large -}\boldsymbol{\kappa}_{2}$ determines a likelihood value for the
correlated extreme point $\mathbf{x}_{1_{i\ast}}$ and $\mathbf{x}_{2_{i\ast}}%
$, where the reproducing kernel for each extreme point $k_{\mathbf{x}%
_{1_{i\ast}}}$ and $k_{\mathbf{x}_{2_{i\ast}}}$ has the preferred form of
either $k_{\mathbf{x}}\left(  \mathbf{s}\right)  =\left(  \mathbf{s}%
^{T}\mathbf{x}+1\right)  ^{2}$ or $k_{\mathbf{x}}\left(  \mathbf{s}\right)
=\exp\left(  -\gamma\left\Vert \mathbf{s}-\mathbf{x}\right\Vert ^{2}\right)
$, wherein $0.01\leq\gamma\leq0.1$.

In addition, let the Wolfe-dual novel principal eigenaxis $\boldsymbol{\psi
}=\boldsymbol{\psi}_{1}+\boldsymbol{\psi}_{2}$ and the primal novel principal
eigenaxis $\boldsymbol{\kappa}=\boldsymbol{\kappa}_{1}-\boldsymbol{\kappa}%
_{2}$ be solutions of the vector algebra locus equation%
\[
\left(  \boldsymbol{\kappa}_{1}-\boldsymbol{\kappa}_{2}\right)
\boldsymbol{\kappa}=\boldsymbol{\psi}_{1}+\boldsymbol{\psi}_{2}-\left(
\sum\nolimits_{i=1}^{l_{1}}\xi_{i}\psi_{1i\ast}+\sum\nolimits_{i=1}^{l_{2}}%
\xi_{i}\psi_{2_{i_{\ast}}}\right)  \text{,}%
\]
so that the geometric locus of the novel principal eigenaxis
$\boldsymbol{\kappa}=\boldsymbol{\kappa}_{1}-\boldsymbol{\kappa}_{2}$
satisfies the geometric locus of the decision boundary of the minimum risk
binary classification system $k_{\mathbf{s}}\boldsymbol{\kappa}+$
$\boldsymbol{\kappa}_{0}\overset{\omega_{1}}{\underset{\omega_{2}}{\gtrless}%
}0$ in terms of a critical minimum eigenenergy $\left\Vert \boldsymbol{\kappa
}_{1}-\boldsymbol{\kappa}_{2}\right\Vert _{\min_{c}}^{2}$ and a minimum
expected risk $\mathfrak{R}_{\mathfrak{\min}}\left(  \left\Vert
\boldsymbol{\kappa}_{1}-\boldsymbol{\kappa}_{2}\right\Vert _{\min_{c}}%
^{2}\right)  $, at which point the total allowed eigenenergy $\left\Vert
\boldsymbol{\kappa}\right\Vert _{\min_{c}}^{2}$ and the expected risk
$\mathfrak{R}_{\mathfrak{\min}}\left(  \left\Vert \boldsymbol{\kappa
}\right\Vert _{\min_{c}}^{2}\right)  $ exhibited by the minimum risk binary
classification system are jointly regulated by the equilibrium requirement on
the dual locus $\boldsymbol{\kappa}=\boldsymbol{\kappa}_{1}-\boldsymbol{\kappa
}_{2}$ of the discriminant function at the geometric locus of decision
boundary of the system, so that the geometric locus of the novel principal
eigenaxis $\boldsymbol{\kappa}=\boldsymbol{\kappa}_{1}-\boldsymbol{\kappa}%
_{2}$ satisfies the law of cosines in the symmetrically balanced manner
\begin{align*}
d\left(  \mathbf{s}\right)   &  :\left\Vert \boldsymbol{\kappa}_{1}\right\Vert
_{\min_{c}}^{2}-\left\Vert \boldsymbol{\kappa}_{1}\right\Vert \left\Vert
\boldsymbol{\kappa}_{2}\right\Vert \cos\theta_{\boldsymbol{\kappa}%
_{1}\boldsymbol{\kappa}_{2}}\\
&  =\left\Vert \boldsymbol{\kappa}_{2}\right\Vert _{\min_{c}}^{2}-\left\Vert
\boldsymbol{\kappa}_{2}\right\Vert \left\Vert \boldsymbol{\kappa}%
_{1}\right\Vert \cos\theta_{\boldsymbol{\kappa}_{2}\boldsymbol{\kappa}_{1}}\\
&  =\frac{1}{2}\left\Vert \boldsymbol{\kappa}_{1}-\boldsymbol{\kappa}%
_{2}\right\Vert _{\min_{c}}^{2}\text{.}%
\end{align*}

Thereby, the minimum risk binary classification system $k_{\mathbf{s}%
}\boldsymbol{\kappa}+$ $\boldsymbol{\kappa}_{0}\overset{\omega_{1}%
}{\underset{\omega_{2}}{\gtrless}}0$ satisfies a state of statistical
equilibrium, so that the total allowed eigenenergy $\left\Vert
\boldsymbol{\kappa}\right\Vert _{\min_{c}}^{2}$ and the expected risk
$\mathfrak{R}_{\mathfrak{\min}}\left(  \left\Vert \boldsymbol{\kappa
}\right\Vert _{\min_{c}}^{2}\right)  $ exhibited by the system are jointly
minimized within the decision space $Z=Z_{1}\cup Z_{2}$ of the system, at
which point the system exhibits the minimum probability of classification error.

It follows that the components of the dual loci of the novel principal
eigenaxes $\boldsymbol{\psi}=\boldsymbol{\psi}_{1}+\boldsymbol{\psi}_{2}$ and
$\boldsymbol{\kappa}=\boldsymbol{\kappa}_{1}-\boldsymbol{\kappa}_{2}$ are both
subject to deep-seated statistical interconnections between the elements
$\left\Vert k_{\mathbf{x}_{i}}\right\Vert \left\Vert k_{\mathbf{x}_{j}%
}\right\Vert \cos\theta_{k_{\mathbf{x_{i}}}k_{\mathbf{x}_{j}}}$ and the
eigenvalues $\lambda_{N}\leq\mathbf{\ldots}\leq\lambda_{1}$ of the joint
covariance matrix $\mathbf{Q}$ of the random quadratic form $\boldsymbol{\psi
}^{T}\mathbf{Q}\boldsymbol{\psi}$, along with the elements and the eigenvalues
$\lambda_{N}^{-1}\leq\mathbf{\ldots}\leq\lambda_{1}^{-1}$ of the inverted
joint covariance matrix $\mathbf{Q}^{-1}$ of the random quadratic form
$\boldsymbol{\psi}^{T}\mathbf{Q}^{-1}\boldsymbol{\psi}$, so that the geometric
locus of the novel principal eigenaxis $\boldsymbol{\kappa}=\boldsymbol{\kappa
}_{1}-\boldsymbol{\kappa}_{2}$ is the principal part of an equivalent
representation of the pair of random quadratic forms $\boldsymbol{\psi}%
^{T}\mathbf{Q}\boldsymbol{\psi}$ and $\boldsymbol{\psi}^{T}\mathbf{Q}%
^{-1}\boldsymbol{\psi}$.

Thereby, the components $\psi_{1_{i\ast}}k_{\mathbf{x}_{1_{i\ast}}}$and
$\psi_{2_{i\ast}}k_{\mathbf{x}_{2_{i\ast}}}$ of the geometric locus of the
novel principal eigenaxis $\boldsymbol{\kappa}=\boldsymbol{\kappa}%
_{1}-\boldsymbol{\kappa}_{2}$ are statistically interconnected with the
components $\psi_{1i\ast}\frac{k_{\mathbf{x}_{1i\ast}}}{\left\Vert
k_{\mathbf{x}_{1i\ast}}\right\Vert }$ and $\psi_{2i\ast}\frac{k_{\mathbf{x}%
_{2i\ast}}}{\left\Vert k_{\mathbf{x}_{2i\ast}}\right\Vert }$ of the principal
eigenvector $\boldsymbol{\psi}_{\max}$ of the joint covariance matrix
$\mathbf{Q}$ and the inverted joint covariance matrix $\mathbf{Q}^{-1}$, along
with the elements and the eigenvalues of the joint covariance matrix
$\mathbf{Q}$ and the inverted joint covariance matrix $\mathbf{Q}^{-1}$, so
that the geometric locus of the novel principal eigenaxis $\boldsymbol{\kappa
}=\boldsymbol{\kappa}_{1}-\boldsymbol{\kappa}_{2}$ is the exclusive principal
eigen-coordinate system of the geometric locus of the decision boundary%
\[
\left(  k_{\mathbf{s}}\boldsymbol{-}\frac{1}{l}\sum\nolimits_{i=1}%
^{l}k_{\mathbf{x}_{i\ast}}\right)  \left(  \boldsymbol{\kappa}_{1}%
-\boldsymbol{\kappa}_{2}\right)  +\frac{1}{l}\sum\nolimits_{i=1}^{l}%
y_{i}\left(  1-\xi_{i}\right)  =0
\]
of the minimum risk binary classification system%
\[
\left(  k_{\mathbf{s}}\boldsymbol{-}\frac{1}{l}\sum\nolimits_{i=1}%
^{l}k_{\mathbf{x}_{i\ast}}\right)  \left(  \boldsymbol{\kappa}_{1}%
-\boldsymbol{\kappa}_{2}\right)  +\frac{1}{l}\sum\nolimits_{i=1}^{l}%
y_{i}\left(  1-\xi_{i}\right)  \overset{\omega_{1}}{\underset{\omega
_{2}}{\gtrless}}0\text{,}%
\]
so that the geometric locus of the novel principal eigenaxis
$\boldsymbol{\kappa}=\boldsymbol{\kappa}_{1}-\boldsymbol{\kappa}_{2}$
satisfies the geometric locus of the decision boundary in terms of a critical
minimum eigenenergy $\left\Vert \boldsymbol{\kappa}_{1}-\boldsymbol{\kappa
}_{2}\right\Vert _{\min_{c}}^{2}$ and a minimum expected risk $\mathfrak{R}%
_{\mathfrak{\min}}\left(  \left\Vert \boldsymbol{\kappa}\right\Vert _{\min
_{c}}^{2}\right)  $ in the following manner%
\begin{align*}
\left\Vert \boldsymbol{\kappa}_{1}-\boldsymbol{\kappa}_{2}\right\Vert
_{\min_{c}}^{2}  &  =\sum\nolimits_{i=1}^{l_{1}}\psi_{1i\ast}\left(  1-\xi
_{i}\right)  +\sum\nolimits_{i=1}^{l_{2}}\psi_{2i\ast}\left(  1-\xi_{i}\right)
\\
&  =\boldsymbol{\psi}_{\max}-\left(  \sum\nolimits_{i=1}^{l_{1}}\xi_{i}%
\psi_{1i\ast}+\sum\nolimits_{i=1}^{l_{2}}\xi_{i}\psi_{2i\ast}\right)  \text{,}%
\end{align*}
at which point the total allowed eigenenergy $\left\Vert \boldsymbol{\kappa
}_{1}-\boldsymbol{\kappa}_{2}\right\Vert _{\min_{c}}^{2}$ and the expected
risk $\mathfrak{R}_{\mathfrak{\min}}\left(  \left\Vert \boldsymbol{\kappa}%
_{1}-\boldsymbol{\kappa}_{2}\right\Vert _{\min_{c}}^{2}\right)  $ exhibited by
the geometric locus of the novel principal eigenaxis $\boldsymbol{\kappa
}=\boldsymbol{\kappa}_{1}-\boldsymbol{\kappa}_{2}$ are both regulated by
values of the scale factors $\psi_{1i\ast}$ and $\psi_{2i\ast}$ for the
components $\psi_{1i\ast}\frac{k_{\mathbf{x}_{1i\ast}}}{\left\Vert
k_{\mathbf{x}_{1i\ast}}\right\Vert }$ and $\psi_{2i\ast}\frac{k_{\mathbf{x}%
_{2i\ast}}}{\left\Vert k_{\mathbf{x}_{2i\ast}}\right\Vert }$ of the principal
eigenvector $\boldsymbol{\psi}_{\max}$ of the joint covariance matrix
$\mathbf{Q}$ of the random quadratic form $\boldsymbol{\psi}^{T}%
\mathbf{Q}\boldsymbol{\psi}$ and the inverted joint covariance matrix
$\mathbf{Q}^{-1}$ of the random quadratic form $\boldsymbol{\psi}%
^{T}\mathbf{Q}^{-1}\boldsymbol{\psi}$, where the regularization parameters
$\xi_{i}=\xi\ll1$ determine negligible constraints.

It follows that the shape of the decision space $Z=Z_{1}\cup Z_{2}$ of the
minimum risk binary classification system is completely determined by the
geometric locus of the novel principal eigenaxis $\boldsymbol{\kappa
}=\boldsymbol{\kappa}_{1}-\boldsymbol{\kappa}_{2}$ of the system, such that
the shape of the geometric locus of the decision boundary that is represented
by the graph of the vector algebra locus equation%
\[
\left(  k_{\mathbf{s}}\boldsymbol{-}\frac{1}{l}\sum\nolimits_{i=1}%
^{l}k_{\mathbf{x}_{i\ast}}\right)  \left(  \boldsymbol{\kappa}_{1}%
-\boldsymbol{\kappa}_{2}\right)  +\frac{1}{l}\sum\nolimits_{i=1}^{l}%
y_{i}\left(  1-\xi_{i}\right)  =0\text{,}%
\]
the shape of the geometric locus of the decision border of the decision region
$Z_{1}$ that is represented by the graph of the vector algebra locus equation%
\[
\left(  k_{\mathbf{s}}\boldsymbol{-}\frac{1}{l}\sum\nolimits_{i=1}%
^{l}k_{\mathbf{x}_{i\ast}}\right)  \left(  \boldsymbol{\kappa}_{1}%
-\boldsymbol{\kappa}_{2}\right)  +\frac{1}{l}\sum\nolimits_{i=1}^{l}%
y_{i}\left(  1-\xi_{i}\right)  =+1\text{,}%
\]
and the shape of the geometric locus of the decision border of the decision
region $Z_{2}$ that is represented by the graph of the vector algebra locus
equation%
\[
\left(  k_{\mathbf{s}}\boldsymbol{-}\frac{1}{l}\sum\nolimits_{i=1}%
^{l}k_{\mathbf{x}_{i\ast}}\right)  \left(  \boldsymbol{\kappa}_{1}%
-\boldsymbol{\kappa}_{2}\right)  +\frac{1}{l}\sum\nolimits_{i=1}^{l}%
y_{i}\left(  1-\xi_{i}\right)  =-1
\]
are all determined by the geometric locus of the novel principal eigenaxis
$\boldsymbol{\kappa}=\boldsymbol{\kappa}_{1}-\boldsymbol{\kappa}_{2}$, wherein
the exclusive principal eigen-coordinate system $\boldsymbol{\kappa
}=\boldsymbol{\kappa}_{1}-\boldsymbol{\kappa}_{2}$ is the principal part of an
equivalent representation of the pair of random quadratic forms
$\boldsymbol{\psi}^{T}\mathbf{Q}\boldsymbol{\psi}$ and $\boldsymbol{\psi}%
^{T}\mathbf{Q}^{-1}\boldsymbol{\psi}$ in such a manner that the geometric
locus of the novel principal eigenaxis $\boldsymbol{\kappa}=\boldsymbol{\kappa
}_{1}-\boldsymbol{\kappa}_{2}$ is the principal eigenaxis of the geometric
loci of the decision boundary and the pair of symmetrically positioned
decision borders of the decision regions $Z_{1}$ and $Z_{2}$ of the minimum
risk binary classification system%
\[
\left(  k_{\mathbf{s}}\boldsymbol{-}\frac{1}{l}\sum\nolimits_{i=1}%
^{l}k_{\mathbf{x}_{i\ast}}\right)  \left(  \boldsymbol{\kappa}_{1}%
-\boldsymbol{\kappa}_{2}\right)  +\frac{1}{l}\sum\nolimits_{i=1}^{l}%
y_{i}\left(  1-\xi_{i}\right)  \overset{\omega_{1}}{\underset{\omega
_{2}}{\gtrless}}0\text{,}%
\]
so that the geometric locus of the novel principal eigenaxis
$\boldsymbol{\kappa}=\boldsymbol{\kappa}_{1}-\boldsymbol{\kappa}_{2}$
represents an eigenaxis of symmetry that spans the decision space $Z=Z_{1}\cup
Z_{2}$ of the minimum risk binary classification system, such that the
magnitude and the direction of the novel principal eigenaxis
$\boldsymbol{\kappa}=\boldsymbol{\kappa}_{1}-\boldsymbol{\kappa}_{2}$ are both
functions of differences between joint variabilities of extreme vectors
$k_{\mathbf{x}_{1_{i\ast}}}$ and $k_{\mathbf{x}_{2_{i\ast}}}$ that belong to
the two classes $\omega_{1}$ and $\omega_{2}$ of random vectors $\mathbf{x\in
}$ $%
\mathbb{R}
^{d}$, at which point the geometric locus of the novel principal eigenaxis
$\boldsymbol{\kappa}=\boldsymbol{\kappa}_{1}-\boldsymbol{\kappa}_{2}$
satisfies the geometric locus of the decision boundary in terms of a critical
minimum eigenenergy $\left\Vert \boldsymbol{\kappa}_{1}-\boldsymbol{\kappa
}_{2}\right\Vert _{\min_{c}}^{2}$ and a minimum expected risk $\mathfrak{R}%
_{\mathfrak{\min}}\left(  \left\Vert \boldsymbol{\kappa}_{1}%
-\boldsymbol{\kappa}_{2}\right\Vert _{\min_{c}}^{2}\right)  $, so that the
uniform properties exhibited by all of the points $\mathbf{s}$ that lie on the
geometric locus of the decision boundary are the critical minimum eigenenergy
$\left\Vert \boldsymbol{\kappa}_{1}-\boldsymbol{\kappa}_{2}\right\Vert
_{\min_{c}}^{2}$ and the minimum risk $\mathfrak{R}_{\mathfrak{\min}}\left(
\left\Vert \boldsymbol{\kappa}_{1}-\boldsymbol{\kappa}_{2}\right\Vert
_{\min_{c}}^{2}\right)  $ exhibited by the geometric locus of the novel
principal eigenaxis $\boldsymbol{\kappa}=\boldsymbol{\kappa}_{1}%
-\boldsymbol{\kappa}_{2}$.
\end{theorem}

\begin{proof}
Theorem \ref{Principal Eigenstructures Theorem} is proved by conditions
expressed by Theorem \ref{Principal Eigen-coordinate System Theorem} and
Corollary \ref{Principal Eigen-coordinate System Corollary}, along with a
constructive proof that demonstrates how a well-posed constrained optimization
algorithm executes the fundamental laws of binary classification expressed by
Theorem \ref{Direct Problem of Binary Classification Theorem}.
\end{proof}

As of now, we have proved each and every one of the fundamental laws of binary
classification---that are expressed by Theorem
\ref{Direct Problem of Binary Classification Theorem}---by means of a
constructive proof that demonstrates how a well-posed constrained optimization
algorithm executes the fundamental laws. An overview of the constructive proof
is presented below.

\section{\label{Section 22}Overview of a Constructive Proof}

We have demonstrated how a well-posed constrained optimization algorithm
transforms a collection of labeled feature vectors%
\[
\left(  \mathbf{x}_{1}\mathbf{,}y_{1}\right)  ,\ldots,\left(  \mathbf{x}%
_{N}\mathbf{,}y_{N}\right)  \in%
\mathbb{R}
^{d}\times Y,Y=\left\{  \pm1\right\}  \text{,}%
\]
wherein $N$ feature vectors $\mathbf{x}\in%
\mathbb{R}
^{d}$ are generated by certain probability density functions $p\left(
\mathbf{x};\omega_{1}\right)  $ and $p\left(  \mathbf{x};\omega_{2}\right)  $,
into a data-driven mathematical model of a minimum risk binary classification
system%
\begin{align*}
&  \left(  k_{\mathbf{s}}\boldsymbol{-}\frac{1}{l}\sum\nolimits_{i=1}%
^{l}k_{\mathbf{x}_{i\ast}}\right)  \left[  \sum\nolimits_{i=1}^{l_{1}}%
\psi_{1_{i_{\ast}}}k_{\mathbf{x}_{1_{i\ast}}}-\sum\nolimits_{i=1}^{l_{2}}%
\psi_{2_{i_{\ast}}}k_{\mathbf{x}_{2_{i\ast}}}\right] \\
&  +\frac{1}{l}\sum\nolimits_{i=1}^{l}y_{i}\left(  1-\xi_{i}\right)
\overset{\omega_{1}}{\underset{\omega_{2}}{\gtrless}}0\text{,}%
\end{align*}
such that the fundamental unknowns are the scale factors $\psi_{1i\ast}$ and
$\psi_{2i\ast}$ for the components $\psi_{1i\ast}\frac{k_{\mathbf{x}_{1i\ast}%
}}{\left\Vert k_{\mathbf{x}_{1i\ast}}\right\Vert }$ and $\psi_{2i\ast}%
\frac{k_{\mathbf{x}_{2i\ast}}}{\left\Vert k_{\mathbf{x}_{2i\ast}}\right\Vert
}$ of a Wolfe-dual novel principal eigenaxis%
\begin{align*}
\boldsymbol{\psi}  &  =\sum\nolimits_{i=1}^{l_{1}}\psi_{1i\ast}\frac
{k_{\mathbf{x}_{1i\ast}}}{\left\Vert k_{\mathbf{x}_{1i\ast}}\right\Vert }%
+\sum\nolimits_{i=1}^{l_{2}}\psi_{2i\ast}\frac{k_{\mathbf{x}_{2i\ast}}%
}{\left\Vert k_{\mathbf{x}_{2i\ast}}\right\Vert }\\
&  =\boldsymbol{\psi}_{1}+\boldsymbol{\psi}_{2}%
\end{align*}
whose structure and behavior and properties are symmetrically and equivalently
related to the structure and behavior and properties of the primal novel
principal eigenaxis%
\begin{align*}
\boldsymbol{\kappa}  &  =\sum\nolimits_{i=1}^{l_{1}}\psi_{1_{i_{\ast}}%
}k_{\mathbf{x}_{1_{i\ast}}}-\sum\nolimits_{i=1}^{l_{2}}\psi_{2_{i_{\ast}}%
}k_{\mathbf{x}_{2_{i\ast}}}\\
&  =\boldsymbol{\kappa}_{1}-\boldsymbol{\kappa}_{2}%
\end{align*}
of the minimum risk binary classification in such a manner that the
discriminant function%
\[
d\left(  \mathbf{s}\right)  =\left(  k_{\mathbf{s}}\boldsymbol{-}\frac{1}%
{l}\sum\nolimits_{i=1}^{l}k_{\mathbf{x}_{i\ast}}\right)  \left(
\boldsymbol{\kappa}_{1}-\boldsymbol{\kappa}_{2}\right)  +\frac{1}{l}%
\sum\nolimits_{i=1}^{l}y_{i}\left(  1-\xi_{i}\right)
\]
is represented by a geometric locus of a novel principal eigenaxis%
\begin{align*}
\boldsymbol{\kappa}  &  =\sum\nolimits_{i=1}^{l_{1}}\psi_{1_{i_{\ast}}%
}k_{\mathbf{x}_{1_{i\ast}}}-\sum\nolimits_{i=1}^{l_{2}}\psi_{2_{i_{\ast}}%
}k_{\mathbf{x}_{2_{i\ast}}}\\
&  =\boldsymbol{\kappa}_{1}-\boldsymbol{\kappa}_{2}\text{,}%
\end{align*}
at which point a dual locus $\boldsymbol{\kappa}=\boldsymbol{\kappa}%
_{1}-\boldsymbol{\kappa}_{2}$ of likelihood components and principal eigenaxis
components $\psi_{1_{i_{\ast}}}k_{\mathbf{x}_{1_{i\ast}}}$ and $\psi
_{2_{i_{\ast}}}k_{\mathbf{x}_{2_{i\ast}}}$ represents an exclusive principal
eigen-coordinate system of the geometric locus of the decision boundary of the
system, and also represents an eigenaxis of symmetry that spans the decision
space of the system, wherein each scale factor $\psi_{1_{i_{\ast}}}$ or
$\psi_{2_{i_{\ast}}}$ determines a scaled extreme vector $\psi_{1_{i_{\ast}}%
}k_{\mathbf{x}_{1_{i\ast}}}$ or $\psi_{2_{i_{\ast}}}k_{\mathbf{x}_{2_{i\ast}}%
}$ that represents a principal eigenaxis component---on the exclusive
principal eigen-coordinate system $\boldsymbol{\kappa}=\boldsymbol{\kappa}%
_{1}-\boldsymbol{\kappa}_{2}$---that determines a likely location for a
correlated extreme point $\mathbf{x}_{1_{i\ast}}\mathbf{\sim}$ $p\left(
\mathbf{x};\omega_{1}\right)  $ or $\mathbf{x}_{2_{i\ast}}\mathbf{\sim}$
$p\left(  \mathbf{x};\omega_{2}\right)  $, along with a likelihood component
that determines a likelihood value for the correlated extreme point
$\mathbf{x}_{1_{i\ast}}$ or $\mathbf{x}_{2_{i\ast}}$, where the reproducing
kernel for each extreme point $k_{\mathbf{x}_{1_{i\ast}}}$ and $k_{\mathbf{x}%
_{2_{i\ast}}}$ has the preferred form of either $k_{\mathbf{x}}\left(
\mathbf{s}\right)  =\left(  \mathbf{s}^{T}\mathbf{x}+1\right)  ^{2}$ or
$k_{\mathbf{x}}\left(  \mathbf{s}\right)  =\exp\left(  -\gamma\left\Vert
\mathbf{s}-\mathbf{x}\right\Vert ^{2}\right)  $, wherein $0.01\leq\gamma
\leq0.1$.

By the process of determining the scale factors $\psi_{1i\ast}$ and
$\psi_{2i\ast}$ in accordance with the vector algebra locus equations in
(\ref{Wolf-dual Comp 1}) and (\ref{Wolf-dual Comp 2}), we have demonstrated
how each scale factor $\psi_{1i\ast}$ and $\psi_{2i\ast}$ maps covariance and
distribution information---for a correlated extreme vector $k_{\mathbf{x}%
_{1_{i\ast}}}$ and $k_{\mathbf{x}_{2_{i\ast}}}$---onto the correlated extreme
vector $k_{\mathbf{x}_{1_{i\ast}}}$ and $k_{\mathbf{x}_{2_{i\ast}}}$ in such a
manner that each scale factor $\psi_{1_{i_{\ast}}}$ or $\psi_{2_{i_{\ast}}}$
determines a scaled extreme vector $\psi_{1_{i_{\ast}}}k_{\mathbf{x}%
_{1_{i\ast}}}$ or $\psi_{2_{i_{\ast}}}k_{\mathbf{x}_{2_{i\ast}}}$ that
represents a principal eigenaxis component on the exclusive principal
eigen-coordinate system $\boldsymbol{\kappa}=\boldsymbol{\kappa}%
_{1}-\boldsymbol{\kappa}_{2}$, along with a likelihood component, so that each
principal eigenaxis component $\psi_{1_{i_{\ast}}}k_{\mathbf{x}_{1_{i\ast}}}$
or $\psi_{2_{i_{\ast}}}k_{\mathbf{x}_{2_{i\ast}}}$ on $\boldsymbol{\kappa
}=\boldsymbol{\kappa}_{1}-\boldsymbol{\kappa}_{2}$ determines a likely
location for a correlated extreme point $\mathbf{x}_{1_{i\ast}}$ or
$\mathbf{x}_{2_{i\ast}}$, and each likelihood component $\psi_{1_{i_{\ast}}%
}k_{\mathbf{x}_{1_{i\ast}}}$ or $\psi_{2_{i_{\ast}}}k_{\mathbf{x}_{2_{i\ast}}%
}$ on $\boldsymbol{\kappa}=\boldsymbol{\kappa}_{1}-\boldsymbol{\kappa}_{2}$
determines a likelihood value for the correlated extreme point $\mathbf{x}%
_{1_{i\ast}}$ or $\mathbf{x}_{2_{i\ast}}$.

By the process of determining the vector algebra locus equations of
(\ref{Decision Boundary}) - (\ref{Decision Border 2}), we have demonstrated
how the geometric locus of the novel principal eigenaxis $\boldsymbol{\kappa
}=\boldsymbol{\kappa}_{1}-\boldsymbol{\kappa}_{2}$ of a minimum risk binary
classification system $k_{\mathbf{s}}\boldsymbol{\kappa}+$ $\boldsymbol{\kappa
}_{0}\overset{\omega_{1}}{\underset{\omega_{2}}{\gtrless}}0$ represents an
exclusive principal eigen-coordinate system of the geometric locus of the
decision boundary of the system and also represents an eigenaxis of symmetry
for the decision space $Z=Z_{1}\cup Z_{2}$ of the system.

By the conditions expressed in Theorem \ref{Principal Eigenstructures Theorem}%
, we have demonstrated how a geometric locus of a novel principal eigenaxis
$\boldsymbol{\kappa}=\boldsymbol{\kappa}_{1}-\boldsymbol{\kappa}_{2}$ is the
principal part of an equivalent representation of a pair of random quadratic
forms $\boldsymbol{\psi}^{T}\mathbf{Q}\boldsymbol{\psi}$ and $\boldsymbol{\psi
}^{T}\mathbf{Q}^{-1}\boldsymbol{\psi}$ associated with a joint covariance
matrix $\mathbf{Q}$ and the inverted joint covariance matrix $\mathbf{Q}^{-1}%
$, so that the novel principal eigenaxis $\boldsymbol{\kappa}%
=\boldsymbol{\kappa}_{1}-\boldsymbol{\kappa}_{2}$ is the principal eigenaxis
of the decision boundary of a minimum risk binary classification system
$k_{\mathbf{s}}\boldsymbol{\kappa}+$ $\boldsymbol{\kappa}_{0}\overset{\omega
_{1}}{\underset{\omega_{2}}{\gtrless}}0$, at which point the geometric locus
of the novel principal eigenaxis $\boldsymbol{\kappa}=\boldsymbol{\kappa}%
_{1}-\boldsymbol{\kappa}_{2}$ satisfies the geometric locus of the decision
boundary in terms of the critical minimum eigenenergy and the minimum expected
risk exhibited by the minimum risk binary classification system.

Theorem \ref{Principal Eigenstructures Theorem} also identifies critical
interconnections between the intrinsic components of a minimum risk binary
classification system that determine the statistical structure and the
functionality of the discriminant function of the system, so that the
discriminant function of any given minimum risk binary classification system
generalizes in a significant manner and thereby extrapolates.

By the process of determining the integral equation in (\ref{AIE1}), we have
demonstrated how a data-driven version of the general form of the integral
equation in (\ref{TIE1}) is determined by the constrained optimization
algorithm that resolves the inverse problem of the binary classification of
random vectors. Accordingly, we have demonstrated that any given minimum risk
binary classification system satisfies the law of total allowed eigenenergy
for minimum risk binary classification systems expressed by (\ref{TIE1}) in
Theorem \ref{Direct Problem of Binary Classification Theorem}.

Thereby, we have also demonstrated that the geometric locus of the novel
principal eigenaxis $\boldsymbol{\kappa}=\boldsymbol{\kappa}_{1}%
-\boldsymbol{\kappa}_{2}$ of any given minimum risk binary classification
system $k_{\mathbf{s}}\boldsymbol{\kappa}+$ $\boldsymbol{\kappa}%
_{0}\overset{\omega_{1}}{\underset{\omega_{2}}{\gtrless}}0$ satisfies the law
of cosines in a symmetrically balanced manner, so that the minimum risk binary
classification system satisfies the law of symmetry for minimum risk binary
classification systems expressed by (\ref{Law of Cosines T}) in Theorem
\ref{Direct Problem of Binary Classification Theorem}.

By the process of determining the integral equation in (\ref{AIE2}), we have
demonstrated how a data-driven version of the general form of the integral
equation in (\ref{TIE2}) is determined by the constrained optimization
algorithm that resolves the inverse problem of binary classification. Thereby,
we have demonstrated that any given minimum risk binary classification system
satisfies the law of statistical equilibrium for minimum risk binary
classification systems expressed by (\ref{TIE2}) in Theorem
\ref{Direct Problem of Binary Classification Theorem}.

In conclusion, we have demonstrated how data-driven versions of the general
forms of the vector algebra locus formulae in (\ref{Theoretical System}) -
(\ref{Law of Cosines T})---which are expressed in the direct problem of the
binary classification of random vectors by Theorem
\ref{Direct Problem of Binary Classification Theorem}---are determined by the
constrained optimization algorithm that resolves the inverse problem of the
binary classification of random vectors.

Thereby, we are now in a position to express the inverse problem of the binary
classification of random vectors.

\section{\label{Section 23}The Inverse Problem}

We have proved Theorem \ref{Inverse Problem of Binary Classification Theorem}
by means of a constructive proof that demonstrates how a well-posed
constrained optimization algorithm executes the fundamental laws of binary
classification expressed by Theorem
\ref{Direct Problem of Binary Classification Theorem}.

\begin{theorem}
\label{Inverse Problem of Binary Classification Theorem}Let%
\begin{equation}
\left(  k_{\mathbf{s}}\boldsymbol{-}\frac{1}{l}\sum\nolimits_{i=1}%
^{l}k_{\mathbf{x}_{i\ast}}\right)  \left(  \boldsymbol{\kappa}_{1}%
-\boldsymbol{\kappa}_{2}\right)  +\frac{1}{l}\sum\nolimits_{i=1}^{l}%
y_{i}\left(  1-\xi_{i}\right)  \overset{\omega_{1}}{\underset{\omega
_{2}}{\gtrless}}0 \tag{23.1}\label{Applied System}%
\end{equation}
be any given minimum risk binary classification system that is subject to
random inputs $\mathbf{x\in}$ $%
\mathbb{R}
^{d}$ such that $\mathbf{x\sim}$ $p\left(  \mathbf{x};\omega_{1}\right)  $ and
$\mathbf{x\sim}$ $p\left(  \mathbf{x};\omega_{2}\right)  $, where either
$\xi_{i}=\xi=0$ or $\xi_{i}=\xi\ll1$, $y_{i}=\pm1$, and $p\left(
\mathbf{x};\omega_{1}\right)  $ and $p\left(  \mathbf{x};\omega_{2}\right)  $
are certain probability density functions for two classes $\omega_{1}$ and
$\omega_{2}$ of random vectors $\mathbf{x\in}$ $%
\mathbb{R}
^{d}$, where$\ \omega_{1}$ or $\omega_{2}$ is the true category, satisfying
the following geometrical and statistical criteria:

$1$. The discriminant function%
\begin{equation}
d\left(  \mathbf{s}\right)  =\left(  k_{\mathbf{s}}\boldsymbol{-}\frac{1}%
{l}\sum\nolimits_{i=1}^{l}k_{\mathbf{x}_{i\ast}}\right)  \left(
\boldsymbol{\kappa}_{1}-\boldsymbol{\kappa}_{2}\right)  +\frac{1}{l}%
\sum\nolimits_{i=1}^{l}y_{i}\left(  1-\xi_{i}\right)  \tag{23.2}%
\label{Decision Function A}%
\end{equation}
is represented by a geometric locus of a novel principal eigenaxis%
\begin{align}
\boldsymbol{\kappa} &  =\boldsymbol{\kappa}_{1}-\boldsymbol{\kappa}%
_{2}\tag{23.3}\label{Novel Principal Eigenaxis A}\\
&  =\sum\nolimits_{i=1}^{l_{1}}\psi_{1_{i_{\ast}}}k_{\mathbf{x}_{1_{i\ast}}%
}-\sum\nolimits_{i=1}^{l_{2}}\psi_{2_{i_{\ast}}}k_{\mathbf{x}_{2_{i\ast}}%
}\nonumber
\end{align}
structured as a locus of signed and scaled extreme vectors $\psi_{1_{i_{\ast}%
}}k_{\mathbf{x}_{1_{i\ast}}}$ and $-\psi_{2_{i_{\ast}}}k_{\mathbf{x}%
_{2_{i\ast}}}$, so that a dual locus of likelihood components and principal
eigenaxis components $\psi_{1_{i_{\ast}}}k_{\mathbf{x}_{1_{i\ast}}}$ and
$\psi_{2_{i_{\ast}}}k_{\mathbf{x}_{2_{i\ast}}}$ represents an exclusive
principal eigen-coordinate system of the geometric locus of the decision
boundary of the system, and also represents an eigenaxis of symmetry that
spans the decision space $Z=Z_{1}\cup Z_{2}$ of the system, such that each
scale factor $\psi_{1_{i_{\ast}}}$ or $\psi_{2_{i_{\ast}}}$ determines a
scaled extreme vector $\psi_{1_{i_{\ast}}}k_{\mathbf{x}_{1_{i\ast}}}$ or
$\psi_{2_{i_{\ast}}}k_{\mathbf{x}_{2_{i\ast}}}$ that represents a principal
eigenaxis component that determines a likely location for a correlated extreme
point $\mathbf{x}_{1_{i\ast}}\mathbf{\sim}$ $p\left(  \mathbf{x};\omega
_{1}\right)  $ or $\mathbf{x}_{2_{i\ast}}\mathbf{\sim}$ $p\left(
\mathbf{x};\omega_{2}\right)  $, along with a likelihood component that
determines a likelihood value for the correlated extreme point $\mathbf{x}%
_{1_{i\ast}}$ or $\mathbf{x}_{2_{i\ast}}$, where the reproducing kernel for
each extreme point $k_{\mathbf{x}_{1_{i\ast}}}$ and $k_{\mathbf{x}_{2_{i\ast}%
}}$ has the preferred form of either $k_{\mathbf{x}}\left(  \mathbf{s}\right)
=\left(  \mathbf{s}^{T}\mathbf{x}+1\right)  ^{2}$ or $k_{\mathbf{x}}\left(
\mathbf{s}\right)  =\exp\left(  -\gamma\left\Vert \mathbf{s}-\mathbf{x}%
\right\Vert ^{2}\right)  $, wherein $0.01\leq\gamma\leq0.1$;

$2$. The geometric locus of the novel principal eigenaxis $\boldsymbol{\kappa
}=\boldsymbol{\kappa}_{1}-\boldsymbol{\kappa}_{2}$ is the solution of the
vector algebra locus equation%
\begin{equation}
\left(  k_{\mathbf{s}}\boldsymbol{-}\frac{1}{l}\sum\nolimits_{i=1}%
^{l}k_{\mathbf{x}_{i\ast}}\right)  \left(  \boldsymbol{\kappa}_{1}%
-\boldsymbol{\kappa}_{2}\right)  +\frac{1}{l}\sum\nolimits_{i=1}^{l}%
y_{i}\left(  1-\xi_{i}\right)  =0 \tag{23.4}%
\label{Equation of Decision Boundary A}%
\end{equation}
that represents the geometric locus of the decision boundary of the system,
where the expression $\frac{1}{l}\sum\nolimits_{i=1}^{l}k_{\mathbf{x}_{i\ast}%
}$ represents a locus of average risk in the decision space $Z=Z_{1}\cup
Z_{2}$ of the system, and the statistic $\frac{1}{l}\sum\nolimits_{i=1}%
^{l}y_{i}\left(  1-\xi_{i}\right)  :y_{i}=\pm1$ represents an expected
likelihood of observing $l$ extreme vectors $\left\{  k_{\mathbf{x}_{i\ast}%
}\right\}  _{i=1}^{l}$ within the decision space $Z=Z_{1}\cup Z_{2}$, so that
all of the points $\mathbf{s}$ that lie on the geometric locus of the decision
boundary exclusively reference the novel principal eigenaxis
$\boldsymbol{\kappa}=\boldsymbol{\kappa}_{1}-\boldsymbol{\kappa}_{2}$, as well
as the vector algebra locus equations%
\begin{equation}
\left(  k_{\mathbf{s}}\boldsymbol{-}\frac{1}{l}\sum\nolimits_{i=1}%
^{l}k_{\mathbf{x}_{i\ast}}\right)  \left(  \boldsymbol{\kappa}_{1}%
-\boldsymbol{\kappa}_{2}\right)  +\frac{1}{l}\sum\nolimits_{i=1}^{l}%
y_{i}\left(  1-\xi_{i}\right)  =+1 \tag{23.5}\label{Equation of Border 1 A}%
\end{equation}
and%
\begin{equation}
\left(  k_{\mathbf{s}}\boldsymbol{-}\frac{1}{l}\sum\nolimits_{i=1}%
^{l}k_{\mathbf{x}_{i\ast}}\right)  \left(  \boldsymbol{\kappa}_{1}%
-\boldsymbol{\kappa}_{2}\right)  +\frac{1}{l}\sum\nolimits_{i=1}^{l}%
y_{i}\left(  1-\xi_{i}\right)  =-1 \tag{23.6}\label{Equation of Border 2 A}%
\end{equation}
that represent the geometric loci of the decision borders of the corresponding
decision regions $Z_{1}$ and $Z_{2}$ of the system, so that all of the points
$\mathbf{s}$ that lie on the geometric loci of the decision borders
exclusively reference the novel principal eigenaxis $\boldsymbol{\kappa
}=\boldsymbol{\kappa}_{1}-\boldsymbol{\kappa}_{2}$.

Thereby, the geometric locus of the novel principal eigenaxis
$\boldsymbol{\kappa}=\boldsymbol{\kappa}_{1}-\boldsymbol{\kappa}_{2}$
represents an eigenaxis of symmetry that spans the decision space $Z=Z_{1}\cup
Z_{2}$ of the minimum risk binary classification system%
\[
\left(  k_{\mathbf{s}}\boldsymbol{-}\frac{1}{l}\sum\nolimits_{i=1}%
^{l}k_{\mathbf{x}_{i\ast}}\right)  \left(  \boldsymbol{\kappa}_{1}%
-\boldsymbol{\kappa}_{2}\right)  +\frac{1}{l}\sum\nolimits_{i=1}^{l}%
y_{i}\left(  1-\xi_{i}\right)  \overset{\omega_{1}}{\underset{\omega
_{2}}{\gtrless}}0\text{,}%
\]
at which point the shape of the decision space $Z=Z_{1}\cup Z_{2}$ is
completely determined by the exclusive principal eigen-coordinate system
$\boldsymbol{\kappa}=\boldsymbol{\kappa}_{1}-\boldsymbol{\kappa}_{2}$;

$3$. The discriminant function is the solution of the integral equation%
\begin{align}
f_{1}\left(  d\left(  \mathbf{s}\right)  \right)   &  :\int_{Z_{1}%
}\boldsymbol{\kappa}_{1}d\boldsymbol{\kappa}_{1}+\int_{Z_{2}}%
\boldsymbol{\kappa}_{1}d\boldsymbol{\kappa}_{1}+\delta\left(  y\right)
\lambda_{1}^{-1}\sum\nolimits_{i=1}^{l_{1}}k_{\mathbf{x}_{1_{i\ast}}}\left(
\boldsymbol{\kappa}_{1}-\boldsymbol{\kappa}_{2}\right)  \tag{23.7}%
\label{AIE_One}\\
&  =\int_{Z_{1}}\boldsymbol{\kappa}_{2}d\boldsymbol{\kappa}_{2}+\int_{Z_{2}%
}\boldsymbol{\kappa}_{2}d\boldsymbol{\kappa}_{2}+\delta\left(  y\right)
\lambda_{1}^{-1}\sum\nolimits_{i=1}^{l_{2}}k_{\mathbf{x}_{2_{i\ast}}}\left(
\boldsymbol{\kappa}_{2}-\boldsymbol{\kappa}_{1}\right)  \text{,}\nonumber
\end{align}
over the decision space $Z=Z_{1}\cup Z_{2}$ of the minimum risk binary
classification system%
\[
\left(  k_{\mathbf{s}}\boldsymbol{-}\frac{1}{l}\sum\nolimits_{i=1}%
^{l}k_{\mathbf{x}_{i\ast}}\right)  \left(  \boldsymbol{\kappa}_{1}%
-\boldsymbol{\kappa}_{2}\right)  +\frac{1}{l}\sum\nolimits_{i=1}^{l}%
y_{i}\left(  1-\xi_{i}\right)  \overset{\omega_{1}}{\underset{\omega
_{2}}{\gtrless}}0\text{,}%
\]
where $\delta\left(  y\right)  =\frac{1}{l}\sum\nolimits_{i=1}^{l}y_{i}\left(
1-\xi_{i}\right)  $, so that the total allowed eigenenergy $\left\Vert
\boldsymbol{\kappa}_{1}-\boldsymbol{\kappa}_{2}\right\Vert _{\min_{c}}^{2}$
and the expected risk $\mathfrak{R}_{\mathfrak{\min}}\left(  \left\Vert
\boldsymbol{\kappa}_{1}-\boldsymbol{\kappa}_{2}\right\Vert _{\min_{c}}%
^{2}\right)  $ exhibited by the system are jointly regulated by the
equilibrium requirement on the dual locus $\boldsymbol{\kappa}%
=\boldsymbol{\kappa}_{1}-\boldsymbol{\kappa}_{2}$ of the discriminant function
at the geometric locus of the decision boundary of the system%
\begin{align*}
d\left(  \mathbf{s}\right)   &  :\left\Vert \boldsymbol{\kappa}_{1}\right\Vert
_{\min_{c}}^{2}-\left\Vert \boldsymbol{\kappa}_{1}\right\Vert \left\Vert
\boldsymbol{\kappa}_{2}\right\Vert \cos\theta_{\boldsymbol{\kappa}%
_{1}\boldsymbol{\kappa}_{2}}\\
&  =\left\Vert \boldsymbol{\kappa}_{2}\right\Vert _{\min_{c}}^{2}-\left\Vert
\boldsymbol{\kappa}_{2}\right\Vert \left\Vert \boldsymbol{\kappa}%
_{1}\right\Vert \cos\theta_{\boldsymbol{\kappa}_{2}\boldsymbol{\kappa}_{1}}\\
&  \equiv\frac{1}{2}\left\Vert \boldsymbol{\kappa}_{1}-\boldsymbol{\kappa}%
_{2}\right\Vert _{\min_{c}}^{2}\text{,}%
\end{align*}
at which point the dual locus $\boldsymbol{\kappa}=\boldsymbol{\kappa}%
_{1}-\boldsymbol{\kappa}_{2}$ of the discriminant function satisfies the
geometric locus of the decision boundary in terms of a critical minimum
eigenenergy $\left\Vert \boldsymbol{\kappa}\right\Vert _{\min_{c}}^{2}$ and a
minimum expected risk $\mathfrak{R}_{\mathfrak{\min}}\left(  \left\Vert
\boldsymbol{\kappa}\right\Vert _{\min_{c}}^{2}\right)  $ in such a manner that
regions of counter risks of the system are symmetrically balanced with regions
of risks of the system, so that critical minimum eigenenergies $\left\Vert
\psi_{1_{i\ast}}k_{\mathbf{x}_{1_{i\ast}}}\right\Vert _{\min_{c}}^{2}$
exhibited by principal eigenaxis components $\psi_{1_{i_{\ast}}}%
k_{\mathbf{x}_{1_{i\ast}}}$ on side $\boldsymbol{\kappa}_{1}$ of the novel
principal eigenaxis $\boldsymbol{\kappa}=\boldsymbol{\kappa}_{1}%
-\boldsymbol{\kappa}_{2}$---that determine probabilities of finding extreme
points $\mathbf{x}_{1_{i\ast}}$ located throughout the decision space
$Z=Z_{1}\cup Z_{2}$ of the system, are symmetrically balanced with critical
minimum eigenenergies $\left\Vert \psi_{2_{i_{\ast}}}k_{\mathbf{x}_{2_{i\ast}%
}}\right\Vert _{\min_{c}}^{2}$ exhibited by principal eigenaxis components
$\psi_{2_{i_{\ast}}}k_{\mathbf{x}_{2_{i\ast}}}$ on side $\boldsymbol{\kappa
}_{2}$ of the novel principal eigenaxis $\boldsymbol{\kappa}%
=\boldsymbol{\kappa}_{1}-\boldsymbol{\kappa}_{2}$---that determine
probabilities of finding extreme points $\mathbf{x}_{2_{i\ast}}$ located
throughout the decision space $Z=Z_{1}\cup Z_{2}$ of the system;

$4$. The discriminant function minimizes the integral equation%
\begin{align}
f_{2}\left(  d\left(  \mathbf{s}\right)  \right)   &  :\int_{Z_{1}%
}\boldsymbol{\kappa}_{1}d\boldsymbol{\kappa}_{1}-\int_{Z_{1}}%
\boldsymbol{\kappa}_{2}d\boldsymbol{\kappa}_{2}+\delta\left(  y\right)
\lambda_{1}^{-1}\sum\nolimits_{i=1}^{l_{1}}k_{\mathbf{x}_{1_{i\ast}}}\left(
\boldsymbol{\kappa}_{1}-\boldsymbol{\kappa}_{2}\right)  \tag{23.8}%
\label{AIE_Two}\\
&  =\int_{Z_{2}}\boldsymbol{\kappa}_{2}d\boldsymbol{\kappa}_{2}-\int_{Z_{2}%
}\boldsymbol{\kappa}_{1}d\boldsymbol{\kappa}_{1}+\delta\left(  y\right)
\lambda_{1}^{-1}\sum\nolimits_{i=1}^{l_{2}}k_{\mathbf{x}_{2_{i\ast}}}\left(
\boldsymbol{\kappa}_{2}-\boldsymbol{\kappa}_{1}\right)  \text{,}\nonumber
\end{align}
over the decision regions $Z_{1}$ and $Z_{2}$ of the minimum risk binary
classification system $\left(  k_{\mathbf{s}}\boldsymbol{-}\frac{1}{l}%
\sum\nolimits_{i=1}^{l}k_{\mathbf{x}_{i\ast}}\right)  \left(
\boldsymbol{\kappa}_{1}-\boldsymbol{\kappa}_{2}\right)  +\frac{1}{l}%
\sum\nolimits_{i=1}^{l}y_{i}\left(  1-\xi_{i}\right)  \overset{\omega
_{1}}{\underset{\omega_{2}}{\gtrless}}0$, where $\delta\left(  y\right)
=\frac{1}{l}\sum\nolimits_{i=1}^{l}y_{i}\left(  1-\xi_{i}\right)  $, so that
the system satisfies a state of statistical equilibrium such that the total
allowed eigenenergy $\left\Vert \boldsymbol{\kappa}_{1}-\boldsymbol{\kappa
}_{2}\right\Vert _{\min_{c}}^{2}$ and the expected risk $\mathfrak{R}%
_{\mathfrak{\min}}\left(  \left\Vert \boldsymbol{\kappa}_{1}%
-\boldsymbol{\kappa}_{2}\right\Vert _{\min_{c}}^{2}\right)  $ exhibited by the
system are jointly minimized within the decision space $Z=Z_{1}\cup Z_{2}$ of
the system, at which point critical minimum eigenenergies $\left\Vert
\psi_{1_{i\ast}}k_{\mathbf{x}_{1_{i\ast}}}\right\Vert _{\min_{c}}^{2}$ and
$\left\Vert \psi_{2_{i_{\ast}}}k_{\mathbf{x}_{2_{i\ast}}}\right\Vert
_{\min_{c}}^{2}$ exhibited by corresponding principal eigenaxis components
$\psi_{1_{i_{\ast}}}k_{\mathbf{x}_{1_{i\ast}}}$ and $\psi_{2_{i_{\ast}}%
}k_{\mathbf{x}_{2_{i\ast}}}$ on side $\boldsymbol{\kappa}_{1}$ and side
$\boldsymbol{\kappa}_{2}$ of the novel principal eigenaxis $\boldsymbol{\kappa
}=\boldsymbol{\kappa}_{1}-\boldsymbol{\kappa}_{2}$ are minimized throughout
the decision regions $Z_{1}$ and $Z_{2}$ of the system, so that regions of
counter risks and risks of the system---located throughout the decision region
$Z_{1}$ of the system---are symmetrically balanced with regions of counter
risks and risks of the system---located throughout the decision region $Z_{2}$
of the system.

Thereby, the minimum risk binary classification system%
\[
\left(  k_{\mathbf{s}}\boldsymbol{-}\frac{1}{l}\sum\nolimits_{i=1}%
^{l}k_{\mathbf{x}_{i\ast}}\right)  \left(  \boldsymbol{\kappa}_{1}%
-\boldsymbol{\kappa}_{2}\right)  +\frac{1}{l}\sum\nolimits_{i=1}^{l}%
y_{i}\left(  1-\xi_{i}\right)  \overset{\omega_{1}}{\underset{\omega
_{2}}{\gtrless}}0
\]
satisfies a state of statistical equilibrium so that the total allowed
eigenenergy and the expected risk exhibited by the system are jointly
minimized within the decision space $Z=Z_{1}\cup Z_{2}$ of the system, at
which point the system exhibits the minimum probability of classification
error for any given random vectors $\mathbf{x\in}$ $%
\mathbb{R}
^{d}$ such that $\mathbf{x\sim}$ $p\left(  \mathbf{x};\omega_{1}\right)  $ and
$\mathbf{x\sim}$ $p\left(  \mathbf{x};\omega_{2}\right)  $;

$5$. The geometric locus of the novel principal eigenaxis $\boldsymbol{\kappa
}=\boldsymbol{\kappa}_{1}-\boldsymbol{\kappa}_{2}$ satisfies the law of
cosines in the symmetrically balanced manner%
\begin{align}
\frac{1}{2}\left\Vert \boldsymbol{\kappa}\right\Vert _{\min_{c}}^{2}  &
=\left\Vert \boldsymbol{\kappa}_{1}\right\Vert _{\min_{c}}^{2}-\left\Vert
\boldsymbol{\kappa}_{1}\right\Vert \left\Vert \boldsymbol{\kappa}%
_{2}\right\Vert \cos\theta_{\boldsymbol{\kappa}_{1}\boldsymbol{\kappa}_{2}%
}\tag{23.9}\label{Law of Cosines A}\\
&  =\left\Vert \boldsymbol{\kappa}_{2}\right\Vert _{\min_{c}}^{2}-\left\Vert
\boldsymbol{\kappa}_{2}\right\Vert \left\Vert \boldsymbol{\kappa}%
_{1}\right\Vert \cos\theta_{\boldsymbol{\kappa}_{2}\boldsymbol{\kappa}_{1}%
}\text{,}\nonumber
\end{align}
where $\theta$ is the angle between $\boldsymbol{\kappa}_{1}$ and
$\boldsymbol{\kappa}_{2}$, so that the geometric locus of the novel principal
eigenaxis $\boldsymbol{\kappa}=\boldsymbol{\kappa}_{1}-\boldsymbol{\kappa}%
_{2}$ represents an eigenaxis of symmetry that exhibits symmetrical dimensions
and densities, such that the magnitude and the direction of the novel
principal eigenaxis $\boldsymbol{\kappa}=\boldsymbol{\kappa}_{1}%
-\boldsymbol{\kappa}_{2}$ are both functions of differences between joint
variabilities of extreme vectors $k_{\mathbf{x}_{1_{i\ast}}}$ and
$k_{\mathbf{x}_{2_{i\ast}}}$ that belong to the two classes $\omega_{1}$ and
$\omega_{2}$ of random vectors $\mathbf{x\in}$ $%
\mathbb{R}
^{d}$, at which point the critical minimum eigenenergy $\left\Vert
\boldsymbol{\kappa}_{1}\right\Vert _{\min_{c}}^{2}$ exhibited by side
$\boldsymbol{\kappa}_{1}$ is symmetrically balanced with the critical minimum
eigenenergy $\left\Vert \boldsymbol{\kappa}_{2}\right\Vert _{\min_{c}}^{2}$
exhibited by side $\boldsymbol{\kappa}_{2}$%
\[
\left\Vert \boldsymbol{\kappa}_{1}\right\Vert _{\min_{c}}^{2}=\left\Vert
\boldsymbol{\kappa}_{2}\right\Vert _{\min_{c}}^{2}\text{,}%
\]
the length of side $\boldsymbol{\kappa}_{1}$ equals the length of side
$\boldsymbol{\kappa}_{2}$%
\[
\left\Vert \boldsymbol{\kappa}_{1}\right\Vert =\left\Vert \boldsymbol{\kappa
}_{2}\right\Vert \text{,}%
\]
and counteracting and opposing forces and influences of the minimum risk
binary classification system%
\[
\left(  k_{\mathbf{s}}\boldsymbol{-}\frac{1}{l}\sum\nolimits_{i=1}%
^{l}k_{\mathbf{x}_{i\ast}}\right)  \left(  \boldsymbol{\kappa}_{1}%
-\boldsymbol{\kappa}_{2}\right)  +\frac{1}{l}\sum\nolimits_{i=1}^{l}%
y_{i}\left(  1-\xi_{i}\right)  \overset{\omega_{1}}{\underset{\omega
_{2}}{\gtrless}}0
\]
are symmetrically balanced with each other about the geometric center of the
locus of the novel principal eigenaxis $\boldsymbol{\kappa}=\boldsymbol{\kappa
}_{1}-\boldsymbol{\kappa}_{2}$%
\begin{align*}
&  \left\Vert \boldsymbol{\kappa}_{1}\right\Vert \left(  \sum\nolimits_{i=1}%
^{l_{1}}\operatorname{comp}_{\overrightarrow{\boldsymbol{\kappa}_{1}}}\left(
\overrightarrow{\psi_{1_{i_{\ast}}}k_{\mathbf{x}_{1_{i\ast}}}}\right)
-\sum\nolimits_{i=1}^{l_{2}}\operatorname{comp}%
_{\overrightarrow{\boldsymbol{\kappa}_{1}}}\left(  \overrightarrow{\psi
_{2_{i_{\ast}}}k_{\mathbf{x}_{2_{i\ast}}}}\right)  \right) \\
&  =\left\Vert \boldsymbol{\kappa}_{2}\right\Vert \left(  \sum\nolimits_{i=1}%
^{l_{2}}\operatorname{comp}_{\overrightarrow{\boldsymbol{\kappa}_{2}}}\left(
\overrightarrow{\psi_{2_{i_{\ast}}}k_{\mathbf{x}_{2_{i\ast}}}}\right)
-\sum\nolimits_{i=1}^{l_{1}}\operatorname{comp}%
_{\overrightarrow{\boldsymbol{\kappa}_{2}}}\left(  \overrightarrow{\psi
_{1_{i_{\ast}}}k_{\mathbf{x}_{1_{i\ast}}}}\right)  \right)  \text{,}%
\end{align*}
whereon the statistical fulcrum of the novel principal eigenaxis
$\boldsymbol{\kappa}=\boldsymbol{\kappa}_{1}-\boldsymbol{\kappa}_{2}$ is located.

Thereby, counteracting and opposing components of critical minimum
eigenenergies related to likely locations of extreme points from class
$\omega_{1}$ and class $\omega_{2}$ that determine regions of counter risks
and risks of the system---along the dual locus of side $\boldsymbol{\kappa
}_{1}$---are symmetrically balanced with counteracting and opposing components
of critical minimum eigenenergies related to likely locations of extreme
points from class $\omega_{2}$ and class $\omega_{1}$ that determine regions
of counter risks and risks of the system---along the dual locus of side
$\boldsymbol{\kappa}_{2}$;

$6$. The center of total allowed eigenenergy and expected risk of the minimum
risk binary classification system%
\[
\left(  k_{\mathbf{s}}\boldsymbol{-}\frac{1}{l}\sum\nolimits_{i=1}%
^{l}k_{\mathbf{x}_{i\ast}}\right)  \left(  \boldsymbol{\kappa}_{1}%
-\boldsymbol{\kappa}_{2}\right)  +\frac{1}{l}\sum\nolimits_{i=1}^{l}%
y_{i}\left(  1-\xi_{i}\right)  \overset{\omega_{1}}{\underset{\omega
_{2}}{\gtrless}}0
\]
is located at the geometric center of the locus of the novel principal
eigenaxis $\boldsymbol{\kappa}=\boldsymbol{\kappa}_{1}-\boldsymbol{\kappa}%
_{2}$ of the system, whereon the statistical fulcrum of the system is located;

Then the minimum risk binary classification system acts to jointly minimize
its eigenenergy and risk by locating a point of equilibrium, at which point
critical minimum eigenenergies exhibited by the system are symmetrically
concentrated in such a manner that the geometric locus of the novel principal
eigenaxis of the system is an eigenaxis of symmetry that exhibits symmetrical
dimensions and densities, so that the dual locus of the discriminant function
of the system is in statistical equilibrium at the geometric locus of the
decision boundary of the system, such that counteracting and opposing forces
and influences of the system are symmetrically balanced with each
other---about the geometric center of the locus of the novel principal
eigenaxis---whereon the statistical fulcrum of the system is located.

Thereby, the minimum risk binary classification system satisfies a state of
statistical equilibrium so that the total allowed eigenenergy and the expected
risk exhibited by the system are jointly minimized within the decision space
of the system, at which point the system exhibits the minimum probability of
classification error.
\end{theorem}

The general locus formula that resolves the inverse problem of the binary
classification of random vectors---that is expressed by Theorem
\ref{Inverse Problem of Binary Classification Theorem}---is readily
generalized to applied minimum risk multiclass classification systems.

\subsection{Applied Minimum Risk Classification Systems}

Corollary \ref{Applied Multiclass Classification System Corollary} generalizes
the fundamental laws of binary classification expressed by Theorem
\ref{Inverse Problem of Binary Classification Theorem} to applied minimum risk
multiclass classification systems.

\begin{corollary}
\label{Applied Multiclass Classification System Corollary}Any given minimum
risk multiclass classification system that is subject to $M$ sources of random
vectors $\mathbf{x\in}$ $%
\mathbb{R}
^{d}$ is determined by $M$ ensembles of $M-1$ minimum risk binary
classification systems, such that each ensemble is determined by an
architecture wherein one class is compared with all of the other $M-1$
classes, so that every one of the $M-1$ minimum risk binary classification
systems in each and every one of the $M$ ensembles satisfies the geometrical
and statistical criteria expressed by Theorem
\ref{Inverse Problem of Binary Classification Theorem}.

Thereby, the minimum risk multiclass classification system satisfies a state
of statistical equilibrium so that the total allowed eigenenergy and the
expected risk exhibited by the system are jointly minimized within the
decision space of the system, at which point the system exhibits the minimum
probability of classification error.
\end{corollary}

\begin{proof}
Corollary \ref{Applied Multiclass Classification System Corollary} is proved
by Theorem \ref{Inverse Problem of Binary Classification Theorem} and the
superposition principle
\citep{Lathi1998}%
---since any given applied minimum risk multiclass classification system is
based on a \textquotedblleft one versus all\textquotedblright\ architecture.
\end{proof}

We have finally reached a position where we have completed our treatise on the
fundamental problem of the binary classification of random vectors. Our major
findings are presented below.

\section{\label{Section 24}Major Findings}

We have covered a lot of ground in this treatise on the fundamental problem of
the binary classification of random vectors. We now summarize our major findings.

We have proved that Bayes' decision rule---which is considered the gold
standard for binary and multiclass classification tasks---does not satisfy the
conditions of Bayes' theorem, wherein identical random vectors generated by
distinct probability density functions account for the same effect exhibited
by a binary classification system. Thereby, we have demonstrated that Bayes'
decision rule constitutes an ill-posed rule for the direct problem of the
binary classification of random vectors---at which point the direct problem
was recognized to be an ill-posed problem.

We have developed a well-posed rule for the direct problem of the binary
classification of random vectors. Namely, we have derived a general locus
formula that resolves the direct problem of the binary classification of
random vectors by enlarging the complexity of a likelihood ratio test---that
is based on the \emph{maximum likelihood criterion}---which constitutes a
well-posed variant of \textquotedblleft Bayes' decision rule\textquotedblright%
\ for binary classification systems.

Thereby, we have developed a general locus formula for finding discriminant
functions of minimum risk binary classification systems that has the general
form of a system of fundamental locus equations of binary classification,
subject to distinctive geometrical and statistical conditions for a minimum
risk binary classification system in statistical equilibrium, so that certain
random vectors have coordinates that are solutions of the locus equations.

Thus, we have devised a theoretical model of a minimum risk binary
classification system that is based on the mathematical structure of the
operator of the system.

As a result, we have \emph{uncovered the black box }of a theoretical model of
a\emph{ minimum risk binary classification system}.

We have also uncovered a machine learning algorithm that resolves the inverse
problem of the binary classification of random vectors by identifying novel
and extremely unobvious processes---which include a novel principal
eigen-coordinate transform algorithm---that are executed by a well-posed
variant of the constrained optimization algorithm that is used by support
vector machines to learn nonlinear decision boundaries.

It was seen that the machine learning algorithm \emph{finds a system }of
fundamental \emph{locus equations} of binary classification, subject to
distinctive geometrical and statistical conditions for a minimum risk binary
classification system in statistical equilibrium---that is satisfied by
certain random vectors---at which point data-driven versions of the general
forms of the fundamental locus equations were seen to be determined by
distinctive algebraic and geometric interconnections between all of the random
vectors and the components of the minimum risk binary classification system.

It was also seen that the data-driven version of the general locus formula
executes precise mathematical conditions that \emph{statistically}
\emph{pre-wire} important \emph{generalizations} within the \emph{discriminant
function} of a minimum risk binary classification system---so that the
discriminant function generalizes and thereby \emph{extrapolates} in a
significant manner.

Thereby, it was seen that the structure and function of minimum risk binary
classification systems are intimately intertwined, such that the
\emph{structure} of a discriminant function of a minimum risk binary
classification system is essential for its \emph{functionality}---which
includes the ability of the discriminant function to \emph{generalize}.

Moreover, it was seen that the machine learning algorithm finds discriminant
functions---of minimum risk binary classification systems---by executing a
novel principal eigen-coordinate transform algorithm.

Thus, we have revealed a machine learning algorithm that determines the
mathematical structure of an operator of a minimum risk binary classification
system, such that the mathematical structure of the operator of the learning
machine \emph{is aligned with} the mathematical structure of the operator of
the theoretical model.

As a result, we have \emph{uncovered the black box }of an applied model of
a\emph{ minimum risk binary classification system}.

We have demonstrated that reproducing kernels are fundamental components of
minimum risk binary classification systems, such that certain types of
reproducing kernels replace random vectors with second-order curves---formed
by first and second degree vector components---that are more or less sinuous
and thereby preserve topological properties of vectors in Hilbert space. We
have also demonstrated that geometric loci of linear and quadratic decision
boundaries are both well-approximated by such second order curves in certain
reproducing kernel Hilbert spaces.

We have proved, from first principles, that any given minimum risk binary
classification system that is subject to random vectors has a certain
statistical structure and exhibits certain statistical behavior and properties.

Thereby, we have revealed a machine learning algorithm that determines minimum
risk binary classification systems whose statistical structure and behavior
and properties \emph{match} the statistical structure and behavior and
properties exhibited by the theoretical model of a minimum risk binary
classification system.

Correspondingly, we have devised a \emph{mathematical system} whose
statistical structure and behavior and properties \emph{models} fundamental
aspects of a minimum risk binary classification system---which is subject to
random vectors. The model represents a discriminant function, a decision
boundary, an exclusive principal eigen-coordinate system and an eigenaxis of
symmetry---that spans the decision space---of a minimum risk binary
classification system, so that the exclusive principal eigen-coordinate system
connects the discriminant function to the decision boundary of the system, at
which point the discriminant function, the exclusive principal
eigen-coordinate system and the eigenaxis of symmetry are each represented by
a geometric locus of a novel principal eigenaxis---which has the structure of
a dual locus of likelihood components and principal eigenaxis components.

We used the model of a minimum risk binary classification system that is
outlined above to explain how a discriminant function extrapolates---and
thereby generalizes in a significant manner. We also used the model to explain
how a minimum risk binary classification system acts to minimize its risk.
Even more, we used the model to predict error rates exhibited by minimum risk
binary classification systems.

Thereby, we have proved that discriminant functions of minimum risk binary
classification systems extrapolate, and thereby generalize in a very
nontrivial manner---because the important generalizations for a minimum risk
binary classification system are statistically pre-wired within the geometric
locus of the novel principal eigenaxis of the system---by means a system of
fundamental locus equations of binary classification, subject to distinctive
geometrical and statistical conditions for a minimum risk binary
classification system in statistical equilibrium---that is satisfied by
certain random vectors.

Equally important, we used the model to \emph{predict behavior} that we have
not been aware of. We used the model to predict that any given minimum risk
binary classification system acts to jointly minimize its eigenenergy and risk
by locating a point of equilibrium---at which point critical minimum
eigenenergies exhibited by the system are symmetrically concentrated---so that
the discriminant function of the system is in statistical equilibrium at the
decision boundary of the system, such counteracting and opposing forces and
influences of the system are symmetrically balanced with each other---about
the geometric center of the locus of the novel principal eigenaxis of the
system---whereon the statistical fulcrum of the system is located.

Thereby, we have proved that any given minimum risk binary classification
system satisfies a state of statistical equilibrium so that the total allowed
eigenenergy and the expected risk exhibited by the system are jointly
minimized within the decision space of the system, at which point the system
exhibits the minimum probability of classification error.

We have demonstrated that the use of scalar-valued cost functions grossly
oversimplifies the complexity of the fundamental problem of finding
discriminant functions of minimum risk binary classification systems. We have
shown that finding a discriminant function of a minimum risk binary
classification system involves optimizing a vector-value cost function---in
accordance with a well-posed eigenenergy functional---so that the total
allowed eigenenergy exhibited by the minimum risk binary classification system
is regulated by critical minimum eigenenergy constraints on a pair of primal
and dual novel principal eigenaxes that are symmetrically and equivalently
related to each other. Thereby, we have demonstrated that the use of a
vector-valued cost function is essential for finding discriminant functions of
minimum risk binary classification systems---that are subject to random vectors.

We have demonstrated that the overall structure and behavior and properties
exhibited by any given minimum risk binary classification system are
determined by elegant, deep-seated interconnections---between intrinsic
components of the system and the random vectors used to infer the values of
the parameters of the system---such that the parameters of the system
characterize the overall structure and behavior and properties of the system.

Correspondingly we have demonstrated that the essential information content of
any given training data set---that is used to find a discriminant function of
a minimum risk binary classification system---is contained within the
eigenstructures of the data set, such that all of the individual feature
vectors `add up' to a complete and sufficient eigenstructure, so that all of
the individual feature vectors `speak for themselves'---at which point joint
variabilities between all of the feature vectors are `accounted for.'

We have devised a mathematical framework for the direct problem and the
inverse problem of the binary classification of random vectors by devising
novel geometric locus methods in Hilbert\emph{\ }spaces---within statistical
frameworks---that fruitfully treat fundamental locus problems in binary
classification, where the Hilbert spaces are reproducing kernel Hilbert spaces
that have certain reproducing kernels. Thereby, we have demonstrated that
finding discriminant functions of minimum risk binary classification systems
is essentially a deep-seated locus problem in binary classification---situated
far beneath the surface---at which point underlying aspects of the problem are
subtle and extremely unobvious conditions.

We have demonstrated that the well-posed constrained optimization algorithm
that resolves the inverse problem of the binary classification of random
vectors executes novel and elegant processes---which include a novel principal
eigen-coordinate transform algorithm---that represent the solution for finding
discriminant functions of minimum risk binary classification systems, at which
point the direct problem is transformed into a feasible one.

Finally, we have demonstrated that the inverse problem of the binary
classification of random vectors is directly related to the forward problem of
the binary classification of random vectors---by fundamental laws of binary
classification that discriminant functions of minimum risk binary
classification systems are subject to.

The major findings outlined above promote new insights into fundamental
problems in both machine learning and data-driven mathematical modeling applications.

\section{\label{Section 25}New Insights}

We have discovered that the general problem of the binary classification of
random vectors is essentially a deep-seated locus problem in binary
classification that is situated far beneath the surface---at which point
underlying aspects of the problem are subtle and extremely unobvious conditions.

Correspondingly, we have discovered that the general problem of the binary
classification of random vectors is a statistical coordinate transform
problem, so that the general problem is resolved by a suitable change of the
basis of an intrinsic coordinate system of a locus equation of the decision
boundary of a minimum risk binary classification system, so that an exclusive
principal eigen-coordinate system---that provides dual representation of the
discriminant function, the intrinsic coordinate system of the decision
boundary, and an eigenaxis of symmetry that spans the decision space of the
minimum risk binary classification system---is generated by a novel principal
eigen-coordinate transform algorithm.

Even more, we have discovered that the general problem of the binary
classification of random vectors is also a system identification problem, so
that the overall statistical structure and behavior and properties of a binary
classification system are determined by transforming a collection of
observations into a data-driven mathematical model that represents fundamental
aspects of the system.

We have also discovered that solving the system identification problem for the
general problem of the binary classification of random vectors involves
solving a direct (forward) problem---which entails determining a fully
specified mathematical model of a binary classification system whose solution
is used to predict some type of system behavior.

As a result, we have obtained new insights into long-standing and deep-seated
problems in both machine learning and data-driven mathematical modeling applications.

We have discovered that the direct problem of the binary classification of
random vectors is an ill-posed problem, wherein identical random vectors
generated by distinct probability density functions account for the same
effect exhibited by a binary classification system.

We have discovered that resolving the general problem of the binary
classification of random vectors requires solving a system identification
problem, so that a deep-seated statistical dilemma is resolved by a well-posed
constrained optimization algorithm that executes a novel principal
eigen-coordinate transform algorithm.

We have discovered that resolving the bias and variance dilemma---for the
fundamental problem of finding discriminant functions of minimum risk binary
classification systems, subject to random vectors---requires solving a
data-driven mathematical modeling problem that is fruitfully treated by novel
geometric locus methods in Hilbert\emph{\ }spaces---within statistical frameworks.

Correspondingly, we have discovered that resolving the bias and variance
dilemma---for the fundamental problem of finding discriminant functions of
minimum risk binary classification systems, subject to random
vectors---requires finding a natural solution that determines an exclusive
principal eigen-coordinate system of the geometric locus of the decision
boundary of a minimum risk binary classification system.

Thereby, we have discovered that the general problem of the binary
classification of random vectors is a novel principal eigen-coordinate
transformation problem, so that the general problem is resolved by a suitable
change of coordinate system.

We have discovered a data-driven version of a general locus formula for
finding discriminant functions of minimum risk binary classification systems
that has the general form of a system of fundamental locus equations of binary
classification, subject to distinctive geometrical and statistical conditions
for a minimum risk binary classification system in statistical equilibrium, so
that certain random vectors have coordinates that are solutions of the locus
equations. Thereby, we have discovered that the data-driven version of the
general locus formula executes precise mathematical conditions that
\emph{statistically} \emph{pre-wire} important \emph{generalizations} within
the \emph{discriminant function} of a minimum risk binary classification
system---so that the discriminant function generalizes and thereby
\emph{extrapolates} in a significant manner.

Thus, we have discovered that the \emph{structure} of a discriminant function
of a minimum risk binary classification system is essential for its
\emph{functionality}---which includes the ability of the discriminant function
to \emph{generalize}.

We have discovered that the \textquotedblleft cost\textquotedblright\ of
finding a discriminant function of a minimum risk binary classification
system---subject to random vectors---is the critical minimum eigenenergy that
is necessary for the system to achieve a state of statistical equilibrium, at
which point critical minimum eigenenergies exhibited by the system are
symmetrically concentrated in such a manner that the geometric locus of the
novel principal eigenaxis of the system represents an eigenaxis of symmetry
that exhibits symmetrical dimensions and densities, so that counteracting and
opposing forces and influences of the system are symmetrically balanced with
each other---about the geometric center of the locus of the novel principal
eigenaxis---whereon the statistical fulcrum of the system is located.

Correspondingly, we have discovered that the total allowed eigenenergy that is
exhibited by a minimum risk binary classification system accounts for right
and wrong decisions made by the system. Thereby, we have discovered that
vector-valued cost functions provide essential measures for finding
discriminant functions of minimum risk binary classification systems---subject
to random vectors.

We have discovered that the inverse problem of the binary classification of
random vectors is directly related to the forward problem of the binary
classification of random vectors by fundamental laws of binary classification
that discriminant functions---of minimum risk binary classification systems,
subject to random vectors---are subject to. We have also discovered that the
theoretical model of a minimum risk binary classification system expresses
fundamental laws of binary classification, whereas the applied model of a
minimum risk binary classification system explains and executes these laws.

More generally, we have discovered that resolving the bias and variance
dilemma can be regarded as a system identification problem, so that the
overall statistical structure and behavior and properties of a system being
modeled are determined by transforming a collection of observations into a
data-driven mathematical model that represents fundamental aspects of the system.

We have also discovered that the data-driven mathematical model executes
precise mathematical conditions that \emph{statistically} \emph{pre-wire}
important \emph{generalizations} within the \emph{target function }of the
system---so that the target function generalizes and thereby
\emph{extrapolates} in a significant manner.

Thereby, we have discovered that the statistical structure of a target
function is essential for its functionality---which includes the ability of
the target function to generalize.

We have discovered that formulating and solving certain system identification
problems involves determining how and why a particular system locates a point
of equilibrium---so that the energy exhibited by the system is minimized in
such a manner that the system satisfies a state of equilibrium---at which
point the structure and behavior and properties of the system exhibit the
maximum amount of stability.

Correspondingly, we have discovered that formulating and solving certain
system identification problems involves finding a suitable equivalent
representation of a given system---that requires finding a suitable
statistical representation for the transformed basis of an intrinsic
coordinate system of the given system.

We have discovered that resolving the bias and variance dilemma for certain
data-driven mathematical modeling problems involves the identification of
certain processes---which execute certain methods---that represent the
solution for finding the target function of a data-driven mathematical system,
at which point the problem being considered is transformed into a feasible one.

We have discovered that determining the generalization behavior of certain
machine learning algorithms involves solving a system identification problem,
so that the overall statistical structure and behavior and properties of a
system are determined by transforming a collection of observations into a
data-driven mathematical model that represents fundamental aspects of the system.

As a final point, we have discovered that definite claims regarding the
generalization performance that is exhibited by certain machine learning
algorithms requires a proof, from first principles, which demonstrates that a
target function of a system has a certain statistical structure, along with a
constructive proof---which demonstrates that the machine learning algorithm is
aligned with the statistical structure of the target function of the system.

\section{Acknowledgments}

The author is indebted to Oscar Gonzalez and Garry Jacyna. The counsels of
Oscar Gonzalez and Garry Jacyna motivated the author to learn from both
breakthroughs and mistakes---and thereby persevere---and stay the course. The
author's master's thesis
\citep{Reeves1995}
was the primary impetus for this work. The counsel of Oscar Gonzalez motivated
the trailblazer within the author. Some of the material in this treatise
includes portions of the author's Ph.D. dissertation
\citep{Reeves2009}%
. Initial parts of this work would not have occurred without the support of
Garry Jacyna. The counsel of Garry Jacyna sustained the trailblazer within the
author---and also enabled the author to successfully navigate the Ph.D. pipeline.

\bibliographystyle{plainnat}
\bibliography{locus}

\end{document}